\documentclass[oneside,12pt]{IIScthesisPSnPDF}
\usepackage[table,dvipsnames,svgnames,x11names]{xcolor}

\usepackage{amssymb}
\usepackage{amsmath}
\usepackage{latexsym}
\usepackage{multicol}
\usepackage{booktabs}
\usepackage{multirow}
\usepackage{fancyhdr}
\usepackage[numbers]{natbib}
\usepackage{cleveref}
\usepackage{wrapfig}
\usepackage{algorithm}
\usepackage{amsfonts}       %
\usepackage{algorithmic}

\usepackage{mathtools}
\usepackage{enumerate}
\usepackage{cancel}
\usepackage{mathrsfs}
\usepackage{color}
\usepackage[compact]{titlesec}
\usepackage{caption}
\usepackage{mdwlist}
\usepackage{tikz}
\usepackage{varwidth}
\usepackage[vertfit]{breakurl}
\usepackage{datetime}
\usepackage{pdfpages}
\usetikzlibrary{shapes,arrows, trees}
\usepackage{bigstrut} %
\newcommand\Tstrut{\rule{0pt}{2.6ex}}         %
\newcommand\Bstrut{\rule[-0.9ex]{0pt}{0pt}}
\def\checkmark{\tikz\fill[scale=0.4](0,.35) -- (.25,0) -- (1,.7) -- (.25,.15) -- cycle;} 
\newcommand{\Cross}{$\mathbin{\tikz [x=1.4ex,y=1.4ex,line width=.2ex, black] \draw (0,0) -- (1,1) (0,1) -- (1,0);}$}%

\newcommand{\prob}[1]{\mathsf{Pr}\left(#1\right)}

\newtheorem{example}{Example}[chapter] %
\newtheorem{theorem}{Theorem}[chapter]
\newtheorem{lemma}{Lemma}[chapter]

\newtheorem{definition}{Definition}[chapter]
\newtheorem{proposition}{Proposition}[chapter]

\newcommand{\remove}[1]{}

\newtheorem{remark}{\textsc{Remark}}
\newtheorem{assumption}[theorem]{\textsc{Assumption}}

\DeclareMathOperator*{\argmin}{\arg\!\min}
\DeclareMathOperator*{\argmax}{\arg\!\max}

\newenvironment{proof}{\noindent{\bf Proof:} \hspace*{1mm}}{\hfill $\Box$ }
\newcommand{\notes}[1]{}

\newcommand{\first}[1]{$1^{\mathrm{st}}$}
\newcommand{\second}[1]{$2^{\mathrm{nd}}$}

\newcommand{\squishlisttwo}{
\begin{list}{$\blacktriangleright$}
{ \setlength{\itemsep}{0.5pt}
\setlength{\parsep}{0pt}
\setlength{\topsep}{0pt}
\setlength{\partopsep}{0.5pt}
\setlength{\leftmargin}{1em}
\setlength{\labelwidth}{1em}
\setlength{\labelsep}{0.5em} } }

\newcommand{\squishend}{
\end{list} }
\allowdisplaybreaks[1]

\newcommand{\hrnote}[1]{} %

\usepackage{dsfont}

\DeclareMathOperator{\Tr}{Tr}

\newcommand{\RR}{\mathbb{R}}

\newcommand{\ZZ}{\mathbb{Z}}

\newcommand{\EE}{\mathbb{E}}

\newcommand{\diag}{\mathrm{diag}}
\newcommand{\trn}{\top}

\newcommand{\ex}[2][{}]{\mathbb{E}_{#1}\left[#2\right]}
\newcommand{\var}[2][{}]{\mathbb{V}_{#1}\left[#2\right]}

\newcommand{\indc}{\mathds{1}}

\newcommand{\gpl}{F_{\mathrm{pl}}}
\newcommand{\err}{\mathrm{Err}}
\newcommand{\zo}{\text{0-1}}
\newcommand{\gopt}{F_{\mathrm{opt}}}
\newcommand{\gstar}{F^{*}}
\newcommand{\mpl}{\mM_{\mathrm{pl}}}

\newcommand{\rec}{\mathrm{rec}}

\newcommand{\cov}{\mathrm{cov}}

\newcommand{\blambda}{\bm{\lambda}}

\newcommand{\laloss}{\ell^{\mathrm{LA}}}
\newcommand{\hybloss}{\ell^{\mathrm{hyb}}}

\newcommand{\bone}{\bm{1}}
\newcommand{\regub}{\beta}

\newcommand{\softmax}{\textup{\textrm{softmax}}}
\newcommand{\ttt}[1]{\texttt{#1}}
\newcommand{\csst}{\ttt{CSST}}

\newcommand{\fro}{\mathrm{fro}}
\newcommand{\op}{\mathrm{op}}
\newcommand{\mcnorm}{\mC_{\|\cdot\|}}
\newcommand{\mcop}{\mC_{\|\cdot\|_{\op}}}
\newcommand{\mnnorm}{\mN_{\|\cdot\|}}
\newcommand{\pwhat}{\widehat{P}_w}
\newcommand{\otilde}{\widetilde{O}}

\newcommand{\Pw}{\mathcal{P}_w}
\def\coloneqq{\mathrel{\mathop:}=}

\newcommand{\mA}{\mathcal{A}}
\newcommand{\mB}{\mathcal{B}}
\newcommand{\mC}{\mathcal{C}}
\newcommand{\mD}{\mathcal{D}}

\newcommand{\mF}{\mathcal{F}}
\newcommand{\mG}{\mathcal{G}}
\newcommand{\mH}{\mathcal{H}}

\newcommand{\mL}{\mathcal{L}}
\newcommand{\mM}{\mathcal{M}}
\newcommand{\mN}{\mathcal{N}}

\newcommand{\mP}{\mathcal{P}}

\newcommand{\mS}{\mathcal{S}}
\newcommand{\mT}{\mathcal{T}}

\newcommand{\mX}{\mathcal{X}}
\newcommand{\mY}{\mathcal{Y}}

\newcommand{\bD}{\mathbf{D}}
\newcommand{\bE}{\mathbf{E}}

\newcommand{\bG}{\mathbf{G}}

\newcommand{\bM}{\mathbf{M}}

\newcommand{\bP}{\mathbf{P}}

\newcommand{\tC}{\widetilde{C}}

\newcommand{\frb}{\mathrm{F}}
\newcommand{\bmP}{\bm{\mP}}
\newcommand{\selmix}{\mathrm{SelMix}}
\newcommand{\invtemp}{s}
\newcommand{\weightmat}{W}

\newcommand{\vtt}{\widetilde{V}^{(t)}}

\newcommand{\mb}[1]{\mathbf{#1}}
\newcommand{\mc}[1]{\mathcal{#1}}
\newcommand{\mbb}[1]{\mathbb{#1}}

\newcommand{\paptitle}{NoisyTwins}%

\definecolor{mygray}{gray}{0.9} %

\definecolor{Gray}{gray}{0.95}
\definecolor{gray_1}{HTML}{e7edee}
\definecolor{gray_2}{HTML}{cfd4d5}
\definecolor{gray_3}{HTML}{b7bbbc}
\definecolor{gray_4}{HTML}{a2a5a6}
\definecolor{b_1}{HTML}{d2e2df}
\definecolor{b_2}{HTML}{e1ebe9}
\definecolor{b_3}{HTML}{e9f2f1}
\definecolor{b_4}{HTML}{f3faf9}

\newcommand{\veryshortarrow}[1][3pt]{\mathrel{%
   \hbox{\rule[\dimexpr\fontdimen22\textfont2-.2pt\relax]{#1}{.4pt}}%
   \mkern-4mu\hbox{\usefont{U}{lasy}{m}{n}\symbol{41}}}} %

\newtheorem{conjecture}{Conjecture} %

\definecolor{codegreen}{rgb}{0,0.6,0}
\definecolor{codegray}{rgb}{0.5,0.5,0.5}
\definecolor{codepurple}{rgb}{0, 0, 0}
\definecolor{backcolour}{rgb}{0.95,0.95,0.92}

\newcommand{\addedtext}[1]{#1}
\newcommand{\modifiedtext}[1]{#1}
\newcommand{\highlight}[1]{{\cellcolor[gray]{0.8}#1}}

\newcommand{\s}[2]{#1{\tiny$\pm$#2}}

\def\sS{{\mathbb{S}}}
\def\sB{{\mathbb{B}}}
\def\checkmark{\tikz\fill[scale=0.4](0,.35) -- (.25,0) -- (1,.7) -- (.25,.15) -- cycle;}

\definecolor{codegreen}{rgb}{0,0.6,0}
\definecolor{codegray}{rgb}{0.5,0.5,0.5}
\definecolor{codepurple}{rgb}{0, 0, 0}
\definecolor{backcolour}{rgb}{0.95,0.95,0.92}

\usepackage{bbm}
\usepackage{pbox}
\usepackage{minitoc}
\usepackage{adjustbox}

\usepackage{makecell}
\usepackage{setspace}
\usepackage{multirow}
\usepackage{makecell}
\usepackage{longtable}
\usepackage{enumitem}
\usepackage{amssymb}
\usepackage{bm}
\usepackage{hyperref}
\usepackage{amsmath,amssymb} %
\usepackage{sidecap}
\usepackage{appendix}
\usepackage{listings}

\usepackage{tablefootnote}
\usepackage{soul}
\usepackage{diagbox} %

\makeatletter
\DeclareRobustCommand\onedot{\futurelet\@let@token\@onedot}
\def\@onedot{\ifx\@let@token.\else.\null\fi\xspace}

\def\eg{\emph{e.g}\onedot} 
\def\ie{\emph{i.e}\onedot} 
 
\def\etc{\emph{etc}\onedot} 
\def\wrt{w.r.t\onedot} 
\def\etal{\emph{et al}\onedot}
\makeatother

\newcolumntype{C}[1]{>{\centering\arraybackslash\hspace{0pt}}p{#1}}
\usepackage{pifont}%
\newcommand{\cmark}{\ding{51}}
\newcommand{\xmark}{\ding{56}}

\definecolor{dgreen}{rgb}{0.04,0.7,0.13}
\definecolor{maroon}{rgb}{0.75,0.07,0.03}
\definecolor{dblue}{rgb}{0.1,0.07,0.75}

\newcommand{\stdn}[1]{}

\definecolor{NVblue}{rgb}{0.07,0.12,0.83}
\definecolor{BUred}{rgb}{0.8,0.0,0.0}
\definecolor{slategray}{rgb}{0.44, 0.7, 0.5}
\definecolor{ao}{rgb}{0.0, 0.5, 0.0}
\definecolor{my_gray}{gray}{0.95}

\usepackage{epsfig}
\usepackage{graphicx}

\usepackage{subcaption}
\usepackage{lipsum}
\newcommand{\norm}[1]{\left\lVert#1\right\rVert}
\usepackage{array}
\newcolumntype{H}{>{\setbox0=\hbox\bgroup}c<{\egroup}@{}}

\newdateformat{monthyeardate}{ \monthname[\THEMONTH] \THEYEAR}

\newcommand{\blankpage}{
\newpage
\thispagestyle{empty}
\mbox{}
\newpage
}

\crefname{observation}{observation}{observations}
\crefname{algorithm}{algorithm}{algorithms}
\crefname{align}{equation}{equations}
\crefname{eqnarray}{equation}{equations}

\begin{document}
\title{Learning from Limited and Imperfect Data}

\submitdate{July 2024} %
\phd
\dept{Department of Computational and Data Sciences}
\faculty{Faculty of Engineering}
\author{Harsh Rangwani}

\maketitle

\blankpage

\vspace*{\fill}
\begin{center}
\large\bf \textcopyright \ Harsh Rangwani\\
\large\bf July 2024 %
\\
\large\bf All rights reserved
\end{center}
\vspace*{\fill}
\thispagestyle{empty}

\blankpage

\vspace*{\fill}
\begin{center}
DEDICATED TO \\[2em]
\Large\it my parents, Sneha and Rajesh Rangwani,\\
\Large\it my sister, Bhavika Rangwani,~~\\

\vspace{0.3cm}
\it for believing in me and supporting me always!
\end{center}
\vspace*{\fill}
\thispagestyle{empty}

\blankpage

\vspace*{\fill}
\begin{tabular}{p{0.4\columnwidth}p{0.5\columnwidth}}
	{\em Signature of the Author}: & \dotfill \\
	& Harsh Rangwani \\
	& Dept. of Computational and Data Sciences \\ 
	& Indian Institute of Science, Bangalore \vspace{1in}\\
	{\em Signature of the Thesis Supervisor}: & \dotfill \\
	&R. Venkatesh Babu \\
	& Professor \\
	& Dept. of Computational and Data Sciences \\ 
	& Indian Institute of Science, Bangalore
\end{tabular}
\vspace*{\fill}
\thispagestyle{empty}

\blankpage

\setcounter{secnumdepth}{3}
\setcounter{tocdepth}{3}

\frontmatter %
\pagenumbering{roman}

\prefacesection{Acknowledgements}
First I would like to thank my parents and sister for their encouragement and unwavering support while pursuing this PhD. During the course of my PhD, they have supported me unconditionally, despite coming from a societal background where pursuing PhD is a rarity.

Next, I want to express my deepest gratitude towards my advisor, Prof. R. Venkatesh Babu, for his constant encouragement to pursue new ideas and push me out of my comfort zone. Being an exemplary researcher, his dedication and hard-working nature has inspired me to constantly be focused on doing good research. These qualities have shaped me a lot, and I will be carrying this forward in my research career.  I also want to deeply thank Dr. Varun Jampani and Dr. Sho Takemori, for serving as academic mentors and providing me with useful advice to improve my research skills. I also want to thank Prof. Anirban, Prof. Rajiv, Prof. Sastry, Prof. Soma, and Prof. Aditya for their advice during evaluations. 

This thesis presents collaborative works. I am deeply indebted to the friends who have collaborated with me, from whom I have learned a lot: Sumukh Aithal, Arihant Jain, Mayank Mishra, Tejan Karmali, Shrinivas Ramasubramanian, Naman Jaswani, Lavish Bansal, Kartik Sharma, Pradipto Mondal, Ashish Asokan, Soumalya Nandi,  Ankit Dhiman, R Srinath, Yashwanth M, Kunal Samanta, Kato Takashi, Yuhei Umeda, Ankit Sinha, Prof. Anirban Chakraborty. In particular, I would like to thank Sumukh for his energetic approach towards experimenting with things. I would like to thank Shrinivas for collaborating on very challenging deep theoretical and empirical work. I am thankful to Mayank for his perseverance and continued efforts on projects, which have been fruitful lately. The authors sincerely thank Tejan for being a collaborator and being trusted for almost everything near deadlines. In all the papers where I have been a joint first author, I have been co-involved in all the parts of a research paper, including ideation, implementation, analysis, and writing.

I also want to thank all the members of Vision and AI Lab (VAL Lab), for making my stay enjoyable and memorable for the last five years. I have enjoyed and benefited from interactions with VAL members (in addition to collaborators): Ram Prabhakar, Deepak Babu Sam, KL Navaneet, Jogendra Nath Kundu, Sravanti Addepalli, Abhipsa Basu, Sunandini Sanyal, Gaurang Sriramanan, Priyam Dey, Prasanna B, Kaushal Bhogale, Shivangi, Samyak Jain, Badrinath Singhal, Aakash Kumar Singh. I especially want to thank Sravanti Addepalli and Jogendra Nath Kundu, whom I could just run into for any kind of help and advice. I have immensely benefited from their helpful advice on paper and rebuttal writing. I also want to thank Balaji, Srikrishna, and Kuldeep from Adobe for their advice on the project and research.

I want to also thank the PMRF funding scheme for providing me with scholarship and travel funding for attending conferences, which has helped me stay relaxed and productive during this PhD journey. I also want to thank Kotak AI Centre, Google Travel Grant, ACM ARCS, and Pratiksha Trust Travel Grant for sponsoring the international conference trips. I want to thank SERC and VAL Lab for providing me with support with computational resources. Thanks to the CDS support staff for helping with all of the administrative processing of these.

Last but not least, I want to thank my friends who have made my stay enjoyable at IISc: Tejan Karmali, Vikash Kumar, Rishubh Parihar, Naman Jaswani, Alok Kumar, Gaurav Nayak, Sangeeta Yadav,  Ashish Kumar Jayant, Dushyant Raut, Bharati Khanijo, Ruchi Bhoot, Raj Maddipatti, Sumanth Kumar, Shubham Goswami, Manu Ghulyani, Rahul John Roy, Karan Jaswani, Ghanshyam Chandra, Shweta Pandey, Naveen Paluru, Utkarsh Gupta, Subhabrata Basak, Gourab Panigrahi \dots. As listing names is often error-prone, I apologize if some names were accidentally missed. I want to thank Tejan and Vikash for their help during ACL injury.

I am indebted to Dr. Anil Kumar Singh and Dr. Rajeev Sangal for introducing me to research during my undergrad at IIT BHU. Finally, I want to thank the Almighty God for providing me with courage and inspiration when I didn't see the light at the end of the tunnel.

\prefacesection{Abstract}
Deep Neural Networks have demonstrated orders of magnitude improvement in capabilities over the years after AlexNet won the ImageNet challenge in 2012. One of the major reasons for this success is the availability of large-scale, well-curated datasets. These datasets (\eg ImageNet, MSCOCO, etc.)  are often manually balanced across categories (classes) to facilitate learning of all the categories. This curation process is often expensive and requires throwing away precious annotated data to balance the frequency across classes. This is because the distribution of data in the world (\eg, internet, etc.) significantly differs from the well-curated datasets and is often over-populated with samples from common categories. The algorithms designed for well-curated datasets perform suboptimally when used for learning from imperfect datasets with long-tailed imbalances and distribution shifts.To expand the use of deep models, it is essential to overcome the labor-intensive curation process by developing robust algorithms that can learn from diverse, real-world data distributions. Toward this goal, we develop practical algorithms for Deep Neural Networks which \emph{can learn from limited and imperfect} data present in the real world. This thesis is divided into four segments, each covering a scenario of learning from limited or imperfect data. The first part of the thesis focuses on \textbf{Learning Generative Models from Long-Tail Data}, where we mitigate the mode-collapse and enable diverse aesthetic image generations for tail (minority) classes. In the second part, we enable effective generalization on tail classes through \textbf{Inductive Regularization schemes}, which allow tail classes to generalize as effectively as the head classes without requiring explicit generation of images. In the third part, we develop algorithms for \textbf{Optimizing Relevant Metrics} for learning from long-tailed data with limited annotation (semi-supervised), followed by the fourth part, which focuses on the \textbf{Efficient Domain Adaptation} of the model to various domains with very few to zero labeled samples.

\vspace{1mm} \noindent \textbf{Generative Models for Long-Tail Data.} We first evaluate generative models' performance, specifically variants of Generative Adversarial Networks (GANs) on long-tailed datasets. The GAN variants suffer from either mode-collapse or miss-class modes during generation. To mitigate this, we propose \emph{Class Balancing GAN with a Classifier in the Loop}, which uses a classifier to asses the modes in generated images and regularizes GAN to produce all classes equally. To alleviate the dependence on the classifier, we take a closer look at model behavior at mode collapse and observe that spectral norm explosion of Batch Norm parameters correlates with mode collapse. We develop an inexpensive \emph{group Spectral Regularizer (gSR)} which mitigates the spectral collapse and significantly improves the SotA conditional GANs (SNGAN and BigGAN) performance on long-tailed data. However, we observed that class confusion is present in the generated images due to gSR norm regularization for large datasets. For this, in our latest work \emph{NoisyTwins}, we factor the latent space as distinct Gaussian by design for each class, enforcing class consistency and intra-class diversity using a contrastive approach (BarlowTwins). This helps us to scale high-resolution StyleGANs for thousand class long-tailed datasets of ImageNet-LT and iNaturalist2019, achieving State-of-the-Art (SotA) results while maintaining class-consistency.

\vspace{1mm} \noindent \textbf{Inducting Regularization Schemes for Long-Tailed Data.} While Data Generation is promising for improving classification models on tail classes, it often comes with the cost of training an auxiliary generative model. Hence, lightweight techniques such as higher loss weights for tail classes (loss re-weighting) while training CNNs, is practical to improve performance on the minority classes. However, due to this, we observe that the model converges to saddle point instead of minima for tail classes, hindering generalization.  We show that inducing inductive bias of \emph{escaping saddles and converging to minima }for tail classes, using Sharpness Aware Minimization (SAM) significantly improves performance on tail classes. Despite inductive regularizations,   training Vision Transformers (ViTs) for long-tail recognition is still challenging due to the complete lack of inductive biases such as locality of features, making their training data intensive. We propose \emph{DeiT-LT}, which introduces OOD and low-rank distillation from CNNs, to induce CNN-like robustness into scalable ViTs.

\vspace{1mm} \noindent \textbf{Semi-Supervised Learning for Practical Non-Decomposable Objectives.} The above methods work in supervised long-tail learning, where they avoid throwing off the annotated data. However, the real benefit of long-tailed methods could be leveraged when they utilize the extensive unlabeled data present (\ie semi-supervised setting). For this, we introduce a paradigm where we measure the performance using relevant non-decomposable metrics such as worst-case recall and recall H-mean on a held-out set, and we use their feedback to learn in a semi-supervised long-tailed setting. We introduce \emph{Cost-Sensitive Self Training (CSST)}, which generalizes self-training based semi-supervised learning (\eg FixMatch, etc.) to the long-tail setting with strong guarantees and empirical performance. The general trend these days is to use self-supervised pre-training to obtain a robust model and then fine-tune it. In this setup, we introduce \emph{SelMix}, an inexpensive fine-tuning technique to optimize the relevant metrics using pre-trained models. In SelMix, we relax the assumption that the unlabeled distribution is similar to the labeled one, making the models robust to distribution shifts.

\vspace{1mm} \noindent \textbf{Efficient Domain Adaptation.} The long-tail learning algorithms focus on the limited data setup and improving in-distribution generalization. However for practical usage, the model must learn from imperfect data and perform well across various domains. Towards this goal, we develop \emph{Submodular Subset Selection for Adversarial Domain Adaptation}, which carefully selects a few samples to be labeled for maximally improving model performance in the target domain. To further improve the efficiency of the Adaptation procedure, we introduce \emph{Smooth Domain Adversarial Training (SDAT)}, which converges to generalizable smooth minima. The smooth minimum enables efficient and effective model adaptation across domains and tasks.

\prefacesection{Publications based on this Thesis}
    \begin{enumerate}

    \item \textbf{Harsh Rangwani}, Konda Reddy Mopuri, and R.Venkatesh Babu, \href{https://arxiv.org/pdf/2106.09402}{\textit{Class Balancing GAN with a Classifier in the Loop }}, in Uncertainity in Artifical Intelligence (\textbf{UAI}), 2021.
    
    \item \textbf{Harsh Rangwani}, Naman Jaswani, Tejan Karmali, Varun Jampani, R. Venkatesh Babu \\ \href{https://arxiv.org/pdf/2208.09932}{\textit{Improving GANs for Long-Tailed Data through Group Spectral Regularization}}, in European Conference on Computer Vision
    (\textbf{ECCV}), 2022. 

    \item \textbf{Harsh Rangwani}\footnote{Equal contribution authors}, Lavish Bansal\footnotemark[1], Kartik Sharma, Tejan Karmali, Varun Jampani, R. Venkatesh Babu, \href{https://arxiv.org/pdf/2304.05866}{\textit{NoisyTwins: Class-Consistent and Diverse Image Generation through StyleGANs}} in IEEE/CVF Conference on Computer Vision and Pattern Recognition (\textbf{CVPR}), 2023. 
    
    \item  \textbf{Harsh Rangwani}\footnotemark[1], Sumukh K Aithal\footnotemark[1], Mayank Mishra, R. Venkatesh Babu, \href{http://arxiv.org/abs/2212.13827}{\textit{Escaping Saddle Points for Effective Generalization on Class-Imbalanced Data}}, in Advances in Neural Information Processing Systems (\textbf{NeurIPS}), 2022.

    \item \textbf{Harsh Rangwani}\footnotemark[1],  Pradipto Mondal\footnotemark[1], Mayank Mishra\footnotemark[1], Ashish Ramayee Asokan, R. Venkatesh Babu \href{https://arxiv.org/pdf/2404.02900}{\textit{DeiT-LT: Distillation Strikes Back for Vision Transformer Training on Long-Tailed Datasets}}, in IEEE/CVF Conference on Computer Vision and Pattern Recognition (\textbf{CVPR}), 2024.

    \item \textbf{Harsh Rangwani}\footnotemark[1], Shrinivas Ramasubramanian\footnotemark[1], Sho Takemori\footnotemark[1], Kato Takashi, Yuhei Umeda, R. Venkatesh Babu, \href{https://openreview.net/pdf?id=bGo0A4bJBc}{\textit{Cost Sensitive Self-Training for Optimising Non-decomposable Measures}}, in Advances in Neural Information Processing Systems (\textbf{NeurIPS}), 2022.
    
    \item Shrinivas Ramasubramanian\footnotemark[1], \textbf{Harsh Rangwani}\footnotemark[1], Sho Takemori\footnotemark[1], Kunal Samanta, Yuhei Umeda, R. Venkatesh Babu, \href{https://arxiv.org/pdf/2403.18301v1.pdf}{\textit{Selective Mixup Fine-Tuning for Optimizing Non-Decomposable Objectives}}, in International Conference on Learning Representations (\textbf{ICLR}), 2024. \textit{(Spotlight, Top 5\%)}

    \item \textbf{Harsh Rangwani}, Arihant Jain\footnotemark[1], Sumukh K Aithal\footnotemark[1], R. Venkatesh Babu, 
 \href{https://arxiv.org/pdf/2109.08901v1.pdf}{\textit{S3VAADA: Submodular Subset Selection for Virtual Adversarial Active Domain
Adaptation}}, in International Conference on Computer Vision (\textbf{ICCV}), 2021.

    \item \textbf{Harsh Rangwani}\footnote{Equal contribution authors}, Sumukh K Aithal\footnotemark[1], Mayank Mishra, Arihant Jain, R. Venkatesh Babu, \href{https://arxiv.org/pdf/2206.08213}{\textit{A Closer Look at Smoothness in Domain Adversarial Training}}, in International Conference on Machine Learning (\textbf{ICML}), 2022. (\textit{also Spotlight at ECCVW 2022})

\end{enumerate}

\vspace{10mm}
The following publications are done in my Ph.D. but are outside the scope of this thesis.

\begin{enumerate}

    \item Tejan Karmali, Rishubh Parihar, Susmit Agrawal, \textbf{Harsh Rangwani}, Varun Jampani, Maneesh Singh, R.Venkatesh Babu, \href{https://arxiv.org/pdf2208.03764}{\textit{Hierarchical Semantic Regularization of Latent Spaces in StyleGANs}}, in European Conference on Computer Vision (\textbf{ECCV}), 2022. 
    
    \item Ankit Dhiman, Srinath R, \textbf{Harsh Rangwani}, Rishubh Parihar, Lokesh R Boregowda, Srinath Sridhar, R Venkatesh Babu \href{https://arxiv.org/pdf/2308.10337}{\textit{Strata-NeRF: Neural Radiance Fields for Stratified Scenes}}, in International Conference on Computer Vision (\textbf{ICCV}), 2023.

    \item M Yashwanth, Gaurav Kumar Nayak, \textbf{Harsh Rangwani}, Arya Singh, R. Venkatesh Babu, and Anirban Chakraborty, \href{https://openaccess.thecvf.com/content/WACV2024/papers/Yashwanth_Minimizing_Layerwise_Activation_Norm_Improves_Generalization_in_Federated_Learning_WACV_2024_paper.pdf}{\textit{Minimizing Layerwise Activation Norm Improves Generalization in Federated Learning}}, in  IEEE/CVF Winter Conference on Applications of Computer Vision (\textbf{WACV}), 2024. 

    \item Soumalya Nandi, Sravanti Addepalli\footnotemark[1], \textbf{Harsh Rangwani}\footnotemark[1], and R. Venkatesh Babu, \href{https://arxiv.org/pdf/2304.10446.pdf}{\textit{Certified Adversarial Robustness Within Multiple Perturbation Bounds}}, in IEEE/CVF Conference on Computer Vision and Pattern Recognition (\textbf{CVPR}) Workshops, 2023.

    \item \textbf{Harsh Rangwani}, Aishwarya Agarwal, Kuldeep Kulkarni, R. Venkatesh Babu, Srikrishna Karnam {\textit{Crafting Parts for Expressive Object Composition}}, Under Review.

    \item Shrinivas Ramasubramanian\footnotemark[1], Soumyajit Karmakar\footnotemark[1], Harsh Rangwani\footnotemark[1], Yasuhiro Aoki, Genta Suzuki, R.Venkatesh Babu, \textit{Prompt-ReID: Prompt Based Inference for \\ Re-Identification using Vision-Language Models},  Under Review.

\end{enumerate}

\tableofcontents
\listoffigures
\listoftables

\mainmatter %
\setcounter{page}{1}

\chapter{Introduction}
\label{chap:introduction}

\hrnote{ CNNs have improved on public tasks
}
\hrnote{For them to be used all across various tasks, they need to work on messy real-world data}

\hrnote{Can the models be trained effectively using the limited and imperfect data present in the real world?}

Deep Neural Networks (DNNs) have shown significant promise, particularly with the advent of AlexNet~\cite{krizhevsky2012imagenet} that wont the ImageNet Challenge in 2012. Following this, DNNs have demonstrated State-of-the-Art (SotA) performance across various computer vision tasks (such as recognition, detection, segmentation \etc) and have become the de-facto standard in the community. However, in addition to many technical advances, the significant performance improvements also come from the availability of large-scale datasets, starting with ImageNet. The data-curation and annotation process to produce these datasets is often expensive and requires a lot of filtering. This makes it infeasible for small organizations to do this for each task at hand.
This poses an important question, which we aim to address -- \emph{Can neural networks be learned from the messy and imperfect data in the real world, without the expensive data curation? }

\section{The Context}

\hrnote{Deep Learning fueled by balanced datasets}
Computer vision, the field of artificial intelligence that enables computers to derive meaningful information from digital images, videos, and other visual inputs, represents one of the most transformative applications of artificial intelligence, with the potential to revolutionize industries from healthcare and autonomous vehicles to retail and manufacturing. The ability of machines to understand and interpret visual information has become increasingly critical in our digital world, driving remarkable innovations across sectors~\cite{voulodimos2018deep}. Deep Learning has significantly improved the performance of various computer vision tasks such as image recognition~\cite{krizhevsky2012imagenet}, object detection~\cite{ren2015faster}, semantic segmentation~\cite{ronneberger2015u}, etc. However, much of this is fueled by the availability of \emph{large-scale, balanced, and curated datasets} like ImageNet, COCO \etc. The typical paradigm used these days involves fine-tuning large models pre-trained on \emph{large balanced} datasets for various computer vision tasks. Recent models like CLIP~\cite{CLIP}, pre-trained on internet-scale image-caption pairs, require a data pre-processing step involving \emph{balancing the data across classes}~\cite{xu2024demystifying}. The artificial balancing of data leads to the discarding of expensive annotated data present for data-rich classes. 
    
The requirement of balancing is recurring since the natural distribution of categories across species, actions, \etc follows a long-tailed distribution, \ie a majority of samples occur for common (head) classes and there exist a large number of (tail) classes with a very few number of samples each. This follows a pattern described by Zipf's law~\cite{zipf1949principle}, which suggests that the frequency of categories in natural distributions decreases in an exponential decay manner.  Due to their naturalness, long-tailed distributions have been studied extensively across finance and statistics~\cite{ong1995class, anderson2004long}. However, despite long-tail distribution being the natural distribution, most ML works have focused on developing algorithms to learn from balanced datasets. Hence, to reduce the wastage of annotated data in balancing, \emph{Long Tailed Learning} aims to effectively use long-tailed datasets present in nature that include face recognition and medical imaging \etc to learn deep models. In this setting, the final evaluation ensures that the model needs to learn effectively from \emph{limited samples in tail classes} to perform well in all categories.

Another common scenario due to {imperfect data} occurs when a deep model trained on data from a single source domain, fails catastrophically when deployed in a different target domain. For example, an object detector trained on day-time images fails for night-time images. To mitigate this and solve the same task in the target domain (\ie night), producing new labeled data is prohibitively expensive and often redundant in the presence of source data. Hence, effectively making use of source model and data is essential for efficiency. Further, being able to transfer knowledge from source to target allows us to use cheap synthetic data from the source domain for model training, after which the model that can be adapted with limited data to the target domain. This has potential applications in Reinforcement Learning, Autonomous Driving, \etc, due to the availability of synthetic data. This setting of adapting models using the imperfect data in the target domain is studied as the problem of \emph{Domain Adaptation}.

\hrnote{Problems with just using balanced data}

\hrnote{What problems you are trying to address in the thesis}

\section{Notions of Learning}
In this section we explain the learning notions which we aim to optimize in \emph{long-tail learning} and \emph{domain adaptation} settings, in this thesis. This distinguishes this work from existing works that optimize a different notion of the same setting.
\subsection{Learning from Long-Tail Data}
In this section, we aim to clarify the notion of long-tailed learning that we study in this thesis. In long-tailed learning, we assume learning on a data distribution such that the majority of samples are present in a few head classes, and there are a lot of tail classes with few samples. In the world humans can reason about new classes if they have seen a similar class before; for example, a human might be able to recognize red hummingbird easily if they have seen a common hummingbird. Inspired by this, we say that a long-tailed learning algorithm is successful if the algorithm's performance on a particular tail class through a long-tailed algorithm is better than learning through few-shot tail class data. Hence, the long-tail learning algorithm aims to transfer knowledge from head classes to tail.

For long-tailed learning various regularization techniques have been developed to mitigate the bias towards head classes~\cite{zhang2021deep}. However, in our case, we aim to develop specific type of regularization techniques that enforce the useful properties of the converged model for the head classes to be enfored for the tail classes as well. This will ensure that tail classes attain similar generalizable solutions as the head classes by utilizing the regularization prior. For example, let's say an important property $\mathcal{P}$ of the model with parameters $\theta$ when computed on tail data $\mathcal{D}_{t}$ should be similar to the property computed on head class data $\mathcal{D}_h$ after regularization.
\begin{equation}
    \EE_{x \sim \mathcal{D}_{t}}[P_{\theta}(x)] \approx \EE_{x \sim \mathcal{D}_{h}}[P_{\theta}(x)]
\end{equation}
One such example of a useful property is the margin distance to the decision boundary. To develop such a scheme for long-tail learning, the task is to find such a property which can be \textbf{a)} matched without much computation overhead through regularization and \textbf{b)} improves the model's generalization performance under appropriate evaluation criterion. This aligns well with our high-level goal of transferring knowledge from head to tail classes.

We want to clarify that just improving the performance of the models on long-tail benchmarks by using pre-trained models on large balanced datasets like CLIP, etc, doesn't satisfy the long-tail learning criterion described above. 

\subsection{Learning to adapt models with minimal data}
The goal of domain adaptation is to improve the performance of models in a target domain $\mD_{\mT}$, using labeled data from source domain $\mD_{\mS}$ and unlabeled data from the target domain $\mD_{\mathcal{T}}$. We aim to improve the model performance on the target domain by developing algorithms to use the data present in the target domain effectively. The adapted model is useful when the adapted model performance is better than the source model, demonstrating practical usage of target data. Further, just using large-scale pre-trained backbones (\eg CLIP) to improve adaptation performance on benchmarks doesn't imply improved domain adaptation in general.

\section{Problems of Interest}
\label{sec:intro_problems}
This thesis aims to address problems related to learning from limited and imperfect data, ranging from supervised long-tail learning to unsupervised domain adaptation. This covers issues from learning from a few samples to adapting to a different domain in an unsupervised fashion. The focus is to ensure that developed algorithms for limited data converge to solutions akin to ones learnt from a well-curated balanced dataset, in following cases:

\subsection{Generative Models for Long-Tailed Data}
Generative models have become an integral part of computer vision after demonstrating spectacular performance in image generation~\cite{rombach2021highresolution, karras2019style}. The usage of generative models has increased over time in various applications like segmentation~\cite{abdal2021labels4free}, depth estimation~\cite{bhattad2024stylegan} \etc as they learn an excellent image prior from the data.
One of the popular classes of contemporary generative models is Generative Adversarial Networks (GANs)~\cite{goodfellow2014generative}, which learns a function to sample (\ie generate) data for a given data distribution. GANs have demonstrated excellent generation quality with fast sampling, which makes them the right choice for interactive image applications~\cite{Karras2019stylegan2, Sauer2021NEURIPS}. Further, to improve the usability of Diffusion models (\eg Stable-Diffusion), they are being distilled into GANs for faster sampling in creative applications~\cite{yin2024one, sauer2023adversarial}. We consider the challenging setting of training a Generative Adversarial Network on long-tailed data, where the goal is to generate diverse and aesthetic images even for the tail classes.  We evaluate the model in this setting by computing the distribution distance between the generated images and a class-balanced held-out set to ensure the improvement in the quality of generation across all classes.

\subsection{Long-Tailed Recognition}

One can use the generative models to solve the data scarcity issue for the tail classes. However, that involves an additional overhead in training an additional model and training the image recognition model on a much larger data set. This can be mitigated in case we can achieve similar solutions for head and tail classes by introducing specific schemes (like regularization, \etc) during training to induce the desired properties. One example could be that the margin between the centroid of samples and the decision boundary is the same across the head and tail classes. In this setup, we study how we could induce properties like equi-margin, first for the CNNs and then for the ViTs (Vision Transformers) models. Inducing properties via regularization is more challenging for ViTs as they don't have any inductive bias due to their architecture, unlike CNNs, which have biases like locality of features, etc. This leads to CNN being more robust to long-tail imbalances, requiring more specialised techniques for ViTs.

\subsection{Leveraging Unlabeled Data in Long-Tailed Learning}

Most work on long-tailed learning has focused on learning from supervised datasets with label-image pairs. However, the maximal benefit of focusing on long-tailed datasets could come from the effective usage of the large unlabeled data present. The big advantage of a long-tailed setup is that the supervised data distribution coarsely matches the distribution of the unlabeled data (assuming a mild shift in distribution). On the contrary, the balanced dataset distribution differs from the long-tailed unlabeled data, so semi-supervised algorithms designed for balanced data fail in real-world long-tailed settings~\cite{kim2020distribution, oh2022daso}. To improve this, we study the problem of semi-supervised long-tailed learning, where we assume access to a small labeled long-tailed dataset and a large set of unlabeled data. For validation, we assume access to a held-out balanced set, which fairly provides knowledge about the algorithm's performance across all classes. More specifically, on the validation set, we measure relevant metrics for long-tailed data like worst-case recall and H-mean of recall, which are a form of non-decomposable metrics, and update the training algorithm to optimize them. In general \textbf{non-decomposable} objectives are ones, which cannot be expressed as sum of terms (\eg errors) on individual samples and require special techniques for optimization. This problem setup closely resembles the cases we observe in the real world, where we have a constant data source and a small budget for labeling to train a model to be deployed, hence needs to obey the constraints.

\subsection{Learning to Adapt Models with Minimal Supervision}
One crucial assumption above is that the unlabeled data available follows the same distribution as the labeled case. However, with time, the distribution of data being collected and used as unlabeled data might change. This situation can also arise when the model must be deployed in a new target domain, where data distribution might differ from source training distribution. In such cases, we study the problem of how to use minimal data for model adaptation to the target domain. We first look at the problem of converging to generalizable solutions in the loss landscape in the source domain, such that the solution can be adapted effectively using unlabeled data. We further look at the problem of finding a small subset of samples to be labeled to achieve maximal improvement in performance on the target domain (active learning). 

\section{Organization of the thesis}

We discuss the organization of the thesis in this section. The thesis is divided into four parts, where each part broadly addresses one problem setup discussed in Section-\ref{sec:intro_problems}. Each chapter discusses an instance of algorithm and strategies developed to tackle the problem.

    \vspace{1mm} \noindent \textbf{1. Training Generative Adversarial Networks (GANs) on Long-Tailed Data.} Generative Adversarial Networks (GANs) are a class of popular generative models used for modeling image data distribution, as they achieve great photorealism and have the property of fast one-step sampling from the distribution. However, the majority of GANs utilize balanced datasets for training. Due to this, they suffer from mode collapse when trained on long-tailed datasets. In this part, we discuss strategies to mitigate mode-collapse and train variants of GANs on small-scale to large-scale long-tailed datasets.
    
    \begin{itemize}
            \item \textbf{Class Balancing GAN with A Classifier in The Loop.} In Chapter~\ref{chap:cbgan}, we first look at the issues faced by GANs when trained on long-tailed data. We find that conditional GANs that generate images based on class labels suffer from mode collapse. On the other hand, unconditional GANs don't collapse, but they generate arbitrary images with arbitrary label distribution and miss generating certain classes altogether. In this work, we aim to use the extensive work on long-tail recognition to obtain a good classifier and then use it to guide the generations of unconditional GAN (\ie classifier guidance). We propose a \emph{Class Balancing Regularizer} which first estimates the class distribution of the generated samples and then provides feedback to the unconditional GAN (CBGAN) to produce a more uniform generation across all the classes in the dataset. We demonstrate that CBGAN maintains photorealism (no-collapse) and generates near-uniform class distribution without missing any classes.
        
        \item \textbf{Group Spectral Regularization (gSR) for GANs.} The unconditional CBGAN requires an additional classifier at training time, which is an overhead and limits scaling to large GAN models. In Chapter~\ref{chap:gsrgan}, we aim to tackle the problem of mode-collapse in conditional GANs without requiring any external classifier feedback. Our important observation is that \emph{spectral norm} of tail class-specific parameters explodes and deviates from the distribution of parameters of head classes. This phenomenon leads to a mode-collapse in generated images for the popular cGANs such as SNGAN~\cite{miyato2018cgans} and BigGAN~\cite{brock2018large}. To mitigate this spectral collapse, we propose an inexpensive group Spectral Regularizer (gSR) based on power iteration to mitigate the collapse. We find that gSR can mitigate collapse effectively and helps BigGAN to generate high-resolution results ($256\times256$), which is 4 $\times$ of CBGAN, on the long-tailed LSUN scene dataset.

        \item \textbf{Diverse and Class Consistent Generation.} In the gSR work, we were able to mitigate the mode-collapse by group Spectral Regularizer. However, this led to the problem of class confusion (with large no. of classes) as the regularizer enforced all class-specific parameter vectors to be low norm. This prevented the model from scaling on datasets like ImageNet-LT with 1000 classes. In Chapter~\ref{chap:NoisyTwins}, we aim to generate class-consistent outputs from a conditional StyleGAN while maintaining diversity and avoiding collapse. This is achieved through a self-supervised learning-inspired NoisyTwins regularizer, which enforces distinct clusters for each class in the latent space through augmentation based regularization. We show that NoisyTwins can train StyleGAN on large-scale ImageNet-LT and iNaturalist 2019, achieving diversity with high fidelity across all ($\sim$ 1000) classes. 
    \end{itemize}
   \vspace{1mm} \noindent  \textbf{2. Inductive Regularization for Long-Tailed Recognition.} With generative models, we can augment the long-tailed training data to improve the performance of the downstream recognition models. However, this comes with an overhead of training a generative model and extra training to process the generated data. In Part~\ref{part:inductive_LT}, we aim to improve the recognition model performance with regularization techniques instead of generation. In this line of work, we mainly aim to induce generalization properties achieved by head classes, like convergence to flat minima, into the tail classes through regularization.

    \begin{itemize}

            \item \textbf{Escaping Saddle Points for Generalization on Long-Tailed Data.} In Chapter~\ref{chap:SaddleSAM}, we look closely at the curvature of the converged solution in the loss landscape.  When we analyze the loss landscape independently for each class in a long-tail setting, we obtain an interesting insight. When the loss is re-weighted to focus on the tail classes, the solution converges to a saddle point instead of the minima for tail classes. We find that convergence to saddle points hinders generalization, which can be improved by using optimization techniques to escape saddle points. We demonstrate that Sharpness Aware Minimization (SAM)~\cite{foret2020sharpness} can be effectively used to escape saddle points theoretically and empirically, significantly improving the tail class recognition performance across various datasets and re-weighting techniques.

        \item \textbf{Inducing CNN Inductive Bias into ViTs for Long-Tailed Recognition.} In Chapter~\ref{chap:DeiT-LT}, we aim to train the recent scalable Vision Transformer (ViT) model on long-tailed data. Due to a lack of inductive bias like locality of features and parameter sharing, the ViTs overfit on tail classes and perform worse in comparison to CNNs. To induce the inductive bias for robustness to long-tail imbalances,  we propose \emph{DeiT-LT}, a data-efficient distillation method using CNN as a teacher. \emph{DeiT-LT is a scheme to train ViTs through distillation from low-rank CNN teachers using Out-of-Distribution (OOD) images}. We find that distillation jointly with ground truth training allows the benefit of both - the scalable ViT architecture learns better on head classes from ground truth, and the necessary robustness on tail classes comes from distilling from CNN. DeiT-LT enables training ViTs from scratch on long-tailed datasets ranging from small-scale CIFAR to large-scale iNaturalist-2018 datasets, demonstrating significant improvements over prior arts.
        
    \end{itemize}

    \vspace{1mm} \noindent  \textbf{3. Semi Supervised Long-Tail Learning for Non-Decomposable Objectives.} In Part~\ref{part:SemiSL_LT}, we aim to develop algorithms for a more practical setting where we have access to a real-world long-tail dataset but only have a small budget to label a fraction of the data. This corresponds to a Semi-Supervised Long-Tail Learning setting where we must effectively utilize both the unlabeled and labeled data to achieve optimal performance. We work in a paradigm where we keep a balanced held-out set to measure model performance by using relevant non-decomposable metrics like Worst-Case Recall and H-Mean of Recall in a long-tail setting. We then use the feedback from the held-out set to guide the training algorithm in optimizing the non-decomposable metric objective. 

    \begin{itemize}
        \item \textbf{Cost Sensitive Self-Training (CSST) for Non-Decomposable Metric Optimization.} In Chapter~\ref{chap:csst}, we aim to generalize the semi-supervised self-training based methods like FixMatch~\cite{sohn2020fixmatch} for long-tail semi-supervised learning. Using self-training algorithms naively on long-tailed data leads to learning only from head classes. In our work, we introduce \emph{Cost Sensitive Self-Training (CSST)}, which theoretically aims to optimize objectives like worst-case recall, leading to a cost-sensitive regularization technique that focuses on all classes. We introduce an online algorithm that periodically measures the model performance on the held-out set and then dynamically adjusts the self-training regularizer to effectively optimize the desired metric objective. We demonstrate that \emph{CSST} can generalize to self-training methods such as FixMatch and UDA~\cite{xie2020self} on computer vision and natural language processing (NLP) tasks.

        \item \textbf{Selective Mixup Fine-Tuning for Metric Optimization.} In recent years, it has been established that general-purpose self-supervised pre-training leads to more robust models~\cite{caron2021emerging, hendrycks2019using}, which could be fine-tuned for downstream tasks. However, in long-tail semi-supervised settings, many works tends to optimize a single metric objective from scratch. This leads to lack of robustness and a computation overhead of training model for each objective from scratch. To mitigate this in Chapter~\ref{chap:selmix}, we introduce \emph{SelMix}, a fine-tuning technique based on mixup~\cite{zhang2017mixup} to optimize the desired non-decomposable metric objective. Our main idea is to find a distribution of class labels to be mixed up for maximal gain in the desired objective. After finding the optimal mixup distribution using a held-out set, we train the model using the mixup distribution. We demonstrate the effectiveness of SelMix across various benchmarks and settings in long-tailed learning.
        
    \end{itemize}

    \vspace{1mm} \noindent  \textbf{4. Efficient Domain Adaptation.} In earlier work, the main assumption is that the distribution of the samples at test time is the same as one used for training. This is impractical, as the data distributions continuously change over time and when the model is deployed in various domains. In Part~\ref{part:ef_DA}, we study the problem of efficiently adapting the model to generalize to the target domain. We first discuss a technique to select a subset of samples to maximize target adaptation performance (semi-supervised). We then introduce a regularizer so the model learns generalizable solutions without requiring target labels.  

    \begin{itemize}
          \item \textbf{Submodular Subset Selection for Domain Adaptation.} In Chapter~\ref{chap:S3VAADA}, we aim to select a subset of samples to adapt the model to the target domain for maximally increasing target performance. To select a subset, we introduce submodular subset criteria based on the uncertainty, diversity, and representativeness of the selected samples. Due to submodularity, we can perform set selection in linear time through a greedy algorithm. After selecting the samples, we utilize a Virtual Adversarial Domain Adaptation technique to ensure that the model can gain from the target samples maximally.

        \item \textbf{Smooth Domain Adversarial Training.} In Chapter~\ref{chap:SDAT}, we take a closer look at the curvature properties of domain adversarial training. In domain adversarial training, in addition to cross-entropy loss on the source domain, there is also an adversarial loss on a discriminator, which ensures that the target domain and source domain have similar feature distribution. We find that converging to a flat curvature region for the cross-entropy loss in the source domain, stabilizes training and leads to better adaptation in the target domain. On the contrary, smoothing the adversarial loss leads to suboptimal results. Based on this, we introduce \emph{Smooth Domain Adversarial Training (SDAT)} for domain adaptation, demonstrating significant improvement in DA tasks across benchmarks.

    \end{itemize}

\vspace{1mm} \noindent \textbf{Conclusion} In Chapter~\ref{chap:conclusion}, we provide a summary of the works and the conclusions made in each part of this thesis. We also discuss some open problems and avenues that could be explored in further research, related to the research contributions presented in this thesis.

\part{Training GANs on Long-Tailed Datasets}
\label{part:GAN_LT}

\chapter{Class Balancing GAN with a Classifier in the Loop}
\label{chap:cbgan}

\begin{changemargin}{7mm}{7mm} 

Generative Adversarial Networks (GANs) have swiftly evolved to imitate increasingly complex image distributions. However, majority of the developments focus on performance of GANs on balanced datasets. We find that the existing GANs and their training regimes which work well on balanced datasets fail to be effective in case of imbalanced (i.e. long-tailed) datasets. In this work we introduce a novel theoretically motivated Class Balancing regularizer for training GANs. Our regularizer makes use of the knowledge from a pre-trained classifier to ensure balanced learning of all the classes in the dataset. This is achieved via modelling the effective class frequency based on the exponential forgetting observed in neural networks and encouraging the GAN to focus on underrepresented classes. We demonstrate the utility of our regularizer in learning representations for long-tailed distributions via achieving better performance than existing approaches over multiple datasets. Specifically, when applied to an unconditional GAN, it improves the FID from $13.03$ to $9.01$ on the long-tailed iNaturalist-$2019$.
\end{changemargin}

\section{Introduction}
\begin{figure*}[t]
\centering
\includegraphics[width=0.98\textwidth, height=3cm]{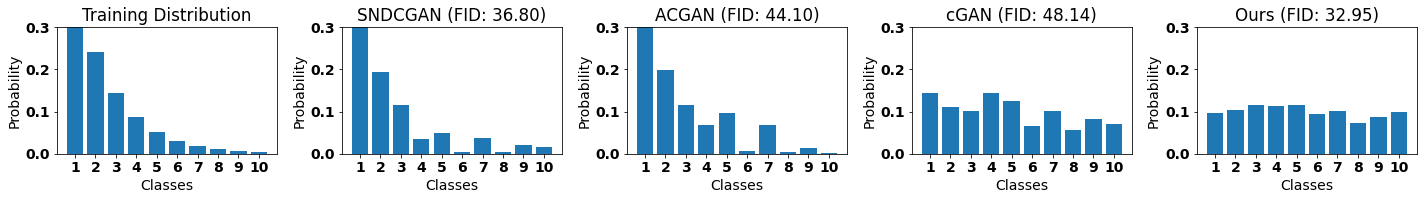}
\caption{Distribution of classes and corresponding FID scores on long-tailed CIFAR-10. SNDCGAN and ACGAN tend to produce arbitrary distributions which are biased towards majority classes, whereas cGAN samples suffer in quality with high FID. Our method achieves low FID and a balanced distribution at the same time. }
\label{cbgan_fig:dist_stats}
\end{figure*}
\label{cbgan_sec:introduction}
Image Generation witnessed unprecedented success in recent years following the invention of Generative Adversarial Networks (GANs) by~\citet{goodfellow2014generative}. GANs have improved significantly over time with the introduction of better architectures~\citep{gulrajani2017improved,radford2015unsupervised}, formulation of superior objective functions~\citep{jolicoeur2018relativistic,arjovsky2017wasserstein}, and regularization techniques~\citep{miyato2018spectral}. An important breakthrough for GANs has been the ability to effectively class conditioning for synthesizing images~\citep{mirza2014conditional,miyato2018cgans}. Conditional GANs have been shown to scale to large datasets such as ImageNet~\citep{imagenet_cvpr09} dataset with $1000$ classes~\citep{miyato2018cgans}. 

One of the major issues with unconditional GANs has been their inability to produce 
balanced distributions over all the classes present in the dataset. This is seen as 
problem of missing modes in the generated distribution. A version of the
missing modes problem, known as the `covariate shift' problem was studied by~\citet{santurkar2018classification}. One possible reason is the absence of knowledge about the class distribution $P(Y|X)$\footnote{Here Y represents labels and X represents data.} of the generated samples during training. Conditional GANs on the other hand, do not suffer from this issue since the class labels $Y$ are supplied to the GAN during training. However, it has been recently found by ~\citet{ravuri2019classification} that despite being able to do well on metrics such as Inception Score (IS)~\citep{salimans2016improved} and Fr\`{e}chet Inception Distance (FID)~\citep{heusel2017gans}, the samples generated from the state-of-the-art conditional GANs lack the diversity in comparison to the underlying training datasets. Further, we observe that, although conditional GANs work well in balanced case, they suffer performance degradation in the imbalanced case (Table-\ref{cbgan_tab:gan_result}). \\
In order to address these shortcomings, we propose a novel method of inducing  the information about the class distribution. We estimate the class distribution $P(Y|X)$ of generated samples in the GAN framework using a pre-trained classifier. The regularizer utilizes the estimated class distribution to penalize excessive generation of samples from the majority classes, thereby enforcing the GAN to also generate samples from minority classes. Our regularizer involves a novel method of modelling the forgetting of samples by GANs, based on the exponential forgetting observed in neural networks~\citep{kirkpatrick2017overcoming}. We show the implications of our regularizer by a theoretical upper bound in Section~\ref{cbgan_sec:method}.

We also experimentally demonstrate the effectiveness of the proposed class balancing regularizer in the scenario of training GANs for image generation on long-tailed datasets, including the large scale iNaturalist-2019 \citep{inat19} dataset. Generally, even in the long-tailed distribution tasks, the test set is balanced despite the imbalance in the training set. This is because it is important to develop Machine Learning systems that generalize well across all the support regions of the data distribution, avoiding undesired over-fitting to the majority (or head) classes.  

In summary, our contributions can be listed as follows:

\begin{itemize}
\itemsep=0em
    
    \item We propose a `class-balancing' regularizer that makes use of the statistic $P(Y|X)$ of generated samples to promote uniformity while sampling from an unconditional GAN. The effect of our regularizer is depicted both theoretically (Section~\ref{cbgan_sec:method}) and empirically (Section~\ref{cbgan_sec:Experiments}).
    \item We show that our regularizer enables GANs to learn uniformly across classes even when the training distribution
    is long-tailed. We observe consistent gains in FID and accuracy of a
    classifier trained on the generated samples.
    
    \item Our method is able to scale to large and naturally occurring datasets such as iNaturalist-$2019$ and, achieves state-of-the-art FID score of 9.01.
\end{itemize}

\section{Background}
\label{cbgan_sec:background}
\subsection{Generative Adversarial Networks (GANs)}
Generative Adversarial Network (GAN) is a two player game in which the discriminator network $D$ tries to classify images into two classes: real and fake. The generator network $G$ tries to
generate images (transforming a noise vector $z \sim P_z$ ) which fool the discriminator $D$ into classifying them as real. The game can be 
formulated as the following mathematical objective:
\begin{equation}
    \underset{G}{min} \; \underset{D}{max}\; E_{x \sim P_{r}}[\log(D(x))] + E_{z \sim P_{z}}[\log(1 - D(G(z))].
\end{equation}
The exact inner optimization for training $D$ is computationally prohibitive in large networks; hence the GAN is trained through alternate minimization of loss functions.
Multiple
loss functions have been proposed for stabilizing GAN training. In our work we use the relativistic loss function~\citep{jolicoeur2018relativistic} which is
formulated as:
\begin{equation}
    L_{D}^{rel} = - E_{(x, z) \sim (P_r, P_z)}[\log(\sigma(D(x) - D(G(z)))]
\end{equation}
\begin{equation}
L_{G}^{rel} = - E_{(x, z) \sim (P_r, P_z)}[\log(\sigma(D(G(z)) - D(x))].
\end{equation}
\noindent \textbf{Issue in the long-tailed scenario}: This unconditional GAN formulation does not use any class information $P(Y|X)$ about the images and tends to produce different number of samples from different classes~\citep{santurkar2018classification}. In other words, the generated distribution is not balanced (uniform) across different classes. This issue is more severe when the training data is long-tailed, where the GAN might completely ignore learning some (minority) classes (as shown in the SNDCGAN distribution of Figure~\ref{cbgan_fig:dist_stats}).
\subsection{Conditional GAN}
The conditional GAN~\citep{mirza2014conditional} generates images associated with a given input label
$y$ using the following objective:
\begin{equation}
    \label{cbgan_equation:cgan}
    \underset{G}{\min} \; \underset{D}{\max}\; E_{x \sim P_{r}}[\log(D(x|y))] + E_{z \sim P_{z}}[\log(1 - D(G(z|y))].
\end{equation}
The Auxillary Classifier GAN (ACGAN) ~\citep{odena2017conditional} uses an auxiliary classifier for classification along with a discriminator to enforce high confidence samples from the conditioned class $y$. Whereas cGAN with projection~\citep{miyato2018cgans} uses 
Conditional Batch Norm~\citep{de2017modulating} in the generator and a projection step in the discriminator
to provide class information to the GAN. We refer to this method as cGAN in the subsequent sections.\\
\begin{figure}[!t]
    \centering\includegraphics[width=.75\linewidth]{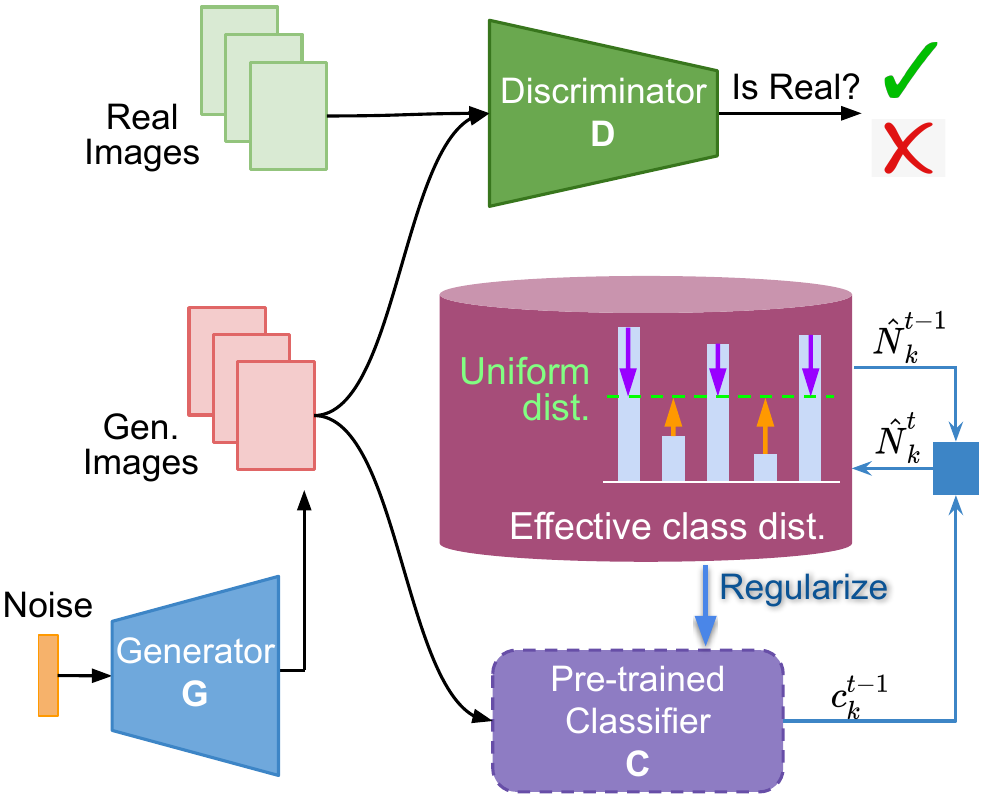}
  \caption{{Class Balancing Regularizer} aims to help GANs generate balanced distribution of samples across classes even when they are trained on imbalanced datasets. The method achieves this by keeping an estimate of effective class distribution of generated images using a pre-trained classifier. The GAN is then incentivized to generate images from underrepresented classes, which moves the GAN towards generating uniform distribution.}
  \label{cbgan_fig:approach_overview}
\end{figure}

\noindent \textbf{Issue with Conditional GAN in Long-tailed Setting}:
The objective in eq.(\ref{cbgan_equation:cgan}) can be seen as learning
a different $G(z|y)$ and $D(x|y)$ for each of the $K$ classes. In this
case the tail classes with fewer samples can suffer from poor learning of $G(z|y)$ as there are very few samples for learning. However, in practice there is parameter sharing among different class generators, yet there are class specific parameters present in the form of Conditional BatchNorm. We find that performance of conditional GANs degrades more in comparison to unconditional GANs in the long-tailed scenario (observed in Table~\ref{cbgan_tab:gan_result}).

\section{Method}
\label{cbgan_sec:method}
In our method we use a pretrained classifier $(C)$ to provide feedback to the generator about the label distribution $P(Y|X)$ through the proposed regularizer. One can train the classifier $(C)$ before the GAN, on the underlying long-tailed dataset. However, if the full set of labels are not available, we show (Section~\ref{cbgan_subsec:semi-supervised}) that a limited set of labeled data is sufficient for training the classifier. The proposed regularizer term is then added to the generator loss and trained using backpropogation. 
We first describe the method of modelling the class distribution in Section~\ref{cbgan_class_statistics}. Exact formulation of the regularizer and its theoretical properties are described in Section~\ref{cbgan_sub:formulation}. The overview of our method is presented in Figure~\ref{cbgan_fig:approach_overview}.
\subsection{Class Statistics for GAN}
\label{cbgan_class_statistics}
GAN is a dynamic system in which the generator $G$ has to continuously adapt itself in a way that it is able
to fool the discriminator $D$. During the training, discriminator $D$ updates itself, which changes the objective for the generator $G$. This change in objective can
be seen as learning of different tasks for the generator $G$. In this context, we draw motivation from the 
seminal work on catastrophic forgetting in neural networks~\citep{kirkpatrick2017overcoming} which shows that a neural network trained using SGD suffers from exponential forgetting of
earlier tasks when trained on a new task. 
Based on this, we define \emph{effective class frequency} $\hat{N_k^t}$ of class $k$ at a training cycle $t$ as:
\begin{equation}
    \hat{N_k^t} = (1 - \alpha) \hat{N_k}^{t-1} + \beta c_k^{t-1} 
    \label{cbgan_eq:update_equation}
\end{equation}
here $c_k^{t -1}$ is the number of samples of class $k$ produced by the GAN in cycle $(t-1)$ and $\hat{N_k}^0$ is initialized to a constant for all $k$ classes. The class to which the generated sample belongs to is determined by a pretrained classifier $C$. We find that using exponential decay, using either ($\beta = 1$) or a convex combination ($\beta=\alpha$) is sufficient for all the experiments. Although $D$ gets updated continuously, the update is slow and requires some iterations to change the form of $D$. Hence we update the  statistics after certain number of iterations which compose a cycle. Here $\alpha$ is the exponential forgetting factor which is set to $0.5$ as default in our experiments. We normalize the $\hat{N_k^t}$ to obtain discrete \textit{effective class distribution $N_k^t$}:
\begin{equation}
    N_k^t =  \frac{\hat{N_k^{t}}}{\sum_k \hat{N_k^{t}}}.
\end{equation}

\subsection{Regularizer Formulation}
\label{cbgan_sub:formulation}
The regularizer objective is defined as the minimization of the term ($L_{reg}$) below:
\begin{equation}
    \underset{\hat{p} }{\min}  \; \sum_{k} \frac{\hat{p}_k\log(\hat{p}_k)}{N_k^t}  
     \label{cbgan_equation:regularizer}
\end{equation}
where $ \hat{p} = \sum_{i = 1}^{n} \frac{C(G(z_i))}{n}$ and $z_i \sim P_z$. In other words, $\hat{p}$ is the average softmax
vector (obtained from the classifier $C$) over the batch of $n$ samples and $\hat{p_k}$ is its $k^{th}$ component corresponding to class $k$. If the classifier $C$ recognizes the samples confidently with probability $\approx 1$,  $\hat{p}_k$ can be seen as the approximation to the ratio of the number of samples that belong to class $k$ to the total number of samples in the batch $n$. $N_k^t$ in the regularizer term is obtained through the update rule in Section \ref{cbgan_class_statistics} and is a constant during backpropagation. The regularizer objective in Eq.~(\ref{cbgan_equation:regularizer}) when multiplied with a negative unity, can also be interpreted as maximization of the weighted entropy computed over the batch.

\noindent \textbf{Proposition 1:}
The minimization of the proposed objective in (\ref{cbgan_equation:regularizer}) leads to the following bound on $\hat{p_k}$: 
\begin{equation}
    \hat{p_k} \leq e^{-K(log(K) - 1)\frac{N_k^t}{\sum_{k}{N_k^t}} -1}
\end{equation}

where $K$ is the number of distinct class labels produced by the classifier $C$.%

\noindent \textbf{Proof:}
\begin{equation}
    \displaystyle \underset{\hat{p}}{\min} \; \sum_{k} \frac{\hat{p_k}\log(\hat{p_k})}{N_k^t} 
\end{equation}
Introducing the probability constraint and the Lagrange multiplier $\lambda$:
\begin{equation}
    \displaystyle L(\hat{p}, \lambda) = \sum_{k} \frac{\hat{p_k}\log(\hat{p_k})}{N_k^t} - \lambda (\sum{\hat{p_k}} - 1) 
\end{equation}
On solving the equations obtained by setting $\displaystyle \frac{\partial L}{\partial \hat{p_k}} = 0:$
\begin{equation}
    \label{cbgan_eq:lambda_val}
    \frac{1}{N_k^t} + \frac{\log(\hat{p_k})}{N_k^t} - \lambda = 0 \implies \hat{p_k} = e^{\lambda N_k^t - 1}
\end{equation}
Using the constraint  $\displaystyle \frac{\partial L}{\partial \lambda} = 0$ we get:
\begin{equation}
    \sum_{k} \hat{p_k} = 1 \implies \sum_{k} e^{\lambda N_k^t - 1} = 1 \implies \sum_{k} e^{\lambda N_k^t} = e
\end{equation}
Now we normalize both sides by $K$, the number of distinct labels produced by classifier and apply Jensen's inequality for concave function $\psi (\frac{\sum a_ix_i}{\sum a_i}) \geq \frac{\sum a_i\psi(x_i)}{\sum a_i}$ and take $\psi$ as $\log$ function:
\begin{equation}
\displaystyle    \frac{e}{K} = \sum_{k}\frac{e^{\lambda N_k^t}}{K} \implies \log (\frac{e}{K}) = \log(\sum_{k}\frac{e^{\lambda N_k^t}}{K}) \geq \sum_{k} \frac{\lambda N_k^t}{K}
\end{equation}
On substituting the value of $\displaystyle \lambda$ in inequality from Eq. \ref{cbgan_eq:lambda_val}:
\begin{align}
    K(1 - \log(K))  \geq \lambda \sum_{k} N_k^t \implies  K(1 -\log(K))  \geq  (\sum_{k} N_k^t) \frac{1 + \log(\hat{p_k})}{N_k^t}
\end{align}
On simplifying and exponentiation we get the following result:
\begin{equation}
    \displaystyle \hat{p_k} \leq e^{-K(\log(K) - 1) \frac{N_k^t}{\sum_k N_k^t} - 1}.
\end{equation}

The penalizing factor $K(\log(K) - 1)$ is increasing
in terms of number of classes $K$ in the dataset. This helps the overall objective since we need a large penalizing factor to compensate for as $N_k^t/\sum_k N_k^t$ will be smaller when number of classes is large in the dataset. Also, in case of generating a balanced distribution, $\hat{p_k} = 1/K$ which
leads to the exponential average $N_k^t = 1/K$ given sufficient iterations. In this case the upper bound value will be  $1/K$ which equals the value of $\hat{p_k}$, proving that the given bound is tight. We would like to highlight that the proof is valid for any $N_k^t > 0$ but in our case $\sum_k N_k^t = 1$. 
\\

\noindent \textbf{Implications of the proposition 1}:
  \begin{figure}[!t]%
    \centering\includegraphics[width=0.75\linewidth]{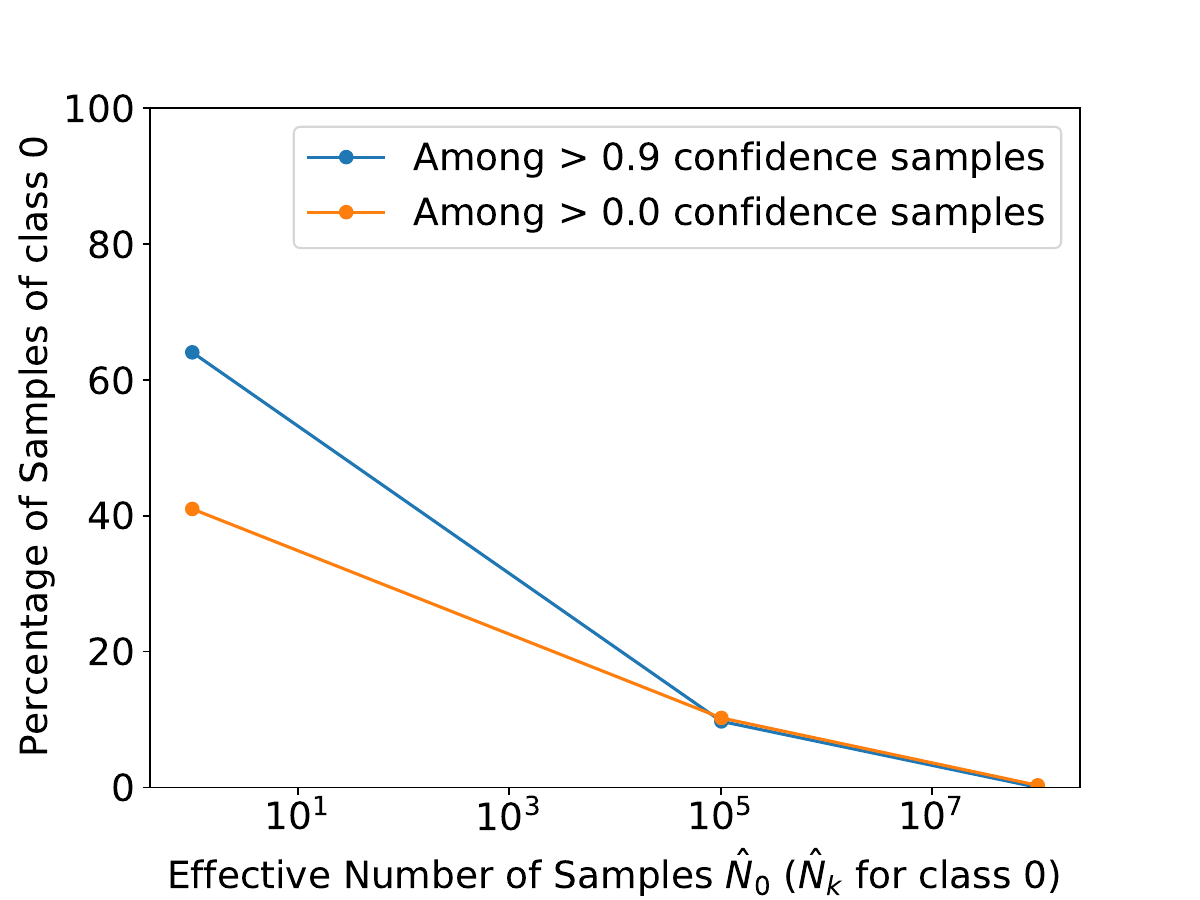}
    \caption{\label{cbgan_fig:toy_exp} Shows the percentage of generated samples for class $0$ by SNDCGAN on CIFAR-10 for varying values of effective class frequency $\hat{N_0}$. When $\hat{N_0}$ is large, the network tries to decrease fraction of class 0 samples whereas when $\hat{N_0}$ is small it tries to increase fraction of class 0 samples among the generated samples. The blue and orange lines respectively correspond to the percentage of class $0$ samples, in $>0.9$ confidence samples and all the samples.}
  \end{figure}
The bound on $\hat{p_k}$ is inversely proportional to the exponent of the fraction of effective class distribution $N_k^t$ for a given class $k$. To demonstrate the effect of our regularizer empirically, we construct two extreme case
examples:

\begin{itemize}
\itemsep0em
\item If $N_k^t \gg N_i^t$, $\forall i \neq k, N_k^t \approx 1$, then the bound on $\hat{p_k}$ would approach $e^{-K(\log(K) - 1) - 1}$. Hence the network is expected to decrease the proportion of class $k$ samples.
    
\item If $N_k^t \ll N_i^t$, $\forall i \neq k, N_k^t \approx 0$, then the bound on $\hat{p_k}$ will be $e^{-1}$.  Hence the network might increase the proportion of class $k$ samples.
\end{itemize}
We verified these two extreme cases by training a SNDCGAN~\citep{miyato2018spectral} (DCGAN with spectral normalization, hyperparameters defined in Appendix Table~\ref{cbgan_app: hyperparams})  on CIFAR-10 and fixing $\hat{N_k^t}$ (unnormalized version of $N_k^t$) across time steps and denote it as $\hat{N_k}$. We run two experiments by initialising $\hat{N_k}$ to a very large value and 
a very small value. Results presented in Figure~\ref{cbgan_fig:toy_exp} show that the GAN increases the proportion of samples of class $k$ in case of low $\hat{N_k}$
and decreases the proportion of samples in case of large $\hat{N_k}$. This shows the balancing behaviour of proposed regularizer. The diversity regularizer proposed in by authors of DeGAN~\cite{addepalli2020degan} is a unique case for the proposed regularizer with ($N_k^t = 1$).\\

\noindent \textbf{Proposition 2 \citep{guiacsu1971weighted}:} If each  $\hat{p_k} = e^{\lambda N_k^t - 1}$ where $\lambda$ is obtained from solution of $\sum_{k}e^{\lambda N_k^t - 1} = 1$, then the regularizer objective in  Eq. \ref{cbgan_equation:regularizer} attains the optimal minimum value of $\lambda -\sum_{k} \frac{e^{\lambda N_k^t - 1}}{N_k^t}$. For proof please refer to Appendix~\ref{cbgan_app:proof2}.

\noindent \textbf{Implications of Proposition 2:} As we only use necessary conditions to prove the bound in proposition 1, the optimal solution can be a maxima, minima or a saddle point. Prop. 2 result shows that the optimal solution found in Prop. 1 (i.e. $\hat{p_k} = e^{\lambda N_k^t - 1}$) is inded optimal. 
\subsection{Combining the Regularizer and GAN Objective}
The regularizer can be combined with the generator loss in the following way:
\begin{equation}
    L_{g} = - E_{(x, z) \sim (P_r, P_z)}[\log(\sigma(D(G(z)) - D(x)))] + \lambda L_{reg}
\end{equation}
It has been recently shown~\citep{jolicoeur2018rfdiv} that the first term of the loss leads to minimization of $D_f(P_g, P_r)$ that is f-divergence
between real ($P_r$) and generated data distribution ($P_g$). The regularizer term ensures that the distribution of classes across generated
samples is uniform. The combined objective provides insight into the working of framework, as the first term ensures that the generated images fall in the image distribution and the second term ensures that the
distribution of classes is uniform. So the first term leads to \textit{divergence minimisation} to real data while satisfying the \textit{constraint} of class balance (i.e. second term), hence overall it can be seen as \textit{constained optimization}. As $P_r$ comprises of diverse samples from majority class the first objective
term ensures that $P_g$ is similarly diverse. The second term in the objective ensures that the discriminative properties
of all classes are present uniformly in the generated distribution, which ensures that 
minority classes get benefit of diversity present in the majority classes. This is analogous to approaches that transfer knowledge from
majority to minority classes for long-tailed classifier learning~\citep{Liu_2019_CVPR,wang2017learning}.
\begin{table}[t]%
    \centering
    \resizebox{\columnwidth}{!}{%
    \begin{tabular}{lccccccc}
    \hline
    Imbalance Ratio &  \multicolumn{3}{c}{100} & \multicolumn{3}{c}{10} &  \multicolumn{1}{c}{1}\\\hline
                & FID ($\downarrow$)&      KLDiv($\downarrow$)&  Acc.($\uparrow$)& FID($\downarrow$)&      KLDiv($\downarrow$)&  Acc.($\uparrow$)& FID ($\downarrow$)    \\\hline %
   \multicolumn{8}{c}{CIFAR-10} \\ \hline
         SNDCGAN	&36.97 $\pm$ 0.20&	0.31 $\pm$ 0.0& 68.60 & 32.53 $\pm$ 0.06& 	0.14 $\pm$ 0.0 &  80.60&	27.03 $\pm$ 0.12\\
         ACGAN	&44.10 $\pm$ 0.02 &	0.33 $\pm$ 0.0&	43.08& 38.33 $\pm$ 0.10&	0.12 $\pm$ 0.0&	60.01& 24.21 $\pm$ 0.08 \\
         cGAN	& 48.13 $\pm$ 0.01&	0.02 $\pm$ 0.0&	47.92& 26.09 $\pm$ 0.04&	0.01 $\pm$ 0.0&	68.34& 18.99 $\pm$ 0.03\\
         \rowcolor[gray]{0.85}
         Ours	&32.93 $\pm$ 0.11 &	0.06 $\pm$ 0.0&	72.96& 30.48 $\pm$ 0.07&	0.01 $\pm$ 0.0& 82.21& 25.68$\pm$ 0.07 \\

         \hline 
        \multicolumn{8}{c}{LSUN} \\ \hline    
         SNResGAN	&37.70 $\pm$ 0.10 &	0.68 $\pm$ 0.0 &	75.27& 33.28 $\pm$ 0.02 & 	0.29 $\pm$ 0.0& 79.20&	28.99 $\pm$ 0.03 \\
         ACGAN	&43.76 $\pm$ 0.06&	0.39 $\pm$ 0.0 &	62.33& 31.98 $\pm$ 0.02&	0.05 $\pm$ 0.0&	75.47& 26.43 $\pm$ 0.04\\
         cGAN	&75.39 $\pm$ 0.12&	0.01 $\pm$ 0.0   & 	44.40 & 30.68 $\pm$ 0.04&	0.00 $\pm$ 0.0&	72.93 & 27.59 $\pm$  0.03 \\
         \rowcolor[gray]{0.85}
         Ours	&35.04 $\pm$ 0.19&	0.06 $\pm$ 0.0&	77.93& 28.78 $\pm$ 0.01&	0.01 $\pm$ 0.0&	82.13& 28.15 $\pm$ 0.05\\

      \hline

    \end{tabular}}
    \caption{Results on CIFAR-10 (top panel) and $5$ class subset of LSUN (bottom panel) datasets with varying imbalance. In the last column FID values in balanced scenarios are present for
    ease of reference. FID, KL Div. and Acc. are calculated on $50$K sampled images from each.}
    \label{cbgan_tab:gan_result}
\end{table}

\label{cbgan_formulation}
\section{Experiments}
\label{cbgan_sec:Experiments}
For evaluating the effectiveness of our balancing regularizer, we conduct image generation experiments over long-tailed distributions. %
In these experiments we aim to train a GAN using data from a long-tailed dataset, which is common in the real world setting. For achieving good performance on all classes in the dataset, the method requires to transfer knowledge from majority to minority classes. Several works have focused on learning classifiers on long-tailed distributions~\citep{cao2019learning,cui2019classbalancedloss, Liu_2019_CVPR}. Yet, works focusing on Image Generation using long-tailed dataset are limited. Generative Minority Oversampling (GAMO)~\citep{Mullick_2019_ICCV} attempts to solve the problem
by introducing a three player framework, which is an encoder-decoder network and not a GAN. We do not compare our results
with GAMO as it is not
trivial to extend GAMO to use schemes such as spectral normalization \citep{miyato2018spectral}, and ResGAN-like architecture~\citep{gulrajani2017improved} which impede a fair comparison to our experiments.

\noindent \textbf{Datasets}: We perform extensive experimentation on CIFAR-10 and a subset of LSUN, as these are widely used for evaluating GANs. The LSUN subset consists of $250$K training images and $1.5$K validation images. 
The LSUN subset is composed of $5$ balanced classes. \citet{santurkar2018classification} identified this subset to be a challenging case for GANs to generate uniform distribution of classes.
The original CIFAR-10 dataset is composed of $50$K training images and $10$K validation images. We construct the
long-tailed version of the datasets by following the same 
procedure as~\citet{cao2019learning}.
Here, images are removed from training set to convert it
to a long-tailed distribution while the validation set is kept unchanged.
The imbalance ratio $(\rho)$ determines the ratio of number of samples in most populated
class to the least populated one, which can be mathematically stated as: $\displaystyle \rho = {max_{k}\{n_k\}}/{min_{k}\{n_k\}}$.%

\noindent \textbf{Pre-Trained Classifier}:
\label{cbgan_sec:pretrained_classifier}
An important component of our framework is the pre-trained classifier. All the pre-trained classifiers in our experiments use a ResNet32~\citep{he2016deep} architecture.
The classifier is trained using Deferred Re-Weighting (DRW) scheme~\citep{cao2019learning,cui2019classbalancedloss} on the long-tailed data. We use the available open source code\footnote{https://github.com/kaidic/LDAM-DRW}. We use the following learning 
rate schedule: initial learning rate of $0.01$ and multiplying by
$0.01$ at epoch $160$ and $180$. We train the models for 
$200$ epochs and start reweighting (DRW) at epoch $160$. We give a summary of the validation accuracy of the models for various imbalance ratios $(\rho)$ in Table~\ref{cbgan_tab:clf_acc}.
\begin{figure*}[!t]
\centering
\begin{subfigure}{.4\textwidth}
  \centering
  \includegraphics[width=.8\linewidth]{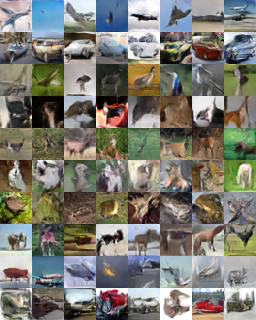}  
  \caption{ACGAN (Conditional)}
  \label{cbgan_fig:sub-first}
\end{subfigure}%
\begin{subfigure}{.4\textwidth}
  \centering
  \includegraphics[width=.8\linewidth]{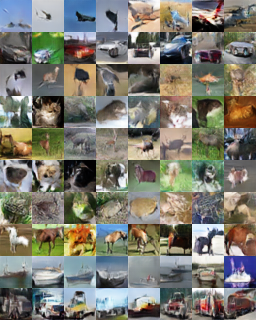}  
  \caption{cGAN (Conditional)}
  \label{cbgan_fig:sub-second}
\end{subfigure}

\begin{subfigure}{.4\textwidth}
  \centering
  \includegraphics[width=.8\linewidth]{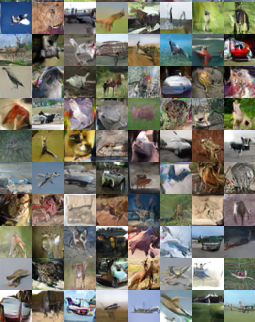}  
  \caption{SNDCGAN (Unconditional)}
  \label{cbgan_fig:sub-third}
\end{subfigure}
\begin{subfigure}{.4\textwidth}
  \centering
  \includegraphics[width=.8\linewidth]{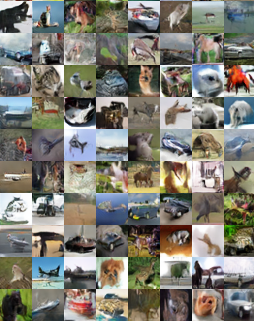}  
  \caption{Ours (Unconditional)}
  \label{cbgan_fig:sub-fourth}
\end{subfigure}
\caption{Images generated by different GANs for CIFAR-10 with imbalance ratio ($\rho = 10$).}
\label{cbgan_fig:cifar10}
\end{figure*}

\begin{table}[h]
    \centering
    \begin{tabular}{lccc}
         \hline
         Imbalance Ratio $(\rho)$ & 100 & 10 & 1  \\ \hline
         CIFAR-10 & 76.67 & 87.70 & 92.29  \\ 
         LSUN & 82.40 & 88.07 &  90.53\\ \hline
    \end{tabular}
    \caption{Validation Accuracy of the PreTrained Classifiers used with GANs. The balanced classifier also serves as an annotator.}
    \label{cbgan_tab:clf_acc}
\end{table}

\noindent \textbf{GAN Architecture}: 
We use the SNDCGAN architecture for experiments on CIFAR-10 with images of size of $32 \times 32$ and SNResGAN (ResNet architecture with spectral normalization) structure for experiments on LSUN dataset with images of size $64 \times 64$. For the conditional GAN baselines we conditioned the generator using Conditional BatchNorm.
We compare our method to two widely used conditional GANs: ACGAN and cGAN. The other baseline we use is the unconditional GAN (SNDCGAN \& SNResGAN) without our regularizer. All the GANs were trained with spectral normalization in the discriminator for stabilization~\citep{miyato2018spectral}.

\noindent \textbf{Training Setup:}
We use the learning rate of $0.0002$
for both generator and discriminator. We use Adam optimizer with $\beta_1 = 0.5$
and $\beta_2 = 0.999$ for SNDCGAN and $\beta_1 = 0$ and $\beta_2 = 0.999$
for SNResGAN. We use a batch size of $256$ and perform
$1$ discriminator update for every generator update. As a sanity check, we use
the FID values and visual inspection of images on the balanced dataset and verify the range of values from~\citet{kurach2019large}. We update the statistics $N_k^t$ using Eq. \ref{cbgan_eq:update_equation} after every 2000 iterations, for all experiments in Table \ref{cbgan_tab:gan_result}. The code is implemented with PyTorch Studio GAN framework \cite{kang2020contrastive}. Further details and ablations are present in the Appendix. %

\noindent \textbf{Evaluation} We used the following evaluation metrics:

\noindent\textbf{1. KL Divergence w.r.t. Uniform Distribution of labels}: Labels for the generated samples are obtained by using the pre-trained classifier (trained on balanced data) as an annotator. The annotator is just used for evaluation on long-tailed data. Low values of this metric signify that the generated samples are uniformly distributed across classes. \\
\noindent \textbf{2. Classification Accuracy (CA)}: We use the $\{(X, Y)\}$ pairs from the GAN generated samples to train a ResNet32 classifier and test it on real data. For unconditional GANs, the label $Y$ is obtained from the classifier trained on long-tailed data. Note that this is similar to Classifier Accuracy Score of \citep{ravuri2019classification}.  \\
\noindent \textbf{3. Fr\`{e}chet Inception Distance (FID)}: It measures the 2-Wasserstein Distance on distributions obtained from Inception Network~\citep{heusel2017gans}. We use $10$K samples from CIFAR-10 validation set and $10$K ($2$K from each class) fixed random images from
LSUN dataset for measuring FID.\\

\noindent \textbf{Discussion of Results:} We present our results below: \\
1) \textbf{Stability}: In terms of stability, we find that cGAN suffers from 
early collapse in case of high imbalance ($\rho = 100$) and stops improving within $10$K iterations. Therefore, although cGAN is stable in balanced scenario, it
is unstable in case of long-tailed version of the given dataset.\\
2) \textbf{Biased Distribution}: Contrary to cGAN, we find that the distribution of classes generated by ACGAN, SNDCGAN and SNResGAN is imbalanced. The images obtained by sampling uniformly and labelling by annotator, suffers from a high KL 
divergence to the uniform distribution. Some classes are almost absent from the distribution of generated samples as shown in Figure \ref{cbgan_fig:dist_stats}. In this case, in Table~\ref{cbgan_tab:gan_result} we observe FID score
just differs by a small margin even if there is a large imbalance in class distribution. This happens for ACGAN as its loss is composed of GAN loss and classification loss terms, therefore in the long-tailed setting the ACGAN gets biased towards optimising GAN loss ignoring classification loss, hence tends to produce arbitrary distribution.  Contrary to this, our GAN produces class samples uniformly as it is evident from the low KL Divergence.\\
3) \textbf{Comparison with State-of-the-Art Methods}: In this work we observe that classification accuracy is weakly correlated with FID score
which is in agreement with~\citet{ravuri2019classification}. We achieve better classifier accuracy when compared to
cGAN in all cases, which is the current state-of-the-art for Classifier Accuracy Score (CAS). Our method shows minimal degradation in FID in comparison to the corresponding balanced case. It is also able to achieve the best FID in $3$ out of $4$ long-tailed cases. Hence we expect that methods such as Consistency Regularization~\citep{zhang2019consistency} and Latent Optimization~\citep{wu2019logan} can be applied in conjunction with our method to further improve
the quality of images. However in this work we specifically focused
on techniques used to provide class information $(Y)$ of the images $(X)$ to the GAN. Several state-of-the-art GANs use the approach of cGAN~\citep{wu2019logan,brock2018large} for conditioning the discriminator
and the generator. We provide a qualitative comparison of the images produced in Fig.~\ref{cbgan_fig:cifar10}, in which it can be observed that CBGAN produces diverse and high-quality images compared to other state-of-the-art baselines.

\begin{figure*}[t]
     \centering
     \begin{subfigure}[b]{0.22\textwidth}
         \centering
         \includegraphics[width=\textwidth]{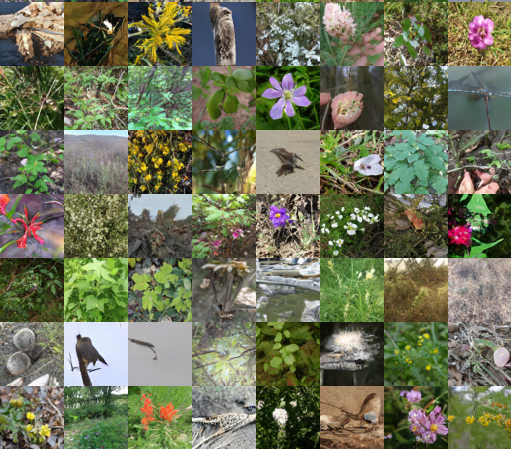}
         \caption{SNResGAN(13.03)}

     \end{subfigure}
     \hfill
     \begin{subfigure}[b]{0.22\textwidth}
         \centering
         \includegraphics[width=\textwidth]{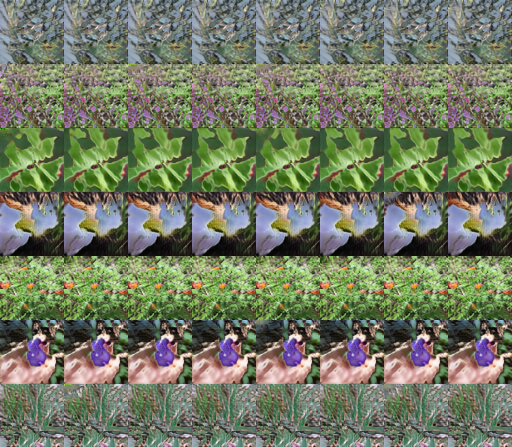}
         \caption{ACGAN(47.15)}
         \label{cbgan_fig:three sin x}
     \end{subfigure}
     \hfill
     \begin{subfigure}[b]{0.22\textwidth}
         \centering
         \includegraphics[width=\textwidth]{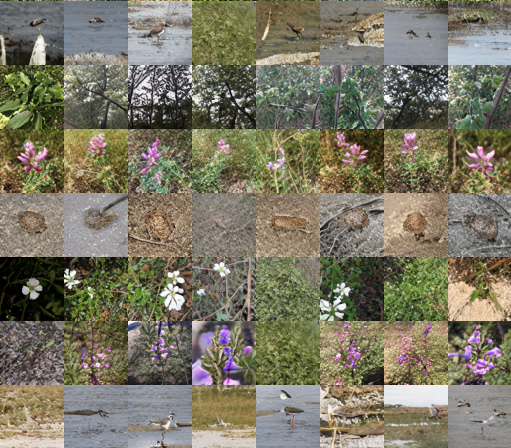}
         \caption{cGAN (21.53)}
         \label{cbgan_fig:five over x}
     \end{subfigure}
     \hfill
      \begin{subfigure}[b]{0.22\textwidth}
         \centering
         \includegraphics[width=\textwidth]{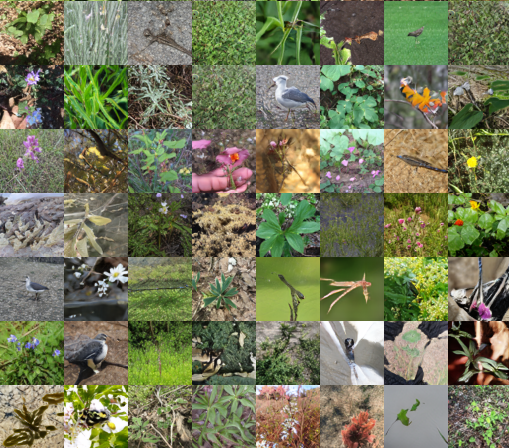}
         \caption{Ours (9.01)}
         
     \end{subfigure}
        \caption{Shows the $64 \times 64$ generated images for the iNat-2019 dataset for different baselines.}
        \label{cbgan_fig:inat_visual}
    
\end{figure*}

\subsection{Experiments on Datasets with Large Number of Classes}
For showing the effectiveness of our method on datasets with large number of classes, we show results on long-tailed CIFAR-100 ($\rho$ = 10) and iNaturalist-2019~\citep{inat19} (1010 classes). The iNaturalist dataset consists of image of species which naturally have a long-tailed distribution. For the iNaturalist dataset we use a batch size of 256 which is significantly less than number of classes present (1010), and use $\beta = \alpha$ for updating the \textit{effective class distribution} in Eq.~\ref{cbgan_eq:update_equation}. We use a SNResGAN based architecture for both datasets.  Additional hyperparameters and training details are present in Appendix Section \ref{cbgan_app:largedatasets}. We use CIFAR-100 validation set and a balanced subset of (16160) iNaturalist images for calculation of FID with $50$K generated images in each case.

Table \ref{cbgan_tab:cifarinat} summarizes the results on the above two datasets. Our method clearly outperforms all the other baselines in terms of FID. For the CIFAR-100 dataset our method achieves balanced distribution similar to the cGAN. In the case of iNaturalist our method achieves KL DIV of $0.6$ which is significantly better than other baselines. Other baselines cause significant imbalance in the generated distribution and hence are unsuitable for real world long-tailed datasets such as iNaturalist (examples are shown in Figure \ref{cbgan_fig:inat_visual}. The superior results on iNaturalist demonstrate that our method is also effective  in the case when batch size is less than the number of classes present in the dataset. 

\begin{table}[t]
    \centering
    \resizebox{0.75\columnwidth}{!}{%
    \begin{tabular}{lcccc}
    \hline
    & FID ($\downarrow$)&      KLDiv($\downarrow$) & FID ($\downarrow$)&      KLDiv($\downarrow$) \\ \hline
    Imbalance Ratio & \multicolumn{2}{c}{100} & \multicolumn{2}{c}{10} \\ \hline
    \multicolumn{5}{c}{CIFAR-10} \\ \hline 
        \pbox[c][0.5 cm] [c] {10cm}{SNDCGAN} & 36.97 $\pm$ 0.20&	0.31 $\pm$ 0.0&  32.53 $\pm$ 0.06& 	0.14 $\pm$ 0.0 \\ \hline
        \rowcolor[gray]{0.85}
        \pbox[c][1. cm] [c] {10cm}{Ours\\ (Supervised)}  	&32.93 $\pm$ 0.11 &	0.06 $\pm$ 0.0& 30.48 $\pm$ 0.07&	0.01 $\pm$ 0.0\\ \hline
        \rowcolor[gray]{0.85}
        \pbox[c][1. cm] [c] {10cm}{Ours\\ (Semi Supervised)} & 33.32 $\pm$ 0.03 & 0.14 $\pm$ 0.0 &  30.37 $\pm$ 0.14 &  0.04 $\pm$ 0.0 \\ \hline
        \multicolumn{5}{c}{LSUN} \\ \hline 
        
        \pbox[c][0.5 cm] [c] {10cm}{SNResGAN} & 37.70 $\pm$ 0.10 &	0.68 $\pm$ 0.0 &   33.28 $\pm$ 0.02 & 	0.29 $\pm$ 0.0 \\ \hline
        \rowcolor[gray]{0.85}
        \pbox[c][1. cm] [c] {10cm}{Ours\\ (Supervised)} &  35.04 $\pm$ 0.19&	0.06 $\pm$ 0.0& 28.78 $\pm$ 0.01&	0.01 $\pm$ 0.0  \\ \hline
        \rowcolor[gray]{0.85}
        \pbox[c][1. cm] [c] {10cm}{Ours\\ (Semi Supervised)}  & 35.95 $\pm$ 0.05 & 0.15 $\pm$ 0.0 & 30.96 $\pm$ 0.07 & 0.06 $\pm$ 0.0 \\ \hline
    \end{tabular}}
    \caption{Comparison of results in Semi Supervised Setting. The pretrained classifier used in our framework is fine-tuned with 0.1\% of labelled data. The same classifier trained on balanced dataset is used as annotator for calculating KL Divergence for all baselines.}
    \label{cbgan_tab:semi-sup}
    
\end{table}
\subsection{Other baselines using Pre-trained Classifier}
Since the proposed generative framework utilizes a pretrained classifier, we believe the comparison (against other generative model) should be performed via providing the same classifier. Therefore, in this subsection, we choose ACGAN framework and build a baseline by adding the pretrained classifier. Note that in ACGAN discriminator not only performs the real-fake distinction but also serves as an auxiliary classifier and labels the sample into the underlying classes in the dataset. In this baseline, we replace the latter part with the pre-trained classifier used in the proposed framework. Hence both the frameworks are even with respect to the availability of the $P(Y/X)$ information. 

The resulting generator can avail the label information of the generated samples from this classifier. In other words, if the generator intends to produce a sample of class $y$ via conditioning, the pretrained classifier can provide the required feedback in the form of cross entropy loss. However, we find that this baseline of ACGAN that employs a pre-trained classifier suffers from mode collapse ($42.28$ FID) and only generates extremely limited within class diversity in images (e.g. in Fig. \ref{cbgan_fig:other_baseline_images}). On the contrary, our method ($9.01$ FID) using the same pre-trained classifier doesn't suffer from mode collapse and also works in case of iNaturalist dataset. This shows that the proposed framework and regularizer are non-trivial and prevent the GAN from mode collapse. We believe there is scope for understanding the nature of the involved optimization in this future. For the exact details of implementation please refer to the Appendix Section~\ref{cbgan_app:other_baseline}. An overview of the baseline is depicted in Figure \ref{cbgan_fig:approach_other_baseline}.
\subsection{Semi-supervised class-balancing GAN}
\label{cbgan_subsec:semi-supervised}
In case of conditional GANs, class labels are required for GAN training. However, in our case the stage of classifier learning which requires labels is decoupled from GAN learning. In this section we show how this can be advantageous in practice. Since our framework only requires knowledge of $P(Y/X)$, we find that a classifier trained through any of a variety of sources could be used for providing feedback to the Generator. This feedback allows the GAN to generate class balanced distributions even in cases when the labels for underlying long-tailed distributions are not known. This reduces the need for labelled data in our framework and shows the effectiveness over conditional GAN. Note that the performance of conditional GANs deteriorates~\citep{lucic2019high} when used with limited labelled data. We use a ResNet-50 pretrained model on ImageNet from BiT (Big Image Transfer)~\citep{kolesnikov2019large} and finetune it using $0.1$ \% of labelled data of balanced training set (i.e. $5$ images per class for CIFAR-10 and $50$ images per class for LSUN).

\begin{table}[t]
    \centering
    \resizebox{0.75\columnwidth}{!}{%

    \begin{tabular}{lcc|cc}
        \hline
         & \multicolumn{2}{c}{iNatuarlist 2019} & \multicolumn{2}{c}{CIFAR-100} \\\hline
         & FID ($\downarrow$) & KL Div($\downarrow$)          & FID ($\downarrow$) & KL Div($\downarrow$)  \\ \hline
         SNResGAN & 13.03 $\pm$ 0.07 & 1.33 $\pm$ 0.0 & 30.05 $\pm$ 0.05  & 0.18 $\pm$ 0.0\\  
         ACGAN & 47.15 $\pm$ 0.11 & 1.80 $\pm$ 0.0 & 69.90 $\pm$ 0.13 & 0.40 $\pm$ 0.0\\ 
         cGAN & 21.53 $\pm$ 0.14 & 1.47 $\pm$ 0.0 & 30.87 $\pm$ 0.06 &  0.09 $\pm$ 0.0 \\
         \rowcolor[gray]{0.85}
         Ours & 9.01 $\pm$ 0.08 & 0.60 $\pm$ 0.0 & 28.17 $\pm$ 0.06 & 0.11 $\pm$ 0.0 \\
         \hline

    \end{tabular}}
    \caption{Results on iNaturalist (2019) and CIFAR-100 ($\rho=10$) dataset. Significant performance increase is achieved by our method in comparison to baselines.}
    \label{cbgan_tab:cifarinat}
\end{table}
In Table \ref{cbgan_tab:semi-sup} we use the classifier trained with $0.1 \%$ of labelled data to train GAN on long-tailed version of CIFAR-10 and LSUN datasets ($\rho = 10,100$)  We observe that even with semi-supervised classifier, our method is able to produce an avg reduction of \textbf{0.26} in KL Divergence when compared to unsupervised GAN (SNResGAN) and also achieves better FID score in all cases. Note that this application is unique to our framework as conditional GANs explicitly require labels for whole dataset for training. %
\subsection{Analysis on the Effect of Classifier Performance}

\begin{figure}[!t]
    \centering
    \includegraphics[width=0.75\columnwidth]{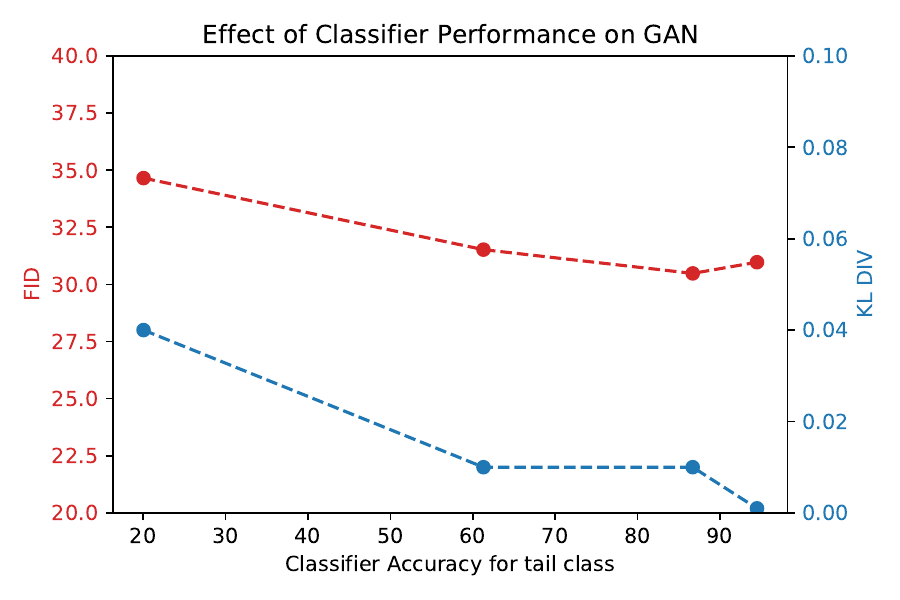}
    \caption{Analysis of Classifier Performance on long-tailed CIFAR-10 ($\rho$ = 10). The data points from left to right correspond to imbalance ratio's ($\rho$) 500, 100, 10 and 1 respectively. The performance is significantly robust after tail class accuracy reaches 60\%.}
    \label{cbgan_fig:clf_analysis}
\end{figure}

For analyzing the effect of classifier performance on the GAN training in our method, we train the GAN on long-tailed CIFAR-10 ($\rho = 10$) with different classifiers. We learn multiple classifiers with different imbalance ratios ($\rho$) of $1, 10, 100, 500$ and $5000$. As the imbalance ratio increases, the accuracy of the resulting classifier decreases, hence, we can have classifiers of varied accuracy for deploying in our framework. Note that in case of high imbalance ratio, it becomes harder for the classifier to learn the tail part of the distribution. Figure~\ref{cbgan_fig:clf_analysis} shows the performance of the resulting GAN (in terms of the FID and KLDIV measures) with respect  to the classifier performance on the tail class (i.e. least populated class). \\
From Figure~\ref{cbgan_fig:clf_analysis} it can be observed that for a large range of classifier accuracies on the tail class, GANs learned in our framework are able to achieve similar FID and KL Divergence performance. Our framework only requires reasonable classifier performance which can be easily achieved via normal cross entropy training on the long-tailed dataset. For further enforcing this claim, we only use a classifier that is trained using cross entropy loss for iNaturalist-2019 experiments, in which our method achieves state-of-the-art FID score. Hence one does not need to explicitly resort to any complex training regimes for the classifier. In the extreme case of $\rho = 5000$, the classifier performance is $0$ on the tail class. In this case, our algorithm diverges and is not able to produce meaningful results. 
However, note that with a classifier accuracy of as small as $20\%$ our framework achieves decent FID and KL Divergence.

\section{Conclusion}
\label{cbgan_section:conclusion}
Long-tailed distributions are a common occurrence in real-world datasets in the context of learning problems such as object recognition. However, a vast majority of the existing contributions consider simplified laboratory scenarios in which the data distributions are assumed to be uniform over classes. In this chapter, we consider learning a generative model (GAN) on long-tailed data distributions with an objective to faithfully represent the tail classes. We propose a class-balancing regularizer to balance class distribution of generated samples while training the GAN. We justify our claims on the proposed regularizer by presenting a theoretical bound and comprehensive experimental analysis. The key idea of our framework is having a classifier in the loop for keeping an uninterrupted check on the GAN's learning which enables it to retain the minority nodes of the underlying data distribution. We demonstrated that the dependency on such a classifier is not arduous. Our experimental analysis clearly brings out the effectiveness of our regularizer in the GAN framework for generating the images from complex long-tailed datasets such as iNaturalist, on which it achieves the state-of-the-art performance.

\chapter{Improving GANs for Long-Tailed Data through Group Spectral Regularization}
\label{chap:gsrgan}

\begin{changemargin}{7mm}{7mm} 
  Deep long-tailed learning aims to train useful deep networks on practical, real-world imbalanced distributions, wherein most labels of the tail classes are associated with a few samples. There has been a large body of work to train discriminative models for visual recognition on long-tailed distribution. In contrast, we aim to train conditional Generative Adversarial Networks, a class of image generation models on long-tailed distributions. We find that similar to recognition, state-of-the-art methods for image generation also suffer from performance degradation on tail classes. The performance degradation is mainly due to class-specific mode collapse for tail classes, which we observe to be correlated with the spectral explosion of the conditioning parameter matrix. We propose a novel group Spectral Regularizer (gSR) that prevents the spectral explosion alleviating mode collapse, which results in diverse and plausible image generation even for tail classes. We find that gSR effectively combines with existing augmentation and regularization techniques, leading to state-of-the-art image generation performance on long-tailed data. Extensive experiments demonstrate the efficacy of our regularizer on long-tailed datasets with \mbox{different} degrees of imbalance. Project Page: \href{https://sites.google.com/view/gsr-eccv22}{https://sites.google.com/view/gsr-eccv22}
\end{changemargin}

\section{Introduction}
\label{gsr_sec:intro}
Generative Adversarial Networks (GAN)~\cite{goodfellow2014generative} are consistently at the forefront of generative models for image distributions, also being used for diverse applications like image-to-image translation~\cite{mao2019mode}, super resolution~\cite{ledig2017photo} etc. One of the classic applications of GAN is class specific image generation, by conditioning on the class label $y$. The generated images in ideal case should associate to class label $y$, be of high quality and exhibit diversity. The conditioning is usually achieved with conditional Batch Normalization (cBN) ~\cite{de2017modulating} layers which induce class-specific ($y$) features at each layer of the generator. The class conditioning enables GAN models like the state-of-the-art (SOTA) BigGAN~\cite{brock2018large} to generate diverse images with high fidelity, in comparison to unconditional models~\cite{kang2020contrastive}.

Recent works~\cite{zhao2020differentiable} demonstrate
that performance of models like BigGAN deteriorates from mode collapse when limited training data is presented. The differentiable data augmentation \mbox{approaches}~\cite{zhao2020differentiable, karras2020training, tran2021data} attempt to mitigate this degradation by enriching the training data through augmentations. On the other hand, model based regularization techniques like LeCam~\cite{tseng2021regularizing} are also proposed to prevent the degradation of image quality in such cases.

\begin{figure*}[t]
    \centering
    
    \includegraphics[width=\textwidth]{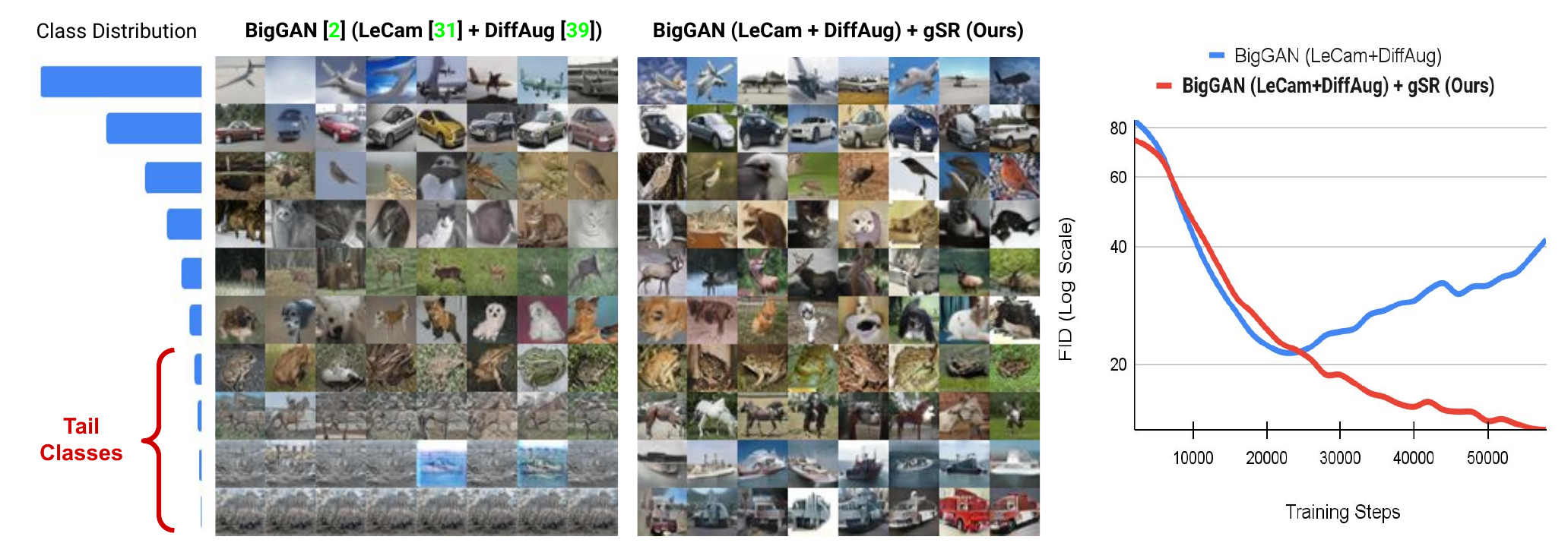}
    \caption{{Regularizing GANs on long-tailed training data.} \emph{(left)} Images generated from BigGAN trained on long-tailed CIFAR-10. \emph{(right)} FID scores vs. training steps. The proposed gSR regularizer prevents mode collapse, for the tail classes~\cite{brock2018large, tseng2021regularizing, zhao2020differentiable}. }\label{gsr_fig:overview}
\end{figure*}%
Although these methods lead to effective increase in image generation quality, they are designed to increase performance when trained on balanced datasets (\ie even distribution of samples across classes). We find that the state-of-the-art methods like BigGAN (LeCam) with augmentation, which are designed for limited data, also suffer from the phenomenon of class-specific mode collapse when trained on long-tailed datasets. By \textit{class-specific mode collapse}, we refer to deteriorating quality of generated images for tail classes, as shown in Fig.~\ref{gsr_fig:overview}.

In this work we aim to investigate the cause of the class-specific mode collapse that is observed in the generated images of tail classes. We find that the class-specific mode collapse is correlated with spectral explosion (\ie sudden increase in spectral norms, ref. Fig.~\ref{gsr_fig:sn_fid}) of the corrosponding class-specific (cBN) parameters (when grouped into a matrix, described in Sec.~\ref{gsr_sec:regularizer}). %
To prevent this spectral explosion, we introduce a novel class-specific \textit{group \mbox{Spectral Regularizer}} (gSR), which constrains the spectral norm of class-specific cBN parameters. Although there are multiple spectral~\cite{yoshida2017spectral, vahdat2020NVAE} regularization (and normalization~\cite{miyato2018spectral}) techniques used in deep learning literature, all of them are specific to model weights $W$, whereas our regularizer focuses on cBN parameters. We, through our analysis, show that our proposed gSR leads to reduced correlation among tail classes' cBN parameters, effectively mitigating the issue of class-specific mode collapse. 

We extensively test our regularizer's effectiveness by combining it with popular SNGAN~\cite{miyato2018spectral} and BigGAN~\cite{brock2018large} architectures. It also complements discriminator specific SOTA regularizer (LeCam+DiffAug~\cite{tseng2021regularizing}), as it's combination with gSR ensures improved quality of image generation even across tail classes (Fig.~\ref{gsr_fig:overview}). 
In summary, we make the following contributions:
\begin{itemize}
\itemsep0em 
    \item We first report the phenomenon of class-specific mode collapse, which is observed when cGANs are trained on long-tailed imbalance datasets. We find that spectral norm explosion of class-specific cBN parameters correlates with its mode collapse.
    \item We find that even existing techniques for limited data~\cite{tseng2021regularizing, liu2021generative, Karras2020ada} are unable to prevent class-specific collapse. Hence we propose a novel group Spectral Regularizer (gSR) which helps in alleviating such collapse.
    \item   Combining gSR with existing SOTA GANs with regularization leads to large average relative improvement (of $\sim25\%$ in FID) for image generation on 5 different long-tailed dataset configurations.
\end{itemize}

\section{Related Work}
\noindent\textbf{Generative Adversarial Networks:} 
Generative Adversarial Networks~\cite{goodfellow2014generative} are a combination of Generator $G$ and Discriminator $D$ aimed at learning a generative model for a distribution. GANs have enabled learning of models for complex distributions like high resolution images etc. One of the inflection point for success was the invention of Spectral Normalization (SN) for GANs (SNGAN) which enabled GANs to scale to datasets like ImageNet~\cite{deng2009imagenet} (1000 classes). The GAN training was further scaled by BigGAN \cite{brock2018large} which demonstrated successful high resolution image generation, using a deep ResNet network. 

\noindent \textbf{Regularization:} Several regularization techniques \cite{zhang2019consistency, zhao2021improved, mao2019mode, kavalerov2019cgans, liu2019spectral, zhou2021omni} are developed to alleviate the problem of mode collapse in GANs like Gradient Penalty~\cite{gulrajani2017improved}, Spectral Normalization~\cite{miyato2018spectral} etc. These include LeCam~\cite{tseng2021regularizing} and Differentiable Augmentations~\cite{zhao2020differentiable, Karras2020ada} which are the regularization techniques specifically designed to prevent mode collapse in limited data scenarios. Commonality among majority of these techniques are that they (a) designed for the data which is balanced across classes, and (b) focus on discriminator network $D$. In this work, we aim to regularize the cBN parameters in generator $G$, which makes our regularizer complementary to earlier works.

\noindent \textbf{Long-Tailed Imbalanced Data:} Long-tailed imbalance is a form of distribution in which majority of the data samples are present in head classes and the occurrence of per class data samples decreases exponentially as we move towards tail classes (Fig.~\ref{gsr_fig:overview}(\textit{left})). This family of distribution represents natural distribution for species' population~\cite{inat19}, objects~\cite{wang2017learning} etc. As these distributions are natural, a lot of work has been done to learn discriminative models (\ie classifiers)~\cite{cao2019learning, cui2019classbalancedloss, menon2021longtail, kang2019decoupling, yang2020rethinking, zhou2020BBN} which work across all classes, despite that training data follows a long-tailed distribution. However, even though there has been a lot of interest, there are still only a handful works which aim to learn generative models for long-tailed distribution. Mullick \etal~\cite{Mullick_2019_ICCV} develop GAMO which aims to learn how to oversample using a generative model. Class Balancing GAN (CBGAN) with a Classifier in the Loop~\cite{rangwani2021class} is the only work which aims to learn a GAN to generate good quality images across classes (in a long-tailed distribution). However their model is an unconditional GAN which requires a trained classifier to guide the GAN. The requirement of such a classifier can be restrictive. In this work we aim to develop conditional GANs which use class labels, does not require external classifier and generate good quality images (even for tail classes) when trained using long-tailed training data.

\section{Approach}
\begin{figure*}[t]    
    \centering
    \includegraphics[width=\textwidth, height=6.5cm]{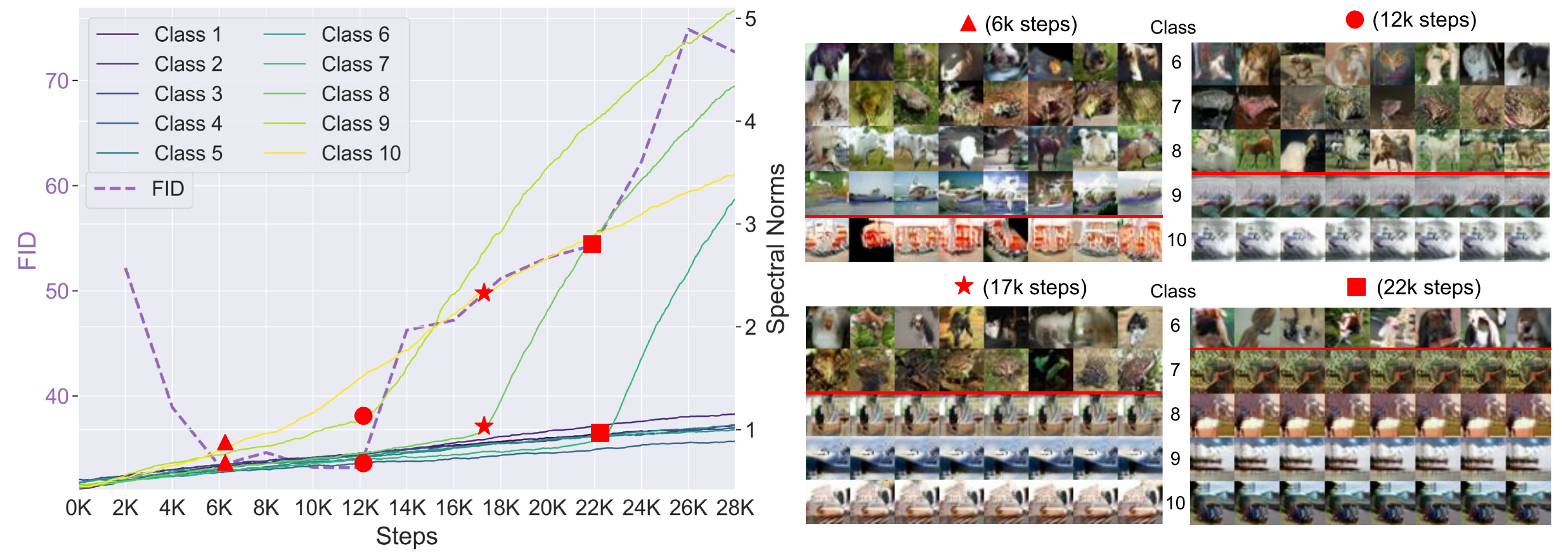}
    \caption{{Correlation between class-specific mode collapse and spectral explosion.} \emph{(left)} FID/Spectral Norms of class-specific gain parameter of conditional BatchNorm layer on CIFAR-10. \textcolor{red}{Symbols} on plot indicate that FID score's increase correlates with onset of spectral explosion on 4 \textcolor{brown}{tail classes} respectively. \emph{(right)} Images generated for \textcolor{brown}{tail classes} at these train steps reveals corresponding class-specific mode collapse. 
    \label{gsr_fig:sn_fid}
    }
\end{figure*}
We start by describing conditional Generative Adversarial Networks (Sec. \ref{gsr_subsec:GAN}) and the associated class-specific mode collapse (Sec. \ref{gsr_subsec:class-specific-collapse}). Following that we introduce our regularizer formulation, and explain the decorrelation among features caused by gSR that mitigates the mode collapse for tail classes (Sec. \ref{gsr_sec:regularizer}).
\subsection{Conditional Generative Adversarial Networks}
\label{gsr_subsec:GAN}
Generative Adversarial Networks (GAN) are a combination of two networks the generator $G$ and discriminator $D$. The discriminator's goal is to classify images from training data distribution ($P_{r}$) as real and the images generated through $G$ as fake. In a conditional GAN (cGAN), the generator and discriminator are also enriched with the information about the class label $y$ associated with the image $\mathbf{x} \in \mathbb{R}^{3\times H \times W}$. The conditioning information $y$ helps the cGAN in generating diverse images of superior quality, in comparison to unconditional GAN. The cGAN objectives can be described as:
\begin{equation}
\begin{split}
      &\max_D V(D) = \underset{x \sim P_{r}}{\mathbb{E}}[f_\mathcal{D} D(x|y)] + \underset{z \sim P_{z}}{\mathbb{E}}[f_{\mathcal{G}}(1 - D(G(z|y)))] \\
     & \min_G \mathcal{L}_{G} = \underset{z \sim P_{z}}{\mathbb{E}}[f_{\mathcal{G}}(1 - D(G(z|y)))] 
\end{split}
\end{equation}
where $p_z$ is the prior distribution of latents, $f_{\mathcal{D}}, f_{\mathcal{G}}$, and $g_\mathcal{{G}}$ refer to the mapping functions from which different formulations of GAN can be derived (ref.~\cite{liu2021generative}). The generator $G(z|y)$ generates images corresponding to the class label $y$. In earlier works the conditioning information $y$ was concatenated with the noise vector $z$, however conditioning each layer using cBN layer has shown improved performance~\cite{miyato2018cgans}. For a feature $\mathbf{x^l_y} \in \mathbb{R}^d$ conditioned on class $y$ (out of $K$ classes) corresponding to layer $l$, the transformation can be described as:
\begin{equation}
    \mathbf{\hat{x}_{y}^{l}} = \frac{\mathbf{{x}_{y}^{l}} - \mathbf{\mu_B^{l}}}{\sqrt{{\mathbf{\sigma^{l}_B}}^2  + \epsilon}} \rightarrow \mathbf{\gamma^l_{y}} \mathbf{\hat{x}_{y}^{l}} + \mathbf{\beta^l_{y}}
\end{equation}

here the $\mathbf{\mu_B^{l}}$ and ${\mathbf{\sigma^{l}_B}}^2$ are the mean and the variance of the batch respectively. The $\mathbf{\gamma^l_{y}} \in \mathbb{R}^d$ and the $\mathbf{\beta^l_{y}} \in \mathbb{R}^d$ are the cBN parameters which enable generation of the image for specific class $y$. We focus on the behaviour of these parameters in the subsequent sections.

\subsection{Class-Specific Mode Collapse}
\label{gsr_subsec:class-specific-collapse}
Due to widespread use of conditional GANs~\cite{brock2018large, miyato2018cgans}, it is important that these models are able to learn across various kinds of training data distributions.
However, while training a conditional GAN on long-tailed data distribution, we observe that 
GANs suffer from model collapse on tail classes (Fig.~\ref{gsr_fig:overview}). This leads to only a single pattern being generated for a particular tail class. To investigate the cause of this phenomenon, we inspect the class-specific parameters of cBN, which are gain $\mathbf{\gamma^l_{y}}$ and bias $\mathbf{\beta^l_{y}}$.
In existing works, characteristics of groups of features have been insightful for analysis of neural networks and have led to development of regularization techniques  \cite{wu2018group, huang2021group}.
Hence for further analysis we also create $n_g$ groups of the $\mathbf{\gamma^y_{l}}$ parameters and stack them to obtain $\mathbf{\Gamma^l_y} \in \mathbb{R}^{n_g \times n_c}$, where $n_c$ are the number of columns after grouping. It is observed that the value of spectral norm ($\sigma_{\max}(\mathbf{\Gamma_y^l}) \in \mathbb{R}$) explodes (\ie increases abnormally) as mode collapse occurs for corresponding tail class $y$  as shown in Fig.~\ref{gsr_fig:sn_fid}. We observe this phenomenon consistently across both the smaller SNGAN~\cite{miyato2018spectral} (Fig. \ref{gsr_fig:sn_fid}) and the larger BigGAN~\cite{brock2018large} (Fig.~\ref{gsr_fig:overview}) model. We observe similar spectral explosion for BigGAN model as in Fig. \ref{gsr_fig:sn_fid} (empirically shown in Fig. \ref{gsr_fig:fid_sn}). In the earlier works, mode collapse could be detected by anomalous behaviour of spectral norm of discriminator (refer to Appendix for details). However in the class-specific mode collapse the discriminator's spectral norms show normal behavior and are unable to signal such collapse. Here, our analysis of $\sigma_{\max}(\mathbf{\Gamma^l_y})$ helps in detecting class-specific mode collapse. \\

\subsection{Group Spectral Regularizer (gSR)}
\label{gsr_sec:regularizer}
\begin{figure}[t]
    \centering
    \includegraphics[width=\textwidth]{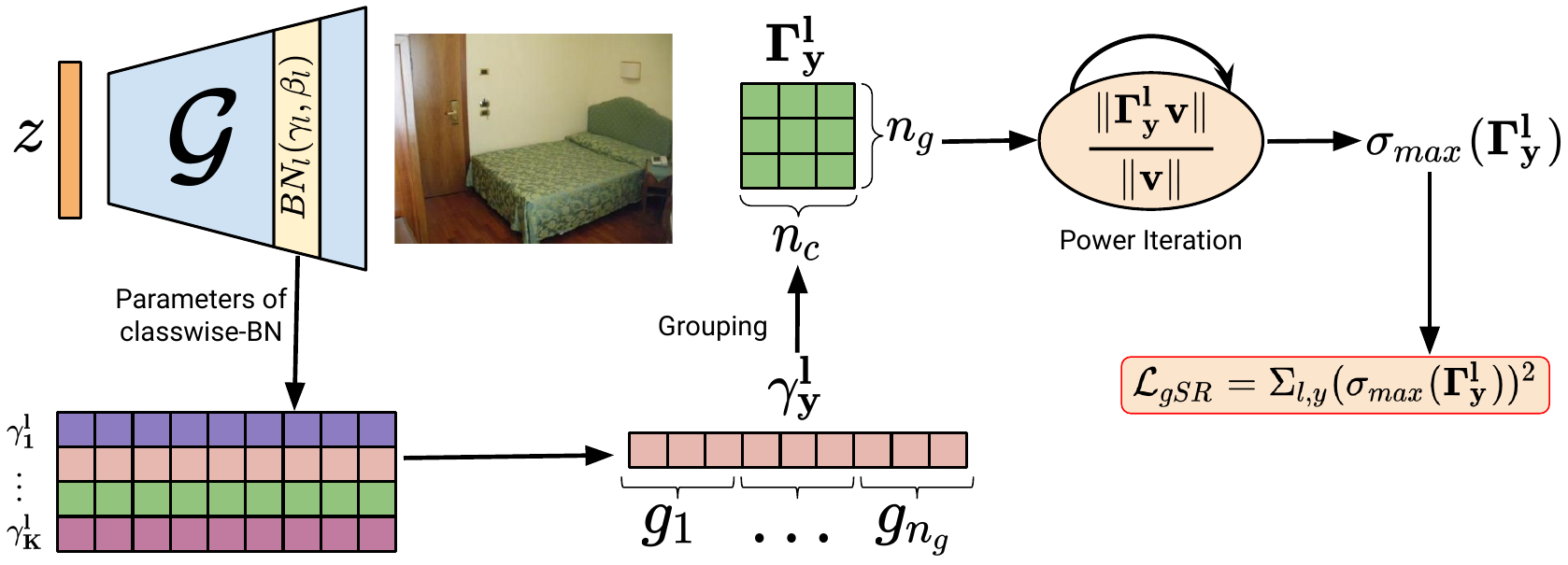}
    \caption{{Algorithmic overview.} During each training step, 1) we extract the gain $\mathbf{\gamma^l_y}$ for each cBN layers in $G$, 2) group them into matrix $\mathbf{\Gamma^l_y}$ and estimate $\sigma_{\max} (\Gamma^l_y)$. 3) We repeat the same procedure with bias $\mathbf{\beta^l_y}$ to obtain $\sigma_{\max} (\mathbf{B^l_y})$. 4) Finally, we regularize both as described in $\mathcal{L}_{gSR}$ (Eq.~\ref{gsr_eq:reg_loss}).}
    \label{gsr_fig:algo}
\end{figure}

Our aim now is to prevent the class-specific mode collapse while training. To achieve this, we introduce a regularizer for the generator $G$ which prevents spectral explosion. We would like to emphasize that earlier works including Augmentations~\cite{zhao2020differentiable}, LeCam~\cite{tseng2021regularizing} regularizer etc. are applied on discriminator, hence our regularizer's focus on G is complementary to those of existing techniques. As we observe that spectral norm explodes for the $\mathbf{\gamma^l_{y}}$ and $\mathbf{\beta^l_{y}}$ we deploy a group Spectral Regularizer (gSR) to prevent mode collapse. Steps followed by gSR for estimation of $\sigma_{\max}$ of $\mathbf{\gamma^l_{y}} (\in \mathbb{R}^d)$ are described below (also given in Fig. \ref{gsr_fig:algo}):
\begin{equation}
    \label{gsr_eq:grouping}
    Grouping: \; \mathbf{\Gamma^l_y} = \Pi(\mathbf{\gamma^l_y}, n_g) \in \mathbb{R}^{n_g \times n_c} 
\end{equation}
\begin{equation}
    \label{gsr_eq:power_iter}
    Power\ Iteration: \; \sigma_{\max}(\mathbf{\Gamma^l_y}) = \max_{\mathbf{v}} {\Vert\mathbf{\Gamma^l_yv}\Vert}/{\Vert \mathbf{v}\Vert}
\end{equation}

where $\mathbf{\mathbf{v}}(\in \mathbb{R}^d)$ is a random vector for power iterations. $n_g$ and $n_c$ are the number of groups and number of columns respectively, such that $n_g \times n_c = d$ . After estimation of $\sigma_{\max}(\mathbf{\Gamma^l_y})$ and similarly  $\sigma_{\max}(\mathbf{B^l_y})$, the regularized loss objective for generator can be written as:
\begin{equation}
\label{gsr_eq:reg_loss}
    \min_{G} \mathcal{L}_{    G} + \lambda_{gSR} \mathcal{L}_{gSR} ; \; \; \text{where} \; \;     \mathcal{L}_{gSR} = \sum_l \sum_{y} \lambda_y (\sigma^2_{\max}(\mathbf{\Gamma_{y}^l}) + \sigma^2_{\max}(\mathbf{B_{y}^l})).
\end{equation}

As the spectral explosion is prominent for the tail classes, we weigh the spectral regularizer term with $\lambda_y$ which has an inverse relation with number of samples $n_y$ in class $y$. As it has been shown in an earlier work~\cite{cao2019learning} that directly using $1/n_y$ can be over-aggressive. Hence, we use the effective number of samples (a soft version of the inverse relation) formally given as ~\cite{cui2019classbalancedloss} (where $\alpha = 0.99$): $\lambda_y = {(1 - \alpha)}/{(1 - \alpha^{n_y})}$.

The regularized objective is used to update weights using backpropagation. Spectral regularizers are used in earlier works~\cite{vahdat2020NVAE, yoshida2017spectral} but they are applied on network weights $W$, whereas to the best of our knowledge, ours is the first work that proposes the regularization of the batch normalization (BatchNorm) parameters. There exist other form of techniques like Spectral Normalization and Spectral Restricted Isometry Property (SRIP)~\cite{bansal2018can} regularizer, which we empirically did not find to be effective for our work (comparison in Sec.~\ref{gsr_sec:srsn}). \\

\noindent\textbf{Decorrelation Effect (Relation with Group Whitening):}
\begin{figure*}[t]    
    \centering
    \includegraphics[width=\textwidth]{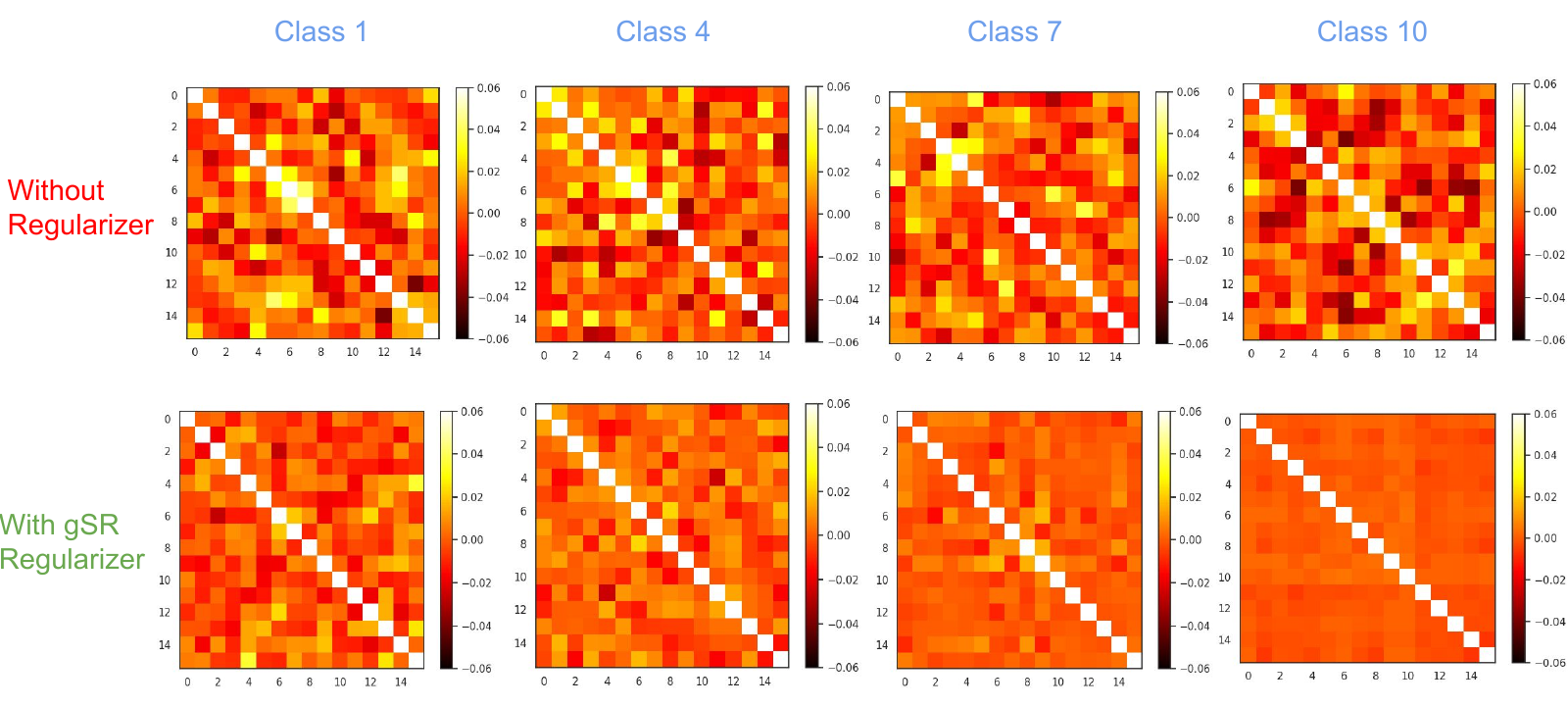}
    \caption{{Covariance matrices of $\mathbf{\Gamma_y^l}$ for (l = 1) for SNGAN baseline.}  After using gSR (for tail classes with high $\lambda$) the covariance matrix converges to a diagonal matrix in comparison to without gSR (where large correlations exist). This demonstrates the decorrelation effect of gSR on $\mathbf{\gamma_y^l}$, which alleviates class-specific mode collapse.}
    \label{gsr_fig:corr_plot}
\end{figure*}
\label{gsr_sec:modecollapse}
Group Whitening~\cite{huang2021group} is a recent work which whitens the activation map $X$ by grouping, normalizing and whitening using Zero-phase Component Analysis (ZCA) \nolinebreak{to obtain $\hat{X}$.} 
Due to whitening, the rows of $\hat{X_g}$ get decorrelated, which can be verified by finding the similarity of covariance matrix $\frac{1}{n_c}\hat{X_g}\hat{X_g}^{\intercal}$ to a diagonal matrix. The Group Whitening transformation significantly improves the generalization performance by learning diverse features. As our regularizer also operates on groups of features, we find that minimizing the $\mathcal{L}_{gSR}$ loss also leads to decorrelation of the rows of $\mathbf{\Gamma_y^l}$. We verify this phenomenon by visualizing the covariance matrix $\frac{1}{n_c}[\mathbf{\Gamma_y^l} - \mathbb{E}[\mathbf{\Gamma_y^l}]] ([\mathbf{\Gamma_y^l} - \mathbb{E}[\mathbf{\Gamma_y^l}]])^{\intercal}$.

In Fig.~\ref{gsr_fig:corr_plot}, we plot the covariance matrices for both the SNGAN and SNGAN with regularizer (gSR). We clearly observe that for tail classes with high $\lambda_y$ the covariance matrix is more closer to a diagonal matrix which confirms the decorrelation of parameters caused by gSR . 
We find that decorrelation is required more in layers with more class-specific information (\ie earlier layers of generator) rather than layers with generic features like edges. We provide the visualizations for more layers in the Appendix. 

Recent theoretical results~\cite{wang2020mma,jin2020does} for supervised learning show that decorrelation of parameters mitigates overfitting, and leads to better generalization. This is analogous to our observation of decorrelation being able to prevent mode collapse and helpful in generating diverse images.

\section{Experimental Evaluation}
We perform extensive experiments on various long-tailed datasets with different resolution. For the controlled imbalance setting, we perform experiments on CIFAR-10~\cite{Krizhevsky09learningmultiple} and LSUN~\cite{yu2015lsun}, which are commonly used in the literature~\cite{cao2019learning, cui2019classbalancedloss, santurkar2018classification} (Sec.~\ref{gsr_sec:synth_dist_results}). We also show results on challenging real-world datasets (with skewed data distribution) of iNaturalist2019~\cite{inat19} and AnimalFaces~\cite{kolouri2016sliced} (Sec.~\ref{gsr_sec:natural_dist_results}). 

\noindent\textbf{Datasets:} We use the CIFAR-10~\cite{Krizhevsky09learningmultiple} and a subset (5 classes) of LSUN dataset~\cite{yu2015lsun} (50k images balanced across classes) for our experiments. 
The choice of 5 class subset is for a direct comparison with related works~\cite{rangwani2021class, santurkar2018classification} which identify this subset as challenging and use that for experiments.
For converting the balanced subset to the long-tailed dataset with imbalance ratio ($\rho$) (\ie ratio of highest frequency class to the lowest frequency class), we remove the additional samples from the training set. Prior works~\cite{cao2019learning, cui2019classbalancedloss, menon2021longtail} follow this standard process to create benchmark long-tailed datasets. We keep the validation sets balanced and unchanged to evaluate the performance by treating all classes equally. We provide additional details about datasets in the Appendix. We perform experiments on the imbalance ratio of 100 and 1000. In case of CIFAR-10 for $\rho = 1000$ the majority class contains 5000 samples whereas the minority class has only 5 samples. For performing well in this setup, the GAN has to robustly learn from many-shot (5000 sample class) as well as at few-shot (5 sample class) together, making this benchmark challenging.

\setlength{\intextsep}{0pt}%
\begin{table*}[t]
    \centering
    \caption{{Quantitative results on the CIFAR-10 and LSUN dataset.} On an average, we observe a relative improvement in FID of 20.33\% and 39.08\% over SNGAN and BigGAN baselines respectively.}
    \label{gsr_tab:main_results}
    \resizebox{\textwidth}{!}{\begin{tabular}{lcccc|cccc}
    \toprule
    & \multicolumn{4}{c}{CIFAR-10} & \multicolumn{4}{c}{LSUN} \\ \hline
         Imb. Ratio ($\rho$) & \multicolumn{2}{c}{100} & \multicolumn{2}{c}{1000} & \multicolumn{2}{c}{100} &\multicolumn{2}{c}{1000}  \\ \hline
         & FID ($\downarrow$)&      IS($\uparrow$)& FID ($\downarrow$)&      IS($\uparrow$) & FID ($\downarrow$)&      IS($\uparrow$)& FID ($\downarrow$)&      IS($\uparrow$)  \\\midrule
         CBGAN~\cite{rangwani2021class} &33.01$_{\pm0.12}$ &6.58$_{\pm0.05}$ &44.82$_{\pm0.12}$ &5.92$_{\pm0.05}$ &37.41$_{\pm0.10}$ &2.82$_{\pm0.03}$ &44.70$_{\pm0.13}$ &2.77$_{\pm0.02}$\\
         LSGAN~\cite{mao2017least} &24.36$_{\pm0.01}$ &7.77$_{\pm0.07}$ &51.47$_{\pm0.21}$ &6.54$_{\pm0.05}$ &37.64$_{\pm0.05}$ &3.12$_{\pm0.01}$ &41.50$_{\pm0.04}$ &2.74$_{\pm0.02}$\\
         SNGAN~\cite{miyato2018spectral} &30.62$_{\pm0.07}$   &6.80$_{\pm0.02}$ &54.58$_{\pm0.19}$ &6.19$_{\pm0.01}$ &38.17$_{\pm0.02}$ &3.02$_{\pm0.01}$ &38.36$_{\pm0.11}$ &2.99$_{\pm0.01}$\\
         \rowcolor{gray!10} \; + gSR (Ours) &18.58$_{\pm0.10}$ &7.80$_{\pm0.09}$ &48.69$_{\pm0.04}$ &5.92$_{\pm0.01}$ &28.84$_{\pm0.09}$ &3.50$_{\pm0.01}$ &35.76$_{\pm0.05}$ &3.56$_{\pm0.01}$\\ \midrule
         BigGAN~\cite{brock2018large} &19.55$_{\pm0.12}$ &8.80$_{\pm0.09}$ &50.78$_{\pm0.23}$ &6.50$_{\pm0.05}$ &38.65$_{\pm0.05}$ &\textbf{4.02$_{\pm0.01}$} &45.89$_{\pm0.30}$ &3.25$_{\pm0.01}$\\
         \rowcolor{gray!10} \; + gSR (Ours) &\textbf{12.03$_{\pm0.08}$} &\textbf{9.21$_{\pm0.07}$} &\textbf{38.38$_{\pm0.01}$} &\textbf{7.24$_{\pm0.04}$} &\textbf{20.18$_{\pm0.07}$} &3.67$_{\pm0.01}$ &\textbf{24.93$_{\pm0.09}$} &\textbf{3.68$_{\pm0.01}$} \\ \bottomrule
    \end{tabular}}
\end{table*}

\noindent\textbf{Evaluation: } We report the standard Inception Score (IS)~\cite{salimans2016improved} and Fr\'echet Inception Distance (FID) metrics for the generated datasets. We report the mean and standard deviation of 3 evaluation runs similar to the protocol followed by DiffAug~\cite{zhao2020differentiable} and LeCam~\cite{tseng2021regularizing}. We use a held out set of 10k images for calculation of FID for both the datasets. The held out sets are balanced across classes for fair evaluation of each class. 

\noindent\textbf{Configuration:} We perform experiments by modifying the PyTorch-StudioGAN implemented  by Kang \etal~\cite{kang2020contrastive}, which serves as the baseline for our framework. We generate 32 $\times$ 32 sized images for the CIFAR-10 dataset and 64 $\times$ 64 sized images for the LSUN dataset. For the CIFAR-10 experiments, we use 5 $D$ steps for each $G$ step. Unless explicitly stated, we by default add the following two SOTA regularizers to obtain strong generic baselines for all the experiments (except CBGAN for which we follow exact setup described by Rangwani \etal~\cite{rangwani2021class}):
\begin{itemize}
\itemsep0em
    \item {DiffAugment~\cite{zhao2020differentiable}:} We apply the differential augmentation technique with the colorization, translation, and cutout augmentations.
    \item {LeCam~\cite{tseng2021regularizing}}: LeCam regularizer prevents divergence of discriminator $D$ by constraining its output through a regularizer term $R_{LC}$ (ref. Appendix).
\end{itemize}
Any improvement over these strong regularizers published recently is meaningful and shows the effectiveness of the proposed methods. We use a batch size of 64 for all our CIFAR-10 and LSUN experiments. For sanity check of the implementation we run the experiments for the balanced dataset (CIFAR-10) case where our FID is similar to the one obtained in LeCam~\cite{tseng2021regularizing}, details are in the Appendix.

\noindent\textbf{Baselines:} We compare our regularizer with the recent work of Class Balancing GAN (CBGAN) ~\cite{rangwani2021class} which uses an auxiliary classifier for long-tailed image generation. We use the public codebase\footnote{https://github.com/val-iisc/class-balancing-gan} and configuration provided by the authors.
The auxiliary classifiers are obtained using the LDAM-DRW as suggested by CBGAN authors. We use the SNGAN~\cite{miyato2018spectral} (with projection discriminator~\cite{miyato2018cgans}) as our base method on which we apply the Augmentation and LeCam regularizer for a strong baseline. We also compare our method with LSGAN~\cite{mao2017least}, which is shown to be effective in preventing the mode-collapse (we use the same configuration as in SNGAN for fairness of comparison). To demonstrate improvement over large scale GANs we also use BigGAN \cite{brock2018large} with LeCam and DiffAug regularizers as baseline.  We then add our group Spectral Regularizer (gSR) in the loss terms for BigGAN and SNGAN, and report the performance in \mbox{Table~\ref{gsr_tab:main_results}.} We don't use ACGAN as our baseline as it leads to image generation which doesn't match the conditioned class label (\ie class confusion)~\cite{rangwani2021class}.

\subsection{Results on CIFAR-10 and LSUN}
\label{gsr_sec:synth_dist_results}

\begin{figure*}[t]
  \centering
\begin{minipage}[c]{0.34\linewidth}
    \centering
    \includegraphics[width=\textwidth]{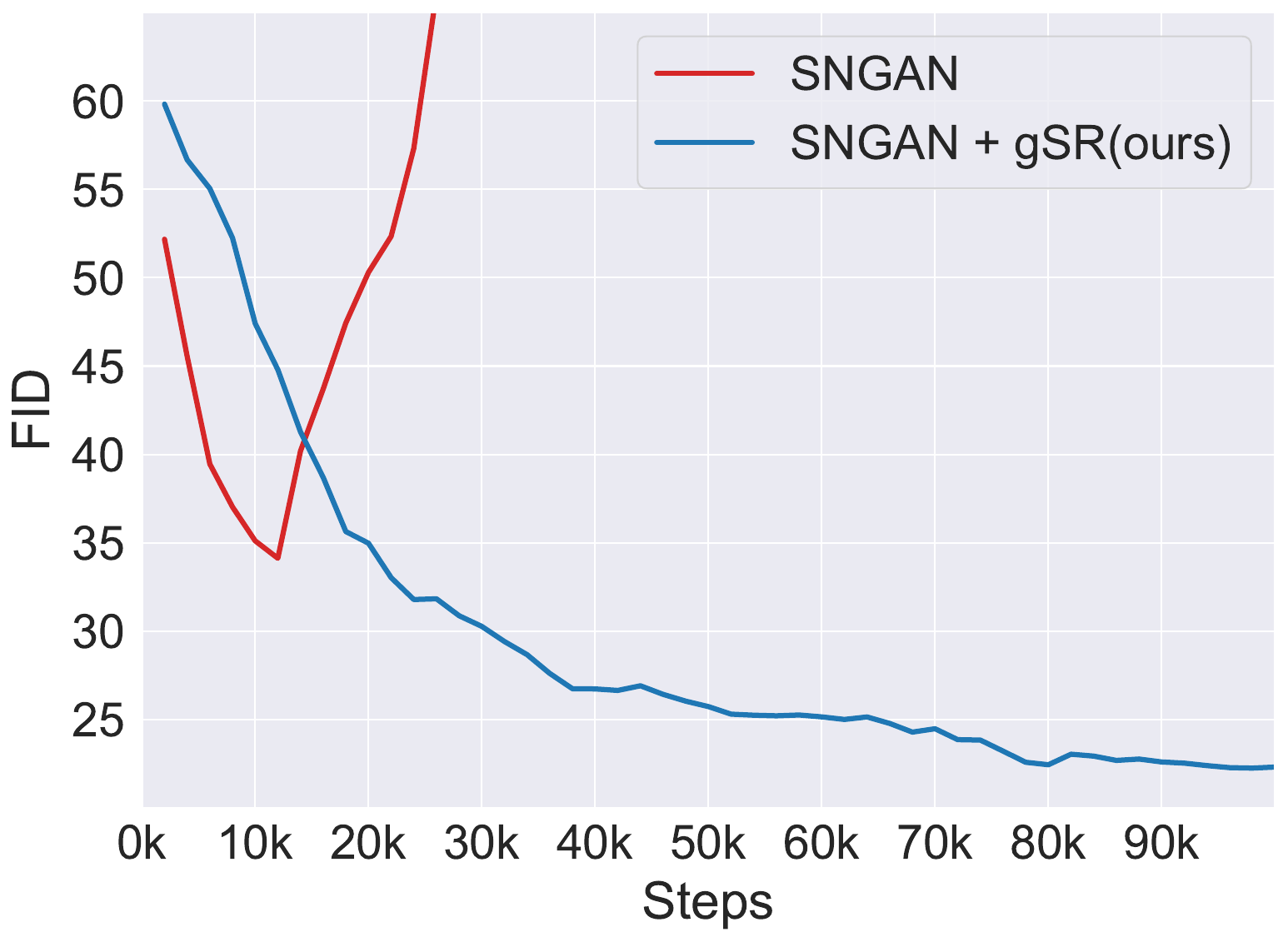}
    \caption{{Stability}. Addition of gSR (to baseline) stabilizes the training to continually improve, as indicated by the FID scores.}
    \label{gsr_fig:fid_gsr}
  \end{minipage}
   \hfill
  \begin{minipage}[c]{0.64\textwidth}
    \centering
    \includegraphics[width=\textwidth]{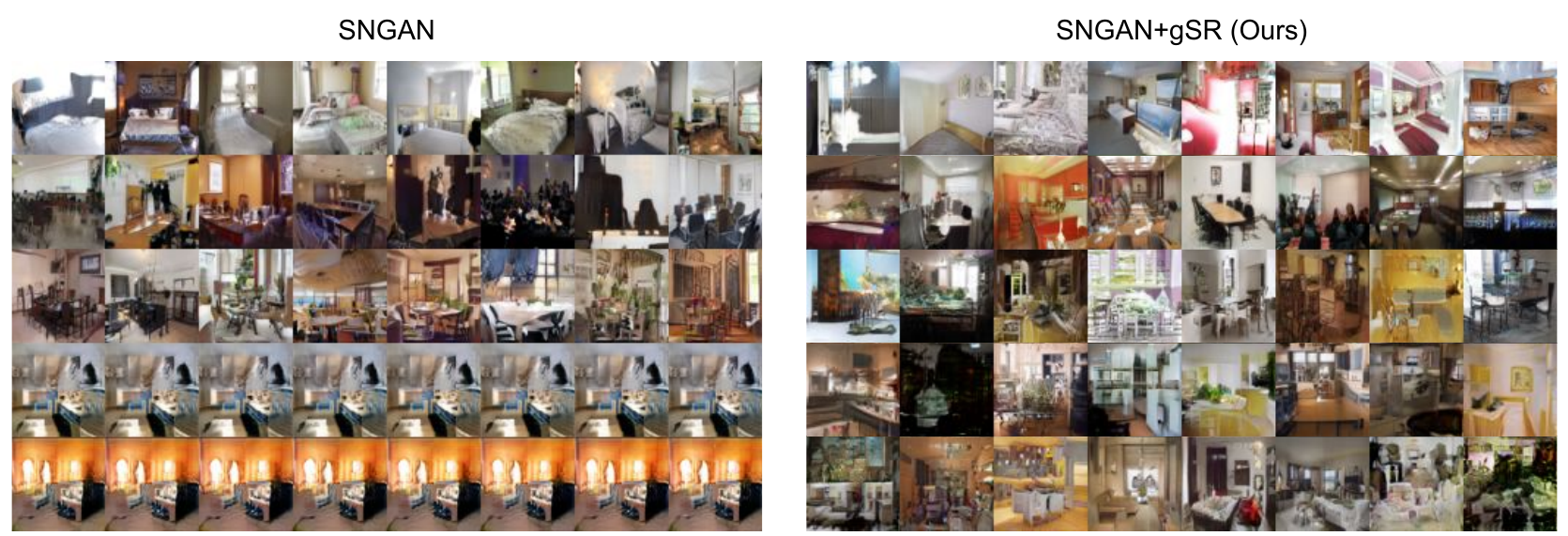}
    \begin{minipage}{1cm}
    \vfill
    \end{minipage}
    \caption{{Qualitative evaluations of SNGAN baseline on LSUN dataset.} Each row represents images from a class. Note the class-specific mode collapse observed in tail-classes in SNGAN (last two rows), which is alleviated after addition of gSR to generate diverse images.}
    \label{gsr_fig:sngan_imgs}
  \end{minipage}
\end{figure*}

\noindent\textbf{Stability:} Fig. \ref{gsr_fig:fid_gsr} shows the FID vs iteration steps for the SNGAN and 
SNGAN +gSR configuration. Using gSR regularizer with SNGAN is able to effectively prevent the class-specific mode collapse, which in turn helps the GAN to improve for a long duration of time. SNGAN without gSR starts degrading quickly and stops improving very soon, this shows the stabilizing effect imparted by gSR regularizer in the training process. The stabilizing effect is similarly observed even for the BigGAN (ref. FID plot in Fig.~\ref{gsr_fig:overview}).

\noindent\textbf{Comparison of Quality}: We observe that application of regularizer
is effectively avoids mode collapse and leads to a large average improvement (of 7.46) in FID for the (SNGAN + gSR) case, in comparison to SNGAN baseline across the four datasets (Table \ref{gsr_tab:main_results}). Our method is also effective on BigGAN where it is able to achieve SOTA FID and IS significant improvement in almost all cases. Although SNGAN and BigGAN baselines are already enriched with SOTA regularizers of (LeCam + DiffAug) to improve results, yet the addition of our gSR regularizer significantly boosts performance by harmonizing with other regularizers. It also shows that our regularizer complements the existing work and effectively reduces mode collapse. Fig.~\ref{gsr_fig:sngan_imgs} shows a comparison of the generated images for the different methods, where gSR is able to produce better images for the tail classes for LSUN dataset (refer Fig.~\ref{gsr_fig:overview} for qualitative comparison on CIFAR-10 ($\rho$ = 100)). To quantify improvement over each class, we compute class-wise FID and mean FID (\ie Intra FID) in Fig. \ref{gsr_fig:intra-fid}. We find that gSR leads to very significant improvement in tail class FID as it prevents the collapse. Due to the stabilizing effect of gSR we find that head class FID are also improved, clearly demonstrating the benefit of gSR for all classes. We also provide additional metrics (precision~\cite{kynkaanniemi2019improved}, recall~\cite{kynkaanniemi2019improved}, density~\cite{ferjad2020icml}, coverage~\cite{ferjad2020icml} and Intra-FID) in Appendix. We find that almost all metrics show similar improvement as seen in FID (Table \ref{gsr_tab:main_results}).

\subsection{Results on Naturally Occurring Distributions}
\label{gsr_sec:natural_dist_results}
\begin{figure*}[t]
  \centering
\begin{minipage}[c]{0.34\linewidth}
    \centering
    \includegraphics[width=\textwidth]{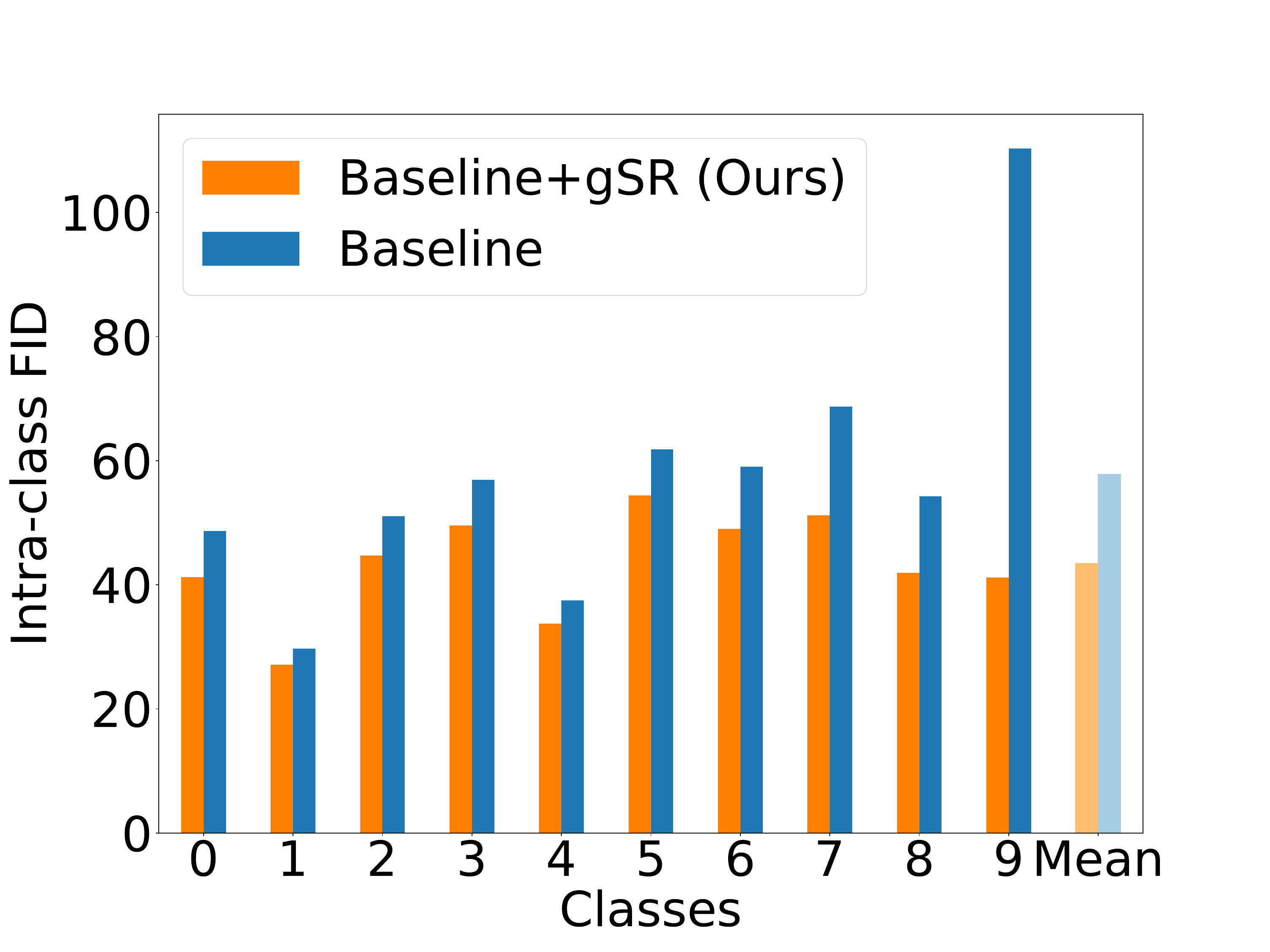}
    \captionof{figure}{Class-Wise FID and mean FID (Intra-FID) of BigGAN on CIFAR-10 over 5$K$ generated images($\rho$ = 100).}
    \label{gsr_fig:intra-fid}
  \end{minipage}
   \hfill
  \begin{minipage}[c]{0.60\textwidth}
    \centering
    \captionof{table}{{Quantitative results on iNaturalist-2019 and AnimalFace Dataset.} We compare mean FID ($\downarrow$) with other existing \mbox{approaches.}}
    \label{gsr_tab:iNaturalist}
    \begin{tabular}{l|c|c|c}
    \toprule
                & \multicolumn{2}{c}{iNaturalist 2019} & AnimalFace \\ \hline
         Method & cGAN & FID($\downarrow$) & FID($\downarrow$)\\ \midrule
         SNResGAN$^*$~\cite{miyato2018spectral}& \xmark & 13.03$_{\pm0.07}$ & -\\
         CBGAN$^*$~\cite{rangwani2021class} & \xmark & 9.01$_{\pm0.08}$ & - \\ 
         ACGAN$^*$~\cite{odena2017conditional} & \cmark & 47.15$_{\pm0.11}$ & - \\
         SNGAN$^*$~\cite{miyato2018cgans} & \cmark & 
         21.53$_{\pm0.03}$ & -\\  \midrule
         
         BigGAN~\cite{brock2018large} & \cmark & 6.87$_{\pm0.04}$ & 38.41$_{\pm0.04}$\\
        \rowcolor{gray!10} \; + gSR (Ours) & \cmark&\textbf{6.36$_{\pm0.04}$} & \textbf{35.76$_{\pm0.04}$} \\ \bottomrule
    \end{tabular}
  \end{minipage}
\end{figure*}

To show the effectiveness of our regularizer on natural distributions we perform experiments on two challenging datasets: iNaturalist-2019~\cite{inat19} and AnimalFace~\cite{si2011learning}. %
The iNaturalist dataset is a real-world long-tailed dataset with 1010 
classes of different species. There is high diversity among the images of each class, due to their distinct sources of origin. The dataset follows a long-tailed distribution with around 260k images. The second dataset we experiment with is the Animal Face Dataset~\cite{kolouri2016sliced} which contains 20 classes with with around 110 samples per class. We generate 64 $\times$ 64 images for both datasets using the BigGAN with a batch size of 256 for iNaturalist and 64 for AnimalFaces. The BigGAN baseline is augmented with LeCam and DiffAug regularizers. We compare our method with the baselines described in~\cite{rangwani2021class}. We evaluate each model using the FID on a training subset which is balanced across classes. For baselines we directly report results from Rangwani \etal \cite{rangwani2021class} (indicated by $^*$) in 
Table~\ref{gsr_tab:iNaturalist}.

The BigGAN baseline achieves an FID of 6.87 on iNaturalist 2019 dataset, which improves relatively by 7.42\% (-0.51 FID) when proposed gSR is combined with BigGAN. Our approach also is able to achieve FID better than SOTA CBGAN on iNaturalist 2019 dataset.
\mbox{Table~\ref{gsr_tab:iNaturalist}} shows the performance of the BigGAN baseline over the AnimalFace dataset, where after combining with our gSR regularizer we see FID improvement by 6.90\% (-2.65 FID). The improvements on both the large long-tailed dataset and few shot dataset of AnimalFace shows that gSR is able to effectively improve performance on real-world datasets. We provide additional experimental details and results in the Appendix.
\setlength{\intextsep}{0pt}%

\section{Analysis}
\subsection{Ablation over Regularizers}
We use the combination of existing regularizers (LeCam + DiffAug) with our regularizer (gSR) to obtain the best performing models. For further analysis of importance of each, we study their effect in comparison to gSR in this section. We perform experiments by independently applying each of them on vanilla SNGAN. Table~\ref{gsr_tab:sngan_abl} shows that existing regularizer in itself is not able to effectively reduce FID, whereas gSR is effectively able to reduce FID independently by 3.8 points. However we find that existing regularizers along with proposed gSR, make an effective combination which further reduces FID significantly (by 9.27) on long-tailed CIFAR-10 ($\rho = 100$). This clearly shows that our regularizer effectively complements the existing regularizers.
\subsection{High Resolution Image Generation} 
\begin{figure*}[t]    
    \centering
    \includegraphics[width=\textwidth]{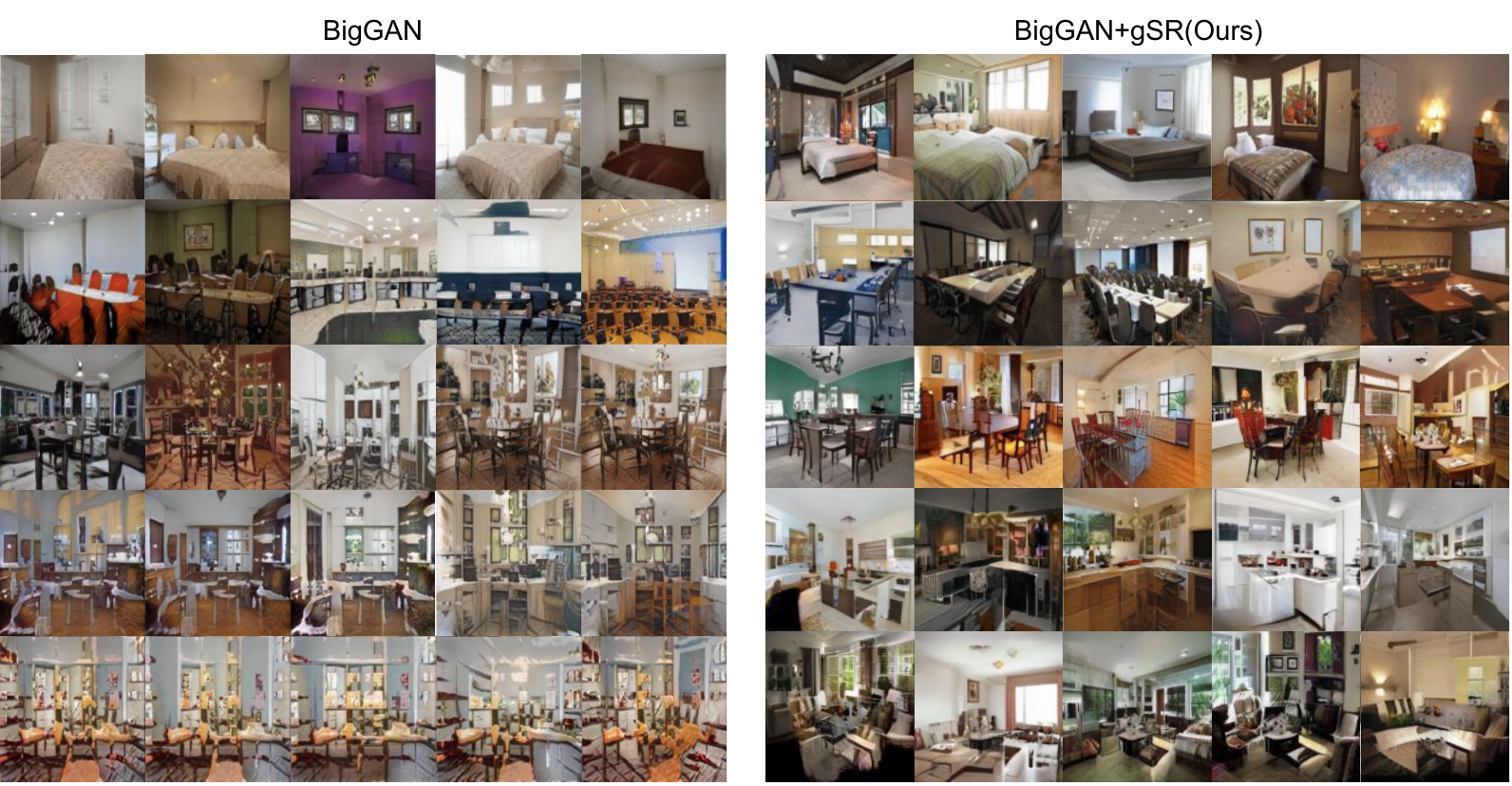}
    \caption{{Qualitative comparison of BigGAN variants on LSUN dataset ($\rho$=100) (128 $\times$ 128).} Each row represents images from a distinct class. }
    \label{gsr_fig:lsun_128}
\end{figure*}

As the LSUN dataset is composed of high resolution scenes we also investigate if the class-specific mode collapse phenomneon when GANs are trained for high resolution image synthesis. For this we train SNGAN and BigGAN baselines for (128 $\times$ 128) using the DiffAugment and LeCAM regularizer (details in Appendix). We find that similar phenomenon of spectral explosion leading to class-specific collapse occurs (as in 64 $\times$ 64 case), which is mitigated when the proposed gSR regularizer is combined with the baselines (Fig. \ref{gsr_fig:lsun_128}). The gSR regularizer leads to significant improvement in FID (Table \ref{gsr_tab:high_res}) also seen in qualitatively in Fig. \ref{gsr_fig:lsun_128}.

\begin{table*}[t]
\parbox{.48\linewidth}{
\centering
    \caption{{Ablation over regularizers on SNGAN.} We report FID and IS on  CIFAR-10 dataset (with $\rho = 100$).}
    \label{gsr_tab:sngan_abl}
    \begin{tabular}{c|c|c|c}
    \toprule
          \begin{tabular}[c]{@{}l@{}}LeCam+\\DiffAug\end{tabular}  & gSR  & FID($\downarrow$) & IS($\uparrow$)\\ \midrule
            \xmark         &      \xmark     &31.73$_{\pm0.08}$                 &7.18$_{\pm0.02}$      \\  
                        \rowcolor{gray!10}  \xmark         &      \cmark     &27.85$_{\pm0.05}$                  &7.09$_{\pm0.02}$    \\
                        \cmark         &      \xmark     &30.62$_{\pm0.07}$                  &6.80$_{\pm0.02}$      \\   
          \rowcolor{gray!10} \cmark        &      \cmark     &\textbf{18.58$_{\pm0.10}$} &\textbf{7.80$_{\pm0.09}$}                      \\ \bottomrule 
    \end{tabular}
}
\hfill
\parbox{.48\linewidth}{
    \centering
    \caption{{Image Generation (128 $\times$ 128)}. We report FID on both SNGAN and BigGAN on LSUN dataset (for $\rho$ = 100 and $\rho$ = 1000).}
    \label{gsr_tab:high_res}
    {\begin{tabular}{lcc}
    \toprule
         Imb. Ratio ($\rho$)  & \multicolumn{1}{c}{100} &\multicolumn{1}{c}{1000}  \\ \hline
         &  FID ($\downarrow$)& FID ($\downarrow$) \\\midrule
         
         SNGAN~\cite{miyato2018spectral} &53.91$_{\pm0.02}$ &72.37$_{\pm0.08}$\\
         \rowcolor{gray!10} \; + gSR (Ours)  &\textbf{25.31$_{\pm0.03}$} &\textbf{31.86$_{\pm0.03}$}\\ \midrule
         BigGAN~\cite{brock2018large}   &61.63$_{\pm0.11}$ &77.17$_{\pm0.18}$\\
         \rowcolor{gray!10} \; + gSR (Ours)   &\textbf{16.56$_{\pm0.02}$} &\textbf{45.08$_{\pm0.10}$} \\ \bottomrule
    \end{tabular}%
}}
\end{table*}

\subsection{Comparison with related Spectral Regularization and Normalization Techniques}
\label{gsr_sec:srsn}
\begin{figure*}[t]
  \centering
  \begin{minipage}[t]{0.45\linewidth}
    \centering
    \includegraphics[width=1\textwidth, height=2.5cm]{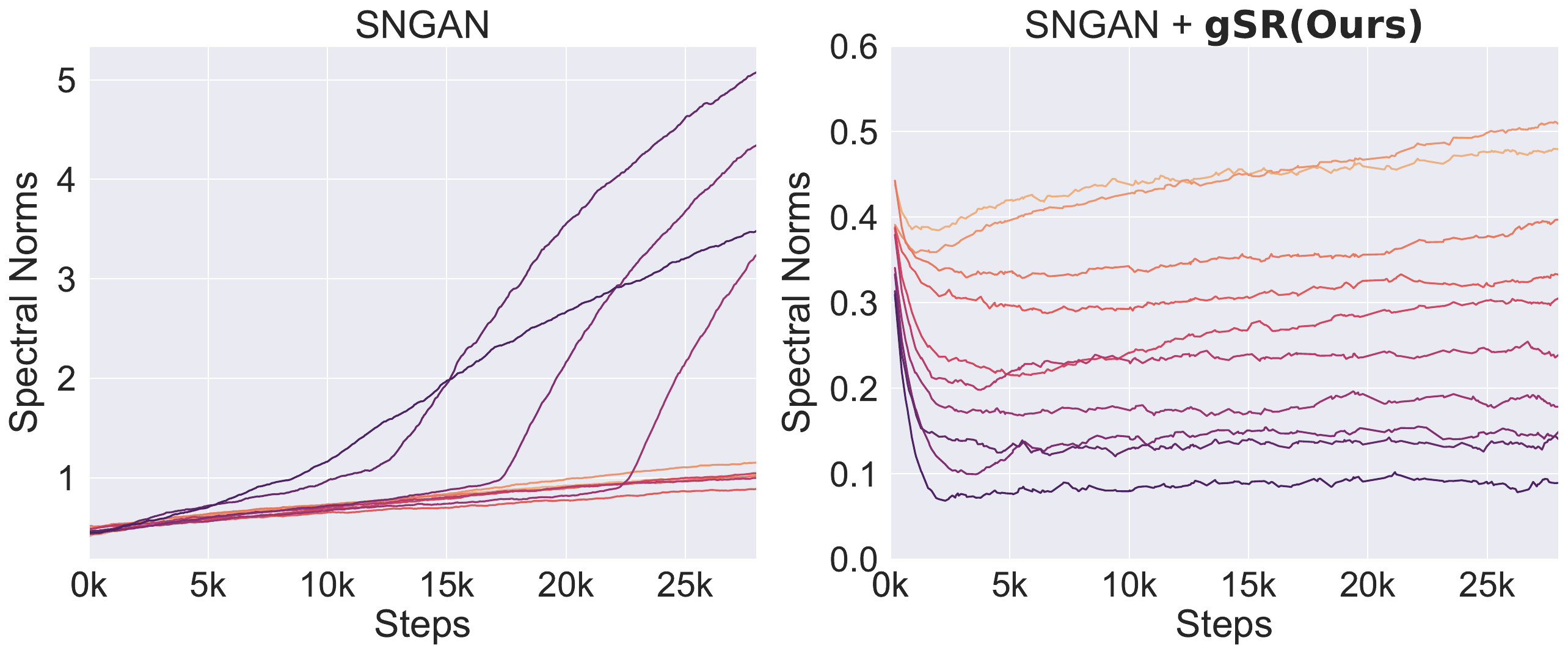}
  \end{minipage}
  \begin{minipage}[t]{0.54\linewidth}
    \centering
    \includegraphics[width=1\textwidth,height=2.5cm]{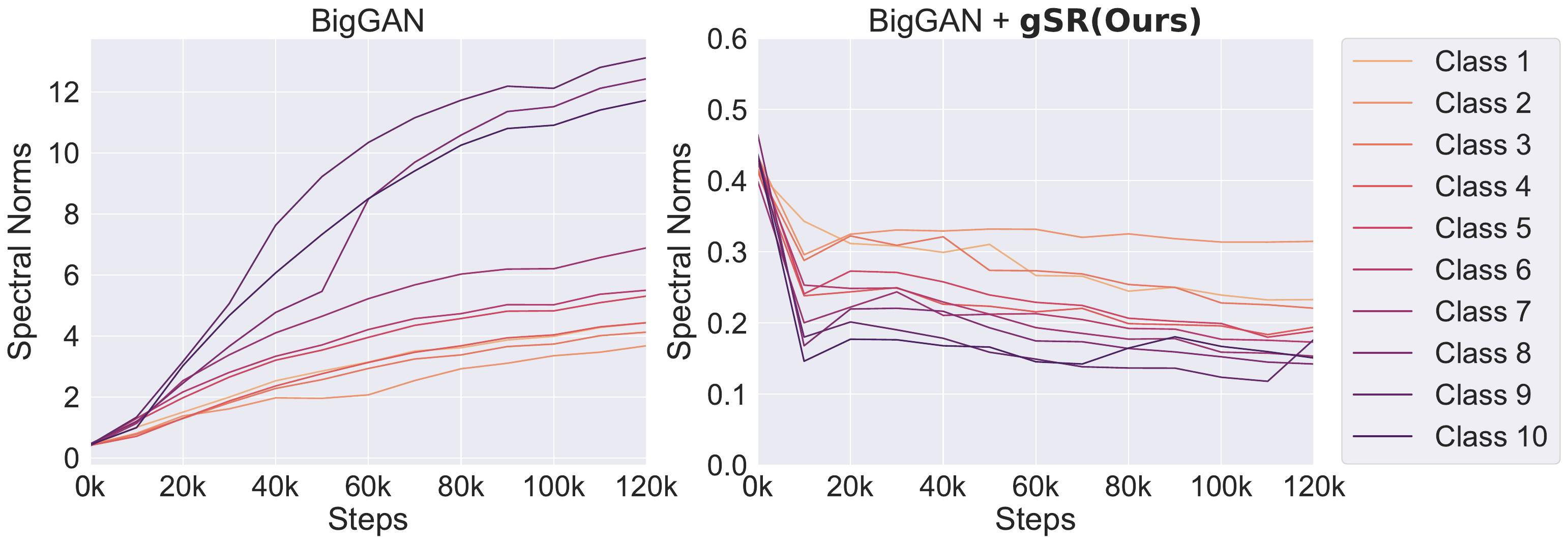}
  \end{minipage}
    \caption{{Effect of gSR on spectral norms of $\mathbf{\Gamma^l_y}$ (CIFAR-10).} We observe a spectral explosion both for SNGAN(\textit{left}) and BigGAN(\textit{right}) baselines of tail classes' cBN parameters. This is prevented by addition of gSR as shown on corresponding right.}
  \label{gsr_fig:fid_sn}
\end{figure*}

As gSR constrains the exploding spectral norms for the cBN parameters (Fig.~\ref{gsr_fig:fid_sn})
to evaluate its effectiveness, we test it against other variants of spectral normalization and regularization techniques on SGAN for CIFAR-10 ($\rho=100$).

\noindent\textbf{Group Spectral Normalization (gSN) of BatchNorm Parameters:} In this setting, rather than using sum of spectral norms (Eq. \ref{gsr_eq:reg_loss}) as regularizer for the class-specific parameters of cBN in gSR, we normalize them by dividing it by group spectral norms (\ie $\frac{\mathbf{\gamma^l_y}}{\sigma_{max}(\mathbf{\Gamma^l_y})}$) \cite{miyato2018spectral}.

\noindent\textbf{Group Spectral Restricted Isometry Property (gSRIP) Regularization:}  Extending SRIP~\cite{bansal2018can}, the class-specific parameters of cBN which are grouped to form a matrix $\mathbf{\Gamma^l_y}$, the regularizer is the sum of square of spectral norms of $(\mathbf{\Gamma^l_y}^\intercal\mathbf{\Gamma^l_y} - \mathbf{I)}$, (instead that of $\mathbf{\Gamma^l_y}$ in gSR (Eq. \ref{gsr_eq:reg_loss})). We report our findings in Table~\ref{gsr_tab:compare_reg_norm}. It can be inferred that all three techniques, namely gSN, gSRIP, and gSR, lead to significant improvements over the baseline. This also confirms our hypothesis that reducing (or constraining) spectral norm of cBN parameters alleviates class-specific mode collapse. However, it is noteworthy that gSR gives the highest boost over the baseline, surpassing gSN and gSRIP by a considerable margin in terms of FID.

\subsection{Analysis of gSR}
In this section we provide ablations of gSR using long-tailed CIFAR-10 ($\rho$=100).  

 \noindent \textbf{Can gSR work with StyleGAN-2?} We train and analyze the StyleGAN2-ADA implementation available~\cite{kang2020contrastive} on long-tailed datasets, where we find it also suffers from class-specific mode collapse.
 
  \begin{figure}
    \centering
    \includegraphics[width=0.5\textwidth]{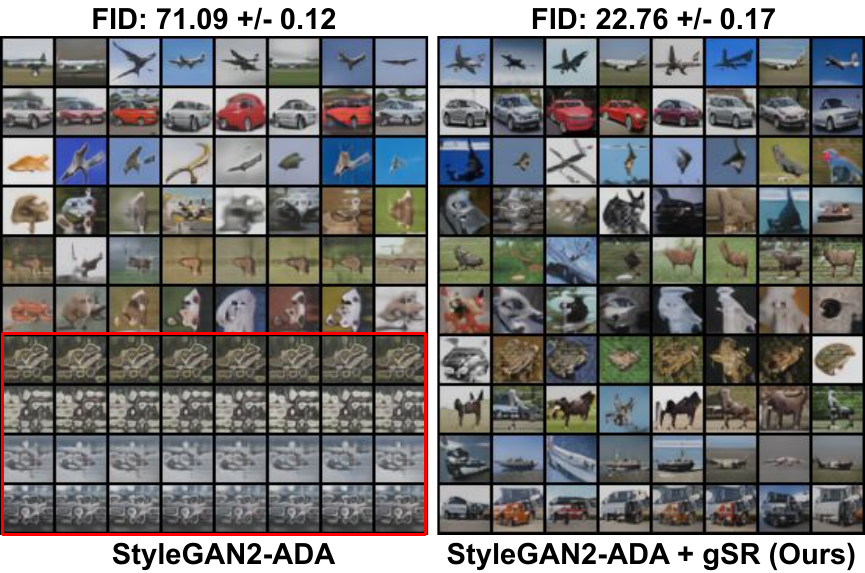}
    \caption{{StyleGAN2-ADA}  On CIFAR-10 ($\rho = 100$), comparison of gSR with the baseline.}
    \label{gsr_fig:qual_results}
\end{figure} We then implement gSR for StyleGAN2 by grouping 512 dimensional class conditional embeddings in mapping network to 16x32 and calculating their spectral norm which is added to loss (Eq. \ref{gsr_eq:reg_loss}) as $R_{gSR}$.We find that gSR is able to effectively prevent the mode collapse (Fig.\ \ref{gsr_fig:qual_results}) and also results in significant improvement in FID in comparison to StyleGAN2-ADA baseline.
 
\noindent\textbf{What is gSR's effect on spectral norms?}  We plot spectral norms of class-specific gain parameter of 1$^{st}$ layer of generator in SNGAN. Spectral norms explode for the tail classes without gSR, while they remain stable when gSR is used. Fig.~\ref{gsr_fig:fid_gsr} for same experiment shows that while using gSR the FID keeps improving, whereas it collapses without using gSR.  This confirms our hypothesis that constraining spectral norms stabilizes the training. We find that a similar phenomenon also occurs for BigGAN (Fig. ~\ref{gsr_fig:fid_gsr}) which uses SN in $G$, which shows that gSR is complementary to SN.

\noindent\textbf{What should be the ideal number of groups?} Grouping of the ($\mathbf{\gamma^l_y}$) into $\mathbf{\Gamma^l_y} \in \mathbb{R}^{{n_g}\times{n_c}}$ is a central operation in our regularizer formulation (Eq. \ref{gsr_eq:grouping}). We group $\mathbf{\gamma^l_y}$ (and $\mathbf{\beta^l_y}$) $\in \mathbb{R}^{256}$ into a matrix $\mathbf{\Gamma^l_y}$ (and $\mathbf{B^l_y}$) ablate over different combinations of $n_g$ and $n_c$. Table \ref{gsr_tab:group_ablations} shows that FID scores do not change much significantly with $n_g$. As we also use power iteration to estimate the spectral norm $\sigma_{\max} (\mathbf{\Gamma^l_y})$, we report iteration complexity (multiplications). Since grouping into square matrix($n_g = 16$) gives slightly better FIDs while also being time efficient we use it for our experiments. We also provide additional mathematical intuition for the optimality of choice of $n_c = n_g$ in Appendix. %

\begin{table*}[t]
\parbox{.4\linewidth}{
\centering
    \centering
    \caption{{Quantitative comparison of spectral regularizers.} Comparison against Different Spectral Norm Regularizers on grouped cBN parameters.}
    \label{gsr_tab:compare_reg_norm}
    \begin{tabular}{l|cc}
    \toprule
         & FID($\downarrow$) & IS($\uparrow$) \\ \midrule
         SNGAN~\cite{miyato2018cgans} & 30.62$_{\pm0.07}$ & 6.80$_{\pm0.07}$ \\
        + gSN~\cite{miyato2018spectral}  & 23.97$_{\pm0.13}$ & 7.49$_{\pm0.05}$ \\
        + gSRIP~\cite{bansal2018can}  & 23.67$_{\pm0.02}$ & 7.79$_{\pm0.06}$ \\
        \rowcolor{gray!10} + gSR (Ours)  & \textbf{18.58}$_{\pm0.10}$ & \textbf{7.80}$_{\pm0.09}$ \\ \bottomrule
    \end{tabular}
}
\hfill
\parbox{.53\linewidth}{
        \centering
    \caption{{Group size ablations.} We report average FID and IS on CIFAR-10 dataset. $n_g=16$ gives the best FID while also being computationally efficient, measured by per Iteration (Iter.) complexity. (Iter. complexity for power iteration method is calculated as ($n_g^2$ + $n_c^2$) x number of power iterations (4 in our setting)).}
    \label{gsr_tab:group_ablations}
    \resizebox{\linewidth}{!}{
    {\begin{tabular}{c|c|c|c|c} 
    \toprule
    $n_g$ & $n_c$ & FID($\downarrow$) & IS($\uparrow$) & Iter. Complexity($\downarrow$) \\ \midrule
         4 & 64 & 20.16$_{\pm0.03}$ & \textbf{7.96$_{\pm0.01}$} & 16448 \\
         8 & 32 & 18.69$_{\pm0.06}$ & 7.80$_{\pm0.01}$ & 4352\\
         16 & 16 & \textbf{18.58$_{\pm0.10}$} & 7.80$_{\pm0.09}$ & \textbf{2048}\\
         32 & 8 & 20.19$_{\pm0.06}$ & 7.85$_{\pm0.01}$ & 4352\\ \bottomrule
    \end{tabular}}
}}
\end{table*}
\section{Conclusion and Future Work}
In this work we identify a novel failure mode of \textit{class-specific mode collapse}, which occurs when conditional GANs are trained on long-tailed data distribution. Through our analysis we find that the class-specific collapse for each class correlates closely with a sudden increase (explosion) in the spectral norm of its (grouped) conditional BatchNorm  (cBN) parameters. To mitigate the spectral explosion we develop a novel group Spectral Regularizer (gSR), which constrains the spectral norms and alleviates mode collapse. The gSR reduces spectral norms (estimated through power iteration) of grouped parameters and leads to decorrelation of parameters, which enables GAN to effectively improve on long-tailed data distribution without collapse. Our empirical analysis shows that gSR:  a) leads to improved image generation from conditional GANs (by alleviating class-specific collapse), and b) effectively complements exiting regularizers on discriminator to achieve state-of-the-art image generation performance on long-tailed datasets. One of the limitations present in our framework is that it introduces additional hyperparameter $\lambda$ for the regularizer, an implicit (i.e. hyperparameter free) decorrelated parameterization for alleviating class-specific mode collapse is a good direction for future work. We hope that this work leads to further research on training better GANs using real-world long-tailed datasets. \\

\chapter{\paptitle: Class-Consistent and Diverse Image Generation through StyleGANs}
\label{chap:NoisyTwins}

\begin{changemargin}{7mm}{7mm} 
StyleGANs are at the forefront of controllable image generation as they produce a latent space that is semantically disentangled, making it suitable for image editing and manipulation. However, the performance of StyleGANs severely degrades when trained via class-conditioning on large-scale long-tailed datasets.  We find that one reason for degradation is the collapse of latents for each class in the $\mathcal{W}$ latent space. With NoisyTwins, we first introduce an effective and inexpensive augmentation strategy for class embeddings, which then decorrelates the latents based on self-supervision in the  $\mathcal{W}$  space. This decorrelation mitigates collapse, ensuring that our method preserves intra-class diversity with class-consistency in image generation. We show the effectiveness of our approach on large-scale real-world long-tailed datasets of ImageNet-LT and iNaturalist 2019, where our method outperforms other methods by $\sim 19\%$ on FID, establishing a new state-of-the-art.
\end{changemargin}

\section{Introduction}
StyleGANs~\cite{karras2019style, karras2020analyzing} have shown unprecedented success in image generation, particularly on well-curated and articulated datasets (eg. FFHQ for face images, etc.). In addition to generating high fidelity and diverse images, StyleGANs also produce a disentangled latent space, which is extensively used for image editing and manipulation tasks~\cite{wu2021stylespace}. As a result, StyleGANs are being extensively used in various applications like face-editing~\cite{shen2020interpreting, harkonen2020ganspace}, video generation~\cite{stylegan_v, digan}, face reenactment~\cite{bounareli2022finding}, etc., which are a testament to their usability and generality. 
However, despite being successful on well-curated datasets, training StyleGANs on in-the-wild and multi-category datasets is still challenging. A large-scale conditional StyleGAN (i.e. StyleGAN-XL) on ImageNet was recently trained successfully by Sauer \etal~\cite{Sauer2021ARXIV} using the ImageNet pre-trained model through the idea of a projection discriminator~\cite{Sauer2021NEURIPS}. While the StyleGAN-XL uses additional pre-trained models, obtaining such models for distinctive image domains like medical, forensics, and fine-grained data may not be feasible, which limits its generalization across domains.

\begin{figure}[!t]
    \centering
    \includegraphics[width=0.75\columnwidth]{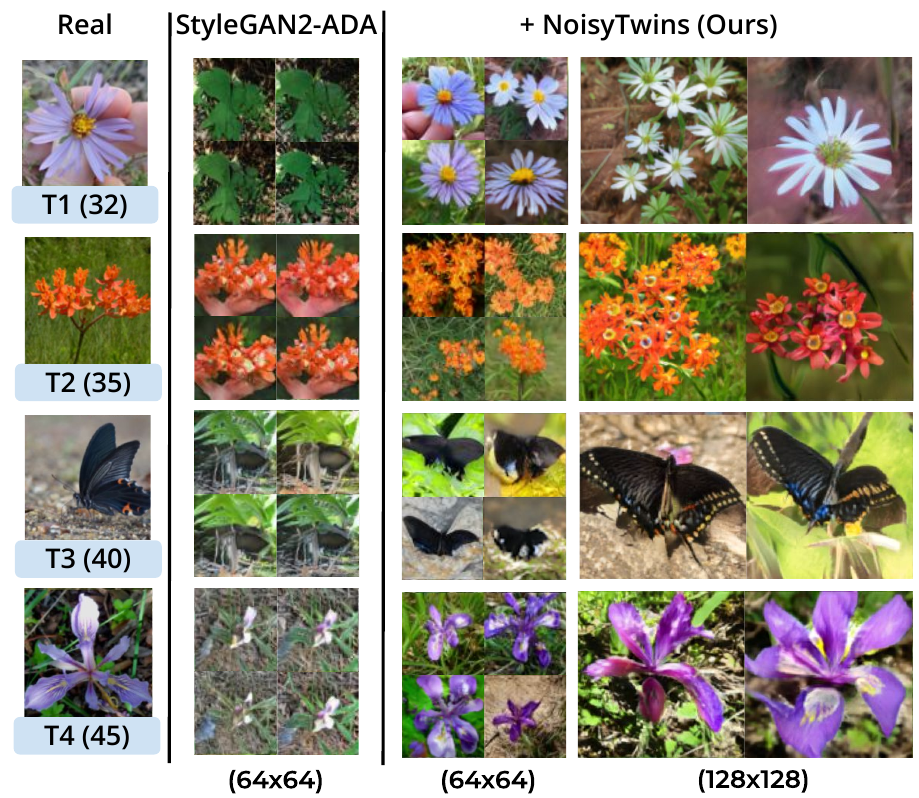}
    \caption{{Qualitative Comparison on tail classes (T1-T4) for iNaturalist 2019.} We provide sample(s) from real class (with class frequency), generated by StyleGAN2-ADA and after adding proposed NoisyTwins. NoisyTwins achieves remarkable diversity, class-consistency and quality by just using ~{38} samples on average.}
    \label{nt_fig:result_teaser}
    \vspace{-5.0mm}
\end{figure}

\begin{figure*}[!t]
    \centering
    \includegraphics[width=\linewidth]{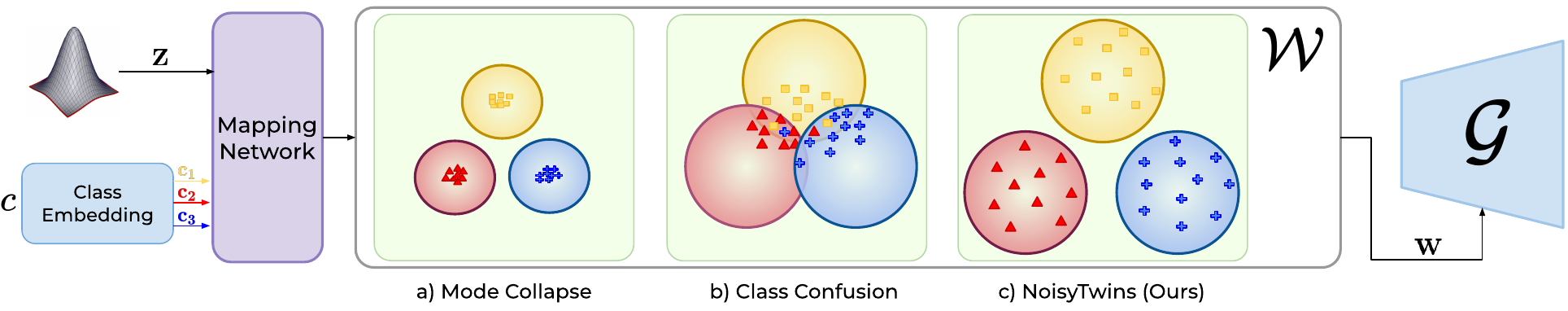}
    \caption{{Schematic illustration of $\mc{W}$ space for different GANs.} Existing conditioning methods either suffer from mode collapse~\cite{Karras2020ada} or lead to class confusion~\cite{rangwani2022gsr} in $\mc{W}$ space. With proposed NoisyTwins, we achieve intra class diversity while avoiding class confusion.}
    \label{nt_fig:teaser}
     \vspace{-5mm}
\end{figure*}

In this work, we aim to train vanilla class-conditional StyleGAN without any pre-trained models on challenging real-world long-tailed data distributions. As training StyleGAN with augmentations~\cite{Karras2020ada, zhao2020diffaugment}  leads to low recall~\cite{improved_pr} (which measures diversity in the generated images) and mode collapse, particularly for minority (i.e. tail) classes.
For investigating this phenomenon further, we take a closer look at the latent $\mc{W}$ space of StyleGAN that is produced by a fully-connected mapping network that takes the conditioning variables $\mb{z}$ (i.e. random noise) and class embedding $\mb{c}$ as inputs. The vectors $\mb{w}$ in $\mc{W}$ space are used for conditioning various layers of the generator (Fig. \ref{nt_fig:teaser}).  We find that output vectors $\mb{w}$ from the mapping network hinge on the conditioning variable $\mb{c}$ and become invariant to random conditioning vector $\mb{z}$. This collapse of latents leads to unstable training and is one of the causes of poor recall (a.k.a. mode collapse) for minority classes. Further, on augmenting StyleGAN with recent conditioning and regularization techniques~\cite{rangwani2022gsr, kang2021ReACGAN}, we find that they either lead to a poor recall for minority classes or lead to class confusion (Fig. \ref{nt_fig:teaser}) instead of mitigating the collapse.

To mitigate the collapse of $\mb{w}$ in $\mc{W}$ space, we need to ensure that the change in conditioning variable $\mb{z}$ leads to the corresponding change in $\mb{w}$. Recently in self-supervised learning, several techniques~\cite{zbontar2021barlow, bardes2022vicreg} have been introduced to prevent the collapse of learned representations by maximizing the information content in the feature dimensions. Inspired by them we propose \paptitle{}, in which we first generate inexpensive twin augmentations for class embeddings and then use them to decorrelate the $\mb{w}$ variables through self-supervision. The decorrelation ensures that $\mb{w}$ vectors are diverse for each class and the GAN is able to produce intra-class diversity among the generated images. 

We evaluate \paptitle{} on challenging benchmarks of large-scale long-tailed datasets of ImageNet-LT~\cite{liu2019large} and iNaturalist 2019~\cite{van2018inaturalist}. These benchmarks are particularly challenging due to a large number of classes present, which makes GANs prone to class confusion.
On the other hand, as these datasets are long-tailed with only a few images per class in tail classes, generating diverse images for those classes is challenging. %
We observe that existing metrics used in GAN evaluations are not able to capture both class confusion and mode collapse.
As a remedy, we propose to use intra-class Frechet Inception Distance (FID)~\cite{heusel2017gans} based on features obtained from pre-trained CLIP~\cite{radford2021learning} embeddings as an effective metric to measure the performance of class-conditional GANs in long-tailed data setups. 
Using \paptitle{} enables StyleGAN to generate diverse and class-consistent images across classes, mitigating the mode collapse and class confusion issues in existing state-of-the-art (SotA) (Fig.~\ref{nt_fig:result_teaser}). Further, with NoisyTwins, we obtain diverse generations for tail classes even with $\leq$ 30 images, which can be attributed to the transfer of knowledge from head classes through shared parameters (Fig.~\ref{nt_fig:result_teaser} and \ref{nt_fig:qualitative_imgnet_lt}). In summary, we make the following contributions:

\begin{enumerate}[topsep=0pt,itemsep=-1ex,partopsep=1ex,parsep=1ex]
    \item We evaluate various recent SotA GAN conditioning and regularization techniques on the challenging task of long-tailed image generation. We find that all existing methods either suffer from mode collapse or lead to class confusion in generations.%
    \item To mitigate mode collapse and class confusion, we introduce \paptitle{}, an effective and inexpensive augmentation strategy for class embeddings that decorrelates latents in the $\mc{W}$ latent space (Sec.~\ref{nt_sec:approach}).
    \item We evaluate NoisyTwins on large-scale long-tailed datasets of ImageNet-LT and iNaturalist-2019, where it consistently improves the StyleGAN2 performance ($\sim 19\%$), achieving a new SotA. Further, our approach can also prevent mode collapse and enhance the performance of few-shot GANs (Sec. \ref{nt_subsec:few_shots_new}).
\end{enumerate}

\section{Related Works}
 \vspace{1mm}\noindent \textbf{StyleGANs.} Karras \etal introduced StyleGAN~\cite{karras2019style} and subsequently improved its image quality in StyleGAN2. StyleGAN could produce high-resolution photorealistic images as demonstrated on various category-specific datasets. It introduced a mapping network, which mapped the sampled noise into another latent space, which is more disentangled and semantically coherent, as demonstrated by its downstream usage for image editing and manipulation~\cite{Alaluf_2022_CVPR, Patashnik_2021_ICCV, shen2020interfacegan, shen2021closedform, parihar2022everything}. Further, StyleGAN has been extended to get novel views from images~\cite{FreeStyleGAN2021, liu2d3d, Shi2021Lifting2S}, thus making it possible to get 3D information from it. These downstream advances are possible due to the impressive performance of StyleGANs on class-specific datasets (such as faces). However, similar photorealism levels are yet uncommon on multi-class long-tailed datasets (such as ImageNet).

\vspace{1mm} \noindent  \textbf{GANs for Data Efficiency and Imbalance.}
Failure of GANs on less data was concurrently reported by Karras \etal ~\cite{Karras2020ada} and Zhao \etal ~\cite{zhao2020diffaugment}. The problem is rooted in the overfitting of the discriminator due to less real data. Since then, the proposed solutions for this problem have relied on a) augmenting the data, b) introducing regularizers, and c) architectural modifications. Karras \etal ~\cite{Karras2020ada} and Zhao \etal ~\cite{zhao2020diffaugment} relied on differentiable data augmentation before passing images into the discriminator to solve this problem. DeceiveD~\cite{jiang2021DeceiveD} proposed to introduce label-noise for discriminator training. LeCamGAN ~\cite{lecamgan} finds that enforcing LeCam divergence as a regularization trick in the discriminator can robustify GAN training under a limited data setting. DynamicD~\cite{yang2022improving} tunes the capacity of the discriminator on-the-fly during training. While these methods can handle the data inefficiency, they are ineffective on class imbalanced long-tailed data distribution~\cite{rangwani2022gsr}.

CBGAN~\cite{rangwani2021class} proposed a solution to train the unconditional GAN model on long-tailed data distribution by introducing a signal from the classifier to balance the classes generated by GAN. In a long-tailed class-conditional setting, gSR~\cite{rangwani2022gsr} proposes to regularize the exploding spectral norms of the class-specific parameters of the GAN. Collapse-by-conditioning~\cite{shahbazi2022collapse} addresses the limited data in classes by introducing a training regime that transitions from an unconditional to a class-conditioned setting, thus exploiting the shared information across classes during the early stages of the training. However, these methods suffer from either class confusion or poor generated image quality on large datasets, which is resolved by \paptitle{}.

\vspace{1mm} \noindent  \textbf{Self-Supervised Learning for GANs.} Ideas from Self-supervised learning have shown their benefits in GAN training. IC-GAN~\cite{casanova2021instanceconditioned} trains GAN conditioned on embeddings on SwAV~\cite{caron2020unsupervised}, which led to remarkable improvement in performance on the long-tailed version of ImageNet. InsGen~\cite{yang2021dataefficient} and ReACGAN~\cite{kang2020ContraGAN, kang2021ReACGAN} introduce the auxiliary task of instance discrimination for the discriminator, thereby making the discriminator focus on multiple tasks and thus alleviating discriminator overfitting. While InsGen relies on both noise space and image space augmentations, ReACGAN and ContraGAN follow only image space augmentations. Contrary to these, \paptitle{} performs augmentations in the class-embedding space and contrasts them in the $\mathcal{W}$-space of the generator instead of the discriminator.
\begin{figure*}[t]
    \centering
    \includegraphics[width=\linewidth]{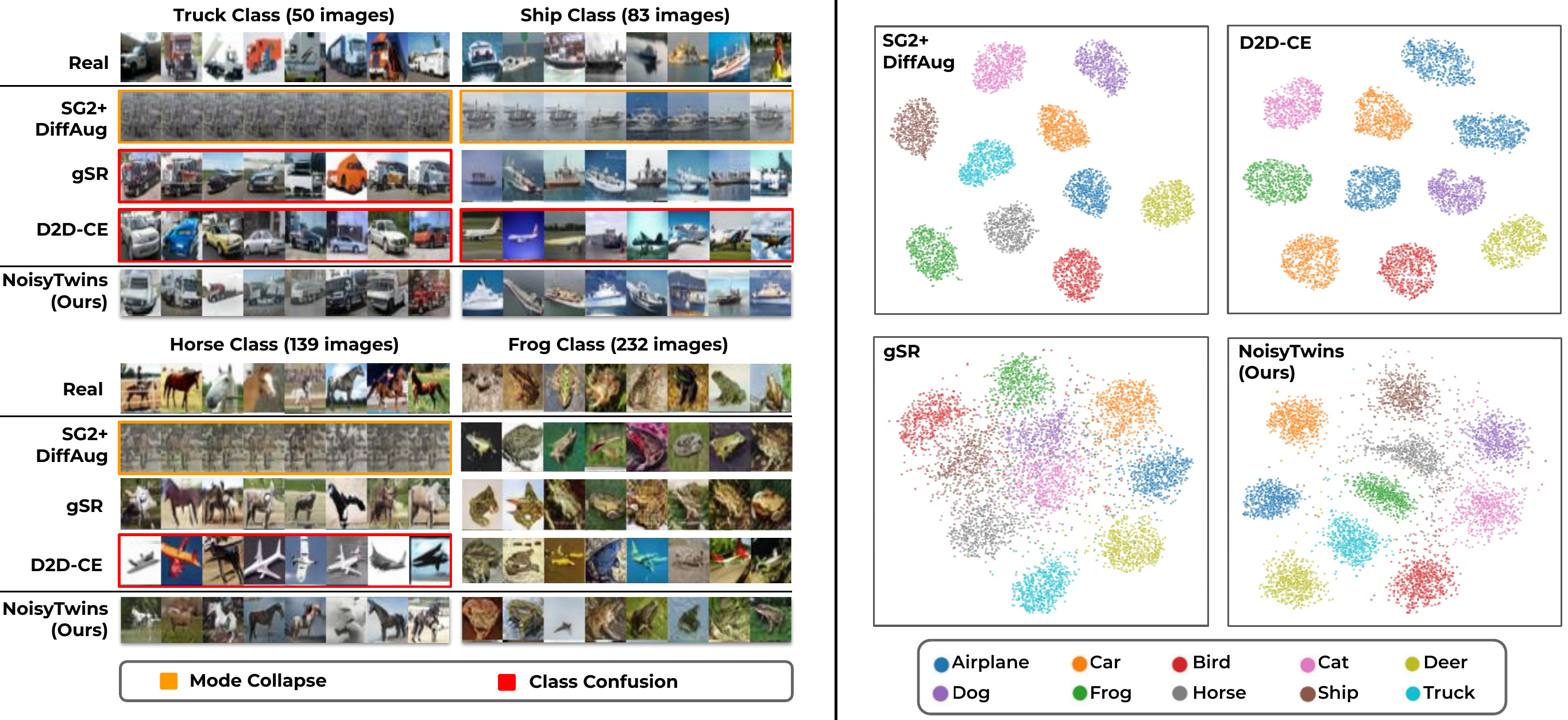}
     \caption{{Comparison of GANs and their $\mc{W}$ space for CIFAR10-LT.} We plot the generated images on (\textit{left}) and generate a t-SNE plot of $\mathbf{w}$ latents for generated images in $\mc{W}$ space (\textit{right}). We find that mode collapse and class confusion in images is linked to the corresponding collapse and confusion in latent $\mc{W}$ space. Our proposed NoisyTwins mitigates both collapse (\textit{left}) and confusion (\textit{right}) simultaneously.}
    \label{nt_fig:cifar10_comparisons_motivation}
    \vspace{-5mm}
\end{figure*}

\section{Preliminaries}

\subsection{StyleGAN}
StyleGAN~\cite{karras2019style} is a Generative Adversarial Network comprising of its unique Style Conditioning Based Generator ($\mc{G}$) and discriminator network ($\mc{D}$) trained jointly. We will focus on the architecture of StyleGAN2~\cite{Karras2019stylegan2} as we use it in our experiments, although our work is generally applicable to all StyleGAN architectures. The StyleGAN2 generator is composed of blocks that progressively upsample the features and resolution inspired by Progressive GAN~\cite{karras2017progressive}, starting from a single root image. The diversity in the images comes from conditioning each block of image generation through conditioning on the latent coming from the mapping network (Fig. \ref{nt_fig:teaser}). The mapping network is a fully connected network that takes in the conditioning variables, the $\mb{z} \in \mbb{R}^d$ coming from a random distribution (e.g., Gaussian, etc.) and class conditioning label $c$ which is converted to an embedding $\mb{c} \in \mbb{R}^d$. The mapping network takes these and outputs vectors $\mb{w}$ in the $\mc{W}$ latent space of StyleGAN, which is found to be semantically disentangled to a high extent~\cite{wu2021stylespace}. The $\mb{w}$ is then processed through an affine transform and passed to each generator layer for conditioning the image generation process through Adaptive Instance Normalization (AdaIN)~\cite{huang2017arbitrary}. The images from generator $\mc{G}$, along with real images, are passed to discriminator $\mc{D}$ for training. The training utilizes the non-saturating adversarial losses~\cite{goodfellow2020generative} for $\mc{G}$ and $\mc{D}$ given as:
\begin{flalign}
    \min_{\mc{D}} \mc{L}_{\mc{D}} &= \sum_{i=1}^{m} \log(\mc{D}(\mb{x}_i)) + \log(1 - \mc{D}(\mc{G}(\mb{z}_i, \mb{c}_i))) \\
     \min_{\mc{G}} \mc{L}_{\mc{G}} &= \sum_{i=1}^{m} -\log(\mc{D}(\mc{G}(\mb{z}_i, \mb{c}_i))).
\end{flalign}
We now describe the issues present in the StyleGANs trained on long-tailed data and their analysis in $\mc{W}$ space.

\subsection{Class Confusion and Class-Specific Mode Collapse in Conditional StyleGANs}
To finely analyze the performance of StyleGAN and its variants on long-tailed datasets, we train them on the CIFAR10-LT dataset. In Fig. \ref{nt_fig:cifar10_comparisons_motivation}, we plot the qualitative results of generated images and create a t-SNE plot for latents in $\mc{W}$ space for each class.  
We first train the StyleGAN2 baseline with augmentations (DiffAug)~\cite{Karras2020ada, zhao2020diffaugment}. We find that it leads to mode collapse, specifically for tail classes (Fig. \ref{nt_fig:cifar10_comparisons_motivation}). In conjunction with images, we also observe that corresponding t-SNE embeddings are also collapsed near each class's mean in $\mc{W}$ space. Further, recent methods which have proposed the usage of contrastive learning for GANs, improve their data efficiency and prevent discriminator overfitting~\cite{kang2020ContraGAN, jeong2021contrad}. We also evaluate them by adding the contrastive conditioning method, which is D2D-CE loss-based on ReACGAN~\cite{kang2021ReACGAN}, to the baseline, where in results, we observe that the network omits to learn tail classes and produces head class images at their place (i.e., class confusion). In Fig.~\ref{nt_fig:cifar10_comparisons_motivation}, it can be seen that the network confuses semantically similar classes, that is, generating cars (head or majority class) in place of trucks and airplanes (head class) instead of ships. In the $\mc{W}$ space, we find the same number of clusters as the number of classes in the dataset. However, the tail label cluster images also belong to the head classes of cars and airplanes. In a very recent work gSR~\cite{rangwani2022gsr}, it has been shown that constraining the spectral norm of $\mc{G}$ embedding parameters can help reduce the mode collapse and lead to stable training. However, we find that constraining the embeddings leads to class confusion, as seen in t-SNE visualization in Fig.~\ref{nt_fig:cifar10_comparisons_motivation}. We find that this class confusion gets further aggravated when StyleGAN is trained on datasets like ImageNet-LT, which contain a large number of classes, along with a bunch of semantically similar classes (Sec.~\ref{nt_subsec:result_inat_imnet}).  
Based on our $\mc{W}$ space analysis and qualitative results above, we observe that the class confusion and mode collapse of images is tightly coupled with the structure of $\mc{W}$ space. Further, the recent SotA methods are either unable to prevent collapse or suffer from class confusion. Hence, this work aims to develop a technique that mitigates both confusion and collapse.

\section{Approach}
\label{nt_sec:approach}
In this section, we present our method \paptitle{}, which introduces noise-based augmentation twins in the conditional embedding space (Sec. \ref{nt_subsec: noise-aug}), and then combines it with the Barlow-Twins-based regularizer from the self-supervised learning (SSL) paradigm to resolve the issue of class confusion and mode collapse (Sec.~\ref{nt_subsec:bt}).

\subsection{Noise Augmentation in Embedding Space}
\label{nt_subsec: noise-aug}
As we observed in the previous section that $\mb{w}$ vectors for each sample become insensitive to changes in $\mb{z}$. This collapse in $\mb{w}$ vectors for each class leads to mode collapse for baselines (Fig. \ref{nt_fig:cifar10_comparisons_motivation}). One reason for this could be the fact that $\mb{z}$ is composed of continuous variables, whereas the embedding vectors $\mb{c}$ for each class are discrete.
Due to this, the GAN converges to the easy degenerate solution where it generates a single sample for each class, becoming insensitive to changes in $\mb{z}$. For inducing some continuity in $\mb{c}$ embeddings vectors, we introduce an augmentation strategy where we add i.i.d. noise of small magnitude in each of the variables in $\mb{c}$. Based on our observation (Fig.~\ref{nt_fig:cifar10_comparisons_motivation}) and existing works~\cite{rangwani2022gsr}, there is a high tendency for mode collapse in tail classes. Hence we add noise in embedding space that is proportional to the inverse of the frequency of samples . We provide the mathematical expression of noise augmentation $\mb{\tilde{c}}$ below:
\begin{equation}
    \mb{\tilde{c}} \sim \mb{c} + \mc{N}(\mathbf{0}, \sigma_c\mbb{I}_d) \; \text{where} \; \sigma_c = \sigma \frac{( 1 - \alpha)}{1 - \alpha^{n_{c}}}.
    \label{nt_eq:noise_aug_eq}
\end{equation}
Here $n_c$ is the frequency of training samples in class $c$, $\mbb{I}$ is the identity matrix of size $d \times d$, and $\alpha, \sigma$ are hyper-parameters. The expression of $\sigma_c$ is from the effective number of samples~\cite{cui2019class}, which is a softer version of inverse frequency proportionality. In contrast to the image space augmentation, these \emph{noise augmentations come for free} as there is no significant additional computation overhead. 
This noise is added in the embedding $\mb{c}$ before passing it to the generator and the discriminator, which ensures that the class embeddings occupy a continuous region in latent space. \\ 
\begin{figure*}[t]
    \centering
    \includegraphics[width=\linewidth]{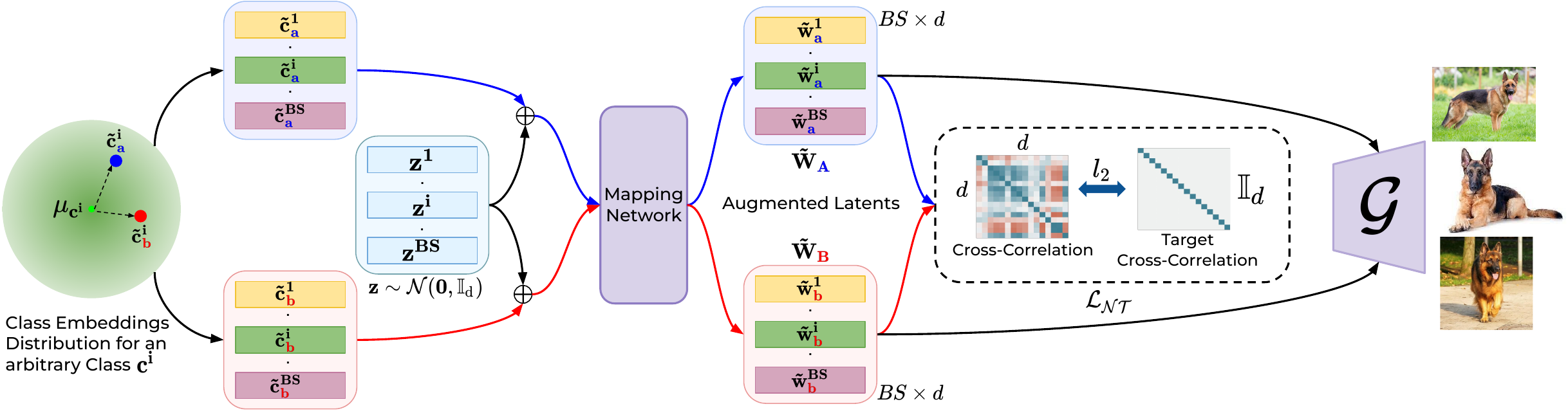}
    \caption{{Overview of NoisyTwins.} For the $i^{th}$ sample of class $c^i$, we create twin augmentations ($\tilde{\mb{c}}^{i}_a$, $\tilde{\mb{c}}^{i}_b$), by sampling from a Gaussian centered at class embedding ($\mu_{\mb{c{^i}}}$).  After this, we concatenate them with the same $\mb{z}^{i}$ and obtain ($\tilde{\mb{w}}^{i}_{a}, \tilde{\mb{w}}^{i}_{b}$) from the mapping network, which we stack in batches of augmented latents ($\mb{\tilde{W}_{A}}$ and $\mb{\tilde{W}_{B}}$). The twin ($\tilde{\mb{w}}^{i}_{a}, \tilde{\mb{w}}^{i}_{b}$) vectors are then made invariant to augmentations (similar) in the latent space by minimizing cross-correlation~\cite{bardes2022vicreg, zbontar2021barlow} between the latents of two augmented batches ($\mb{\tilde{W}_{A}}$ and $\mb{\tilde{W}_{B}}$). }
    \label{nt_fig:overview_noisy}
    \vspace{-5mm}
\end{figure*}
\noindent \textbf{Insight:} The augmentation equation above (Eq. \ref{nt_eq:noise_aug_eq})  can be interpreted as approximating the discrete random variable $\mb{c}$ with a Gaussian with finite variance and the embedding parameters $\mb{c}$ being the mean $\mb{\mu_{c}}$: 
\begin{equation}
    \mb{\tilde{c}} \sim \mc{N}(\mb{\mu_{c}}, \sigma_c \mbb{I}_d).
\end{equation}
This leads to the class-embedding input $\mb{\tilde{c}}$ to mapping network to have a Gaussian distribution, similar in nature to $\mb{z}$. This noise augmentation strategy alone mitigates the degenerate solution of class-wise mode collapse to a great extent (Table~\ref{nt_tab:imageNetLT_iNat}) and helps generate diverse latent $\mb{w}$ for each class. Due to the diverse $\mb{w}$ conditioning of the GAN, it leads to diverse image generation. 

\subsection{Invariance in \texorpdfstring{$\mc{W}$}{W}-Space  with \paptitle{}}
\label{nt_subsec:bt}
The augmentation strategy introduced in the previous section expands the region for each class in $\mc{W}$ latent space. Although that does lead to diverse image generation as $\mb{w}$ are diverse; however, this does not ensure that these $\mb{w}$ will generate class-consistent outputs for augmentations in embedding ($\tilde{\mb{c}}$). To ensure class consistent predictions, we need to ensure invariance in $\mb{w}$ to noise augmentation.

For enforcing invariance to augmentations, a set of recent works~\cite{zbontar2021barlow, bardes2022vicreg, grill2020bootstrap} in self-supervised learning make the representations of augmentations similar through regularization. Among them, we focus on Barlow Twins as it does not require a large batch size of samples. Inspired by Barlow twins, we introduce NoisyTwins (Fig. \ref{nt_fig:overview_noisy}), where we generate twin augmentations $\tilde{\mb{c}}_a$
and $\tilde{\mb{c}}_b$ of the same class embedding ($\mu_{c}$) and concatenate them to same $\mb{z}$. After creating a batch of such inputs, they are passed to the mapping network to get batches of augmented latents ($\tilde{\mb{W}}_A$ and $\tilde{\mb{W}}_B$). These batches are then used to calculate the cross-correlation matrix of latent variables given as: 
\begin{equation}
     \mb{C}_{j,k} = \frac{\underset{(\tilde{\mb{w}}_{a}, \tilde{\mb{w}}_{b}) \in (\tilde{\mb{W}}_{A}, \tilde{\mb{W}}_B)}{\sum} \tilde{\mb{w}}_{a}^{j} \tilde{\mb{w}}_{b}^{k}}{\underset{\tilde{\mb{w}}_{a} \in \tilde{\mb{W}}_{A} }{\sum} {\tilde{\mb{w}}_{a}^{j} \tilde{\mb{w}}_{a}^{j}} \underset{\mb{\tilde{w}}_{b} \in \tilde{\mb{W}}_{B} }{\sum} \tilde{\mb{w}}_{b}^{k} \tilde{\mb{w}}_{b}^{k}} 
\end{equation}
here the  matrix $\mb{C}$ is a square matrix of size same as of latents $\mb{w}$. The final loss based on confusion matrix
is given as:
\begin{equation}
\label{nt_eq:noisytwins}
    \mc{L}_{\mc{NT}} = \sum_{j} (1 - \mb{C}_{jj}^{2}) + \gamma\sum_{j\neq k} \mb{C}_{j,k}^2.
\end{equation}
The first term tries to make the two latents $(\tilde{\mb{w}}_{a}$ and $\tilde{\mb{w}}_{b}$) invariant to the noise augmentation applied (i.e. similar), whereas the second term tries to de-correlate the different variables, thus maximizing the information in $\mb{w}$ 
 vector~\cite{zbontar2021barlow} (See Appendix). The $\gamma$ is the hyper-parameter that determines the relative importance of the two terms. This loss is then added to the generator loss term ($\mc{L_G} + \lambda L_{\mc{NT}}$) and optimized through backpropagation. The above procedure comprises our proposed method, \paptitle{} (Fig. \ref{nt_fig:overview_noisy}), which we empirically evaluate in the subsequent sections.

\section{Experimental Evaluation}

\subsection{Setup}
\vspace{1mm} \noindent  \textbf{Datasets:} We primarily apply all methods on long-tailed datasets, as GANs trained on them are more prone to class confusion and mode collapse. We first report on the commonly used CIFAR10-LT dataset with an imbalance factor (\ie ratio of most to least frequent class) of 100. To show our approach's scalability and real-world application, we test our method on the challenging ImageNet-LT and iNaturalist 2019 datasets. The ImageNet-LT~\cite{liu2019large} is a long-tailed variant of the 1000 class ImageNet dataset, with a plethora of semantically similar classes (e.g., Dogs, Birds etc.), making it challenging to avoid class confusion. iNaturalist-2019~\cite{van2018inaturalist} is a real-world long-tailed dataset composed of 1010 different variants of species, some of which have fine-grained differences in their appearance. For such fine-grained datasets, ImageNet pre-trained discriminators~\cite{Sauer2021NEURIPS, Sauer2021ARXIV} may not be useful, as augmentations used to train the model makes it invariant to fine-grained changes.

\vspace{1mm} \noindent\textbf{Training Configuration:} We use StyleGAN2 architecture for all our experiments. All our experiments are performed using PyTorch-StudioGAN implemented by Kang \etal~\cite{kang2022StudioGAN}, which serves as a base for our framework. 
We use Path Length regularization (PLR) as it ensures changes in $\mc{W}$ space lead to changes in images, using a delayed PLR on ImageNet-LT following~\cite{Sauer2021ARXIV}. We use a batch size of 128 for all our experiments, with one G step per D step. Unless stated explicitly, we used the general training setup for StyleGANs from~\cite{kang2022StudioGAN}, including methods like $R_1$ regularization~\cite{mescheder2018training}, etc. More details on the exact training configuration for each dataset are provided in Appendix.

\begin{figure}[!t]
    \centering
    \includegraphics[width=0.75\linewidth]{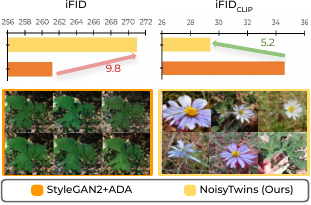}
    \caption{{Choice of Eval. backbone:} intra-FID (iFID) of a class based on InceptionV3 backbone (left plot) is not able to capture the mode collapse (increase in iFID in the absence of mode collapse). This is well-captured by iFID$_\mathrm{CLIP}$ based on CLIP~\cite{radford2021learning} backbone (right plot, decrease in iFID in the absence of mode collapse).}
    \label{nt_fig:iFid Comparison}
    \vspace{-5mm}
\end{figure}

\vspace{1mm} \noindent \textbf{Metrics:}
In this work we use the following metrics for evaluation of our methods: \\
\noindent \textbf{a) FID:} Fr\'echet Inception Distance~\cite{heusel2017gans} is the Wasserstein-2 Distance between the real and validation data. We use a held-out validation set and 50k generated samples to evaluate FID in each case. As FID is biased towards ImageNet and can be arbitrarily manipulated,
we also report FID$_{\mathrm{CLIP}}$.\\ 
\noindent \textbf{b) Precision \& Recall:} As we aim to mitigate the mode collapse and achieve diverse generations across classes, we use improved Precision \& Recall~\cite{kynkaanniemi2019improved} metrics, as poor recall indicates mode collapse~\cite{Sauer2021ARXIV}. \\
\noindent \textbf{d) Intra-Class FID$_\mathrm{CLIP}$ (iFID$_\mathrm{CLIP}$):}
The usage of only FID based on Inception-V3 Networks for evaluation of Generative Models has severe limitations, as it has been found that FID can be reduced easily by some fringe features~\cite{Kynkaanniemi2022}. iFID is computed by taking FID between 5k generated and real samples for the same class. As we want to evaluate both class consistency and diversity, we find that similar limitations exist for intra-class FID (iFID), which has been used to evaluate class-conditional GANs~\cite{kang2022StudioGAN}.  In Fig.~\ref{nt_fig:iFid Comparison}, we show the existence of generated images for a particular class (more in Appendix) from models trained on iNaturalist 2019, where iFID is better for the mode collapsed model than the other model generating diverse images. Whereas the iFID$_\mathrm{CLIP}$, based on CLIP backbone can rank the models correctly with the model having mode collapse having high iFID$_\mathrm{CLIP}$. Further, we find that the mean iFID can be deceptive in detecting class confusion and collapse cases, as it sometimes ranks models with high realism better than models generating diversity (See Appendix). Hence, mean iFID$_\mathrm{CLIP}$ (ref. to as iFID$_\mathrm{CLIP}$ in result section for brevity) can be reliably used to evaluate models for class consistency and diversity.

\begin{table*}[!t]
    \centering
    \parbox{\textwidth}{
    \caption{{Quantitative results on ImageNet-LT and iNaturalist 2019 Datasets.} We compare FID($\downarrow$), FID$_\mathrm{CLIP}$($\downarrow$), iFID$_\mathrm{CLIP}$($\downarrow$), Precision($\uparrow$) and Recall($\uparrow$) with other existing \mbox{approaches on StyleGAN2 (SG2).} We obtain an average $\sim19\%$ relative improvement on FID, $\sim33\%$ on FID$_\mathrm{CLIP}$, and $\sim11\%$ on iFID$_\mathrm{CLIP}$ metrics over the previous SotA on ImageNet-LT and iNaturalist 2019 datasets.
    }
    \vspace{-3mm}
        \label{nt_tab:imageNetLT_iNat}}
    \resizebox{\textwidth}{!}{
    \begin{tabular}{lccccc|ccccc}
    \toprule
         & \multicolumn{5}{c}{ImageNet-LT} & \multicolumn{5}{c}{iNaturalist 2019} \\ \hline
         Method & FID($\downarrow$) & FID$_\mathrm{CLIP}$($\downarrow$) & iFID$_\mathrm{CLIP}$($\downarrow$) & Precision($\uparrow$) & Recall($\uparrow$) & FID($\downarrow$) & FID$_\mathrm{CLIP}$($\downarrow$) & iFID$_\mathrm{CLIP}$($\downarrow$) & Precision($\uparrow$) & Recall($\uparrow$) \\
         \midrule
         SG2~\cite{Karras2019stylegan2}& 41.25 & 11.64 & 46.93 & 0.50 & \underline{0.48} & 19.34 & 3.33 & 38.24 & 0.74 & 0.17\\
         SG2+ADA~\cite{Karras2020ada}& 37.20 & 11.04 & 47.41 & 0.54 & 0.38 & 14.92 & 2.30 & 35.19 & 0.75 & 0.57\\
         SG2+ADA+gSR~\cite{rangwani2022gsr} & 24.78 & 8.21 & 44.42 & 0.63 & 0.35 & 15.17 & 2.06 & 36.22 & 0.74 & 0.46 \\
         \midrule
         
        \rowcolor{gray!10}  SG2+ADA+Noise (Ours) & \underline{22.17} & \underline{7.11} & \underline{41.20} & \textbf{0.72} & 0.33 & \underline{12.87} & \underline{1.37} & \textbf{31.43} & \textbf{0.81} & \underline{0.63}\\
        \rowcolor{gray!10} \; + NoisyTwins (Ours) & \textbf{21.29} & \textbf{6.41} & \textbf{39.74} & \underline{0.67} & \textbf{0.49} & \textbf{11.46} & \textbf{1.14} & \underline{31.50} & \underline{0.79} & \textbf{0.67}\\ \bottomrule
    \end{tabular}
    }
    \vspace{-2.00mm}
\end{table*}

\begin{table*}[t]
\parbox{.6\linewidth}{
\centering
    \caption{{Quantitative results on CIFAR10-LT Dataset.} We compare with other existing \mbox{approaches.} We obtain $\sim26\%$ relative improvement over the existing methods on FID$_\mathrm{CLIP}$ and iFID$_\mathrm{CLIP}$ metrics.}
    \label{nt_tab:CIFAR10LT}
    \resizebox{0.65\textwidth}{!}
    {
    \begin{tabular}{lccccc}
    \toprule
         Method & FID($\downarrow$) & FID$_\mathrm{CLIP}$($\downarrow$) & iFID$_\mathrm{CLIP}$($\downarrow$) & Precision($\uparrow$) & Recall($\uparrow$) \\ \midrule
         SG2+DiffAug~\cite{zhao2020diffaugment}& 31.73 & 6.27 & 11.59 & 0.63 & 0.35\\
         SG2+D2D-CE~\cite{kang2021ReACGAN} & 19.97 & 4.77 & 11.35 & \textbf{0.73} & 0.42 \\ 
         gSR~\cite{rangwani2022gsr} & 22.10 & 5.54 & 9.94 & 0.70 & 0.29 \\
         \midrule
         
        \rowcolor{gray!10}  SG2+DiffAug+Noise (Ours) & 28.90 & 5.26 & 10.65 & 0.71 & 0.38\\
        \rowcolor{gray!10} \; + NoisyTwins(Ours) & \textbf{17.74} & \textbf{3.55} & \textbf{7.24} & 0.70 & \textbf{0.51}\\  \bottomrule
    \end{tabular}
    }
}
\hfill
\parbox{.34\linewidth}{
        \centering
    \caption{{Comparison with SotA approaches on BigGAN.} We compare FID($\downarrow$)  with other existing \mbox{models} on ImageNet-LT (IN-LT) and iNaturalist 2019 (iNat-19).}
    \label{nt_tab:SotA_compare}
    \vspace{-4mm}
    \resizebox{0.28\textwidth}{!}
    {\begin{tabular}{lcc}
    \toprule
         Method &  iNat-19  & IN-LT \\ \midrule
         BigGAN~\cite{brock2018large} & 14.85 & 28.10\\
         \; + gSR~\cite{rangwani2022gsr} & 13.95 & -\\
         ICGAN~\cite{casanova2021instanceconditioned}  & - & 23.40\\
         \midrule
         
        \rowcolor{gray!10}  StyleGAN2-ADA~\cite{Karras2020ada} &  14.92 & 37.20\\
        \rowcolor{gray!10}  + NoisyTwins (Ours) & \textbf{11.46} & \textbf{21.29}\\ \bottomrule
    \end{tabular}}
}
\vspace{-4mm}
\end{table*}

\noindent \textbf{Baselines:}
For evaluating NoisyTwins performance in comparison to other methods, we use the implementations present in StudioGAN~\cite{kang2020ContraGAN}. For fairness, we re-run all the baselines on StyleGAN2 in the same hyperparameter setting. We compare our method to the StyleGAN2 (SG2)
and StyleGAN2 with augmentation (DiffAug~\cite{zhao2020diffaugment} and ADA~\cite{Karras2020ada}) baselines. We further tried to improve the baseline by incorporating the recent LeCam regularization method; however, it resulted in gains only for the iNaturalist 2019 dataset, where we use LeCam for all experiments. Further on StyleGAN2, we also use contrastive D2D-CE loss conditioning (\ie ReACGAN) as a baseline.   However, the D2D-CE baseline completely ignores learning of tail classes (Fig. \ref{nt_fig:cifar10_comparisons_motivation}) for CIFAR10-LT and is expensive to train; hence we do not report results for it for large-scale long-tailed datasets. We also compare our method against the recent SotA group Spectral Normalization (gSR)~\cite{rangwani2022gsr} method, which we implement for StyleGAN2 by constraining the spectral norms of embedding parameters of the generator ($\mc{G}$) as suggested by authors. As a sanity check, we reproduce their results on CIFAR10-LT and find that our implementation matches the reported results correctly. We provide results on all datasets, for both the proposed Noise Augmentation (+ Noise) and the overall proposed NoisyTwins (+NoisyTwins) method.

\subsection{Results on Long-Tailed Data Distributions}
\label{nt_subsec:result_inat_imnet}

\begin{figure*}[t]
    \centering
    \includegraphics[width=\linewidth]{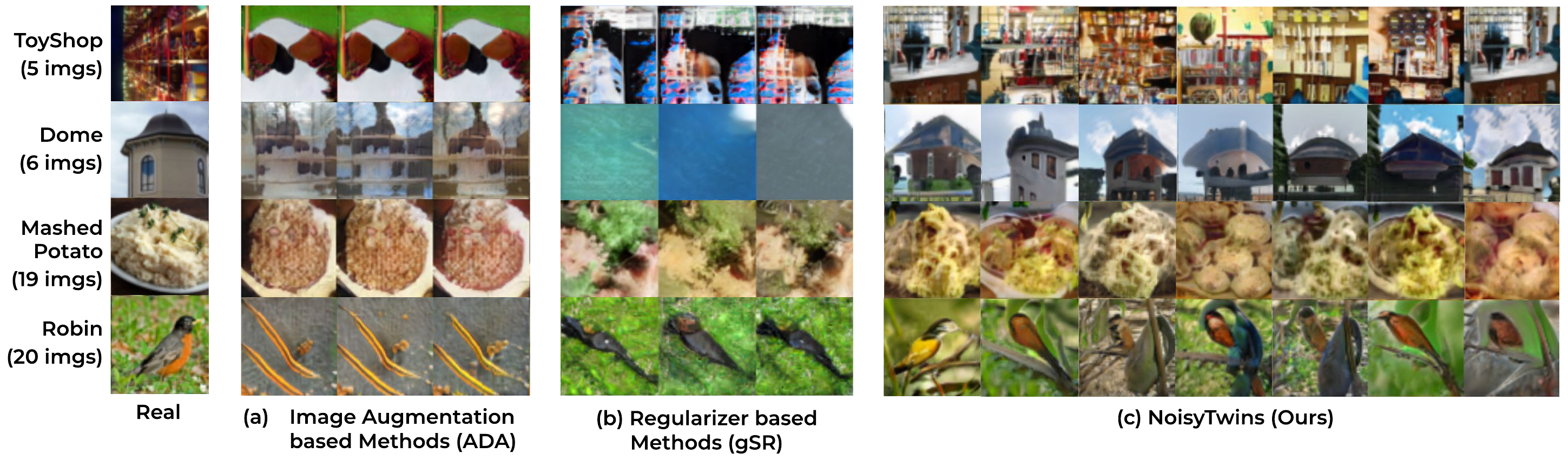}
    \caption{{Qualitative results on ImageNet-LT for tail classes.} We find that existing SotA methods for tail classes show collapsed (a) or arbitrary image generation (b). With NoisyTwins, we observe diverse and class-consistent image generation, even for classes having 5-6 images. The tail classes get enhanced diversity by transferring the knowledge from head classes, as they share parameters.}
    \label{nt_fig:qualitative_imgnet_lt}
    \vspace{-5mm}
\end{figure*}

\noindent\textbf{CIFAR10-LT}. We applied DiffAug~\cite{zhao2020diffaugment} on all baselines, except on gSR, where we found that DiffAug provides inferior results compared to ADA (as also used by authors~\cite{rangwani2022gsr}). It can be observed in Table~\ref{nt_tab:CIFAR10LT} that the addition of NoisyTwins regularization significantly improves over baseline by ($\sim$ 14 FID) along with providing superior class consistency as shown by improved iFID$_\mathrm{CLIP}$. NoisyTwins is also able to outperform the recent gSR regularization method and achieves improved results for all metrics. Further, NoisyTwins improves FID for StyleGAN2-ADA baseline used by  gSR too from 32.08 to 23.02, however the final results are inferior than reported DiffAug baseline results. Further, we observed that despite not producing any tail class images (Fig.~\ref{nt_fig:cifar10_comparisons_motivation}), the D2D-CE baseline has much superior FID in comparison to baselines. Whereas the proposed iFID$_\mathrm{CLIP}$ value is similar for the baseline and D2D-CE model. This clearly demonstrates the superiority of proposed iFID$_\mathrm{CLIP}$ in detecting class confusion. %

\noindent\textbf{Large-scale Long-Tailed Datasets.} We experiment with iNaturalist 2019 and ImageNet-LT. These datasets are particularly challenging as they contain long-tailed imbalances and semantically similar classes, making GANs prone to mode collapse and class confusion. The baselines StyleGAN2 and StyleGAN2-ADA both suffer from mode collapse (Fig. \ref{nt_fig:qualitative_imgnet_lt}), particularly for the tail classes.  Whereas for the recent SotA gSR method, we find that although it undergoes less collapse in comparison to baselines, it suffers from class confusion as seen from similar Intra-FID$_\mathrm{CLIP}$ in comparison to baselines (Table~\ref{nt_tab:imageNetLT_iNat}). Compared to that, our method NoisyTwins improves when used with StyleGAN2-ADA significantly, leading to a relative improvement of $42.7\%$ in FID for ImageNet-LT and  $23.19\%$ on the iNaturalist 2019 dataset when added to StyleGAN2-ADA baseline. Further with Noise Augmentation (+Noise), we observe generations of high-quality class-consistent images, but it also suffers from mode collapse. This can be observed by the high-precision values in comparison to low-recall values. However, adding NoisyTwins regularization over the noise augmentation improves diversity by improving recall  as observed in Table \ref{nt_tab:imageNetLT_iNat}. 

Fig. \ref{nt_fig:qualitative_imgnet_lt} presents the generated images of tail classes for various methods on ImageNet-LT , where NoisyTwins generations show remarkable diversity in comparison to others. The presence of diversity for classes with just 5-6 training images demonstrates successful transfer of knowledge from head classes to tail classes, due to shared parameters. Further, to compare with existing SotA reported results, we compare FID of BigGAN models from gSR~\cite{rangwani2022gsr} and Instance Conditioned GAN (ICGAN)~\cite{casanova2021instanceconditioned}. For fairness, we compare FID on the validation set for which we obtained gSR models from authors and re-evaluate them, as they reported FID on a balanced training set. As  BigGAN models are more common for class-conditioned generation~\cite{kang2022StudioGAN}, their baseline performs superior to StyleGAN2-ADA baselines (Table~\ref{nt_tab:SotA_compare}). However, the addition of NoisyTwins to the StyleGAN2-ADA method improves it significantly, even outperforming the existing methods of gSR (by 18.44\%) and ICGAN (by 9.44\%)  based on BigGAN architecture. This shows that NoisyTwins allows the StyleGAN2 baseline to scale to large and diverse long-tailed datasets.
\subsection{NoisyTwins on Few-Shot Datasets}
\label{nt_subsec:few_shots_new}
We now demonstrate the potential of NoisyTwins in another challenging scenario of class-conditional few-shot image generation from GANs. We perform our experiments using a conditional StyleGAN2-ADA baseline, for which we tune hyper-parameters to obtain a strong baseline. We then apply our method of Noise Augmentation and NoisyTwins over the strong baseline for reporting our results. We use the few-shot dataset of LHI-AnimalFaces~\cite{si2011learning} and a subset of ImageNet Carnivores~\cite{liu2019few, shahbazi2022collapse} to report our results. Table~\ref{nt_tab:few_shot} shows the results of these experiments, where we find that our method, NoisyTwins, significantly improves the FID of StyleGAN2 ADA baseline by (22.2\%) on average for both datasets. Further, combining Noisy Twins with SotA Transitional-cGAN~\cite{shahbazi2022collapse} through official code, also leads to effective improvement in FID. These results clearly demonstrate the diverse potential and applicability of our proposed method NoisyTwins.
\section{Analysis}
We perform analysis of NoisyTwins w.r.t. to its hyperparameters, standard deviation ($\sigma$) of noise augmentation and regularization strength ($\lambda$). We also compare NoisyTwins objective (ref. Eq.~\ref{nt_eq:noisytwins}) with contrastive objective. Finally, we compare NoisyTwins over Latent Diffusion Models for long-tailed class conditional generation task. We perform ablation experiments on CIFAR10-LT, for which additional details and results are present in Appendix. We also present comparison of NoisyTwins for GAN fine-tuning. 
\begin{figure}[!t]
    \centering
    \includegraphics[width=0.75\linewidth]{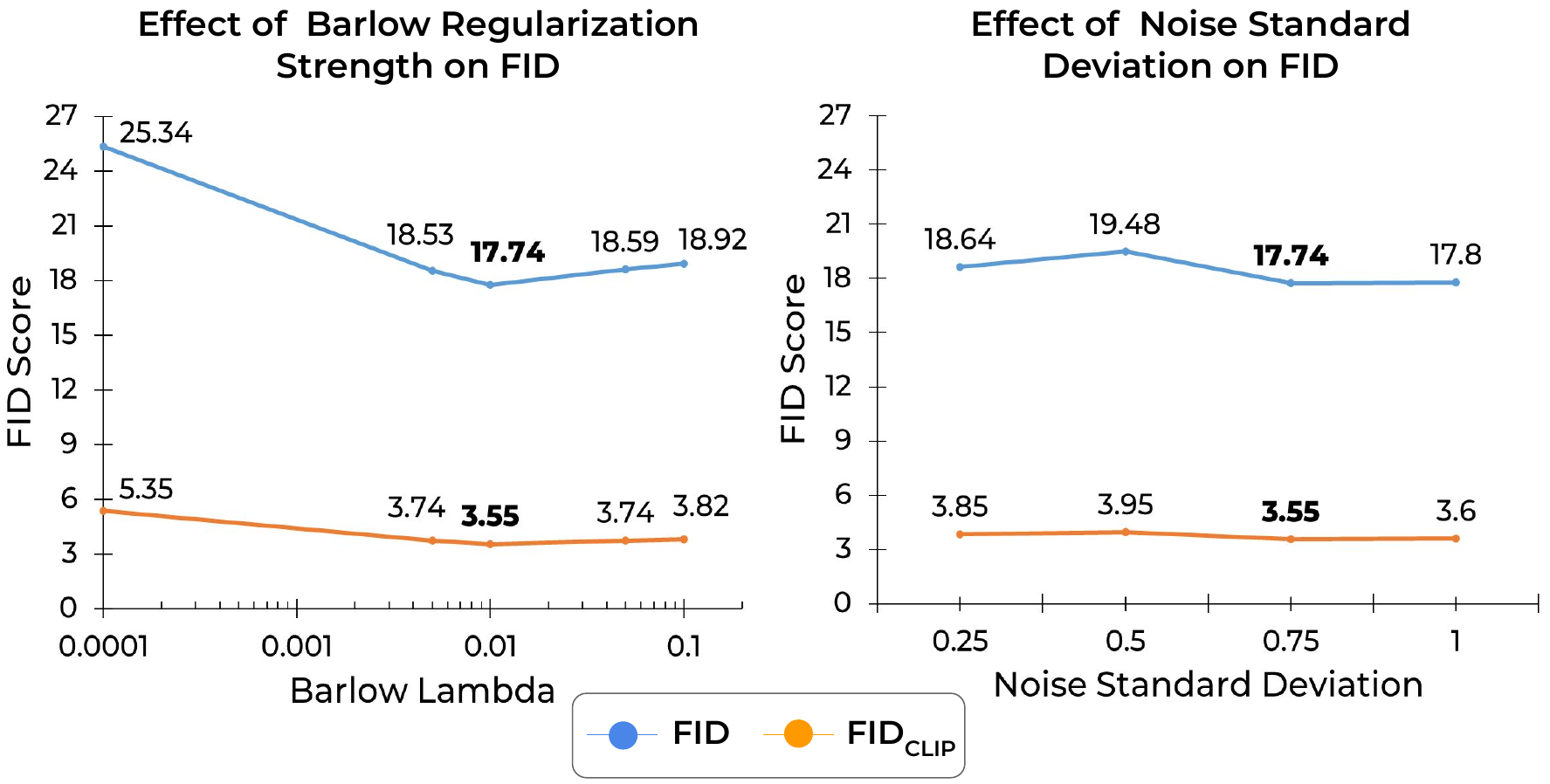}
    \caption{{Ablation of Hyperparameters.} Quantitative comparison on CIFAR10-LT for standard deviation of Noise Augmentation ($\sigma$) and strength ($\lambda$) of NoisyTwins loss.}
    \label{nt_fig:CIFAR10LT_Ablation}
  
\end{figure}
\begin{table}[!t]
    \centering
    \caption{{Quantitative results on ImageNet Carnivore and AnimalFace Datasets.} Our method improves over both StyleGAN2-ADA (SG2-ADA) baseline and SotA Transitional-cGAN .}
    \label{nt_tab:few_shot}
    \resizebox{0.75\textwidth}{!}
    {
    \begin{tabular}{lcc|cc}
    \toprule
        & \multicolumn{2}{c|}{ImageNet Carnivore} & \multicolumn{2}{c}{AnimalFace} \\ \hline
         Method & FID($\downarrow$) & iFID$_\mathrm{CLIP}$($\downarrow$) & FID($\downarrow$) & iFID$_\mathrm{CLIP}$($\downarrow$) \\ \midrule
         SG2~\cite{Karras2019stylegan2}& 111.83 & 36.34 & 94.09 & 29.94 \\
         SG2+ADA~\cite{Karras2020ada}& 22.77 & 12.85 & 20.25 & 11.12 \\
         \midrule
         
        \rowcolor{gray!10}  SG2+ADA+Noise (Ours) & \underline{19.25} & \underline{12.51} & \underline{18.78} & \underline{10.42}\\
        \rowcolor{gray!10} \; + NoisyTwins (Ours) & \textbf{16.01} & \textbf{12.41} & \textbf{17.27} &  \textbf{10.03}\\ \midrule
         & \multicolumn{2}{c|}{FID($\downarrow$)}  & \multicolumn{2}{c}{FID($\downarrow$)} \\  \midrule Transitional-cGAN~\cite{shahbazi2022collapse} & \multicolumn{2}{c|}{14.60} & \multicolumn{2}{c}{20.53} \\
          \rowcolor{gray!10} \; + NoisyTwins (Ours)& \multicolumn{2}{c|}{\textbf{13.65}} & \multicolumn{2}{c}{\textbf{16.15}} \\
        \bottomrule
    \end{tabular}
    }
    \vspace{-5.0mm}
\end{table}

\vspace{1mm} \noindent \textbf{How much noise and regularization strength is optimal?} In Fig. \ref{nt_fig:CIFAR10LT_Ablation}, we ablate over the noise variance parameter $\sigma$ for CIFAR10-LT. We find that a moderate value of noise strength 0.75 leads to optimal results. For the strength of NoisyTwins loss ($\lambda$), we find that the algorithm performs similarly on values near 0.01 and is robust to it (Fig. \ref{nt_fig:CIFAR10LT_Ablation}).

\vspace{1mm} \noindent   \textbf{Which type of self-supervision to use with noise augmentation?}
The goal of our method is to achieve invariance to Noise Augmentation in the $\mc{W}$ latent space. This can be achieved using either contrastive learning-based methods like SimCLR\cite{chen2020simple} or negative-free method like Barlow Twins\cite{zbontar2021barlow}. Contrastive loss (SimCLR based) produces FID of 26.23 vs 17.74 by NoisyTwins (BarlowTwins based). We find that contrastive baseline improves over the noise augmentation baseline (28.90) however falls significantly below the NoisyTwins, as the former requires a large batch size to be effective which is expensive for GANs. 

 \vspace{1mm}\noindent  \textbf{How does NoisyTwins compare with modern Vision and Language models?} For evaluating the effectiveness of modern vision language-based diffusion models, we test the generation of the iNaturalist 2019 dataset by creating the prompt ``a photo of \texttt{S}" where we replace the class name in place of \texttt{S}. We use the LDM~\cite{rombach2021highresolution} model trained on LAION-400M to perform inference, generating 50 images per class. We obtained an FID of 57.04 in comparison to best FID of 11.46 achieved by NoisyTwins. This clearly demonstrates that for specific use cases like fine-grained generation, GANs are still ahead of general-purpose LDM.

\section{Conclusion}

\vspace{-1mm}
In this work, we analyze the performance of StyleGAN2 models on the real-world long-tailed datasets  including iNaturalist 2019 and ImageNet-LT. We find that existing works lead to either class confusion or mode collapse in the image space. This phenomenon is rooted in collapse and confusion in the latent $\mc{W}$ space of StyleGAN2. Through our analysis, we deduce that this collapse occurs when the latents become invariant to random conditioning vectors $\mb{z}$, and collapse for each class. To mitigate this, we introduce inexpensive noise based augmentation for discrete class embeddings. Further, to ensure class consistency, we couple this augmentation technique with BarlowTwins' objective in the latent $\mc{W}$ space which imparts intra-class diversity to latent $\mb{w}$ vectors. The noise augmentation and regularization comprises our proposed NoisyTwins technique, which improves the performance of StyleGAN2 establishing a new SotA on iNaturalist 2019 and ImageNet-LT. The extension of NoisyTwins for conditioning on more-complex attributes for StyleGANs is a good direction for future work. \\

\label{nt_sec:intro}

\part{Inductive Regularization for Long-Tailed Recognition}
\label{part:inductive_LT}
\chapter{Escaping Saddle Points for Effective Generalization on Class-Imbalanced Data}
\label{chap:SaddleSAM}

\begin{changemargin}{7mm}{7mm} 
Real-world datasets exhibit imbalances of varying types and degrees. Several techniques based on re-weighting and margin adjustment of loss are often used to enhance the performance of neural networks, particularly on minority classes. In this work, we analyze the class-imbalanced learning problem by examining the loss landscape of neural networks trained with re-weighting and margin based techniques. Specifically, we examine the spectral density of Hessian of class-wise loss, through which we observe that the network weights converges to a saddle point in the loss landscapes of minority classes. Following this observation, we also find that optimization methods designed to escape from saddle points can be effectively used to improve generalization on minority classes. We further theoretically and empirically demonstrate that Sharpness-Aware Minimization (SAM), a recent technique that encourages convergence to a flat minima, can be effectively used to escape saddle points for minority classes. Using SAM results in a 6.2\% increase in accuracy on the minority classes over the state-of-the-art Vector Scaling Loss, leading to an overall average increase of 4\% across imbalanced datasets. The code is available at \href{https://github.com/val-iisc/Saddle-LongTail}{https://github.com/val-iisc/Saddle-LongTail}.
\end{changemargin}

\section{Introduction}
In recent years, there has been a lot of progress in visual recognition thanks to the availability of well-curated datasets~\cite{krizhevsky2009learning, russakovsky2015imagenet}, which are artificially balanced in terms of the frequency of samples across classes. However, modern real-world datasets are often imbalanced (\ie long-tailed etc.)~\cite{krishna2017visual, thomee2016yfcc100m, van2017devil} and suffer from various kinds of distribution shifts.  Overparameterized models like deep neural networks usually overfit classes with a high frequency of samples ignoring the minority (tail) ones~\cite{buda2018systematic, van2017devil}. In such  scenarios, when evaluated for metrics that focus on performance on minority data, these models perform poorly. These metrics are an essential and practical criterion for evaluating models in various domains like fairness~\cite{cotter2019training}, medical imaging~\cite{zhang2019medical} etc.
 
 Many approaches designed for improving the generalization performance of models trained on imbalanced data, are based on the re-weighting of loss~\cite{cui2019class}. The relative weights for samples of each class are determined, such that the expected loss closely approximates the testing criterion objective~\cite{cao2019learning}. In recent years, re-weighting techniques such as Deferred Re-Weighting (DRW)~\cite{cao2019learning}, and Vector Scaling (VS) Loss~\cite{kini2021label} have been introduced, which improve over the classical re-weighting method of weighting the loss of each class sample proportionally to the inverse of class frequency. However, even these improved re-weighting techniques lead to overfitting on the samples of tail classes. Also, it has been shown that use of re-weighted loss for training deep networks converges to final solutions similar to the un-weighted loss case, rendering it to be ineffective~\cite{byrd2019effect}.

\begin{figure*}[!t]
  \centering
  \includegraphics[width=1\textwidth]{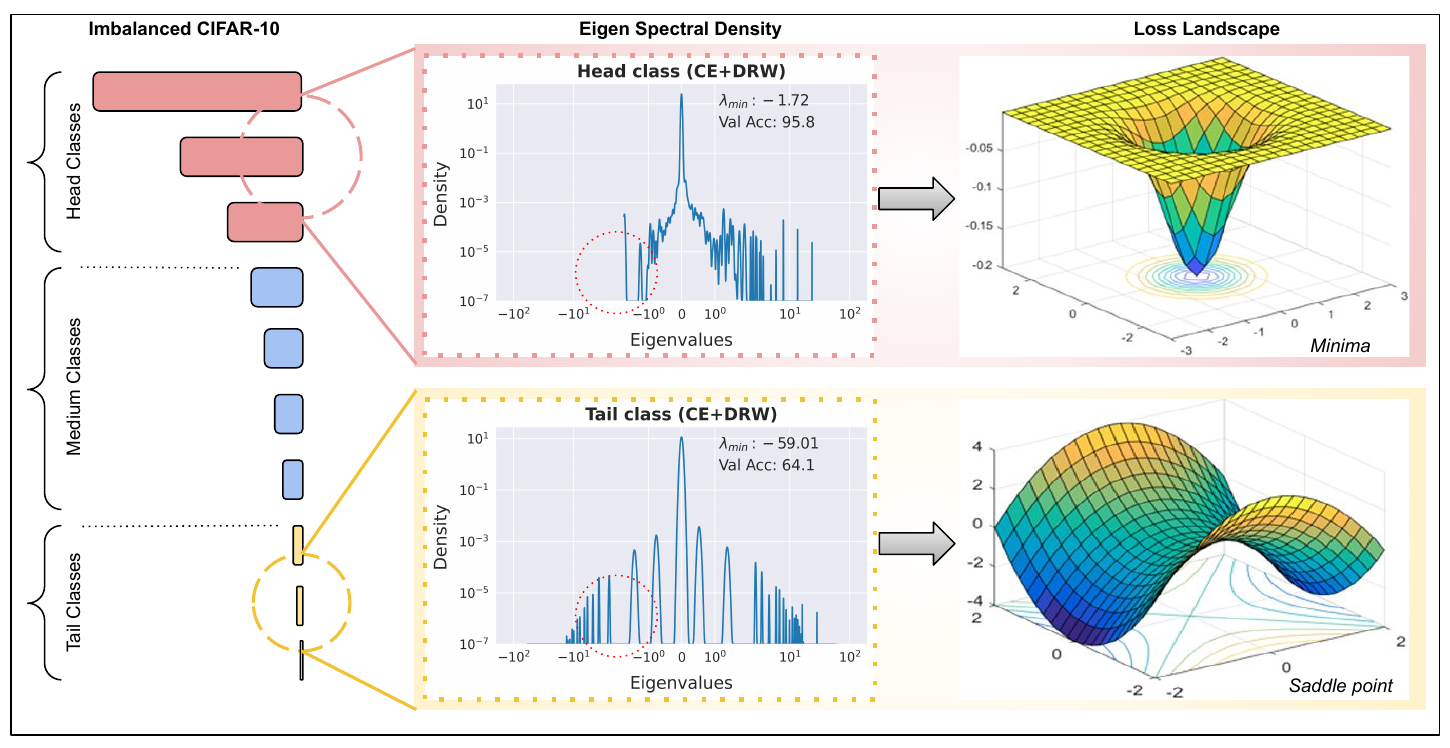}
  \caption{
With class-wise Hessian analysis of loss, we observe that when deep neural networks are trained on class-imbalanced datasets, the final solution for tail classes reach a region of large negative curvature indicating convergence to saddle point (bottom), whereas the head classes converge to a minima (top). The properties of the loss landscape (saddle points or minima) can be observed by analyzing eigen spectral density (centre). \protect\footnotemark{} }
  \label{saddlesam_fig:overview}
\vspace{-1em}
\end{figure*}

This work looks at the loss landscape in weight space around final converged solutions for networks trained with re-weighted loss. We find that the generic Hessian-based analysis of the average loss used in prior works~\cite{pmlr-v97-ghorbani19b, foret2021sharpnessaware}, does not uncover any interesting insights about the sub optimal generalization on tail classes (Sec. \ref{saddlesam_saddle_analysis}). As the frequency of samples is different for each class due to imbalance, we analyze the Hessian of the loss for each class. This proposed way of analysis finds that re-weighting cannot prevent convergence to saddle points in the region of high negative curvature for tail classes, which eventually leads to poor generalization~\cite{dauphin2014identifying}. Whereas for head classes, the solutions converge to a minima with almost no significant presence of negative curvature, similar to networks trained on balanced data. This problem of converging to saddle points has not received much traction in recent times, as the negative eigenvalues disappear when trained on balanced datasets, indicating convergence to local minima~\cite{chaudhari2019entropy, pmlr-v97-ghorbani19b}. However, surprisingly our analysis shows that convergence to saddle points is still a practical problem for neural networks when they are trained on imbalanced (long-tailed) data (Fig. \ref{saddlesam_fig:overview}).

In the community, a lot of techniques optimization methods are designed to be able to escape saddle points efficiently~\cite{ge2015escaping, jin2017escape, Jin2019StochasticGD}, some of which involve adding a component of isotropic noise to gradients. However, these methods have not been able to improve the performance of deep networks in practice, as the implicit noise of SGD in itself mitigates the issue of saddle points when trained on balanced data~\cite{daneshmand2018escaping, Jin2019StochasticGD}. However in the case of imbalanced datasets, we find that the component of SGD along negative curvature (i.e., implicit noise) is insufficient to escape saddle points for minority classes. Thus, learning on imbalanced data can be serve as a practical benchmark for optimization algorithms that can escape saddle points. 

We further demonstrate that Sharpness-Aware Minimization (SAM)~\cite{foret2021sharpnessaware} a recent optimization technique, with re-weighting can effectively enhance the gradient component along the negative curvature, allowing effective escape from saddle points which leads to improved generalization performance. We find that SAM can significantly improve the performance across various re-weighting and margin enhancing methods designed for long-tailed and class-imbalanced learning. The significant improvements are also observed on large-scale datasets of ImageNet-LT and iNaturalist 2018, demonstrating our resutls' applicability at scale. 
We summarize our contributions below:

\begin{itemize}
    \item We propose class-wise Hessian analysis of loss which reveals convergence to saddle points in the loss landscape for minority classes. We find that even loss re-weighting solutions converge to saddle point, leading to sub-optimal generalization on the minority classes. 
    \item We theoretically demostrate that SAM with re-weighting and high regularization factor significantly enhances the component of stochastic gradient along the direction of negative curvature , that results in effective escape from saddle points.
    \item We find that SAM can successfully enhance the performance of even state-of-the-art techniques for learning on imbalanced datasets which have a re-weighting component (\eg VS Loss and LDAM) across various datasets and degrees of imbalance.
\end{itemize}

\footnotetext{Figures for the minima and saddle point are from \protect\citep{arora2020theory} and used for illustration purposes only.}
\section{Related Work \& Background}
In this work, we use $g(x)$ to denote the output of a model,  $\nabla g(x)$ to denote the gradients with respect to parameters, $x$ and $y$ denote the data and labels, respectively. 
We review the re-weighting methods used for training on imbalanced data with distribution shifts, followed by optimization techniques related to our work.

\subsection{Long-Tailed Learning}
Re-sampling \cite{buda2018systematic} and Re-weighting \cite{5128907} are the most commonly used methods to train on class-imbalanced datasets. Oversampling the minority classes \cite{chawla2002smote} and undersampling the majority classes \cite{buda2018systematic} are two approaches to re-sampling. Oversampling leads to overfitting on the tail classes, and undersampling discards a large amount of data, which inevitably results in poor generalization. 
~\citet{Kang2020Decoupling} proposed to decouple representation learning and classifier training to improve performance with the same. Mixup Shifted Label-Aware Smoothing model (MiSLAS) \cite{zhong2021improving} aims to improve the calibration of models trained on long-tailed datasets by mixup and label-aware smoothing and thereby improve performance. RIDE \cite{wang2021longtailed} and TADE \cite{zhang2021test} are ensemble-based methods that achieve state-of-the-art on the long-tailed visual recognition.  \citet{Samuel_2021_ICCV} introduces a new loss, DRO-LT, based on distributionally robust optimization for learning balanced feature representations. We explore the problem of training class-imbalanced datasets through the lens of optimization and loss landscape. We will now describe some representative recent effective methods in detail, which we will use as baselines. Additional discussion on long-tailed learning methods is present in App. \textcolor{red}{H}.

\textbf{LDAM \cite{cao2019learning}}: LDAM introduces optimal margins for each class based on reducing the error through a generalization bound. It results in the following loss function where $\Delta_j$ is the margin for each class:
\begin{equation}
\label{saddlesam_eq:ldam_loss}
    \mathcal{L}_{\text{LDAM}}(y;g(x)) = -\log \frac{e^{g(x)_y - \Delta_y}}{e^{g(x)_y - \Delta_y} + \sum_{j \neq y} e^{g(x)_j}} \; \; \text{where} \; \; \Delta_j = \frac{C}{n_j^{1/4}} \; \text{for} \; j \in \{1, \dots, k\}.
\end{equation}
The core idea of LDAM is to regularize the classes with low frequency (low \ie $n_j$) more, in comparison to the head classes with high frequency.

\textbf{DRW \cite{cao2019learning}}: Deferred Re- Weighting refers to training the model with average loss till certain epochs (K), then introducing weight $w_j$ proportional to  $1/n_j$ to loss term specific to each class $j$ at a later stage. This way of re-weighting has been shown to be effective for improving generalization performance when combined with various losses such as Cross Entropy (CE), LDAM etc. We will be using CE+DRW method as a representative re-weighting method for our analysis. We define CE+DRW loss below for completeness:
\begin{equation}
    \mathcal{L}_{\text{CE}}(y;g(x)) = - w_{y}\log({e^{g(x)_y}}/{\sum_{j = 1}^{k}   e^{g(x)_y})} \; \text{where} \; w_{j} = \frac{1}{1 + (n_j - 1)\mathbbm{1}_{epoch \geq K} }.
\end{equation}

\textbf{VS\cite{kini2021label}}: Vector Scaling loss is a recently proposed loss function which unifies the idea of multiplicative shift (CDT shift \cite{ye2020identifying}), additive shift (i.e Logit Adjustment~\cite{menon2020long}) and loss re-weighting. The final loss has the following form:
\begin{equation}
\label{saddlesam_eq:vs_loss}
    \mathcal{L}_{\text{VS}}(y;g(x)) = -w_{y}\log({e^{\gamma_y g(x)_y + \Delta_y}}/{\sum_{j = 1}^{k}e^{\gamma_j g(x)_j + \Delta_j}}).
\end{equation}
Here the $\gamma_j$ and $\Delta_j$ are the multiplicative and additive logit hyperparameters, respectively.     
\subsection{Loss Landscape}
\textbf{Saddle Points}: Saddle points are regions in loss landscape that usually depict a plateau region with some negative curvature. In the non-convex setting, it has been shown that there is an existence of an exponential number of saddle points in loss landscape and convergence to these points demonstrate poor generalization~\cite{dauphin2014identifying}. There has been a lot of effort in developing methods for effectively escaping saddle points which involve the addition of noise (e.g., Perturbed Gradient Descent (PGD)~\cite{ge2015escaping, jin2017escape, Jin2019StochasticGD}). However, these algorithms have not received much attention in the deep learning community as it has been shown that the implicit noise in SGD can escape saddles easily and converge to local minima~\cite{daneshmand2018escaping}. Also, it has been empirically shown that negative eigenvalues from the Hessian spectrum disappear after a few steps of training, indicating escape from saddle points when neural networks are trained on balanced datasets~\cite{alain2019negative, chaudhari2019entropy, sagun2017empirical}. However, contrary to this, we demonstrate that convergence to saddle points is prevalent in minority class loss landscapes and is a practical problem that can serve as a practical benchmark for the development of algorithms that escape saddle points.

\textbf{Flat Minima based Optimization methods}: Empirically, it has been shown that converging to a flat minima in loss landscape for a deep network leads to improved generalization in comparison to sharp minima ~\cite{hochreiter1997flat, keskar2016large}. Recent works have tried to exploit this connection between the geometry of the loss landscape and generalization to achieve lower generalization error. Sharpness-Aware Minimization (SAM) \cite{foret2021sharpnessaware} is one such algorithm that aims to simultaneously minimize the loss value and sharpness of the loss surface. SAM has shown impressive generalization abilities across various tasks including Natural Language processing \cite{bahri2021sharpness}, meta-learning \cite{abbas2022sharp} and domain adaptation \cite{rangwani2022closer}.
Low-Pass Filtering SGD (LPF-SGD) \cite{bisla2022low} is another recently proposed optimization algorithm that aims to recover flat minima from the optimization landscape. LPF-SGD convolves the loss function with a Gaussian Kernel with variance proportional to the norm of the parameters of each filter in the network. In this work, we aim to explore the effectiveness of such algorithms for the task of escaping saddle points, which is a new direction for these algorithms.

\begin{figure}[t]
\begin{subfigure}{.30\textwidth}
  \centering
  \includegraphics[width=1.0\linewidth]{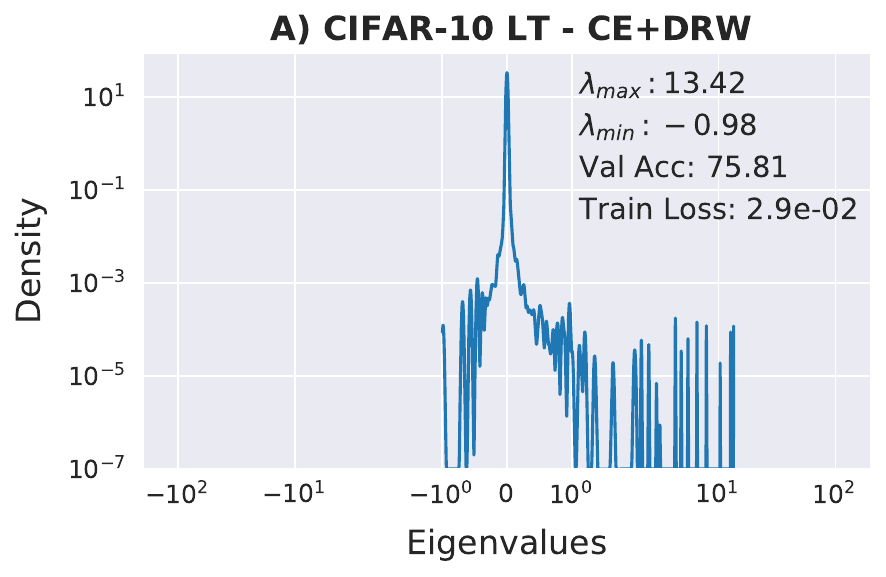}  

  \label{saddlesam_fig:sub-first2}
\end{subfigure}
\begin{subfigure}{.30\textwidth}
  \centering
  \includegraphics[width=1.0\linewidth]{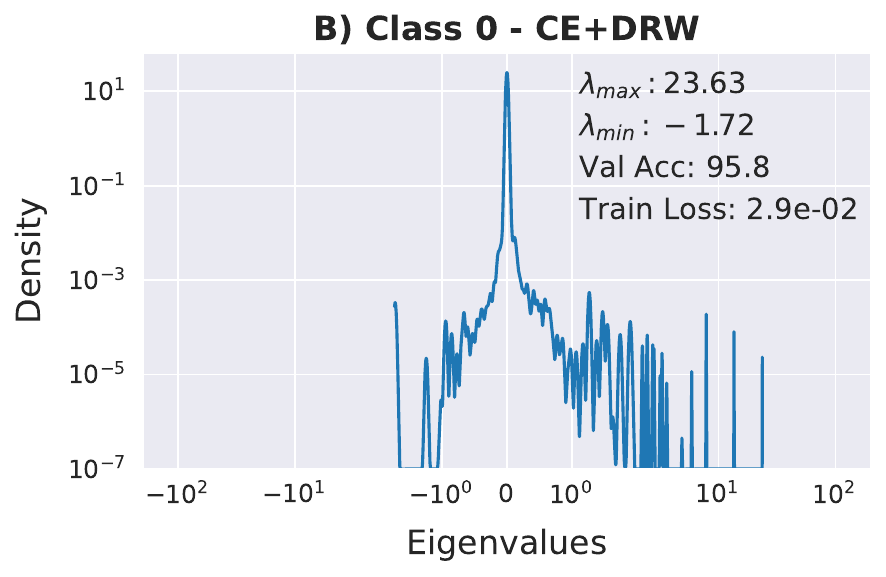}  

  \label{saddlesam_fig:sub-second1}
\end{subfigure}
\begin{subfigure}{.30\textwidth}
  \centering
  \includegraphics[width=1.0\linewidth]{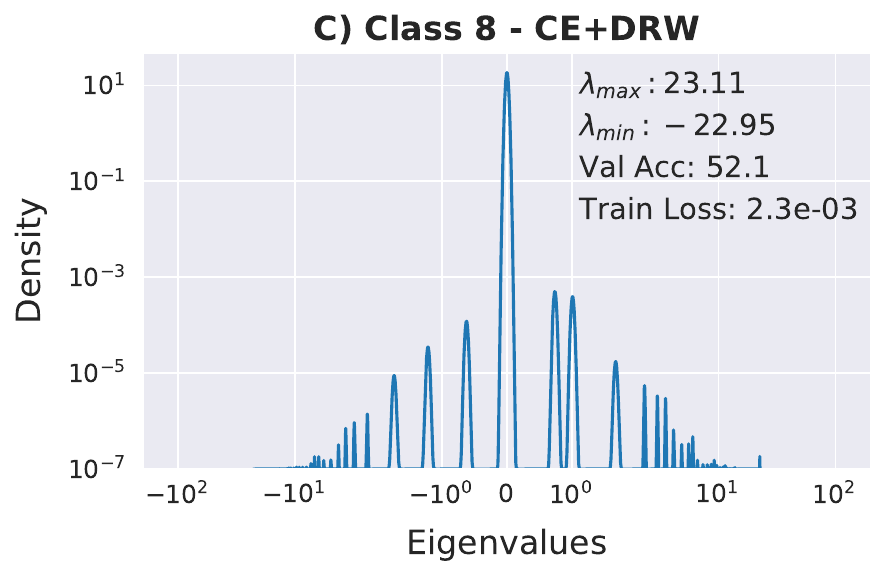}  

  \label{saddlesam_fig:sub-second2}
\end{subfigure}
\newline

\begin{subfigure}{.30\textwidth}
  \centering
  \includegraphics[width=1.0\linewidth]{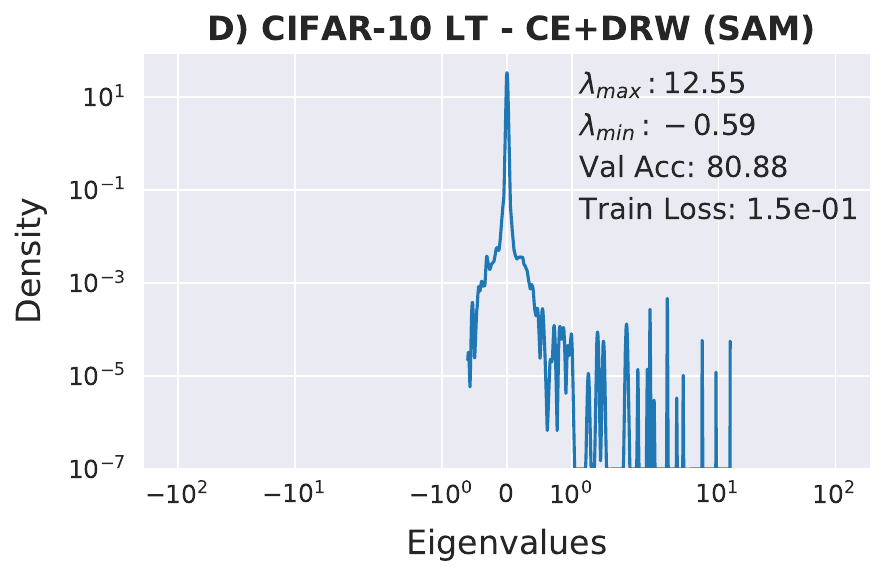}  

  \label{saddlesam_fig:sub-third1}
\end{subfigure}
\begin{subfigure}{.30\textwidth}
  \centering
  \includegraphics[width=1.0\linewidth]{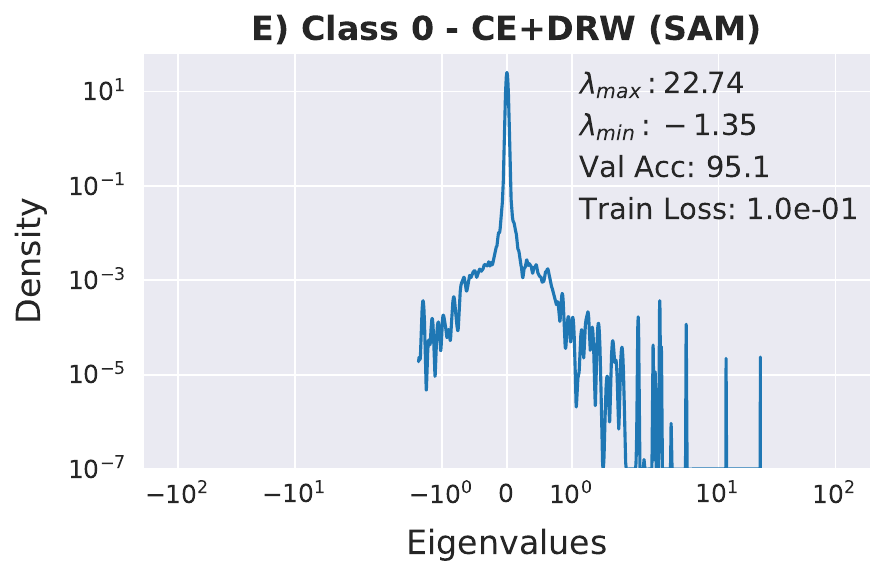}  

  \label{saddlesam_fig:sub-fourth3}
\end{subfigure}
\begin{subfigure}{.30\textwidth}
  \centering
  \includegraphics[width=1.0\linewidth]{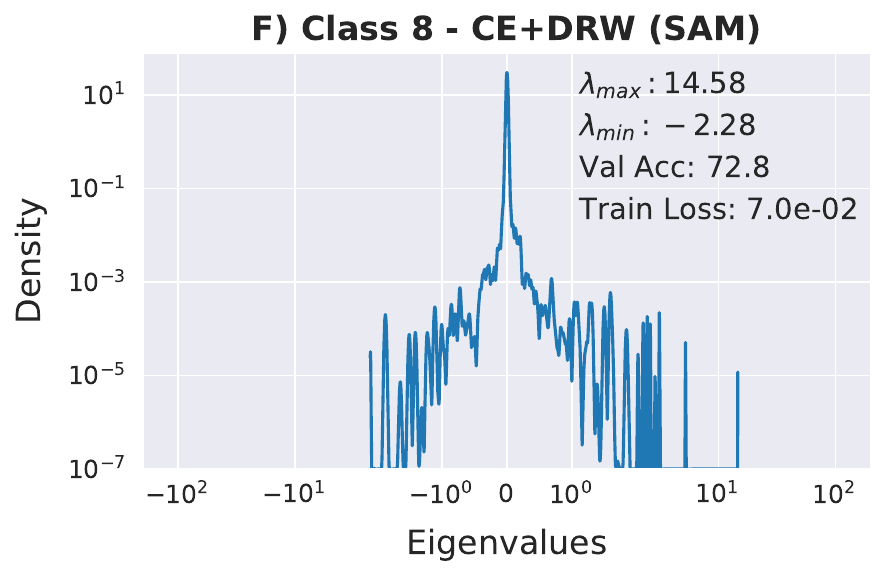}  

  \label{saddlesam_fig:sub-second3}
\end{subfigure}
\caption{Eigen Spectral Density (Class-wise) on the head class (Class 0) and tail class (Class 8) with SGD and SAM. It can be observed that with the head classes, the validation accuracy with SGD (B) and SAM (E) are similar and the density of negative eigenvalues is not significant. On the tail class, SAM (F) escapes the saddle points (large density of negative eigenvalues) in SGD (C), leading to 20\% increase in validation accuracy. A and D show the overall spectral density calculated across all samples in the dataset. Overall spectral density does not indicate the presence of saddles.}
\label{saddlesam_fig:head-tail-esd}
\vspace{-1em}
\end{figure}

\section{Convergence to Saddle Points in Tail Class Loss Landscape}
\label{saddlesam_saddle_analysis}
This section analyzes the dynamics of the loss landscape of neural networks trained on imbalanced datasets. We use the Cross Entropy (CE) loss $\hat{\mathcal{L}}_{\mathrm{CE}}$ to denote the average cross entropy loss for each class. For fine-grained analysis, we focus on average loss on each class $\hat{\mathcal{L}}_{\mathrm{CE}}({y})$. We visualize the loss landscape of the head and tail classes through the computation of Hessian Eigenvalue Density \cite{pmlr-v97-ghorbani19b}. The Hessian of the train loss for each class $H = \nabla^2_{w}\hat{\mathcal{L}}_{\mathrm{CE}}({y})$ contains important properties about the curvature of the classwise loss landscape. The Hessian Eigenvalue Density provides all suitable information regarding the eigenvalues of $H$. In this work, we focus on $\lambda_{max}$(max eigenvalue) and $\lambda_{min}$(min eigenvalue), which depict the extent of positive and negative curvature present. We use the Lanczos algorithm as introduced in \citet{pmlr-v97-ghorbani19b} to compute the Hessian Eigenvalue Density (spectral density) tractably. We further calculate the validation accuracy of a particular class $y$ and its eigen spectral density for analysis. We provide more details for these experiments in the App. \textcolor{red}{D}.

\textbf{Does the proposed class-wise analysis of loss landscape offer any additional insights?} In prior works~\cite{pmlr-v97-ghorbani19b,gilmer2021loss, li2020hessian}, the Hessian of the average loss is used to characterize the nature of the converged point in the loss landscape. However, we find that when particularly trained on imbalanced datasets like CIFAR-10 LT, the eigen spectral density of the Hessian of average loss (Fig. \ref{saddlesam_fig:head-tail-esd}\textcolor{red}{A}) does not differ from that of head class loss (Fig. \ref{saddlesam_fig:head-tail-esd}\textcolor{red}{B}), indicating convergence to a local minima. However, explicitly analyzing the Hessian for the tail class loss (Fig. \ref{saddlesam_fig:head-tail-esd}\textcolor{red}{C}) gives the correct indication of the presence of negative eigenvalues (i.e., curvature), which is in contrast to average loss. Hence, our proposed class-wise analysis of Hessian is essential for characterizing the nature of the converged solution when the training data is imbalanced.

\textbf{What happens when you train a neural network with CE-DRW method on CIFAR-10 LT?}
Fig. \ref{saddlesam_fig:head-tail-esd} shows the spectral density on samples from the head class (Class 0 with 5000 samples) and tail class (Class 8 with 83 samples) at the checkpoint with the best validation accuracy. The spectral density of the head class contains few negative eigenvalues. Most of the eigenvalues are centered around zero, as also observed when training on a balanced dataset \cite{pmlr-v97-ghorbani19b}. On the other hand, for the tail class, there exists a large number of both negative and positive eigenvalues, indicating convergence to a saddle point. We find that at this point, the $\hat{\mathcal{L}}_{CE}(y)$ is low along with the norm of gradient, which indicates a stationary saddle point. We also observe that the spectral density of the tail class contains many outlier eigenvalues, and $\lambda_{max}$ is much larger compared to the head class indicating sharp curvature. These evidences show that \emph{the tail class solution converges to a saddle point instead of a local minimum}. \citet{merkulov2019empirical} indicated the existence of stationary points with low error but poor generalization in the loss landscape of neural networks. Also, the existence of saddle points being associated with poor generalization has been observed for small networks~\cite{dauphin2014identifying}. However, in this work, we show that convergence to saddle points can specifically occur in the loss landscape of tail classes even for the popular ResNet \cite{He_2016_CVPR} family of networks, which is an important and novel observation to the best of our knowledge.

\begin{figure}
\begin{subfigure}{.3\textwidth}
  \centering
  \includegraphics[width=1\textwidth]{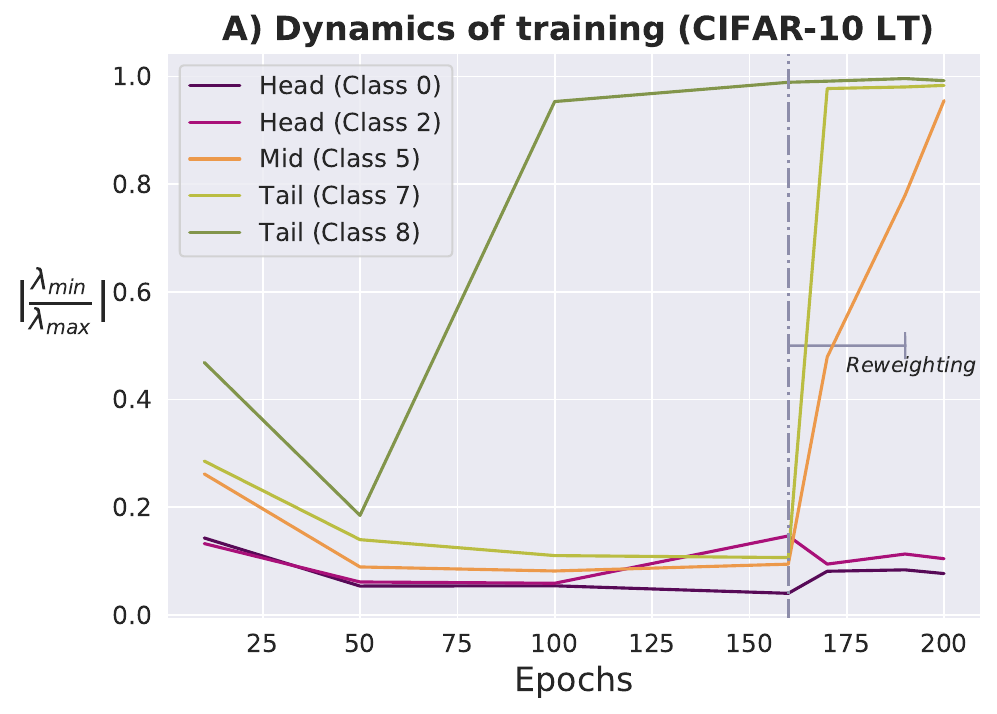}
   \label{saddlesam_fig:conv_dynamics}
\end{subfigure}
\begin{subfigure}{.3\textwidth}
  \centering
  \includegraphics[width=1\textwidth]{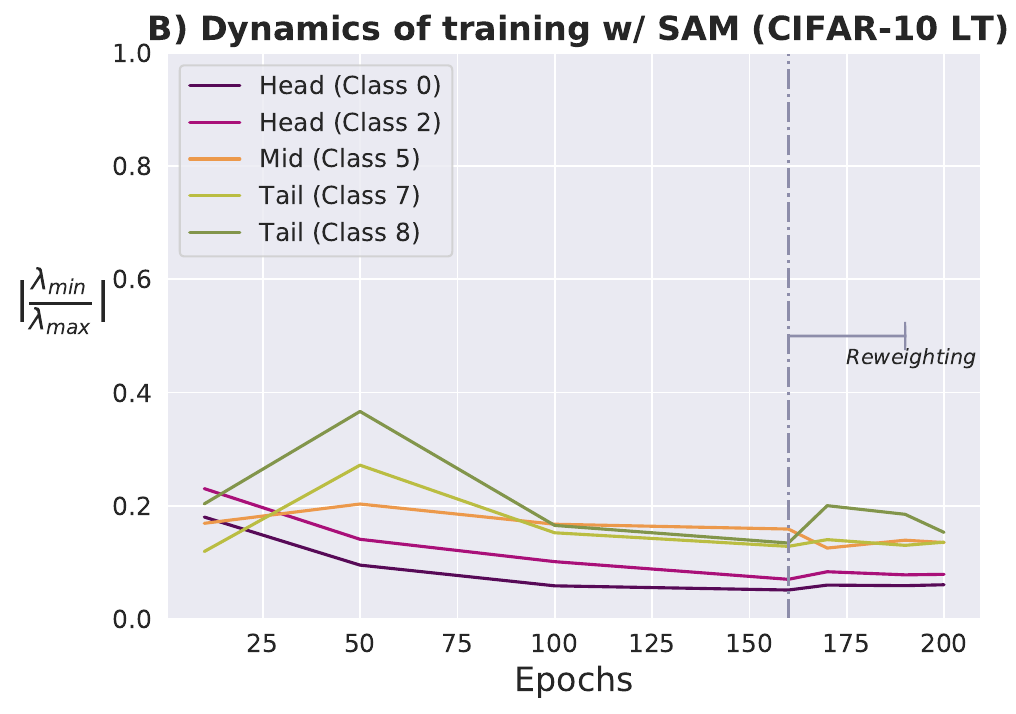}  
  \label{saddlesam_fig:app_dynamics}
\end{subfigure}
\begin{subfigure}{.3\textwidth}
  \centering
  \includegraphics[width=1\textwidth]{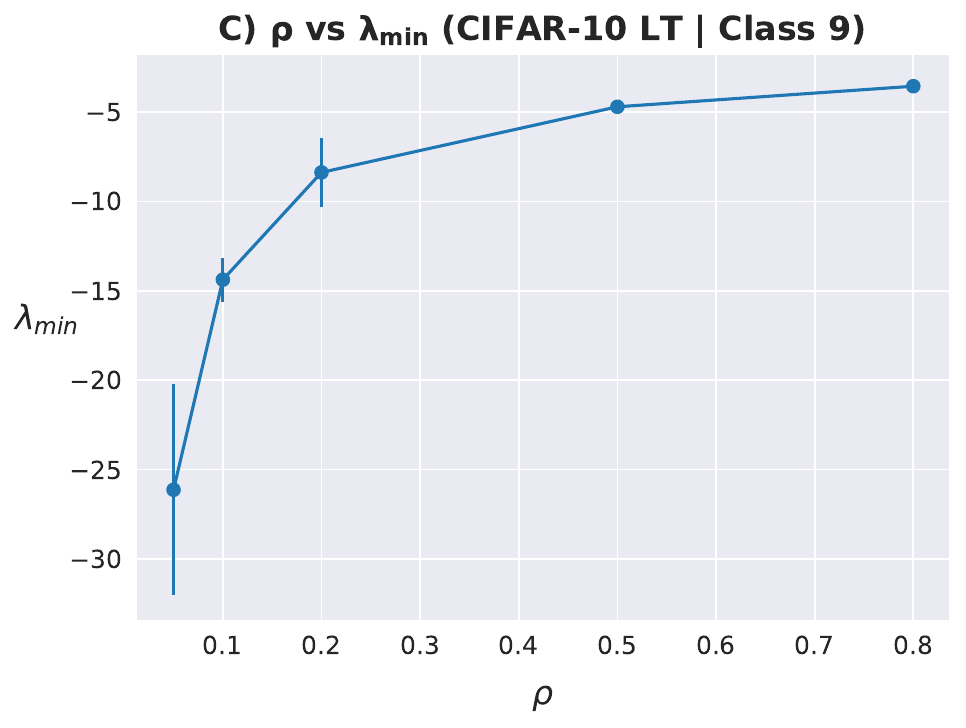}  
  \label{saddlesam_fig:rhovslam}
\end{subfigure}

\caption{A) In CE+DRW, the tail class loss landscapes show significant non-convexity as indicated by the large value of $|{\lambda_{min}}/{\lambda_{max}}|$ whereas head classes (0,2) converge to convex landscapes.
B) When CE+DRW is trained with SAM, it avoids convergence to non-convex regions throughout the training, as indicated by the low value of $|{\lambda_{min}}/{\lambda_{max}}|$.
C) With high $\rho$, the $\lambda_{\min}$ increases, and the model converges to a point with low negative curvature (approx. minima).}
\label{saddlesam_fig:dynamics}
\end{figure}

\textbf{Dynamics of training on Long-Tailed Datasets}: We analyze the $|{\lambda_{min}}/{\lambda_{max}}|$ for the head, mid and tail classes at various epochs (10, 50, 160, 170, 190, 200) across training to understand the dynamics of optimization with CE+DRW on long-tailed data (Fig. \ref{saddlesam_fig:dynamics}\textcolor{red}{A}). $|{\lambda_{min}}/{\lambda_{max}}|$ is a measure of non-convexity of the loss landscape \cite{li2018visualizing}, where a high value of $|{\lambda_{min}}/{\lambda_{max}}|$ conveys non-convexity indicating convergence to points with significant negative curvature. The network converges to non-convex regions with negative curvature for tail classes, showing convergence to the saddle point. Also, we find that for the certain tail (Class 7, 8) and mid classes (Class 5), the network starts converging towards regions with negative curvature after applying loss re-weighting (DRW at 160th epoch). This indicates that DRW leads to convergence to a saddle point rather than preventing it.

\section{Escaping Saddle Points for Improved Generalization}
\label{saddlesam_sec:sam_analysis}
In this section, we analyze the Sharpness-Aware Minimization technique for escaping from saddle points in tail class loss landscape. In existing works~\cite{andriushchenko2022understanding, liu2022towards, zhuang2022surrogate}, the effectiveness of SAM in escaping saddle points has not been explored to the best of our knowledge.

\noindent \textbf{Sharpness-Aware Minimization (SAM)}: 
Sharpness-Aware Minimization is a recent technique which aims to flatten the loss landscape by first finding a sharp maximal point $\epsilon$ in the neighborhood of current weight $w$. It then minimizes the loss at the sharp point ($w + \epsilon$):    
\begin{equation}
    \underset{w}{\min} \;\underset{||\epsilon|| \leq \rho}{\max}\; f(w + \epsilon).
\end{equation}
Here, $f$ is any objective (eg. CE or LDAM loss function) and $\rho$ is the hyperparameter that controls the extent of neighborhood. A high value of $\rho$ leads to convergence to much flat loss landscape. The inner optimization in above objective is first approximated using a first order solution:\vspace{-1em}

\begin{equation}
\begin{split}
            \hat{\epsilon}(w) \approx \underset{||\epsilon|| \leq \rho}{\arg \max} \;  f(w) + \epsilon^T\nabla f(w) \;
            = \rho \nabla f(w) / ||\nabla f(w)||_2.
\end{split}
\end{equation}
After finding $\hat{\epsilon}(w)$, the network weights are updated using the gradient $\nabla f(w)|_{w + \hat{\epsilon}(w)}$. In recent work~\cite{andriushchenko2022understanding}, it has been shown that the normalization of the norm of the gradient for $\hat{\epsilon}(w)$ calculation above leads to oscillation which implies non-convergence theoretically. Also, it has been empirically shown that the unnormalized version of the gradient with adjusted $\rho$ performs better than the normalized version. Hence, we use the approximation i.e. $\hat{\epsilon}(w) = \rho \nabla f(w)$ for our theoretical results. As we will be using the stochastic version of the gradient, we use $z$ as the stochasticity parameter and denote the gradient as $\nabla f_{z}(w)$. With this, we now define the gradient with respect to $w$ that is associated with SAM:
\begin{equation}
    \nabla f_{z}^{\text{SAM}}(w) = \nabla f_{z}(w + \hat{\epsilon}(w)) =  \nabla f_{z}(w + \rho \nabla f_{z}(w)).
\end{equation}
As we are using the same batch for obtaining the gradient to calculate the $\hat{\epsilon}(w)$ and loss, we can use the same $z$ as the argument. We now analyze the component of the SAM gradient in the direction of negative curvature, which is required for escaping saddle points~\cite{daneshmand2018escaping}. 

\subsection{Analysis of SAM for Escaping Saddle Points}
\label{saddlesam_sam_saddle}

\definecolor{darkgreen}{RGB}{30,160,30}

Our analysis is based on the Correlated Negative Curvature (CNC) assumption \citep{daneshmand2018escaping} that states that stochastic gradients have components along the direction of negative curvature, which helps them escape from saddle points. This assumption has been shown to be theoretically valid for the problem of learning half-spaces and also has been empirically verified for a large number of neural networks of different sizes \cite{daneshmand2018escaping}. We now formally state the assumption below:
\begin{assumption}[Correlated Negative Curvature \cite{daneshmand2018escaping}]
\label{saddlesam_ass:cnc}
Let $\mathbf{v_{w}}$ be the minimum eigenvector corresponding to the minimum eigenvalue of the Hessian matrix $\nabla^2 f(w)$. The stochastic gradient $\nabla f_{\bm{z}}(w)$ satisfies the CNC assumption if the second moment of the projection along the direction $\mathbf{v_w}$ is uniformly bounded away from zero, i.e.
\begin{equation}
    \exists \; \gamma \geq 0 \; s.t. \; \forall w : \mathbf{E}[<\mathbf{v_w}, \nabla {f_{\bm{z}}({{w}})>^2}] \geq \gamma.
    \end{equation}

\end{assumption}

\vspace{-0.5em}
It has also been emphasized that the value of $\gamma$ is shown to correlate with the magnitude of $\lambda_{\min}^2$. This shows that with a high negative eigenvalue, there is a large component of gradient along the negative curvature along $\bm{v_{w}}$. This allows the SGD algorithms to escape the saddle points. However, we find that in the case of class imbalanced learning (Fig. \ref{saddlesam_fig:overview}) even stochastic gradients may have an insufficient component in the direction of negative curvature to escape the saddle points. We now show that SAM technique, which aims to reach a flat minima, further amplifies the gradient component along negative curvature and can be effectively used to escape the saddle point. We now formally state our theorem based on the CNC assumption below:
\begin{theorem}
\label{saddlesam_th:sam_rho}
Let $\mathbf{v_{w}}$ be the minimum eigenvector corresponding to the minimum eigenvalue $\lambda_{\min}$ of the Hessian matrix $\nabla^2 f(w)$. The $\nabla f_{\bm{z}}^{\text{SAM}}({{w}})$ satisfies that it's second moment of projection in ${v_{w}}$ is atleast $(1 + \rho\lambda_{min})^2$ times the original (component of $\nabla f_{\bm{z}}({{w}})$): 
\begin{equation}
    \exists \; \gamma \geq 0 \; s.t. \; \forall w : \mathbf{E}[<\mathbf{v_w}, \nabla f_{\bm{z}}^{\text{SAM}}({{w}})>^2] \geq (1 + \rho \lambda_{min})^2\gamma.
\end{equation}
\end{theorem}

\begin{remark}
The above theorem adds the factor $(1 + \rho\lambda_{\min}) ^ 2$ to increase the component in direction of negative curvature ($\gamma$) when $\lambda_{\min} \leq \frac{-2}{\rho}$. Due to this increase, the model will be able to escape from directions with high negative curvature, leading to an increased $\lambda_{\min}$. Also, as the factor $ \frac{-2}{\rho}$ is inversely proportional to $\rho$, the high value of $\rho$ aids in effectively increasing the minimum negative eigenvalue. To empirically verify this, we evaluate the Hessian spectrum for the CIFAR-10 LT dataset using CE-DRW method for different values of $\rho$ (Fig. \ref{saddlesam_fig:dynamics}\textcolor{red}{C}). We find that, as expected from the theorem, in practice, the high values of $\rho$ lead to less negative values of $\lambda_{\min}$. This indicates escaping the saddle points effectively, hence avoiding convergence to regions having negative curvature in loss landscape. The proof of the above theorem and additional details is provided in Appendix \textcolor{red}{B}.

\vspace{1mm} \noindent    We also want to convey that theoretically, techniques like Perturbed Gradient Descent (PGD), and LPF-SGD (Low-Pass Filter SGD), which add Gaussian noise into gradient to escape saddle points can also be used for mitigating negative curvature. Also it has been found that SGD~\cite{daneshmand2018escaping} can also escape the saddle points and converges to solutions with a flat loss landscape. Also, theoretically according to Theorem 2 in \citet{daneshmand2018escaping} the SGD algorithm convergence to a second-order stationary point depends on the $\gamma$ as $\mathcal{O}(\gamma^{-4})$ under some assumptions on $f$. 
As we find that as SAM with high $\rho$ enhances the component of SGD in direction of negative curvature ($\gamma$) by $(1 + \rho \lambda_{\min})^2$, it is reasonable to expect that SAM is able to escape saddle points effectively and converge to solutions with significant less negative curvature quickly implying better generalization. We provide empirical evidence for this in Fig. \ref{saddlesam_fig:dynamics}\textcolor{red}{B} and Sec. \ref{saddlesam_subsec:ablation}.
\end{remark}

\noindent \textbf{What happens when you train a neural network with SAM + DRW?}
With SAM (high $\rho$), the large negative eigenvalues present in the loss landscape of the tail class get suppressed (Fig. \ref{saddlesam_fig:head-tail-esd}\textcolor{red}{F}). In the spectral density for the tail class, it can be seen that $\lambda_{min}$ is much closer to zero for SAM compared to its counterpart with SGD. This aligns with the hypothesis that SAM escapes regions of negative curvature, leading to improved accuracy on the tail classes. However, the spectral density of the head class does not change significantly compared to that of Empirical Risk Minimization (ERM), although the $\lambda_{max}$ is much lower for SAM, indicating a flatter minima for the head class.  

We also analyze the $|{\lambda_{min}}/{\lambda_{max}}|$ across multiple steps of training with SAM (Fig. \ref{saddlesam_fig:dynamics}\textcolor{red}{B}), where
$|{\lambda_{min}}/{\lambda_{max}}|$ is a measure of non-convexity of the loss surface. We observe that SAM does not allow the tail classes to reach a region of high non-convexity. The values of  $|{\lambda_{min}}/{\lambda_{max}}|$ is much lower for SAM compared to SGD (Fig. \ref{saddlesam_fig:dynamics}\textcolor{blue}{A}) throughout training, indicating minimal negative eigenvalues (\ie more convexity) in the loss landscape, especially for the tail and medium classes. This clearly shows that SAM avoids regions of substantial negative curvature in the search of flat minima. Further, we note that once the re-weighting begins, SAM is able to avoid convergence to a saddle point (non-convexity decreases), which is contrary to what we observe with CE+DRW (with SGD). Theorem \ref{saddlesam_th:sam_rho} states that SAM consists of larger component in the direction of negative curvature which allows to reach a solution with minimal negative curvature. Empirically, Fig. \ref{saddlesam_fig:dynamics}\textcolor{blue}{B} also supports the Theorem \ref{saddlesam_th:sam_rho} as we observe that SAM reaches a minima (high convexity) for all the classes.

\begin{table}
  \caption{Results on CIFAR-10 LT and CIFAR-100 LT with $\beta$=100. SAM with re-weighting is able to avoid the regions of negative curvature, leading to major gain in performance on the mid and tail classes with CE, LDAM and VS.}
  \label{saddlesam_tab:cifar_lt}
  \centering
  \begin{adjustbox}{max width=\linewidth}
  \begin{tabular}{l|llll | llll}
    \toprule
    \multicolumn{1}{c}{}  & \multicolumn{4}{c}{CIFAR-10 LT}& \multicolumn{4}{c}{CIFAR-100 LT}\\
    
    \cmidrule(r){1-9}
     &Acc & Head & Mid & Tail & Acc & Head  & Mid & Tail  \\
    \midrule
    CE & 71.7 $_{\pm 0.1}$  & 90.8$_{\pm 3.6}$  & 71.9$_{\pm 0.4}$  & 52.3$_{\pm 3.7}$ & 38.5$_{\pm 0.5}$ & 64.5$_{\pm 0.7}$ & 36.8$_{\pm 1.0}$ & 8.2$_{\pm 1.0}$\\
    \rowcolor{mygray} CE + SAM & 73.1$_{\pm 0.3}$  & 93.3$_{\pm 0.2}$ & 74.1$_{\pm 0.6}$ & 51.7$_{\pm 1.0}$ & 39.6$_{\pm 0.6}$ & 66.5$_{\pm 0.7}$ & 38.1$_{\pm 1.1}$ & 8.0$_{\pm 0.6}$  \Bstrut{}\\
\hline
    CE + DRW \cite{cao2019learning} & 75.5$_{\pm 0.2}$ & 91.6$_{\pm 0.4}$  & 74.1$_{\pm 0.4}$  & 61.4$_{\pm 0.9}$ & 41.0$_{\pm 0.6}$ & 61.3$_{\pm 1.3}$ & 41.7$_{\pm 0.5}$ & 14.7$_{\pm 0.9}$ \Tstrut{}\\
    \rowcolor{mygray}
    \rowcolor{mygray} CE + DRW + SAM & 80.6$_{\pm 0.4}$ & 91.4$_{\pm 0.3}$ & 78.0$_{\pm 0.4}$ & 73.1 $_{\pm 0.9}$ & 44.6$_{\pm 0.4}$ & 61.2$_{\pm 0.8}$ & 47.5$_{\pm 0.6}$ & 20.7$_{\pm 0.6}$  \Bstrut{}\\
    \hline
    LDAM + DRW \cite{cao2019learning} & 77.5$_{\pm 0.5}$ & 91.1$_{\pm 0.8}$ & 75.7$_{\pm 0.7}$ & 66.4$_{\pm 0.2}$ & 42.7$_{\pm 0.3}$ & 61.8$_{\pm 0.6}$ & 42.2$_{\pm 1.5}$ & 19.4$_{\pm 0.9}$ \Tstrut{}\\
    \rowcolor{mygray} LDAM + DRW + SAM & 81.9$_{\pm 0.4}$ & 91.0$_{\pm 0.2}$ & 79.2$_{\pm 0.5}$ & 76.4 $_{\pm 1.1}$ & 45.4$_{\pm 0.1}$ & 64.4$_{\pm 0.3}$ & 46.2$_{\pm 0.2}$ & 20.8 $_{\pm 0.3}$  \Bstrut{}\\
    \hline
    VS \cite{kini2021label} & 78.6$_{\pm 0.3}$ & 90.6$_{\pm 0.4}$ & 75.8$_{\pm 0.5}$  & 70.3$_{\pm 0.5}$ & 41.7$_{\pm 0.5}$ & 54.4$_{\pm 0.2}$ & 41.1$_{\pm 0.6}$ & 26.8$_{\pm 1.0}$  \Tstrut{}\\
    \rowcolor{mygray} VS + SAM & 82.4$_{\pm 0.4}$ & 90.7$_{\pm 0.0}$ & 79.6$_{\pm 0.5}$ & 78.0$_{\pm 01.2}$ & 46.6$_{\pm 0.4}$ & 56.4$_{\pm 0.4}$  & 48.8$_{\pm 0.6}$  & 31.7$_{\pm 0.1}$ \Bstrut{}\\

    \bottomrule
  \end{tabular}

   \end{adjustbox}
\end{table}

\section{Experiments}
\label{saddlesam_sec:experiments}
\subsection{Class-Imbalanced Learning}
 \textbf{Datasets}: We report our results on four long-tailed datasets: CIFAR-10 LT \cite{cao2019learning}, CIFAR-100 LT \cite{cao2019learning}, ImageNet-LT \cite{liu2019large}, and iNaturalist 2018 \cite{van2018inaturalist}. 
\textbf{a) CIFAR-10 LT and CIFAR-100 LT}: The original CIFAR-10 and CIFAR-100 datasets consist of 50,000 training images and 10,000 validation images, spread across 10 and 100 classes, respectively. We use two imbalance versions, i.e., long-tail imbalance and step imbalance, as followed in \citet{cao2019learning}. The imbalance factor, $\beta =\frac{N_{\max }}{N_{\min }}$, denotes the ratio between the number of samples in the most frequent ($N_{\max }$) and least frequent class ($N_{\min}$). For both the imbalanced versions, we analyze the results with $\beta$ = 100.  \textbf{b) ImageNet-LT and iNaturalist 2018}: We use the ImageNet-LT version as proposed by \cite{liu2019large}, which is an class-imbalanced version of the large-scale ImageNet dataset \cite{russakovsky2015imagenet}. It consists of 115.8K images from 1000 classes, with 1280 images in the most frequent class and 5 images in the least. iNaturalist 2018 \cite{van2018inaturalist} is a real-world long-tailed dataset that contains 437.5K images from 8,142 categories. In the case of long-tail imbalance, we segregate the classes of all the datasets into \textit{Head} (Many), \textit{Mid} (Medium), and \textit{Tail} (Few) subcategories, as defined in \cite{zhong2021improving}. For step imbalance experiments on CIFAR datasets, we split the classes into \textit{Head} (Frequent) and \textit{Tail} (Minority), as done in \cite{cao2019learning}. \\
\\
\textbf{Experimental Details}:
We follow the hyperparameters and setup as in \citet{cao2019learning} for CIFAR-10 LT and CIFAR-100 LT datasets. We train a ResNet-32 architecture as the backbone and SGD with a momentum of 0.9 as the base optimizer for 200 epochs. A multi-step learning rate schedule is used, which drops the learning rate by 0.01 and 0.0001 at the 160th and 180th epoch, respectively. For training with SAM, we set a constant $\rho$ value of either 0.5 or 0.8 for most methods. For ImageNet-LT and iNaturalist 2018 datasets, we use the ResNet-50 backbone similar to \cite{zhong2021improving}. An initial learning rate of 0.1 and 0.2 is set for iNaturalist 2018 and ImageNet-LT, respectively, followed by a cosine learning rate schedule. We initialize the $\rho$ value with 0.05 and utilize a step schedule to increase the $\rho$ value during the course of training for SAM experiments. We run every experiment on long-tailed CIFAR datasets with three seeds and report the mean and standard deviation. Additional implementation details are provided in the App. \textcolor{blue}{C}. Algorithm for DRW+SAM is defined in App. \textcolor{blue}{G}.
\\
\\
\textbf{Baselines}: 
\textbf{a) Cross-Entropy (CE)}: CE minimizes the average loss across all samples, and thus, the performance of tail classes is much lower than that of head classes. 
\textbf{b) CE + Deferred Re-Weighting (DRW) \cite{cao2019learning}}: The re-weighting of CE loss inversely by class frequency is done in the later stage of training.
\textbf{c) LDAM + DRW \cite{cao2019learning}}: Label-Distribution-Aware Margin (LDAM) proposes a margin-based loss that encourages larger margins for less-frequent classes.
\textbf{d) Vector Scaling (VS) Loss~\cite{kini2021label}}: VS loss incorporates both additive and multiplicative logit adjustments to modify inter-class margins. 

\begin{table}[t]
	\begin{minipage}{0.6\linewidth}
  \caption{Results on CIFAR-10 and CIFAR-100 with Step Imbalance ($\beta$ = 100). SAM generalizes well across datasets with varied type of imbalance, resulting in substantial gain in tail accuracy in all settings.}
  \label{saddlesam_tab:cifar_lt_step}
  \centering
  \begin{adjustbox}{max width=\linewidth}
  \begin{tabular}{l|lll | lll}
    \toprule
    \multicolumn{1}{c}{}  & \multicolumn{3}{c}{CIFAR-10}& \multicolumn{3}{c}{CIFAR-100}\\
    
    \cmidrule(r){1-7}
     &Acc & Head  & Tail & Acc & Head   & Tail  \\
    \midrule
    CE & 65.1 & 88.6 & 41.7 & 38.6 & 76.3 & 00.9 \\
    \rowcolor{mygray}CE + SAM & 66.1 & 92.9 & 39.4 & 39.3 & 78.6 & 00.0 \Bstrut{} \\
\hline
    CE + DRW \cite{cao2019learning} & 72.2 & 93.1 & 51.2 & 45.8 & 73.9 & 17.8  \Tstrut{}\\
    \rowcolor{mygray}CE + DRW + SAM & 79.3 & 92.7 & 65.8 & 48.3 & 73.1  & 23.4   \Bstrut{}\\
\hline
    LDAM + DRW \cite{cao2019learning} & 77.6 & 89.2 & 66.0 & 45.3 & 70.3 & 20.4 \Tstrut{}\\
    \rowcolor{mygray}LDAM + DRW + SAM & 81.0 & 90.5 & 71.5 & 49.2 & 74.0 & 24.4 \Bstrut{}\\
    \hline
    VS \cite{kini2021label} & 77.0 & 91.7 & 62.3 & 46.5 & 69.0 & 24.1 \Tstrut{}\\
    \rowcolor{mygray}VS + SAM & 82.0 & 91.7 & 72.3 & 48.3 & 70.4 & 26.2 \Bstrut{}\\

    \bottomrule
  \end{tabular}
   \end{adjustbox}
	\end{minipage}\hfill
	\begin{minipage}{0.35\linewidth}
      \centering
      \includegraphics[width=1.0\textwidth]{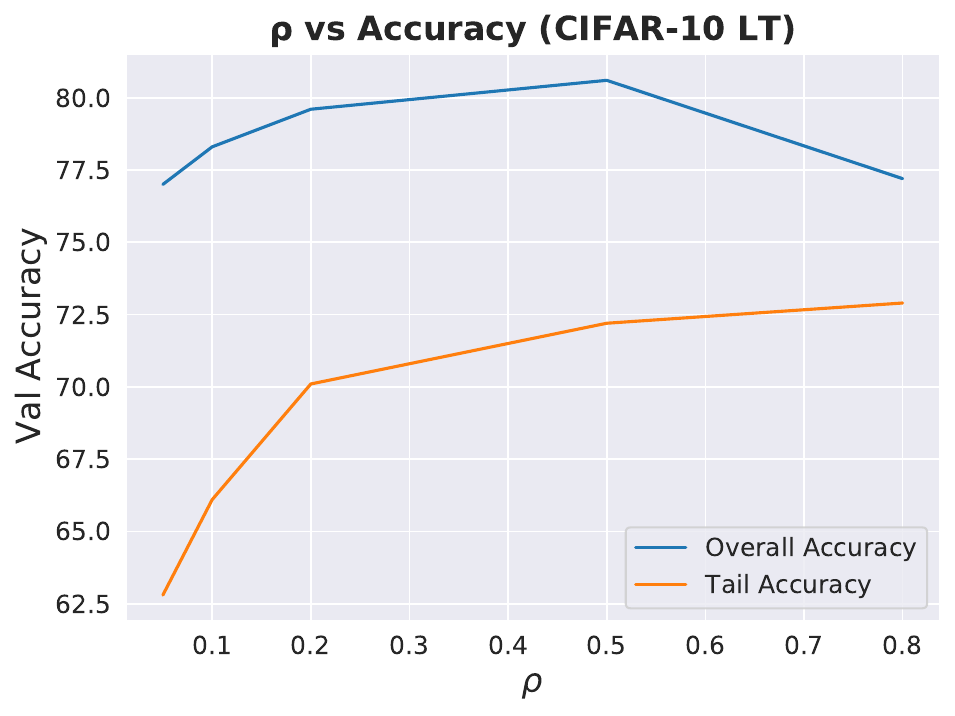}
      \captionof{figure}{Impact of $\rho$ (regularization factor) on Overall Accuracy and Tail Accuracy (CIFAR-10 LT).}
      \label{saddlesam_fig:rhovsacc}

	\end{minipage}
\end{table}

\noindent \textbf{Results}:
Table \ref{saddlesam_tab:cifar_lt} summarizes our results on CIFAR-10 LT and CIFAR-100 LT with $\beta$ of 100. It can be observed that SAM with re-weighting significantly improves the accuracy on mid and tail classes while preserving the accuracy on head classes. SAM improves upon the overall performance of CE+DRW by 5.1\% on CIFAR-10 LT and 3.6\% on CIFAR-100 LT datasets, with the tail class accuracy increasing by 11.7\% and 7.7\% respectively. %
These results empirically show that escaping saddle points with SAM leads to a notable increase in overall accuracy primarily due to the major gain in the accuracy on the tail classes. The addition of SAM to recently proposed long-tail learning methods like LDAM and VS loss leads to a significant increase in performance, which indicates that the role of SAM is orthogonal to the margin-based methods. On the other hand, SAM without re-weighting (CE+SAM) improves accuracy on the head and mid classes rather than the tail class. This can be attributed to the fact that standard ERM minimizes the average loss across all the samples without re-weighting such that the weightage of tail class samples in the overall loss is minimal. This shows that naive application of SAM is ineffective in improving tail class performance, in comparison to proposed combination of re-weighting methods with SAM. We also show improved results with various imbalance factors ($\beta$) in App. \textcolor{blue}{F}.

We also show results with step imbalance ($\beta$ = 100) on CIFAR-10 and CIFAR-100 datasets (Table \ref{saddlesam_tab:cifar_lt_step}). With step imbalance on CIFAR-10, the first five classes have 5000 samples each, and the remaining classes have 50 samples each. The addition of SAM improves the overall performance of CE+DRW on CIFAR-10 by 7.1\%, with the tail class accuracy increasing by 14.6\%. We observe that on most tail classes, the density of negative eigenvalues in the spectral density is much lower with SAM. This indicates that despite multiple classes with few samples, SAM with DRW can avoid the saddle points. SAM systematically improves performance with LDAM and VS loss leading to state-of-the-art performance on both CIFAR-10 and CIFAR-100 in the step imbalance setting.

\textbf{Do these observations scale to large-scale datasets?}
We report the results on ImageNet-LT dataset in Table \ref{saddlesam_tab:imagenet_lt}. We also compare with recent long-tail learning methods:  cRT \cite{Kang2020Decoupling}, MisLAS \cite{zhong2021improving}, DisAlign \cite{zhang2021distribution} and DRO-LT \cite{Samuel_2021_ICCV}. The observations on CIFAR-10 LT and CIFAR-100 LT hold good even on ImageNet-LT. For example, the accuracy on tail classes increases by 6.5\% with the introduction of SAM on CE + DRW, which is similar to the gain observed in CIFAR-100 LT with CE + DRW. We observe that LDAM+DRW+SAM surpasses the performance of two-stage training methods including MisLAS, cRT, LWS, and DisAlign. Compared to these two-stage methods, our method is a single stage method and outperforms these two-stage methods. These observations point out that the problem of saddle points also exists in large datasets and convey that SAM is easily generalizable to large-scale imbalanced datasets without making any significant changes. On iNaturalist 2018 \cite{van2018inaturalist} too, the accuracy on tail classes gets boosted by more than 3\% with SAM (Table \ref{saddlesam_tab:imagenet_lt}). 

\begin{table}
  \caption{Results on iNaturalist 2018 and ImageNet-LT datasets with LDAM+DRW and comparison with other methods. The numbers for methods marked with $\dag$ are taken from \cite{zhong2021improving}.}
  \label{saddlesam_tab:imagenet_lt}
  \centering
  \begin{adjustbox}{max width=\linewidth}
  \begin{tabular}{l|c|llll|llll}
    \toprule

    \multicolumn{2}{c}{Method}  & \multicolumn{4}{c}{iNaturalist 2018} &
    \multicolumn{4}{c}{ImageNet-LT}\\
    \hline
     &Two stage & Acc & Head & Mid & Tail & Acc & Head & Mid & Tail  \Tstrut{}\\
    \midrule
    CE &\Cross& 60.3 & 72.8 & 62.7 & 54.8 & 42.7 & 62.5 & 36.6 & 12.5  \\
    \hline
    cRT \cite{Kang2020Decoupling} $\dag$&\checkmark &68.2 & \underline{73.2} & 68.8 & 66.1 &50.3 & \underline{62.5} & 47.4 & 29.5 \Tstrut{}\\
    LWS \cite{Kang2020Decoupling} $\dag$&\checkmark &69.5 & 71.0 & 69.8 & 68.8 & 51.2 & 61.8 & 48.6 & 33.5\\
    MiSLAS \cite{zhong2021improving} &\checkmark &\textbf{71.6} & \underline{73.2} & \textbf{72.4} & \underline{70.4} & 52.7 &61.7 & 51.3 & \textbf{35.8} \\
    DisAlign \cite{zhang2021distribution} &\checkmark &69.5 & 61.6 & \underline{70.8} & 69.9 & 52.9 & 61.3 & \textbf{52.2} & 31.4\\
    DRO-LT \cite{Samuel_2021_ICCV} &\Cross &69.7 & \textbf{73.9} & 70.6 & 68.9 &\textbf{53.5} & \textbf{64.0} & 49.8 & 33.1\\

\hline
    CE + DRW & \Cross & 63.0 & 59.8 & 64.4 & 62.3 & 44.9 & 57.9 & 42.2 &21.6 \Tstrut{} \\
    \rowcolor{mygray} CE + DRW + SAM & \Cross& 65.3 & 60.5 & 66.2 & 65.5 & 47.1 & 56.6 & 45.8 & 28.1  \\
\hline
    LDAM + DRW & \Cross&67.5 & 63.0 & 68.3 & 67.8 &49.9 & 61.1 & 48.2 & 28.3 \Tstrut{} \\
    \rowcolor{mygray} LDAM + DRW + SAM & \Cross&\underline{70.1} & 64.1 & 70.5 & \textbf{71.2} & \underline{53.1} & 62.0 & \underline{52.1} & \underline{34.8}\\
    \bottomrule
  \end{tabular}

  \end{adjustbox}

\end{table}

\textbf{Comparison with SOTA}: VS loss \cite{kini2021label} is a recently proposed margin-based method that achieves state-of-the-art performance on class-imbalanced datasets with single-stage training without strong augmentations \cite{zhong2021improving}, ensembles \cite{zhang2021test} or self-supervision \cite{yang2020rethinking}. SAM significantly improves upon the performance of VS on both CIFAR-10 LT and CIFAR-100 LT. For the practitioners, we suggest \emph{using high $\rho$ SAM with re-weighting or margin based methods} for effective learning on long-tailed data. We also integrate SAM with more recent IB-Loss \cite{Park_2021_ICCV} and Parametric Contrastive Learning (PaCo) \cite{cui2021parametric} methods and report the results in App. \textcolor{blue}{E}. We find that SAM is also effectively able to improve performance of these recent methods.

\subsection{Ablation Studies}
\label{saddlesam_subsec:ablation}
\textbf{A note on $\rho$ value}:
We observe that as we increase the smoothness parameter ($\rho$) in SAM, the accuracy on the tail classes increases significantly (Fig. \ref{saddlesam_fig:rhovsacc}). The accuracy on tail classes increases from 63\% for $\rho$ = 0.05 to 73\% for $\rho$ = 0.8 on CIFAR-10 LT with CE+DRW. This can be ascribed to the correlation between $\lambda_{min}$ and $\rho$ as discussed in Sec. \ref{saddlesam_sam_saddle}. As the $\rho$ increases, the negative curvature in the tail classes disappears because SAM aims to find a flat minima with a large neighborhood with a low loss value. A very large $\rho$ (0.8) leads to a drop in the head accuracy because it restricts the solution space of the head class, resulting in a drop in the overall accuracy. This also emphasizes that a high $\rho$ is necessary for escaping saddle points and achieving the best results.

\textbf{Other methods to escape saddle points}: In Table \ref{saddlesam_tab:lpf_sgd}, we show that other methods developed to escape saddle points, such as PGD, can be used for improving generalization on tail classes. LPF-SGD, an algorithm promoting convergence to flat landscape, inherently adds Gaussian noise to the network parameters and could be considered similar to PGD. We can see that the addition of PGD and LPF-SGD to CE+DRW leads to a substantial gain in the performance of tail classes on CIFAR-10 LT and CIFAR-100 LT. It can also be observed that CE+DRW+SAM outperforms both PGD and LPF-SGD by 2\% on average. This further highlights that various methods in literature developed to escape saddle points efficiently can be directly used to improve the performance of minority classes when training on class-imbalanced datasets.
\begin{table}
  \caption{Results on CIFAR-10 LT and CIFAR-100 LT with various methods that escape saddle points.}
  \label{saddlesam_tab:lpf_sgd}
  \centering
  \begin{adjustbox}{max width=\linewidth}
  \begin{tabular}{l|llll | llll}
    \toprule
    \multicolumn{1}{c}{}  & \multicolumn{4}{c}{CIFAR-10 LT}& \multicolumn{4}{c}{CIFAR-100 LT}\\
    
    \cmidrule(r){1-9}
     &Acc & Head & Mid & Tail & Acc & Head  & Mid & Tail  \\
    \midrule

    CE + DRW & 75.5&91.6&74.1&61.4&41.0&61.3&41.7&14.7 \Tstrut{}\\
    CE + DRW + PGD \cite{jin2017escape} & 77.2&92.0&75.2&65.0&42.2&63.0&41.6&17.0\\
    CE + DRW + LPF-SGD \cite{bisla2022low} & 78.5&90.8&77.7&67.2&42.9&64.0&43.7&15.8\\
    CE + DRW + SAM & 80.6  & 91.4 & 78.0 & 73.1  & 44.6& 61.2 & 47.5 & 20.7
    \Bstrut{}\\

    \bottomrule
  \end{tabular}

   \end{adjustbox}
\end{table}

\section{Conclusion}
In this work, we show that training on imbalanced datasets can lead to convergence to points with sufficiently large negative curvature in the loss landscape for minority classes. We find that this is quite common when neural networks are trained with loss functions that are re-weighted or modified to enhance the focus on minority classes. Due to the occurrence of saddle points, we observe that the network suffers from poor generalization on minority classes. We propose to use Sharpness-Aware Minimization (SAM) with a high regularization factor $\rho$ as an effective method to escape regions of negative curvature and enhance the generalization performance. We theoretically and empirically demonstrate that SAM with high $\rho$ is able to escape saddle points faster than SGD and converge to better solutions, which is a novel observation to the best of our knowledge. We show that combining SAM with state-of-the-art techniques for learning with imbalanced data leads to significant gains in performance on minority classes. We hope that our work leads to further research in studying the effect of negative curvature in generalization as we show they are a practical issue for class-imbalanced learning using deep neural networks.

\chapter{DeiT-LT: Distillation Strikes Back for Vision Transformer Training on Long-Tailed Datasets}
\label{chap:DeiT-LT}

\begin{changemargin}{7mm}{7mm} 

Vision Transformer (ViT) has emerged as a prominent architecture for various computer vision tasks. In ViT, we divide the input image into patch tokens and process them through a stack of self-attention blocks. However, unlike Convolutional Neural Network (CNN), ViT’s simple architecture has no informative inductive bias (e.g., locality, etc.). Due to this, ViT requires a large amount of data for pre-training. Various data-efficient approaches (DeiT) have been proposed to train ViT on balanced datasets effectively. However, limited literature discusses the use of ViT for datasets with long-tailed imbalances. In this work, we introduce DeiT-LT to tackle the problem of training ViTs from scratch on long-tailed datasets. In DeiT-LT, we introduce an efficient and effective way of distillation from CNN via distillation \texttt{DIST} token by using out-of-distribution images and re-weighting the distillation loss to enhance focus on tail classes. This leads to the learning of local CNN-like features in early ViT blocks, improving generalization for tail classes.
Further, to mitigate overfitting, we propose distilling from a flat CNN teacher, which leads to learning low-rank generalizable features for \texttt{DIST} tokens across all ViT blocks.  With the proposed DeiT-LT scheme, the distillation \texttt{DIST} token becomes an expert on the tail classes, and the classifier \texttt{CLS} token becomes an expert on the head classes. The experts help to effectively learn features corresponding to both the majority and minority classes using a distinct set of tokens within the same ViT architecture. We show the effectiveness of DeiT-LT for training ViT from scratch on datasets ranging from small-scale CIFAR-10 LT to large-scale iNaturalist-2018.
Project Page: \href{https://rangwani-harsh.github.io/DeiT-LT}{https://rangwani-harsh.github.io/DeiT-LT}.

\end{changemargin}

\section{Introduction}
\label{deit-lt_sec:intro}

A plethora of recent works~\cite{li2022nested, cao2019learning, cui2019class, zhou2020BBN, menon2020long} focus on training deep neural networks for recognition on long-tailed datasets, such that networks perform reasonably well across all classes, including the minority classes. Loss manipulation-based techniques~\cite{cao2019learning, cui2019class,kinivs} enhance the network's focus toward learning tail classes by enforcing a large margin or increasing the weight for loss for these classes. As these techniques enhance the focus on the tail classes, they often lead to some performance degradation in the head (majority) classes. To mitigate this, State-of-the-Art (SotA) techniques currently train multiple expert networks~\cite{wang2020long, li2022nested} that specialize in different portions of the data distribution. The predictions from these experts are then aggregated to produce the final output, which improves the performance over individual experts. However, all these efforts have been restricted to Convolutional Neural Networks (CNNs), particularly ResNets~\cite{he2016deep}, with little attention to the architectures such as Transformers~\cite{dosovitskiy2015discriminative, vaswani2017attention}, MLP-Mixers~\cite{tolstikhin2021mlp} etc.

\begin{figure*}[t]
    \centering
    \includegraphics[width=\textwidth]{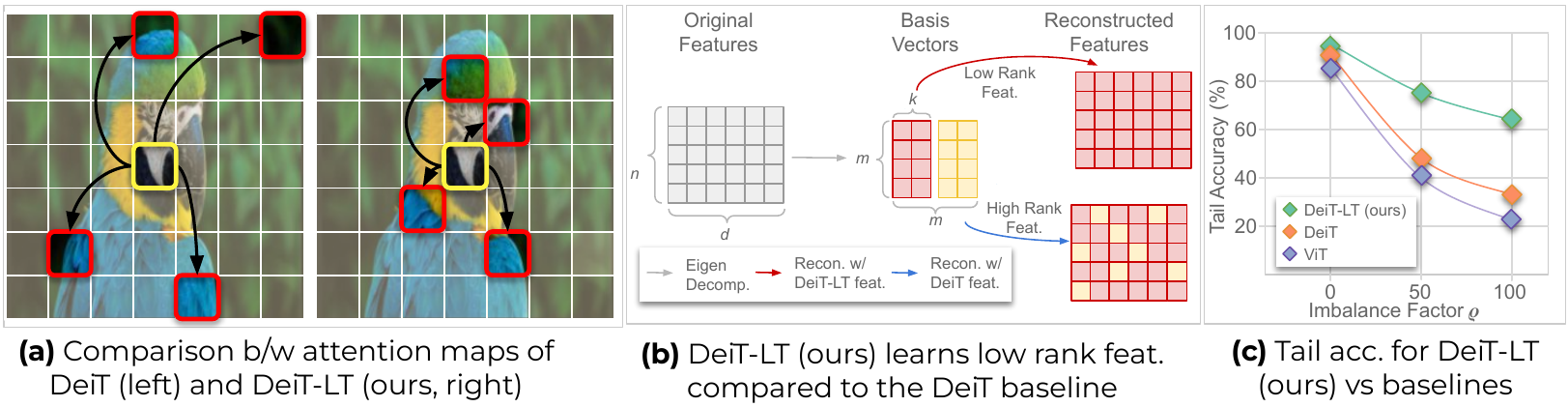}
    \caption{We propose DeiT-LT (Fig.~\ref{deit-lt_fig:overview}, a distillation scheme for Vision Transformer (ViT), tailored towards long-tailed data). In Deit-LT, \textbf{a)} we introduce OOD distillation from CNN, which leads to learning local generalizable features in early blocks. \textbf{b)} we propose to distill from teachers trained via SAM~\cite{foret2020sharpness} which induces low-rank features across blocks in ViT to improve generalization. \textbf{c)} In comparison to other SotA ViT baselines, Deit-LT (ours) demonstrates significantly improved performance for minority classes, with increasing imbalance.}
    \label{deit-lt_fig:teaser-fig}
    \vspace{-1em}
\end{figure*}

Recently, the transformer architecture adapted for computer vision, named as Vision Transformer (ViT)~\cite{dosovitskiyimage}, has gained popularity due to its scalability and impressive performance on various computer vision tasks~\cite{carion2020end, strudel2021segmenter}. One caveat behind its impressive performance is the requirement for pre-training on large datasets~\cite{dosovitskiy2015discriminative}. The data-efficient transformers  (DeiT)~\cite{touvron2021deit} aimed to reduce this requirement for pre-training by distilling information from a pre-trained CNN. Subsequent efforts have further improved the data and compute efficiency ~\cite{touvron2022deit, Touvron2022ThreeTE} of ViTs. However, all these improvements have been primarily based on increasing performance on the balanced ImageNet dataset. We find that these improvements are still insufficient for robust performance on long-tailed datasets (Fig.~\ref{deit-lt_fig:teaser-fig}\textcolor{red}{c}).

In this work, we aim to investigate and improve the \emph{training of Vision Transformers from scratch without the need for large-scale pre-training} on diverse long-tailed datasets, varying in image size and resolution. Recent works show improved performance for ViTs on long-tailed recognition tasks, but they often need expensive pre-training on large-scale datasets~\cite{chen2022reltransformer, long2022retrieval}. The requirement of pre-training is computationally expensive and restricts their application to specialized domains such as medicine, satellite, speech, etc. Furthermore, the large-scale pre-trained datasets often contain biases that might be inadvertently induced with their usage~\cite{agarwal2021evaluating, ousidhoum2021probing, wang2023overwriting}. To mitigate these shortcomings, we introduce \textit{\textbf{Data-efficient Image Transformers for Long-Tailed Data (DeiT-LT)}} - a scheme for training ViTs from scratch on small and large-scale long tailed datasets. DeiT-LT is based on the following important design principles:
\begin{itemize}
    \item DeiT-LT involves distilling knowledge from low-resolution teacher networks using out-of-distribution (OOD) images generated through strong augmentations. Notably, this method proves effective even if the CNN teacher wasn't originally trained on such augmentations. The outcome is the successful induction of CNN-like feature locality in the ViT student network, ultimately enhancing generalization performance, particularly for minority (tail) classes (Fig.~\ref{deit-lt_fig:teaser-fig}\textcolor{red}{a},~\ref{deit-lt_fig:similarity_plot} and Sec.~\ref{deit-lt_sebsec:out-of-dis-dist}).
    
    \item Further, to improve the generality of features, we 
    propose to distill knowledge via flat CNN teachers trained through Sharpness Aware Minimization (SAM)~\cite{foret2020sharpness}. This results in low-rank generalizable features for long-tailed setup across all ViT blocks  (Fig.~\ref{deit-lt_fig:teaser-fig}\textcolor{red}{b} and Sec.~\ref{deit-lt_subsec:low-rank-sam}).
    
    \item In DeiT~\cite{touvron2021deit}, the classification and distillation tokens produce similar predictions. However, in proposed DeiT-LT, we ensure their divergence such that the classification token becomes an expert on the majority classes. Whearas, the distillation token learns local low-rank features, becoming an expert on the minority. Hence, DeiT-LT can focus on both the majority and minority effectively, which is not possible with vanilla DeiT training (Fig.~\ref{deit-lt_fig:attention-vis} and Sec.~\ref{deit-lt_sebsec:out-of-dis-dist}).
\end{itemize} 
We demonstrate the effectiveness of DeiT-LT across diverse small-scale (CIFAR-10 LT, CIFAR-100 LT) as well as large-scale datasets (ImageNet-LT, iNaturalist-2018). We find that DeiT-LT effectively improves over the teacher CNN across all datasets and achieves performances superior to SotA CNN-based methods without requiring any pre-training.

\section{Background}
\vspace{-1mm}
\label{deit-lt_sec:related_works}
\begin{figure*}[!t]
    \centering
    \includegraphics[width=\textwidth]{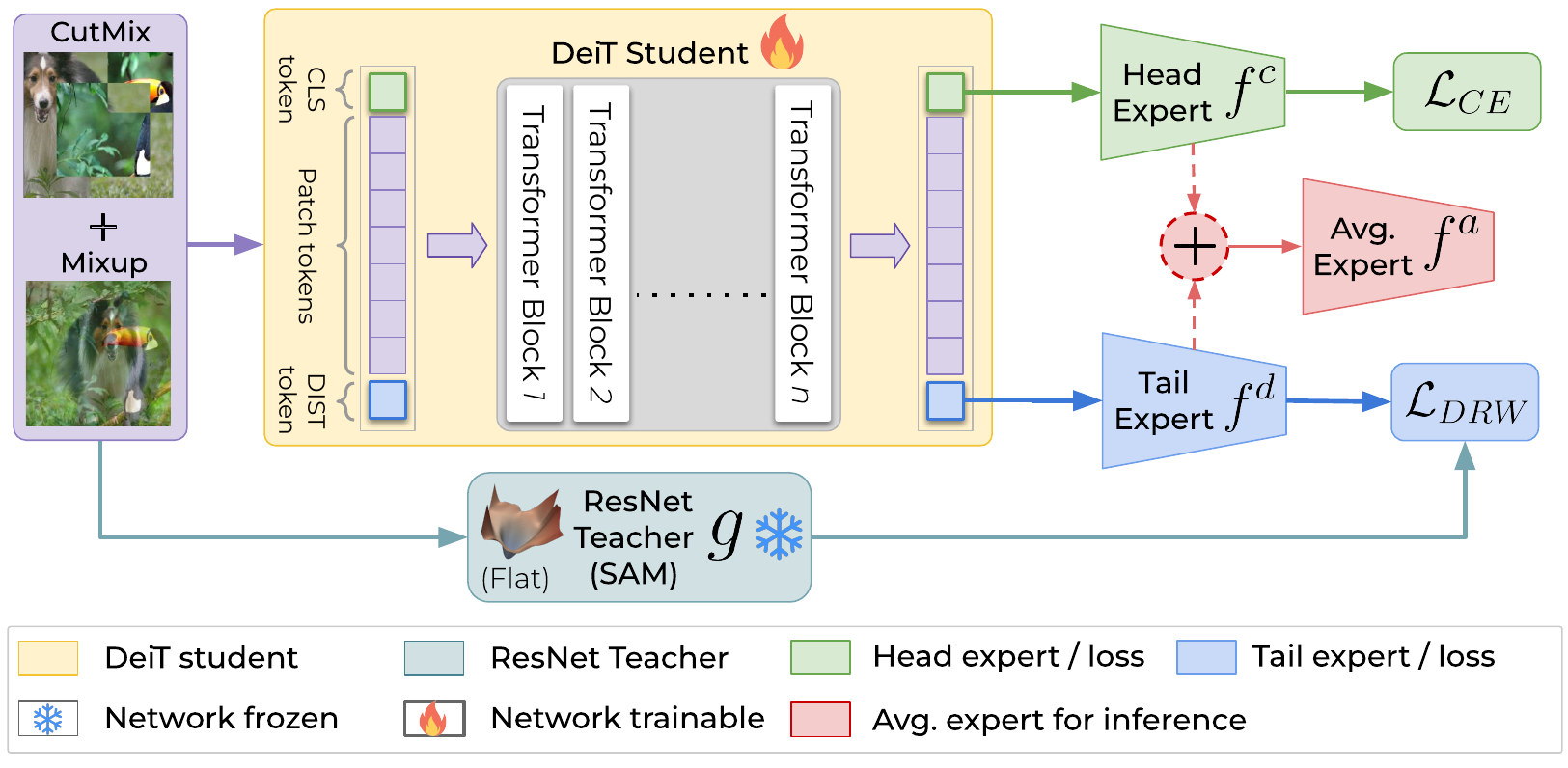}
    \caption{ Overview of DeiT-LT. 
 The Head Expert classifier trains using CE loss against ground truth, whereas the Tail Expert classifier trains using DRW loss against hard-distillation targets from the flat ResNet teacher trained via SAM~\cite{foret2020sharpness}. The distillation is performed using out-of-distribution images created using strong augmentations and Mixup.
 }
    \label{deit-lt_fig:overview}
    \vspace{-1.5 em}
\end{figure*}
\vspace{1mm} \noindent \textbf{Long-Tailed Learning.} With the increased scale of deep learning, large crowd-sourced long-tailed datasets have become common. A plethora of techniques are developed to learn machine learning models using such datasets, where the objective is improved performance, particularly on tail classes. The methods can be broadly divided into three categories: a) loss re-weighting b) decoupled classifier and representations and c) expert-based classifier training. In addition, there are some techniques based on the synthetic generation for long-tailed recognition~\cite{kim2020m2m, rangwani2021class, rangwani2023noisytwins, hrangwani2022gsr}, which are orthogonal to this study. The loss re-weighting-based techniques include margin based techniques like LDAM~\cite{cao2019learning}, and Logit-Adj~\cite{menon2020long}, which enforce a higher margin for tail classes. The other set (eg. CB-Loss~\cite{cui2019class}, VS-Loss~\cite{kinivs} etc.) introduce re-weighting factors in cross entropy loss based on the training set label distribution. The other set of techniques propose to decouple the learning of representations with classifier learning, as it's observed that margin based losses lead to sub-optimal representations~\cite{kang2019decoupling}. The classifier is then learned using Learnable Weight Scaling (LWS), $\tau$-normalization, which improves performance on the tail classes~\cite{kang2019decoupling}. Further, after this follow-up works~\cite{ye2020identifying, wang2021contrastive} like MiSLAS~\cite{zhou2017places} proposed Mixup~\cite{zhang2018mixup} based improved representation learning and LADE~\cite{hong2021disentangling} proposes improved classifier training by adapting to target label distribution. Further, contrastive methods, including PaCo~\cite{cui2021parametric} and BCL~\cite{ren2020balanced}, have demonstrated improved performance with contrastive learning. However, all these methods lead to performance degradation on head classes to improve performance on tail classes.  To mitigate this degradation, the techniques (like RIDE~\cite{wang2020long} etc.) learn different experts on different parts of the data distribution. These experts are learned in a way that makes them diverse in their predictions and can be combined efficiently to obtain improved predictions. However, these methods require additional computation to combine experts at the inference time. In our work, we can efficiently learn experts on majority and minority using a single ViT backbone, the predictions of which we average to prevent any additional inference overhead at the deployment time.

\vspace{1mm} \noindent \textbf{Vision Transformer.} In recent literature, Vision Transformers~\cite{dosovitskiyimage} have emerged as strong competitors for ResNets as they are easier to scale and lead to improved generalization. DeiT ~\cite{touvron2021deit} developed a data-efficient way to train these models by distilling through Convolutional Neural Networks. However, despite being data efficient, these models still produce sub-optimal performance on long-tailed data. RAC~\cite{long2022retrieval} utilizes pre-trained transformer for data-efficiency on long-tailed data. However, these pre-trained models are often domain specific and do not generalize well to other domains like medical, synthetic etc. In our work, we train Vision Transformers from scratch, even for small datasets like CIFAR-10 LT, CIFAR-100 LT, which makes them free from biases due to pre-training on large datasets~\cite{wang2023overwriting}.%

\vspace{1mm} \noindent \textbf{Data Efficient Vision Transformers (DeiT).}
The Vision Transformer (ViT) architecture~\cite{dosovitskiyimage} consists of transformer architecture stacked with Multi-Headed Self-Attention~\cite{vaswani2017attention}. To provide input to the Vision Transformer architecture, we first convert the image into patches. These image patches are passed through a linear layer to convert them into tokens that are then passed to the attention blocks. The attention blocks learn the relationship between these tokens for performing a given task. In addition to this, the ViT architecture also contains one classifier (\texttt{CLS}) token that represents the features to be used for classification. In the Data Efficient Transformer (DeiT)~\cite{touvron2021deit}, there is an additional distillation (\texttt{DIST}) token in the ViT backbone that learns via distillation from the teacher CNN. For the classification head and the distillation head,  $\mathcal{L}_{CE}$ is used for training (Fig.~\ref{deit-lt_fig:overview}). The final loss function for the network is:
\begin{equation}
    \mathcal{L}= \mathcal{L}_{CE}(f^{c}(x), y) + \mathcal{L}_{CE} (f^{d}(x), y_{t}), y_t = \arg \max_{i} g(x)_{i}
\end{equation}
here $f^{c}(x)$ is output from the classifier of student $\texttt{CLS}$ token, $f^{d}(x)$ is output from the classifier of student $\texttt{DIST}$ token, $g(x)$ denotes the output of the teacher CNN network, $y \in [K]$ is the ground truth, $y_t$ is the label produced by the teacher corresponding to the sample $x$, and $N_i$ is the number of samples in class $i$. At the time of inference in DeiT, we obtain logit outputs from the two heads $f^{d}(x)$ and $f^{c}(x)$, and average them to produce the final prediction.

\section{DeiT-LT (DeiT for Long-Tailed Data)}

\begin{table}
\centering
\setlength{\tabcolsep}{7pt}
\caption{{Effect of augmentations:} Comparison of teacher (\textit{Tch}) and student (\textit{Stu}) accuracy (\%) and training time (in hours) on CIFAR-10 LT ($\rho$ = 100) using various augmentation strategies with mixup (\cmark) and without mixup (\xmark). Despite low teacher training accuracy on the out-of-distribution images, the student (Stu.) performs better on the validation set.}
\label{deit-lt_tab:augs}
\vspace{-2mm}
\resizebox{0.75\linewidth}{!}{%
\begin{tabular}{@{}c|cc|cc|c@{}}
\toprule
\textbf{\begin{tabular}[c]{@{}c@{}}Tch\\ Model\end{tabular}} &
  \textbf{\begin{tabular}[c]{@{}c@{}}Stu\\ Augs.\end{tabular}} &
  \textbf{\begin{tabular}[c]{@{}c@{}}Tch\\ Augs.\end{tabular}} &
  \textbf{\begin{tabular}[c]{@{}c@{}}Tch\\ Acc.\end{tabular}} &
  \textbf{\begin{tabular}[c]{@{}c@{}}Stu\\ Acc.\end{tabular}} &
  \textbf{\begin{tabular}[c]{@{}c@{}}Train\\ Time\end{tabular}} \\ \midrule
\begin{tabular}[c]{@{}c@{}}RegNetY\\ 16GF\end{tabular} & Strong (\cmark) & Strong (\cmark) & 79.1 & 70.2 & 33.3 \\ \midrule
\multirow{3}{*}{ResNet-32}                             & Strong (\xmark) & Weak (\xmark)   & 97.2 & 54.2 & 17.8 \\
                                                       & Strong (\xmark) & Strong (\xmark) & 71.9 & 69.6 & 17.8 \\
                                                       & Strong (\cmark) & Strong (\cmark) & 56.6 & 79.4 & 19.0 \\ \bottomrule
\end{tabular}}
\vspace{-4mm}
\end{table}    

\label{deit-lt_sec:deit_lt}
In this section, we introduce DeiT-LT - the Data-efficient Image Transformer that is specialized to be effective for Long-Tailed data. We start with a DeiT transformer-based architecture which, in addition to the classification (\ttt{CLS}) token, also contains a distillation (\ttt{DIST}) token (Fig.~\ref{deit-lt_fig:overview}) that learns via distillation from a CNN.
The DeiT-LT introduces three particular design components, which are: \textbf{a)} the effective distillation via out-of-distribution (OOD) images, which induces local features and leads to the creation of experts,  \textbf{b)} training Tail Expert classifier using DRW loss and \textbf{c) }learning of low-rank generalizable features from flat teachers via distillation. In the following sections, we analyze our design choices in detail. We analyze CIFAR-10 LT using LDAM+DRW+SAM ResNet-32~\cite{rangwani_escapingsaddle} CNN teacher, to justify the rationale behind each design component.

\subsection{ Distillation via Out of Distribution Images}
\label{deit-lt_sebsec:out-of-dis-dist}
We now focus on how to distill knowledge from a CNN architecture to a ViT effectively. In the original DeiT work \cite{touvron2021deit}, the authors first train a large CNN, specifically RegNetY~\cite{radosavovic2020designing}, with strong augmentations ($\mathcal{A}$) as used by a ViT for distillation. However, this incurs the additional expense of training a large CNN for subsequent training of the ViT through distillation. In contrast, we propose to train a small teacher CNN (ResNet-32) with the usual weak augmentations, but during distillation, we pass strongly augmented images to obtain predictions to be distilled. 

\begin{figure}[!t]
\centering
    \includegraphics[width=0.60\linewidth]{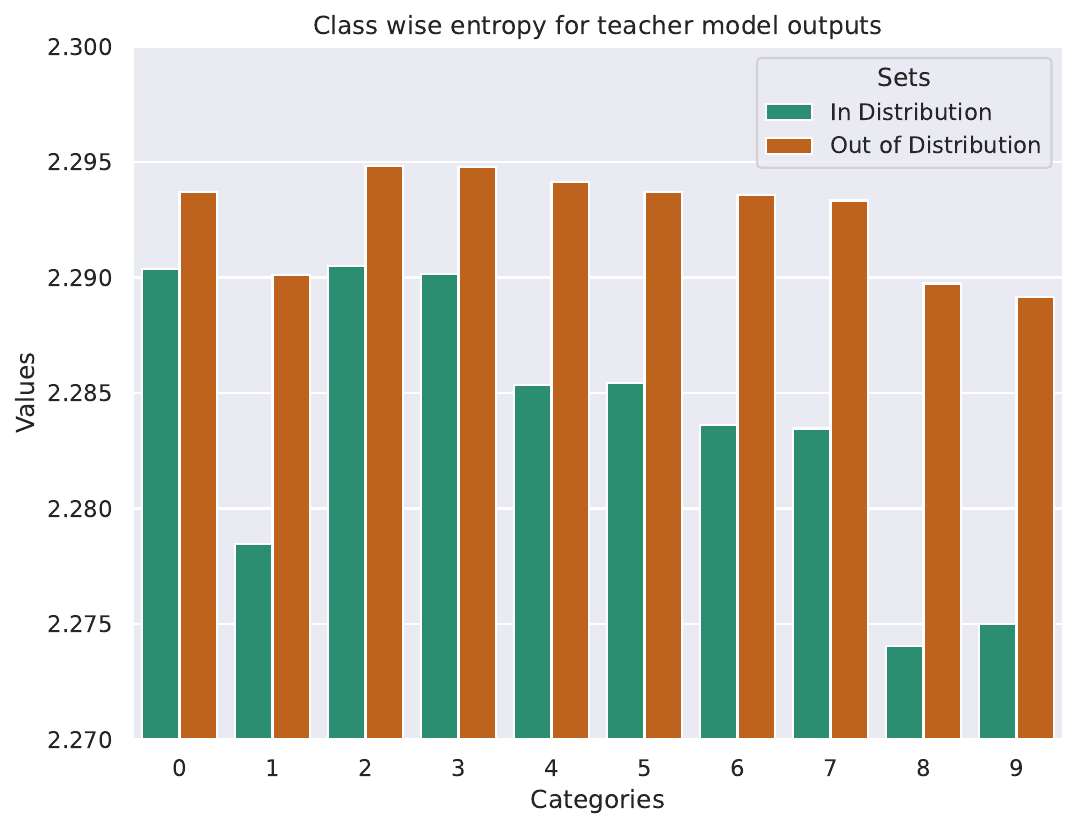}
    \vspace{-2mm}
    \caption{{Entropy of teacher outputs:} Comparison of the entropy of in-distribution samples and out-of-distribution samples with the ResNet-32 teacher on CIFAR-10 LT. We observe a higher accuracy in Table-\ref{deit-lt_tab:augs} corresponding to out-of-distribution samples.}
    \label{deit-lt_fig:entropy}
    \vspace{-2mm}
\end{figure}

These strongly augmented images are \emph{out-of-distribution (OOD)} images for the ResNet-32 CNN as the model's accuracy on these training images is low, as seen in Table \ref{deit-lt_tab:augs}. However, despite the low accuracy, the strong augmentations lead to effective distillation in comparison to the weak augmentations on which the original ResNet was trained (Table \ref{deit-lt_tab:augs}). This works because the ViT student learns to mimic the incorrect predictions of the CNN teacher on the out-of-distribution images, which in turn enables the student to learn the inductive biases of the teacher:

\begin{equation}
    f^{d} (X) \approx g (X) , X \sim A(x).
\end{equation}
Further, we find that creating additional out-of-distribution samples by mixing up images from two classes~\cite{yun2019cutmix, zhang2018mixup} improves the distillation performance. This can also be seen

\noindent
from the entropy of predictions on teacher, which are high (\ie more informative) for OOD samples (Fig.~\ref{deit-lt_fig:entropy}). \textit{In general, we find that increasing diverse amount of out-of-distribution~\cite{nayak2021effectiveness} data while distillation helps improve performance and leads to effective distillation from the CNN.} Details regarding the augmentations are in Appendix Sec. \ref{deit-lt_suppl:training_aug}.

\begin{figure*}[!t]
     \centering
     \begin{subfigure}[b]{0.32\textwidth}
         \centering
         \includegraphics[width=\textwidth]{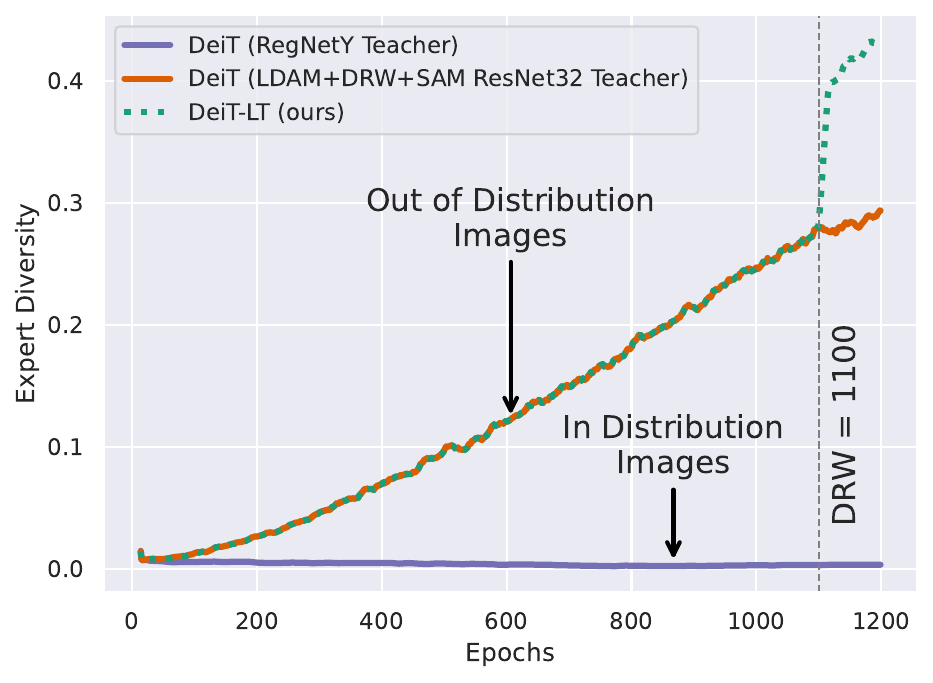}
         \caption{Diversity for \ttt{CLS} and \ttt{DIST} experts}
         \label{deit-lt_fig:similarity_plot}
     \end{subfigure}
     \begin{subfigure}[b]{0.29\textwidth}
         \centering
         \includegraphics[width=\textwidth]{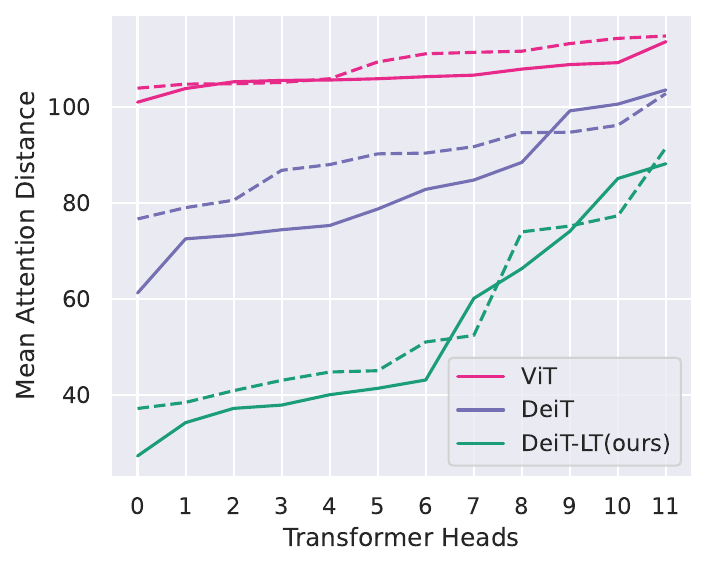}
         \caption{Locality of Attention Heads}
         \label{deit-lt_fig:ViT_locality}
     \end{subfigure}
    \begin{subfigure}[b]{0.32\textwidth}
         \centering
         \includegraphics[width=\textwidth]{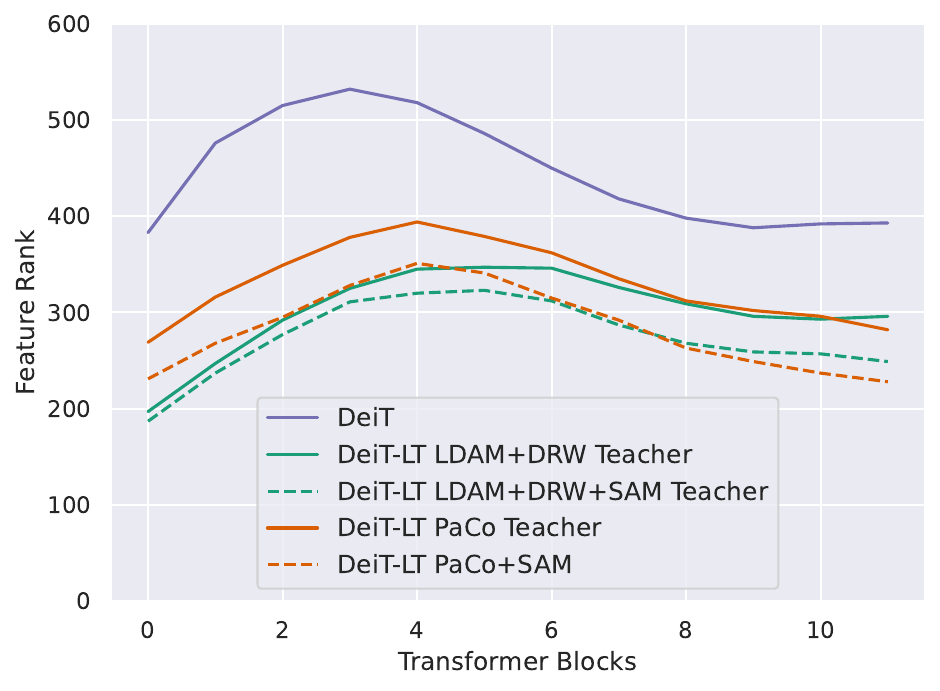}
         \caption{Rank of ViT from Distillation of CNN teachers.}
         \label{deit-lt_fig:ViT_rank}
     \end{subfigure}
     
        \caption{Effect of Distillation in DeiT-LT.  In \textbf{a)} we train DeiT-B with teachers trained on in-distribution images (RegNetY-16GF) and out-of-distribution images (ResNet32). The out-of-distribution distillation leads to diverse experts, which become more diverse with deferred re-weighting on the distillation token (DRW). In \textbf{b)} we plot the \emph{Mean Attention Distance} for the patches across the early self attention block 1 (solid) and  block 2 (dashed) for baselines, where we find that DeiT-LT leads to highly local and generalizable features. In \textbf{c)} we show the rank of features for \ttt{DIST} token, where we demonstrate that students trained with SAM are more low-rank in comparison to baselines}
        \label{deit-lt_fig:dist_sim}
\vspace{-2mm}
\end{figure*}

 Due to distillation via out-of-distribution images, the teacher predictions $y_t$ often differ from the ground truth $y$. Hence, the classification token (\texttt{CLS}) and distillation token (\texttt{DIST}) representations diverge while training. This phenomenon can be observed in Fig. \ref{deit-lt_fig:similarity_plot}, where the cosine distance between the representation of the \texttt{CLS} and \texttt{DIST} tokens increases as the training progresses. This leads to the \texttt{CLS} token being an expert on head classes, while the \texttt{DIST} token specializes in tail class predictions. Our observation debunks the \emph{myth that it is required for the \texttt{CLS} token predictions to be similar to \texttt{DIST}} for effective distillation in transformer, as observed by \citet{touvron2021deit}. 

 \noindent \textbf{Tail Expert with DRW loss.} Further in this stage, we also introduce Deferred Re-Weighting (DRW)~\cite{cao2019learning} for distillation loss, where we weigh the loss for each class using a factor $w_y = {1}/\{1 + (e_y - 1)\mathds{1}_{\mathrm{epoch \geq K}}\}$, where $e_y=\frac{1-\beta^{N_y}}{1-\beta}$ is the effective number of samples in class $y$~\cite{cui2019class}, after $K$ number of epochs~\cite{cao2019learning}. Hence the overall loss is given as:  

\begin{align*}
    \mathcal{L} = \frac{1}{2} \mathcal{L}_{CE}(f^{c}(x), y) + \frac{1}{2}\mathcal{L}_{DRW} (f^{d}(x), y_{t}),
     \text{where} ~ \mathcal{L}_{DRW} = -w_{y_t} \; log (f^{d}(x)_{y_t}).
\end{align*}
The DRW stage further enhances the focus of the distillation head (\texttt{DIST}) on the tail classes, leading to improved performance. This is also observed in Fig. \ref{deit-lt_fig:similarity_plot}, where the diversity between the two tokens improves after the introduction of the DRW stage. This leads to the creation of diverse \ttt{CLS} and \ttt{DIST} tokens, which are experts on the majority and minority classes respectively.

\noindent \textbf{Induction of Local Features: } To gain insights into the generality and effectiveness of OOD Distillation, we take a closer look at the tail features produced by DeiT-LT. In Fig.~\ref{deit-lt_fig:ViT_locality}, we plot the mean attention distance for each patch across ViT heads~\cite{raghu2021vision} (Details in Appendix Sec. \ref{deit-lt_suppl:local_connectivity}).  

\vspace{1mm} \noindent \textbf{Insight 1:} DeiT-LT contains heads that attend locally, like CNN, in the neighborhood of the patch in early blocks (1,2). 

\vspace{1mm} \noindent Due to this learning of local generalizable class agnostic features, we observe improved generalization on minority classes (Fig. \ref{deit-lt_fig:teaser-fig}\textcolor{red}{c}). Without the OOD distillation, we find that the vanilla DeiT-III and ViT baselines overfit only on the spurious global features (Fig.~\ref{deit-lt_fig:ViT_locality}) and do not generalize well for tail classes. Hence, this makes OOD distillation in DeiT-LT a well-suitable method for long-tailed scenarios.

\subsection{Low-Rank Features via SAM teachers}
\label{deit-lt_subsec:low-rank-sam}
To further improve the generalizability of the features, particularly for classes with less data, we propose to distill via \emph{teacher CNN models trained via Sharpness Aware Minimization (SAM) objective}~\cite{foret2020sharpness}. Models trained via SAM  converge to flat minima~\cite{rangwani_escapingsaddle} and lead to low-rank features~\cite{andriushchenko2023sharpness}. For analyzing the rank of features for the ViT student in LT case, we calculate rank specifically for the features of tail classes~\cite{andriushchenko2023sharpness}. We detail the procedure of our rank calculation in Appendix Sec. \ref{deit-lt_suppl:low_rank}.
We confirm our observations across diverse teacher models trained via LDAM and PaCo.
We find the following insight for distillation via \ttt{DIST} token: 

\vspace{1mm} \noindent \textbf{Insight 2.} We observe that distilling into ViT via predictions made using SAM teacher leads to low-rank generalizable (\texttt{DIST}) token features across blocks of ViT (Fig. \ref{deit-lt_fig:ViT_rank}). 

\vspace{1mm}  \noindent {This transfer of a CNN teacher's characteristic (low-rank) to the student, by just distilling via final logits, is a significant novel finding in the context of distillation for ViTs.}

\vspace{1mm} \noindent \textbf{Training Time.} In the original DeiT formulation, the authors~\cite{touvron2022deit} propose training a large CNN RegNetY-16GF at a high resolution (224 $\times$ 224) for distillation to the ViT. We find that competitive performance can be achieved even with training a smaller ResNet-32 CNN (32 $\times$ 32) at a lower resolution, as seen in Table~\ref{deit-lt_tab:augs}. This significantly reduces compute requirement and overall training time by 13 hours, as the ResNet-32 model can be trained quickly (Table~\ref{deit-lt_tab:augs}). Further, we find that with SAM teachers, the student converges much faster than vanilla teacher models, demonstrating the efficacy of SAM teachers for low-rank distillation (Appendix Sec. \ref{deit-lt_suppl:sam_convergence}).

\section{Experiments}

\subsection{Datasets}

We analyze the performance of our proposed method on four datasets, namely \textbf{CIFAR-10 LT}, \textbf{CIFAR-100 LT}, \textbf{ImageNet-LT}, and \textbf{iNaturalist-2018}. We follow \cite{cao2019learning} to create long-tailed versions of CIFAR~\cite{krizhevsky2009learning} datasets, where the number of samples is exponentially decayed using an imbalance factor $\rho =\frac{\max_i N_{i}}{\min_j N_j}$ (number of samples in the most frequent class by that in the least frequent class). For ImageNet-LT, we create an imbalanced version of the ImageNet~\cite{russakovsky2015imagenet} dataset as described in \cite{liu2019large}. We also report performance on iNaturalist-2018~\cite{van2018inaturalist}, a real-world long-tailed dataset. We divide the classes into three subcategories: \textbf{Head} (\textit{Many}), \textbf{Mid} (\textit{Medium}), and \textbf{Tail} (\textit{Few}) classes. More details regarding the datasets can be found in Appendix Sec. \ref{deit-lt_suppl:datasets}.
\begin{table}[!t]
    \centering
    
    \caption{Results on CIFAR-10 LT and CIFAR-100 LT datasets with $\rho$=50 and $\rho$=100. We report the \emph{overall} accuracy for available methods. (The teacher used to train the respective student (DeiT-LT) model can be identified by matching superscripts)}
    \resizebox{0.75\linewidth}{!}{%
    \begin{tabular}{l|c|c|c|c}
        \toprule[1pt]
         \multirow{2}{*}{\begin{tabular}{c}\textbf{Method}\end{tabular}} & \multicolumn{2}{c|}{CIFAR-10 LT} & \multicolumn{2}{c}{CIFAR-100 LT} \Tstrut\Bstrut\\
         \cline{2-5}
         & \multicolumn{1}{c|}{$\rho$ = 100} & \multicolumn{1}{c|}{$\rho$ = 50} & \multicolumn{1}{c|}{$\rho$ = 100} & \multicolumn{1}{c}{$\rho$ = 50} \Tstrut\Bstrut\\
         \midrule
         \rowcolor{Gray} \multicolumn{5}{c}{ResNet32 Backbone} \Tstrut\\
         \midrule
         
         CB Focal loss~\cite{cui2019class} & 74.6 & 79.3 & 38.3 & 46.2\\ 
         LDAM+DRW~\cite{cao2019learning} & 77.0 & 79.3 & 42.0 & 45.1 \\ 
         LDAM+DAP~\cite{jamal2020rethinking} & 80.0 & 82.2 & 44.1 & 49.2 \\
         BBN~\cite{zhou2020BBN} & 79.8 & 82.2 & 39.4 & 47.0 \\
         
        CAM~\cite{zhang2021bag} & 80.0 & 83.6  & 47.8 & 51.7 \\
        Log. Adj.~\cite{menon2020long} & 77.7 & -  & 43.9 & - \\
        RIDE~\cite{wang2020long}  & - & -  & 49.1 & - \\
        MiSLAS~\cite{zhong2021improving} & 82.1 & 85.7  & 47.0 & 52.3 \\
        Hybrid-SC~\cite{wang2021contrastive} & 81.4 & 85.4  & 46.7 & 51.9 \\
        SSD~\cite{li2021self_iccv} & - & -  & 46.0 & 50.5\\
        ACE~\cite{cai2021ace} & 81.4 & 84.9  & 49.6 & 51.9 \\
        GCL~\cite{gcl} & 82.7 & 85.5 & 48.7 & 53.6\\ 
    
        VS~\cite{kinivs} & 78.6 & -  & 41.7 \\
        VS+SAM~\cite{rangwani_escapingsaddle} & 82.4 & - & 46.6 & - \\
\midrule  
        \rowcolor{Gray}$^{1}$\small{L-D-SAM} \cite{rangwani_escapingsaddle} & 81.9 & 84.8 & 45.4  & 49.4 \\
        \rowcolor{Gray}$^{2}$PaCo+SAM\cite{rangwani_escapingsaddle, cui2021parametric} & 86.8  & 88.6  & 52.8& 56.6  \\
        \midrule

        \rowcolor{Gray} \multicolumn{5}{c}{ViT-B Backbone} \\
         \midrule  
         ViT ~\cite{dosovitskiyimage} & 62.6 & 70.1 & 35.0 & 39.0 \\
          ViT (cRT) ~\cite{kang2019decoupling} & 68.9 & 74.5 & 38.9 & 42.2 \\
         DeiT ~\cite{touvron2021deit} & 70.2 & 77.5 & 31.3 & 39.1 \\
         DeiT-III ~\cite{touvron2022deit} & 59.1 & 68.2 & 38.1 & 44.1\\

         \midrule
         \rowcolor{Gray}$^{1}$DeiT-LT(ours) & 84.8 & 87.5 & 52.0 & 54.1 \\
         \rowcolor{Gray}$^{2}$DeiT-LT(ours) & \textbf{87.5} & \textbf{89.8} & \textbf{55.6} & \textbf{60.5} \\

\bottomrule

    \end{tabular}}
    \label{deit-lt_tab:cifar10_cifar100}
    \vspace{-4mm}
    \end{table}

\subsection{Experimental Setup}
We follow the setup mentioned in DeiT~\cite{touvron2021deit} to create the student backbone for our experiments. We use the DeiT-B student backbone architecture for all the datasets.
We train our teacher models using re-weighting based LDAM-DRW-SAM method~\cite{rangwani_escapingsaddle} and the contrastive PaCo+SAM (training PaCo~\cite{cui2021parametric} with SAM~\cite{foret2020sharpness} optimizer), employing ResNet-32 for small scale datasets (CIFAR-10 LT and CIFAR-100 LT) and ResNet-50 for large scale ImageNet-LT, and iNaturalist-2018.
We train the head expert classifier with CE loss $\mathcal{L}_{CE}$ against the ground truth, while the tail expert classifier is trained with the CE+DRW loss $\mathcal{L}_{DRW}$ against the hard-distillation targets from the teacher network. 

\vspace{1mm} \noindent \textbf{Small scale CIFAR-10 LT and CIFAR-100 LT.} These models are trained for 1200 epochs, where DRW training for the Tail Expert Classifier starts from epoch 1100. Except for the DRW training (last 100 epochs), we use Mixup and Cutmix augmentation for the input images. These datasets are trained with a cosine learning rate schedule with a base LR of $5 \times 10^{-4}$ using the AdamW~\cite{loshchilov2017decoupled} optimizer.

\vspace{1mm} \noindent \textbf{Large scale ImageNet-LT and iNaturalist-2018.} These models are trained for 1400 and 1000 epochs, respectively, with the DRW training for the Tail Expert Classifier starting from 1200 and 900 epochs. We use Mixup and Cutmix throughout training. Both datasets follow a cosine learning rate schedule, with a base LR of $5 \times 10^{-4}$. More details on the experimental process can be found in Appendix Sec~\ref{deit-lt_suppl:experimental}.

\vspace{1mm} \noindent \textbf{Baselines.} We use the popular data-efficient baselines for ViT: \textbf{a) ViT: }The standard Vision Transformer (\textbf{ViT-B})\cite{dosovitskiyimage}   architecture trained with CE Loss against the ground truth. For a fair comparison, we train ViT with the same augmentation strategy used for the DeiT-LT experiments. \textbf{b) DeiT~\cite{touvron2021deit}}: Vanilla DeiT model that uses RegNetY-16GF teacher trained with in-distribution images for distillation. \textbf{c) DeiT-III: } A recent improved version of DeiT (\cite{touvron2022deit}) that focuses on improving the supervised learning of ViT on balanced datasets using three simple augmentations (GrayScale, Solarisation, and Gaussian Blur) and LayerScale \cite{touvron2021going}, also demonstrating the redundancy of distillation in DeiTs. The long-tailed baseline of \textbf{d) ViT (cRT): } a decoupled approach of first training classifier (ViT) and then re-training the classifier for a small number of epochs with class-balanced sampling \cite{kang2019decoupling}. We further attempted training other baselines like LDAM, etc, on ViT. However, we found some optimization difficulties in training ViTs (details in Appendix Sec. \ref{deit-lt_suppl:additional_baselines}).

We want to convey that we do not compare against baselines~\cite{tian2022vl,long2022retrieval, LiVT, xu2023rethink}, which use pre-training, usually on large datasets, to produce results on even CIFAR datasets (Ref. Appendix Sec. \ref{deit-lt_suppl:clip}). Our goal is to develop a generic technique for training ViTs across domains and modalities on long-tailed data without requiring any external supervision.

\vspace{-1mm}
\section{Results}
\vspace{-1mm}
In this section, we present results for DeiT-LT across various datasets. We use re-weighting based LDAM+DRW+SAM (referred to as L-D-SAM in Table \ref{deit-lt_tab:cifar10_cifar100},\ref{deit-lt_tab:imgnet},\ref{deit-lt_tab:inat}) and contrastive PaCo+SAM teachers for training DeiT-LT student models.

\begin{table}[!t]
    \centering
    \setlength{\tabcolsep}{7pt}
    \caption{Results on ImageNet-LT. (The teacher used to train respective student (DeiT-LT) can be identified by matching superscripts) }
    
    \resizebox{0.75\linewidth}{!}{%
        \begin{tabular}{l|c|ccc}
            \toprule[1pt]
            \multirow{3}{*}{Method} & \multicolumn{4}{c}{ImageNet-LT} \Bstrut \\
            \cline{2-5}
            & Overall & Head & Mid & Tail \Tstrut\Bstrut\\
            \midrule
            \rowcolor{Gray} \multicolumn{5}{c}{ResNet50 Backbone} \Tstrut\Bstrut\\
            \midrule
            CB Focal loss~\cite{cui2019class} & 33.2 & 39.6 & 32.7 & 16.8 \\ 
            LDAM~\cite{cao2019learning} & 49.8 & 60.4 & 46.9 & 30.7 \\ 
            c-RT~\cite{kang2019decoupling} & 49.6 & 61.8 & 46.2 & 27.3 \\
            $\tau$-Norm~\cite{kang2020exploring} & 49.4 & 59.1 & 46.9 & 30.7 \\
            Log. Adj.~\cite{menon2020long} & 50.1 & 61.1 & 47.5 & 27.6 \\
            RIDE(3 exps)~\cite{wang2020long} & 54.9 & 66.2 & 51.7 & 34.9 \\
            MiSLAS~\cite{zhong2021improving} & 52.7 & 62.9 & 50.7 & 34.3 \\
            Disalign~\cite{zhang2021distribution} & 52.9 & 61.3 & 52.2 & 31.4 \\ 
            TSC~\cite{li2022targeted} & 52.4 & 63.5 & 49.7 & 30.4 \\ 
            GCL~\cite{gcl} & 54.5 & 63.0 & 52.7 & 37.1 \\
            SAFA~\cite{hong2022safa} & 53.1 & 63.8 & 49.9 & 33.4 \\
            BCL~\cite{ren2020balanced} & 57.1 & 67.9 & 54.2 & 36.6 \\
            ImbSAM~\cite{zhou2023imbsam} & 55.3 & 63.2 & 53.7 & 38.3 \\
            CBD$_{ENS}$~\cite{iscen2021cbd}& 55.6&68.5 &52.7 &29.2 \\
            \midrule
            \rowcolor{Gray}$^{1}$L-D-SAM~\cite{rangwani_escapingsaddle} & 53.1 & 62.0 & 52.1 & 32.8 \\
            \rowcolor{Gray}$^{2}$PaCo+SAM~\cite{rangwani_escapingsaddle, cui2021parametric} & 57.5	& 62.1	& 58.8	& 39.3 \\
            \midrule
            \rowcolor{Gray} \multicolumn{5}{c}{ViT-B Backbone} \\
            \midrule
            ViT ~\cite{dosovitskiyimage} & 37.5 & 56.9 & 30.4 & 10.3 \\
            DeiT-III ~\cite{touvron2022deit} & 48.4 & \textbf{70.4} & 40.9 & 12.8 
            
            \\
            \midrule
            \rowcolor{Gray}$^{1}$DeiT-LT(ours) & 55.6 & 65.2 & 54.0 & 37.1 \\
            \rowcolor{Gray}$^{2}$DeiT-LT(ours) & \textbf{59.1} & 66.6 & \textbf{58.3} & \textbf{40.0} \\
            \bottomrule
        \end{tabular}
    }
    \label{deit-lt_tab:imgnet}
    \vspace{-4mm}
\end{table}

\begin{table}[!t]
    \centering
    \setlength{\tabcolsep}{10pt}
    \caption{Results on iNaturalist-2018. (The teacher used to train student (DeiT-LT) can be identified by matching superscripts) }
    
    \resizebox{0.75\linewidth}{!}{%
        \begin{tabular}{l|c|ccc}
            \toprule[1pt]
            \multirow{3}{*}{Method} & \multicolumn{4}{c}{iNaturalist-2018} \Bstrut \\
            \cline{2-5}
            & Overall & Head & Mid & Tail \Tstrut\Bstrut\\
            \midrule
            \rowcolor{Gray} \multicolumn{5}{c}{ResNet50 Backbone} \Tstrut\Bstrut\\
            \midrule
            c-RT~\cite{kang2019decoupling} & 65.2 & 69.0 & 66.0 & 63.2 \\
            $\tau$-Norm~\cite{kang2020exploring} & 65.6 & 65.6 & 65.3 & 65.9  \\
            RIDE(3 exps)~\cite{wang2020long} & 72.2 & 70.2 & 72.2 & 72.7 \\
            MiSLAS~\cite{zhong2021improving} & 71.6 & \textbf{73.2} & 72.4 & 70.4 \\
            Disalign~\cite{zhang2021distribution} & 70.6 & 69.0 & 71.1 & 70.2 \\ 
            TSC~\cite{li2022targeted} & 69.7 & 72.6 & 70.6 & 67.8 \\ 
            GCL~\cite{gcl} & 71.0 & 67.5 & 71.3 & 71.5 \\
            ImbSAM~\cite{zhou2023imbsam}& 71.1 & 68.2 & 72.5 & 72.9 \\
            CBD$_{ENS}$~\cite{iscen2021cbd} & 73.6 & 75.9 & 74.7 & 71.5 \\
            \midrule
            \rowcolor{Gray}$^{1}$L-D-SAM~\cite{rangwani_escapingsaddle} & 70.1 & 64.1 & 70.5 & 71.2 \\
            \rowcolor{Gray}$^{2}$PaCo+SAM~\cite{rangwani_escapingsaddle} & 73.4 & 66.3 & 73.6 & 75.2 \\
            \midrule
            \rowcolor{Gray} \multicolumn{5}{c}{ViT-B Backbone} \\
            \midrule
            ViT ~\cite{dosovitskiyimage} & 54.2 & 64.3 & 53.9 & 52.1 \\
            DeiT-III ~\cite{touvron2022deit} & 61.0 & 72.9 & 62.8 & 55.8 \\
            \midrule
            \rowcolor{Gray}$^{1}$DeiT-LT(ours) & 72.9 & 69.0 & 73.3 & 73.3 \\
            \rowcolor{Gray}$^{2}$DeiT-LT(ours) & \textbf{75.1} & 70.3 & \textbf{75.2} & \textbf{76.2} \\
            \bottomrule
        \end{tabular}
    }
    \label{deit-lt_tab:inat}
    \vspace{-5mm}
\end{table}

\begin{figure}[!t]
    \centering
    \includegraphics[width=0.85\columnwidth]{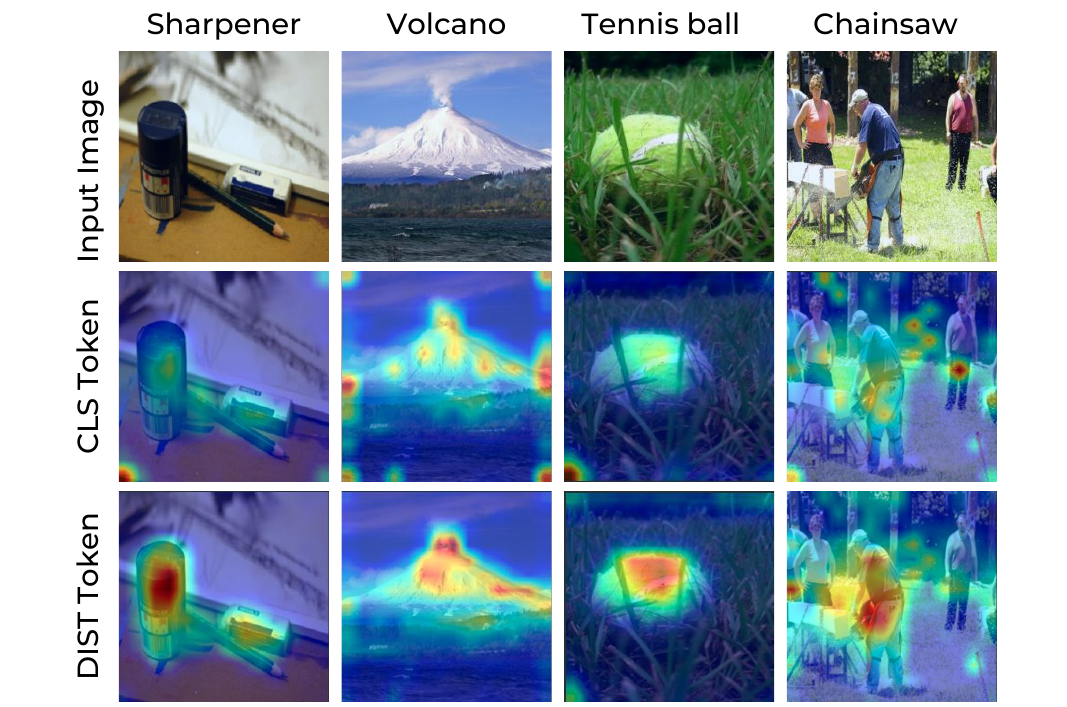}
    \caption{ Visual comparison of the attention maps with respect to the \ttt{CLS} and \ttt{DIST} tokens for \emph{tail} images from the ImageNet-LT dataset. The attention maps are computed by \emph{Attention Rollout} \cite{abnar2020quantifying}.}
    \label{deit-lt_fig:attention-vis}
    \vspace{-5mm}
\end{figure}

\vspace{1mm} \noindent \textbf{Results on Small Scale Datasets.} 
Table \ref{deit-lt_tab:cifar10_cifar100} presents results for the CIFAR-10 LT and CIFAR-100 LT datasets, with varying imbalance factors ($\rho = 100$ and $\rho = 50$). We primarily compare our results to the SotA methods, which train the networks from scratch. The other techniques utilize additional pre-training with extra data~\cite{LiVT, chen2022reltransformer}, making the comparison unfair. Our proposed student network DeiT-LT outperformed the teachers used for their training by an average of 1.9\% and 4.5\% on CIFAR-10 LT and CIFAR-100 LT, respectively. 
This demonstrates the advantage of training the DeiT-LT transformer, which provides additional generalization improvements over the CNN teacher. Further, the DeiT-LT (PaCo+SAM) model significantly improves by 24.9\%  over the ViT baseline (which has the same augmentations as in DeiT-LT) and 28.4\% over the data efficient DeiT-III transformer for CIFAR-10 LT dataset for $\rho = 100$. A similar improvement can also be observed for the CIFAR-100 LT dataset, where DeiT-LT (PaCo+SAM) fares better than ViT baseline and DeiT-III by 20.6\% and 17.5\%, respectively. This shows the effectiveness of the DeiT-LT distillation procedure via CNN teachers. Compared to CNN-based methods, we demonstrate that the transformer-based methods can achieve SotA performance when trained with DeiT-LT distillation procedure, combining both the scalability of transformers on head classes and utilizing inductive biases of CNN for tail classes. To the best of our knowledge, our proposed DeiT-LT for transformers is the \emph{first work in literature that can achieve SotA performance for long-tailed data on small datasets when trained from scratch}. The other works~\cite{LiVT} require transformer pre-training on large datasets, such as ImageNet, to achieve comparable performance on these small datasets.

\vspace{1mm} \noindent \textbf{Results on Large Scale Datasets.}
In this section, we present results attained by DeiT-LT on the large-scale long-tailed datasets of ImageNet-LT and iNaturalist-2018. 
We train all transformer-based methods for similar epochs for a particular dataset, to keep the comparison fair across all baselines (See Appendix Sec. \ref{deit-lt_suppl:training_config}).  Table \ref{deit-lt_tab:imgnet} presents the result on the ImageNet-LT dataset, where we find that when distilling using LDAM+DRW+SAM (L-D-SAM), our DeiT-LT significantly improves by 2.5\% over the teacher network. 
Notably, it can be seen that our DeiT-LT method, when distilling from PaCo+SAM teacher, achieves a 1.6\% performance gain over the already near SotA teacher network. 
Further, the distillation-based DeiT-LT method achieves a significant gain of 21.6\% and 10.7\% over the baseline transformer training methods, ViT and DeiT-III respectively. This demonstrates that improvement due to distillation scales well with an increase in the size of datasets. 
For iNaturalist-2018, we notice an improvement of close to 3\% over the LDAM+DRW+SAM (L-D-SAM) teacher network and an improvement of 1.7\% over the recent PaCo+SAM teacher.
Additionally, we notice a significant improvement over the data-efficient transformer-based baselines. The data-efficient transformer-based methods struggle while modeling the tail classes, which is supplemented via proposed Distillation loss in DeiT-LT. This enables DeiT-LT to work well across all the classes; the head classes benefit from enhanced learning capacity due to scalable Vision Transformer (ViT), and tail classes are learned well via distillation. Our results are superior for both datasets compared to the CNN-based SotA methods, demonstrating the advantage of DeiT-LT. (Refer Appendix Sec. \ref{deit-lt_suppl:detailed_results} for detailed results.) %

\begin{table}[t]
    \centering
    \caption{Table showing ablations for various components in DeiT-LT for CIFAR-10 LT and CIFAR-100 LT.}
\begin{tabular}{ccc|cc}
  \toprule
   \textbf{OOD Distill} & \textbf{DRW} & \textbf{SAM} & \textbf{C10 LT}  & \textbf{C100 LT} \\ \midrule
  \xmark & \xmark & \xmark & 70.2 & 31.3 \\
  \cmark & \xmark & \xmark & 84.5 & 48.9 \\
  \cmark & \cmark & \xmark & 87.3 & 54.5\\
  \cmark & \cmark & \cmark & 87.5 & 55.6\\
  \bottomrule
  \end{tabular}
\label{deit-lt_tab:ablation_table}
\vspace{-4mm}
\end{table}

\section{Analysis and Discussion}

\vspace{1mm} \noindent \textbf{Visualizations of Attentions.}
Our training methodology ensures that the \ttt{CLS} and \ttt{DIST} representations diverge while training. While the \ttt{CLS} token is trained against the ground truth, it cannot learn efficient representation for tail classes' images due to ViT's inability to train well on small amounts of data. Distilling from a teacher via out-of-distribution data and introducing re-weighting loss helps the \ttt{DIST} token to learn better representation for the images of minority classes as compared to the \ttt{CLS} token. We further corroborate this by comparing the attention visualization obtained through \emph{Attention Rollout}~\cite{abnar2020quantifying}, for the \ttt{CLS} and \ttt{DIST} token on tail images, for ImageNet-LT dataset (as CIFAR-10 is too small) using DeiT-LT. As can be seen in Fig.~\ref{deit-lt_fig:attention-vis}, the \ttt{CLS} and the \ttt{DIST} token focus on different parts of the image. The \ttt{DIST} token is able to identify the patches of interest (high red intensity) for images of tail classes, while the \ttt{CLS} token fails to do so. The diversity in localized regions demonstrates the complementary information present across the \ttt{CLS} and \ttt{DIST} experts, which is in contrast with DeiT, where both the tokens \ttt{CLS} and \ttt{DIST} are quite similar. We compare visualization with different methods in Appendix Sec. \ref{deit-lt_suppl:vis_attention}.

\vspace{1mm} \noindent \textbf{Ablation Analysis Across DeiT-LT components.}
We analyze the influence of three key components of our DeiT-LT method, namely OOD distillation, training the Tail Expert classifier with DRW loss, and using SAM teacher for distillation. As can be seen in Table ~\ref{deit-lt_tab:ablation_table}, using OOD distillation brings around 14\% and 18\% improvement over DeiT \cite{touvron2021deit} for CIFAR-10 LT and CIFAR-100 LT, respectively, followed by the other two components, which further improve the accuracy by around 3\% and 6.7\% for CIFAR-10 LT and CIFAR-100 LT, respectively.

\vspace{1mm} \noindent \textbf{Analysis across Transformer Variants.}
In this section, we aim to analyze the performance of DeiT-LT across transformer variants having different capacities. 
\begin{table}[t]
    `\centering
    
    \caption{Analysis across transformer capacity for CIFAR-10 LT  and CIFAR-100 LT for DeiT-LT student($\rho = 100$) with PaCo teacher. }
    \begin{tabular}{c|c|ccc}
        \toprule[1pt]
        \rowcolor{Gray} Model & Overall & Head & Mid & Tail \\
        \midrule 
        \multicolumn{5}{c}{CIFAR-10 LT ($\rho = 100$)} \\
        \midrule 
        DeiT-LT  Tiny (Ti) & 80.8&89.7 &75.1 &79.4 \\
        DeiT-LT  Small (S) & 85.5&92.7 &81.5 &83.7 \\

        DeiT-LT  Base (B) &87.5 &94.5 &84.1 &85.0 \\
        \midrule
        \multicolumn{5}{c}{CIFAR-100 LT ($\rho = 100$)} \\
        \midrule
        DeiT-LT  Tiny (Ti) & 49.3& 66.3& 50.0&27.3 \\
        DeiT-LT  Small (S) & 54.3& 72.6& 54.8&31.1 \\

        DeiT-LT  Base (B) &55.6 &73.1 &56.9 &32.1 \\
\bottomrule      
    \end{tabular}
    \label{deit-lt_tab:cifar10_capacity}
    \vspace{-4mm}
\end{table}
For this, we fix the teacher network and training schedules while varying the network sizes. We experiment with the ViT-Ti, ViT-S, and ViT-B architectures, as introduced in the original ViT work~\cite{dosovitskiyimage}. In Table~\ref{deit-lt_tab:cifar10_capacity}, we observe that the proposed DeiT-LT method scales well with the increased capacity of the Transformer network, and leads to performance improvements.

\label{deit-lt_sec:conclusion}
 \vspace{1mm} 
 \noindent\textbf{Limitations.} One limitation of our 
framework is that the learning for tail classes is done mostly through distillation. Hence, the performance on tail classes remains similar (Table ~\ref{deit-lt_tab:imgnet} and ~\ref{deit-lt_tab:inat}) to that of the CNN classifier. Future works can aim to develop adaptive methods that can shift their focus from CNN to ground truth labels, as the CNN feedback saturates.
\vspace{-1mm}
\section{ Conclusion}
\vspace{-1mm}
In this work, we introduce DeiT-LT, a training scheme to train ViTs from scratch on real-world long-tailed datasets efficiently. We reintroduce the idea of knowledge distillation into ViT students via teacher CNN, as it enables effective learning on the tail classes. This distillation component was found to be redundant and removed from the latest DeiT-III.  Further, in DeiT-LT, we introduce out-of-distribution (OOD) distillation via the teacher, in which we pass strongly augmented images to teachers originally trained via mild augmentations for distillation. The distillation loss is re-weighted to enhance the focus on learning from tail classes. This helps make the classification token an expert on the head classes and the distillation token an expert on the tail classes. To improve generality in minority classes, we induce low-rank features in ViT by distilling from teachers trained from Sharpness Aware Minimization (SAM). The proposed DeiT-LT scheme allows ViTs to be trained from scratch as CNNs and achieve performance competitive to SotA without requiring any pre-training on large-datasets.

\part{Semi-Supervised Long-Tail Learning for Non-Decomposable Objectives}
\label{part:SemiSL_LT}

 \chapter{Cost-Sensitive Self-Training for Optimizing Non-Decomposable Metrics}
\label{chap:csst}

\begin{changemargin}{7mm}{7mm} 
Self-training based semi-supervised learning algorithms have enabled the learning of highly accurate deep neural networks, using only a fraction of labeled data. However, the majority of work on self-training has focused on the objective of improving accuracy whereas practical machine learning systems can have complex goals (e.g. maximizing the minimum of recall across classes etc.) that are non-decomposable in nature. In this work, we introduce the Cost-Sensitive Self-Training (\ttt{CSST}) framework which generalizes the self-training-based methods for optimizing non-decomposable metrics. We prove that our framework can better optimize the desired non-decomposable metric utilizing unlabeled data, under similar data distribution assumptions made for the analysis of self-training.  Using the proposed \ttt{CSST} framework we obtain practical self-training methods (for both vision and NLP tasks) for optimizing different non-decomposable metrics using deep neural networks.  Our results demonstrate that \ttt{CSST} achieves an improvement over the state-of-the-art in most cases across datasets and objectives.

\end{changemargin}

\section{Introduction}

In recent years, semi-supervised learning algorithms are increasingly being used for training deep neural networks~\cite{chapelle2009semi,kingma2014semi,sohn2020fixmatch,xie2020self}. These algorithms lead to accurate models by leveraging the unlabeled data in addition to the limited labeled data present. For example, it’s possible to obtain a model with minimal accuracy degradation ($\leq 1\%$) using 5\% of labeled data with semi-supervised algorithms compared to supervised models trained using 100\% labeled data~\cite{sohn2020fixmatch}. Hence, the development of these algorithms has resulted in a vast reduction in the requirement for expensive labeled data.

Self-training is one of the major paradigms for semi-supervised learning. It involves obtaining targets (\eg pseudo-labels) from a network from the unlabeled data, and using them to train the network further. The modern self-training methods also utilize additional regularizers that enforce prediction consistency across input transformations (e.g., adversarial perturbations~\cite{miyato2018virtual}, augmentations~\cite{xie2020unsupervised,sohn2020fixmatch}, etc.) , enabling them to achieve high performance using only a tiny fraction of labeled data. Currently, the enhanced variants of self-training with consistency regularization~\cite{zhang2021flexmatch,pham2021meta} are among the state-of-the-art (SOTA) methods for semi-supervised learning. 

Despite the popularity of self-training methods, most of the works~\cite{xie2020unsupervised, berthelot2019mixmatch, sohn2020fixmatch} have focused on the objective of improving prediction accuracy. However, there are nuanced objectives in real-world based on the application requirements. Examples include minimizing the worst-case recall~\cite{mohri2019agnostic} used for federated learning, classifier coverage for minority classes for ensuring fairness~\cite{goh2016satisfying}, etc. These objectives are complex and cannot be expressed just by using a loss function on the prediction of a single input (i.e., non-decomposable). There has been a considerable effort in optimizing non-decomposable objectives for different supervised machine learning models~\cite{narasimhan2021training,sanyal2018optimizing}. 
\begin{figure*}

\begin{minipage}[c]{0.55\linewidth}
    \includegraphics[width=\textwidth]{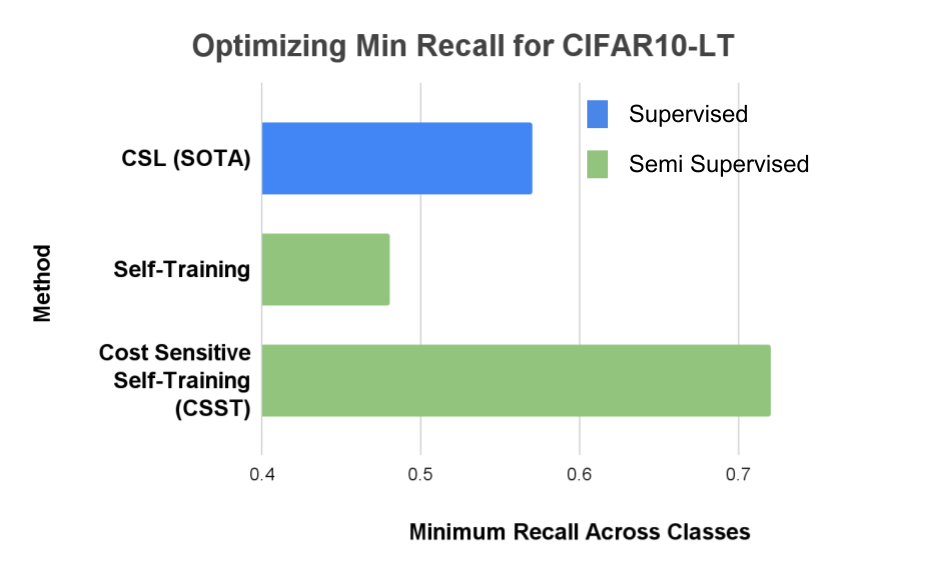}
\end{minipage}\hfill
\begin{minipage}[c]{0.45\linewidth}
  \caption{We show a comparison of the SOTA CSL~\cite{narasimhan2021training} method with the Self-training-based Semi-Supervised methods, for optimizing the minimum recall objective on the CIFAR10-LT dataset. Our proposed \ttt{CSST} framework produces significant gains in the desired metric leveraging additional unlabeled data through our proposed weighted novel consistency regularizer and thresholding mechanism.}
  \label{csst_fig:overview}
\end{minipage}
\vspace{-5mm}
\end{figure*}
However, as supervision can be expensive, in this work we aim to answer the question:
\emph{Can we optimize non-decomposable objectives using self-training-based methods developed for semi-supervised learning?}  

 We first demonstrate that vanilla self-training methods (e.g., FixMatch~\cite{sohn2020fixmatch}, UDA~\cite{xie2020unsupervised}, etc.) can produce unsatisfactory results for non-decomposable metrics (Fig. \ref{csst_fig:overview}). We then generalize the Cost-Sensitive Loss for Self-Training by introducing a novel weighted consistency regularizer, for a particular non-decomposable metric. Further, for training neural networks we introduce appropriate loss functions and pseudo label selection (thresholding) mechanisms considering the non-decomposable metric we aim to optimize. We also prove that we can achieve better performance on desired non-decomposable metric through our framework utilizing self-training, under similar assumptions on data distributions as made for theoretical analysis of self-training~\cite{wei2020theoretical}. We demonstrate the practical application by optimizing various non-decomposable metrics by plugging existing methods (\eg FixMatch~\cite{sohn2020fixmatch} etc.) into our framework. Our framework leads to a significant average improvement in desired metric of minimizing worst-case recall while maintaining similar accuracy (Fig. \ref{csst_fig:overview}).

 In summary: \textbf{a)} we introduce a Cost-Sensitive Self-Training (\texttt{CSST})  framework for optimizing non-decomposable metrics that utilizes unlabeled data in addition to labeled data. (Sec.~\ref{csst_sec:proposed-method})
 \textbf{b)} we provably demonstrate that our \texttt{CSST} framework can leverage unlabeled data to achieve better performance over baseline on desired non-decomposable metric (Sec.~\ref{csst_sec:theoretical_results}) \textbf{c)} we show that by combining \ttt{CSST} with self-training frameworks (\eg FixMatch~\cite{sohn2020fixmatch}, UDA~\cite{xie2020unsupervised} etc.)  leads to effective optimization of non-decomposable metrics, resulting in significant improvement over vanilla baselines. (Sec.~\ref{csst_sec:expt})

\section{Preliminaries}
\label{csst_sec:preliminaries}
\subsection{Non-Decomposable Objectives and Reduction to Cost-Sensitive Learning}
\label{csst_sec:ndo-loss-fun}
\begin{wraptable}[13]{r}{7.0cm}
\vspace{-6.8mm}
\caption{\addedtext{Metrics defined using entries of a confusion matrix.}}\label{csst_wrap-tab:metrics}
\vspace{-5mm}
\addedtext{\begin{tabular}{cc}\\\midrule  
Metric & Definition  \\\midrule
\midrule
 Recall ($\mathrm{rec}_i[F]$) & $\frac{C_{i,i}[F]}{ \sum_j{C_{i,j}[F]} } $ \\  \midrule
Coverage  ($\mathrm{cov}_i[F]$) & $\sum_j{C_{j,i}[F]} $ \\   \midrule
Precision  ($\mathrm{prec}_i[F]$) & $\frac{C_{i,i}[F]}{ \sum_k{C_{k,i}[F]} } $ \\  \midrule
Worst Case Recall &  $ \min_{i} \frac{C_{i,i}[F]}{ \sum_j{C_{i,j}[F]} } $ \\  \midrule
Accuracy  & $\sum_{i}{C_{i,i}[F]}$ \\ \bottomrule 
\end{tabular}}
\vspace{1mm}
\end{wraptable} 

We consider the $K$-class classification problem
with an instance space $\mX$ and the set of labels $\mY = [K]$.
The data distribution on $\mX \times [K]$ is denoted by $D$.
For $i \in [K]$, we denote by $\pi_i$ the class prior $\prob(y = i)$. 
Notations commonly used across chapter are in  Table \ref{csst_tab:notations} present in Appendix.
For a classifier $F : \mX\rightarrow [K]$, 
we define confusion matrix $\mathbf{C}[F] \in \RR^{K \times K}$ by 
$C_{ij}[F] = \ex[(x, y) \sim D]{\indc(y=i, F(x)=j)}$.
Many metrics relevant to classification can be defined as functions of entries of confusion matrices such as class-coverage, recall and accuracy to name a few. We introduce more complex metrics,  which are of practical importance in the case of imbalanced distributions~\cite{cotter2019optimization} (Tab. \ref{csst_wrap-tab:metrics}).

A classifier often tends to suffer low recalls on tail (\addedtext{minority}) classes in such cases. Therefore, one may want to maximize the worst case recall, \begin{equation*}
    \label{csst_eq:min-recall-obj}
    \modifiedtext{\max_F \min_{i \in [K]} \text{rec}_i[F]}.
\end{equation*} 

Similarly, on long-tailed datasets, the tail classes suffer from low coverage, lower than their respective priors. An interesting objective in such circumstances is to maximise the mean recall, subject to the coverage being within a given margin.
\begin{equation}
   \label{csst_eq:coverage-constraint-obj}
  \max_{F} \frac{1}{K}
  \sum_{i\in [K]}\text{rec}_i[F] \quad \text{s.t. } \cov_j[F] \ge \frac{0.95}{K}, \forall j \in [K].
\end{equation} 
Many of these metrics are \textbf{non-decomposable},
i.e., one cannot compute these metrics 
by simply calculating the average of scores on individual examples.
Optimizing for these metrics can be regarded as instances of cost-sensitive learning (CSL). More specifically, \emph{optimization problems of the form which can be written as a linear combination of $G_{i,j}$ and $C_{ij}[F]$ will be our focus in this work} where $\bG$ is a $K \times K$ matrix.
\begin{equation}
   \label{csst_eq:csl-obj}
   \max_F \sum_{i, j \in [K]}G_{ij} C_{ij}[F],
\end{equation}
The entry $G_{ij}$ represents the reward associated with predicting class $j$ when the true class is $i$.
The matrix $\bG$ is called a gain matrix \cite{narasimhan2021training}.
Some more complex non-decomposable objectives for classification can be reduced to CSL \cite{narasimhan2015consistent,tavker2020consistent,narasimhan2021training}.
For instance, 
the aforementioned two complex objectives
can be reduced to CSL using continuous relaxation or a Lagrange multiplier as bellow.
  Let $\Delta_{K- 1} \subset \RR^{K}$ be the $K-1$-dimensional probability simplex.
Then, maximizing the minimum recall is equivalent to the 
saddle-point optimization problem:
\begin{equation}
   \max_F \min_{\blambda \in \Delta_{K-1}} \sum_{i \in [K]}\lambda_i \frac{C_{ii}[F]}{\pi_i}.
\end{equation}
Thus, for a fixed $\blambda$, 
the corresponding gain matrix is given as a diagonal matrix $\diag(G_1, \dots, G_K)$
with $G_i = \lambda_i / \pi_i$ for $1 \le i \le K$.
Similarly, using Lagrange multipliers $\blambda \in \RR_{\ge 0}^K$, 
Eq. \eqref{csst_eq:coverage-constraint-obj} is rewritten as a max-min optimization problem \citep[Sec. 2]{narasimhan2021training}:
\begin{equation}
 \label{csst_eq: cov-const-obj}
  \addedtext{ \max_F \min_{\blambda \in \RR^K_{\ge 0}} 
   \frac{1}{K}
   \sum_{i\in [K]} C_{ii}[F]/\pi_i +\sum_{j \in [K]}  \lambda_j \left(
      \sum_{i \in [K]} C_{ij}[F] - 0.95/K
   \right).}
\end{equation}
In this case,
the corresponding gain matrix $\mathbf{G}$ is given as $G_{ij} = \frac{\delta_{ij}}{K\pi_i} + \lambda_j$,
where $\delta_{ij}$ is the Kronecker's delta.
One can solve these max-min problems by alternatingly updating $\blambda$ 
(using exponented gradient or projected gradient descent) and 
optimizing the cost-sensitive objectives~\cite{narasimhan2021training}.

\subsection{Loss Functions for Non-Decomposable Objectives}
\label{csst_sec:loss-fn-for-ndo}
\label{csst_loss func: NDO}
The cross entropy loss function is appropriate for optimizing accuracy for deep neural networks, 
however, learning with CE can suffer low performance for cost-sensitive objectives \cite{narasimhan2021training}.
Following \cite{narasimhan2021training}, we introduce calibrated loss functions for given gain matrix $\mathbf{G}$.
We let 
$p_m: \mX \rightarrow \Delta_{K-1} \subset \RR^{K}$
be a prediction function of a model, where $\Delta_{K-1}$
is the $K-1$-dimensional probability simplex.
For a gain matrix $\mathbf{G}$, 
the corresponding loss function is 
given as a combination of logit adjustment \cite{menon2020long} and loss re-weighting \cite{patrini2017making}.
We decompose the gain matrix $\bG$ as $\bG = \bM \bD$, where 
$\bD = \diag(G_{11}, \dots, G_{KK})$ be a diagonal matrix, with $D_{ii} > 0, \forall i \in [K]$
and $\bM \in \RR^{K \times K}$.
For $y \in [K]$ and model prediction $p_m(x)$, the hybrid loss is defined as follows:
\begin{equation}
   \hybloss(y, p_m(x)) = -\sum_{i \in [K]} M_{yi} \log\left(
      \frac{\left(p_m(x)\right)_i/D_{ii}}{\sum_{j \in [K]} \left(p_m(x)\right)_j/D_{jj}}
   \right).   \label{csst_loss:hyb}
\end{equation}
To make the dependence of $\bG$ explicit, we also denote $\hybloss(y, p_m(x))$ as $ \hybloss(y, p_m(x); \bG)$.
The average loss on training sample $S \subset \mX$ is defined as 
$\mathcal{L}^{\mathrm{hyb}}(\mX) = \frac{1}{| S |}\sum_{x \in S}{\ell^{\mathrm{hyb}}(u, p_m(x))}.$
\citet{narasimhan2021training} proved that the hybrid loss is calibrated,
that is learning with $\hybloss$ gives the Bayes optimal classifier for $\bG$ (c.f., \cite[Proposition 4]{narasimhan2021training}, 
\addedtext{of which we provide a formal statement in Sec. \ref{csst_sec:appendix-formal-statement}}).
If $\bG$ is a diagonal matrix (i.e., $\bM = \bone_K$), 
the hybrid loss is called the logit adjusted (LA) loss
and $\hybloss(y, p_m(x))$ is denoted by $\laloss(y, p_m(x))$.

\subsection{Consistency Regularizer for Semi-Supervised Learning}
Modern self-training methods not only leverage pseudo labels, but also 
forces consistent predictions of a classifier on augmented examples or neighbor examples
\cite{wei2020theoretical,miyato2018virtual,xie2020unsupervised,sohn2020fixmatch}.
More formally, a classifier $F$ is trained so that the consistent regularizer $R(F)$ is small 
while a supervised loss or a loss between pseudo labeler are minimized \cite{wei2020theoretical,sohn2020fixmatch}.
Here the consistency regularizer $R(F)$ is defined as
\begin{equation*}
   \ex[x]{\indc(F(x) \neq F(x'), 
   \exists x' \text{ s.t. } x'\text{ is a neighbor of an  augmentation of } x)}.
\end{equation*}
In existing works, consistency regularizers are considered for optimization of accuracy.
In the subsequent sections,
we consider consistency regularizers for cost-sensitive objectives.

\section{Cost-Sensitive Self-Training \ for Non-Decomposable Metrics}
\label{csst_sec:theoretical_results}
\subsection{CSL and Weighted Error}
\label{csst_sec:csl-weighted-err}
In the case of accuracy or $\zo$-error,
a self-training based SSL algorithm using a consistency regularizer achieves the state-of-the-art performance
across a variety of datasets
\cite{sohn2020fixmatch} and 
its effectiveness has been proved theoretically \cite{wei2020theoretical}.
This section provides theoretical analysis of a self-training based SSL algorithm for non-decomposable objectives
by generalizing \cite{wei2020theoretical}.
More precisely, 
the main result of this section (Theorem \ref{csst_thm:main-err-bd}) states that
an SSL method using consistency regularizer improves a given pseudo labeler for non-decomposable objectives.
We provide all the omitted proofs in Appendix for theoretical results in the chapter.

In Sec. \ref{csst_sec:preliminaries}, we considered non-decomposable metrics and their reduction to
cost-sensitive learning objectives defined by Eq. \eqref{csst_eq:csl-obj} using a gain matrix.
In this section, we consider an equivalent objective using the notion of weighted error.
For weight matrix $w = (w_{ij})_{1 \le i, j\le K}$ 
and a classifier $F : \mX \rightarrow [K]$,
a weighted error is defined as follows:
\begin{align*}
\err_w(F) = \sum_{i, j \in [K]}w_{ij} \ex[x \sim P_i]{\indc (F(x) \ne j)},
\end{align*}
where, $P_i(x)$ denotes the class conditional distribution $\prob(x \mid y = i)$.
If $w = \diag(1/K, \dots, 1/K)$, then this coincides with the usual balanced error \cite{menon2020long}.
Since
\begin{math}
    C_{ij}[F] = \ex[(x, y) \sim D]{\indc(y = i, F(x) = j)}
    = \prob(y = i) - \prob(y = i)\ex[x \sim P_i]{\indc(F(x) \ne j)},
\end{math} we can write:
\begin{equation*}
    G_{ij} C_{ij}[F] = G_{ij} (\prob(y = i) - \prob(y = i)\ex[x \sim P_i]{\indc(F(x) \ne j)}) = G_{ij} (\pi_{i} - \pi_{i} \ex[x \sim P_i]{\indc (F(x) \ne j)})
\end{equation*}
Here $\pi_i$ is the class prior $\prob(y = i)$ for $1\le i \le K$ as before. Hence  CSL objective (\ref{csst_eq:csl-obj}) i.e. $\max_{F} \sum_{i,j} G_{ij} C_{ij}[F]$ is equivalent to minimizing the negative term above i.e.
$ G_{ij} \pi_{i} \ex[x \sim P_i]{\indc (F(x) \ne j)}$ which is same as $\err_{w}(F)$ with 
$w_{ij} = G_{ij} \pi_i$ for $1 \le i, j \le K$. Hence, the \emph{notion of weighted error is equivalent to CSL}, which we will also use later for deriving loss functions. We further note that if we add a matrix with the same columns ($c\indc \geq 0$)  to the gain matrix $\bG$, still
the maximizers of CSL \eqref{csst_eq:csl-obj} are the same as the original problem.
Hence, without loss of generality, we assume $w_{ij}\ge 0$.
We assume $w \ne \bm{0}$, i.e., $|w|_1 > 0$ for avoiding degenerate solutions.

\begin{figure}[!t]
    \centering 
    \includegraphics[width=\linewidth]{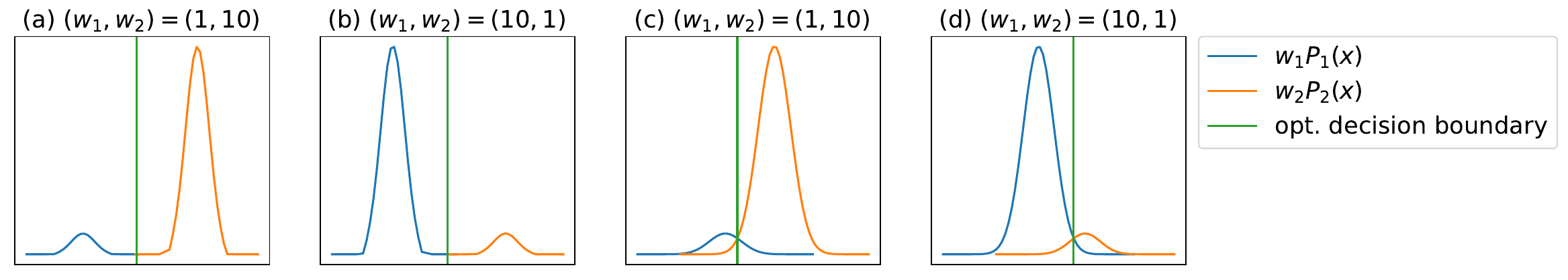}
    \caption{
    Using a simple example, we explain a difference in theoretical assumptions compared to 
    \cite{wei2020theoretical} that assumes $\{P_i\}_{1\le i \le K}$ have disjoint supports.
    Here, we consider the case when $K=2$, $w = \diag(w_1, w_2)$, and 
    $P_1$ and $P_2$ are distributions on $\mX \subset \RR$.
            (a), (b): In a perfect setting where two distributions $P_1$ and $P_2$ have disjoint supports, 
the Bayes optimal classifier for the CSL is identical to the ground truth classifier 
($x \mapsto \argmax_i P_i(x)$) for any choices of weights $(w_1, w_2)$.
(c), (d): 
In a more generalized setting, 
the Bayes optimal classifier $x \mapsto \argmax_{i}w_i P_i(x)$ depends on the choice of weights (i.e., gain matrix). 
The optimal decision boundary (the green line) for the CSL moves to the right
as $w_1/w_2$ increases.
    }
    \label{csst_fig:disjoint_non_disjoint_supp}
\end{figure}

In the previous work \cite{wei2020theoretical}, it is assumed that 
there exists a ground truth classifier $F^\star: \mX \rightarrow [K]$
and the supports of distributions $\{P_i\}_{1 \le i \le K}$ are disjoint. 
However, if supports \addedtext{of distributions $\{P_i\}_{1 \le i \le K}$} are disjoint, a solution of the minimization problem $\min_F \err_w(F)$
is independent of $w$ in some cases.
More precisely, if $w = \diag(w_1, \dots, w_K)$ i.e. a diagonal matrix and $w_i > 0, \forall i$, 
then the optimal classifier is given as $x \mapsto \argmax_{k\in[K]} w_k P_k(x)$ 
(this follows from \cite[Proposition 1]{narasimhan2021training}).
If supports are disjoint, 
then the optimal classifier is the same as $x \mapsto \argmax_{k \in [K]} P_k(x)$, which coincides with the 
ground truth classifier.
Therefore, we do not assume the supports of $P_i$ are disjoint nor a ground truth classifier exists unlike \cite{wei2020theoretical}.
See Fig. \ref{csst_fig:disjoint_non_disjoint_supp} for an intuitive explanation.

\subsection{Weighted Consistency Regularizer}
\label{csst_subsec:weighted-consistency-def}
For improving accuracy, the consistency in prediction is equally important across the distributions $P_i$
for $1 \le i \le K$ \cite{sohn2020fixmatch,wei2020theoretical}.
However, for our case of weighted error, 
if the entries of the $i_0^{th}$ row of the weight matrix $w$ are larger
than the other entries for some $i_0$,
then the consistency of model predictions on examples drawn from the distribution $P_{i_0}$ are more important 
than those on the other examples. 
In this case, we require more restrictive consistency regularizer 
for distribution for $P_ {i_0}$.
Thus, we require a weighted (or cost-sensitive) consistency regularizer, which we define below.

We assume that the instance space $\mX$ is a normed vector space with norm $|\cdot|$
and $\mT$ is a set of augmentations, i.e., each $T \in \mT$ is a map from $\mX$ to itself.
For a fixed $r > 0$, we define $\mB(x)$ by 
$\{x' \in \mX : \exists T \in \mT \text{ s.t. } | x' - T(x)| \le r\}$.
For each $i \in [K]$, we define conditional consistency regularizer by 
\begin{math}
    R_{\mB, i}(F) = \ex[x\sim P_i]{\indc\left(\exists x' \in \mB(x) \text{ s.t. } F(x) \ne F(x') \right)}.
\end{math}
Then, we define the weighted consistency regularizer by  %
\begin{math}
    R_{\mB, w}(F) = \sum_{i, j \in [K]}w_{ij} R_{\mB, i}(F).
\end{math}
\addedtext{If the prediction of classifier $F$ is robust to augmentation $T \in \mT$ and small noise, then $R_{\mB, w}(F)$ is small.}
For $\regub > 0$, we consider the following optimization objective for finding a classifier $F$:
\begin{equation}
    \label{csst_eq:cons-cnst-obj}
    \min_{F}\err_w(F) \quad \text{subject to } R_{\mB, w}(F) \le \regub.
\end{equation}
A solution of the problem \eqref{csst_eq:cons-cnst-obj} is denoted by $\gstar$.
\addedtext{We let $\gpl: \mX \rightarrow [K]$ a pseudo labeler (a classifier) with reasonable performance (we elaborate on this in Section \ref{csst_subsec:weighted_consistency}).}
The following mathematically informal assumption below is required to interpret our main theorem.
\begin{assumption}
    \label{csst_assump:mu-small}
    We assume both $\regub$ and $\err_w(\gstar)$ are sufficiently small so that they are 
    negligible compared to $\err_w(\gpl)$.
\end{assumption}
\addedtext{Assumption 1 assumes existence of an optimal classifier $F^*$ that  minimizes the error $\err_w(F)$ (i.e. Bayes Optimal) among the class of classifiers which are robust to data augmentation (i.e. low weighted consistency $R_{\mB, w}(F)$). As we operate in case of overparameterized neural networks such a classifier $F^*$ is bound to exist, but is unknown in our problem setup.}
In the case of the balanced error, 
the validity of this assumption is justified by the fact that 
the existing work \cite{sohn2020fixmatch} using consistency regularizer on data augmentation obtains classifier $F$ , that achieves 
high accuracy (i.e., low balanced errors) for balanced datasets.
Also in Appendix \ref{csst_sec:appendix-examples-for-theoretical}, we provide an example that supports the validity of the assumption in the case of 
Gaussian mixtures and diagonal weight matrices.

\subsection{Expansion Property}
For $x \in \mX$, we define the neighborhood $\mN(x)$ of $x$ by $\{x' \in \mX: \mB(x) \cap \mB(x') \ne \emptyset\}$.
For a subset $S \subseteq \mX$, neighborhood of $S$ is defined as $\mN(S) = \cup_{x \in S} \mN(x)$.
Similarly to \cite{wei2020theoretical}, we consider the following property on distributions.
\begin{definition}
    \label{csst_def:c-exp-def}
    \addedtext{Let $c: (0, 1] \rightarrow [1, \infty)$ be a non-increasing function.}
    For a distribution $Q$ on $\mX$ %
    we say $Q$ has $c$-expansion property if
    $Q(\mN(S)) \ge c(Q(S)) Q(S)$ for any measurable $S \subseteq \mX$.
\end{definition}

The $c$-expansion property implies that if $Q(S)$ decreases, then the 
``expansion factor'' ${Q(\mN(S))}/{Q(S)}$ increases. 
This is a natural condition, because it roughly requires that 
if $Q(S)$ is small, then $Q(\mN(S))$ is large compared to $Q(S)$. \addedtext{For intuition 
let us consider a ball of radius $l$ depicting $S \subset \RR^d$ with volume $Q(S)$ 
and its neighborhood $\mN(S)$ expands to a ball with radius $l+1$. 
The expansion factor here would be $((l + 1)/l)^d$, hence as $l$  (i.e. $Q(S)$) 
increases $(1 + 1/l)^d$ (i.e. ${Q(\mN(S))}/{Q(S)}$) decreases. 
Hence, it's natural to expect $c$ to be a non-decreasing function.}
\modifiedtext{The $c$-expansion property (on each $P_i$) considered here is equivalent to the $(a, \widetilde{c})$-expansion property,
which is shown to be realistic for vision and used for 
theoretical analysis of self-training in \cite{wei2020theoretical}, where $a \in (0, 1)$ and $\widetilde{c} > 1$
(see Sec. \ref{csst_sec:appendix-a-c-expansion} in Appendix).} 
In addition, we also show that it is also satisfied for mixtures of Gaussians and 
mixtures of manifolds (see Example \ref{csst_exa:c-exp-mix-gauss} in Appendix for more details). 
Thus, the $c$-expansion property is a general property satisfied for a wide class of distributions.
\begin{assumption}
    \label{csst_assump:c-exp}
    For weighted probability measure $\Pw$ on $\mX$ by 
    \begin{math}
        \Pw(U) = 
        \frac{\sum_{i, j \in [K]}w_{ij}P_i(U)}{\sum_{i,j \in [K]} w_{ij}}
    \end{math}
    for $U \subseteq \mX$.
    We assume $\Pw$ satisfies $c$-expansion
    for a non-increasing function $c: (0, 1] \rightarrow [1, \infty)$.
\end{assumption}

\subsection{Cost-Sensitive Self-Training with Weighted Consistency Regularizer}
In this section we first introduce the assumptions on the pseudo labeler $\gpl$ and then introduce the theoretical Cost-Sensitive Self-Training (\texttt{CSST}) objective.
\label{csst_subsec:weighted_consistency}
$\gpl$ can be any classifier satisfying the following assumption, however, typically it is a classifier trained on a labeled dataset.
\begin{assumption}
    \label{csst_assump:pseudo-labeler}
    We assume that $\err_w(\gpl) + \err_w(\gstar) \le |w|_1$.
    Let $\gamma = c(p_w)$, where 
    \begin{math}
    p_w = \frac{\err_w(\gpl) + \err_w(\gstar)}{|w|_1}. 
    \end{math}
    We also assume $\gamma > 3$.
\end{assumption}
Since $c$ is non-increasing, $\gamma$ (as a function of $\err_w(\gpl)$) is a non-increasing function of $\err_w(\gpl)$ (and $\err_w(\gstar)$).
Therefore, the assumption $\gamma > 3$ roughly requires that $\err_w(\gpl)$ is ``small''.
We provide concrete conditions for $\err_w(\gpl)$ that satisfy $\gamma > 3$ 
in the case of mixture of isotropic $d$-dimenional Gaussians for a region $\mB(x)$ defined by $r$ in 
Appendix \addedtext{ (Example \ref{csst_exa:gamma-3})}.
In the example, we show that 
the condition $\gamma > 3$ 
is satisfied if $\err_w(\gpl) < 0.17$
in the case when $r = 1/(2\sqrt{d})$
and satisfied if $\err_w(\gpl) < 0.33$ in the case when $r = 3/(2\sqrt{d})$, where $\mX \subseteq \RR^d$.
Since we assume $\err_w(\gstar)$ is negligible compared to $\err_w(\gpl)$ (Assumption \ref{csst_assump:mu-small}),
the former condition in Assumption \ref{csst_assump:pseudo-labeler} is approximately equivalent to 
$\err_w(\gpl) \le |w|_1$ which is satisfied by the definition of $\err_w$.

We define $L^{(i)}_{\zo}(F, F') = \ex[x \sim P_i]{\indc(F(x) \ne F'(x))}$.
Then, we consider the following objective:
\begin{equation}
    \label{csst_eq:ssl-non-decomp-obj-theoretical}
    \min_F \mL_w(F), \quad 
    \text{where } \mL_w(F) =\frac{\gamma + 1}{\gamma - 1}L_w(F, \gpl) + \frac{2\gamma}{\gamma - 1}R_{\mB, w}(F).
\end{equation}
Here $L_w(F, \gpl)$ is defined as 
\begin{math}
    \sum_{i, j \in [K]}w_{ij} L^{(i)}_\zo(F, \gpl).
\end{math}
The above objective corresponds to cost-sensitive self-training (with $\gpl$) objective with weighted consistency regularization. We provide following theorem which relates the weighted error of classifier $\hat{F}$ learnt using the above objective to the weighted error of the pseudo labeler ($\gpl$).
\begin{theorem}
    \label{csst_thm:main-err-bd}
    \modifiedtext{Any learnt classifier $\widehat{F}$ using the loss function $\mL_w$
    (i.e.,  $\argmin_F\mL_w(F)$)} satisfies:
    \begin{align*}
        \err_w(\widehat{F}) 
        &\le
        \frac{2}{\gamma - 1}
        \err_w(\gpl)
        + \frac{\gamma + 1}{\gamma - 1} \err_w(\gstar)
        + \frac{4\gamma}{\gamma - 1} R_{\mB, w}(\gstar).
    \end{align*}
\end{theorem}
\begin{remark}
    Since both $ \err_w(\gstar)$ and $R_{\mB, w}(\gstar) \le \regub$ are negligible compared to $\err_w(\gpl)$ 
    and $\gamma > 3$,
    Theorem~\ref{csst_thm:main-err-bd} asserts that \addedtext{$\hat{F}$} learnt 
    by minimizing semi-supervised loss $L_w(F, \gpl)$ with the consistency regularizer $R_{\mB, w}(F)$ can 
    achieve superior performance than the pseudo labeler \addedtext{$\gpl$}
    in terms of the weighted error $\err_w$. The above theorem  is a generalization of \cite[Theorem 4.3]{wei2020theoretical}, which provided
a similar result for balanced $\zo$-error in the case of distributions with disjoint supports.
    In Appendix \addedtext{Sec. \ref{csst_sec:appendix-all-layer}}, following \cite{wei2020theoretical,wei2019improved}, we also provide a generalization bound for $\err_w(F)$ using all-layer margin \cite{wei2019improved}
    in the case when classifiers are neural networks.
\end{remark}

\section{\texttt{CSST} in Practice}

In the previous section, we proved that by using self-training (\texttt{CSST}), we can achieve a superior classifier $\hat{F}$ in comparison to pseudo labeler $\gpl$ through weighted consistency regularization. As we have established the equivalence of the weighted error $Err_{w}$ to the CSL objective expressed in terms of $\bG$ (Sec. \ref{csst_sec:csl-weighted-err}) , we can theoretically optimize a given non-decomposable metric expressed by $\bG$ better using \ttt{CSST}, utilizing the additional unlabeled data via self-training and weighted consistency regularization.  We now show how \ttt{CSST} can be used in practice for optimizing non-decomposable metrics in the case of neural networks.

The practical self-training methods utilizing consistency regularization (e.g., FixMatch~\cite{sohn2020fixmatch}, etc.) for semi-supervised learning have supervised loss $\mathcal{L}_s$ for labeled and consistency regularization loss for unlabeled samples (i.e., $\mathcal{L}_u$) with a thresholding mechanism to select unlabeled samples. The final loss for training the network is $\mathcal{L}_s + \lambda_{u}\mathcal{L}_u$, where $\lambda_{u}$ is the hyperparameter. The supervised loss $\mathcal{L}_s$ can be modified conveniently based on the desired non-decomposable metric by using suitable $\bG$ (Sec. \ref{csst_sec:ndo-loss-fun}). We will now introduce the novel weighted consistency loss and its corresponding thresholding mechanism for unlabeled data in \ttt{CSST}, used for optimizing desired non-decomposable metric.   

\vspace{1mm} \noindent \textbf{Weighted Consistency Regularization.}
As the idea of consistency regularization is to enforce consistency between model prediction on different augmentations of input, this is usually achieved by minimizing some kind of divergence $\mathcal{D}$. A lot of recent works~\cite{miyato2018virtual, sohn2020fixmatch,xie2020unsupervised} in semi-supervised learning  use $\mathcal{D}_{\mathrm{KL}}$ to enforce consistency between the model's prediction on unlabeled data and its augmentations, $p_{m}(x)$ and $p_{m}(\mathcal{A}(x))$. Here $\mathcal{A}$ usually denotes a form of strong augmentation. Across these works, the distribution of confidence of the model's prediction is either sharpened or used to get a hard pseudo label to obtain $\hat{p}_{m}(x)$. As we aim to optimized the cost-sensitive learning objective, we aim to match the distribution of normalized distribution (i.e. $\norm(\bG^{\mathbf{T}}\hat{p}_{m}(x)) = \bG^{\mathbf{T}}\hat{p}_{m}(x)/\sum_i{(\bG^{\mathbf{T}}\hat{p}_{m}(x))_i}$) (Proposition 2~\cite{narasimhan2021training} also in Prop.  \ref{csst_prop:prop2-narashimhan} ) with the $p_{m}(\mathcal{A}(x))$ by minimizing the KL-Divergence between these. We now propose to use the following weighted consistency regularizer loss function for optimizing the same:
\begin{equation}
  \label{csst_eq:wt-consistency-reg}
    \ell^{\mathrm{wt}}_{u}(\hat{p}_{m}(x), p_{m}(\mathcal{A}(x)), \bG) = -\sum_{i=1}^{K}(\bG^{\mathbf{T}}\hat{p}_{m}(x))_i \log(p_{m}(\mathcal{A}(x))_i)
\end{equation}
\begin{proposition}
The minimizer of $\mathcal{L}^{wt}_{u} =\frac{1}{|B_u|}\sum_{x \in B_u} \ell^{wt}_{u}(\hat{p}_{m}(x), p_{m}(\mathcal{A}(x)), \bG)$ 
leads to minimization of KL Divergence i.e. $\mathcal{D}_{KL}(\norm(\bG^{\mathbf{T}}\hat{p}_{m}(x)) || p_{m}(\mathcal{A}(x))) \; \forall x \in \mX $  .
\label{csst_prop:kl-weighted}
\end{proposition}
As the above loss is similar in nature to the cost sensitive losses introduced by~\citet{narasimhan2021training} (Sec. \ref{csst_sec:ndo-loss-fun}) we can use the logit-adjusted variants (i.e. $\ell^{\mathrm{LA}}$ and $\ell^{\mathrm{hyb}}$ based on type of $\bm{G}$) of these in our final loss formulations ($\mathcal{L}^{wt}_{u}$) for training overparameterized deep networks.  We further show in Appendix Sec. \ref{csst_sec:appendix-conn-between} that the above loss $\ell^{\mathrm{wt}}_{u}$ approximately minimizes the theoretical weighted consistency regularization term $R_{\mB, w}(F)$ defined in Sec. \ref{csst_subsec:weighted_consistency}.

\vspace{1mm} \noindent \textbf{Threshold Mechanism for \ttt{CSST}.}
\label{csst_subsec:threshold}
In the usual semi-supervised learning formulation~\cite{sohn2020fixmatch} we use the confidence threshold ($\max_{i}(p_{m}(x)_i) > \tau$) as the function to select samples for which consistency regularization term is non-zero. We find that this leads to sub-optimal results in particularly the case of non-diagonal $\mathbf{G}$ as only a few samples cross the threshold (Fig. \ref{csst_fig:mask_rate}).  As in the case of cost-sensitive loss formulation the samples may not achieve the high confidence to cross the threshold of consistency regularization. This is also theoretically justified by the following proposition:  
\begin{proposition}[\cite{narasimhan2021training} Proposition 2]
   \label{csst_prop:prop2-narashimhan}
   Let $p_{m}^{opt}(x)$ be the optimal softmax model function obtained by optimizing the cost-sensitive objective in Eq. \eqref{csst_eq:csl-obj} by averaging weighted loss function $\ell^{wt}(y,p_m(x)) = - \sum_{i=1}^{K} G_{y,i} \log{\frac{(p_m(x)_i)}{\sum_{j}p_m(x)_j}}$. Then optimal $p_{m}^{opt}(x)$ is: $p_{m}^{opt}(x) = \frac{G_{y,i}}{\sum_{j} G_{y,j}} = \norm(\bG^{T}\mathbf{y})  \forall (x,y)$.
\end{proposition}
Here $\mathbf{y}$ is the one-hot representation vector for a label $y$. This proposition demonstrates that for a particular sample the high confidence $p_{m}(x)$ may not be optimal based on $\bG$. We now propose our novel way of thresholding samples for which consistency regularization is applied in \texttt{CSST}. Our thresholding method takes into account the objective of optimizing the non-decomposable metric by taking $\bG$ into account. We propose to use the threshold on KL-Divergence of the softmax of the sample $ {p}_{m}(x)$ with the optimal softmax (i.e. $\norm(\bG^{T}\hat{{p}}_{m}(x))$) for a given $\mathbf{G}$ corresponding to the pseudo label (or sharpened) $\hat{{p}}_{m}(x)$, using which we modify the consistency regularization loss term:
\begin{equation}
    \label{csst_loss:CSST-KL}
    \mathcal{L}_u^{wt}(B_u) = \frac{1}{|B_u|}\sum_{x \in B_u}\indc_{(\mathcal{D}_{KL} (\norm(\bG^{T}\hat{p}_{m}(x))  \; || \; p_m(x))\leq \tau)}
     \ell^{\mathrm{wt}}_{u}(\hat{p}_{m}(x), p_{m}(\mathcal{A}(x)), \bG).
\end{equation}
We name this proposed combination of KL-Thresholding and weighted consistency regularization as $\texttt{CSST}$ in our experimental results. We find that for non-diagonal gain matrix $\bG$ the proposed thresholding plays a major role in improving performance over supervised learning. This is demonstrated by comparison of $\ttt{CSST}$ and \ttt{CSST} w/o KL-Thresholding (without proposed thresholding mechanism) in Fig. \ref{csst_fig:mask_rate} and Tab. \ref{csst_tab:coverage}. We will now empirically incorporate \ttt{CSST} by introducing consistency based losses and thresholding mechanism for unlabeled data, into the popular semi-supervised methods of FixMatch~\cite{sohn2020fixmatch} and Unsupervised Data Augmentation for Consistency Training (UDA)~\cite{xie2020unsupervised}. The exact expression for the weighted consistency losses utilized for UDA and FixMatch have been provided in the Appendix \addedtext{ Sec. \ref{csst_sec:appendix-fixmatch-obj} and \ref{csst_sec:appendix-uda-obj}} .

\label{csst_sec:proposed-method}

\section{Experiments}
\label{csst_sec:expt}

\begin{figure*}[!t]
  \centering
\begin{minipage}[c]{0.68\textwidth}
    \centering
    \captionof{table}{Results of maximizing the worst-case recall over 
        \emph{all classes} (col 2--3) and
        over just the head and tail classes (col 4--7). }
    \label{csst_tab:min-recall}
    \centering
        \adjustbox{width=\textwidth}{
            \begin{tabular}{ccccccccc}
            \hline
            \multicolumn{1}{c}{\textbf{Method}} & 
            \multicolumn{2}{c}{\textbf{CIFAR10-LT ($\rho = 100$)}} & 
            \multicolumn{2}{c}{\textbf{CIFAR100-LT ($\rho = 10$)}} &
            \multicolumn{2}{c}{\textbf{Imagenet100-LT ($\rho = 10$)}} &
            \\
            & \textbf{Avg. Rec} & \textbf{Min. Rec} & 
            \textbf{Avg. Rec } & \textbf{Min. HT Rec} &
            \textbf{Avg. Rec } & \textbf{Min. HT Rec} &
            \\
            \hline
            \texttt{ERM} &  
                0.52 &	0.26 & 
                0.36 &	0.14 & 0.40 & 0.30
                \\
            \texttt{LA} &  
                0.51 &	0.38 & 
                0.36 &	0.35 & 0.48 & 0.47
                \\
            \texttt{CSL} &  
                0.64 &	0.57 & 
                0.43 &	0.43 & 0.52 & 0.52
                \\
            \hline
            \begin{tabular}{@{}c@{}}\texttt{Vanilla} \\ (FixMatch)\end{tabular}    
            \
                & \highlight{0.78} &	0.48 &
                \highlight{0.63} &	0.36 & 0.58 & 0.49
                \\
            \begin{tabular}{@{}c@{}}\texttt{CSST} \\ (FixMatch)\end{tabular}
                & 0.76 &	\highlight{0.72} &
                \highlight{0.63} &	\highlight{0.61} & \highlight{0.64} & \highlight{0.63}
                \\
            \hline
        
        \end{tabular}}
  \end{minipage}  
\begin{minipage}[c]{0.3\linewidth}
    \centering
    \includegraphics[width=\linewidth]{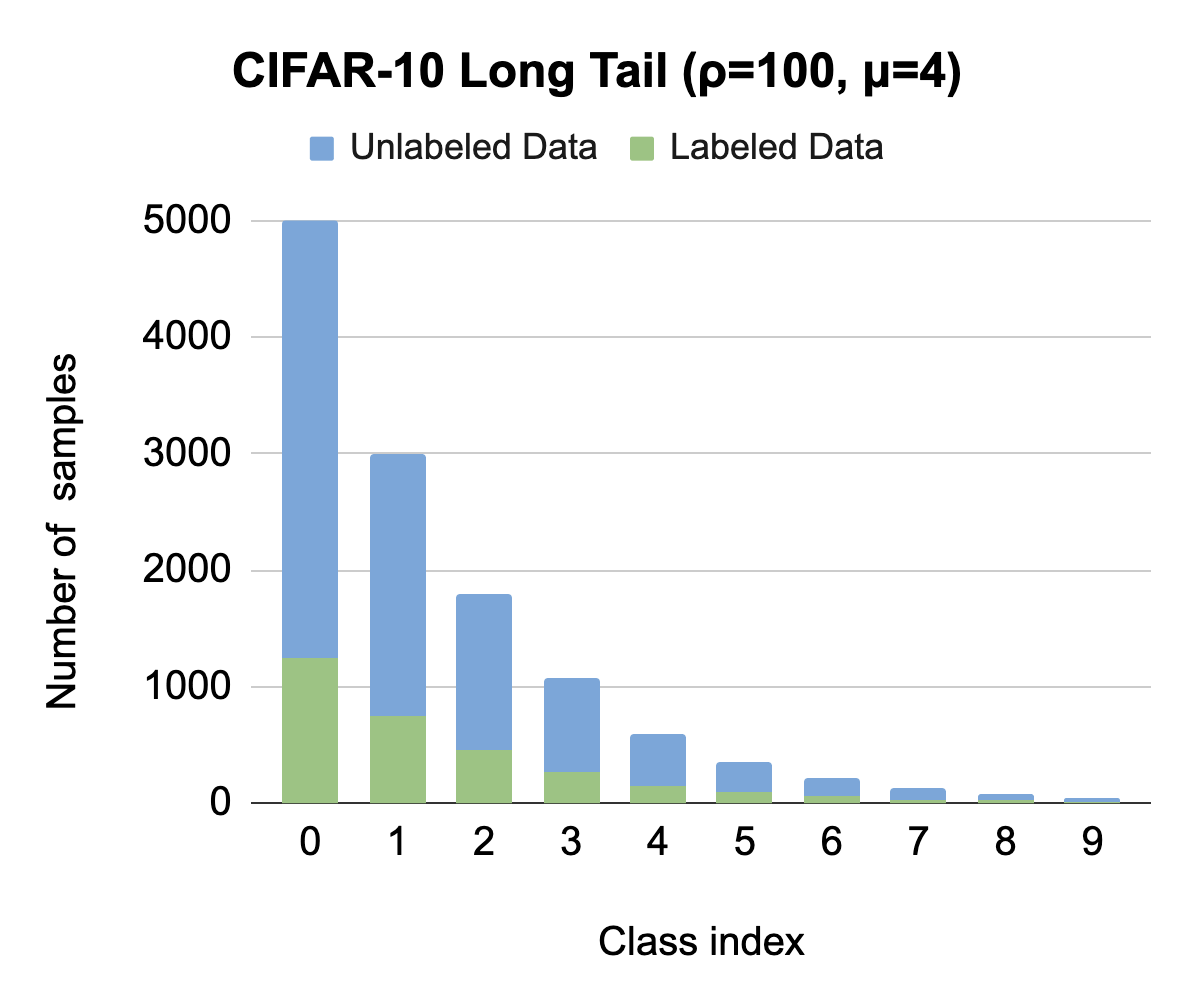}
    \captionof{figure}{CIFAR-10 Long tail distribution $\rho=100,\mu=4$.}
    \label{csst_fig:intra-fid}
  \end{minipage}

\end{figure*}
We demonstrate that the proposed \ttt{CSST} framework shows significant gains in performance on both vision and NLP tasks on imbalanced datasets, with an imbalance ratio defined on the training set as $\rho = \frac{\max_i P(y=i)}{\min_i P(y=i)}$. We assume the labeled and unlabeled samples come from a similar data distribution and the unlabeled samples are much more abundant ($\mu$ times) the labeled. The frequency of samples follows an exponentially decaying long-tailed distributed as seen in  Fig.~\ref{csst_fig:intra-fid}, which closely imitates the distribution of real-world data~\cite{thomee2016yfcc100m, krishna2017visual}.
 For CIFAR-10~\cite{krizhevsky2009learning}, IMDb~\cite{maas-EtAl:2011:ACL-HLT2011} and DBpedia-14~\cite{lehmann2015dbpedia}, we use $\rho = 100$ and $\rho = 10$ for CIFAR-100~\cite{krizhevsky2009learning} and ImageNet-100~\cite{russakovsky2015imagenet}~\footnote{https://www.kaggle.com/datasets/ambityga/imagenet100} datasets.
We compare our method against supervised methods of ERM, Logit Adjustment (\texttt{LA})~\cite{menon2020long} and Cost Sensitive Learning (\texttt{CSL})~\cite{narasimhan2014statistical} trained on the same number of labeled samples as used by semi-supervised learning methods, along with vanilla semi-supervised methods of FixMatch (for vision) and UDA (for NLP tasks). We use WideResNets(WRN) \cite{BMVC2016_87}, specifically WRN-28-2 and WRN-28-8 for CIFAR-10 and CIFAR-100 respectively. For ImageNet, we use a ResNet-50~\cite{he2016deep} network for our experiments and finetuned DistilBERT(base uncased)~\cite{sanh2019distilbert} for IMDb and DBpedia-14 datasets. We divide the balanced held-out set for each dataset equally into validation and test sets.  A detailed list of hyper-parameters and additional experiments can be found in the Appendix \addedtext{Tab. \ref{csst_tab:hyperparams}} and Sec. \ref{csst_sec: additional-experiments} respectively.

\label{csst_subsec:worst-case-recall}
For CIFAR-10, IMDb, and DBpedia-14 datasets, we maximize the minimum recall among all classes (Eq. \eqref{csst_eq:min-recall-obj}).
Given the low number of samples per class for datasets with larger number of classes like CIFAR-100 and ImageNet-100, we pick objective \eqref{csst_eq:min-HT-recall-obj}. We define the head classes ($\mH$) and tail classes ($\mT$) as the first 90 classes and last 10 classes respectively. 
The Min. HT recall objective can be mathematically formulated as: 
\begin{equation}
    \label{csst_eq:min-HT-recall-obj}
  \max_F \min_{(\lambda_{\mathcal{H}}, \lambda_{\mathcal{T}}) \in \Delta_{1}} \frac{\lambda_{\mathcal{H}}}{|\mathcal{H}|} \sum_{i \in \mathcal{H}}\frac{C_{ii}[F]}{\pi_i} + \frac{\lambda_{\mathcal{T}}}{|\mathcal{T}|} \sum_{i \in \mathcal{T}}\frac{C_{ii}[F]}{\pi_i}.
\end{equation}
The corresponding gain matrix $\bf{G}$ is diag($\frac{\lambda_{\mH}}{\pi_1 |\mH|}, \frac{\lambda_{\mH}}{\pi_2 |\mH|}, \dots , \frac{\lambda_{\mT}}{\pi_{K-1} |\mT|}, \frac{\lambda_{\mT}}{\pi_{K} |\mT|}$). 
Since $\bf{G}$ is diagonal here, we use \texttt{CSST}(FixMatch) loss function Eq. \eqref{csst_loss:CSST-KL} with the corresponding $\ell_{u}^{\mathrm{wt}}$ being substituted by $\ell_{u}^{\mathrm{LA}}$ as define in Sec. \ref{csst_sec:ndo-loss-fun}. Also for labeled samples we use $\mathcal{L}_{s}^{\mathrm{LA}}$ as $\bG$ is diagonal, we then combine the loss and train network using SGD. Each few steps of SGD, were followed by an update on the $\blambda$ and $\bG$ based on the uniform validation set (See Alg. \textcolor{red}{1} in Appendix).  We find that \texttt{CSST}(FixMatch) significantly outperforms the other baselines in terms of the  Min. recall and Min. Head-Tail recall for all datasets, the metrics which we aimed to optimize (Tab. \ref{csst_tab:min-recall}), which shows effectiveness of \ttt{CSST} framework. 
Despite optimizing worst-case recall we find that our framework is still able to maintain reasonable average (Avg.) recall in comparison to baseline \ttt{vanilla}(FixMatch), which demonstrates it's practical applicability. We find that optimizing Min. recall across NLP tasks of classification on long-tailed data by plugging UDA into  \ttt{CSST(UDA)} framework shows similar improvement in performance (Tab. \ref{csst_tab:nlp_results}). This establishes the generality of our framework to even self-training methods across domain of NLP as well. 
As the $\mathbf{G}$ is a diagonal matrix for this objective, the proposed KL-Based Thresholding here is equivalent to the confidence based threshold of FixMatch in this case. Despite the equivalence of thresholding mechanism, we see significant gains in min-recall (Tab. \ref{csst_tab:min-recall}) just using the regularization term. We discuss their equivalance in Appendix Sec. \ref{csst_Diagonal-G-hard-PL}.

\label{csst_tab:post-shift}

\begin{table}[!t]
    \centering
    \small
    \caption{Results of maximizing the mean recall subject to coverage constraint 
    \emph{all classes} (col 2--3) and
    over the head and tail classes (col 4--7). 
   Proposed \texttt{CSST}(FixMatch) approach compares favorably to \ttt{ERM},\ttt{LA},\ttt{CSL} \ttt{vanilla}(FixMatch) and \texttt{CSST}(FixMatch) w/o KL-Thresh.. It is the best at both maximizing mean recall and coming close to satisfying the coverage constraint.
    }
    \adjustbox{width=\textwidth}{
    \begin{tabular}{lcc|cc|cccc}
        \hline
        \multicolumn{1}{c}{\textbf{Method}} & 
        \multicolumn{2}{c}{\textbf{CIFAR10-LT }} & 
        \multicolumn{2}{c}{\textbf{CIFAR100-LT }} &
        \multicolumn{2}{c}{\textbf{ImageNet100-LT }} &
        \\
        & \multicolumn{2}{c}{Per-class Coverage } & 
        \multicolumn{2}{c}{Head-Tail Coverage } &
        \multicolumn{2}{c}{Head-Tail Coverage } &
        \\
        & \multicolumn{2}{c}{($\rho = 100$, tgt : 0.1)} & 
        \multicolumn{2}{c}{($\rho = 10$, tgt : 0.01)} &
        \multicolumn{2}{c}{($\rho = 10$, tgt : 0.01)} &
        \\
        \hline
        & \textbf{Avg. Rec} & \textbf{Min. Cov} & 
        \textbf{Avg. Rec } & \textbf{Min. HT Cov} &
        \textbf{Avg. Rec } & \textbf{Min. HT Cov} &
        \\
        \hline
        \texttt{ERM} &  
            0.52 &	0.034 & 
            0.36 &	0.004 & 0.40 & 0.006
            \\
        \texttt{LA} &  
            0.51 &	0.039 & 
            0.36 &	0.009 & 0.48 & 0.009
            \\
        \texttt{CSL} &  
            0.60 &	0.090 & 
            0.45 &	0.010 & 0.48 & \highlight{0.010}
            \\
        \hline
        \begin{tabular}{@{}c@{}}\texttt{Vanilla} (FixMatch)
        \end{tabular}
            & 0.78 &	0.055 &
            \highlight{0.63} &	0.004 & \highlight{0.58} & 0.007
            \\
        \texttt{CSST}(FixMatch) w/o & & \highlight{} &  & & &  \\
        KL-Thresh.
            & \multirow{-2}{*}{0.67} &	\multirow{-2}{*}{\highlight{0.093}} &
            \multirow{-2}{*}{0.47} &	 \multirow{-2}{*}{0.010} &  \multirow{-2}{*}{0.26} &  \multirow{-2}{*}{0.010}
            \\
        \begin{tabular}{@{}c@{}}\texttt{CSST}(FixMatch) 
        \end{tabular}
            & \highlight{0.80} &	0.092 &
            \highlight{0.63} &	\highlight{0.010} & \highlight{0.58} & \highlight{0.010}
            \\
        \hline
    \end{tabular}}
    \vspace{0.5em}

    \label{csst_tab:coverage}
    
\end{table}

\begin{table}[!t]
    \centering
    \small
    \caption{Results of maximizing the min recall over
    \emph{all classes} for classification on NLP datasets. 
   Proposed \texttt{CSST}(UDA) approach outperforms ERM and \ttt{vanilla}(UDA) baselines.
    }
    \begin{tabular}{lcccccccc}
        \hline
        \multicolumn{1}{c}{\textbf{Method}} & 
        \multicolumn{2}{c}{\textbf{IMDb ($\rho = 10$)}} & 
        \multicolumn{2}{c}{\textbf{IMDb ($\rho = 100$)}} &
        \multicolumn{2}{c}{\textbf{DBpedia-14 ($\rho = 100$)}} &
        \\
        \hline
        & \textbf{Avg Rec} & \textbf{Min Rec} & 
        \textbf{Avg Rec } & \textbf{Min Rec} &
        \textbf{Avg Rec } & \textbf{Min Rec} &
         \\\hline
        \texttt{ERM} &  
            0.79 &	0.61 & 
            0.50 &	0.00 & 0.95 & 0.58
            \\
        \midrule
        \texttt{vanilla(UDA)}
            & 0.82 &	0.66 &
            0.50 &	0.00 & 0.96 & 0.65
            \\
        \texttt{CSST(UDA)}
            & \highlight{0.89} &	\highlight{0.88} &
            \highlight{0.77} &	\highlight{0.75} & \highlight{0.99} & \highlight{0.97}
            \\
        \hline
    \end{tabular}
    \vspace{-5mm}

    \label{csst_tab:nlp_results}
    
\end{table}

\vspace{1mm} \noindent \textbf{Maximizing Mean Recall Under Coverage Constraints.}
\label{csst_subsec:max-recall-under-cov}
Maximizing mean recall under coverage constraints objective seeks to result in a model with good average recall, yet at the same time constraints the proportion of predictions for each class to be uniform across classes.
The ideal target coverage under a balanced evaluation set(or such circumstances) is given as
$ 
\label{csst_perfect-coverage}
\text{cov}_{i}[F] = \frac{1}{K}, \ \forall \  i \in [K]
$. 
Along similar lines to objective \eqref{csst_eq:min-HT-recall-obj} we modify the objective \eqref{csst_eq: cov-const-obj} to maximize the average recall subject to the both the average head and tail class coverage (HT Coverage) being above a given threshold of $\frac{0.95}{K}$. The $\bG$ is non-diagonal here and $G_{ij} = (\indc_{j \in \mathcal{H}}  \frac{\lambda_{\mathcal{H}}}{|\mathcal{H}|} + \indc_{j \in \mathcal{T}} \frac{\lambda_{\mathcal{T}}}{|\mathcal{T}|}) + \frac{\delta_{ij}}{K \pi_i} $.
\begin{equation}
 \label{csst_eq: HT-cov-const-obj}
    \max_F \min_{(\lambda_{\mathcal{H}}, \lambda_{\mathcal{T}}) \in \RR^2_{\ge 0}} 
   \sum_{i\in [K]} \frac{C_{ii}[F]}{K\pi_i} + \lambda_{\mathcal{H}}\left(
      \sum_{i \in [K], j \in \mathcal{H}} \frac{C_{ij}[F]}{|\mathcal{H}|}  - \frac{0.95}{K}
   \right)  + \lambda_{\mathcal{T}}  \left(
      \sum_{i \in [K], j \in \mathcal{T}} \frac{C_{ij}[F]}{|\mathcal{T}|} - \frac{0.95}{K}
   \right).
\end{equation}
As these objectives corresponds to a non-diagonal $\bG$ as shown in Eq. \eqref{csst_eq: cov-const-obj} in Sec. \ref{csst_sec:ndo-loss-fun}. Hence, for introducing $\ttt{CSST}$ into FixMatch we replace first supervised loss $\mathcal{L}_{s}$ with $\mathcal{L}_{s}^{\mathrm{hyb}}$. For the unlabeled data we introduce $\ell^{\mathrm{hyb}}$ in $\mathcal{L}^{\mathrm{wt}}_{u}$ (Eq. \eqref{csst_loss:CSST-KL}).  Hence, the final objective $\mathcal{L}$ is defined as, $\mathcal{L} =  \mathcal{L}_s^{\mathrm{hyb}} + \lambda_u \mathcal{L}^{\mathrm{wt}}_u $. We update the parameters of the cost-sensitive loss ($\bG$ and $\blambda$) periodically after few of SGD on the model parameters (Alg. \textcolor{red}{2} in Appendix). In this case our proposed thresholding mechanism in \ttt{CSST}(FixMatch)  introduced in Sec. \ref{csst_subsec:threshold}, leads to effective utilization of unlabled data resulting in improved performance over the naive \ttt{CSST}(FixMatch) without (w/o) KL-Thresholding (Tab. \ref{csst_tab:coverage}). In these experiments, the mean recall of our proposed approach either improves or stays same to the \ttt{vanilla}(FixMatch) implementation but only ours is the one that comes close to satisfying the coverage constraint. We further note that among all the methods, the only methods that come close to satisfying the coverage constraints are the ones with $\ell^{\mathrm{hyb}}$ included in them.

\begin{figure*}[t]
  \centering
\begin{minipage}[c]{0.34\linewidth}
        \centering
         \includegraphics[width=\textwidth]{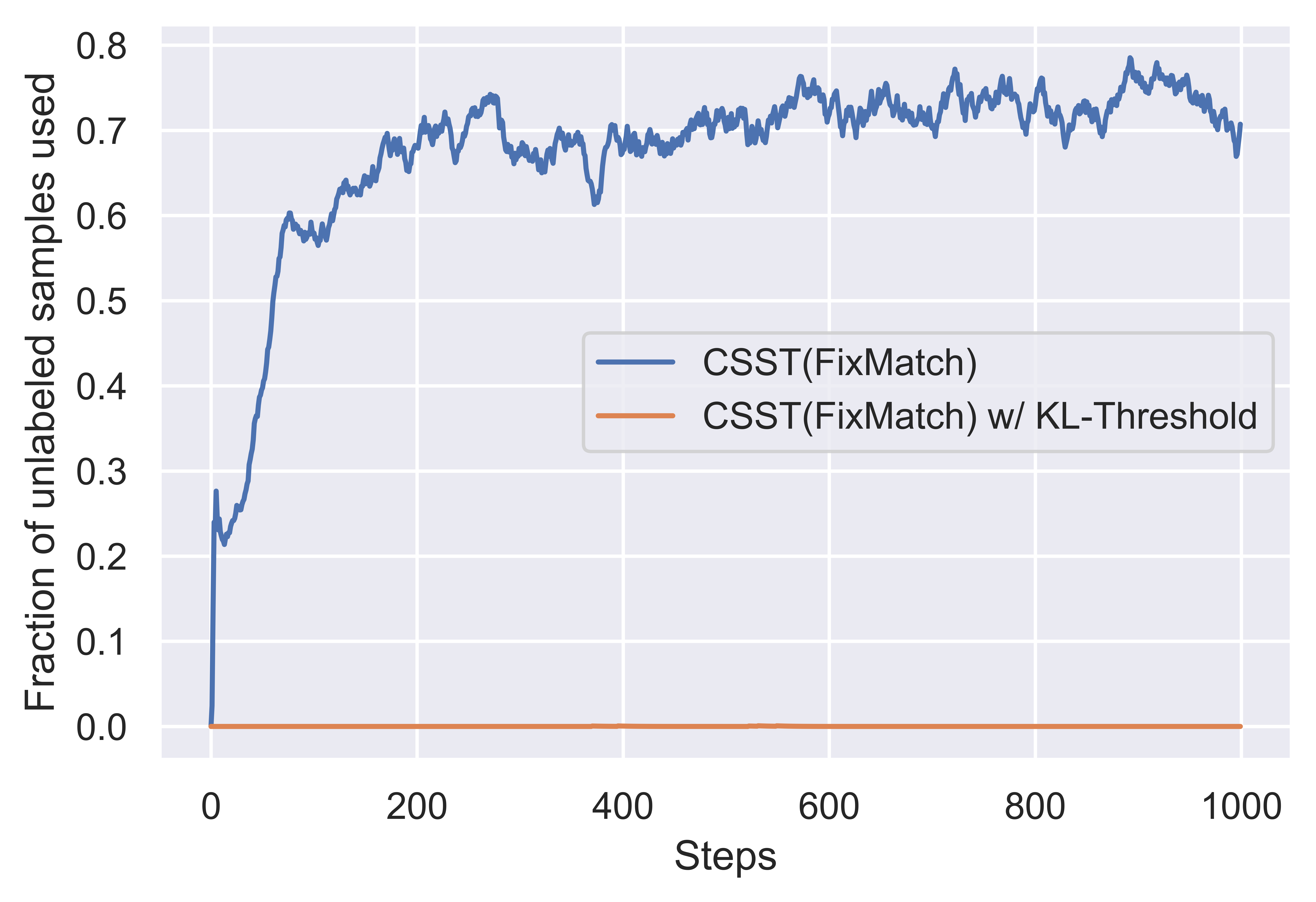}
         \caption{Fraction of unlabelled data used for maximizing average recall under coverage constraints for CIFAR-100 ($\rho=10,\mu=4$) (Sec. \ref{csst_subsec:max-recall-under-cov}).}
         \label{csst_fig:mask_rate}
  \end{minipage}
   \hfill
  \begin{minipage}[c]{0.64\textwidth}

\begin{subfigure}{.5\textwidth}
  \centering
  \includegraphics[width=\linewidth]{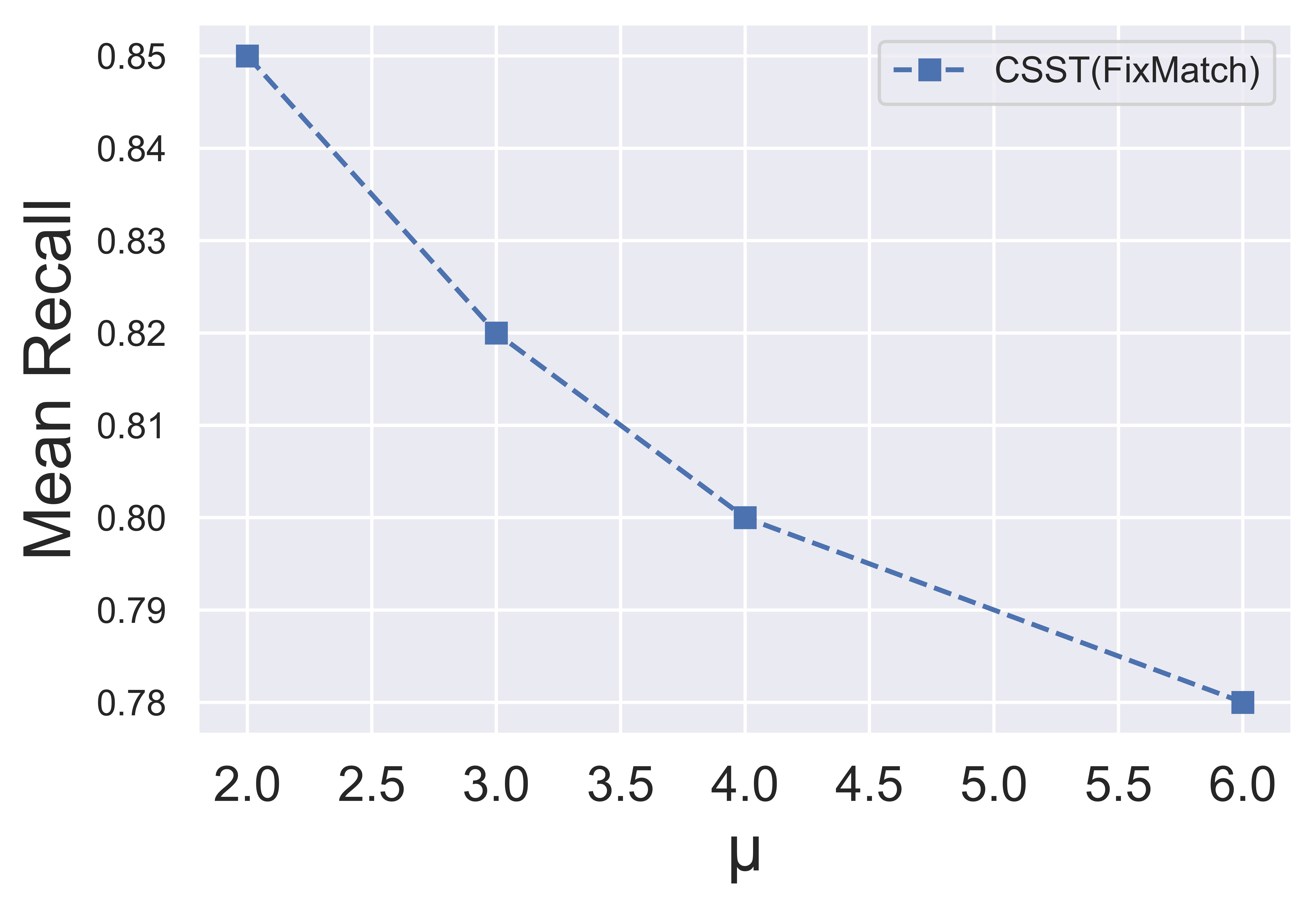}
  \caption{Data Ablation}
  \label{csst_fig:data-abl}
\end{subfigure}%
\begin{subfigure}{.5\textwidth}
  \centering
  \includegraphics[width=\linewidth]{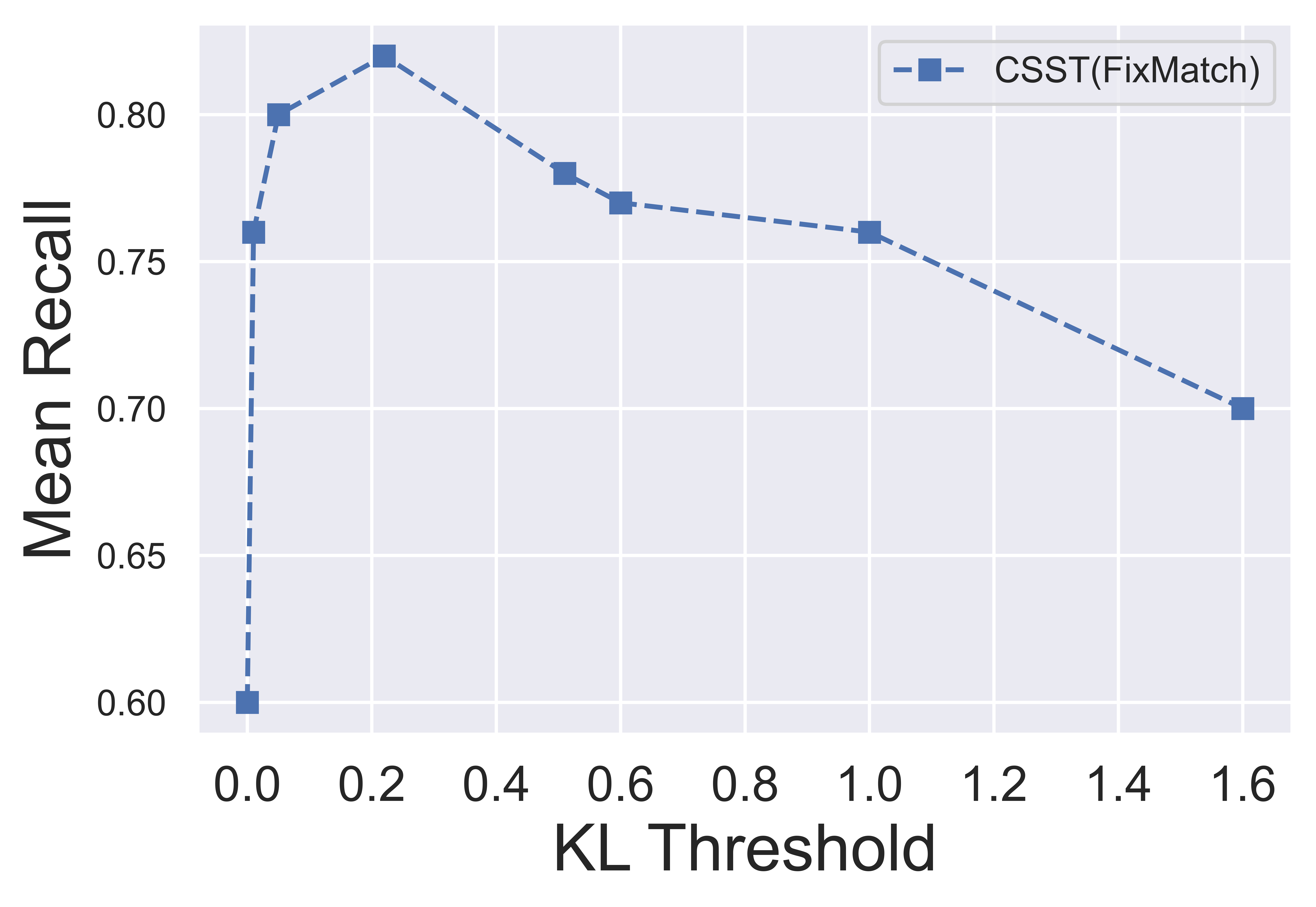}
  \caption{Threshold Ablation}
  \label{csst_fig:thresh-abl}
\end{subfigure}
\caption{Maximizing average recall under coverage constraints for CIFAR-10 Long tail ($\rho=100$) (Sec. \ref{csst_subsec:max-recall-under-cov}). Fig.~shows comparison of (a) increasing the ratio of unlabeled samples to labeled samples given fixed number of unlabeled samples (b) Ablation on KL diveregence based threshold for \texttt{CSST}(FixMatch)}
\label{csst_fig:ablation_main}
  \end{minipage}
\end{figure*}

\vspace{1mm} \noindent \textbf{Ablation on amount of unlabeled data.} Here, we ablated the total number of labeled sampled while keeping the number of unlabeled samples constant.
We observe (in Fig.~\ref{csst_fig:data-abl}) that as we increase the number of labeled samples the mean recall improved. This is because more labeled samples helps in better pseudo-labeling on the unlabeled samples and similarly as we decrease the number of labeled sample, the models errors on the pseudo-labels increase causing a reduction in mean recall. Hence, any addtional labeled data can be easily used to improve \ttt{CSST} performance.

\vspace{1mm} \noindent \textbf{Ablation on $\tau$ threshold.} Fig.~\ref{csst_fig:thresh-abl} shows that when the KL divergence threshold is too high,  a large number of samples with very low degree of distribution match are used for generating the sharpened target (or pseudo-labels), this leads to worsening of mean recall as many targets are incorrect. We find that keeping a conservative of $\tau = 0.3$ works well across multiple experiments.

\section{Related Work}
\vspace{1mm}\noindent \textbf{Self-Training.}
Self-training algorithms have been popularly used for the tasks of semi-supervised learning~\cite{berthelot2019mixmatch,xie2020self,sohn2020fixmatch,laine2016temporal} and unsupervised domain adaptation~\cite{saito2017asymmetric,zou2019confidence}. In recent years several regularizers which enforce consistency in the neighborhood (either an adversarial perturbation~\cite{miyato2018virtual} or augmentation~\cite{xie2020unsupervised}) of a given sample have further enhanced the applicability and performance of self-training methods, when used in conjunction. However, these works have focused mostly on improving the generic metric of accuracy, unlike the general non-decomposable metrics we consider.

\noindent \textbf{Cost-Sensitive Learning.} It refers to problem settings where the cost of error differs for a sample based on what class it belongs to. These settings are very important for critical real world applications like disease diagnosis, wherein mistakenly classifying a diseased person as healthy can be disastrous. There have been a pletheora of techniques proposed for these which can be classified into: importance weighting~\cite{lin2002support,zadrozny2003cost,cui2019class} and adaptive margin~\cite{cao2019learning,zadrozny2003cost} based techniques. For overparameterized models \citet{narasimhan2015consistent} show that loss weighting based techniques are ineffective and propose a logit-adjustment based cost-sensitive loss which we also use in our framework.

\noindent \textbf{Complex Metrics for Deep Learning.} There has been a prolonged effort on optimizing more complex metrics that take into account practical constraints~\cite{narasimhan2014statistical,puthiya2014optimizing,natarajan2016optimal}. However most work has focused on linear models leaving scope for works in context of deep neural networks. \citet{sanyal2018optimizing} train DNN using reweighting strategies for optimizing metrics,  \citet{huang2019addressing} use a reinforcement learning strategy to optimize complex metrics, and \citet{kumar2021implicit}
optimize complex AUC (Area Under Curve) metric for a deep neural network. However, all these works have primarily worked in supervised learning setup and are not designed to effectively make use of available unlabeled data.

\section{Conclusion}
In this work, we aim to optimize practical non-decomposable metrics readily used in machine learning through self-training with consistency regularization, a class of semi-supervised learning methods. We introduce a cost-sensitive self-training framework (\texttt{CSST}) that involves minimizing a cost-sensitive error on pseudo labels and consistency regularization. We show theoretically that we can obtain classifiers that can better optimize the desired non-decomposable metric than the original model used for obtaining pseudo labels, under similar data distribution assumptions as used for theoretical analysis of Self-training. We then apply \texttt{CSST} to practical and effective self-training method of FixMatch and UDA, incorporating a novel regularizer and thresholding mechanism based on a given non-decomposable objective. We find that \texttt{CSST} leads to a significant gain in performance of desired non-decomposable metric, in comparison  to vanilla self-training-based baseline. Analyzing the \texttt{CSST} framework when the distribution of unlabeled data significantly differs from labeled data is a good direction to pursue.

 \chapter{Selective Mixup Fine-Tuning for Optimizing Non-Decomposable Objectives}
\label{chap:selmix}

\begin{changemargin}{7mm}{7mm} 
The rise in internet usage has led to the generation of massive amounts of data, resulting in the adoption of various supervised and semi-supervised machine learning algorithms, which can effectively utilize the colossal amount of data to train models. However, before deploying these models in the real world, these must be strictly evaluated on performance measures like worst-case recall and satisfy constraints such as fairness. We find that current state-of-the-art empirical techniques offer sub-optimal performance on these practical, non-decomposable performance objectives. On the other hand, the theoretical techniques necessitate training a new model from scratch for each performance objective. To bridge the gap, we propose \textbf{SelMix}, a selective mixup-based inexpensive fine-tuning technique for pre-trained models, to optimize for the desired objective. The core idea of our framework is to determine a sampling distribution to perform a mixup of features between samples from particular classes such that it optimizes the given objective.  We comprehensively evaluate our technique against the existing empirical and theoretically principled methods on standard benchmark datasets for imbalanced classification. We find that proposed SelMix fine-tuning significantly improves the performance for various practical non-decomposable objectives across benchmarks.
    Code is available here: \href{https://github.com/val-iisc/SelMix/}{github.com/val-iisc/SelMix/}.

\end{changemargin}

\section{Introduction}
The rise of deep networks has shown great promise by reaching near-perfect performance across computer vision tasks~\cite{he2022masked, kolesnikov2020big, kirillov2023segment, girdhar2023imagebind}. It has led to their widespread deployment for practical applications, some of which have critical consequences~\cite{castelvecchi2020facial}. Hence, these deployed models must perform robustly across the entire data distribution and not just the majority part. These failure cases are often overlooked when considering only accuracy as our primary performance metric. 
Therefore, more practical metrics like Recall H-Mean~\cite{sun2006boosting}, Worst-Case (Min) Recall~\cite{narasimhan2021training, mohri2019agnostic}, etc., should be used for evaluation. However, optimizing these practical metrics directly for deep networks is challenging as they cannot be expressed as a simple average of a function of label and prediction pairs calculated for each sample~\cite{narasimhan2021training}. Optimizing such metrics with constraints is termed formally as \textbf{Non-Decomposable Objective} Optimization.

\begin{figure*}[t]
\begin{minipage}[c]{0.55\linewidth}
    \includegraphics[width=\textwidth]{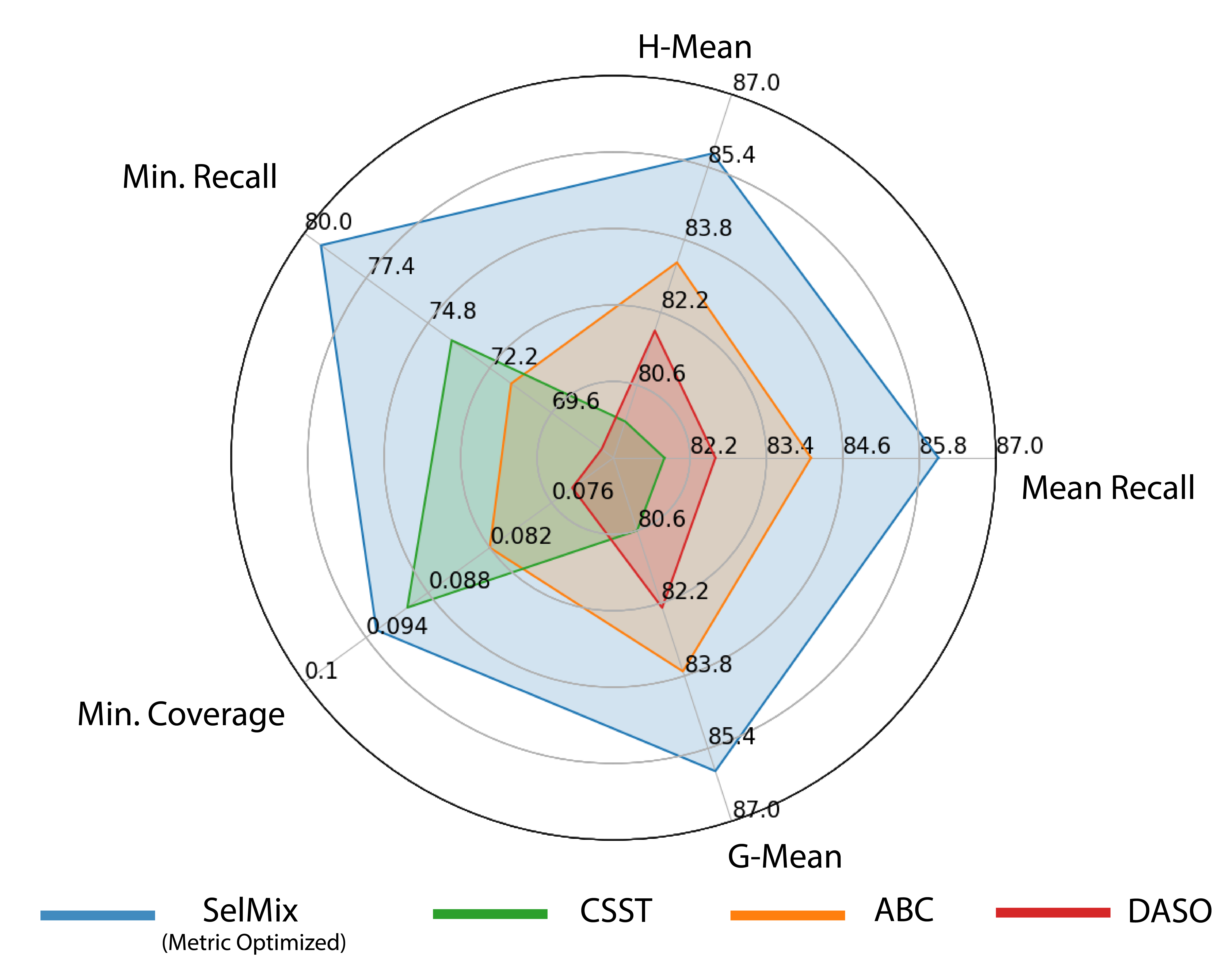}
\end{minipage}\hfill
\begin{minipage}[c]{0.43\linewidth}
  \caption{Overview of Results on CIFAR-10 LT (Semi-supervised). We evaluate the models from SotA Semi-supervised techniques of DASO~\cite{oh2022daso}, ABC~\cite{lee2021abc}, CSST~\cite{rangwani2022costsensitive} and proposed SelMix on different non-decomposable objectives. We find that SelMix produces the best performance for the non-decomposable metric and constraints it is optimized for (\textcolor{RoyalBlue}{\textbf{blue}}). Further, SelMix is an inexpensive fine-tuning technique compared to other expensive full pre-training-based baselines. }
    \label{fig:radar-results}    
\end{minipage}

\end{figure*}

In prior works, techniques exist to optimize such non-decomposable objectives, but their scope has mainly been restricted to linear models~\cite{narasimhan2014statistical, narasimhan2015optimizing}. \citet{narasimhan2021training}, recently developed consistent logit-adjusted loss functions for optimizing non-decomposable objectives for deep neural networks. After this work in supervised setup, Cost-Sensitive Self-Training (CSST) ~\citep{rangwani2022costsensitive} extends it to practical semi-supervised learning (SSL) setup, where both unlabeled and labeled data are present. As these techniques optimize non-decomposable objectives like Min-Recall, the \emph{ long-tailed (LT) imbalanced datasets} serve as perfect benchmarks for these techniques. However, CSST pre-training on long-tailed data leads to sub-optimal representations and hurts the mean recall of the models (Fig.~\ref{fig:radar-results}). Further, these methods require re-training the model from scratch to optimize for each non-decomposable objective, which decreases applicability.
Practical methods based on empirical insights have been developed to improve the mean performance of methods on long-tailed class-imbalanced datasets  (DASO ~\cite{oh2022daso}, ABC~\cite{lee2021abc}, CoSSL~\cite{fan2021cossl} etc.). These methods mainly generate debiased pseudo-labels for consistency regularization, leading to better semi-supervised classifiers. Despite their impressive performance, these classifiers perform suboptimally for the non-decomposable objectives (Fig. \ref{fig:radar-results}). 

 In this chapter, we develop SelMix, a technique \emph{that utilizes a pre-trained model for representations and optimizes it for improving the desired non-decomposable objective through fine-tuning}. We fine-tune a pre-trained model that provides good representations with Selective Mixups (SelMix) between data across different classes. The core contribution of our work is to develop a selective sampling distribution on class samples to selectively mixup, such that it optimizes the given non-decomposable objective (or metric) (Fig. \ref{fig:mixup_variants}). This SelMix distribution of mixup is updated periodically based on feedback from the validation set so that it steers the model in the direction of optimization of the desired metric.
 SelMix improves the decision boundaries of particular classes to optimize the objective, unlike standard Mixup ~\cite{zhang2018mixup} that applies mixups uniformly across all class samples. Further, the SelMix framework can also \textbf{optimize for non-linear objectives}, addressing a shortcoming of existing works \citep{rangwani2022costsensitive, narasimhan2021training}.

To evaluate the performance of SelMix, we perform experiments to optimize several different non-decomposable objectives. These objectives span diverse categories of linear objectives (Min Recall, Mean Recall), non-linear objectives (Recall G-mean, Recall H-mean), and constrained objectives (Recall under Coverage Constraints). 
We find that the proposed SelMix fine-tuning strategy significantly improves the performance on the desired objective, outperforming both the empirical and theoretical state-of-the-art (SotA) methods in most cases (Fig. \ref{fig:radar-results}).  In practical scenarios where the distribution of unlabeled data differs from the labeled data, we find that the adaptive design of SelMix with proposed logit-adjusted FixMatch (LA) leads to a significant $5\%$ improvement over the state-of-the-art methods, demonstrating its robustness to data distribution. Further, our SSL framework extends easily to supervised learning and leads to improvement in desired metrics over the existing methods. We summarize our contributions below: 
\setlist{nolistsep}
\begin{itemize}[noitemsep, leftmargin=*]
    \item We evaluate existing theoretical frameworks ~\cite{rangwani2022costsensitive} and empirical methods 
    ~\cite{oh2022daso, wei2021crest, lee2021abc, fan2021cossl}  on multiple practical non-decomposable metrics. We find that empirical methods perform well on mean recall but poorly on other practical metrics (Fig. \ref{fig:radar-results}) and vice-versa for the theoretical method.
    
    \item We propose \textbf{SelMix}, a mixup-based fine-tuning technique that uses selective mixup over classes to mix up samples to optimize the desired non-decomposable objective. (Fig. \ref{fig:mixup_variants}).

    \item We evaluate SelMix in various supervised and semi-supervised settings, including ones where the unlabeled label distribution differs from that of labeled data. We observe that SelMix with the proposed FixMatch (LA) pre-training outperforms existing SotA methods (Sec.~\ref{sec:empirical_results}).
\end{itemize}%

\section{Related Works}

\textbf{Semi-Supervised Learning in Class Imbalanced Setting.} Semi-Supervised Learning are algorithms that effectively utilizes unlabeled data and the limited labeled data present. A line of work has focused on using consistency regularization and pseudo-label-based self-training on unlabeled data to improve performance (e.g., FixMatch~\cite{sohn2020fixmatch}, MixMatch~\cite{berthelot2019mixmatch}, ReMixMatch~\cite{DBLP:journals/corr/abs-1911-09785}, etc.). However, when naively applied, these methods lead to biased pseudo-labels in class-imbalanced (and long-tailed) settings. Various recent~\cite{fan2021cossl, oh2022daso} methods have been developed that mitigate pseudo-label bias towards the majority classes. These include an auxiliary classifier trained on balanced data (ABC~\cite{lee2021abc}), a semantic similarity-based classifier to de-bias pseudo-label (DASO~\cite{oh2022daso}),  oversampling minority pseudo-label samples (CReST~\cite{Wei_2021_CVPR}). Despite their impressive accuracy, these algorithms compromise performance measures focusing on the minority classes.

\textbf{Non-Decomposible Metric Optimization.} 
Despite their impressive accuracy, there still seems to be a wide gap between the performance of majority and minority classes, especially for semi-supervised algorithms. For such cases, suitable metrics like worst-case recall across classes~\cite{mohri2019agnostic} and F-measure~\cite{eban2017scalable} provide a much better view of the model performance. However, these metrics cannot be expressed as a sum of performance on each sample; hence, they are non-decomposable. Several approaches have been developed to optimize the non-decomposable metrics of interest~\cite{kar2016online, narasimhan2014statistical, narasimhan2015optimizing, sanyal2018optimizing, narasimhan2021training}. In the recent work of CSST~\cite{rangwani2022costsensitive}, cost-sensitive learning with logit adjustment has been generalized to a semi-supervised learning setting for deep neural networks. However, these approaches excessively focus on optimizing desired non-decomposable metrics, leading to a drop in the model's average performance (mean recall). 

\textbf{Variants of MixUp.} After MixUp, several variants of mixup have been proposed in literature like CutMix~\cite{yun2019cutmix}, PuzzleMix~\cite{kim2020puzzle}, TransMix~\cite{chen2022transmix}, SaliencyMix~\cite{uddin2020saliencymix}, AutoMix~\cite{zhu2020automix} etc. However, all these methods have focused on creating mixed-up samples, whereas in our work SelMix, we concentrate samples from which classes $(y,y')$ are mixed up to improve the desired metric. Hence, it is complementary to others and can be combined with them. \addedtext{We also provide further discussion in App. ~\ref{app:rel_work_extra}.} \vspace{-3mm}

\section{Problem Setup} \label{sec:problem_setup}

\begin{figure*}[!t]
        \centering
    \includegraphics[width=\textwidth]{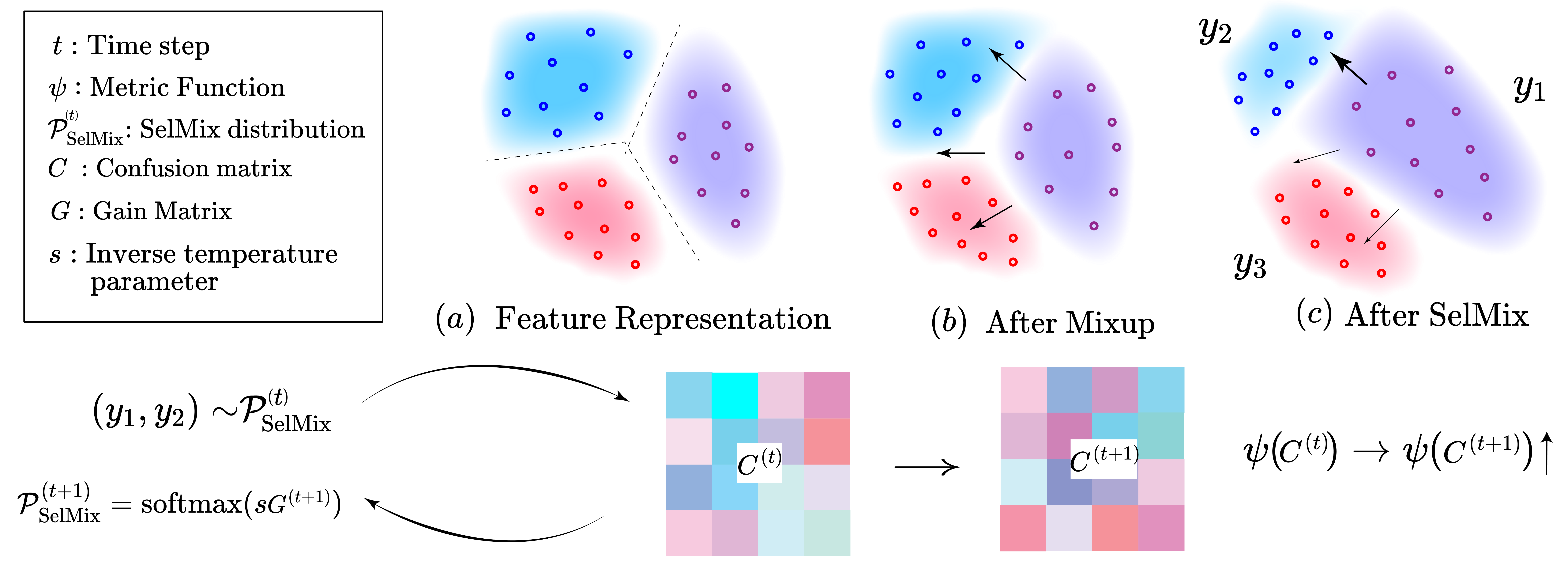}
    \caption{We demonstrate the effect of the variants of mixup on feature representations (a). With Mixup, the feature representation gets equal contribution in all directions of other classes (b).  Unlike this, in SelMix (c), certain class mixups are selected at a timestep $t$ such that they optimize the desired metric.  Below is an overview of how the SelMix distribution is obtained at timestep $t$.}
    \vspace{-1.25em}
    \label{fig:mixup_variants}
\end{figure*}

\textbf{Notation:} For two matrices ${A}, {B} \in \RR^{m \times n}$ with the same size, we define an inner product 
$\left\langle {A}, {B} \right\rangle$ as $\left\langle {A}, {B} \right\rangle = \Tr {A} {B}^\trn = \sum_{i=1}^m \sum_{j=1}^n A_{ij} B_{ij}$. For a general function $f: \RR^{m \times n} \rightarrow \addedtext{\RR}$ for a matrix variable $X \in \RR^{m \times n}$, the directional derivative w.r.t $V \in \RR^{m \times n}$ is defined as:
    \begin{equation}
    \label{defn:dir-derivative}
        \nabla_{V} f(X) = \lim_{\eta \rightarrow 0} \frac{f(X + \eta V) - f(X)}{\eta} 
    \end{equation}
which implies $ f(X + \eta V) \approx f(X) + \eta \nabla_{V} f(X)$ for small $\eta$. In case the function $f$ is differentiable, we depict the gradient (derivative) matrix w.r.t $X \in \RR^{m \times n}$ as $\frac{\partial f}{\partial X} \in \RR^{m\times n}$ with each entry given as:  $(\frac{\partial f}{\partial X})_{ij} = \frac{\partial f}{\partial X_{ij}}$. For a comprehensive list of variable definitions used, please refer Table. \ref{tab:notations}.

Let's examine the classification problem involving $K$ classes, where the data points denoted as $x$ are drawn from the instance space $\mX$, and the labels belong to the set $\mY$ defined as $[K]$. In this classification task, we define a classifier $F$ as a function that maps data points to labels. This function is constructed using a neural network-based scoring function $h$, which consists of two parts: a feature extractor $g$ that maps instances $x$ to a feature space $\RR^{d}$ and a linear layer parameterized by weights $W \in \RR^{d \times K }$. 

\begin{wraptable}{r}{10cm}
\caption{Objectives defined by confusion matrix.}\label{wrap-tab:objectives}
 \centering
\scriptsize
\begin{tabular}{lc}\\\midrule  
\qquad \qquad Objective & Definition  \\ [0.5ex] 
 \hline
 $(\psi^{\text{AM}})$ Mean Recall  &$\frac{1}{K}\sum_{i \in [K]}\frac{C_{ii}[h]}{\sum_{j\in[K]}{C_{ij}[h]}}$  \\ 
 $(\psi^{\text{MR}})$ Min.  Recall & $  \min_{\bm{\lambda} \in\Delta_{K-1}} \sum_{i \in [K]}\lambda_i \frac{C_{ii}[h]}{\sum_{j\in[K]}{C_{ij}[h]}}  $  \\
  $(\psi^{\text{HM}})$ H-mean &  $K\left( \sum_{i \in [K]} \frac{\sum_{j\in [K]} C_{ij}[h]}{C_{ii}[h]}\right)^{-1}$ \\
\multirow{2}{*}{ $(\psi^{\text{AM}}_{\text{cons.}})$ Mean Recall s.t.   } & 
$\min_{\bm{\lambda} \in \mathbb{R}^K_+} \sum_{i \in [K]} \frac{C_{ii}[h]}{\sum_{j\in[K]}{C_{ij}[h]}} + $\\ per class coverage $\geq \tau $  &  $  \sum_{j \in [K]}\lambda_j \left(\sum_{i \in [K]} C_{ij}[h] - \tau \right)$\\
\bottomrule
\end{tabular}
\end{wraptable}

The classification decision is made by selecting the label that corresponds to the highest score in the output vector: $F(x) = \argmax_{i\in[K]} $ $h(x)_i$. The output of this scoring function $h(x)$ in $\mathbb{R}^{K}$ is expressed as $h(x) = W^{\trn}g(x)$, in terms of network parameters. We assume access to samples from the data distribution $\mD$ for training and evaluation. We denote the prior distribution over labels as $\pi$, where $\pi_i = \bP(y=i)$ for $i=1,\dots, K$. Now, let's introduce the concept of the confusion matrix, denoted as $C[h]$, which is a key tool for assessing the performance of a classifier, defined as $C_{ij}[h] = \bE_{x,y \sim \mD}[\indc(y=i, \argmax_l h(x)_l=j)]$. For brevity, we introduce the confusion matrix in terms of scoring function $h$.  The confusion matrix characterizes how well the classifier assigns instances to their correct classes. An objective function $\psi$ is termed ``decomposable'' if it can be expressed as a function $\Phi: \mY \times \mY \rightarrow \RR$. Specifically, 
$\psi$ is decomposable if it can be written as $\ex[x, y]{\Phi(F(x), y)}$. If such a function $\Phi$ doesn't exist, the objective is termed ``non-decomposable''.
In this context, we introduce the non-decomposable objective $\psi$, represented as $\psi: \Delta_{K \times K - 1} \xrightarrow{} \mathbb{R}$, which is a function on the set of confusion matrices $C[h]$, and expressed as $\psi(C[h])$. Our primary aim is to maximize this objective $\psi(C[h])$ which can be used to express various practical objectives found in prior research works~\cite{cotter2019optimization, narasimhan2022consistent}, examples of which are provided in Table~\ref{wrap-tab:objectives} and their real-world usage is described below.

In real-world datasets, a common challenge arises from the inherent long-tailed and imbalanced nature of the distribution of data. In such scenarios, relying solely on accuracy can lead to a deceptive assessment of the classifier. This is because a model may excel in accurately classifying majority classes but fall short when dealing with minority ones. To address this issue effectively, holistic evaluation metrics like H-mean~\cite{kennedy2010learning}, G-mean~\cite{wang2012multiclass, lee2021abc}, and Minimum (worst-case) recall~\cite{narasimhan2021training} prove to be more suitable. These metrics offer a comprehensive perspective, highlighting performance disparities between majority and minority classes that accuracy might overlook. Specifically, the G-mean of recall can be expressed in terms of the confusion matrix ($C[h]$) as:
\begin{math}
\psi^{\text{GM}}({C[h]}) = \left ( \prod_{i \in [K]} \frac{C_{ii}[h]}{\sum_{j\in [K]} C_{ij}[h]} \right )^{\frac{1}{K}}.
\end{math}
For the minimum recall ($\psi^{\text{MR}}$), we use the continuous relaxation as used by \citep{narasimhan2021training}. By writing the overall  objective as min-max optimization over $\blambda \in \Delta_{K-1}$, we have
\begin{math}
    \max_{h} \psi^{\mathrm{MR}} (C[h]) = \max_{h} \min_{\blambda \in \Delta_{K-1}} \sum_{i \in [K]} \lambda_i \frac{C_{ii}[h]}{\sum_{j\in[K]}{C_{ij}[h]}} .
\end{math}
Fairness is another area where such complex objectives are beneficial. For example, prior works~\cite{cotter2019optimization, goh2016satisfying} on fairness consider optimizing the mean recall while constraining the predictive coverage ($\text{cov}_i [C[h]] = \sum_{j} C_{ji}$) that is the proportion of class $i$ predictions on test data given as
\begin{math}
    \max_{h} \frac{1}{K} \sum_{i=1}^K \rec_{i}[h] \; \; \text{s.t.} \; \; \cov_{i}[h]   \geq \frac{\alpha}{K} \; \; \forall i \in [K]. 
\end{math}
Optimization of the above-constrained objectives is possible by using the Lagrange Multipliers ($\blambda \in \RR^{K}_{\geq 0}$) as done in Sec. 2 of~\citet{narasimhan2021training}. By expressing this above expression in terms of $C[h]$ and through linear approximation, the constrained objective $\psi_{\text{cons.}}(C[h])$ can be considered as: 
\begin{math}
    \max_{h} \psi^{\text{AM}}_{\text{cons.}}(C[h]) = \max_{h} \min_{\blambda \in \RR^{K}_{\geq 0}} \frac{1}{K} \sum_{i \in [K]}  \frac{C_{ii}[h]}{\sum_{j\in[K]}{C_{ij}[h]}}
    + \sum_{j \in [K]}\lambda_j \left(\sum_{i \in [K]} C_{ij}[h] - \frac{\alpha}{K} \right).
\end{math}
The $\blambda$ for calculating the value of $\psi_{\text{cons.}}(C[h])$ and $\psi^{\text{MR}}(C[h])$, is periodically updated using exponentiated or projected gradient descent as done in \citep{narasimhan2021training}.
We summarize $\psi(C[h])$ for all non-decomposable objectives we consider in this work in Table \ref{wrap-tab:objectives}.  Unlike existing frameworks~\cite{narasimhan2021training, rangwani2022costsensitive} in addition to objectives that are linear functions of $C[h]$, we can also optimize for non-linear functions like (G-mean and H-mean) for neural networks.

\section{Selective Mixup}

\vspace{3mm}

This section introduces the proposed SelMix (Selective Mixup) procedure for optimizing desired non-decomposable objective $\psi$. We \textbf{a)} first introduce a notion of selective $\mathbf{(i,j)}$ mixup~\cite{zhang2018mixup} between samples of class $i$ and $j$, \textbf{b)} we then define the change (i.e., Gain) in desired objective $\psi$ produced due to $\mathbf{(i,j)}$ mixup, \textbf{c)} to prefer $\mathbf{(i,j)}$ mixups which lead to large Gain in objective $\psi$, we introduce a distribution $\mP_{\selmix}$ from which we sample $\mathbf{(i,j)}$ to form training batches \textbf{d)} we then conclude by introducing a tractable approximation of change (i.e., gain) in metric for the network $W^{\mathrm{T}}g$, \textbf{e)} we summarize the SelMix procedure by providing the Algorithm for training. We provide theoretical results for the optimality of \addedtext{the} SelMix procedure in Sec.~\ref{sec:theoretical_analysis}. We elucidate each part of SelMix in detail below and provide an overview of the proposed algorithm in Fig. \ref{fig:mixup_variants}.

\noindent \textbf{Feature Mixup.} In this work, we aim to optimize non-decomposable objectives with a framework utilizing mixup~\cite{zhang2018mixup}. Mixup minimizes the risk for a linear combination of samples and labels (i.e. ($x,y$) and ($x', y'$)) to achieve better generalization. Manifold Mixup~\cite{verma2019manifold} extends this idea to have \textbf{mixups in feature space, which we use in our work}. However, in vanilla mixup, the samples for mixing up are chosen randomly over the classes. This may be useful in the case of accuracy but can be sub-optimal when we aim to optimize for specific non-decomposable objectives (Table \ref{tab:policy-comparison})  that may require a specific selection of classes in the mixup. Hence, this work focuses on selective mixups between classes and uses them to optimize non-decomposable objectives. 
Consider a $K$ class classification dataset $D$ containing sample pairs ($x,y$). For convenience of notation, we denote the subset of instances with a particular label $y=i$ as $D_{i} = \{x \mid (x, y) \in D, y = i \}$. For the unlabelled part of dataset $\Tilde{D}$ containing $x$, we generate these subsets $\Tilde{D}_{i} = \{x' \in \Tilde{D} \mid i = \argmax_{l} h(x)_{l} \}$ based on the pseudo label $y'=\argmax_{l} h(x)_{l}$ from the model $h$. In addition to these training data, we also assume access to $D^{\mathrm{val}}$ containing ($x,y$).  Following semi-supervised mixup framework \citep{fan2021cossl}  in our work, the mixup between a labeled and pseudo labeled pair of samples (i.e. ($x,y$) from $D$ and $x'$ from $\Tilde{D}$), the features $g(x)$ and $g(x')$ are mixed up, while the label is kept as $y$. The mixup loss for our model with feature extractor $g$ followed by a linear layer with weights $W$ defined as:
\begin{align*}
    \label{eq:selmix-loss-simple}
    \mathcal{L}_{\text{mixup}}(g(x), g(x'), y; W) &=  \mathcal{L}_{\text{SCE}} (W^{\mathrm{T}}(\beta g(x) + (1 - \beta) g(x')), y).
\end{align*}
Here $\beta \sim U[\beta_{min}, 1], \beta_{min} \in [0,1]$ and $\mathcal{L}_{\text{SCE}}$ is the softmax cross entropy loss.  We define \emph{\textbf{(i, j) mixups}} for classes $i$ and $j$ to be the mixup of samples $x \sim D_i$ and $x' \sim \Tilde{D}_j$ and minimization of the corresponding $\mathcal{L}_{\text{mixup}}$ via SGD .
For analyzing the effect of ($i, j$) mixups on the model, we use the loss incurred by mixing the centroids of class samples ($z_i$ and $z_j$), which are defined as $ z_k = \mathbb{E}_{x \sim D_{k}^{\mathrm{val}}}[g(x)]$ for each class $k$. This representative of the expected loss due to $(i, j)$ mixup can be expressed as: \begin{equation}
        \label{eq:mixup-loss-centroid}
        \mathcal{L}_{(i, j)}^{\text{mix}} =  \mathcal{L}_{\text{mixup}}(z_i, z_j, i; W) \; \; \; \forall i,j \in [K]\times [K].
\end{equation}
\textbf{Directional vectors using mixup loss.}  We define $K^2$ directions as the derivative of the expected mixup loss for each of the $(i, j)$ mixup respectively w.r.t the weights $W$ as $V_{i,j} = -\partial{\mathcal{L}_{(i, j)}^{\text{mix}}}/\partial{W}$.
These directions correspond to the small change in weights $W$ upon the minimization of $\mathcal{L}_{(i, j)}^{\text{mix}}$ by stochastic gradient descent. Now we want to calculate the change in the non-decomposable objective $\psi$ along these directions $V_{i,j}$. Assuming the existence of directional derivatives in \addedtext{the} span of $K^2$ directions and fixed feature extractor $g$, we can write the following using Taylor Expansion (Eq. \eqref{defn:dir-derivative}): 
\begin{equation}
    \label{eq:psi-taylor-exps}
    \psi(C[W^\trn g + \eta V_{i,j}^\trn g]) = 
    \psi(C[W^\trn g]) + \eta
        \nabla_{V_{ij}} \psi(C[W^\trn g])
    + O(\eta^2\| V_{i,j}\| ^2).
\end{equation}
In Eq. \eqref{eq:psi-taylor-exps}, since $\eta$ is a small scalar, $ O(\eta^2\| V_{i,j}\| ^2)$ is negligible. Hence the second term in Eq. \eqref{eq:psi-taylor-exps} denotes the major change in objective $\psi$ due to minimization of the loss due to $(i,j)$ mixup $\mathcal{L}_{(i, j)}^{\text{mix}}$ via SGD. We define this as Gain ($G_{ij}$) or increase in desired objective $\psi$ for the $(i,j)$ mixup:
\begin{equation}
        \label{eq:exact-gains}
        G_{ij} =      
        \nabla_{V_{ij}} \psi(C[W^\trn g])
    ,  \quad \text{where} \; 
    {V}_{ij}  = - \frac{\partial \mathcal{L}_{(i, j)}^{\text{mix}}}{\partial W}\quad \forall(i, j)\in [K] \times [K].
\end{equation}
Using this, we define the gain matrix as  $G = [G_{i,j}]_{1 < i,j< K}$ corresponding to each of the $(i, j)$ mixups. We now define a general direction $V$, which is induced by mixing up samples from classes $(i, j)$ respectively, according to $\mathcal{P}_{Mix}(i, j)$ distribution defined over $[K] \times [K]$.  The expected weight change induced by this distribution can be given as (due to linearity of derivatives): $V = \sum_{i,j} \mathcal{P}_{Mix}(i, j) V_{i,j}$. The change in objective induced by the $\mathcal{P}_{Mix}(i, j)$ distribution can be similarly approximated using the Taylor Expansion for the direction $\addedtext{V} = \sum_{i,j} \mathcal{P}_{Mix}(i, j) V_{i,j}$: 
\begin{equation}
    \label{eq:psi-taylor-exps-v}
    \psi(C[W^\trn g + \eta \addedtext{V}^\trn g]) = 
    \psi(C[W^\trn g]) + \eta 
    \sum_{i,j} \mathcal{P}_{Mix}(i, j)  
     \nabla_{V_{ij}} \psi(C[W^\trn g])
    +O(\eta^2\| \addedtext{V}\| ^2).
\end{equation}
To maximize change in objective (LHS), we maximize the second term (RHS), as the first term is constant and the third is negligible for a small step $\eta$. On substituting 
     $   \nabla_{V_{ij}} \psi(C[W^\trn g])
    = G_{ij}$, the second term corresponds to $\mathbb{E}[G] = \sum_{i,j} G_{i,j} \mathcal{P}_{Mix}(i, j)$, which is expectation of gain over $\mathcal{P}_{Mix}$. Thus, maximization of the objective is equivalent to finding optimal $\mathcal{P}_{Mix}$ to maximize expected gain $\mathbb{E}[G]$. In practice to maximize objective $\psi$, we will first sample labels to mixup $(y_1, y_2)$ from optimal $\mathcal{P}_{Mix}$ (described below), then will pick $x_1, x_2 $ from $D_{y_1}, D_{y_2}$ uniformly to form a batch for training. 

\noindent \textbf{Selective Mixup Sampling Distribution.}
In our work, we introduce a novel sampling distribution $\mP_{\text{SelMix}}$ for practically optimizing the gain in objective defined as follows: 
\begin{equation}
    \label{eq:selmix-softmax}
    \mP_{\text{SelMix}}(i, j) = \mathrm{softmax} (sG)_{ij}.
\end{equation}
We aim to maximize $\mathbb{E}[G]$. One strategy could be to only mixup samples from classes $(i, j)$ respectively, corresponding to $\max_{i,j}{G_{ij}}$ or equivalently $s \rightarrow \infty$. However, this doesn't work in practice as we do $n$ steps of SGD based on $\mP_{\text{SelMix}}$  the linear approximation in Eq. \eqref{eq:psi-taylor-exps} becomes invalid\addedtext{,} and later the approximation in Thm. \ref{thm:gain-approx} (See Table~\ref{tab:policy-comparison} for empirical evidence). Hence, we select the $\mP_{\text{Mix}}$ to be the scaled softmax of the gain matrix as our strategy, with $s > 0$ given as $\mP_{\text{SelMix}}$. The proposed $\mP_{\text{SelMix}}$ is the distribution which is an intermediate strategy between the random exploratory uniform $(s = 0)$ and greedy $(s \rightarrow \infty)$ strategies which serves as a good sampling function for maximizing gain. 
We provide theoretical results regarding the optimality of the proposed $\mP_{\text{SelMix}}$ in Sec. \ref{sec:theoretical_analysis}.

\noindent \textbf{Estimation of Gain Matrix.} This notion of gain, albeit accurate, is not practically tractable since $\psi$ is not differentiable w.r.t $W$ in general, as the definition of $ C_{ij}[h] = \bE_{x,y \sim \mD}[\indc(y=i, \argmax_l h(x)_l=j)]$  uses a non-smooth indicator function (Sec. \ref{app:sec-proof-of-approx}). To proceed further with this limitation, we introduce a reformulation of $C$ by defining the $i^{\text{th}}$ row $C_i[h]$ of the confusion matrix in terms of the $i^{\text{th}}$ row $\tilde{C}_i[h]$ of the unconstrained matrix $\Tilde{C}[h] \in \RR^{K \times K}$ as $C_{i}[h] = \pi_i \cdot \text{softmax}(\tilde{C}_i[h])$. This reformulation of the confusion matrix $C$ by design satisfies the necessary constraints, given as $\sum_{j}C_{i,j}[h] = \pi_i \; \text{and} \; \sum_{i,j}C_{i,j}[h] = 1 \quad \text{where} \quad 0\leq C_{i,j}[h] \leq 1$. We can now calculate the same objective $\psi(\tilde{C}[W^{\mathrm{T}}g])$ in terms of $\tilde{C}$. 
The entries of gain matrix ($G$) with reformulation $\tilde{C}$ can be analogously defined (Eq. \eqref{eq:exact-gains}) in terms of $\Tilde{C}$:
\begin{equation}
        \label{eq:reform-gains}
          G_{ij} =      
         \nabla_{V_{ij}} \psi(\Tilde{C}[W^\trn g])
    ,  \quad \text{where} \; 
    {V}_{ij}  = - \frac{\partial \mathcal{L}_{(i, j)}^{\text{mix}}}{\partial W}\quad \forall (i, j)\in [K] \times [K].
\end{equation}
The exact computation of $G_{ij}$ would be given as $ \langle \frac{\partial{\widetilde{C}}}{\partial{W}}, V_{ij} \rangle $ in case  $\frac{\partial{\widetilde{C}}}{\partial{W}}$ was defined. However, this is not defined despite introducing the re-formulation due to the non-differentiability of $\tilde{C}$ w.r.t W. However, with re-formulation under some mild assumptions, given in Theorem \eqref{thm:gain-approx}, we can approximate $G_{ij}$ (first RHS term). We refer readers to Theorem \ref{thm:approx-formula} for a more mathematically precise statement. Further, we want to convey one advantage despite the proposed reformulation: we do not require the actual computation of  $\Tilde{C}$ for gain calculation in Eq. \eqref{eq:gain-approx-exact} (first term). 
As all the terms of $\frac{\partial \psi(\Tilde{C})}{\partial \tilde{C}_{lk}}$ which we require, can be computed analytically in terms of $C$, which makes this operation inexpensive. We provide the $\frac{\partial \psi(\Tilde{C})}{\partial \tilde{C}_{lk}}$ for all $\psi$ in Appendix Sec. \ref{app:unconstrained-derivative}. 
\begin{theorem}
\label{thm:gain-approx}
Assume that $\| V_{ij}\|$ is sufficiently small.
Then, the gain for the $(i,j)$ mixup ($G_{ij}$) can be approximated using the following expression:
\begin{equation}
        \label{eq:gain-approx-exact}
            G_{ij} =  \sum_{k,l}\frac{\partial \psi(\Tilde{C})}{\partial \tilde{C}_{kl}}\left((V_{ij})^{\top}_l \cdot z_k \right) + O\left(\varepsilon(\tC, W) + \|V_{ij}\|^2\right).
\end{equation}
where $z_k = \mathbb{E}_{x \sim D^{\mathrm{val}}_{k}}[g(x)] $ is the mean of the features of the validation samples belonging to class $k$, \addedtext{used to characterize (i,j) mixups (Eq.~\eqref{eq:mixup-loss-centroid})}. 
The error term $\varepsilon(\tC, W)$ does not depend on $V_{ij}$, and 
under reasonable assumptions we can regard this term small (we refer readers to Sec.~\ref{thm:approx-formula}). 
\end{theorem}

In the above theorem formulation\addedtext{,} we approximate the change in the entries of the unconstrained confusion matrix $\tilde{C}_{i,j}[h]$ with the change in logits for the classifier $W^{\mathrm{T}}g$ with weight $W$ along the direction $V_{i,j}$ as $({V}_{ij})^{\top}_l \cdot z_k$.
The most non-trivial assumption of the theorem is that 
for each $k$,
the random vector ${V}_{ij}^\trn g(x)$ has a small variance,
where $x \sim D_k^{\mathrm{val}}$.
Intuitively, if $g$ is a sufficiently good feature extractor, 
then feature vectors $g(x)$ ($x \sim D_k^{\mathrm{val}}$) are distributed near its mean,
hence its linear projections ${V}_{ij}^\trn g(x)$ has a small variance.
Therefore, it is a natural assumption if $g$ is sufficiently good. 
Moreover, this approximation works well in practice, as demonstrated empirically in Sec. \ref{sec:empirical_results}.

\noindent \textbf{Algorithm for training through $\bm{\mathcal{P}_{\text{SelMix}}}$.}
We provide an algorithm for training a neural network through SelMix.
Our algorithm shares \addedtext{a} high-level framework as used by ~\cite{narasimhan2015consistent, narasimhan2022consistent}. The idea is to perform training cycles, in which \addedtext{one} estimate\addedtext{s} the gain matrix $G$ through a validation set ($D^{\text{(val}}$) and use\addedtext{s} it to train the neural network for a few Stochastic Gradient Descent (SGD) steps.

As our expressions of gain are based on a linear classifier, we primarily fine-tune  the linear classifier. The backbone is fine-tuned at a lower learning rate $\eta$ for slightly better empirical results (Sec. \ref{sec:empirical_results}).
Formally, we introduce the time-dependent notations for gain ($G^{(t)}$), the classifier $h^{(t)}$, the SelMix distribution $\mP_{\text{SelMix}}^{(t)}$, weight-direction change $V_{i,j}^{(t)}$ due to the minimization of $\mathcal{L}_{ij}^{\text{mix}}$. As SelMix is a fine-tuning procedure, we assume a feature extractor $g$ and its linear classification layer, pre-trained with an algorithm like FixMatch\addedtext{,} to be provided as input.
Another important step in our algorithm is the update of the pseudo-labels of the unlabeled set, $\Tilde{D}^{(t)}$. 
\begin{wrapfigure}[9]{r}{0.5\textwidth}
    \begin{minipage}{0.5\textwidth}
        \begin{algorithm}[H]
            \centering
            \caption{Training through SelMix}
            \label{alg:ours}
            \begin{algorithmic}
                \scriptsize
                \FOR{$t=1$ {\bfseries to} $T$}
                \STATE Compute $\mP^{(t)}_{\text{SelMix}} = \text{softmax}(sG^{(t)})$ using Thm. \ref{thm:gain-approx}
                \FOR{$n$ SGD steps}
                \STATE $Y_1, Y_2 \sim \mP^{(t)}_{\text{SelMix}}$, $X_1 \sim \mathcal{U}(D_{Y_1})$, $X_2 \sim \mathcal{U}(\tilde{D}_{Y_2})$
                \STATE $h^{(t+1)} := \text{SGD-Update}(h^{(t)},\mathcal{L}_{\text{mixup}},(X_1, Y_1, X_2))$
                \ENDFOR
                \ENDFOR
                \RETURN $h^{(T)}$
            \end{algorithmic}
        \end{algorithm}
    \end{minipage}
\end{wrapfigure}The pseudo-labels are updated with predictions from the $h^{(t)}$. The algorithm is summarized in Alg.~\ref{alg:ours} and detailed in App. Alg. ~\ref{alg:ours-sup}, and an overview is provided in Fig. \ref{fig:mixup_variants}. In our practical implementation, we mask out entries of \addedtext{the} gain matrix with negative gain before performing the softmax operation (Eq. \eqref{eq:selmix-softmax}). We further compare the SelMix framework to others~\cite{rangwani2022costsensitive, narasimhan2021training}, in the App. ~\ref{app:comparison}.

\subsection{Theoretical Analysis of SelMix} 
\textbf{Convergence Analysis.}
For each iteration $t=1, \dots, T$, Algorithm \ref{alg:ours} updates the parameter $W$ of our network $h = W^{T}g$ as follows: 
(a) It selects a mixup pair $(i, j)$ from the distribution $\mP_{\mathrm{SelMix}}^{(t)}$,
(b) and updates the parameter $W$ by $W^{(t + 1)} = W^{(t)} + \eta_t \vtt$, where $\vtt = V_{ij}^{(t)}/\|V_{ij}^{(t)}\|$.
Here, we consider the normalized directional vector $\vtt$ instead of $V_{ij}^{(t)}$.
We denote by $\ex[t-1]{\cdot}$ the conditional expectation conditioned on randomness with respect to 
mixup pairs up to time step $t-1$.
In the convergence analysis, we make the following assumptions for the analysis.
We assume that the objective $\psi$ as a function of $W$ is concave, differentiable and the gradient is $\gamma$-Lipschitz, where $\gamma > 0$ ~\cite{wright2015coordinate,narasimhan2022consistent}.
    We assume that there exists a constant $c > 0$ independent of $t$ satisfying the following condition
    \begin{math}
        \ex[t-1]{\vtt} \cdot 
        \frac{\partial \psi(W^{(t)}) }{\partial W}
         > c\|\frac{\partial \psi(W^{(t)}) }{\partial W}\|,
    \end{math}
    that is, $\vtt$  \addedtext{vector has sufficient alignment with the gradient vector} to maximize the objective $\psi$ in expectation.
    Here, we regard $\psi$ as a function of $W$ in $ \frac{\partial \psi(W^{(t)}) }{\partial W}$.
    Moreover, we assume that in the optimization process, 
    $\|W^{(t)}\|$ does not diverge, i.e., we assume that for any $t \ge 1$, we have
    \begin{math}
        \| W^{(t)} \| < R
    \end{math}
    with a constant $R> 0$.
    In practice, this can be satisfied by adding $\ell^2$-regularization of $W$ to the optimization.
    We define a constant $R_0>0$ as $R_0 = \| W^*\| + R$,
    where $W^* = \argmax_W \psi(W)$. Using the above mild assumptions, we have the following  result (we provide proof in Sec. \ref{sec:app-convergence-analysis-proof}):
    \vspace{-2mm}
\begin{theorem}
    \label{thm:convergence-analysis}
    For any $t > 1$, we have
\begin{math}
  \psi(W^*) -  \ex{\psi(W^{(t)})} \le \frac{4\gamma R_0^2}{c^2 (t-1)},
\end{math}
with an appropriate choice of the learning rate $\eta_t$.
\end{theorem}
\vspace{-2mm}

\label{sec:theoretical_analysis}
Theorem \ref{thm:convergence-analysis} states that the proposed Algorithm \ref{alg:ours} leads to convergence to the optimal metric value $\psi(W^*)$
if $\ex{V_{ij}^{(t)}}$ is a reasonable directional vector for optimization of $\psi$.

\noindent \textbf{Validity of the mixup sampling distribution.}
 By formalizing the optimization process as an online learning problem (a similar setting to that of Hedge \citep{freund1997decision}),
we state that our sampling method is valid.
For conciseness, we only provide an informal statement here and refer to Sec. \ref{sec:app-analysis-of-var-of-selmix-policy} 
for a more precise formulation.
We suppose that 
a mixup pair $(i, j)$ is sampled by a distribution $\mathcal{P}^{(t)}$ on $[K] \times [K]$
for $t=1,\dots, T$.
We call a sequence of sampling distributions $\boldsymbol{\mathcal{P}} = (\mathcal{P}^{(t)})_{t=1}^{T}$ a policy,
and call a policy $\boldsymbol{\mathcal{P}}$ non-adaptive if $\mathcal{P}^{(t)}$ is the same for all $1 \le t \le T$.
For example, if $\mathcal{P}^{(t)}$ is the uniform distribution for any $t$, then $\boldsymbol{\mathcal{P}}$ is non-adaptive.
Then, in Sec. \ref{sec:app-analysis-of-var-of-selmix-policy}, we shall prove the following statement regarding the optimality of $\boldsymbol{\mathcal{P}}_\text{SelMix}$:
\begin{theorem}[Informal]
The SelMix policy $\boldsymbol{\mathcal{P}}_\text{SelMix}$ is approximately better than any non-adaptive policy $\boldsymbol{\mathcal{P}_{\mathrm{na}}} = (\mathcal{P})_{t=1}^{T}$ in terms of the average gain if $T$ is sufficiently large.
\end{theorem}

\section{Experiments}
\label{sec:empirical_results}

\begin{table*}[!t]

    \caption{Comparison of metric values with various Semi-supervised Long-Tailed methods on CIFAR-10/100 LT under $\rho_l$ = $\rho_u$ setup. The best results are indicated in bold.}
    \label{tab:matched-results}
    \vspace{-2mm}
    \begin{adjustbox}{width=\columnwidth,center}
    \begin{tabular}{lccccc|cccccc}\toprule
    & \multicolumn{5}{c}{\textbf{CIFAR-10} \quad $(\rho_l=\rho_u=100, N_1=1500, M_1=3000)$} & \multicolumn{5}{c}{\textbf{CIFAR-100} \quad $(\rho_l=\rho_u=10, N_1=150, M_1=300)$}
    \\  \cmidrule(lr){2-6}\cmidrule(lr){7-12}
               & Mean Rec.  & Min Rec. & HM & GM & Mean Rec./Min Cov.  & Mean Rec.  & Min H-T Rec. & HM & GM & Mean Rec./Min H-T Cov. \\\midrule
    
    DARP~\cite{10.5555/3495724.3496945} & \s{83.3}{0.4} & \s{66.4}{3.1} & \s{81.9}{0.5} & \s{82.6}{0.4} & \s{83.3}{0.4}/\s{0.070}{3e-3} &  \s{56.5}{0.2} & \s{39.6}{1.1} & \s{48.7}{1.3} & \s{55.4}{0.5} & \s{56.5}{0.2}/\s{0.0040}{2e-3} & \\
    CReST~\cite{wei2021crest} & \s{82.1}{0.6} & \s{68.2}{3.2} & \s{81.0}{0.7} & \s{81.6}{0.7} & \s{82.1}{0.6}/\s{0.073}{5e-3} &  \s{58.2}{0.2} & \s{40.7}{0.7} & \s{48.3}{0.2} & \s{54.1}{0.1} & \s{58.2}{0.2}/\s{0.0083}{2e-4} \\
    CReST+~\cite{wei2021crest}   & \s{83.1}{0.3} & \s{71.3}{1.5} & \s{82.2}{0.2} & \s{82.6}{0.3} & \s{83.1}{0.3}/\s{0.076}{2e-3}  & \s{57.8}{0.8} & \s{42.1}{0.7} & \s{48.2}{0.6} & \s{53.8}{0.9} & \s{57.8}{0.8}/\s{0.0088}{1e-4} \\
    
    ABC~\cite{lee2021abc} & \s{85.1}{0.5} & \s{74.1}{0.6} & \s{84.6}{0.5} & \s{84.9}{0.6} & \s{85.1}{0.5}/\s{0.086}{3e-3}  & \s{59.7}{0.2} & \s{46.4}{0.6} & \s{50.1}{1.2} & \s{55.6}{0.4} & \s{59.7}{0.2}/\s{0.0089}{3e-4} \\
    CoSSL~\cite{fan2021cossl} & \s{82.0}{0.3} & \s{70.6}{0.9} & \s{81.3}{0.5} & \s{81.6}{0.3} & \s{82.0}{0.3}/\s{0.074}{4e-3}  & \s{57.9}{0.4} & \s{46.3}{0.5} & \s{53.7}{0.8} & \s{55.2}{0.7} & \s{57.9}{0.4}/\s{0.0051}{3e-4} \\
    DASO~\cite{oh2022daso} & \s{84.1}{0.3} & \s{72.6}{2.1} & \s{83.5}{0.3} & \s{83.8}{0.3} & \s{84.1}{0.3}/\s{0.083}{1e-3} &   \s{\textbf{60.6}}{0.2} & \s{40.9}{0.4} & \s{49.1}{0.7} & \s{55.9}{0.1} & \s{60.6}{0.2}/\s{0.0063}{3e-4} \\
    CSST~\cite{rangwani2022costsensitive} & \s{81.1}{0.2} & \s{71.7}{0.2} & \s{76.9}{0.2} & \s{77.7}{0.7} & \s{81.1}{0.2}/\s{0.090}{2e-4} & \s{57.2}{0.2} & \s{48.4}{0.3} & \s{47.7}{0.8} & \s{53.5}{0.4} & \s{57.2}{0.2}/\s{\textbf{0.0099}}{2e-3}  \\
    
    FixMatch(LA) & \s{79.7}{0.6} & \s{55.9}{1.9} & \s{76.7}{0.1} & \s{78.3}{0.1} & \s{79.7}{0.6}/\s{0.056}{3e-3} & \s{58.8}{0.1} & \s{34.6}{0.6} & \s{45.5}{2.1} & \s{53.4}{0.4} & \s{58.8}{0.1}/\s{0.0053}{1e-5} \\
    
    \quad w/SelMix (Ours) & \s{\textbf{85.4}}{0.1} & \s{\textbf{79.1}}{0.1} & \s{\textbf{85.1}}{0.1} & \s{\textbf{85.3}}{0.1} & \s{\textbf{85.7}}{0.2}/\s{\textbf{0.095}}{1e-3} & \s{\textbf{59.8}}{{0.2}} & \s{\textbf{57.8}}{0.5} & \s{\textbf{53.8}}{0.5} & \s{\textbf{56.7}}{0.4} & \s{\textbf{59.6}}{0.5}/\s{\textbf{0.0098}}{5e-5} \\\bottomrule
    \end{tabular}
    \end{adjustbox}
    \vspace{-5mm} 
\end{table*}

 We demonstrate the effectiveness of SelMix in optimizing various Non-Decomposable objectives across different labeled and unlabeled data distributions. Following conventions for Long-Tail (LT) classification, $N_i$ and $M_i$ represent the number of samples in the $i^{\text{th}}$ class for the labeled and unlabeled sets, respectively. The label distribution is exponential in nature, and the imbalance factor $\rho$ characterizes it. We define it as $\rho_l = N_1/N_K, \addedtext{\rho_u} = M_1/M_K$.
In our experiments, we consider the LT semi-supervised version for CIFAR-10,100, Imagenet-100, and STL-10 datasets as done by \cite{fan2021cossl, oh2022daso, 10.5555/3495724.3496945, lee2021abc, rangwani2022costsensitive}. For the experiments on the long-tailed supervised dataset, we consider the Long-Tailed versions of CIFAR-10, 100, and ImageNet-1k. The parameters for the datasets are available in Tab. \ref{tab:hyperparams}.

\noindent \textbf{Training Details:} Our classifier comprises a feature extractor $g: \mX \rightarrow \mathbb{R}^d$ and a linear layer with weight $W$ (see Sec. \ref{sec:problem_setup}). In semi-supervised learning, we use the pre-trained Wide ResNet-28-2~\cite{DBLP:journals/corr/ZagoruykoK16} with FixMatch~\cite{sohn2020fixmatch}, replacing the loss function with the logit adjusted (LA) cross-entropy loss~\cite{menon2020long} for debiased pseudo-labels. Fine-tuning with SelMix (Alg. \ref{alg:ours}) includes cosine learning rate and SGD optimizer. In supervised learning, we pre-train models with MiSLAS on ResNet-32 for CIFAR-10, CIFAR-100, and ResNet-50 for ImageNet-1k. We freeze batch norm layers and fine-tune the feature extractor with a low learning rate to maintain stable mean feature statistics $z_k$, as per our theoretical findings. Further details and hyperparameters are provided in appendix Table \ref{tab:hyperparams}.

\noindent \textbf{Evaluation Setup.} We evaluate our work against baselines CReST, CReST+~\cite{wei2021crest}, DASO~\cite{oh2022daso}, DARP~\cite{10.5555/3495724.3496945}, and ABC~\cite{lee2021abc} in semi-supervised long-tailed learning.  We assess the methods based on two sets of metric objectives: a) \textbf{Unconstrained objectives}, including G-mean, H-mean, Mean (Arithmetic Mean), and worst-case (Min.) Recall. b) \textbf{Constrained objectives}, involving maximizing recalls under coverage constraints. The constraint requires coverage $\geq \frac{0.95}{K}$ for all classes. For CIFAR-100, we optimize Min Head-Tail Recall/Min Head-Tail coverage instead of Min Recall/Coverage due to its small size. The tail corresponds to the least frequent 10 classes, and the head the rest 90 classes. For detailed metric objectives and definitions, refer to Table \ref{tab:metric_full}. We present results as mean and standard deviation across three seeds.

\noindent \textbf{Matched Label Distributions.} We report results for $\rho_l = \rho_u$, signifying matched labeled and unlabeled class label distributions. SelMix outperforms FixMatch (LA), achieving a 5\% Min Recall boost for CIFAR-10 and a 9.8\% improvement in Min HT Recall for CIFAR-100. SelMix also excels in mean recall, akin to accuracy. Its strategy starts with tail class enhancement, transitioning to uniform mixups (App. \ref{app:gain-matrix-train}).We delve into optimizing coverage-constrained objectives ($\cov_{\text{i}}[h] \geq \frac{0.95}{K}$). Initially, we emphasize mean recall with coverage constraints, supported by CSST. However, SelMix, a versatile method, accommodates objectives like H-mean with coverage (App. \ref{app:hmean-coverage}). Table \ref{tab:matched-results} reveals that most SotA methods miss minimum coverage values, except CSST and SelMix. SelMix outperforms CSST in mean recall while meeting constraints, as confirmed in our detailed analysis (App. \ref{app:detailed-analysis}).
\begin{figure*}[!t]
    \centering
    \includegraphics[width=\textwidth]{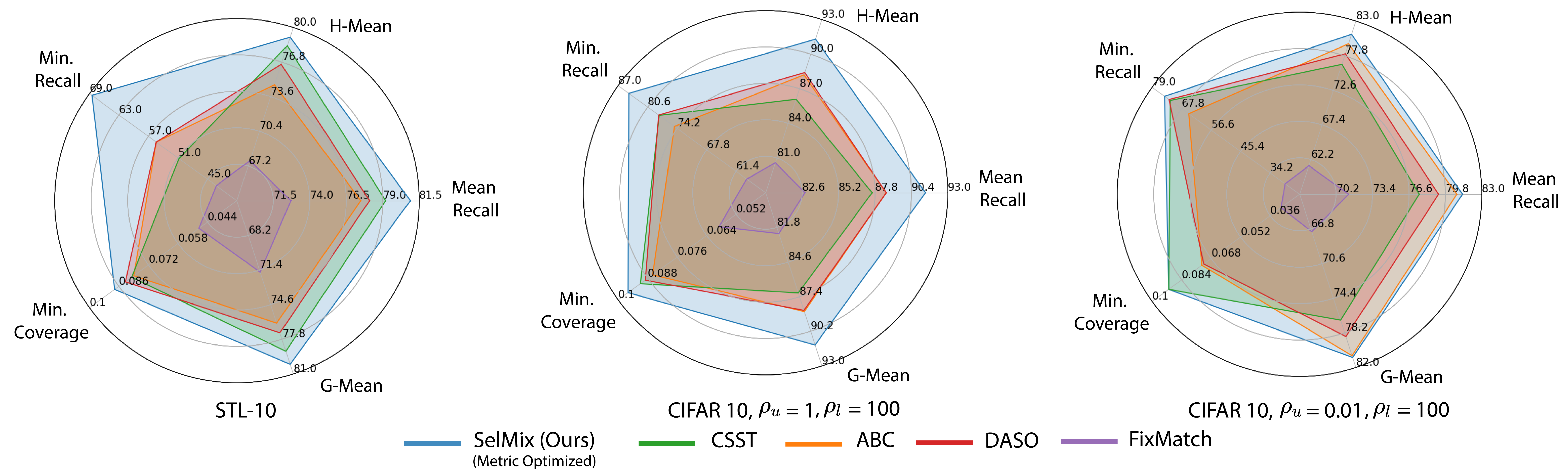}
    \caption{Comparison of metric for semi-supervised CIFAR-10 LT under $\rho_l \neq \rho_u$ and STL-10 $\rho_u = NA$  assumption. For CIFAR-10-LT (semi-supervised) involve $\rho_l = 100, \rho_u = 1$, (uniform) and $\rho_l = 100, \rho_u = \frac{1}{100}$ (inverted). SelMix achieves significant gains over other baselines.}
    \label{fig:summary_mismatched_radar}
    \vspace{-2mm}
\end{figure*}

\begin{table*}[!t]
\vspace{-2mm}
\caption{Comparison results in supervised case for CIFAR-10,100 LT ($\rho=100$). We use the pre-trained model of MiSLAS \cite{zhong2021mislas} in stage-1 and fine-tune using SelMix.}
\vspace{-2mm}
\begin{adjustbox}{width=\columnwidth,center}
\begin{tabular}{lccccc|cccccc}\toprule
& \multicolumn{5}{c}{\textbf{CIFAR-10} \quad $(\rho_l=100)$} & \multicolumn{5}{c}{\textbf{CIFAR-100} \quad $(\rho_l=10)$}
\\  \cmidrule(lr){2-6}\cmidrule(lr){7-12}
           & Mean Rec.  & Min Rec. & HM & GM & Mean Rec./Min Cov.  & Mean Rec.  & Min H-T Rec. & HM & GM & Mean Rec./Min H-T Cov. \\\midrule
MisLaS (Stage 1)~\cite{zhong2021mislas}   &  \s{72.7}{0.3} & \s{45.6}{2.3} & \s{70.3}{1.4} & \s{72.5}{0.9} & \s{72.7}{0.3}/\s{0.045}{2e-3} & \s{39.5}{0.2} & \s{1.2}{0.5} & \s{0.0}{0.0} & \s{0.0}{0.0} & \s{39.5}{0.2}/\s{0.0001}{2e-5}  \\
\quad w/ Stage 2~\cite{zhong2021mislas}  & \s{81.9}{0.1} & \s{72.5}{0.8} & \s{81.3}{0.9} & \s{81.6}{0.1} & \s{81.9}{0.1}/\s{0.077}{0.003} & \s{47.0}{0.4} & \s{15.2}{1.1} & \s{30.9}{0.6} & \s{39.9}{0.5} & \s{47.0}{0.4}/\s{0.0055}{2e-4} \\
\quad w/ SelMix (Ours)& \s{\textbf{83.3}}{0.2} & \s{\textbf{79.2}}{0.7} & \s{\textbf{82.6}}{0.5} & \s{\textbf{82.8}}{0.3} & \s{\textbf{82.8}}{0.2}/\s{\textbf{0.095}}{0.002} &  \s{\textbf{48.3}}{0.1} & \s{\textbf{41.3}}{1.4} & \s{\textbf{38.2}}{0.8} & \s{\textbf{42.3}}{0.5} & \s{\textbf{47.8}}{0.2}/\s{\textbf{0.0095}}{2e-4} &
 \\\bottomrule
\end{tabular}
\label{tab:cifar10/100-sup}
\end{adjustbox}
\vspace{-2.5mm}
 \end{table*}

\begin{table*}[!t]
\caption{Results for scaling SelMix to large datasets ImageNet-1k LT and ImageNet100 LT. }
\vspace{-3mm}
\begin{adjustbox}{width=\columnwidth,center}
\begin{tabular}{lccc|lccc}\toprule
& \multicolumn{3}{c}{\textbf{ImageNet100-LT ($\rho = 10$)}}& & \multicolumn{3}{c}{\textbf{ImageNet1k-LT }} \\  
\midrule
 & Mean Rec. & Min Rec. & Mean Rec./Min H-T Cov.&   & Mean Rec. & Min HT Rec. &  Mean Rec./Min H-T Cov. \\ \midrule
  CSST~\cite{rangwani2022costsensitive} & 59.1 & 12.1 &  59.1/0.003 & MiSLAS (Stage 1)~\cite{zhong2021mislas} & 45.4 & 4.1 & 45.4/0.00000 \\
  Fixmatch (LA) & 69.9 & 6.0 & 69.9/0.002 &  \quad w/ Stage 2 & 52.4 & 29.7 & 52.4/0.00068\\
 \quad  w/ SelMix & \textbf{73.5} & \textbf{24.0} & \textbf{73.1}/\textbf{0.009} &  \quad w/ SelMix & \textbf{52.8} & \textbf{45.1} & \textbf{52.5}/\textbf{0.00099}\\
 \bottomrule

\end{tabular}
\end{adjustbox}
\label{tab:large-data}
\vspace{-5mm}
 \end{table*}

\noindent \textbf{Unknown Label Distributions.} We address the practical scenario where the labeled data's label distribution differs from that of the unlabeled data ($\gamma_l \neq \gamma_u$). We assess two cases: \textit{a) Mismatched Distributions.} We evaluate various techniques on CIFAR-10 with two mismatched unlabeled distributions: balanced ($\rho_u=1$) and inverse ($\frac{1}{100}$). SelMix consistently outperforms all methods, especially in min. Recall and coverage-constrained objectives (Fig. \ref{fig:summary_mismatched_radar}). \textit{b) Real World Unknown Label Distributions.} STL-10 provides an additional 100k samples with an unknown label distribution, emulating scenarios where data is abundant but labels are scarce. SelMix, with no distributional assumptions, outperforms SotA methods like CSST and CRest (which assume matched distribution) in min-recall by a substantial 12.7\% margin (Fig. \ref{fig:summary_mismatched_radar}). Detailed results can be found in App. \ref{app:mismatched_comparison}.

\noindent \textbf{Results on SelMix in Supervised Learning.} To further demonstrate the generality of SelMix, we test it for optimizing non-decomposable objectives via fine-tuning a recent SotA work MisLaS~\cite{zhong2021mislas} for supervised learning. In comparison to fine-tuning stage-2 of MisLaS, SelMix-based fine-tuning achieves better performance across all objectives as in Table~\ref{tab:cifar10/100-sup}, for both CIFAR 10,100-LT.

\noindent \textbf{Analysis of SelMix.} We demonstrate SelMix's scalability on large-scale datasets like Imagenet-1k LT and Imagenet-100 LT and its ability to improve the objective compared to the baseline Tab. \ref{tab:large-data}, with minimal additional compute cost ($\sim$ 2 min.) (see Table \ref{tab:time-required}), through Thm. \ref{thm:regbound-variant}, we show the advantage of SelMix over uniform random sampling and the limitations of a purely greedy policy (ref. Table ~\ref{tab:policy-comparison} for empirical evidence). We observe improved feature extractor learning by comparing a trainable backbone to a frozen one (Tab. \ref{tab:backbone-scaling}). Additionally, our work can be combined with other mixup variants like \cite{kim2020puzzle, yun2019cutmix}, leading to performance enhancements (Tab. \ref{tab:mixup_variants}) demonstrating the diverse applicability for the proposed SelMix method. We refer readers to the Appendix for details on complexity (Sec.\ ~\ref{app:complexity},~\ref{app:computation}), analysis (Sec.\ ~\ref{app:analysis}), and limitations (Sec. ~\ref{app:limitations}).

\section{Conclusion and Discussion}
 We study the optimization of practical but complex metrics like the G-mean and H-mean of Recalls, along with objectives with fairness constraints in the case of neural networks. We find that SotA techniques achieve sub-optimal performance in terms of these practical metrics, notably on worst-case recall. These metrics and constraints are non-decomposable objectives, for which we propose a Selective Mixup (SelMix) based fine-tuning algorithm for optimizing them. The algorithm selects samples from particular classes to a mixup to improve a tractable approximation of the non-decomposable objective.  Our method\addedtext{,} SelMix, can improve on the majority of objectives in comparison to both theoretical and empirical SotA methods, bridging the gap between theory and practice. We expect \addedtext{the} SelMix fine-tuning technique to be used for improving existing models by improving on worst-case and fairness metrics inexpensively.

\part{Efficient Domain Adaptation}
\label{part:ef_DA}

\chapter{S$^3$VAADA: Submodular Subset Selection for Virtual Adversarial Active Domain Adaptation }
\label{chap:S3VAADA}

\begin{changemargin}{7mm}{7mm} 
   Unsupervised domain adaptation (DA) methods have focused on achieving maximal performance through aligning features from source and target domains without using labeled data in the target domain. Whereas, in the real-world scenario’s it might be feasible to get labels for a small proportion of target data. In these scenarios, it is important to select maximally-informative samples to label and find an effective way to combine them with the existing knowledge from source data. Towards achieving this, we propose S$^3$VAADA which i) introduces a novel submodular criterion to select a maximally informative subset to label and ii) enhances a cluster-based DA procedure through novel improvements to effectively utilize all the available data for improving generalization on target. Our approach consistently outperforms the competing state-of-the-art approaches on datasets with varying degrees of domain shifts. The project page with additional details is available here: \href{https://sites.google.com/iisc.ac.in/s3vaada-iccv2021/}{https://sites.google.com/iisc.ac.in/s3vaada-iccv2021/}.
\end{changemargin}

\section{Introduction}

Deep Neural Networks have shown significant advances in image classification tasks by utilizing large amounts of labeled data. Despite their impressive performance, these networks produce spurious predictions hence suffer from performance degradation when used on images that come from a different domain~\cite{saenko2010adapting} (e.g., model trained on synthetic data (source domain) being used on real-world data (target domain)). Unsupervised Domain Adaptation (DA)~\cite{ganin2015unsupervised, long2018conditional, chen2020adversarial, saito2018adversarial,Kundu_2020_CVPR, Kundu_2020_CVPR_usfda} approaches aim to utilize the labeled data from source domain along with the unlabeled data from the target domain to improve the model’s generalization on the target domain. However, it has been observed that the performance of Unsupervised DA models often falls short in comparison to the supervised methods \cite{tsai2018learning}, which leads to their reduced usage for performance critical applications. In such cases, it might be possible to label some of the target data to improve the performance of the model. 
\begin{figure}[t]
    \centering
    \includegraphics[width=0.75\textwidth]{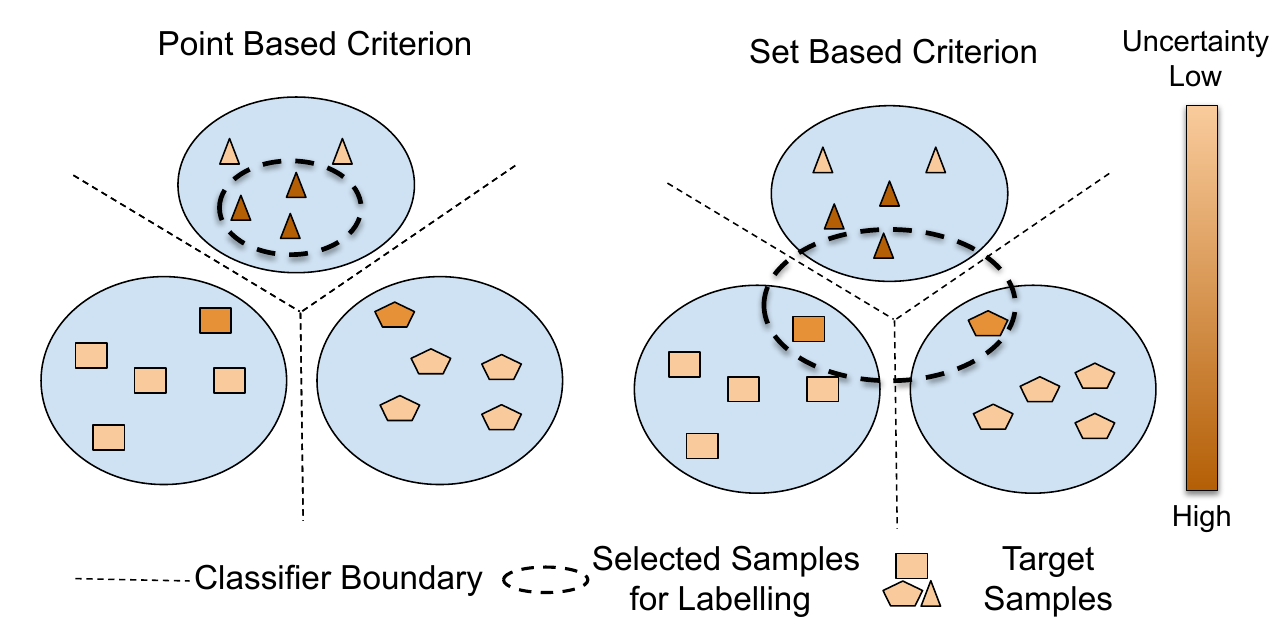}
    \caption{We pose sample selection  for labeling in \textbf{Active Domain Adaptation} as an informative subset selection problem. We propose an information criterion to provide \textit{score for each subset} of samples for labeling. Prior works (See t-SNE for AADA~\cite{Su_2020_WACV} in Sec. 1 of App.~\ref{app:s3vaada}) which use a point-based criterion \textit{(i.e. score each sample independently)} to select samples suffer from redundancy. As our set-based criterion is aware of the other samples in the set, it avoids redundancy and selects diverse samples. }
    \label{s3vaada_fig:inuitive_explanation}
\end{figure}

\begin{figure*}[htp]
  \centering
  \includegraphics[width=\linewidth]{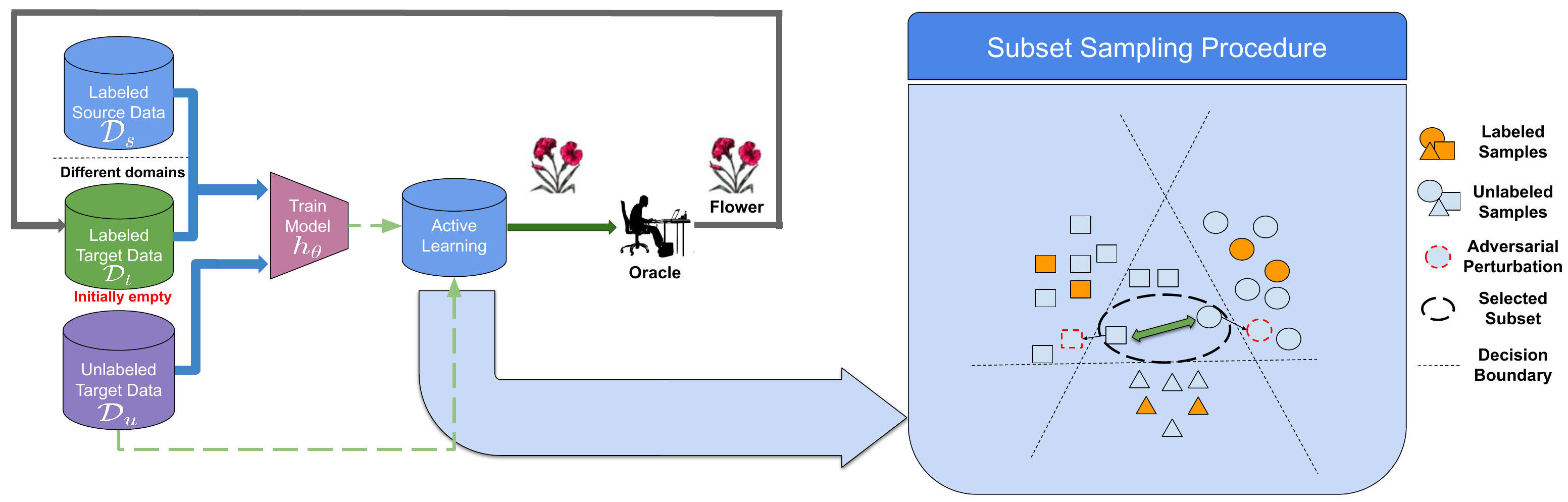}
  \caption{Overview of Submodular Subset Selection for Virtual Adversarial Active Adaptation (S$^3$VAADA). Step 1: We select a subset of samples which are uncertain (i.e., prediction can change with small perturbation), diverse and representative in feature space (see Fig. \ref{s3vaada_fig:sampling-al} for details). Step 2: The labeled samples and the unlabeled samples are used by proposed VAADA adaptation procedure to obtain the final model. The above two steps are iteratively followed selecting $B$ samples in each cycle, till the annotation budget is exhausted.}
  \label{s3vaada_fig:al}
\end{figure*}

In such a case, the dilemma is, \textit{“Which samples from the target dataset should be selected for labeling?”}. Active Learning (AL)~\cite{cohn1994improving, settles2009active} approaches aim to provide techniques to select the maximally informative set for labeling, which is then used for training the model. However, these approaches do not effectively use the unlabeled data and labeled data present in various domains. This objective contrasts with Unsupervised DA objective that aims to use the unlabeled target data effectively. In practice, it has been found that just naively using AL and fine-tuning offers sub-optimal performance in presence of domain shift~\cite{Su_2020_WACV}. 

Another question that follows sample selection (or sampling) is, \textit{“How to effectively use all the data available to improve model performance?”}. Unsupervised DA approaches based on the idea of learning invariant features for both the source and target domain have been known to be ineffective in increasing performance when additional labeled data is present in target domain~\cite{saito2019semi}. Semi-Supervised DA (SSDA)~\cite{saito2019semi, li2020online} methods have been developed to mitigate the above issue, but we find their performance plateau’s as additional data is added (Sec. \ref{s3vaada_sec:results}). This is likely due to the assumption in SSDA of only having a small amount of labeled data per-class (i.e., few shot) in target domain which is restrictive.

The Active Domain Adaptation (Active DA) paradigm introduced by Rai et al. \cite{rai2010domain} aims to first effectively select the informative-samples, which are then used by a complementary DA procedure for improving model performance on target domain. The state-of-the-art work of AADA~\cite{Su_2020_WACV} aim to select samples with high value of $p_{target}(x)/p_{source}(x)$ from domain discriminator, multiplied by the entropy of classifier which is used by DANN \cite{ganin2015unsupervised} for adaptation. As the AADA criterion is a point-estimate, it is unaware of other selected samples; hence the samples selected can be redundant, as shown in Fig. \ref{s3vaada_fig:inuitive_explanation}.

In this work, we introduce  Submodular Subset Selection for Virtual Adversarial Active Domain Adaptation (S$^3$VAADA) which proposes a set-based informative criterion that provides scores for each of the subset of samples rather than a point-based estimate for each sample. As the information criteria is aware of other samples in the subset, it tends to avoid redundancy. Hence, it is able to select diverse and uncertain samples (shown in Fig. \ref{s3vaada_fig:inuitive_explanation}). Our subset criterion is based on the cluster assumption, which has shown to be widely effective in DA ~\cite{deng2019cluster, lee2019drop}. The subset criterion is composed of a novel uncertainty score (Virtual Adversarial Pairwise (VAP)) which is based on the idea of the sensitivity of model to small adversarial perturbations. This is combined with a distance based metrics such that the criterion is submodular (defined in Sec. \ref{s3vaada_sec:notations}). The submodularity of the criterion allows usage of an efficient algorithm \cite{nemhauser1978analysis} to obtain the optimal subset of samples.  After obtaining the labeled data, we use a cluster based domain adaption scheme based on VADA \cite{shu2018dirt}. Although VADA, when naively used is not able to effectively make use of the additional target labeled data \cite{saito2018maximum}, we mitigate this via two modifications (Sec. \ref{s3vaada_section:VAADA}) which form our Virtual Active Adversarial Domain Adaptation (VAADA) procedure. \\
In summary, our contributions are: 
\begin{itemize}[noitemsep]
\item We propose a novel set-based information criterion which is aware of other samples in the set and aims to select uncertain, diverse and representative samples.
\item For effective utilization of the selected samples, we propose a complementary DA procedure of VAADA which enhances VADA's suitability for active DA.%
\item Our method demonstrates state-of-the-art active DA results on diverse domain adaptation benchmarks of Office-31, Office-Home and VisDA-18.
\end{itemize}

\section{Related Work}
\label{s3vaada_sec:rel_work}
\textbf{Domain Adaptation}:  One of the central ideas in DA is minimizing the discrepancy in two domains by aligning them in feature space. 
DANN~\cite{ganin2015unsupervised} achieves this by using domain classifier which is trained through an adversarial min-max objective to align the features of source and target domain. MCD~\cite{saito2018maximum} tries to minimize the discrepancy by making use of two classifiers trained in an adversarial fashion for aligning the features in two domains. The idea of semi-supervised domain adaptation by using a fraction of labeled data is also introduced in MME~\cite{saito2019semi} approach, which induces the feature invariance by a MinMax Entropy objective. Another set of approaches uses the cluster assumption to cluster the samples of the target domain and source domain. In our work, we use ideas from VADA (Virtual Adversarial Domain Adaptation)~\cite{shu2018dirt} to enforce cluster assumption. \\
\textbf{Active Learning (AL):} The traditional AL methods are iterative schemes which obtain labels (from oracle or experts) for a set of informative data samples. The newly labeled samples are then added to the pool of existing labeled data and the model is trained again on the labeled data. The proposed techniques can be divided into two classes: 1) \textbf{Uncertainty Based Methods}: In this case, model uncertainty about a particular sample is measured by specific criterion like entropy~\cite{wang2014new} etc. 2) \textbf{Diversity or Coverage Based Methods}: These methods focus on selecting a diverse set of points to label in order to improve the overall performance of the model. One of the popular methods, in this case, is Core-Set~\cite{sener2018active} which selects samples to maximize the coverage in feature space. However, recent approaches like BADGE~\cite{Ash2020Deep} which use a combination of uncertainty and diversity, achieve state-of-the-art performance.
A few task-agnostic active learning methods \cite{sinha2019variational, yoo2019learning} have also been proposed.\\\\
\textbf{Active Domain Adaptation:} The first attempt for active domain adaptation was made by Rai et al. \cite{rai2010domain}, who use linear classifier based criteria to select samples to label for sentiment analysis. Chattopadhyay et al. \cite{chattopadhyay2013joint} proposed a method to perform domain adaptation and active learning by solving a single convex optimization problem. AADA (Active Adversarial Domain Adaptation)~\cite{Su_2020_WACV} for image based DA is a method which proposes a hybrid informativeness criterion based on the output of classifier and domain discriminator used in DANN. The criterion used in AADA for selecting a batch used is a point estimate, which might lead to redundant sample selection. We introduce a set-based informativeness criterion to select samples to be labeled. CLUE~\cite{prabhu2020active} is a recent concurrent work which selects samples through uncertainty-weighted clustering for Active DA.

\begin{figure*}[!t]
  \centering
  \includegraphics[width=\linewidth]{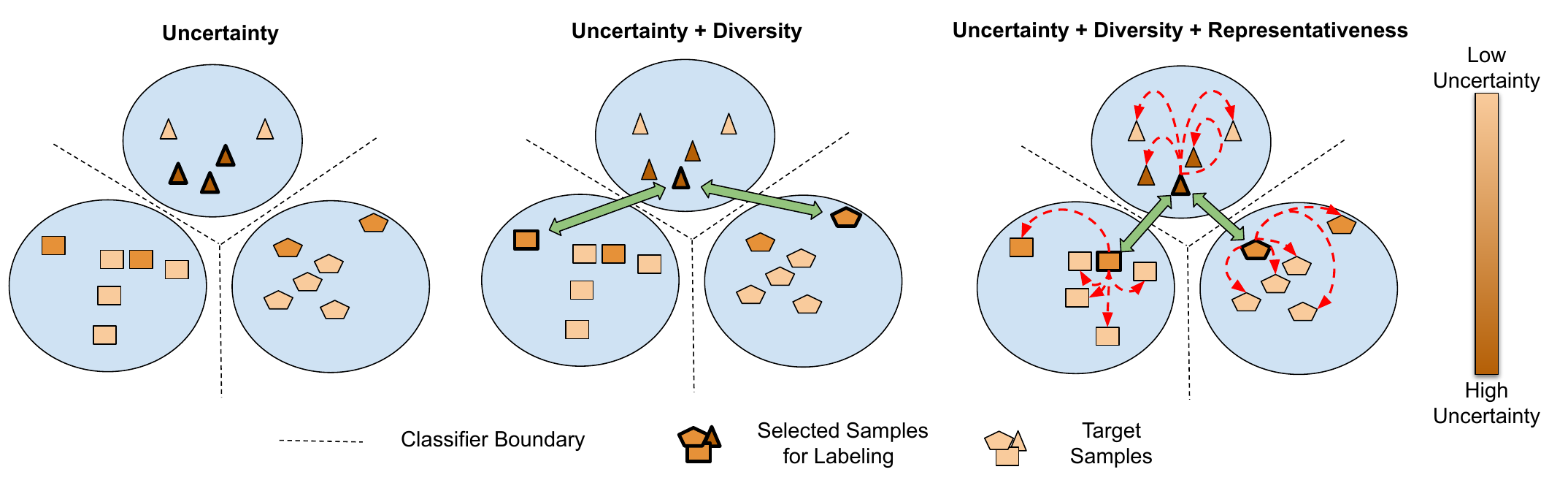}
  \caption{Our sampling technique incorporates uncertainty, diversity and representativeness. Just using {\color{brown} uncertainty} can lead to \textit{redundant} sample selection as shown on left. Whereas, incorporating diversity to ensure a large {\color{YellowGreen} distance} between selected samples may lead to selection of \textit{outliers}. Our sampling technique avoids outliers by selecting uncertain samples which are {\color{red} representative} of the clusters.}
  \label{s3vaada_fig:sampling-al}
\end{figure*}
\label{s3vaada_sec:S^3}

\section{Background}
\subsection{Definitions and Notations}
\label{s3vaada_sec:notations}
\textbf{Definitions:} We first define a set function $f(S)$ for which input is a set $S$. A submodular function is a set function $f:2^{\Omega}\rightarrow \mathbb{R}$, where $2^{\Omega}$ is the power set of set $\Omega$ which contains all elements. The submodular functions are characterized by the property of diminishing returns i.e., addition of a new element to smaller set must produce a larger increase in $f$ in comparison to addition to a larger set. This is mathematically stated as for every $S_1,S_2 \subseteq \Omega$ having $S_1 \subseteq S_2$ then for every $x \in \Omega \backslash S_2$ the following property holds:
\begin{equation}
\begin{split}
    f(S_1 \cup \{x\}) - f(S_1)  \geq  f(S_2\cup \{x\}) - f(S_2). %
\end{split}
\end{equation}
This property is known as the \textit{diminishing returns} property. 
\\
\textbf{Notations Used:} In the subsequent sections we use $h_{\theta}(x)$ as softmax output of the classifier, $h_{\theta}(x)$ is a composition of $f_{\theta} \circ g_{\theta} (x)$ where, $g_{\theta}(x)$ is the function that maps input to embedding and $f_{\theta}$ does final classification. The domain discriminator is a network $D_{\phi}(g_{\theta}(x))$ which classifies the sample into source and target domain which adversarially aligns the domains. We use $\mathcal{D}$ for combined data from both domains and use symbols of $\mathcal{D}_s$ and $\mathcal{D}_t$ for labeled data from source and target domain respectively. $\mathcal{D}_u$ denotes the unlabeled target data. In active DA, we define budget $B$ as number of target samples selected from $\mathcal{D}_u$ and added to $\mathcal{D}_t$ in each cycle.
\\
\textbf{Active Domain Adaptation:}
In each cycle, we first perform DA using $\mathcal{D}_s$ and $\mathcal{D}_t$ as the source and $\mathcal{D}_u$ as the target. Active Learning techniques are then utilized to select $B$ most informative samples from $\mathcal{D}_u$ which is then added to $\mathcal{D}_t$. This is performed for $C$ cycles.

\subsection{Cluster Assumption}
\label{s3vaada_sec:cluster}
Cluster assumption states that the decision boundaries should not lie in high density regions of data samples, which is a prime motivation for our approach. For enforcing cluster assumption we make use of two additional objectives from VADA ~\cite{shu2018dirt} method. The first objective is the minimization of conditional entropy on the unlabeled target data $\mathcal{D}_u$. This is enforced by using the following loss function:
\begin{equation}
    \label{s3vaada_eq:Conditional Entropy}
    L_c(\theta; \mathcal{D}_u) = - \mathbb{E}_{x \sim \mathcal{D}_u}[h_{\theta}(x)^T\ln{  h_{\theta}(x)}].
\end{equation}
The above objective ensures the formation of clusters of target samples, as it enforces high-confidence for classification on target data. 
However due to large capacity of neural networks, the classification function learnt can be locally non Lipschitz which can allow for large change in function value with small change in input. This leads to unreliable estimates of the conditional entropy loss $L_c$. For enforcing the local Lipschitzness we use the second objective, which was originally proposed in Virtual Adversarial Training (VAT) ~\cite{miyato2019vat}. It ensures smoothness in the $\epsilon$ norm ball enclosing the samples. The VAT objective is given below: 
\begin{equation}
    \label{s3vaada_eq: Vat}
    L_{v}(\theta; \mathcal{D}) = \mathbb{E}_{x \sim \mathcal{D}}[\underset{||r||\leq \epsilon}\max{D_{KL}(h_{\theta}(x) || h_{\theta}(x + r))}].
\end{equation}
\section{Proposed Method}
In Active Domain Adaptation, there are two distinct steps i.e., sample selection (i.e. sampling) followed by Domain Adaptation which we describe below: 
\subsection{Submodular Subset Selection}

\subsubsection{Virtual Adversarial Pairwise (VAP) Score}
\label{s3vaada_sec:vapscore}
In our model architecture, we only use a linear classifier and a softmax over domain invariant features $f_\theta(x)$ for classification. Due to the linear nature, we draw inspiration from SVM (Support Vector Machines) based AL methods which demonstrate that the samples near the boundary are likely to be support vectors, hence are more informative than other samples. There is also the theoretical justification behind choosing samples which are near the boundary in case of SVMs \cite{tong2001support}.
Hence we also aim to find the vectors which are near to the boundary by adversarially perturbing each sample $x$. We use the following objective to create perturbation:
\begin{equation}
    \underset{||r_i||\leq \epsilon}{max} D_{KL}(h_{\theta}(x) || h_{\theta}(x + r_i)).
\end{equation}

Power Method \cite{Goluba2000EigenvalueCI} is used for finding the adversarial perturbation $r_i$ which involves initialization of random vector. As we aim to find vectors which are near the decision boundary, there can be cases where a particular sample may lie close to multiple decision boundaries as we operate in the setting of multi-class classification. Hence we create the perturbation $r_i$ for $N$ number of random initializations. This is done to select samples which can be easily perturbed to a diverse set of classes and also increase the reliability of the uncertainty estimate. We use the mean pairwise KL divergence score of probability distribution as a metric for measuring the uncertainty of sample. This is defined as Virtual Adversarial Pairwise (VAP) score given formally as:
\begin{equation} 
    \begin{split}
            VAP(x) = \frac{1}{N^2} \Bigg( \sum_{i = 1}^{N} D_{KL}(h_{\theta}(x) || h_{\theta}(x + r_i))  +\ \sum_{i=1}^{N}\sum_{j=1, i \neq j}^{N} D_{KL}(h_{\theta}(x + r_i) || h_{\theta}(x + r_j)) \Bigg).
    \end{split}
    \label{s3vaada_eq:vap score}
\end{equation}
The first term corresponds to divergence between perturbed input and the original sample $x$, the second term corresponds to diversity in the output of different perturbations. The approach is pictorially depicted on the right side in Fig. \ref{s3vaada_fig:al}. For VAP score to be meaningful we assume that cluster assumption holds and the function is smooth, which makes VAADA a complementary DA approach to our method. 
\subsubsection{Diversity Score}
Just using VAP score for sampling can suffer from the issue of multiple similar samples being selected from the same cluster. For selecting the diverse samples in our set $S$ we propose to use the following diversity score for sample $x_i$ which is not present in $S$.
\begin{equation}
    d(S, x_i) = \underset{x \in S }{\min}D(x, x_i)
\end{equation}
where $D$ is a function of divergence. In our case we use the KL Divergence function:
\begin{equation}
    D(x_j, x_i) = D_{KL}(h_{\theta}(x_j)|| h_{\theta}(x_i)).
\end{equation}
\subsubsection{Representativeness Score}
The combination of above two scores can ensure that the diverse and uncertain samples are selected. But this could still lead to selection of outliers as they can also be uncertain and diversely placed in feature space. For mitigating this we use a term based on facility location problem \cite{enwiki:1012600046} which ensures that selected samples are placed such that they are representative of unlabeled set. The score is mathematically defined as:
\begin{equation}
R(S, x_i) = \underset{x_k \in \mathcal{D}_u}{\sum} \max(0, s_{ki} - \underset{x_j \in S}{\max} \; s_{kj}) 
\end{equation}
here the $s_{ij}$ corresponds to the similarity between sample $x_i$ and $x_j$. We use the similarity function $-\ln(1 - BC(h_{\theta}(x_i),h_{\theta}(x_j))$ where $BC(p,q)$ is the Bhattacharya coefficient~\cite{10.2307/25047882} defined as $\sum_{k} \sqrt{p_kq_k}$ for probability distributions $p$ and $q$.

\subsubsection{Combining the Three Score Functions}
\label{s3vaada_sec: Combinaton}

We define the set function $f(S)$ by defining the gain as a convex combination of $VAP(x_i)$ , $d(S,x_i)$ and $R(S,x_i)$.
\begin{equation}
    \begin{split}
    \label{s3vaada_eq:submod}
     f(S \cup \{x_i\}) - f(S) = \alpha VAP(x_i) + \beta d(S, x_i) 
     +\ (1 - \alpha - \beta) R(S,x_i).
\end{split}
\end{equation}

Here $0\leq \alpha, \beta, \alpha + \beta \leq 1$ are hyperparameters which control relative strength of uncertainty, diversity and representativeness terms. We normalize the three scores before combining them through Eq.~\ref{s3vaada_eq:submod}.\\
\textbf{Lemma 1:} The set function $f(S)$ defined by Eq.~\ref{s3vaada_eq:submod} is submodular. \\
\textbf{Lemma 2:} The set function $f(S)$ defined by Eq.~\ref{s3vaada_eq:submod} is a non decreasing, monotone function.%

We provide proof of the above lemmas in Sec. 2 of App. \ref{app:s3vaada}. Overview of the overall sampling approach is present in Fig. \ref{s3vaada_fig:sampling-al}.

\subsubsection{Submodular Optimization}
\label{s3vaada_sec:submod}
As we have shown in the previous section that the set function $f(S)$ is submodular, we aim to select the set $S$ satisfying the following objective:
\begin{equation}
    \label{s3vaada_eq:submod_obj}
    \underset{S: |S| = B}{max} f(S).
\end{equation}
For obtaining the set of samples $S$ to be selected, we use the greedy optimization procedure. We start with an empty set $S$ and add each item iteratively. For selecting each of the sample ($x_i$) in the unlabeled set, we calculate the gain of each the sample $f(S \cup \{x_i\}) - f(S)$. The sample with the highest gain is then added to set $S$. The above iterations are done till we have exhausted our labeling budget $B$. %
The performance guarantee of the greedy algorithm is given by the following result:
\\
\textbf{Theorem 1}: Let $S^*$ be the optimal set that maximizes the objective in Eq. \ref{s3vaada_eq:submod_obj} then the solution $S$ found by the greedy algorithm has the following guarantee (See Supp. Sec. 2):
\begin{equation}
    f(S) \geq \left(1 - \frac{1}{e}\right)f(S^*).
\end{equation}

\noindent \textbf{Insight for Diversity Component:} The optimization algorithm with $\alpha = 0$ and 
$\beta=1$ degenerates to greedy version of diversity based Core-Set~\cite{sener2016learning} (i.e., $K$-Center Greedy) sampling. Diversity functions based on similar ideas have also been explored for different applications in \cite{joseph2019submodular, chakraborty2014adaptive}. Further details are provided in the Sec. 3 of App. \ref{app:s3vaada}.

\subsection{Virtual Adversarial Active Domain Adaptation}
\label{s3vaada_section:VAADA}
Discriminator-alignment based Unsupervised DA methods fail to effectively utilize the additional labeled data present in target domain \cite{saito2018maximum}. This creates a need for modifications to existing methods which enable them to effectively use the additional labeled target data, and improve generalization on target data. In this work we introduce VAADA (Virtual Adversarial Active Domain Adaptation) which enhances VADA through modification which allow it to effectively use the labeled target data.

We have given our subset selection procedure to select samples to label (i.e., $\mathcal{D}_t$) in Algo. \ref{s3vaada_alg:main_algo} and in Fig. \ref{s3vaada_fig:al}. 
For aligning the features of labeled ($\mathcal{D}_s \cup \mathcal{D}_t$) with $\mathcal{D}_u$, we make use of
domain adversarial training (DANN) loss functions given below:
\begin{equation}
    L_{y}(\theta; \mathcal{D}_s, \mathcal{D}_t) = \mathbb{E}_{(x,y) \sim (\mathcal{D}_s \cup \mathcal{D}_t)}[y^T\; \ln h_{\theta}(x)]
\end{equation}
\begin{equation}
\begin{split}
        L_{d}(\theta; \mathcal{D}_{s}, \mathcal{D}_t, \mathcal{D}_u) = \underset{D_{\phi}}{sup} \; \mathbb{E}_{x \sim \mathcal{D}_s \cup \mathcal{D}_t}[\ln D_{\phi}(f_{\theta}(x))]  +\  \mathbb{E}_{x \sim \mathcal{D}_u} [\ln(1 - D_{\phi}(f_{\theta}(x)))].
\end{split}
\end{equation}
As our sampling technique is based on cluster assumption, for enforcing it we add the Conditional Entropy Loss defined in Eq.~\ref{s3vaada_eq:Conditional Entropy}. Additionally, for enforcing Lipschitz continuity by Virtual Adversarial Training, we use the loss defined in Eq.~\ref{s3vaada_eq: Vat}. 
The final loss is obtained as:
\begin{equation}
\label{s3vaada_eq:final-loss}
\begin{split}
    L(\theta ; \mathcal{D}_{s}, \mathcal{D}_t, \mathcal{D}_u) = L_y(\theta;{\mathcal{D}_s , \mathcal{D}_t})\ +  \lambda_dL_d(\theta; \mathcal{D}_s, \mathcal{D}_t, \mathcal{D}_u)\ + \\ \lambda_sL_{v}(\theta; \mathcal{D}_s \cup \mathcal{D}_t)\ + \lambda_t(L_v(\theta; \mathcal{D}_u) + L_c(\theta; \mathcal{D}_u)).
\end{split}
\end{equation}

The $\lambda$-values used are the \textit{same for all our experiments} are mentioned in the Sec. 5 of App. \ref{app:s3vaada}.

\noindent \textbf{Differences between VADA and VAADA}:
We make certain important changes to VADA listed below, which enables VADA \cite{shu2018dirt}  to effectively utilize the additional supervision of labeled target data and for VAADA procedure:
\\ \textbf{1) High Learning Rate for All Layers}: In VAADA, we use the same learning rate for all layers. In a plethora of DA methods \cite{saito2018maximum, long2018conditional} a lower learning rate for initial layers is used to achieve the best performance. We find that although this practice helps for Unsupervised DA it hurts the Active DA performance (experimentally shown in Sec. 5 of App. \ref{app:s3vaada}).
\\  \textbf{2) Using Gradient Clipping in place of Exponential Moving Average (EMA)}: We use gradient clipping for all network weights to stabilize training whereas VADA uses EMA for the same. We find that clipping makes training of VAADA stable in comparison to VADA and achieves a significant performance increase over VADA.

We find VAADA is able to work robustly across diverse datasets. It has been shown in \cite{saito2018maximum} that VADA, when used out of the box, is unable to get gains in performance when used in setting where target labels are also available for training. This also agrees with our observation that VAADA significantly outperforms VADA in Active DA scenario's (demonstrated in Fig. \ref{s3vaada_fig:training-ablation}, with additional analysis in Sec. 5 of App. \ref{app:s3vaada}).

\begin{algorithm}

\caption{S$^3$VAADA: Submodular Subset Selection for Virtual Adversarial Active Domain Adaptation}
\label{s3vaada_alg:main_algo}
\begin{algorithmic}[1]
\REQUIRE Labeled source $\mathcal{D}_s$; Unlabeled target  $\mathcal{D}_u$; Labeled target $\mathcal{D}_t$; Budget per cycle $B$; Cycles $C$; Model with parameters $\theta$; Parameters $\alpha$, $\beta$
\ENSURE Updated model parameters with improved generalization ability on target domain
\STATE {Train the model according to final loss eq.}
\FOR{cycle $\gets$ 1 to $C$}
\STATE {$S$ $\leftarrow$ $\emptyset$}
\FOR{iter $\gets$ 1 to $B$}
\STATE{$x^* = \underset{x \in \mathcal{D}_u \setminus S}{argmax} \; f(S \cup \{x\}) - f(S)$}
\STATE{$S$ $\leftarrow$ $S$ $\cup$ \{$x^*$\}}
\ENDFOR
\STATE{Get ground truth labels $l_{S}$ for samples in $S$ from oracle}
\STATE{$\mathcal{D}_t \leftarrow \mathcal{D}_t \cup (S, l_{S})$}
\STATE{$\mathcal{D}_u \leftarrow \mathcal{D}_u \setminus S$}
\STATE {Train the model according to final loss eq.}
\ENDFOR
\end{algorithmic}
\end{algorithm}

\section{Experiments}
\label{s3vaada_sec:experiments}
\subsection{Datasets}
We perform experiments across multiple source and target domain pairs belonging to 3 diverse datasets, namely Office-31 \cite{saenko2010adapting}, Office-Home \cite{venkateswara2017Deep}, and VisDA-18 \cite{8575439}. We have specifically not chosen any DA task using real world domain as in those cases the performance maybe higher due to ImageNet initialization not due to adaptation techniques. In \textbf{Office-31} dataset, we evaluate the performance of various sampling techniques on DSLR $\rightarrow$ Amazon and Webcam $\rightarrow$ Amazon, having 31 classes. 
The \textbf{Office-Home} consists of 65 classes and has 4 different domains belonging to Art, Clipart, Product and Real World. We perform the active domain adaptation on Art $\rightarrow$ Clipart, Art $\rightarrow$ Product and Product $\rightarrow$ Clipart. 
\textbf{VisDA-18} image classification dataset consists of two domains (synthetic and real) with 12 classes in each domain. %

\subsection{Experimental Setup}
Following the common practice in AL literature, we first split the target dataset into train set and validation set with 80\% data being used for training and 20\% for validation. 
In all the experiments, we set budget size $B$ as 2\% of the number of images in the target training set, and we perform five cycles ($C$ = 5) of sampling in total. At the end of all cycles, 10\% of the labeled target data will be used for training the model. This setting is chosen considering the practicality of having a small budget of labeling in the target domain and having access to unlabeled data from the target domain. We use ResNet-50 \cite{He_2016_CVPR} as the feature extractor $g_{\theta}$, which is initialized with weights pretrained on ImageNet.
We use SGD Optimizer with a learning rate of 0.01 and momentum (0.9) for VAADA training. 
For the DANN experiments, we follow the same architecture and training procedure as described in \cite{long2018conditional}.
In all experiments, we set $\alpha$ as 0.5 and $\beta$ as 0.3.
We use PyTorch \cite{NEURIPS2019_9015} with automatic mixed-precision training for our implementation. 
Further experimental details are described in the Sec. 6 of  App. \ref{app:s3vaada}.
We report the mean and standard error of the accuracy of the 3 runs with different random seeds.

In AADA \cite{Su_2020_WACV} implementation, the authors have used a different architecture and learning schedule for DANN  which makes comparison of increase in performance due to Active DA, over unsupervised DA intricate. In contrast we use ResNet-50 architecture and learning rate schedule of DANN used in many works \cite{long2018conditional, chen2020adversariallearned}. We first train DANN to reach optimal performance on Unsupervised Domain Adaptation and then start the active DA process. This is done as the practical utility of Active DA is the performance increase over Unsupervised DA. 
\subsection{Baselines}
\begin{figure}[!t]
  \centering
  \includegraphics[width=0.80\linewidth]{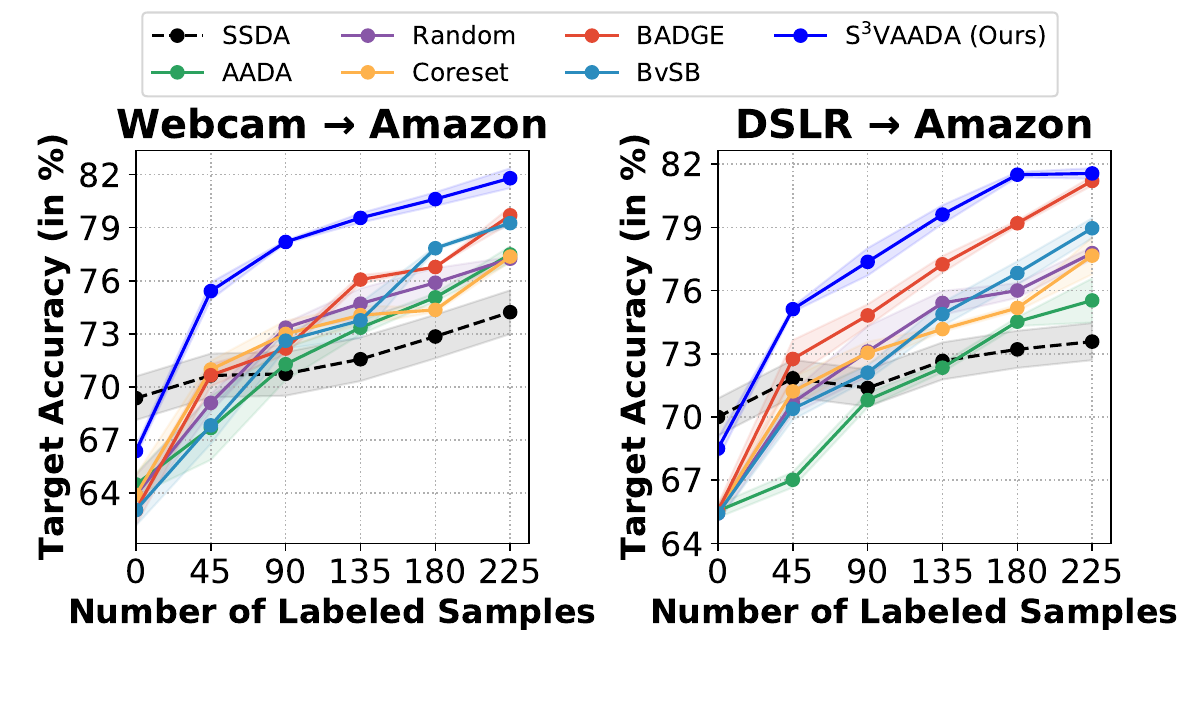}
  \caption{Active DA target accuracy on two adaptation tasks from Office-31 dataset. S$^3$VAADA consistently outperforms BADGE~\cite{Ash2020Deep}, AADA~\cite{Su_2020_WACV} and SSDA (MME$^*$ \cite{saito2019semi}) techniques.}
  \label{s3vaada_fig:office-31-results}
\end{figure}
It has been shown by Su et al. \cite{Su_2020_WACV} that for active DA performing adversarial adaptation through DANN, with adding newly labeled target data into source pool works better than fine-tuning. Hence, we use DANN for all the AL baselines described below: 

\begin{figure*} [!t]
  \centering
  \includegraphics[width=\linewidth]{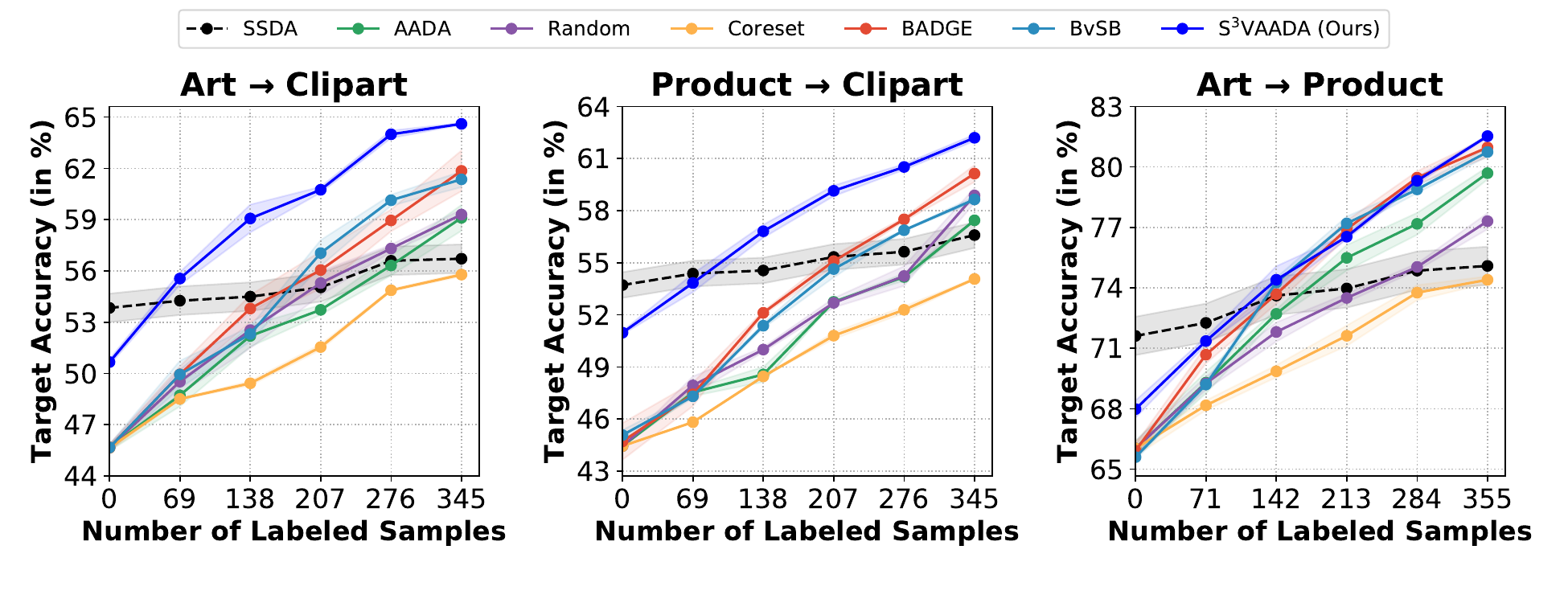}
  \caption{Active DA performance on three different Office-Home domain shifts. We see a significant improvement through S$^3$VAADA in two difficult adaptation tasks of  Art $\rightarrow$ Clipart (left) and Product $\rightarrow$ Clipart (middle) .}
  \label{s3vaada_fig:office-home-results}
\end{figure*}

\begin{enumerate}[noitemsep]
\itemsep0em
    \item \textbf{AADA} (Importance weighted sampling) \cite{Su_2020_WACV}: This state-of-the-art active DA method incorporates uncertainty by calculating entropy and incorporates diversity by using the output of the discriminator.
    \item \textbf{BvSB} (Best vs Second Best a.k.a. margin) \cite{5206627}: It uses the difference between the probabilities of the highest and second-highest class prediction as the the metric of uncertainty, on which low value indicates high uncertainty.
    \item \textbf{BADGE} \cite{Ash2020Deep}: BADGE incorporates uncertainty and diversity by using the gradient embedding on which k-MEANS++ \cite{vassilvitskii2006k} algorithm is used to select diverse samples. BADGE method is currently one of the state-of-the-art methods for AL.
    \item \textbf{$K$-Center (Core-Set)} \cite{sener2018active}: Core-Set selects samples such that the area of coverage is maximized. We use greedy version of Core-Set on the feature space ($g_{\theta}$). It is a diversity-based sampling technique.
    \item \textbf{Random}: Samples are selected randomly from the pool of unlabeled target data.

\end{enumerate}
\noindent \textbf{Semi-Supervised DA:} We compare our method against recent method of MME$^*$~\cite{saito2019semi} with ResNet-50 backbone on Office datasets, using the author's implementation\footnote{https://github.com/VisionLearningGroup/SSDA\_MME}. In each cycle target samples are randomly selected, labeled and provided to the MME$^*$ method for DA.

\subsection{Results}

\begin{figure*}[!t]
\begin{minipage}{0.73\linewidth}
    \centering
    \subcaptionbox{Uncertainity (\textbf{U})\label{s3vaada_fig:subfig1}}[0.3\linewidth]{\includegraphics[width=\linewidth, height=4cm]{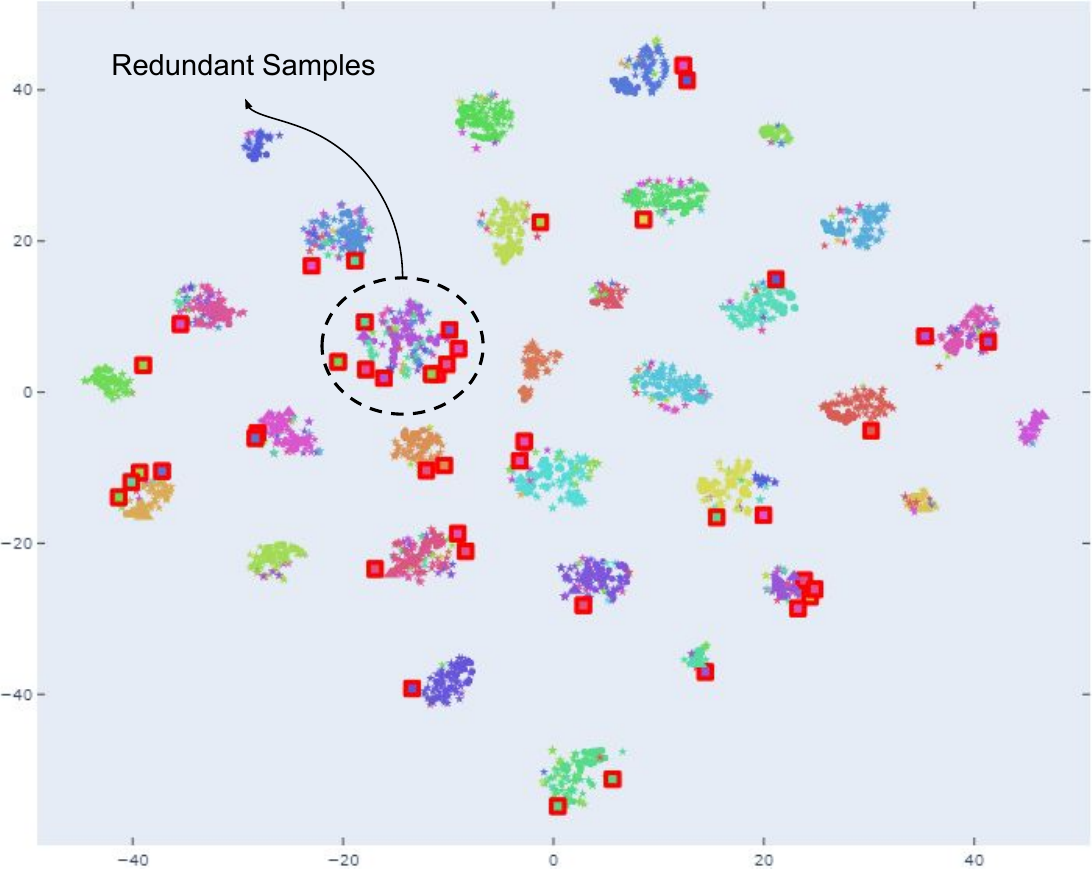}}
    \subcaptionbox{Diversity (\textbf{D})\label{s3vaada_fig:subfig2}}[0.3\linewidth]{\includegraphics[width=\linewidth,height=4cm]{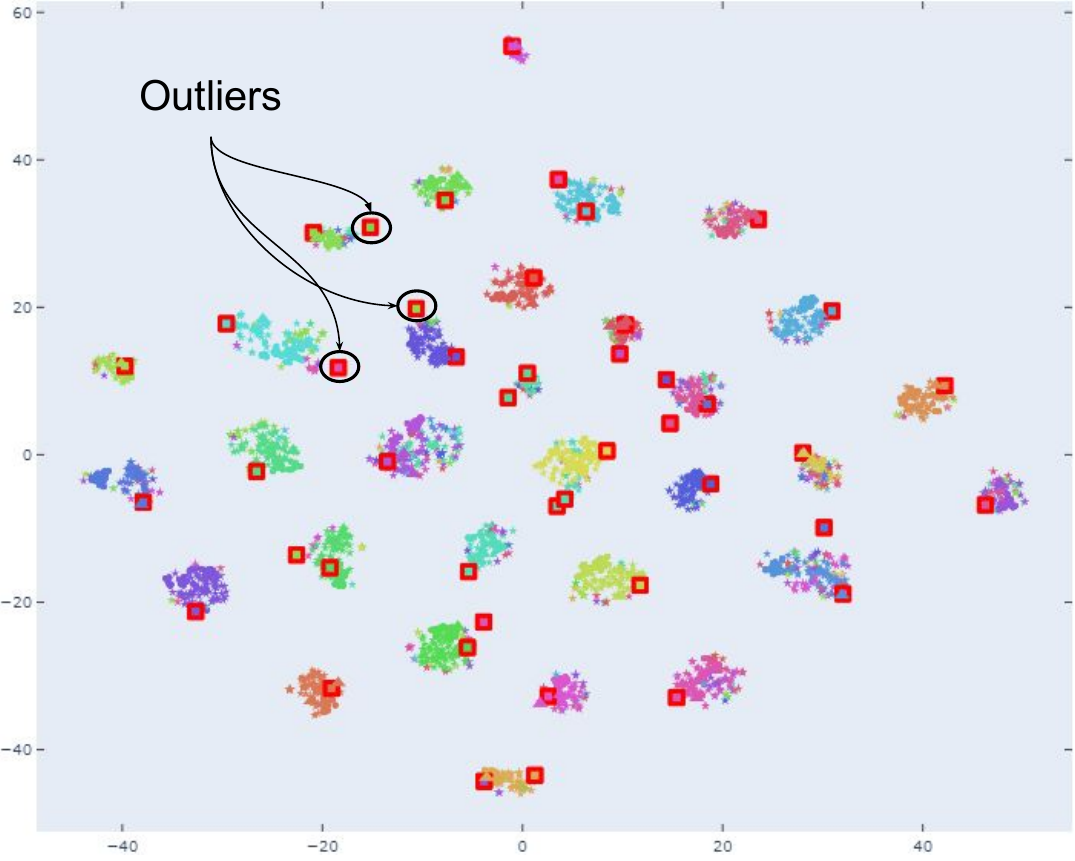}}
    \subcaptionbox{Representativeness (\textbf{R})\label{s3vaada_fig:subfig3}}[0.3\linewidth]{\includegraphics[width=\linewidth,height=4cm]{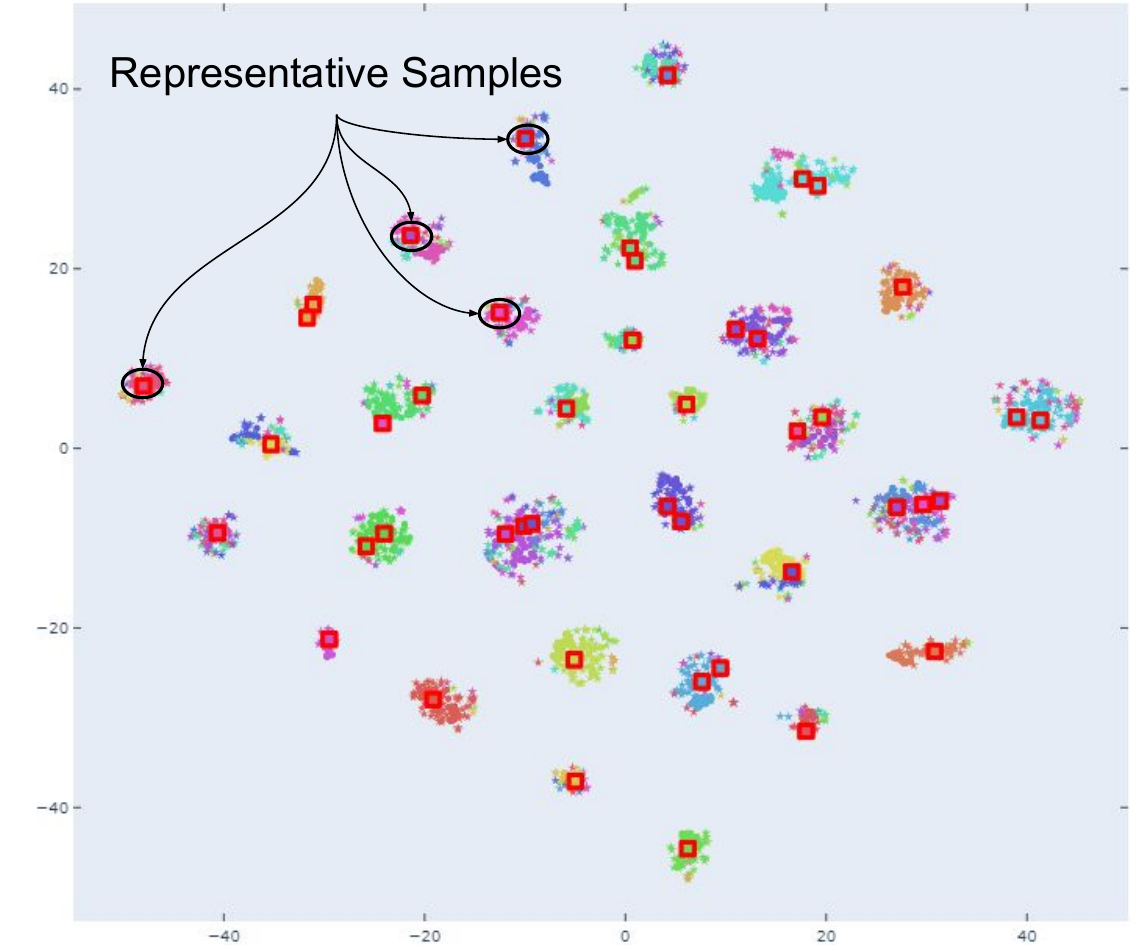}}
    \caption{Feature space visualization using t-SNE with selected samples in Red. Using \textbf{Uncertainty} leads to \textit{redundant} samples from same cluster, whereas using \textbf{Diversity} leads to only diverse boundary samples being selected which maybe \textit{outliers}. Sampling using \textbf{Representativeness} prefers samples near the cluster center, hence we use a combination of these complementary criterion as our criterion.}
    \label{s3vaada_fig:tsne-plot}
\end{minipage}%
\hspace{1mm}
\begin{minipage}{0.24\textwidth}
    \centering
    \includegraphics[width=\linewidth]{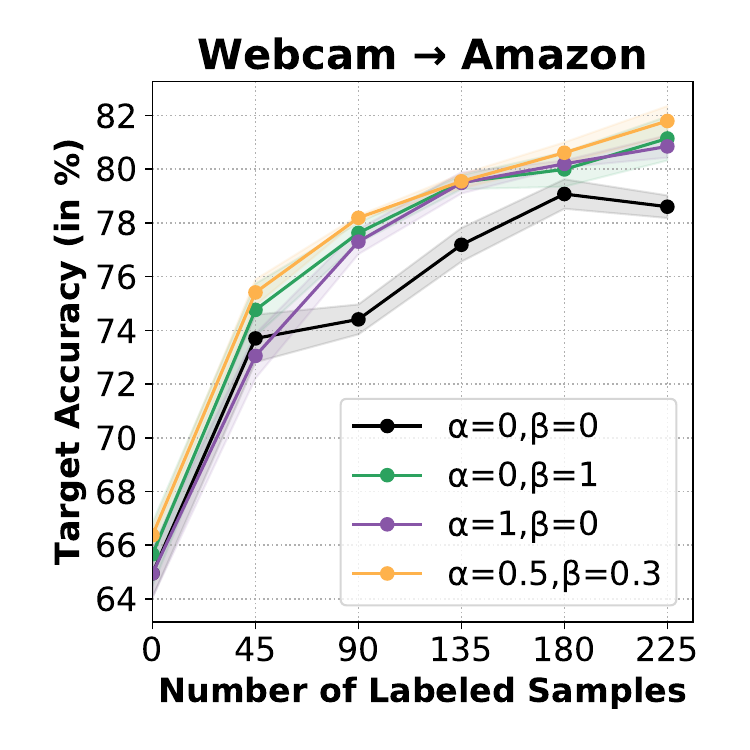}
    \caption{Trade off between Uncertainty, Diversity and Representativeness (i.e., Parameter sensitivity to $\alpha, \beta$).}
    \label{s3vaada_fig:alpha-beta}
\end{minipage}
\end{figure*}

\begin{figure}[!t]
  \centering
  \includegraphics[width=0.8\linewidth]{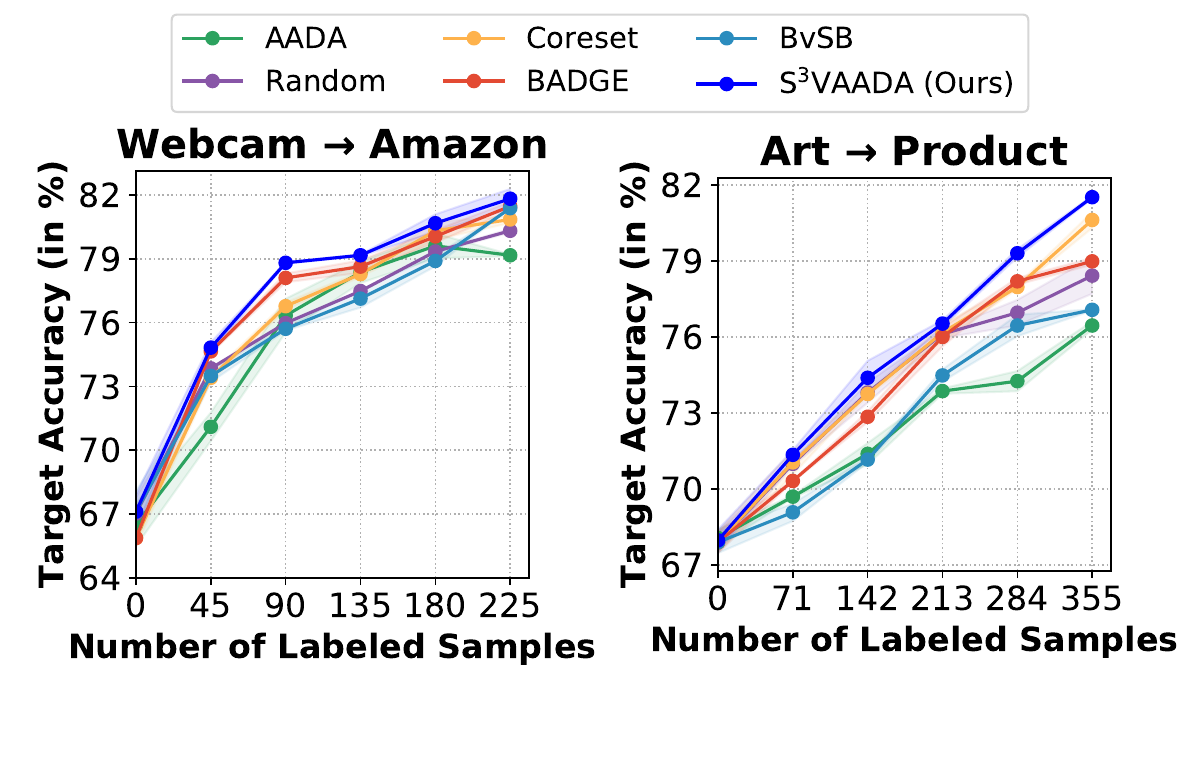}
  \caption{Ablation on sampling methods on different domain shifts. In both cases, we train the sampling techniques via VAADA. Our method (S$^3$VAADA) consistently outperforms all other sampling methods.}
  \label{s3vaada_fig:sampling-ablation}
\end{figure}

\begin{figure}[!t]
  \centering
  \includegraphics[width=0.4\linewidth]{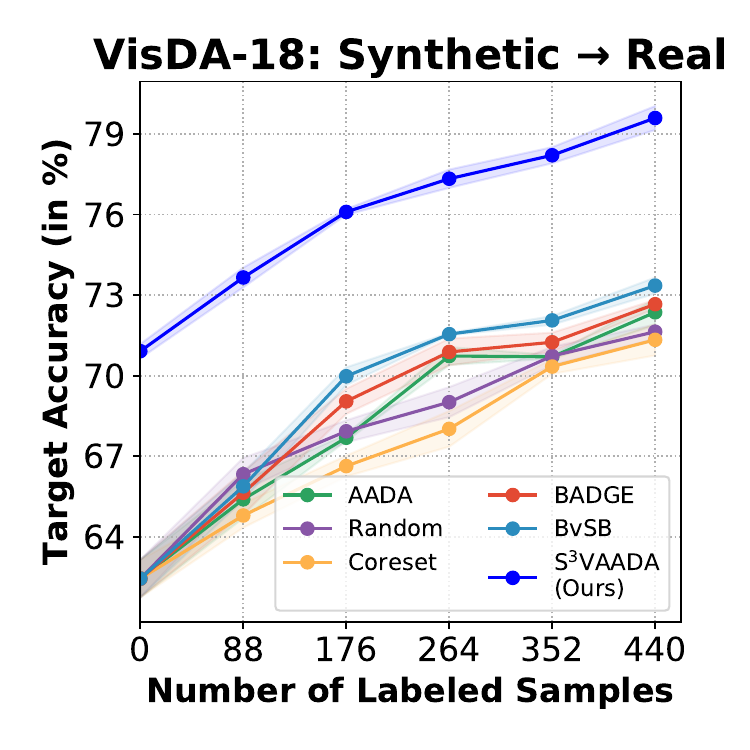}
  \caption{Active DA Results on VisDA-18 dataset.}
  \label{s3vaada_fig:visda-results}
\end{figure}

\label{s3vaada_sec:results}
Fig. \ref{s3vaada_fig:office-31-results} shows the results on Office-31 dataset, S$^3$VAADA outperforms all other techniques. On Webcam $\rightarrow$ Amazon shift, it shows significant improvement of 9\% in the target accuracy with just 45 labeled samples. S$^3$VAADA gets 81.8\% accuracy in the last cycle which is around 15\% more than the unsupervised DA performance, by using just 10\% of the labeled data. On DSLR $\rightarrow$ Amazon shift, VAADA follows a similar trend and performs better than all other sampling techniques. On Office-Home dataset on the harder domain shifts of Art $\rightarrow$ Clipart and Product $\rightarrow$ Clipart, S$^3$VAADA produces a significant increase of 3\%-5$\%$ and 2\%-6\% respectively across cycles, in comparison to other methods (Fig. \ref{s3vaada_fig:office-home-results}). On the easier Art $\rightarrow$ Product shift, our results are comparable to other methods.

\noindent \textbf{Large Datasets:} On the VisDA-18 dataset, where the AADA method is shown to be ineffective \cite{Su_2020_WACV} due to a severe domain shift. Our method (Fig. \ref{s3vaada_fig:visda-results}) is able to achieve significant increase of around 7\% averaged across all cycles, even in this challenging scenario. For demonstrating the scalability of our method to DomainNet~\cite{peng2019moment}, we also provide the results of one adaptation scenario in Sec. 10 of App. \ref{app:s3vaada}. 

\noindent \textbf{Semi-Supervised DA:} From Figs. \ref{s3vaada_fig:office-31-results} and \ref{s3vaada_fig:office-home-results} it is observed that performance of MME$^*$ saturates as more labeled data is added, in contrast, S$^3$VAADA continues to improve as more target labeled data is added. 

\section{Analysis of S$^3$VAADA}
\textbf{Visualization:} Fig. \ref{s3vaada_fig:tsne-plot} shows the analysis of samples selected for uncertainty \textbf{U}, diversity \textbf{D} and representativeness \textbf{R} criterion, which depicts the \textit{complementary} preferences of the three criterion. \\
\textbf{Sensitivity to $\alpha$ and $\beta$ }: Fig. \ref{s3vaada_fig:alpha-beta} shows experiments for probing the effectiveness of each component (i.e., uncertainty, diversity and representativeness) in the information criteria.
We find that just using Uncertainty ($\alpha = 1$) and Diversity ($\beta = 1$) provide reasonable results when used individually. However, the individual performance remain sub-par with the hybrid combination (i.e., $\alpha = 0.5$, $\beta=0.3$). We use value of $\alpha=0.5$ and $\beta=0.3$ across all our experiments, hence our sampling does not require parameter-tuning specific to each dataset.

\noindent \textbf{Comparison of Sampling Methods:}
For comparing the different sampling procedures we fix the adaptation technique to VAADA and use different sampling techniques. Fig. \ref{s3vaada_fig:sampling-ablation} shows that our sampling method outperforms others in both cases. In general we find that hybrid approaches i.e., Ours and BADGE perform \textit{robustly} across domain shifts.

\begin{figure}[!t]
  \centering
  \includegraphics[width=0.8\linewidth]{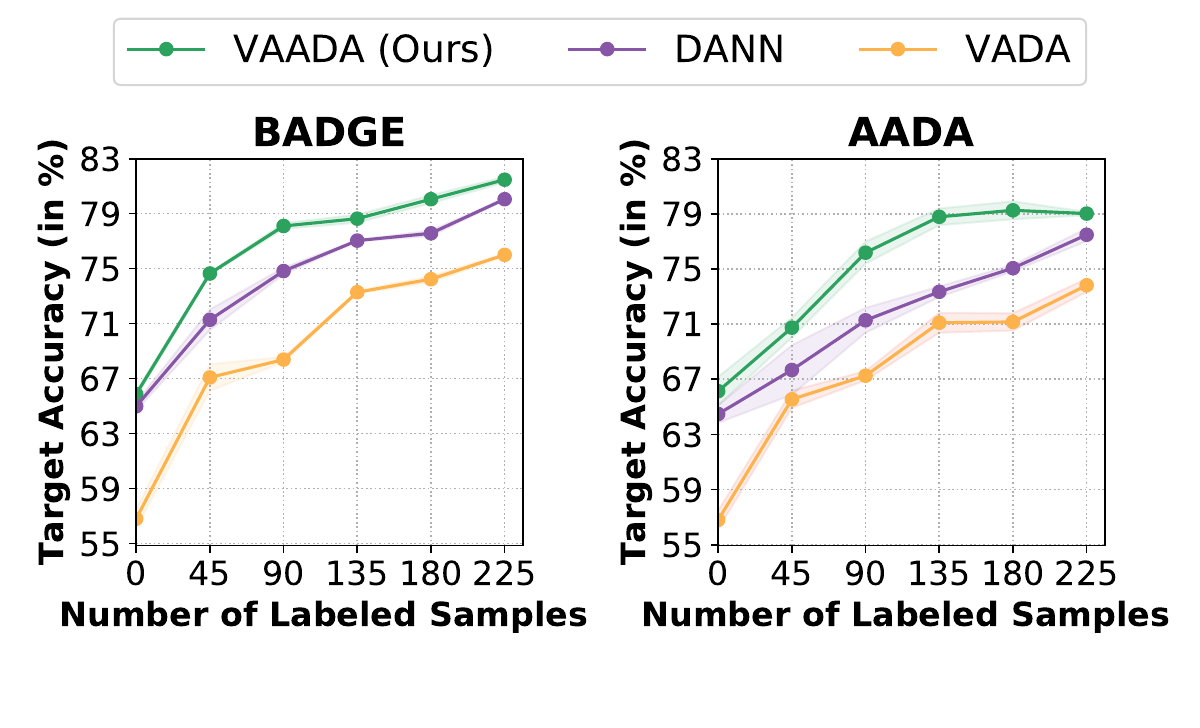}
  \caption{Comparison of different DA methods for Active DA on Webcam $\rightarrow$ Amazon.}
  \label{s3vaada_fig:training-ablation}
\end{figure}

\noindent \textbf{Comparison of VAADA:} Fig. \ref{s3vaada_fig:training-ablation} shows performance of VAADA, DANN and VADA when used as adaptation procedure for two sampling techniques. We find that a significant improvement occurs for all the sampling techniques in each cycle for VAADA comparison to DANN and VADA. The $\geq 5\%$ improvement in each cycle,  shows the importance of proposed improvements in VAADA over VADA.

We provide additional analysis on \textit{convergence, budget size and hyper parameters} in the Sec. 4 of App. \ref{app:s3vaada}. Across our analysis we find that S$^3$VAADA works robustly in various scenarios.

\section{Conclusion}
We formulate the sample selection in Active DA as optimal informative subset selection problem, for which we propose a novel submodular information criteria. The information criteria takes into account the uncertainty, diversity and representativeness of the subset. The most informative subset obtained through submodular optimization is then labeled and used by the proposed adaptation procedure VAADA. We find that the optimization changes introduced for VAADA significantly improve Active DA performance across all sampling schemes. The above combination of sampling and adaptation procedure constitutes S$^3$VAADA, which consistently provides improved results over existing methods of Semi-Supervised DA and Active DA.  \\

\newpage

 \chapter{A Closer Look at Smoothness in Domain Adversarial Training}
\label{chap:SDAT}

\begin{changemargin}{7mm}{7mm} 

Domain adversarial training has been ubiquitous for achieving invariant representations and is used widely for various domain adaptation tasks. In recent times, methods converging to smooth optima have shown improved generalization for supervised learning tasks like classification.  In this work, we analyze the effect of smoothness enhancing formulations on domain adversarial training, the objective of which is a combination of {task loss (eg.\ classification, regression etc.)} and adversarial terms. We find that converging to a smooth minima with respect to (w.r.t.) task loss stabilizes the adversarial training leading to better performance on target domain. In contrast to {task} loss, our analysis shows that {converging to smooth minima w.r.t. adversarial loss leads to sub-optimal generalization on the target domain}. Based on the analysis, we introduce the Smooth Domain Adversarial Training (SDAT) procedure, which effectively enhances the performance of existing domain adversarial methods for both classification and object detection tasks.  Our analysis also provides insight into the extensive usage of SGD over Adam in the community for domain adversarial training. 

\end{changemargin}

\begin{figure*}[!t]
  \centering
  \includegraphics[width=0.95\textwidth]{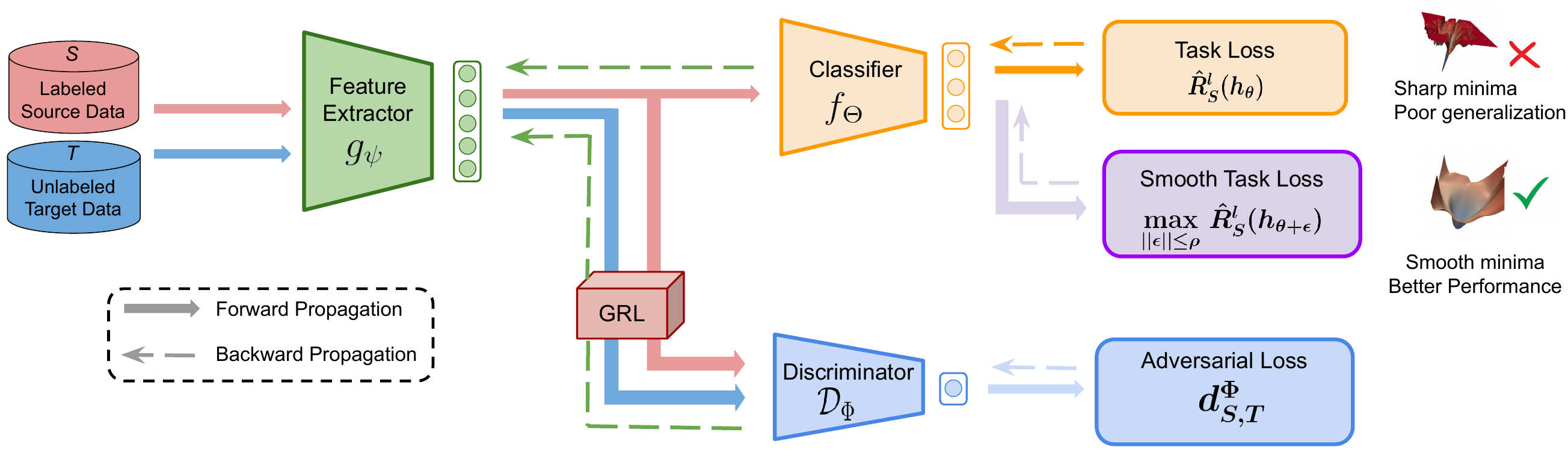}
  \caption{Overview of Smooth Domain Adversarial Training (SDAT). We demonstrate that converging to smooth minima w.r.t. adversarial loss leads to sub-optimal DAT. Due to this conventional approaches which smooth combination of task loss and adversarial loss lead to sub-optimal results. Hence, we propose SDAT which only focuses on smoothing {task} loss, leading to stable training which results in effective generalization on target domain.  }
  \label{sdat_fig:overview}
\end{figure*}
\section{Introduction}
    Domain Adversarial Training \citep{ganin2015unsupervised} (DAT) refers to adversarial learning of neural network based feature representations that are invariant to the domain. For example, the learned feature representations for car images from the Clipart domain should be similar to that from the Web domain. DAT has been widely useful in diverse areas (cited 4200 times) such as recognition \citep{long2018conditional,cui2020gvb,Rangwani_2021_ICCV}, fairness \citep{adel2019one}, object detection \citep{saito2019strong}, domain generalization \citep{li2018domain}, image-to-image translation \citep{liu2017unsupervised} etc. The prime driver of research on DAT is its application in unsupervised Domain Adaptation (DA), which aims to learn a classifier using labeled source data and unlabeled target data, such that it generalizes well on target data. Various enhancements like superior objectives \citep{acuna2021f, zhang2019bridging}, architectures \citep{long2018conditional}, etc. have been proposed to improve the effectiveness of DAT for unsupervised DA.

As DAT objective is a combination of Generative Adversarial Network (GAN) \citep{goodfellow2014generative} (adversarial loss)  and Empirical Risk Minimization (ERM) \citep{vapnik2013nature} (task loss) objectives, there has not been much focus on explicitly analyzing and improving the nature of optimization in DAT. In optimization literature, one direction that aims to improve the generalization focuses on developing algorithms that converge to a smooth (or a flat) minima \citep{foret2021sharpnessaware, keskar2017improving}. However, we find that these techniques, when directly applied for DAT, do not significantly improve the generalization on the target domain (Sec. \ref{sdat_sec:discussion}). \\ \\
In this work, we analyze the loss landscape near the optimal point obtained by DAT to gain insights into the nature of optimization. We first focus on the eigen-spectrum of Hessian (i.e.\ curvature) of the  {task loss (ERM term for classification)} where we find that using Stochastic Gradient Descent (SGD) as optimizer converges to a smoother minima in comparison to Adam \citep{kingma2014adam}. Further, we find that \textit{smoother minima w.r.t. {task} loss results in stable DAT leading to better generalization on the target domain}. Contrary to {task} loss, we find that smoothness enhancing formulation for adversarial components worsens the performance, rendering techniques \cite{cha2021swad} which enhance smoothness for all loss components ineffective. Hence we introduce Smooth Domain Adversarial Training (SDAT) (Fig. \ref{sdat_fig:overview}), which aims to reach a smooth minima only w.r.t. task loss and helps in generalizing better on the target domain. SDAT requires an additional gradient computation step and can be combined easily with existing methods. 
We show the soundness of the SDAT method theoretically through a generalization bound (Sec.\ \ref{sdat_smoothness}) on target error. We extensively verify the empirical efficacy of SDAT over DAT across various datasets for classification (i.e., DomainNet, VisDA-2017 and Office-Home) with ResNet and Vision Transformer \cite{dosovitskiy2020image} (ViT) backbones. We also show a prototypical application of SDAT in DA for object detection, demonstrating it's diverse applicability. In summary, we make the following contributions:
\begin{itemize}
    \item We demonstrate that converging to smooth minima w.r.t. task loss leads to stable and effective domain alignment through DAT, whereas smoothness enhancing formulation for adversarial loss leads to sub-optimal performance via DAT.
    \item For enhancing the smoothness w.r.t. task loss near optima in DAT, we propose a simple, novel, and theoretically motivated SDAT formulation that leads to stable DAT resulting in improved generalization on the target domain. 
    \item We find that SDAT, when combined with the existing state-of-the-art (SOTA) baseline for DAT, leads to significant gains in performance. Notably, with ViT backbone, SDAT leads to a significant effective average gain of \textbf{3.1\%} over baseline, producing SOTA DA performance without requiring any additional module (or pre-training data) using only a 12 GB GPU. The source code used for experiments is available at: \url{https://github.com/val-iisc/SDAT}.
\end{itemize}

\section{Related Work}
\footnotetext{Figures for the smooth minima and sharp minima are from \citep{foret2021sharpnessaware} and used for illustration purposes only.}
{ \textbf{Unsupervised Domain Adaptation}: It refers to a class of methods that enables the model to learn representations from the source domain's labeled data that generalizes well on the unseen data from the target domain \citep{long2018conditional, acuna2021f, zhang2019bridging, Kundu_2021_ICCV, Kundu_2020_CVPR}. One of the most prominent lines of work is based on DAT \citep{ganin2015unsupervised}. 
It involves using an additional discriminator to distinguish between source and target domain features. A Gradient Reversal layer (GRL) is introduced to achieve the goal of learning domain invariant features. The follow-up works have improved upon this basic idea by introducing a class information-based discriminator (CDAN \citep{long2018conditional}), introducing a transferable normalization function \citep{wang2019transferable}, using an improved Margin Disparate Discrepancy~\cite{zhang2019bridging} measure between source and target domain, etc. In this work, we focus on analyzing and improving such methods.}\\

\noindent{\textbf{Smoothness of Loss Landscape:} As neural networks operate in the regime of over parameterized models, low error on training data does not always lead to better generalization \citep{keskar2017large}. Often it has been stated \citep{hochreiter1997flat, hochreiter1994simplifying, he2019asymmetric, dziugaite2017computing}  that smoother minima does generalize better on unseen data. But until recently, this was practically expensive as smoothing required additional costly computations. Recently, a method called Sharpness Aware Minimization (SAM) \citep{foret2021sharpnessaware} for improved generalization has been proposed which finds a smoother minima with an additional gradient computation step. However, we find that just using SAM naively does not lead to improved generalization on target domain (empirical evidence in Tab. \ref{sdat_tab:diff_smooth},\ref{sdat_tab:officehome_erm} and \ref{sdat_table:visda_uda_vit_erm}). In this work, we aim to develop solutions which converge to a smooth minima but at same time lead to better generalization on target domain, which is not possible just by using SAM. }
\begin{figure*}[!t]
\centering
\begin{tabular}{c c c c}
\hspace{-0.4cm}\includegraphics[width=0.23\linewidth, height = 3.7cm]{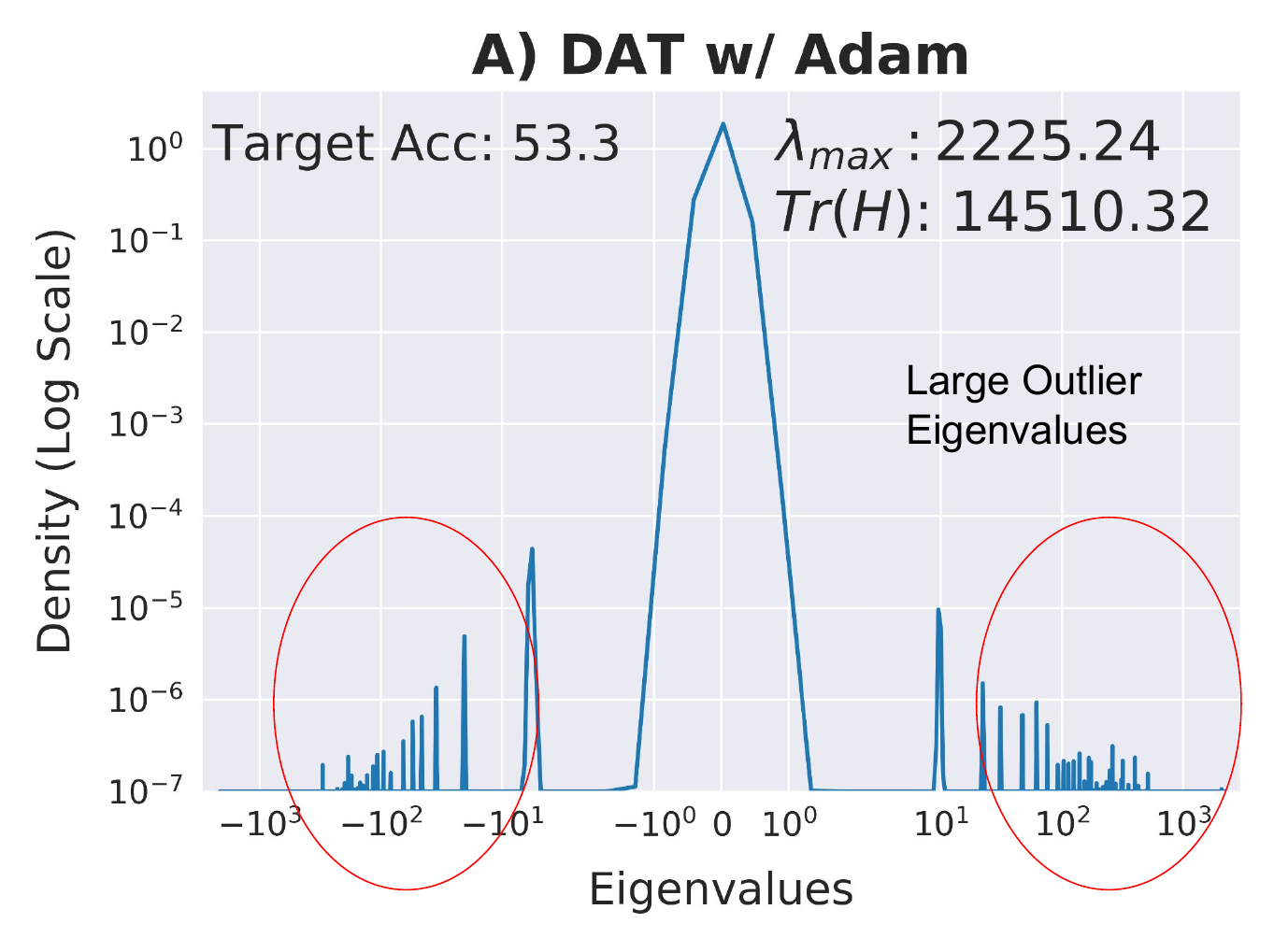} &   \hspace{-0.4cm}\includegraphics[width=0.23\linewidth, height = 3.7cm]{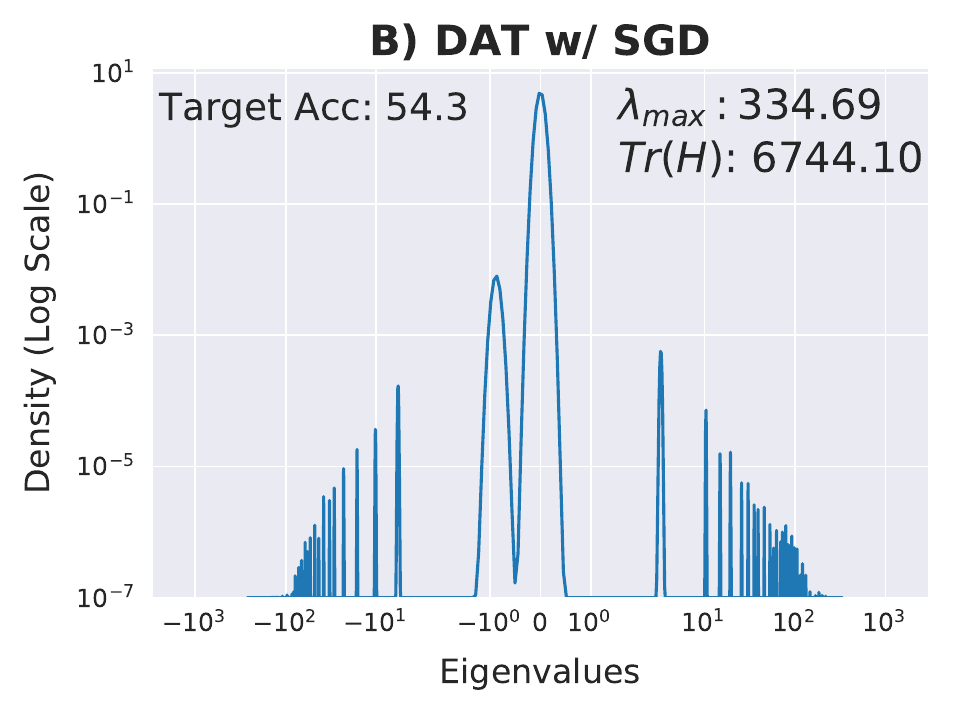} &
\hspace{-0.4cm}\includegraphics[width=0.23\linewidth, height = 3.7cm]{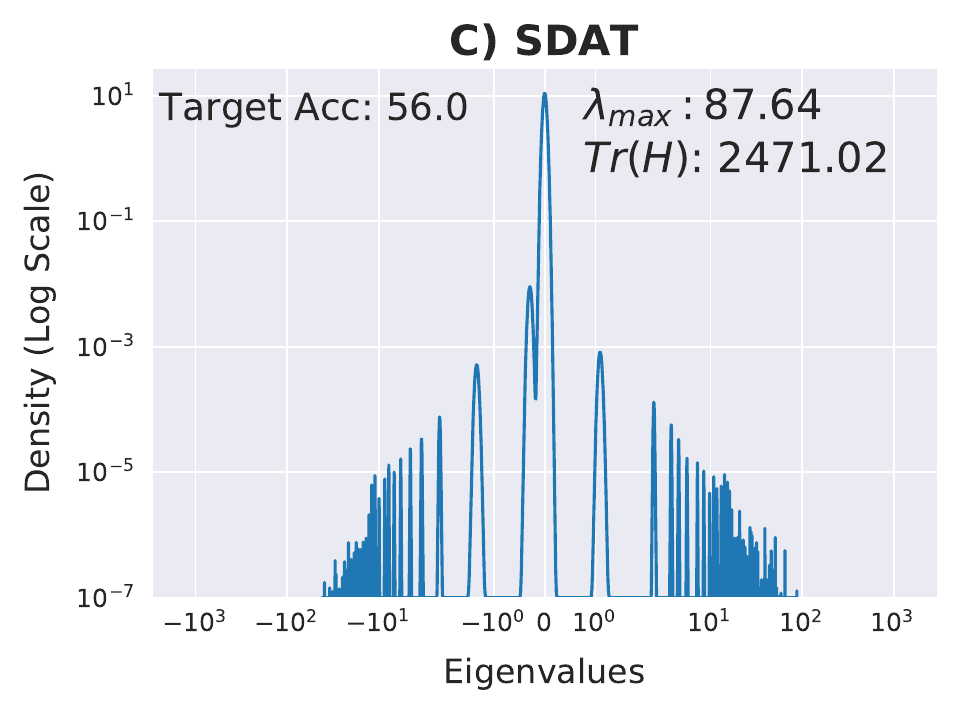} & \hspace{-0.4cm}\includegraphics[width=0.23\linewidth, height = 3.7cm]{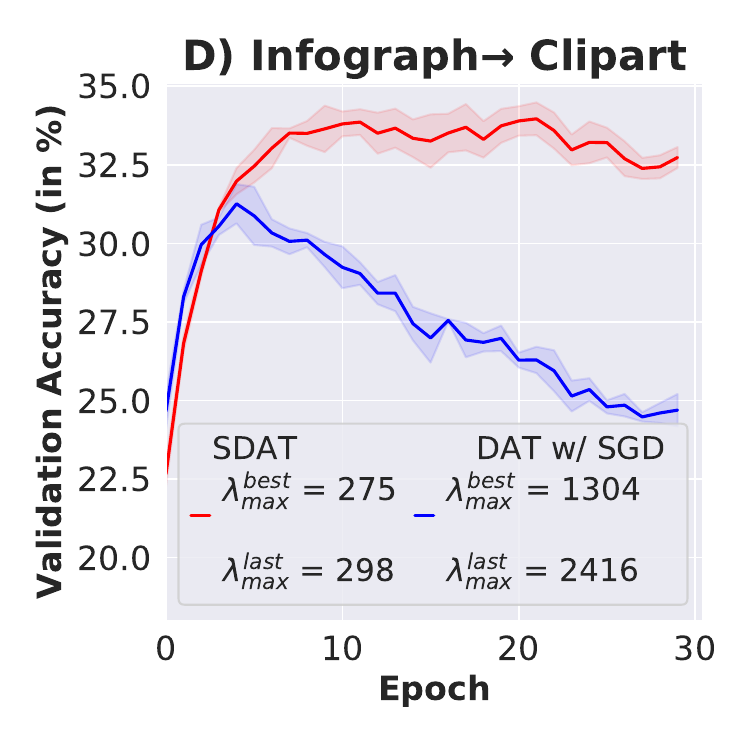} \\
\end{tabular}
\caption{Eigen Spectral Density plots of Hessian ($\nabla^2 \hat{R}_{S}^l(h_{\theta})$) for Adam (\textbf{A}), SGD (\textbf{B}) and SDAT (\textbf{C}) on Art $\veryshortarrow$ Clipart. Each plot contains the maximum eigenvalue ($\lambda_{max}$) and the trace of the Hessian ($Tr(H)$), which are indicators of the smoothness (Low $Tr(H)$ and $\lambda_{max}$ indicate the presence of smoother loss surface). Low range of eigenvalues (x-axis), $Tr(H)$ and $\lambda_{max}$ for SGD compared to Adam indicates that it reaches a smoother minima, which leads to a higher target accuracy. D) Validation accuracy and $\lambda_{\max}$ comparison for SDAT and DAT across epochs, SDAT shows significantly stable training with low $\lambda_{\max}$. }
\label{sdat_fig:hessian}
\end{figure*}
\section{Background}
\subsection{Preliminaries} 
We will primarily focus on unsupervised DA where we have labeled source data $S = \{ (x_i^s, y_i^s) \} $ and unlabeled target data $T = \{ (x_i^t) \}$. The source samples are assumed to be sampled i.i.d. from source distribution $P_S$ defined on input space $\mathcal{X}$, similarly target samples are sampled i.i.d. from $P_T$. $\mathcal{Y}$ is used for denoting the label set which is $\{1, 2,  \dots, k\}$ in our case as we perform multiclass ($k$) classification. We denote $y : \mathcal{X} \rightarrow \mathcal{Y}$ a mapping from images to labels. Our task is to find a hypothesis function $h_{\theta}$ that {has a low risk on the target distribution}. The source risk (a.k.a expected error) of the hypothesis $h_{\theta}$ is defined with respect to loss function $l$ as: $R_{S}^{l}(h_{\theta}) = \mathbb{E}_{x\sim P_S}[l(h_{\theta}(x), y(x))]$. The target risk $R_T^l(h_\theta)$ is defined analogously. The empirical versions of source and target risk will be denoted by $\hat{R}_S^l(h_\theta)$ and $\hat{R}_T^l(h_\theta)$. All notations used in chapter are summarized in App. \ref{sdat_app:notn_tab}. In this work we build on the DA theory of \citep{acuna2021f} {which is a generalization of~\citet{ben2010theory}}. 
We first define the discrepancy between the two domains.

\begin{definition}[$ D_{h_{\theta},\mathcal{H}}^\phi$ discrepancy] The discrepancy between two domains $P_S$ and $P_T$ is defined as following:
\begin{equation}
    \begin{split}
    D_{h_{\theta}, \mathcal{H}}^{\phi}(P_S || P_T) := \sup_{h' \in \mathcal{H}} [\mathbb{E}_{x \sim P_S}[l(h_{\theta}(x),h'(x))]] -  [\mathbb{E}_{x \sim P_T}[\phi^*(l(h_{\theta}(x),h'(x)))]]
\end{split}
\end{equation}
here $\phi^*$ is a frenchel conjugate of a lower semi-continuous convex function $\phi$  that satisfies $\phi(1) = 0$, and $\mathcal{H}$ is the set of all possible hypothesis (i.e. Hypothesis Space).
\end{definition}

This discrepancy distance $D_{h_{\theta}, \mathcal{H}}^\phi$ is based on variational formulation of f-divergence \citep{nguyen2010estimating} for the convex function $\phi$. The $D_{h_{\theta}, \mathcal{H}}^{\phi}$ is the lower bound estimate of the f-divergence function $D^{\phi}(P_S||P_T)$ (Lemma 4 in \citep{acuna2021f}). We state a bound on target risk $R_{T}^l(h_{\theta})$ based on $\mathcal D_{h_{\theta},\mathcal{H}}^\phi$ discrepancy \citep{acuna2021f}:
\begin{theorem}[\textbf{Generalization bound}]
\label{sdat_th:gen-bound}
Suppose $l: \mathcal{Y} \times \mathcal{Y} \rightarrow [0,1] \subset dom \; \phi^*$. Let $h^*$ be the ideal joint classifier with {least} $\lambda^* = R_S^l(h^*) +  R_T^l(h^*)$ {(i.e. joint risk)} in $\mathcal{H}$. We have the following relation between source and target risk:
\begin{equation}
    R_{T}^l(h_{\theta}) \leq R_{S}^{l}(h_{\theta}) + D_{h_{\theta}, \mathcal{H}}^{\phi} (P_S || P_T) + \lambda^*.
\end{equation}
\end{theorem}
The above generalization bound shows that the target risk $R_T^l(h_{\theta})$ is upper bounded by the source risk $R_S^l(h_{\theta})$ and the discrepancy term $D_{h_{\theta}, \mathcal{H}}^\phi$ along with an irreducible constant error $\lambda^*$. Hence, this infers that reducing source risk and discrepancy lead a to reduction in target risk. Based on this, we concretely define the unsupervised adversarial adaptation procedure in the next section.

\subsection{Unsupervised Domain Adaptation}
In this section we first define the components of the framework we use for our purpose: $h_{\theta} = f_{\Theta} \circ g_{\psi}$ where $g_{\psi}$ is the feature extractor  and $f_{\Theta}$ is the classifier. The domain discriminator $\mathcal{D}_{\Phi}$, used for estimating the discrepancy between $P_S$ and $P_T$ is a classifier whose goal is to distinguish between the features of two domains. For minimizing the target risk (Th. \ref{sdat_th:gen-bound}), the optimization problem is as follows:
\begin{equation}
    \underset{\theta}{\min} \; \mathbb{E}_{x \sim P_S}[l(h_{\theta}(x), y(x))] + D_{h_{\theta}, \mathcal{H}}^{\phi}(P_S||P_T)
    \label{sdat_eq:uda_obj_min}
\end{equation}
The discrepancy term under some assumptions (refer App. \ref{sdat_app:discrepancy}) can be upper bounded by a tractable term:
\begin{equation}
\begin{split}
        D_{h_{\theta}, \mathcal{H}}^{\phi}(P_S||P_T) \leq \underset{\Phi}{\max} \; d_{S,T}^{\Phi}
\end{split}
\label{sdat_eq:diver_disc}
\end{equation}
where 
$d_{S,T}^{\Phi} = \mathbb{E}_{x \sim P_S}[\log(\mathcal{D}_{\Phi}(g_{\psi}(x)))] + \mathbb{E}_{x \sim P_T}\log[1 - \mathcal{D}_{\Phi}(g_{\psi}(x))]$.
This leads to the final optimization objective of:
\begin{equation}
        \underset{\theta}{\min} \; \underset{\Phi}{\max} \; \mathbb{E}_{x \sim P_S}[l(h_{\theta}(x), y(x))] +  \; d_{S,T}^{\Phi}
\label{sdat_eq:uda_obj}
\end{equation}
The first term in practice is empirically approximated by using finite samples $\hat{R}_S^l(h_\theta)$ and used as {task} loss (classification) for minimization. The empirical estimate of the second term is adversarial loss which is optimized using GRL as it has a min-max form. (Overview in Fig. \ref{sdat_fig:overview}) The above procedure composes DAT, and we use CDAN \citep{long2018conditional} as our default DAT method.

\section{Analysis of Smoothness} \label{sdat_smoothness}
In this section, we analyze the curvature properties of the task loss with respect to the parameters ($\theta$). Specifically, we focus on analyzing the Hessian of empirical source risk $H = \nabla^2_{\theta} \hat{R}_S^l(h_{\theta})$ which is the Hessian of classification ({task}) loss term. For quantifying the smoothness, we measure the trace $Tr(H)$ and maximum eigenvalue of Hessian ($\lambda_{max}$) as a proxy for quantifying smoothness. 
This is motivated by analysis of 
which states that the low value of $\lambda_{max}$ and $Tr(H)$ are indicative of highly smooth loss landscape \citep{Jastrzebski2020The}. Based on our observations we articulate our conjecture below: 
\begin{conjecture}
  Low $\lambda_{\max}$ for Hessian  of empirical source risk (i.e. task loss) $\nabla^2_{\theta}\hat{R}_S^l(h_{\theta})$ leads to stable and effective DAT, resulting in reduced risk on target domain  $\hat{R}^l_{T}(h_{\theta})$.
  
\end{conjecture}

\begin{figure*}[t]
\centering
\begin{tabular}{c c c}
\includegraphics[width=0.3\linewidth]{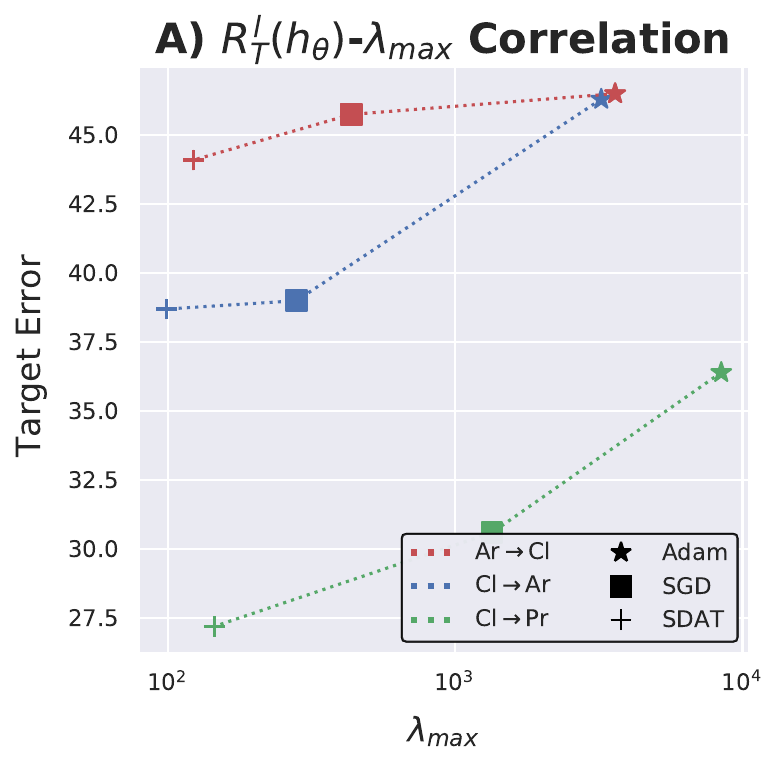} &   \includegraphics[width=0.3\linewidth]{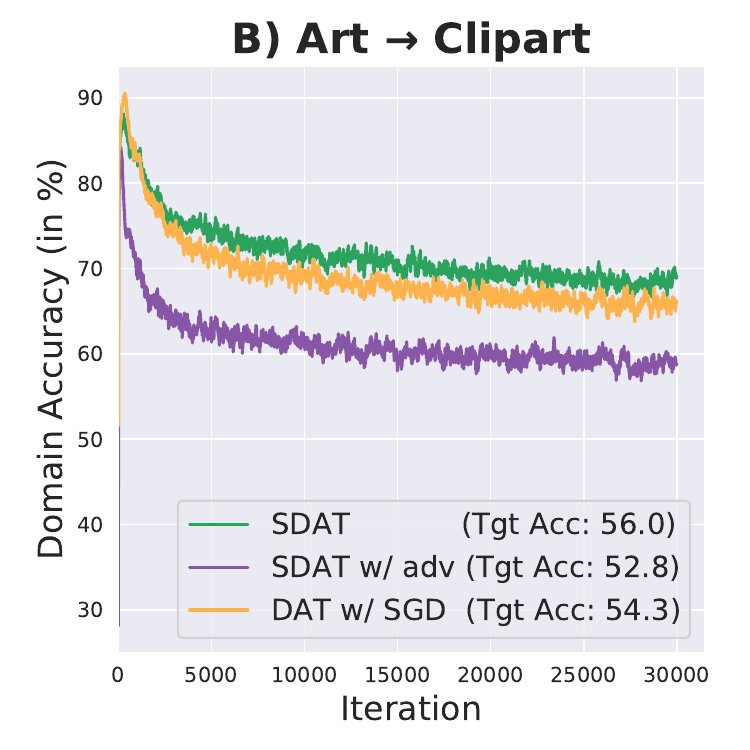} & \includegraphics[width=0.3\linewidth]{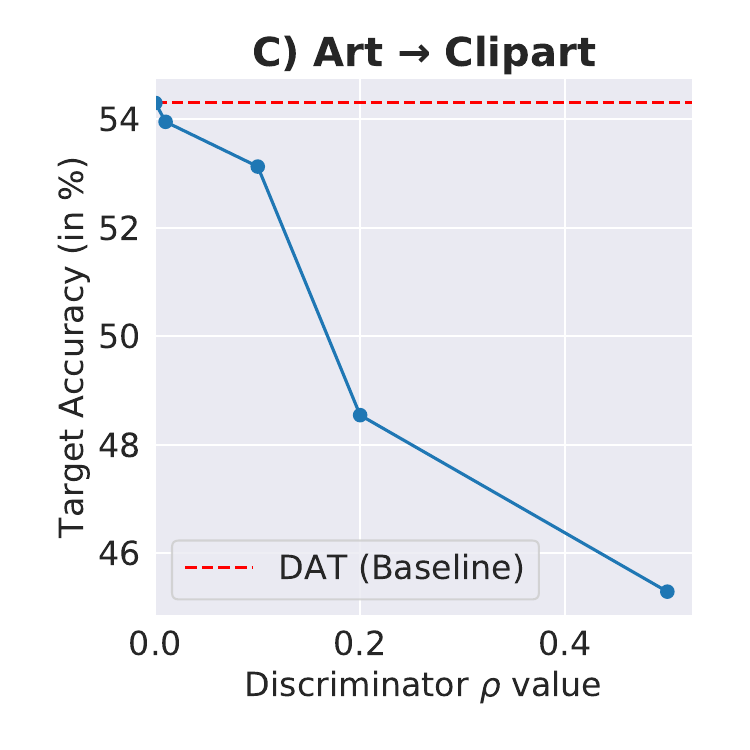} \\
\end{tabular}
\caption{\textbf{A)} Error on Target Domain (y-axis) for Office-Home dataset against maximum eigenvalue $\lambda_{max}$ of classification loss in DAT. 
When compared to SGD, Adam converges to a non-smooth minima (high $\lambda_{max}$), leading to a high error on target. 
Using Adam in comparison to SGD, converges to a non-smooth minima (high $\lambda_{max}$) leading to high error on target.
\textbf{B)} Domain Accuracy (vs iterations), it is lower when adversarial loss is smooth (i.e. SDAT w/ adv), which indicates suboptimal discrepancy estimation $d_{s,t}^{\Phi}$ \textbf{C)} Target Accuracy on Art $\rightarrow$ Clipart vs smoothness of the adversarial component. As the smoothness increases ($\rho$), the target accuracy decreases indicating that smoothing adversarial loss leads to sub-optimal generalization.}
\label{sdat_fig:2}
\end{figure*}

For empirical verification of our conjecture, we obtain the Eigen Spectral Density plot for the Hessian $\hat{R}^l_{T}(h_{\theta})$. %
We show the $\lambda_{max}$, $Tr(H)$ and Eigen Spectrum for different algorithms, namely DAT w/ Adam, DAT w/ SGD and our proposed SDAT (which is described in detail in later sections) in Fig.\ \ref{sdat_fig:hessian}. We find that \textit{high smoothness leads to better generalization on the target domain} (Additional empirical evidence in Fig.\ \ref{sdat_fig:2}A). We hypothesize that enforcing smoothness of classifier $h_{\theta}$ leads to a smooth landscape for discrepancy ($d_{S,T}^{\Phi}$)  as it is also a function of $h_{\theta}$. The smooth landscape ensures stable minimization (of Eq. \ref{sdat_eq:uda_obj}), ensuring a decrease in ($d_{S,T}^{\Phi}$)  with each SGD step even for a large step size (similar to \citep{chu2020smoothness}), this explains the enhanced stability and improved performance of adversarial training. For verifying the stabilization effect of smoothness, empirically we obtain $\lambda_{\max}$ for SGD and proposed SDAT at both best ($\lambda_{\max}^{best}$) and last epoch ($\lambda_{\max}^{last}$) for adaptation from Infographic to Clipart Domain (Fig.\ \ref{sdat_fig:hessian}\textcolor{blue}{D}). We find that as $\lambda_{\max}$ increases (decrease in smoothness of landscape), the training becomes unstable for SGD leading to a drop in validation accuracy. Whereas in the case of the proposed SDAT, the $\lambda_{\max}$ remains low across epochs, leading to stable and better validation accuracy curve. We also provide additional validation accuracy curves for more adaptation tasks where we also observe a similar phenomenon in Fig. \ref{sdat_fig:plotss}. To the best of our knowledge, our analysis of the effect of smoothness of task loss on the stability of DAT is novel. \\
\\
We also find that SGD leads to low $\lambda_{\max}$ (high smoothness w.r.t. task loss) in comparison to Adam leading to better performance. This also explains the widespread usage of SGD for DAT algorithms \citep{ganin2015unsupervised, long2018conditional, saito2018adversarial}, instead of Adam. More details about Hessian analysis is provided in App. \ref{sdat_app:hess}.

\subsection{Smoothing Loss Landscape}
\label{sdat_sec:subopt_dis}
In this section we first introduce the losses which are based on Sharpness Aware Minimization \citep{foret2021sharpnessaware} (SAM). The basic idea of SAM is to find a smoother minima (i.e. low loss in $\epsilon$ neighborhood of $\theta$) by using the following objective given formally below:
\begin{equation}
    \underset{\theta}{\min} \;\underset{||\epsilon|| \leq \rho}{\max}\; L_{obj}(\theta + \epsilon) 
\end{equation}
here $L_{obj}$ is the objective function to be minimized and $\rho\geq0$ is a hyperparameter which defines the maximum norm for $\epsilon$. Since finding the exact solution of inner maximization is hard, SAM maximizes the first order approximation:
\begin{equation}
\begin{split}
            \hat{\epsilon}(\theta) \approx \underset{||\epsilon|| \leq \rho}{\arg \max} \;  L_{obj}(\theta) + \epsilon^{\mathbf{T}}\nabla_{\theta} L_{obj}(\theta) \\
            = \rho \nabla_{\theta} L_{obj}(\theta) / ||\nabla_{\theta} L_{obj}(\theta)||_2.
\end{split}
\end{equation}

The $\hat{\epsilon}(\theta)$ is added to the weights $\theta$. The gradient update for $\theta$ is then computed as $\nabla_{\theta} L_{obj}(\theta)|_{\theta + \hat{\epsilon}(\theta)}$. The above procedure can be seen as a generic smoothness enhancing formulation for any $L_{obj}$. We now analogously introduce the sharpness aware source risk for finding a smooth minima:
\begin{equation}
      \underset{||\epsilon|| \leq \rho}{\max} \; {R}_{S}^l(h_{\theta + \epsilon}) =  
      \underset{||\epsilon|| \leq \rho}{\max} \mathbb{E}_{x \sim P_S}[\; l(h_{\theta + \epsilon}(x), f(x))].
\end{equation}
We also now define the sharpness aware discrepancy estimation objective below:
\begin{equation}
    \max_{\Phi} \min_{||\epsilon|| \leq \rho} d_{S,T}^{\Phi + \epsilon}.
    \label{sdat_eq:smooth_disc}
\end{equation}
As $d_{S,T}^{\Phi}$ is to be maximized the sharpness aware objective will have $\underset{||\epsilon|| \leq \rho}{\min}$ instead of $\underset{||\epsilon|| \leq \rho}{\max}$, as it needs to find smoother maxima. We now theoretically analyze the difference in discrepancy estimation for smooth version $d_{S,T}^{\Phi''}$ (Eq. \ref{sdat_eq:smooth_disc}) in comparison to non-smooth version $d_{S,T}^{\Phi'}$ (Eq. \ref{sdat_eq:diver_disc}). Assuming $\mathcal{D}_{\Phi}$ is a $L$-smooth function {(common assumption for non-convex optimization~\citep{carmon2020lower})}, 
$\eta$ is a small constant and $d_{S,T}^*$ the optimal discrepancy, the theorem states:

\begin{theorem}
\label{sdat_th:suboptimality}
For a given classifier $h_{\theta}$ and one step of (steepest) gradient ascent i.e. $\Phi' = \Phi + \eta (\nabla d_{S,T}^{\Phi}/||\nabla d_{S,T}^{\Phi}||)$ and $\Phi'' = \Phi + \eta (\nabla d_{S,T}^{\Phi}|_{\Phi + \hat{\epsilon}(\Phi)}/||\nabla d_{S,T}^{\Phi}|_{\Phi + \hat{\epsilon}(\Phi)}||)$ 
\begin{equation}
\begin{split}
         d_{S,T}^{\Phi'} - d_{S,T}^{\Phi''} \leq  \eta(1 - \cos \alpha)\sqrt{2L(d^*_{S,T} - d^{\Phi}_{S,T}) }
\end{split}
\end{equation}
where $\alpha$ is the angle between $\nabla d_{S,T}^{\Phi}$ and $\nabla d_{S,T}^{\Phi}|_{\Phi + \hat{\epsilon}(\Phi)}$. 
\end{theorem}

The $d_{S,T}^{\Phi'}$ (non-smooth version) can exceed $d_{S,T}^{\Phi''}$ (smooth discrepancy) significantly, as the term $d_{S,T}^* -  d_{S,T}^{\Phi}  \not\to 0$, as the $h_{\theta}$ objective is to oppose the convergence of $d_{S,T}^{\Phi}$ to optima $d_{S,T}^*$ (min-max training in Eq. \ref{sdat_eq:uda_obj}). Thus $d_{S,T}^{\Phi'}$ can be a better estimate of discrepancy in comparison to $d_{S,T}^{\Phi''}$. A better estimate of $d_{s,t}^{\Phi}$ helps in effectively reducing the discrepancy between $P_S$ and $P_T$, hence leads to reduced $R_{T}^l(h_{\theta})$. This is also observed in practice that smoothing the discriminator's adversarial loss (SDAT w/ adv in Fig.\ \ref{sdat_fig:2}\textcolor{blue}{B}) leads to low domain classification accuracy (proxy measure for $d_{s,t}^{\Phi}$) in comparison to DAT. 
Due to ineffective discrepancy estimation, SDAT w/ adv results in sub-optimal generalization on target domain i.e. high target error $R_{T}^l(h_{\theta})$ (Fig.\ \ref{sdat_fig:2}\textcolor{blue}{B}). We also observe that further increasing the smoothness of the discriminator w.r.t. adversarial loss (increasing $\rho$) leads to lowering of performance on the target domain (Fig.\ \ref{sdat_fig:2}\textcolor{blue}{C}). A similar trend is observed in GANs (App.\ \ref{sdat_app:gan_exp}) which also has a similar min-max objective. The proof of the above theorem and additional experimental details is provided in App. \ref{sdat_app:proof}.

\subsection{Smooth Domain Adversarial Training (SDAT)}
We propose smooth domain adversarial training which only focuses on converging to smooth minima w.r.t. {task} loss (i.e. empirical source risk), whereas preserves the original discrepancy term. We define the optimization objective of proposed Smooth Domain Adversarial Training:

\begin{equation}\min_{\theta} \max_{\Phi} \underset{||\epsilon|| \leq \rho}{\max}  \mathbb{E}_{x \sim P_S}[ l(h_{\theta + \epsilon}(x), y(x))] + d_{S,T}^{\Phi}.
    \label{sdat_eq:uda_obj_smooth}
\end{equation}
The first term is the sharpness aware risk, and the second term is the discrepancy term which is not smooth in our procedure. The term $d_{S,T}^{\Phi}$ estimates $D_{h_{\theta}, H}^{\phi}(P_S||P_T)$ discrepancy. 
We now show that optimizing Eq.\ \ref{sdat_eq:uda_obj_smooth} reduces $R_T^l(h_\theta)$ through a generalization bound. {This bound establishes that our proposed SDAT procedure is also consistent (i.e. in case of infinite data the upper bound is tight) similar to the original DAT objective (Eq. \ref{sdat_eq:uda_obj})}.  

\begin{theorem}

Suppose l is the loss function, we denote $\lambda^* := R_S^l(h^*) + R_T^l(h^*)$ and let $h^*$ be the ideal joint hypothesis:
\begin{equation}
\begin{split}
         R_{T}^l(h_{\theta}) \leq \; \max_{||\epsilon|| \leq \rho}\hat{R}_S^l(h_{\theta + \epsilon}) + D_{h_{\theta}, H}^{\phi}(P_S||P_T)   + \\ \gamma(||\theta||_2^2/\rho^2) + \lambda^* .
\end{split}
\end{equation}
where $\gamma: \mathbb{R}^{+} \rightarrow \mathbb{R}^{+}$ is a strictly increasing function.
\end{theorem}

\begin{table*}[!t]
  \centering     
      \caption{Accuracy (\%) on {Office-Home} for unsupervised DA (with ResNet-50 and ViT backbone). CDAN+MCC w/ SDAT outperforms other SOTA DA techniques. CDAN w/ SDAT improves over CDAN by 1.1\% with ResNet-50 and 3.1\% with ViT backbone. }  
    \vskip 0.15in
  \resizebox{\textwidth}{!}{%
  \begin{tabular}{l|c|cccccccccccc|c}
    \hline
    \textbf{Method} && \textbf{Ar$\veryshortarrow$Cl} & \textbf{Ar$\veryshortarrow$Pr} & \textbf{Ar$\veryshortarrow$Rw} & \textbf{Cl$\veryshortarrow$Ar} & \textbf{Cl$\veryshortarrow$Pr} & \textbf{Cl$\veryshortarrow$Rw} & \textbf{Pr$\veryshortarrow$Ar} & \textbf{Pr$\veryshortarrow$Cl} & \textbf{Pr$\veryshortarrow$Rw} & \textbf{Rw$\veryshortarrow$Ar} & \textbf{Rw$\veryshortarrow$Cl} & \textbf{Rw$\veryshortarrow$Pr} & \textbf{Avg} \Tstrut
    \Bstrut\\\hline \hline

    ResNet-50 \citep{he2016deep} &\parbox[t]{2mm}{\multirow{10}{*}{\rotatebox[origin=c]{90}{ResNet-50}}}& 34.9 & 50.0 & 58.0 & 37.4 & 41.9 & 46.2 & 38.5 & 31.2 & 60.4 & 53.9 & 41.2 & 59.9 & 46.1 \\
    DANN \citep{ganin2016domain} && 45.6 & 59.3 & 70.1 & 47.0 & 58.5 & 60.9 & 46.1 & 43.7 & 68.5 & 63.2 & 51.8 & 76.8 & 57.6 \\
    CDAN* \citep{long2018conditional} && 49.0 & {69.3} & 74.5 & 54.4 & {66.0} & {68.4} & {55.6} & 48.3 & 75.9 & 68.4 & 55.4 & {80.5} & 63.8 \\
    MDD \citep{zhang2019bridging} && 54.9 & 73.7 & 77.8 & 60.0 & 71.4 & 71.8 & 61.2 & 53.6 & 78.1 & 72.5 & 60.2 & 82.3 & 68.1 \\
    f-DAL \citep{acuna2021f}  && {56.7} & {\underline{77.0}} & {81.1} & {63.1} & {72.2} & {75.9} & {\underline{64.5}} & {54.4} & {81.0} & {72.3} & {58.4} & {83.7} & {70.0} \\ 
    
    SRDC \citep{tang2020unsupervised}  && {52.3} & {76.3} & {81.0} & {\textbf{69.5}} & {\underline{76.2}} & {\textbf{78.0}} & {\textbf{68.7}} & {53.8} & {\underline{81.7}} & {\textbf{76.3}} & {57.1} & {85.0} & {\underline{71.3}}\Bstrut\\\cline{1-1}\cline{3-15}
    CDAN && 54.3 & {70.6} & 76.8 & 61.3 & 69.5 & 71.3 & {61.7} & 55.3 & {80.5} & 74.8 & 60.1 & {84.2} & 68.4\\

    \cellcolor{mygray}CDAN w/ SDAT && \cellcolor{mygray}56.0 & \cellcolor{mygray}{72.2} & \cellcolor{mygray}{78.6} & \cellcolor{mygray}{62.5} & \cellcolor{mygray}{73.2} & \cellcolor{mygray}{71.8} & \cellcolor{mygray}{62.1} & \cellcolor{mygray}{55.9} & \cellcolor{mygray}80.3 & \cellcolor{mygray}\underline{75.0} & \cellcolor{mygray}61.4 & \cellcolor{mygray}{84.5} & \cellcolor{mygray}69.5\Bstrut\\\cline{1-1}\cline{3-15}
    {CDAN + MCC} && \underline{57.0} & {76.0} & \underline{81.6} & {64.9} & {75.9} & {75.4} & {63.7} & \underline{56.1} & 81.2 & {74.2} & \underline{63.9} & \underline{85.4} & \underline{71.3} \\
    \cellcolor{mygray}{CDAN + MCC w/ SDAT} && \cellcolor{mygray}\textbf{58.2} & \cellcolor{mygray}\textbf{77.1} & \cellcolor{mygray}\textbf{82.2} & \cellcolor{mygray}\underline{66.3} & \cellcolor{mygray}\textbf{77.6} & \cellcolor{mygray}\underline{76.8} & \cellcolor{mygray}{63.3} & \cellcolor{mygray}\textbf{57.0} & \cellcolor{mygray}\textbf{82.2} & \cellcolor{mygray}{74.9} & \cellcolor{mygray}\textbf{64.7} & \cellcolor{mygray}\textbf{86.0} & \cellcolor{mygray}\textbf{72.2}\Bstrut\\
    \hline \hline
	TVT \cite{yang2021tvt}  &\parbox[t]{2mm}{\multirow{5}{*}{\rotatebox[origin=c]{90}{ViT}}}& \textbf{74.9} & \underline{86.8} & 89.5 & 82.8 & \textbf{87.9} & 88.3 & 79.8 & \textbf{71.9} & \underline{90.1} & 85.5 & \underline{74.6} & 90.6 & \underline{83.6}\Tstrut \\
    \cline{1-1}\cline{3-15} 

    {CDAN}& & 62.6 & 82.9 & 87.2  & 79.2 & 84.9  & 87.1 & 77.9 & 63.3 & 88.7 & 83.1 & 63.5 & 90.8 & 79.3 \\
     \cellcolor{mygray}{CDAN w/ SDAT} && \cellcolor{mygray}69.1 & \cellcolor{mygray}86.6 & \cellcolor{mygray}88.9 & \cellcolor{mygray}81.9 & \cellcolor{mygray}86.2 & \cellcolor{mygray}88.0 & \cellcolor{mygray}\underline{81.0} & \cellcolor{mygray}66.7 & \cellcolor{mygray}89.7 & \cellcolor{mygray}86.2 & \cellcolor{mygray}72.1 & \cellcolor{mygray}\underline{91.9} & \cellcolor{mygray}82.4\Bstrut\\\cline{1-1}\cline{3-15}
    {CDAN + MCC}& & 67.0 & 84.8 & \underline{90.2} & \underline{83.4} & \underline{87.3} & \underline{89.3} & 80.7 & 64.4 & 90.0 & \underline{86.6} & 70.4 & \underline{91.9} & 82.2 \\
    \cellcolor{mygray}{CDAN + MCC w/ SDAT} && \cellcolor{mygray}\underline{70.8} & \cellcolor{mygray}\textbf{87.0} & \cellcolor{mygray}\textbf{90.5} & \cellcolor{mygray}\textbf{85.2} & \cellcolor{mygray}\underline{87.3} & \cellcolor{mygray}\textbf{89.7} & \cellcolor{mygray}\textbf{84.1} & \cellcolor{mygray}\underline{70.7} & \cellcolor{mygray}\textbf{90.6} & \cellcolor{mygray}\textbf{88.3} & \cellcolor{mygray}\textbf{75.5} & \cellcolor{mygray}\textbf{92.1} &\cellcolor{mygray} \textbf{84.3}
    \Bstrut\\\hline \hline
  \end{tabular}%
      }

  \label{sdat_tab:officehome}
\end{table*}

The bound is similar to generalization bounds for domain adaptation \citep{ben2010theory, acuna2021f}. The main difference is the sharpness aware risk term $\max_{||\epsilon|| \leq \rho}\hat{R}^l_S(h_{\theta})$ in place of source risk $R^{l}_S(h_{\theta})$, and an additional term that depends on the norm of the weights $\gamma(||\theta||_2^2/\rho^2)$. 
The first is minimized by decreasing the empirical sharpness aware source risk by using SAM loss shown in Sec.\ \ref{sdat_smoothness}. The second term is reduced by decreasing the discrepancy between source and target domains. The third term, as it is a function of norm of weights $||\theta||_2^2$, can be reduced by using either L2 regularization or weight decay. Since we assume that the $\mathcal{H}$ hypothesis class we have is rich, the $\lambda^*$ term is small. \\ \\
Any DAT baseline can be modified to use SDAT objective just by using few lines of code. \emph{We observe that the proposed SDAT objective (Eq. \ref{sdat_eq:uda_obj_smooth}) leads to significantly lower generalization error compared to the original DA objective (Eq. \ref{sdat_eq:uda_obj}), which we empirically demonstrate in the following sections}.
\section{Adaptation for classification}

We evaluate our proposed method on three datasets: Office-Home, VisDA-2017, and DomainNet, as well as by combining SDAT with two DAT based DA techniques: CDAN and CDAN+MCC. We also show results with ViT backbone on Office-Home and VisDA-2017 dataset.
\subsection{Datasets}
\textbf{Office-Home} \citep{venkateswara2017Deep}: Office-Home consists of 15,500 images from 65 classes and 4 domains: Art (Ar), Clipart (Cl), Product (Pr) and Real World (Rw).

\noindent \textbf{DomainNet }\citep{peng2019moment}: DomainNet consists of 0.6 million images across 345 classes belonging to six domains. The domains are infograph (inf), clipart (clp), painting (pnt), sketch (skt), real and quickdraw.

\noindent \textbf{VisDA-2017 }\citep{visda2017}:  VisDA is a dataset that focuses on the transition from simulation to real world and contains approximately 280K images across 12 classes. %

\subsection{Domain Adaptation Methods}
\textbf{CDAN }\citep{long2018conditional}: Conditional Domain Adversarial network is a popular DA algorithm that improves the performance of the DANN algorithm. CDAN introduces the idea of multi-linear conditioning to align the source and target distributions better. CDAN in Table \ref{sdat_tab:officehome} and \ref{sdat_table:visda_uda} refers to our implementation of CDAN* \cite{long2018conditional} method. 

\noindent \textbf{CDAN + MCC }\citep{jin2020minimum}: The minimum class confusion (MCC) loss term is added as a regularizer to CDAN. MCC is a non-adversarial term that minimizes the pairwise class confusion on the target domain, hence we consider this as an additional minimization term which is added to empirical source risk. This method achieves close to SOTA accuracy among adversarial adaptation methods. 
\subsection{Implementation Details} \label{sdat_imple}
We implement our proposed method in the Transfer-Learning-Library \citep{dalib} toolkit developed in PyTorch \citep{NEURIPS2019_9015}. The difference between the performance reported in CDAN* and our implementation CDAN is due to the batch normalization layer in domain classifier, which enhances performance.

\begin{table*}[ht!]
	\centering
	\caption{Accuracy (\%) on VisDA-2017 for unsupervised DA (with ResNet-101 and ViT backbone). The \textbf{mean} column contains mean across all classes. SDAT particularly improves the accuracy in classes that have comparatively low CDAN performance. 
}
 \vskip 0.15in
	\addtolength{\tabcolsep}{-3pt}
	\label{sdat_table:visda_uda}
	\begin{adjustbox}{max width=\textwidth}

	\begin{tabular}{l|c|cccccccccccc|c}
        \hline
		\textbf{Method} && \textbf{plane} & \textbf{bcycl} & \textbf{bus} & \textbf{car} & \textbf{horse} & \textbf{knife} & \textbf{mcyle} & \textbf{persn} & \textbf{plant} & \textbf{sktb} & \textbf{train} & \textbf{truck} & \textbf{mean} \Tstrut\Bstrut\\
		\hline \hline
		ResNet \citep{he2016deep} &\parbox[t]{2mm}{\multirow{10}{*}{\rotatebox[origin=c]{90}{ResNet-101}}}   &  55.1 & 53.3 & 61.9 & 59.1 &  80.6 & 17.9 & 79.7 & 31.2 & 81.0 & 26.5 & 73.5 & 8.5 & 52.4\\
		DANN \citep{ganin2016domain}& & 81.9 & 77.7 & 82.8 & 44.3 & 81.2 & 29.5 & 65.1 & 28.6 & 51.9 & 54.6 & 82.8 & 7.8 & 57.4\\
		MCD \citep{saito2018maximum}&&   87.0 &  60.9 & \textbf{83.7} & 64.0 & 88.9 & 79.6 & 84.7 & 76.9 & 88.6 & 40.3 & 83.0 & 25.8 & 71.9\\
		CDAN* \citep{long2018conditional}& &   85.2 & 66.9 & \underline{83.0} & 50.8 & 84.2 & 74.9 & 88.1 & 74.5 & 83.4 & 76.0 & 81.9 & 38.0 & 73.9\\
		MCC \citep{jin2020minimum} &&   88.1 & 80.3 & 80.5 & \textbf{71.5} & 90.1 & 93.2 & 85.0 & 71.6 & 89.4 & 73.8 & 85.0 & 36.9 & 78.8\\
		\cline{1-1}\cline{3-15}
		{CDAN} &&   94.9 & 72.0 & \underline{83.0} & 57.3 & 91.6 & 95.2 & \underline{91.6} & 79.5 & 85.8 & 88.8 & \underline{87.0} & 40.5 & 80.6\\
		\cellcolor{mygray}{CDAN w/ SDAT} & &  \cellcolor{mygray}94.8 & \cellcolor{mygray}77.1 & \cellcolor{mygray}82.8 & \cellcolor{mygray}60.9 & \cellcolor{mygray}92.3 & \cellcolor{mygray}95.2 & \cellcolor{mygray}\textbf{91.7} & \cellcolor{mygray}\textbf{79.9} & \cellcolor{mygray}\underline{89.9} & \cellcolor{mygray}\underline{91.2} & \cellcolor{mygray}\textbf{88.5} & \cellcolor{mygray}41.2 & \cellcolor{mygray}82.1\\
		\cline{1-1}\cline{3-15}
		{CDAN+MCC} & &  \underline{95.0} & \underline{84.2} & 75.0 & 66.9 & \textbf{94.4} & \underline{97.1} & 90.5 & \underline{79.8} & 89.4 & 89.5 & 86.9 & \underline{54.4} & \underline{83.6}\\
		\cellcolor{mygray}{CDAN+MCC w/ SDAT} &&   \cellcolor{mygray}\textbf{95.8} & \cellcolor{mygray}\textbf{85.5} & \cellcolor{mygray}76.9 &\cellcolor{mygray}\underline{69.0} & \cellcolor{mygray}\underline{93.5} & \cellcolor{mygray}\textbf{97.4} & \cellcolor{mygray}88.5 & \cellcolor{mygray}78.2 & \cellcolor{mygray}\textbf{93.1} & \cellcolor{mygray}\textbf{91.6} & \cellcolor{mygray}86.3 & \cellcolor{mygray}\textbf{55.3} & \cellcolor{mygray}\textbf{84.3}\\ \hline\hline
		TVT \cite{yang2021tvt} &\parbox[t]{2mm}{\multirow{5}{*}{\rotatebox[origin=c]{90}{ViT}}} & 92.9 & 85.6 & 77.5 & 60.5 & 93.6 & 98.2 & 89.3 & 76.4 & 93.6 & 92.0 & \underline{91.7} & 55.7 & 83.9 \\
        \cline{1-1}\cline{3-15}
        \cline{1-1}\cline{3-15}
		{CDAN} & & 94.3 & 53.0 & 75.7 & 60.5 & 93.9 & 98.3 & \textbf{96.4} & 77.5 & 91.6 & 81.8 & 87.4 & 45.2 & 79.6\\
		\cellcolor{mygray}{CDAN w/ SDAT} & &  \cellcolor{mygray}96.3 & \cellcolor{mygray}80.7 & \cellcolor{mygray}74.5 & \cellcolor{mygray}65.4 & \cellcolor{mygray}95.8 & \cellcolor{mygray}\textbf{99.5} & \cellcolor{mygray}92.0 & \cellcolor{mygray}\underline{83.7} & \cellcolor{mygray}93.6 & \cellcolor{mygray}88.9 & \cellcolor{mygray}85.8 & \cellcolor{mygray}\underline{57.2} & \cellcolor{mygray}84.5\\
		\cline{1-1}\cline{3-15}
		{CDAN+MCC} &&   \underline{96.9} & \underline{89.8} & \underline{82.2} & \underline{74.0} & \underline{96.5} & \underline{98.5} & 95.0 & 81.5 & \underline{95.4}& \underline{92.5} & 91.4 & \textbf{58.5} & \underline{87.7}\\
		\cellcolor{mygray}{CDAN+MCC w/ SDAT} && \cellcolor{mygray}\textbf{98.4} & \cellcolor{mygray}\textbf{90.9} & \cellcolor{mygray}\textbf{85.4} & \cellcolor{mygray}\textbf{82.1} & \cellcolor{mygray}\textbf{98.5} & \cellcolor{mygray}97.6 & \cellcolor{mygray}\underline{96.3} & \cellcolor{mygray}\textbf{86.1} & \cellcolor{mygray}\textbf{96.2} & \cellcolor{mygray}\textbf{96.7} & \cellcolor{mygray}\textbf{92.9} & \cellcolor{mygray}56.8 & \cellcolor{mygray}\textbf{89.8} \\ 
		\hline
		
	\end{tabular}%
	\end{adjustbox}
\end{table*}

We use a ResNet-50 backbone for Office-Home experiments and a ResNet-101 backbone for VisDA-2017 and DomainNet experiments. Additionally, we also report the performance of ViT-B/16 \cite{dosovitskiy2020image} backbone on Office-Home and VisDA-2017 datasets. All the backbones are initialised with ImageNet pretrained weights. We use a learning rate of 0.01 with batch size 32 in all of our experiments with ResNet backbone. We tune $\rho$ value in SDAT for a particular dataset split and use the same value across domains. The $\rho$ value is set to 0.02 for the Office-Home experiments, 0.005 for the VisDA-2017 experiments and 0.05 for the DomainNet experiments. More details are present in Appendix (refer App. \ref{sdat_app:experimental_deets}).

\subsection{Results}

\textbf{Office-Home}: For the Office-Home dataset, we compare our method with other DA algorithms including DANN, SRDC, MDD and f-DAL in Table \ref{sdat_tab:officehome}.
We can see that the addition of SDAT improves the performance on both CDAN and CDAN+MCC across majority of source and target domain pairs. CDAN+MCC w/ SDAT achieves SOTA adversarial adaptation performance on the Office-Home dataset with ResNet-50 backbone. With ViT backbone, the increase in accuracy due to SDAT is more significant. This may be attributed to the observation that ViTs reach a sharp minima compared to ResNets \cite{chen2021vision}. CDAN + MCC w/ SDAT outperforms TVT \cite{yang2021tvt}, a recent ViT based DA method and achieves SOTA results on both Office-Home and VisDA datasets (Table \ref{sdat_table:visda_uda}). Compared to the proposed method, TVT is computationally more expensive to train, contains additional adaptation modules and utilizes a backbone that is pretrained on ImageNet-21k dataset (App. \ref{sdat_app:comp_tvt}). On the other hand, SDAT is conceptually simple and can be trained on a single 12 GB GPU with ViT (pretrained on ImageNet). 
With ViT backbone, SDAT particularly improves the performance of source-target pairs which have low accuracy on the target domain (Pr$\veryshortarrow$Cl, Rw$\veryshortarrow$Cl, Pr$\veryshortarrow$Ar, Ar$\veryshortarrow$Pr in Table \ref{sdat_tab:officehome}). 

\begin{table}[H]
\centering
\caption{Results on DomainNet with CDAN w/ SDAT. The number in the parenthesis refers to the increase in Acc. w.r.t. CDAN.}
\vskip 0.15in
\begin{adjustbox}{max width=\linewidth}
\begin{tabular}{c | c  c  c  c  c | c } 
 \hline
  \textbf{Target \textbf{($\rightarrow$)}} & \multirow{2}{*}{\textbf{clp}} & \multirow{2}{*}{\textbf{inf}} & \multirow{2}{*}{\textbf{pnt}} & \multirow{2}{*}{\textbf{real}} & \multirow{2}{*}{\textbf{skt}} & \multirow{2}{*}{\textbf{Avg}} \Tstrut\\  \textbf{Source ($\downarrow$)} &&&&&& \Bstrut\\
 \hline\hline
\multirow{2}{*}{\textbf{clp}} & - & 22.0  & 41.5 & 57.5 &  47.2 & 42.1 \Tstrut\\&& \textcolor{ForestGreen}{(+1.4)}& \textcolor{ForestGreen}{(+2.6)}&
 \textcolor{ForestGreen}{(+1.5)} & \textcolor{ForestGreen}{(+2.3)} & \textcolor{ForestGreen}{(+2.0)} \\ 

\multirow{2}{*}{\textbf{inf}} & 33.9 & - & 30.3 & 48.1 & 27.9 & 35.0 \Tstrut\\& \textcolor{ForestGreen}{(+2.3)}&&
  \textcolor{ForestGreen}{(+1.0)} & \textcolor{ForestGreen}{(+4.5)} & \textcolor{ForestGreen}{(1.5)} & \textcolor{ForestGreen}{(+2.3)} \\

\multirow{2}{*}{\textbf{pnt}} &  47.5 & 20.7 & - & 58.0 & 41.8 & 42.0 \Tstrut\\ & \textcolor{ForestGreen}{(+3.4)}& 
\textcolor{ForestGreen}{(+0.9)} 
 &&
 \textcolor{ForestGreen}{(+0.8)} &
\textcolor{ForestGreen}{(+1.8)} &
 \textcolor{ForestGreen}{(+1.7)} \\

\multirow{2}{*}{\textbf{real}} &  56.7 & 25.1 & 53.6 & - & 43.9 & 44.8 \Tstrut \\ & \textcolor{ForestGreen}{(+0.9)} &
 \textcolor{ForestGreen}{(+0.7)} &
 \textcolor{ForestGreen}{(+0.4)} &
 &
 \textcolor{ForestGreen}{(+1.6)} &
 \textcolor{ForestGreen}{(+1.0)} \\

\multirow{2}{*}{\textbf{skt}} &  58.7 & 21.8 & 48.1 & 57.1 & - & 46.4 \Tstrut \\ &  \textcolor{ForestGreen}{(+2.7)} & 
 \textcolor{ForestGreen}{(+1.1)} &
 \textcolor{ForestGreen}{(+2.8)} &
 \textcolor{ForestGreen}{(+2.2)} &
&
 \textcolor{ForestGreen}{(+2.2)} 
\\\hline

\multirow{2}{*}{\textbf{Avg}} &  49.2 & 22.4 & 43.4 & 55.2 & 40.2 & 42.1 \Tstrut \\ & \textcolor{ForestGreen}{(+2.3)} &
 \textcolor{ForestGreen}{(+1.0)} &
 \textcolor{ForestGreen}{(+1.7)} &  \textcolor{ForestGreen}{(+2.2)} &
 \textcolor{ForestGreen}{(+1.8)} &
 \textcolor{ForestGreen}{(+1.8)}

\end{tabular}
\label{sdat_table:domainnet}
 \end{adjustbox}
\end{table}

\noindent \textbf{DomainNet}: Table \ref{sdat_table:domainnet} shows the results on the large and challenging DomainNet dataset across five domains. The proposed method improves the performance of CDAN significantly across all source-target pairs. On specific source-target pairs like inf $\veryshortarrow$ real, the performance increase is 4.5\%. The overall performance of CDAN is improved by nearly 1.8\% which is significant considering the large number of classes and images present in DomainNet. The improved results are attributed to stabilized adversarial training through proposed SDAT which can be clearly seen in Fig. \ref{sdat_fig:hessian}\textcolor{blue}{D}.

\noindent\textbf{VisDA-2017}: CDAN w/ SDAT improves the overall performance of CDAN by more than 1.5\% with ResNet backbone and by 4.9\% with ViT backbone on VisDA-2017 dataset (Table \ref{sdat_table:visda_uda}). Also, on CDAN + MCC baseline SDAT leads to 2.1\% improvement over baseline, leading to SOTA accuracy of 89.8\% across classes.  Particularly, SDAT significantly improves the performance of underperforming minority classes like bicycle, car and truck. 
Additional baselines and results are reported in Appendix (App. \ref{sdat_app:add_results}) {along with a discussion on statistical significance (App. \ref{sdat_app:stats_sig})}.

\section{Adaptation for object detection}
To further validate our approach's generality and extensibility, we did experiments on DA for object detection. We use the same setting as proposed in DA-Faster \citep{chen2018domaindafaster} with all domain adaptation components and use it as our baseline. We use the mean Average Precision at 0.5 IoU (mAP) as our evaluation metric. In object detection, the smoothness enhancement can be achieved in two ways (empirical comparison in Sec.~\ref{sdat_sec:od_results}) :

\begin{enumerate}
    \item \textbf{ DA-Faster w/ SDAT-Classification:} Smoothness enhancement for classification loss.
    \item \textbf{ DA-Faster w/ SDAT:} Smoothness enhancement for the combined classification and regression loss.
\end{enumerate}

\subsection{Experimental Setup}
We evaluate our proposed approach on object detection on two different domain shifts:

\noindent\textbf{Pascal to Clipart} ($P \rightarrow C$): 
Pascal \citep{everingham2010pascal} is a real-world image dataset which consists images with 20 different object categories. Clipart \citep{inoue2018crossclipart} is a graphical image dataset with complex backgrounds and has the same 20 categories as Pascal. We use ResNet-101 \citep{he2016deep} backbone  for {Faster R-CNN} \citep{ren2015faster} following \citet{saito2019strong}. 

\noindent \textbf{Cityscapes to Foggy Cityscapes} ($C \rightarrow Fc$): Cityscapes \citep{cordts2016cityscapes} is a street scene dataset for driving, whose images are collected in clear weather. Foggy Cityscapes \citep{sakaridis2018semanticfoggycityscapes} dataset is synthesized from Cityscapes for the foggy weather. We use ResNet-50 \citep{he2016deep} as the backbone for  {Faster R-CNN} for experiments on this task. Both domains have the same 8 object categories with instance labels.

\begin{figure*}[!t]
    \centering
    \begin{subfigure}[b]{0.3\linewidth}
  \centering
  \includegraphics[width=\textwidth, height = 4.1cm]{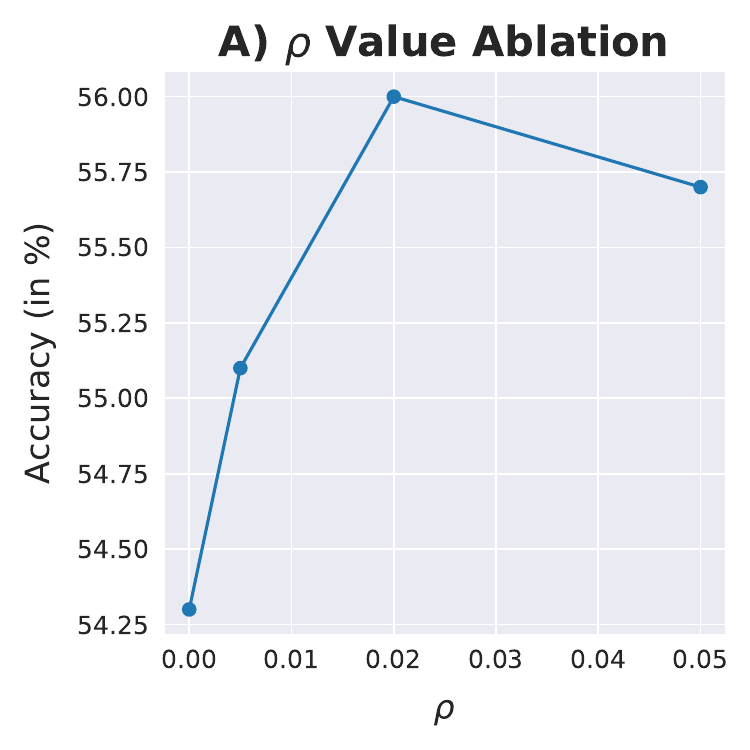}
  \label{sdat_fig:rho_ablation}
\end{subfigure}
\begin{subfigure}[b]{0.3\linewidth}
  \centering
  \includegraphics[width=\textwidth, height=4.1cm ]{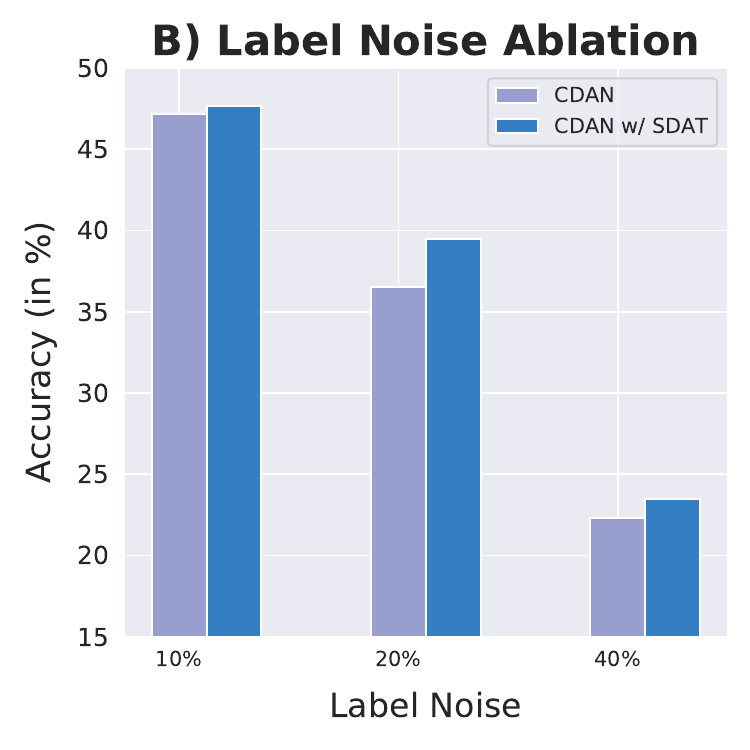}
  \label{sdat_fig:labelnoise}
\end{subfigure}
\begin{subfigure}[b]{0.3\linewidth}
  \centering
  \subfloat{
  \begin{adjustbox}{max width=\linewidth}
  \begin{tabular}{ccc}
 \multicolumn{3}{c}{\large \textbf{C) Smoothing Loss Components}}\\
  \hline
  \textbf{Smooth Task}&\textbf{Smooth Adv} &\textbf{Accuracy}\Tstrut\\ \hline \hline
  \textcolor{red}{\xmark} &\textcolor{red}{\xmark} & 54.3\\\hline
 \textcolor{red}{\xmark}&\textcolor{green}{\cmark}&51.0\\\hline
  \textcolor{green}{\cmark}&\textcolor{red}{\xmark}&\textbf{55.7}\\\hline  
  \textcolor{green}{\cmark}&\textcolor{green}{\cmark}&54.9\\\hline 
  \end{tabular} \end{adjustbox}}
  \vspace{5em}
\end{subfigure}%
\caption{{ Analysis of SDAT for Ar $\rightarrow$ Cl split of Office-Home dataset. \textbf{A)} Variation of target accuracy with maximum perturbation $\rho$. \textbf{B)} Comparison of accuracy of SDAT with DAT for different ratio of label noise. \textbf{C)} Comparison of accuracy when smoothing is applied to various loss components.  }}
\label{sdat_fig:smooth}
\end{figure*}

\subsection{Results}
\label{sdat_sec:od_results}
Table~\ref{sdat_tab:object_detection} shows the results on two domain shifts with varying batch size (\textit{bs}) during training. We find that only smoothing w.r.t. classification loss is much more effective (SDAT-Classification) than smoothing w.r.t. combined classification and regression loss (SDAT). On average, SDAT-Classification produces an mAP gain of {2.0}\% compared to SDAT, and {2.8}\% compared to DA-Faster baseline. 

\begin{table}[!ht]
    \caption{Results on DA for object detection. }
    \vskip 0.15in
    \centering
    \begin{adjustbox}{max width=\columnwidth}
    \begin{tabular}{l|ccc}
    \hline
    \multirow{2}{*}{Method} & $C \rightarrow Fc$  & $P \rightarrow C$ &  $P \rightarrow C$ \Tstrut\\ & \textit{ (bs=2)} & \textit{(bs=2)} &  \textit{ (bs=8)}\\
    \hline \hline
        DA-Faster \citep{chen2018domaindafaster} & 35.21 & 29.96 &  26.40 \\ 
        DA-Faster w/ SDAT & 37.47 & 29.04 &  27.64 \\ 
        DA-Faster w/ SDAT-Classification & \textbf{38.00} & \textbf{31.23} &  \textbf{30.74}
    \end{tabular}
    \end{adjustbox}
    \label{sdat_tab:object_detection}
    
 \end{table}

The proposed SDAT-Classification significantly outperforms DA-Faster baseline  and improves mAP by {1.3}\% on $P \rightarrow C$ and by {2.8}\% on $C \rightarrow Fc$. It is noteworthy that increase in performance of SDAT-Classification is consistent even after training with higher batch size $(bs=8)$ achieving improvement of {4.3}\% in mAP. Table \ref{sdat_tab:object_detection} also shows that even DA-Faster w/ SDAT (i.e. smoothing both classification and regression) outperforms DA-Faster by {0.9} \% on average. The improvement due to SDAT on adaptation for object detection shows the generality of SDAT across techniques that have some form of adversarial component present in the loss formulation.

\section{Discussion}
\label{sdat_sec:discussion}
\textbf{How much smoothing is optimal?}: Figure \ref{sdat_fig:smooth}\textcolor{blue}{A} shows the ablation on $\rho$ value (higher $\rho$ value corresponds to more smoothing) on the Ar$\veryshortarrow$Cl from Office-Home dataset with CDAN backbone. The performance of the different values of $\rho$ is higher than the baseline with $\rho$ = 0. It can be seen that $\rho$ = 0.02 works best among all the different values and outperforms the baseline by at least 1.5\%. We found that the same $\rho$ value usually worked well across domains in a dataset, but different $\rho$ was optimal for different datasets. More details about optimum $\rho$ is in App. \ref{sdat_app:opt_rho}.

\noindent \textbf{Which components benefit from smooth optima?}:
Fig. \ref{sdat_fig:smooth}C shows the effect of introducing smoothness enhancement for different components in DAT.
For this we use SAM on a) task loss (SDAT) b) adversarial loss (SDAT w/ adv) c) both task and adversarial loss (SDAT-all). It can be seen that smoothing the adversarial loss component (SDAT w/ adv) reduces the performance to 51.0\%, which is significantly lower than even the DAT baseline.

\noindent\textbf{Is it Robust to Label Noise?}: In practical, real-world scenarios, the labeled datasets are often corrupted with some amount of label noise. Due to this, performing domain adaptation with such data is challenging. We find that smoother minima through SDAT lead to robust models which generalize well on the target domain. Figure \ref{sdat_fig:smooth}\textcolor{blue}{B} provides the comparison of SGD vs. SDAT for different percentages of label noise injected into training data.

\begin{table}
\centering
\caption{Performance comparison across different loss smoothing techniques on Office-Home. SDAT (with ResNet-50 backbone) outperforms other smoothing techniques in each case consistently.}
\vskip 0.15in
    \begin{adjustbox}{max width=\linewidth}
    \begin{tabular}{l|cccc|cc}
    \hline
    {Method} & Ar$\veryshortarrow$Cl & Cl$\veryshortarrow$Pr & Rw$\veryshortarrow$Cl &  Pr$\veryshortarrow$Cl & \multicolumn{2}{c}{{Avg}}\\
    \hline \hline
       {DAT} & 54.3 & 69.5 & 60.1 & 55.3 &  59.2 &\\
         
         {VAT} & 54.6 &70.7&  60.8 & 54.4 & 60.1 &  \textcolor{ForestGreen}{(+0.9)}\\
       {SWAD} & 54.6& 71.0 & 60.9 & 55.2 & 60.4 & \textcolor{ForestGreen}{(+1.2)} \\
        {LS}  & 53.6  & 71.6 & 59.9 &  53.4 & 59.6 & \textcolor{ForestGreen}{(+0.4)}  \\
        {SAM}  & 54.9  & 70.9 & 59.2 &  53.9 & 59.7 & \textcolor{ForestGreen}{(+0.5)}  \\
        {SDAT} &\textbf{56.0}&  \textbf{73.2} & \textbf{61.4} &\textbf{55.9} & \textbf{61.6} &\textcolor{ForestGreen}{(+2.4)}
  
    \end{tabular}
    \end{adjustbox}
    
    \label{sdat_tab:diff_smooth}
\end{table}

\noindent \textbf{Is it better than other smoothing techniques?}
To answer this question, we compare SDAT with different smoothing techniques originally proposed for ERM. We specifically compare our method against DAT, Label Smoothing (LS) \citep{szegedy2016rethinking}, SAM \cite{foret2021sharpnessaware} and VAT \citep{miyato2019vat}. \citet{stutz2021relating} recently showed that these techniques produce a  significantly smooth loss landscape in comparison to SGD. We also compare with a very recent SWAD \citep{cha2021swad} technique which is shown effective for domain generalization. For this, we run our experiments on four different splits of the Office-Home dataset and summarize our results in Table \ref{sdat_tab:diff_smooth}. We find that techniques for ERM (LS, SAM and VAT) fail to provide significant consistent gain in performance which also confirms the requirement of specific smoothing strategies for DAT. We find that SDAT even outperforms SWAD on average by a significant margin of 1.2\%. Additional details regarding the specific methods are provided in App. \ref{sdat_app:smooth_tech}.

\begin{table}[t!]
    \centering

    \setlength{\tabcolsep}{4pt}
 \caption{Analysing the effect of SDAT on DANN (on DomainNet and Office-Home with ResNet-50 and ViT-B/16 respectively) and GVB-GD (on Office-Home with ResNet-50).}
 \vskip 0.15in
    \begin{adjustbox}{max width=\columnwidth}
    \begin{tabular}{l|c|cccc}
    \hline
    {\textbf{DomainNet}} && \textbf{clp$\veryshortarrow$pnt} &  \textbf{skt$\veryshortarrow$pnt} & \textbf{inf$\veryshortarrow$real} & \textbf{skt$\veryshortarrow$clp} \\
    \hline \hline
    \textbf{DANN}&\parbox[t]{2mm}{\multirow{2}{*}{\rotatebox[origin=c]{90}{RN-50}}} & 37.5 & 43.9 & 37.7 & 53.8 \Bstrut\Tstrut{}\\
    
     \textbf{DANN w/ SDAT}&& 38.9 \textcolor{ForestGreen}{(+1.4)} & 45.7 \textcolor{ForestGreen}{(+1.8)} & 39.6 \textcolor{ForestGreen}{(+1.9)} & 56.3 \textcolor{ForestGreen}{(+2.5)} \Bstrut\Tstrut\\
     
     \hline\hline
    {\textbf{Office-Home}} && \textbf{Ar$\veryshortarrow$Cl} &  \textbf{Cl$\veryshortarrow$Pr} & \textbf{Rw$\veryshortarrow$Cl} & \textbf{Pr$\veryshortarrow$Cl} \Tstrut\\
    \hline 
        \textbf{GVB-GD}&\parbox[t]{2mm}{\multirow{4}{*}{\rotatebox[origin=c]{90}{RN-50}}} &56.4 & 74.2& 59.0& 55.9\Bstrut\Tstrut\\
    
     \textbf{GVB-GD w/ SDAT}&& 57.6 \textcolor{ForestGreen}{(+1.2)} & 75.4 \textcolor{ForestGreen}{(+1.2)}  & 60.0 \textcolor{ForestGreen}{(+1.0)}  &  56.6 \textcolor{ForestGreen}{(+0.7)}   \Bstrut\\
     
     \cline{1-1}
     \cline{3-6}

    \textbf{DANN}&\parbox[t]{2mm}{} & 52.6 & 65.4 & 60.4 & 52.3 \Bstrut\Tstrut\\
    
     \textbf{DANN w/ SDAT}&& 53.4 \textcolor{ForestGreen}{(+0.8)}  & 66.4 \textcolor{ForestGreen}{(+1.0)} & 61.3 \textcolor{ForestGreen}{(+0.9)} &  53.8 \textcolor{ForestGreen}{(+1.5)} \Bstrut\\
     
     \hline

    \textbf{DANN}&\parbox[t]{2mm}{\multirow{2}{*}{\rotatebox[origin=c]{90}{ViT}}} & 62.7&81.8&68.5&66.5 \Bstrut\Tstrut\\
    
     \textbf{DANN w/ SDAT}&& 68.0 \textcolor{ForestGreen}{(+5.3)}  & 82.4 \textcolor{ForestGreen}{(+0.6)} & 73.4 \textcolor{ForestGreen}{(+4.9)} &  68.8 \textcolor{ForestGreen}{(+2.3)} \Bstrut\\
     
     \hline\hline

    \end{tabular}
    
    \label{sdat_tab:dann_domainnet}
\end{adjustbox}
\end{table}
\noindent\textbf{Does it generalize well to other DA methods?}: We show results highlighting the effect of smoothness (SDAT) on DANN\cite{ganin2016domain} and GVB-GD \cite{cui2020gvb} in Table \ref{sdat_tab:dann_domainnet} with ResNet-50 and ViT backbone. DANN w/ SDAT leads to gain in accuracy on both DomainNet and Office-Home dataset. We observe a significant increase (average of +3.3\%) with DANN w/ SDAT (ViT backbone) on Office-Home dataset. SDAT leads to a decent gain in accuracy on Office-Home dataset with GVB-GD despite the fact that GVB-GD is a much stronger baseline than DANN. We primarily focused on CDAN and CDAN + MCC for the main results as we wanted to establish that SDAT can improve on even SOTA DAT methods for showing its effectiveness. Overall, we have shown results with four DA methods (CDAN, CDAN+MCC, DANN, GVB-GD) and this shows that SDAT is a generic method that can applied on top of any domain adversarial training based method to get better performance.

\section{Conclusion}
In this work, we analyze the curvature of loss surface of DAT used extensively for Unsupervised DA. We find that converging to a smooth minima w.r.t. {task} loss (i.e., empirical source risk) leads to stable DAT which results in better generalization on the target domain. We also theoretically and empirically show that smoothing adversarial components of loss lead to sub-optimal results, hence should be avoided in practice. We then introduce our practical and effective method, SDAT, which only increases the smoothness w.r.t. {task} loss, leading to better generalization on the target domain. SDAT leads to an effective increase for the latest methods for adversarial DA, achieving SOTA performance on benchmark datasets. One limitation of SDAT is presence of no automatic way of selecting $\rho$ (extent of smoothness) which is a good future direction to explore.

\chapter{Conclusion}
\label{chap:conclusion}
This thesis explores the problem of learning neural networks from limited and imperfect data. In Section~\ref{chap:introduction} we have introduced the following challenges faced while training deep networks on limited and imperfect datasets: i) generative models suffer from missing modes or mode-collapse for long-tailed datasets, ii) recognition models overfit to small set of images in tail classes, iii) overemphasis on just validation accuracy leads to poor tail class performance in long-tailed recognition and iv) distribution shift between training domain and testing domain at time of deployment. Towards overcoming these challenges we have introduced parts that aim towards i) training generative models for diverse and class-consistent generation even for tail classes, ii) inductive regularization schemes to ensure tail classes also behave like head classes avoiding overfitting, iii) optimizing practical metrics like worst-case recall, etc. for robust long-tail learning and iv) efficiently adapting to target domain via minimal supervision. Each part of the thesis is comprised of chapters that develop techniques to address the goal of each part. The conclusions from each chapter are summarised in Table \ref{tab:conclusion_table}. We provided a high-level summary and conclusions for each part of the thesis in the section below. We conclude by providing some future directions and avenues that can be worked on in the future.

\hrnote{1. Challenges 
2. Goals 
3. Chapters.
4. Parts.

}

\section{Summary of the Thesis}

 Each part of the thesis covers scenarios starting from training GANs on long-tailed data, followed by learning downstream classification models with inductive regularization without generated data. In the third part, we look at the scenario of semi-supervised long-tail learning, followed by learning models from related domains and efficiently adapting to the target domain. We present the high-level summary of each part below:

\begin{itemize}
    \item 
    
    In the first part of the thesis, we look at the problem of training Generative Adversarial Networks (GANs) on long-tailed datasets. We find that unconditional GANs and some conditional GANs (cGANs) start generating arbitrary distribution. On the other hand, most conditional GANs lead to mode-collapse and produce poor-quality generations. To improve quality and balance them across classes, we demonstrate that pre-trained classifiers for long-tailed data can provide balancing feedback to the GAN training through the proposed class balancing regularizer. Following this, we take a closer look at the mode-collapse in conditional GAN to remove the overhead of the additional classifier in the previous approach. We observe that mode-collapse is closely correlated to the occurrence of spectral explosion for class-specific conditioning parameters. Hence, we propose an inexpensive group Spectral Regularizer (gSR), which ensures that class-specific conditioning batch norm parameters follow the same distribution. This leads to mitigation of collapse and diverse images for cGANs like BigGAN and SNGAN. However, with a group Spectral Regularizer, a side effect is that the model confuses similar classes with each other, as the conditioning parameters have the same distribution for large datasets. To mitigate the problem of confusion and collapse simultaneously, in our next work, NoisyTwins, we 
    characterize the conditioning parameters space as Gaussians with a learnable mean. We induce a high variance for tail classes to mitigate collapse. Inspired by self-supervised contrastive learning, we introduce a NoisyTwins regularizer to promote inter-class distance while ensuring that the intra-class is clustered diversly for class consistency. We can scale this method to StyleGAN, producing 
    SotA results for image generation on ImageNet-LT and iNaturalist 2019 datasets. 
    
    \item In the second part of the thesis, we focus on using inductive regularization to improve long-tail recognition tasks. We want to induce the properties of the head classes to the tail classes without explicitly augmenting the dataset with generated samples. In Chapter~\ref{chap:SaddleSAM}, we observe through class-wise hessian analysis that the model trained with re-weighting converges to a minimum for head classes, whereas it gets stuck in a saddle point for tail classes. We demonstrate that escaping from the saddle points improves tail class generalization. Further, we show that Sharpness Aware Minimization (SAM) can be used to escape from Saddle Points efficiently compared to other methods, significantly improving the performance on tail classes for SotA re-weighting methods. Further, we find that the disparity between tail and head classes in performance widens for Vision Transformer (ViT) as ViT learns all its inductive biases from data, which is scarce for tail classes. On the other hand, CNNs have inductive biases like the locality of features, making them robust to the imbalance in the dataset. To induce the robustness of ViT,in Chapter~\ref{chap:DeiT-LT} we introduce an efficient distillation scheme (Deit-LT) to distill from CNN using out-of-distribution data. To improve the generalization further, we introduce distillation from the flat CNNs trained from SAM, which ensures the induction of low-rank features. We find that learning inductive biases via distillation from CNN for tail classes and from ground truth for head classes leads to the best of both worlds, where it can leverage its scalability while maintaining robustness to imbalances.

    \item In the third part of the thesis, we explore the more practical setting where we assume access to long-tailed data. However, only a fraction of it could be labeled practically due to annotation costs (\ie semi-supervised). In this setup, self-training methods like FixMatch obtain decent accuracy but perform much worse on metrics like worst-case recall, etc. Such metrics are non-decomposable, which we aim to optimize in a semi-supervised setting. In Chapter~\ref{chap:csst}, we generalize self-training to cost-sensitive self-training (CSST), where the cost is different for each class of misclassification. We can formulate several non-decomposable metric optimizations as equivalent cost-sensitive problems. We introduce a cost-sensitive consistency regularization loss in self-training to optimize the desired metric. We prove the optimality of the proposed regularizer to optimize the desired metric theoretically, and empirically demonstrate its effectiveness in practice using FixMatch for vision and the UDA method for NLP. We find that CSST effectively optimizes the desired metric while maintaining the accuracy of the baseline method. With CSST, a limitation is that the models need to be trained from scratch to optimize each desired objective; in SelMix (Chapter~\ref{chap:selmix}), we work in a setting where we are provided a pre-trained model for optimizing the preferred metric. The main idea is to selectively mix up tail and head classes to optimize the desired metric. For this, we approximate the gain in metric due to each mixup and then prefer mixups that maximally improve the metric performance. The SelMix distribution is obtained by model performance on a held-out set, which provides metric feedback. SelMix is able to optimize a much broader set of non-decomposable objectives (\ie, non-linear) in comparison to CSST while only being a fine-tuning technique. We demonstrate the effectiveness of SelMix by testing it under various settings where it consistently outperforms the baselines significantly. We believe that techniques like CSST and SelMix can help significantly in the future, as AI systems need to be scrutinized through various metrics to be deployed in critical applications.
    
    \item In the fourth part of the thesis, we shift our focus from the iid setting to learning the model from source domain labeled data and adapting it to the target domain (\ie domain adaptation). We look at the problem similar to a semi-supervised setup, where we carefully select a subset to label to improve the performance in the target domain (\ie active learning). We formulate a submodular criterion based on the selected subset's uncertainty, representativeness, and diversity. This enables us to select optimal subsets in linear time in terms of sample size. We introduce a high learning rate and full-finetuning DA procedure to use these samples for effective adaptation. The overall procedure S3VAADA enables effective learning on datasets of Office-Home, Office-31, VisDa-18, and even the large-scale DomainNet. As this procedure requires some data, we also explore improving the generalization of unsupervised domain adaptation. In this DA setting with task loss and adversarial loss, we demonstrate that converging to a smooth minimum for task loss is beneficial; however, smooth minima for adversarial loss degrade performance. Based on this, we introduce a Smooth Domain Adversarial Training (SDAT) formulation. The SDAT formulation significantly improves performance across various SotA domain adversarial methods, demonstrating the generality of SDAT. These efficient ways enable effective adaptation in the presence of distribution shifts.

\end{itemize}
\begin{table}[!ht]

    \caption{A table depicting problem setup, method and key contributions for each chapter}
    \label{tab:conclusion_table}
    \scalebox{0.7}{
    \renewcommand{\arraystretch}{1.08}{
        \begin{tabular}{ >{\columncolor{gray!10}} p{0.25\linewidth} | p{0.57\linewidth}| >{\columncolor{gray!10}} p{0.57\linewidth}}
            \toprule
            \multicolumn{1}{>{\columncolor{gray!10}}l|}{\textbf{Chapter}}
            &  & 
               \\
            \textbf{Problem Setup} & \multicolumn{1}{c|}{\multirow{-2}{*}{\textbf{Proposed Approach}}}& \multicolumn{1}{>{\columncolor{gray!10}}c}{\multirow{-2}{*}{\textbf{Key Contributions}}}\\
            
             \midrule
            
            \multicolumn{3}{c}{\textbf{Part I: Training Generative Adversarial Networks (GANs) on Long-Tailed Datasets}} \\
            \midrule 
           
            \begin{enumerate}[left=0.1em, start=2]
            \item \parbox[t]{\linewidth}{Class Balancing GAN with a Classifier in the Loop. \\ \\
    Train a GAN on a long-tailed dataset to generate images with a class-balanced data distribution.} 
            \end{enumerate}

            &
  \vspace{-0.3cm}
  \begin{itemize}[left=0.5em]
      \item Our main idea is first to estimate the class distribution $P(Y|X)$ of the generated samples using a classifier trained specifically for long-tailed data. 
      
      \item We propose a `class-balancing' regularizer that uses the estimated statistic $P(Y|X)$ of generated samples, to promote uniformity while sampling from an unconditional GAN. 
      
      \item  The proposed class-balancing regularizer minimizes a form of weighted entropy, and is added to generator loss while GAN training.
  
  \end{itemize} &
  \vspace{-0.3cm}
  \begin{itemize}[left=0.5em]
      \item We first introduce the idea of controlling the idea of generative models using classifier guidance, which has been extensively used for diffusion models.
      
      \item Easy to compute weighted entropy regularizer, which has a theoretical and empirical guarantee.
      
      \item Leads to improvement in FID and balanced generations, also scales to iNaturalist-2019 dataset.
      \item Code: \url{https://github.com/val-iisc/class-balancing-gan}
  \end{itemize} 

 \\ \hline

 \begin{enumerate}[left=0.1em, start=3]
            \item \parbox[t]{\linewidth}{Improving GANs for Long-Tailed Data through Group Spectral Regularization \\ \\ Train a conditional GAN (BigGAN, SNGAN) without mode collapse on long-tailed data with classifier feedback.}
            \end{enumerate} &
  \vspace{-0.3cm}
  \begin{itemize}[left=0.5em]
      \item We observe that the mode-collapse for each tail class is closely coupled with the explosion of spectral norm of class-specific batch norm parameters.

      \item We find that mitigating spectral norm explosion by adding an explicit spectral regularizer mitigates tail class mode collapse.
        
      \item For doing this inexpensively, we propose a group Spectral Regularizer (gSR) which estimates the spectral norm using single power iteration and adds it to the generator loss while training.
  
  \end{itemize} &
  \vspace{-0.3cm}
  \begin{itemize}[left=0.5em]
      \item We find that even existing techniques for training GANs on limited data~\cite{tseng2021regularizing, liu2021generative, Karras2020ada} cannot prevent tail class-specific mode collapse. 
      
      \item Combining gSR with existing SOTA GANs with regularization leads to large average relative improvement (of $\sim25\%$ in FID) for image generation on 5 different long-tailed dataset configurations.
      \item The proposed gSR regularizer helps train GANs at a resolution of 256$\times$256 on the LSUN dataset, which requires large GPU memory with the class balancing regularizer approach.
      
      \item Project Page (w/ Code): \url{https://sites.google.com/view/gsr-eccv22}
  \end{itemize} \\ \hline

  \begin{enumerate}[left=0.1em, start=4]
            \item \parbox[t]{\linewidth}{NoisyTwins: Class-Consistent and Diverse Image Generation through StyleGANs \\ \\ Train a conditional GAN (specifically StyleGAN) without mode collapse ensuring class consistency for large-scale datasets.} 
            \end{enumerate} &
  \vspace{-0.3cm}
  \begin{itemize}[left=0.5em]

    \item  Various recent SotA GAN conditioning and regularization techniques on the challenging task of long-tailed image generation. All existing methods either suffer from mode collapse or lead to class confusion in generations.%
    
    \item To mitigate mode collapse and class confusion, we introduce NoisyTwins, in which we first characterize the class embedding as a gaussian with learnable mean and fixed variance across all classes.

    \item We then sample two embedding vectors from the same class (\ie augmentations) and ensure that their representation in $\mathcal{W}$ latent space are clustered together, away from other class clusters inspired by self-supervised learning. This ensures class consistency while mitigating mode collapse.

  \end{itemize} &
  \vspace{-0.3cm}
  \begin{itemize}[left=0.5em]

    \item NoisyTwins is an inexpensive self-supervised technique to prevent mode-collapse while ensuring the class consistency of the generated images.

    \item We evaluate NoisyTwins on large-scale long-tailed datasets of ImageNet-LT and iNaturalist-2019, where it consistently improves the StyleGAN2 performance ($\sim 19\%$ FID), achieving a new SotA.

    \item NoisyTwins can also prevent collapse in the few-shot GAN setting, improving over SotA across AnimalFace and ImageNet Carnivore datasets.

      \item Project Page (w/Code): \url{https://rangwani-harsh.github.io/NoisyTwins/}
  \end{itemize} \\ \hline

         \end{tabular} 
 }   
}
\end{table}

\begin{table}[htp]

    \scalebox{0.7}{
    \renewcommand{\arraystretch}{1.08}{
        \begin{tabular}{ >{\columncolor{gray!10}} p{0.25\linewidth} | p{0.57\linewidth}| >{\columncolor{gray!10}} p{0.57\linewidth}}
            \toprule
            \multicolumn{1}{>{\columncolor{gray!10}}l|}{\textbf{Chapter}}
            &  & 
               \\
            \textbf{Problem Setup} & \multicolumn{1}{c|}{\multirow{-2}{*}{\textbf{Proposed Approach}}}& \multicolumn{1}{>{\columncolor{gray!10}}c}{\multirow{-2}{*}{\textbf{Key Contributions}}}\\
            
             \midrule
            
            \multicolumn{3}{c}{\textbf{Inductive Regularization for Improving Long-Tailed Recongnition}} \\
            \midrule 
           
            \begin{enumerate}[left=0.1em, start=5]
            \item \parbox[t]{\linewidth}{
            Escaping Saddle Points for Effective Generalization on Class-Imbalanced Data \\ \\

           Training CNNs on long-tailed data to achieve improved generalization on tail classes.
            
            }
            \end{enumerate} &
  \vspace{-0.3cm}
  \begin{itemize}[left=0.5em]
      \item We propose a class-wise Hessian analysis of loss landscape to study the curvature properties of the converged solution. 

      \item We find that all re-weighting methods, including popular ones like LDAM~\cite{cao2019learning} etc., converge to a saddle point for tail classes instead of minima. This hinders the generalization performance of the model on tail classes.

      \item We find that the optimization methods designed to escape from saddle points can be effectively be used to improve generalization performance on tail classes.

      \item We demonstrate that Sharpness Aware Minimization (SAM) ~\cite{foret2020sharpness} being used with re-weighting methods has a larger component in the direction of negative curvature, which helps it to escape saddle points and improve generalization significantly in comparison to other methods.

  \end{itemize} &
  \vspace{-0.3cm}
  \begin{itemize}[left=0.5em]
      \item We provide a key tool of class-wise Hessian to analyze the loss curvature for models trained on long-tailed datasets, leading to uncovered insights.

      \item We find that SAM can successfully enhance the performance of even state-of-the-art learning techniques on imbalanced datasets with a re-weighting component (\eg, VS Loss and LDAM). The major improvement ($\sim$ 6\%) comes from tail classes.

      \item We find that these observations scale to large datasets like ImageNet-LT and iNaturalist2018 along recent techniques like GLMC~\cite{du2023global}, where SAM improves the performance of re-weighting methods by a large margin.

      \item Code (w/ Models): \url{https://github.com/val-iisc/Saddle-LongTail}
  \end{itemize} 

 \\ \hline

 \begin{enumerate}[left=0.1em, start=6]
            \item \parbox[t]{\linewidth}{DeiT-LT:Distillation Strikes Back for Vision Transformer Training on Long-Tailed Datasets \\ \\
            Train a Vision Transformer (ViT) on Long-Tailed Data from scratch without pre-training, having improved generalization on tail classes.}
            \end{enumerate} &
  \vspace{-0.3cm}
  \begin{itemize}[left=0.5em]
      \item We observe that ViT training methods achieve similar accuracy as CNNs on balanced datasets. However, their performance drops drastically in comparison to CNNs on long-tailed data. This can be attributed to the lack of inductive bias present in ViTs, due to which they overfit on tail classes.

      \item To mitigate overfitting and induce generalization like CNNs, we introduce a distillation scheme using a \ttt{DIST} token that uses out-of-distribution data with an enhanced focus on tail classes through re-weighting. 

      \item To further learn a generalizable hypothesis, we propose distilling from CNNs with low-rank features trained via SAM~\cite{foret2020sharpness}. 
  
  \end{itemize} &
  \vspace{-0.3cm}
  \begin{itemize}[left=0.5em]
      \item We make \ttt{DIST} token specific to predict on tail classes, whereas the classification token \ttt{CLS} predicts well on head classes learning from ground truth. This enables the scalability of ViT to be used for learning head classes from \ttt{CLS} and learning robustly on tail classes from the CNN.

      \item The DeiT-LT scheme allows distillation from small CNNs trained on basic weak augmentations rather than large ones like DeiT~\cite{touvron2021deit}, making the distillation process compute efficient.

      \item We demonstrate the effectiveness of DeiT-LT across diverse
small-scale (CIFAR-10 LT, CIFAR-100 LT) and large-scale datasets (ImageNet-LT, iNaturalist-2018). The DeiT-LT improves over the SotA CNN from which it is distilled, demonstrating DeiT-LT's effectiveness.
      
      \item Code: \url{https://github.com/val-iisc/DeiT-LT}
  \end{itemize} \\ \hline

         \end{tabular} 
 }   
 }
\end{table}

\begin{table}[htp]

    \scalebox{0.7}{
       \renewcommand{\arraystretch}{1.08}{
        \begin{tabular}{ >{\columncolor{gray!10}} p{0.25\linewidth} | p{0.57\linewidth}| >{\columncolor{gray!10}} p{0.57\linewidth}}
            \toprule
            \multicolumn{1}{>{\columncolor{gray!10}}l|}{\textbf{Chapter}}
            &  & 
               \\
            \textbf{Problem Setup} & \multicolumn{1}{c|}{\multirow{-2}{*}{\textbf{Proposed Approach}}}& \multicolumn{1}{>{\columncolor{gray!10}}c}{\multirow{-2}{*}{\textbf{Key Contributions}}}\\
            
             \midrule
            
            \multicolumn{3}{c}{\textbf{Part III: Semi-Supervised Long-Tail Learning for Non-Decomposable Objectives}} \\
            \midrule 
           
            \begin{enumerate}[left=0.1em, start=7]
             \item \parbox[t]{\linewidth}{Cost Sensitive Self-Training for Optimising Non-decomposable Measures \\ \\ Optimize metrics like Worst-Case Recall In Semi Supervised Long-Tailed setup}
            \end{enumerate} &
  \vspace{-0.3cm}
  \begin{itemize}[left=0.5em]
      \item We find that self-training frameworks are very effective for optimizing accuracy for balanced data, but fail to perform well with imbalanced settings.

      \item We generalize self-training based on Consistency Regularization (\eg FixMatch~\cite{sohn2020fixmatch}, \etc) to a cost-sensitive setting to optimize non-decomposable metrics like worst-case recall. 
      
      \item The main idea is to introduce a cost-sensitive consistency regularizer, which gives more weight to regularization for classes that are underperforming in performance on a held-out set.

  \end{itemize} &
  \vspace{-0.3cm}
  \begin{itemize}[left=0.5em]
      \item We theoretically prove that Cost-Sensitive Self-Training
        (CSST) can optimize the non-decomposable metric by effectively utilizing the unlabeled data over Self-Training.
    
      \item We demonstrate the effectiveness of the CSST framework for optimizing non-decomposable metrics on both Vision and NLP tasks. 
      
      \item We show the scalability of the proposed method to optimize objectives for datasets like ImageNet. 
      
      \item Code: \url{https://github.com/val-iisc/CSST}
  \end{itemize} 

 \\ \hline

 \begin{enumerate}[left=0.1em, start=8]
             \item \parbox[t]{\linewidth}{Selective Mixup Fine-Tuning for Optimizing Non-Decomposable Objectives \\ \\ 
            Fine-tune a pre-trained model to optimize a particular non-decomposable objective. 
            }
            \end{enumerate} &
  \vspace{-0.3cm}
  \begin{itemize}[left=0.5em]
      \item We evaluate the non-decomposable metric performance for the existing theoretical and practical methods for learning from class-imbalanced data. These methods perform well on accuracy (mean recall) but poorly on other practical non-decomposable metrics.

      \item To optimize the non-decomposable metric performance of the model, we propose a fine-tuning technique based on selective mixup, that is, mixing up classes that improve the desired metric.

      \item The distribution of which classes are to be mixed up is determined by approximating the gain in metric by performing a certain class mixup and then taking a softmax to prefer large-gain mixups.

  \end{itemize} &
  \vspace{-0.3cm}
  \begin{itemize}[left=0.5em]
      \item We provide a tractable approximation of the gain in metric due to various mixups, even when the non-decomposable metric is non-differentiable with respect to neural network weights.
      
      \item We show that the SelMix sampling distribution is optimal compared to any other fixed mixup sampling distribution and prove a convergence result for the SelMix algorithm.

      \item We empirically demonstrate gains for various class-imbalanced settings where even the unlabeled data distribution differs from the labeled data ones. 
      
      \item Code: \url{https://github.com/val-iisc/SelMix}
  \end{itemize} \\ \hline

         \end{tabular} 
 }   
 }
\end{table}

\begin{table}[htp]

    \scalebox{0.7}{
           \renewcommand{\arraystretch}{1.08}{
        \begin{tabular}{ >{\columncolor{gray!10}} p{0.25\linewidth} | p{0.57\linewidth}| >{\columncolor{gray!10}} p{0.57\linewidth}}
            \toprule
            \multicolumn{1}{>{\columncolor{gray!10}}l|}{\textbf{Chapter}}
            &  & 
               \\
            \textbf{Problem Setup} & \multicolumn{1}{c|}{\multirow{-2}{*}{\textbf{Proposed Approach}}}& \multicolumn{1}{>{\columncolor{gray!10}}c}{\multirow{-2}{*}{\textbf{Key Contributions}}}\\
            
             \midrule
            
            \multicolumn{3}{c}{\textbf{Part IV: Efficient Domain Adaptation}} \\
            \midrule 
           
            \begin{enumerate}[left=0.1em, start=9]
            \item \parbox[t]{\linewidth}{ S3VAADA: Submodular Subset Selection for Virtual Adversarial Active Domain
Adaptation \\ \\ Select the optimal subset of samples to be labeled to improve the performance in target domain.}
            \end{enumerate} &
  \vspace{-0.3cm}
  \begin{itemize}[left=0.5em]
      \item We propose a set-based criterion for active learning, that provides a score for a subset, considering the aspects of uncertainty, diversity, and representativeness.
      
      \item We demonstrate that the criterion introduced is submodular and can be used to select near optimal subset in linear time \wrt samples, through a greedy algorithm.

      \item We introduce the Virtual Active Adversarial Domain Adaptation (VAADA) procedure to effectively utilize the selected samples to improve performance on the target domain.

  \end{itemize} &
  \vspace{-0.3cm}
  \begin{itemize}[left=0.5em]
      \item We find that the S3VAADA procedure can effectively improve performance on target by huge margins (10-15 \%) by just labeling 5\% of effectively chosen data.

      \item S3VAADA improves target adaptation performance across various domain adaptation datasets of Office-Home, Office-31 and large-scale ones like VisDa-18 and DomainNet.
      
      \item Project Page (w/ Code):  \url{https://sites.google.com/iisc.ac.in/s3vaada-iccv2021/.}
  \end{itemize} 

 \\ \hline

 \begin{enumerate}[left=0.1em, start=10]
            \item \parbox[t]{\linewidth}{ A Closer Look at Smoothness in Domain Adversarial Training \\ \\ Train a model that generalizes well on a target domain by enforcing a generalizable smooth function.
            }
            \end{enumerate} &
  \vspace{-0.3cm}
  \begin{itemize}[left=0.5em]
      \item We demonstrate that converging to a flat minimum for the task loss (\ie, cross-entropy for classification) leads to stable and improved adversarial domain adaptation. 

      \item Contrary to task loss, we theoretically and empirically demonstrate that converging to flat minima for adversarial discriminator loss leads to unstable training and poor adaptation to the target domain. 
      
      \item Based on the above analysis, we introduce Smooth Domain Adversarial Training (SDAT), which proposes to converge to minima for task loss, leaving the adversarial loss intact. This is achieved by selectively applying Sharpness Aware Minimization to all the task losses in domain adaptation.

  \end{itemize} &
  \vspace{-0.3cm}
  \begin{itemize}[left=0.5em]
      \item We find that replacing Domain Adversarial Training with a Smooth version \ie SDAT in existing state-of-the-art (SOTA) methods,  lead to a significant improvement in performance. Notably, with ViT
backbone, SDAT leads to a significant effective average gain of 3.1 \%.

\item SDAT performance improvement comes without requiring any additional module (or pre-training data) using only a 12 GB GPU.

\item SDAT can be integrated in any existing code-base with only a few lines of code-change making it a very practical option for improving generalization on target.

      \item Code: \url{https://github.com/val-iisc/SDAT}
  \end{itemize} \\ \hline

         \end{tabular} 
 }   
 }
\end{table}

\clearpage

\section{Future Directions}

In this thesis, we propose efficient and effective methods to learn from limited and imperfect data. Based on our findings in this thesis, we present below a few prospective research directions that could be explored in the future. 

\begin{itemize}
    \item \textbf{Long Tail Learning in Foundational Generative Models}
    Recently, the foundation generative models like Stable Diffusion, DALLE-2, etc. have shown
    promise, with the ability to generate images based on the text description. These models are trained on internet-scale datasets using large-scale computational resources. The Internet scale datasets also follow a long-tailed data distribution due to the inherent nature of the data on the web. Hence, a good future direction is to quantify the effect of long-tailed data on text-2-image generations. Further improving the data efficiency of these models across long-tail categories is a good avenue for future work.

    \item \textbf{Quantification of Learning from Head Classes to Tail Classes}
    One of the major focuses of long-tail learning is to be able to transfer knowledge from the populated head classes to tail classes. Despite this being the main goal, it's hard to easily quantify that how much knowledge from head classes is infused into tail classes. As for tail classes the model can either learn from head classes or learn from few-shot tail data. Hence, quantifying the learning of the tail classes in terms of head classes and few-shot learning is a necessity. Further, formalizing this notion of long-tail learning generalization bounds in terms of head-class errors and few-shot errors is a good direction to work theoretically. This makes this direction promising, having avenues of contribution.

    \item \textbf{Compositional Generalization for Long-Tailed (Few-Shot) Data.} One of the foundational models' features is their capability to generalize using few-shot samples. This behavior results from the model's ability to utilize the knowledge from head categories to identify tail categories with only a few samples. For example, let's say we give the model a task to identify a rare species like a blue hummingbird; in such a case, if the model has learned a generic representation (e.g., semantic parts) of a common hummingbird, it can just do some specific modifications to parts for tailoring it to a rare hummingbird. However, most theoretical works have only focused on analyzing the i.i.d. (independently and identically distributed) case. Hence, there is a requirement to re-think generalization by keeping the compositional properties in mind, particularly in the long-tailed setting. 
    
\end{itemize}

\appendix

\counterwithin{figure}{section}
\counterwithin{table}{section}

\makeatletter
\renewcommand\section{\@startsection {subsection}{2}{\z@}%
                                   {-3.5ex \@plus -1ex \@minus -.2ex}%
                                   {2.3ex \@plus.2ex}%
                                   {\normalfont\large\bfseries}}
\makeatother

\makeatletter
\renewcommand\subsection{\@startsection {subsubsection}{3}{\z@}%
                                      {-3.25ex\@plus -1ex \@minus -.2ex}%
                                      {1.5ex \@plus .2ex}%
                                      {\normalfont\normalsize\bfseries}}
\makeatother

\makeatother

\newcommand{\supersection}[1]{%
    \par\addvspace{1.5ex} %
    \refstepcounter{section}%
    \addcontentsline{toc}{section}{\protect\numberline{\thesection}#1} %
    \noindent{\Large\bfseries \thesection. #1\par}%
    \nobreak\smallskip
}
\makeatother

\chapter*{\Huge{Appendix}}
\pagestyle{plain}
\addcontentsline{toc}{chapter}{Appendix}

\renewcommand{\thesection}{A}

\supersection{Class Balancing GAN with a Classifier in the Loop (Chapter-2)}

\section{Proof of Proposition 2}
\label{cbgan_app:proof2}
\textbf{Proposition 2 \citep{guiacsu1971weighted}:} If each $\hat{p_k}$ satisfies Eq. \ref{cbgan_eq:lambda_val}, i.e. $\hat{p_k} = e^{\lambda N_k^t - 1}$, then the regularizer objective in  Eq. \ref{cbgan_equation:regularizer} attains the optimal minimum value of $\lambda -\sum_{k} \frac{e^{\lambda N_k^t - 1}}{N_k^t}$. \\
\textbf{Proof:} We wish to optimize the following objective:
\begin{equation}
    \displaystyle \underset{\hat{p}}{\min} \; \sum_{k} \frac{\hat{p_k}\log(\hat{p_k})}{N_k^t} \quad \text{such that} \quad \sum_{k} \hat{p_k} =1
\end{equation}
Introducing the probability constraint via the Lagrange multiplier $\lambda$:
\begin{equation}
    \displaystyle L(\hat{p}, \lambda) = \sum_{k} \frac{\hat{p_k}\log(\hat{p_k})}{N_k^t} - \lambda (\sum{\hat{p_k}} - 1) 
\end{equation}
\begin{equation}
    \displaystyle L(\hat{p}, \lambda) - \lambda = \sum_{k} \frac{\hat{p_k}\log(\hat{p_k})}{N_k^t} - \lambda (\sum{\hat{p_k}}) 
\end{equation}
\begin{equation}
    \displaystyle L(\hat{p}, \lambda) - \lambda = \sum_{k} \frac{\hat{p_k}\log(\hat{p_k}e^{-\lambda N_k^t})  }{N_k^t} 
\end{equation}
\begin{equation}
    \displaystyle L(\hat{p}, \lambda) - \lambda = \sum_{k} \frac{\hat{p_k}\log(\hat{p_k}e^{-\lambda N_k^t})  }{N_k^t} 
\end{equation}
\begin{equation}
    \displaystyle L(\hat{p}, \lambda) - \lambda = \sum_{k} \frac{e^{\lambda N_k^t}}{N_k^t} ({\hat{p_k}e^{-\lambda N_k^t}\log(\hat{p_k}e^{-\lambda N_k^t}))  }
\end{equation}
Now, for $x>0$, we know that $x\log(x) \geq -\frac{1}{e}$:
\begin{equation}
    \displaystyle L(\hat{p}, \lambda)  \geq \lambda -\sum_{k} \frac{e^{\lambda N_k^t - 1}}{N_k^t}
\end{equation}
The $xlog(x)$ attains the minimal value at only $x = \frac{1}{e}$, which is the point corresponding to $\hat{p_k}e^{-\lambda N_k^t} = 1/e$, for every $k$. This shows that at $\hat{p_k} = e^{\lambda N_k^t - 1}$ is the optimal point where the objective attains it's minimum value. This result is derived from Theorem 2 in \citet{guiacsu1971weighted}.

\label{cbgan_appendix}
\subsection{Datasets}
\label{cbgan_app: datasets}
We use CIFAR-10 \citep{krizhevsky2009learning} dataset for our experiments which has $50$K training
images and $10$K validation images. For the LSUN \citep{yu2015lsun} dataset we use a fixed
subset of $50$K training images for each of bedroom, conference room, dining room, kitchen and living room classes. In total we have $250$K training images and $1.5$K validation set of images for LSUN dataset. The imbalanced versions of the datasets are created by removing images from the training set. For the large dataset experiments, we make use of CIFAR-100 \citep{Krizhevsky09learningmultiple} and iNaturalist-2019 \citep{inat19}. The CIFAR-100 dataset is composed of the 500 training images and 100 testing images for each class. The iNaturalist-2019 is a long-tailed dataset composed of the 268,243 images present across 1010 classes in the training set, the validation set is composed of 3030 images balanced across classes.

\subsection{Architecture Details for GAN}
\label{cbgan_apx:gan_arch}
We use the SNDCGAN architecture for experiments on CIFAR-10 and SNResGAN architecture for experiments
on LSUN, CIFAR-100 and iNaturalist-2019 datasets \citep{gulrajani2017improved, miyato2018spectral}. The notation for the architecture tables
are as follows: m is the batch size, FC(dim\_in, dim\_out) is a fully connected Layer, CONV(channels\_in, channels\_out, kernel\_size, stride) is convolution layer, TCONV(channels\_in, channel\_out, kernel\_size, stride) is the transpose convolution layer, BN is BatchNorm \citep{ioffe2015batch} Layer in case of unconditonal
GANs and conditional BatchNorm in case of conditional GANs. LRelu is the leaky relu activation function and
GSP is the Global Sum Pooling Layer. The DIS\_BLOCK(channels\_in, channels\_out, downsampling) and GEN\_BLOCK(channels\_in, channels\_out, upsampling) correspond to the Discriminator and
Generator block used in \citet{gulrajani2017improved}. The SNResGAN architecture for CIFAR-100 differs by a small amount as it has $32 \times 32$ image size, for which we use the exact same architecture described in \citet{miyato2018spectral}. The architectures are presented in detail
in Tables \ref{cbgan_SNDCGAN_G}, \ref{cbgan_SNDCGAN_D}, \ref{cbgan_SNResGAN_G}  and \ref{cbgan_SNResGAN_D}.

\begin{table*}[!t]
  \centering
  \begin{tabular}{llrc}
    \toprule
    \textbf{Layer} & \textbf{Input}&\textbf{Output} & \textbf{Operation}\\
    \midrule
    Input Layer & (m, 128)&(m, 8192)&\textsc{FC}(128, 8192)\\
    \midrule
    Reshape Layer & (m, 8192)&(m, 4, 4, 512)&\textsc{Reshape}\\
    Hidden Layer & (m, 4, 4, 512)&(m, 8, 8, 256)&\textsc{TConv}(512, 256, 4, 2),\textsc{BN},\textsc{LRelu} \\
    Hidden Layer & (m, 8, 8, 256)&(m, 16, 16, 128)&\textsc{TConv}(256, 128, 4, 2),\textsc{BN},\textsc{LRelu} \\
    Hidden Layer & (m, 16, 16, 128)&(m, 32, 32, 64)&\textsc{TConv}(128, 64, 4, 2),\textsc{BN},\textsc{LRelu} \\
    Hidden Layer & (m, 32, 32, 64)&(m, 32, 32, 3)&\textsc{Conv}(64, 3, 3, 1) \\
    \midrule
    Output Layer & (m, 32, 32, 3)&(m, 32, 32, 3)&\textsc{Tanh} \\
    \bottomrule
  \end{tabular}
   \caption{Generator of SNDCGAN~\citep{miyato2018spectral, radford2015unsupervised} used for CIFAR10 image synthesis.}
   \label{cbgan_SNDCGAN_G}
\end{table*}
 
\begin{table*}[!t]

  \centering
  \begin{tabular}{llrc}
    \toprule
    \textbf{Layer} & \textbf{Input} & \textbf{Output} & \textbf{Operation}\\
    \midrule
    Input Layer & (m, 32, 32, 3)  & (m, 32, 32, 64) & \textsc{Conv(3, 64, 3, 1)}, \textsc{LRelu}\\
    \midrule
    Hidden Layer & (m, 32, 32, 64)  & (m, 16, 16, 64) & \textsc{Conv(64, 64, 4, 2)}, \textsc{LRelu}\\
    Hidden Layer & (m, 16, 16, 64)  & (m, 16, 16, 128) & \textsc{Conv(64, 128, 3, 1)}, \textsc{LRelu}\\
    Hidden Layer & (m, 16, 16, 128)  & (m, 8, 8, 128) & \textsc{Conv(128, 128, 4, 2)}, \textsc{LRelu}\\
    Hidden Layer & (m, 8, 8, 128)  & (m, 8, 8, 256) & \textsc{Conv(128, 256, 3, 1)}, \textsc{LRelu}\\
    Hidden Layer & (m, 8, 8, 256)  & (m, 4, 4, 256) & \textsc{Conv(256, 256, 4, 2)}, \textsc{LRelu}\\
    Hidden Layer & (m, 4, 4, 256)  & (m, 4, 4, 512) & \textsc{Conv(256, 512, 3, 1)}, \textsc{LRelu}\\
    Hidden Layer & (m, 4, 4, 512)  & (m, 512) & \textsc{GSP}\\
    \midrule
    Output Layer & (m, 512)  & (m, 1) & \textsc{FC}(512, 1)\\
    \bottomrule
  \end{tabular}
  \caption{Discriminator of SNDCGAN~\citep{miyato2018spectral} used for CIFAR10 image synthesis.}
  \label{cbgan_SNDCGAN_D}
\end{table*}

\begin{table*}[!t]

  \centering
  \begin{tabular}{llrc}
    \toprule
    \textbf{Layer} & \textbf{Input} & \textbf{Output} & \textbf{Operation}\\
    \midrule
    Input Layer & (m, 128)  & (m, 16384) & \textsc{FC(128, 16384)}\\
    \midrule
    Reshape Layer & (m, 16384)  & (m, 4, 4, 1024) & \textsc{Reshape}\\
    Hidden Layer & (m, 4, 4, 1024) & (m, 8, 8, 512) & \textsc{Gen\_Block}(1024, 512, True) \\
    Hidden Layer & (m, 8, 8, 512) & (m, 16, 16, 256) & \textsc{Gen\_Block}(512, 256, True) \\
    Hidden Layer & (m, 16, 16, 256) & (m, 32, 32, 128) & \textsc{Gen\_Block}(256, 128, True) \\
    Hidden Layer & (m, 32, 32, 128) & (m, 64, 64, 64) & \textsc{Gen\_Block}(128, 64, True) \\
    
    Hidden Layer & (m, 64, 64, 64) & (m, 64, 64, 3) & \textsc{BN}, \textsc{ReLU}, \textsc{Conv(64, 3, 3, 1)} \\
    \midrule
    Output Layer & (m, 64, 64, 3)  & (m, 64, 64, 3) & \textsc{Tanh} \\
    \bottomrule
    
  \end{tabular}
  \caption{Generator of SNResGAN used for LSUN and iNaturalist-2019 image synthesis.}
  \label{cbgan_SNResGAN_G}
\end{table*}
\begin{table*}[!t]

  \centering
  \begin{tabular}{llrc}
    \toprule
    \textbf{Layer} & \textbf{Input} & \textbf{Output} & \textbf{Operation}\\
    \midrule
    Input Layer & (m, 64, 64, 3)  & (m, 32, 32, 64) & \textsc{Dis\_Block}(3, 64, True)\\
    \midrule
    Hidden Layer & (m, 32, 32, 64)  & (m, 16, 16, 128) & \textsc{Dis\_Block}(64, 128, True)\\
    Hidden Layer & (m, 16, 16, 128)  & (m, 8, 8, 256) & \textsc{Dis\_Block}(128, 256, True)\\
    Hidden Layer & (m, 8, 8, 256)  & (m, 4, 4, 512) &      \textsc{Dis\_Block}(256, 512, True)\\
    Hidden Layer & (m, 4, 4, 512)  & (m, 4, 4, 1024) & \textsc{Dis\_Block}(512, 1024, False), \textsc{ReLU}\\
    Hidden Layer & (m, 4, 4, 1024)  & (m, 1024) & \textsc{GSP}\\
    \midrule
    Output Layer & (m, 1024)  & (m, 1) & \textsc{FC(1024, 1)}\\
    \bottomrule
    
  \end{tabular}
    \caption{Discriminator of SNResGAN \citep{miyato2018spectral, gulrajani2017improved} used for LSUN and iNaturalist-2019 image synthesis.}
    \label{cbgan_SNResGAN_D}
\end{table*}

\subsection{Hyperparameter Configuration (Image Generation Experiments)}

\vspace{1mm} \noindent \textbf{Lambda  the Regularizer coeffecient}
The $\lambda$ hyperparameter is the only hyperparameter that we change
across different imbalance scenarios. As the overall objective is 
composed of the two terms:
\begin{equation}
    L_{g} = - E_{(x, z) \sim (P_r, P_z)}[\log(\sigma(D(G(z)) - D(x))] + \lambda L_{reg}
\end{equation}
As the number of terms in the regularizer objective can increase
with number of classes $K$. For making the regularizer term invariant of $K$
and also keeping the scale of regularizer term similar to GAN loss, we normalize it by $K$. Then the loss is multiplied by $\lambda$. Hence
the effective factor that gets multiplied with regularizer term is $\frac{\lambda}{K}$.

The presence of pre-trained classifier which provides labels for generated images makes it easy to determine the value of $\lambda$. Although the pre-trained classifier is trained on long-tailed
data its label distribution is sufficient to provide a signal for balance in generated distribution. We use the KL Divergence of labels with respect to uniform
distribution for $10$k samples in validation stage to check for balance in distribution and choose $\lambda$ accordingly.
We use the FID implementation available here \footnote{https://github.com/mseitzer/pytorch-fid}.

\begin{figure*}
\begin{subfigure}{.5\textwidth}
  \centering
  \includegraphics[width=.8\linewidth]{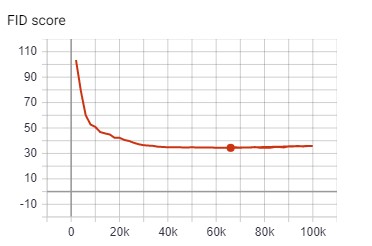}
   \caption{GAN}
  \label{cbgan_fig:fid_gan}
\end{subfigure}%
\begin{subfigure}{.5\textwidth}
  \centering
  \includegraphics[width=.8\linewidth]{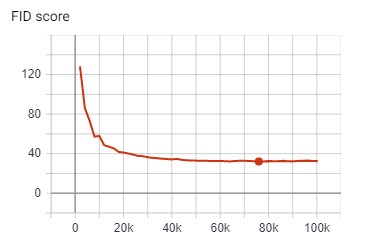}
  \caption{GAN + Proposed Regularizer}
  \label{cbgan_fig:fid_our_gan}
\end{subfigure}
\caption{Plots of FID (y axis) vs Number of iteration steps. We observe a similar curve in both the cases.}
\label{cbgan_fig:fid_curve}
\end{figure*}

\vspace{1mm} \noindent \textbf{Other Hyperparmeters:}
\begin{figure}[!t]
    \centering
    \includegraphics[width=0.75\linewidth]{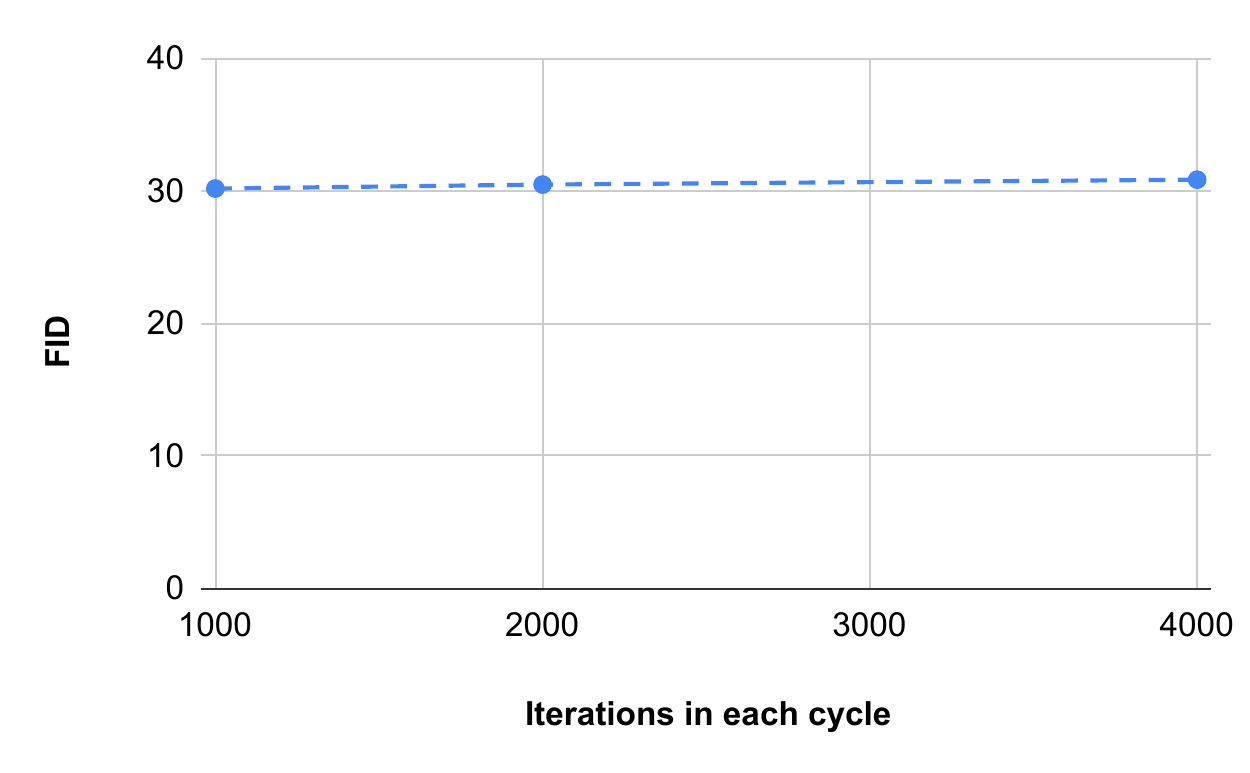}
    \caption{Effect on FID with change in number of steps in each cycle. After each cycle effective class statistics are updated (For CIFAR-10 imbalance ratio $\rho = 10$).}
    \label{cbgan_fig:fid_stats}
\end{figure}
We update the effective class distribution periodically after $2$k updates (i.e. each cycle defined in section \ref{cbgan_sec:method} consists of 2k iteration). We find the algorithm performance to be stable for a large 
range of update frequency depicted in Figure \ref{cbgan_fig:fid_stats}.
We also apply Exponential Moving Average on generator weights after $20$k steps for better generalization. The hyperparameters are present in detail in Table \ref{cbgan_tab:hyperparams}.
\\
\textbf{Validation Step:} We obtain the FID on 10k generated samples after each $2$k iterations and choose the checkpoint with best FID for final sampling and FID present in
Table \ref{cbgan_tab:gan_result}. 
\\
\textbf{Convergence of Network}: We find that our GAN + Regularizer setup also achieves similar convergence in FID value to the GAN without the regularizer. We show the FID curves for the CIFAR-10 (Imbalance Ratio = 10) experiments in Figure \ref{cbgan_fig:fid_curve}. \\
\textbf{Ablation on $\beta$:} We find that for the CIFAR-10 dataset ($\rho = 10$) the choice of $\beta=1$ obtains similar FID ($30.48$) to $\beta=\alpha$ which obtains FID of $30.46$. The KL Divergence is also approximately the same for both cases i.e. $0.01$.

\begin{table}[!ht]
    \centering
    \begin{tabular}{lcc}
    \hline
         Parameter & Values(CIFAR-10) & Values(LSUN)\\ \hline
         Iterations & 100k & 100k\\ 
         $\beta$   & 1 & 1 \\
         Generator lr&  0.002&  0.002 \\ 
         Discriminator lr& 0.002 & 0.002\\ 
         Adam ($\beta_1$) & 0.5 & 0.0\\ 
         Adam ($\beta_2$) & 0.999 & 0.999\\ 
         Batch Size & 256 & 256\\ 
         EMA(Start After) & 20k & 20k\\ 
         EMA(Decay Rate) & 0.9999 & 0.9999\\ \hline
    \end{tabular}
    \caption{Hyperparameter Setting for Image Generation Experiments.}
    \label{cbgan_tab:hyperparams}
\end{table}

\vspace{1mm} \noindent \textbf{Hyperparameters for the Semi Supervised GAN Architecture:}
\label{cbgan_sub:semi-sup-hyper}
We use a ImageNet and ImageNet-21k pre-trained model with ResNet 50 architecture as the base model. The fine tuning of the model on CIFAR-10 and LSUN has been done by using the code of notebook present here
\footnote{https://github.com/google-research/big\_transfer/blob/master \\/colabs/big\_transfer\_pytorch.ipynb}. The accuracy of the classifiers fine-tuned on validation data, trained with $0.1\%$ of labelled data is 84.96 \% for CIFAR-10 and 82.40 \% for LSUN respectively.
The lambda (regularizer coefficient) values are present in the table below:

\begin{table}[!h]
    \centering
    \begin{tabular}{lccc}
        \hline
        Imbalance Ratio ($\rho$) & 100  &  10 \\ \hline
        CIFAR-10 & 10 & 7.5\\ 
        LSUN & 10 & 7.5 \\ \hline
        
    \end{tabular}
    \caption{Values of $\lambda$ for different imbalance cases. For LSUN the $\lambda$ gets divide by 5 and for $\lambda$ it gets divided by 10 before multiplication to regularizer term.}
    \label{cbgan_tab:my_label}
\end{table}

The training hyper parameters are same as the ones present in the Table \ref{cbgan_tab:hyperparams}. Only in case of LSUN semi supervised experiments we use a batch size of 128 to fit into GPU memory for semi supervised experiments.

\label{cbgan_app: hyperparams}

\subsection{Experimental Details on CIFAR-100 and INaturalist-2019}
\label{cbgan_app:largedatasets}
\vspace{1mm} \noindent \textbf{CIFAR-100:}In this section we show results on CIFAR-100 dataset which has 100 classes having 500 training images for each class. We use SNResGAN architecture from \cite{miyato2018spectral} for generating $32 \times 32$ size images, which is similar to SNResGAN  architecture used for LSUN experiments. We use the same hyperparameters used for LSUN experiments listed in Table \ref{cbgan_tab:hyperparams}. We use a $\lambda$ value of 0.5 for CIFAR-100 experiments, the effective value that will get multiplied with $L_{reg}$ is $\frac{0.5}{100}$. The results in Table \ref{cbgan_tab:cifarinat} show that our method on long-tailed CIFAR100 of using GAN + Regularizer achieves the best FID and also have class balance similar to cGAN (conditional GAN). The labels for the samples generated by GAN are obtained by a classifier trained on balanced CIFAR-100 dataset for KL Divergence calculation. The KL Divergence between the GAN label distribution and uniform distribution is present in Table \ref{cbgan_tab:cifarinat}. The classifier for obtaining class labels for KL Divergence evaluation is trained on balanced CIFAR-100 with setup described in~\ref{cbgan_sec:pretrained_classifier} which serves as annotator for all methods. The balanced classifier achieves an accuracy of 70.99\% and the pre-trained classifier trained on long-tailed data achieves an accuracy of 57.63\%. The pre-trained classifier is used in the process of GAN training. 

\vspace{1mm} \noindent \textbf{iNaturalist-2019:} We use SNResGAN architecture described in Table \ref{cbgan_SNResGAN_D} and Table \ref{cbgan_SNDCGAN_G} for generating $64 \times 64$ images for the iNaturalist 2019 dataset. In case of iNaturalist all batch-norms are conditional batch norms (cBN) in Generator, in case of our method and the unconditional baseline (SNResGAN) we use random labels as conditioning labels. As in our method have access to class distribution $N_k^t$ we sample random labels with weight distribution proportional to $1/N_k^t$. With this change we see an FID decrease from 11.58 to 9.01. The $\lambda$ value of 0.5 and $\alpha = 0.005$ is used for the experiments. The effective value of $\lambda$ that will be multiplied with $L_{reg}$ is $\frac{0.5}{1010}$. The statistics $N_k$ is updated for each iteration in this case. We follow SAGAN \citep{pmlr-v97-zhang19d} authors recommendation and use spectral normalization in generator in addition to the discriminator for stability on large datasets.  Other hyperparameters are present in Table \ref{cbgan_tab:hyperparams_cifarinat}.

\begin{table}[h]
    \centering
    \begin{tabular}{lcc}
    \hline
         Parameter & Values(CIFAR-100) & Values(iNat19)\\ \hline
         Iterations & 100k & 200k\\ \
         $\beta$   & 1 & $\alpha$ \\ \
         Generator lr&  0.002&  0.002 \\ \
         Discriminator lr& 0.002 & 0.002\\ \
         Adam ($\beta_1$) & 0.5 & 0.0\\ \
         Adam ($\beta_2$) & 0.999 & 0.999\\ \
         Batch Size & 256 & 256\\ \
         EMA(Start After) & 20k & 20k\\ \
         EMA(Decay Rate) & 0.9999 & 0.9999\\ \hline
    \end{tabular}
    \caption{Hyperparameter Setting for Image Generation Experiments on CIFAR-100 and iNatuaralist-2019}
    \label{cbgan_tab:hyperparams_cifarinat}
\end{table}
The pre-trained classifier for iNaturalist-2019 is ResNet-32 trained with usual cross entropy loss with learning schedule as described in Appendix~\ref{cbgan_sec:pretrained_classifier}. The classifier is trained on resolution of $224 \times 224$ and achieves an accuracy of 46.90 \% on the validation set. As in iNaturalist 2019 case we don't have balanced training set available we use this classifier to get the KL Divergence from uniform distribution.
\newpage

\begin{figure*}[!h]
    \centering
    \includegraphics[width=\textwidth]{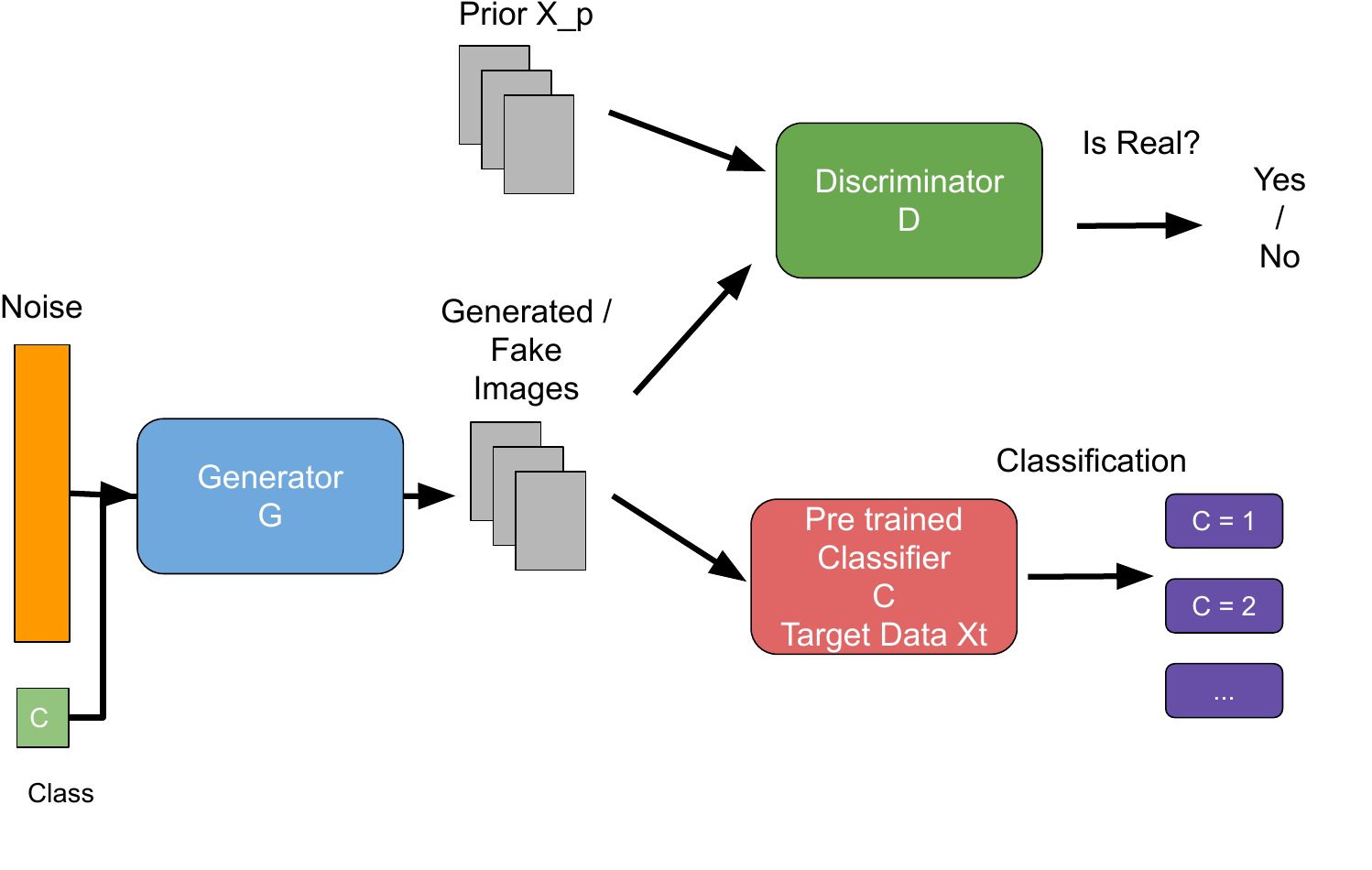}
    \caption{Diagram showing the overview of the other baseline.}
    \label{cbgan_fig:approach_other_baseline}
\end{figure*}

\subsection{Pre-trained Classifier Baseline}
\label{cbgan_app:other_baseline}
\begin{figure*}[!t]
\centering
\begin{subfigure}{0.5\columnwidth}
   \centering
  \includegraphics[width=0.9\columnwidth]{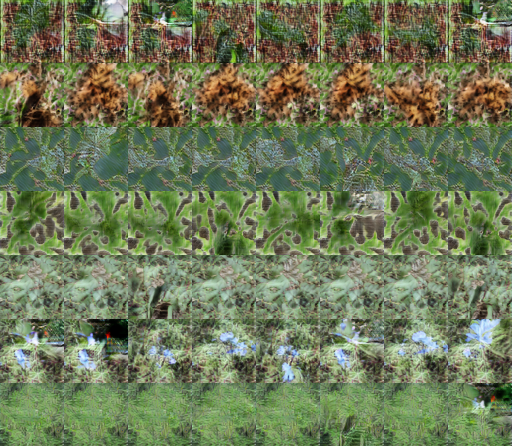}
  \caption{Classifier \\ Baseline}
\end{subfigure}%
\begin{subfigure}{0.5\columnwidth}
   \centering
  \includegraphics[width=0.9\columnwidth]{figures/1_a_cbgan/inat_ours.png}
  \caption{Our \\ method}
\end{subfigure}

\caption{Qualitative comparison of images generated by ACGAN like baseline and our method.}
\label{cbgan_fig:other_baseline_images}
\end{figure*}

In these experiments we introduce a pre-trained classifier in place of the classifier being jointly learned by ACGAN on iNaturalist-2019 dataset. This makes the comparison fair as both approaches use a pre-trained classifier and unlabelled images. The hyperparameters used are same as present in Table \ref{cbgan_tab:hyperparams_cifarinat}. We use a classification loss (i.e. cross entropy loss) in addition to the GAN loss similar to ACGAN. The $\lambda$ value for the classification loss is set to $0.5$. The illustration of the approach is present in the Figure \ref{cbgan_fig:approach_overview}. We find that this approach leads to mode collapse and is not able to produce diverse samples within each class. The comparison of the generated images is present in the Figure \ref{cbgan_fig:other_baseline_images}. \\

\clearpage
\begin{figure*}[!t]
\centering
\begin{subfigure}{.4\textwidth}
  \centering
  \includegraphics[width=.8\linewidth]{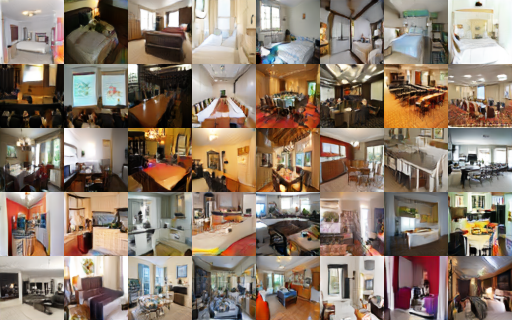}  
  \caption{ACGAN (Conditional)}
  
\end{subfigure}
\begin{subfigure}{.4\textwidth}
  \centering
  \includegraphics[width=.8\linewidth]{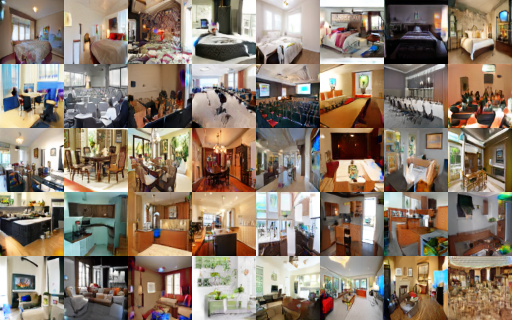}  
  \caption{cGAN (Conditional)}
  
\end{subfigure}

\begin{subfigure}{.4\textwidth}
  \centering
  \includegraphics[width=.8\linewidth]{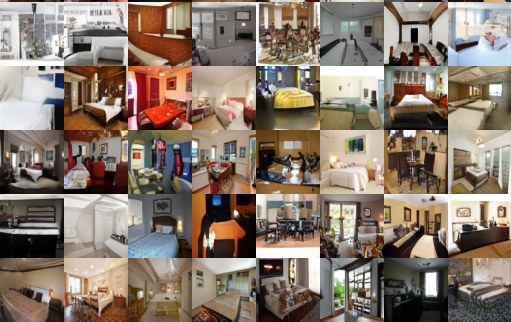}  
  \caption{SNResGAN (Unconditional)}
  
\end{subfigure}
\begin{subfigure}{.4\textwidth}
  \centering
  \includegraphics[width=.8\linewidth]{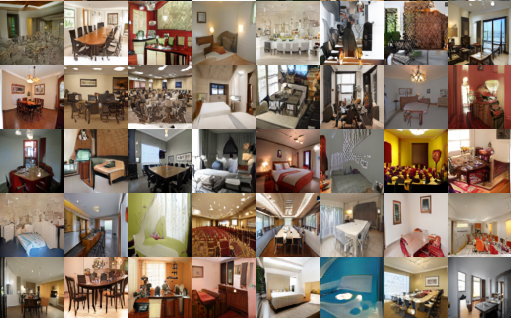}  
  \caption{Ours (Unconditional)}
  
\end{subfigure}
\caption{Images from different GANs with imbalance ratio ($\rho = 10$)}
\label{cbgan_fig:fig}
\end{figure*}

\begin{figure*}[!ht]
    \centering
    \includegraphics{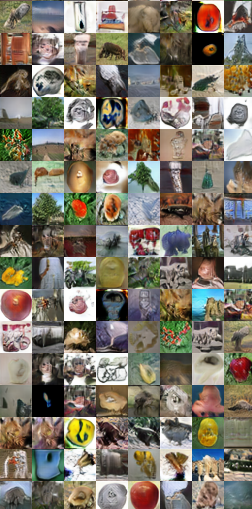}
    \caption{Images generated for CIFAR-100 dataset with our method (GAN + Regularizer).}
    \label{cbgan_fig:cifar100}
\end{figure*}

\renewcommand{\thesection}{B}
\setcounter{subsection}{0}%

\supersection{Improving GANs for Long-Tailed Data through Group Spectral Regularization (Chapter 3)}

\section{Notations}

\noindent We summarize the notations used in the chapter in Table~\ref{gsr_tab:notations}.

\begin{table}[!b]{

\vspace{-0.4cm}
\caption{{Additional metrics on CIFAR-10 dataset.}}
\label{gsr_tab:intra-fid}}
\resizebox{\linewidth}{!}
{ 
\begin{tabular}{lccccc|ccccc}
\toprule
Imb. Ratio ($\rho$)                & \multicolumn{5}{c|}{100}                                                                          & \multicolumn{5}{c}{1000}                                                                          \\
                                  & \multicolumn{1}{l}{Intra-class FID} & Precision & Recall & \multicolumn{1}{l}{Density} & Coverage & \multicolumn{1}{l}{Intra-class FID} & Precision & Recall & Density & \multicolumn{1}{l}{Coverage} \\ \cmidrule(l){2-11} 
SNGAN   & 78.36                               & 0.69      & 0.53   & 0.67                        & 0.51     & {121.57}          & 0.60      & \textbf{0.40}    & 0.43    & 0.32                         \\
 \; + gSR (Ours) & \textbf{55.71}                               & \textbf{0.71}      & \textbf{0.56}   & \textbf{0.76}                        & \textbf{0.67}     & \textbf{108.12}                              & \textbf{0.63}      & 0.39   & \textbf{0.53}    & \textbf{0.34}                         \\ \midrule
BigGAN       & 57.82                               & 0.65      & \textbf{0.58}   & 0.63                        & 0.67     & 109.29                              & 0.56     & 0.50   & 0.40    & 0.40                          \\
 \; + gSR (Ours) & \textbf{43.41}                               & \textbf{0.74}      & 0.56   & \textbf{0.93}                        & \textbf{0.80}      & \textbf{98.59}                               & \textbf{0.59}      & \textbf{0.51}   & \textbf{0.49}    & \textbf{0.51}                         \\ \bottomrule
\end{tabular}
}

\vspace{-0.5cm}
\end{table}

\begin{table}[t]
\centering
\caption{{Notation Table}}
\label{gsr_tab:notations}
\resizebox{\linewidth}{!}{%
\begin{tabular}{p{0.1\linewidth} p{0.225
\linewidth} p{0.65\linewidth}}
\toprule
Symbol                & Space                              & Meaning                                                                               \\ \midrule
$K$                     & $\mathbb{N}$                       & Number of Classes                                                                     \\
$y$                     & \{1, 2, ..., $K$\}                     & Class label                                                                           \\
$\mathbf{z}$            & $\mathbb{R}^{256}$                   & Noise vector                                                                          \\
$D$                     &                                    & Discriminator                                                                         \\
$G$                     &                                    & Generator                                                                             \\
$\mathbf{x}$            & $\mathbb{R}^{3 \times H \times W}$ & Image                                                                                 \\
$\mathbf{x^l_y}$        & $\mathbb{R}^d$                     & Feature vector from the Generator's l$^{th}$ cBN's input feature map                    \\
$\mathbf{\mu^l_B}$      & $\mathbb{R}^d$                     & Mean of incoming features to Generator's l$^{th}$ cBN from minibatch $B$                \\
$\mathbf{\sigma^l_B}$ & $\mathbb{R}^d$                     & Std. dev. of incoming features to Generator's l$^{th}$ cBN from minibatch $B$           \\
$\mathbf{\gamma^l_y}$   & $\mathbb{R}^d$                     & Gain parameter for $y^{th}$ class of $l^{th}$ cBN layer of Generator                  \\
$\mathbf{\beta^l_y}$    & $\mathbb{R}^d$                     & Bias parameter for $y^{th}$ class of $l^{th}$ cBN layer of Generator                  \\
$n_g$                   & $\mathbb{R}$                       & Number of groups                                                                      \\
$n_c$                   & $\mathbb{R}$                       & Number of columns                                                                     \\
$\mathbf{\Gamma^l_y}$    & $\mathbb{R}^{n_g \times n_c}$      & $\mathbf{\gamma^l_y}$ after grouping                                                  \\
$\mathbf{B^l_y}$       & $\mathbb{R}^{n_g \times n_c}$      & $\mathbf{\beta^l_y}$ after grouping                                                   \\
$\sigma_{max}$          & $\mathbb{R}^+$                     & Spectral norm                                                                         \\
$n_y$                   & $\mathbb{N}$                       & Number of samples in class $y$                                                        \\
$\rho$                  & $\mathbb{R}$                       & Imbalance ratio: Ratio between the most and the least frequent classes of the dataset \\ \bottomrule
\end{tabular}}
\end{table}

\section{Addtional Metrics}
In addition to FID and IS reported for experiments in chapter, we also evaluate additional metrics of Precision \cite{kynkaanniemi2019improved}, Recall \cite{kynkaanniemi2019improved}, Density~\cite{yu2020inclusive} and Coverage~\cite{yu2020inclusive} and Intra-FID for CIFAR-10 dataset.
We observe that across all the 4 different imbalance configurations (as in chapter Table {\color{red} 2}) there is significant improvement in all metrics but Recall (which is comparable to baseline in all cases.

\section{Correlations between Spectral Norms and Class-Specific Mode Collapse}
\begin{figure*}[ht]
     \centering
     \begin{subfigure}[b]{\textwidth}
         \centering
         \includegraphics[width=\textwidth, ]{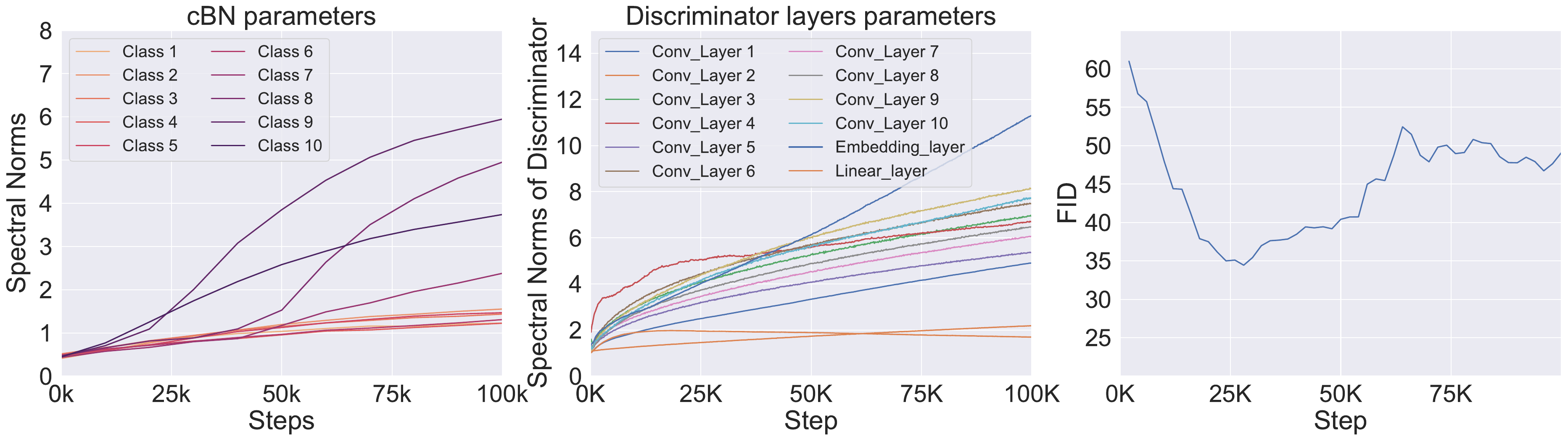}
         \caption{Without gSR}
         \label{gsr_fig:wo_gsr}
     \end{subfigure}
     \hfill
     \begin{subfigure}[b]{\textwidth}
         \centering
         \includegraphics[width=\textwidth, ]{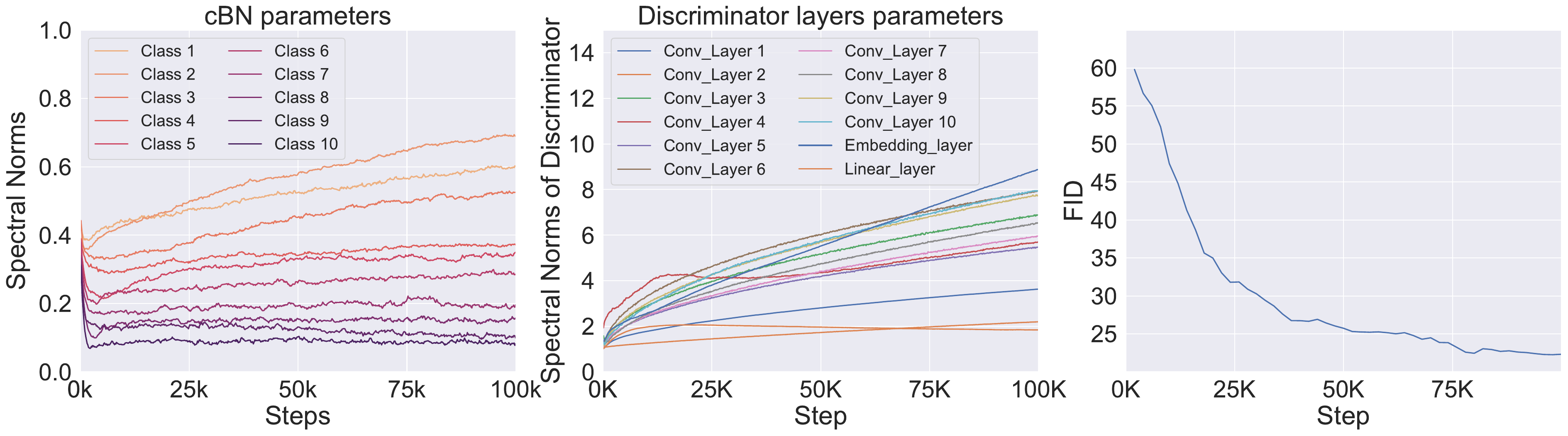}
         \caption{With gSR}
         \label{gsr_fig:w_gsr}
     \end{subfigure}
     
    \caption{{Class-specific mode collapse exhibits unique behaviour with respect to cBN parameters.} Class-specific mode collapse leads to spectral explosion in Generator's cBN parameters' spectral norms (left), which correlates with explosion of FID (right), while having little effect on discriminator's parameters' spectral norms (middle). Class-specific mode collapse is remedied by gSR which keeps the cBN parameters' spectral norms under control.}
        \label{gsr_fig:disc_specnorms}
        \vspace{-4mm}
\end{figure*}

In this section, we provide additional details and comparisons to emphasize the differences between class-specific mode collapse and the usual mode collapse (as described in chapter Sec. \textcolor{red}{3.2}).
In SNGAN~\cite{miyato2018spectral} and  BigGAN~\cite{brock2018large}, the discriminator's (D) weights' spectral norms tend to explode as the mode collapse occurs for balanced data. To determine if this also occurs in long-tailed case we train a SNGAN on CIFAR-10 ($\rho$ = 100) (with and without gSR) and plot the spectral norm of weights of discriminator layers. We find that spectral explosion for discriminator weights is not observed in the class-specific mode collapse (without gSR case), as we report in Fig.~\ref{gsr_fig:disc_specnorms}. Discriminator's layers' spectral norms do not show significant change before and after applying gSR . On the other hand, before applying gSR the spectral norms of class-specific parameters of cBN explode (at step 25k and 50k). At the same stage FID suddenly increases, whereas there is no anomaly in Discriminator's spectral norms'.  Thus, the class-specific mode collapse behaviour is different as compared to that of the mode collapse previously reported in the literature \cite{miyato2018cgans, brock2018large}, and cannot be detected through discriminator spectral norms. Hence, it's detection requires the analysis of spectral norms of grouped parameters in cBN  which we propose in this chapter. 
 
 The above spectral explosion of the generator's cBN motivates us to formulate gSR (Sec.\textcolor{red}{3.3}). We find (Fig. \ref{gsr_fig:disc_specnorms}) that after applying gSR there is no spectral collapse and training is stabilized (decreasing FID).

\section{Analysis of Covariance of grouped cBN Parameters}

\begin{figure*}[!ht]
     \centering
     \begin{subfigure}[b]{\textwidth}
         \centering
         \includegraphics[width=\textwidth, height=140pt]{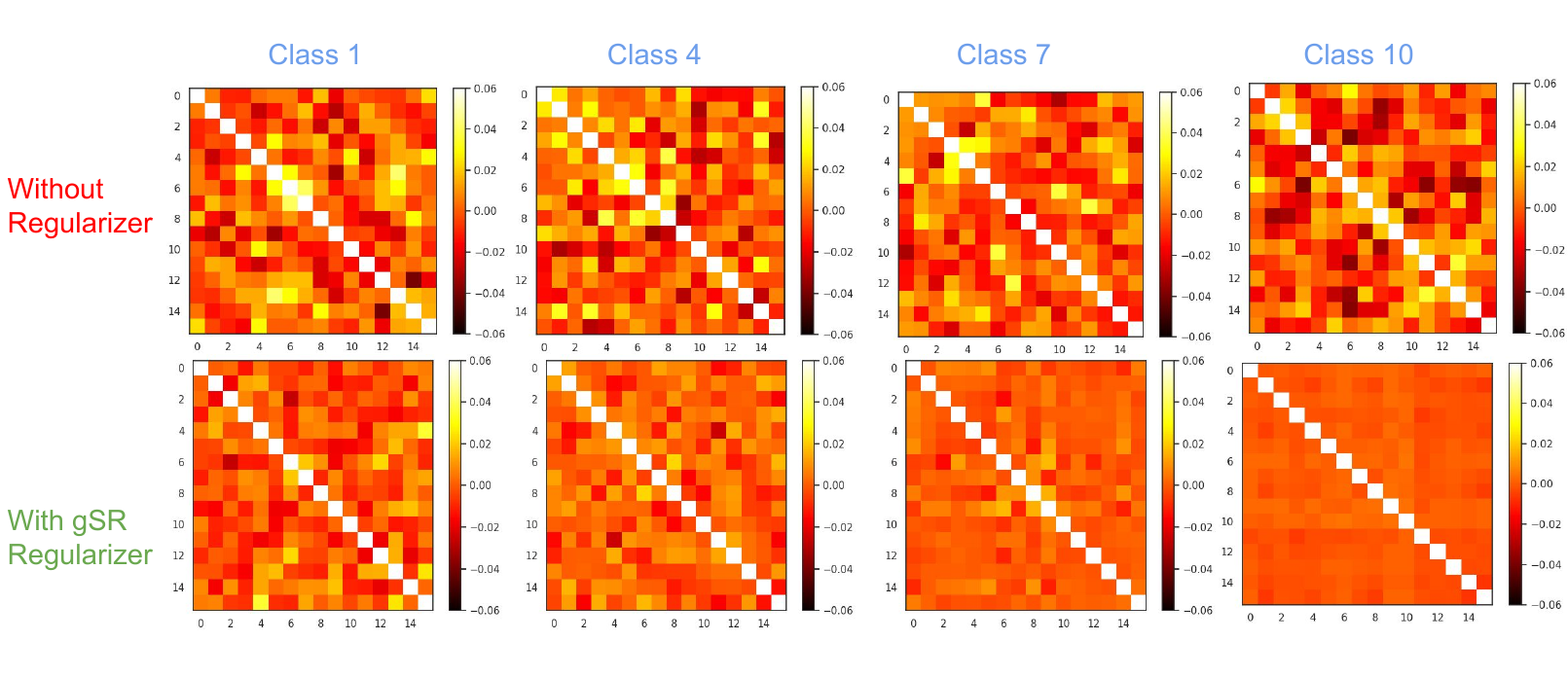}\\[-3ex]
         \caption{Covariance matrices of $\mathbf{\Gamma_y^l}$ for \textbf{(l = 1)} for SNGAN baseline.}
         \label{gsr_fig:y equals x}
     \end{subfigure}
     \hfill
     \begin{subfigure}[b]{\textwidth}
         \centering
         \includegraphics[width=\textwidth, height=140pt]{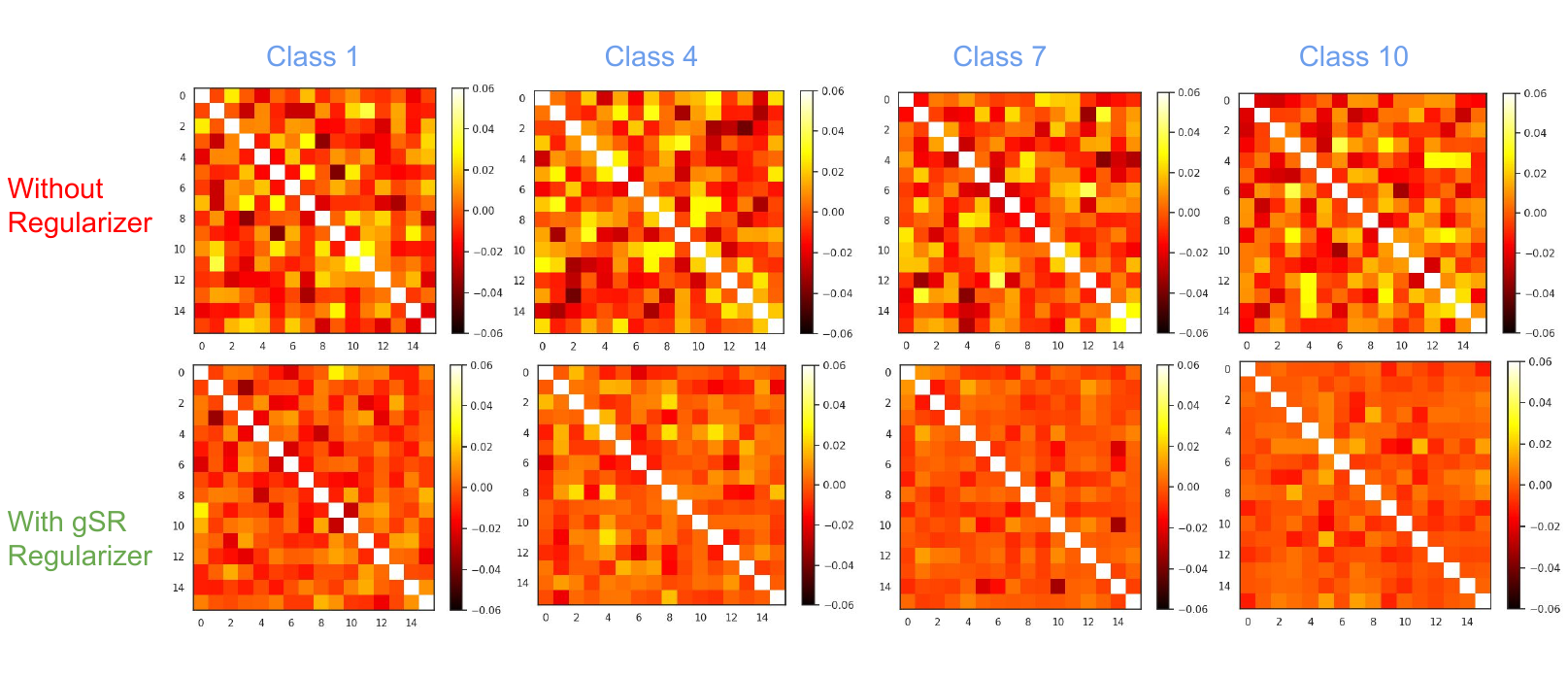}\\[-3ex]
         \caption{Covariance matrices of $\mathbf{\Gamma_y^l}$ for \textbf{(l = 3)} for SNGAN baseline.}
         \label{gsr_fig:three sin x}
     \end{subfigure}
     \hfill
     \begin{subfigure}[b]{\textwidth}
         \centering
         \includegraphics[width=\textwidth, height=140pt]{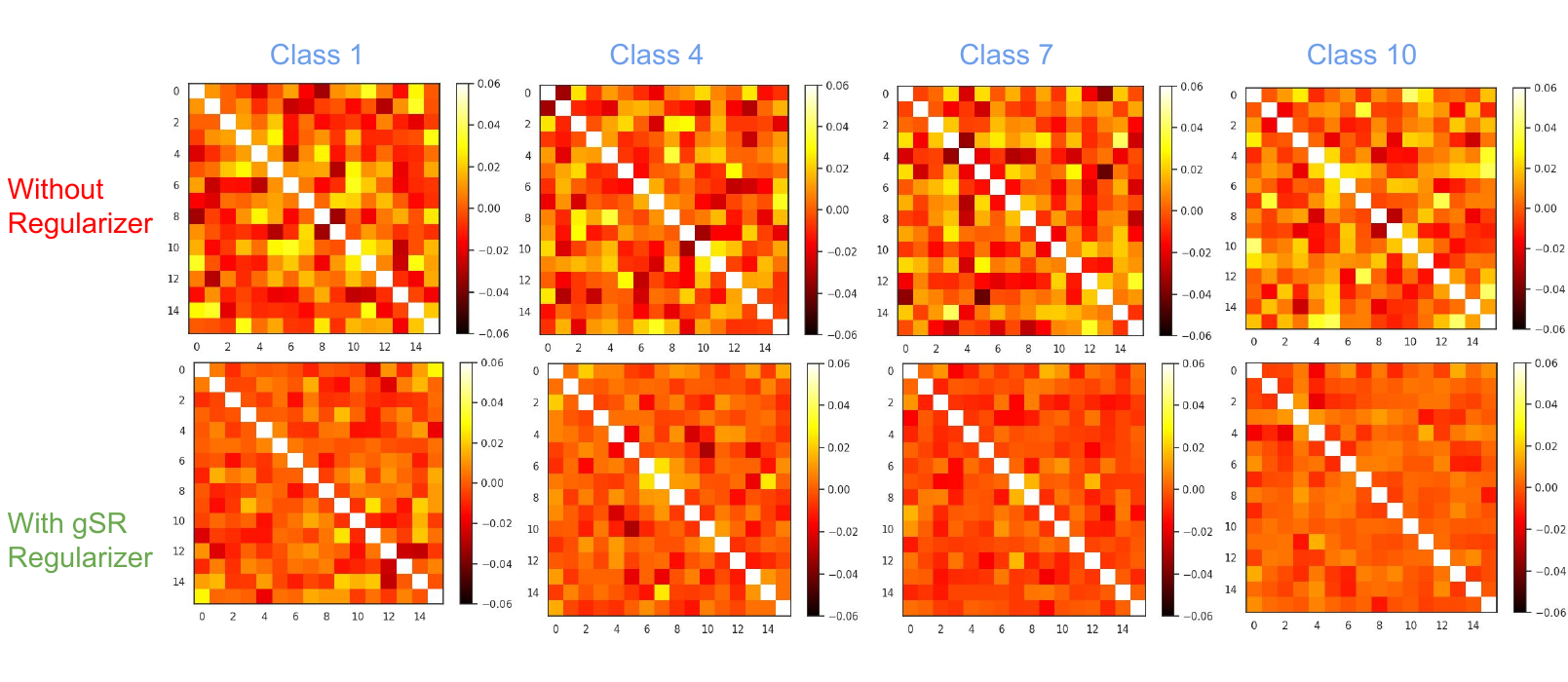}\\[-3ex]
         \caption{Covariance matrices of $\mathbf{\Gamma_y^l}$ for \textbf{(l = 5)} for SNGAN baseline.}
         \label{gsr_fig:five over x}
     \end{subfigure}
    \caption{{Covariance matrices of $\mathbf{\Gamma_y^l}$ for SNGAN baseline on CIFAR-10 ($\rho = 100$).}}
    \label{gsr_fig:three graphs}
\end{figure*}
\vspace{-3mm}
For analyzing the decorrelation effect of gSR (explained in Sec. \textcolor{red}{3.3}), we train a SNGAN on CIFAR-10 ($\rho$=100) with gSR. 
We then visualize the covariance matrices of $\mathbf{\Gamma^l_y}$ (grouped $\mathbf{\gamma^l_y}$) across cBN at different layers $\mathbf{l}$ in the generator. gSR leads to suppression of covariance between off-diagonal features of $\mathbf{\Gamma^l_y}$ belonging to the tail classes, implying decorrelation of parameters (Sec. \textcolor{red}{3.3}). As we go from initial to final cBN layers of the Generator, we see that this suppression is reduced in the case when gSR is applied. This leads to increased similarity between the covariance matrices of the head class and tail class. This effect can be attributed to the features learnt at the respective layers. The initial layers (in G) are responsible for more abstract and class-specific features, whereas the final layers produce features while are more fine-grained and generic across different classes. This is in contrast to what is observed for a classifier, as the generator is an inverted architecture in comparison to a classifier.

\section{Qualitative Results}
\begin{figure*}[t]    
    \centering
    \includegraphics[width=0.75\textwidth]{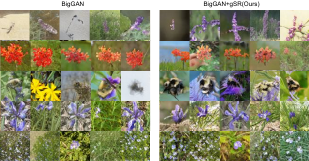}
    \caption{{Qualitative comparison of BigGAN variants on Tail classes from iNaturalist 2019 dataset ($\rho$=100) (64 $\times$ 64).} Each row represents images from a distinct class. }
    \label{gsr_fig:inat_64}
    \vspace{-3mm}
\end{figure*}

\begin{figure*}[!t]
    \centering
    \includegraphics[width=\linewidth]{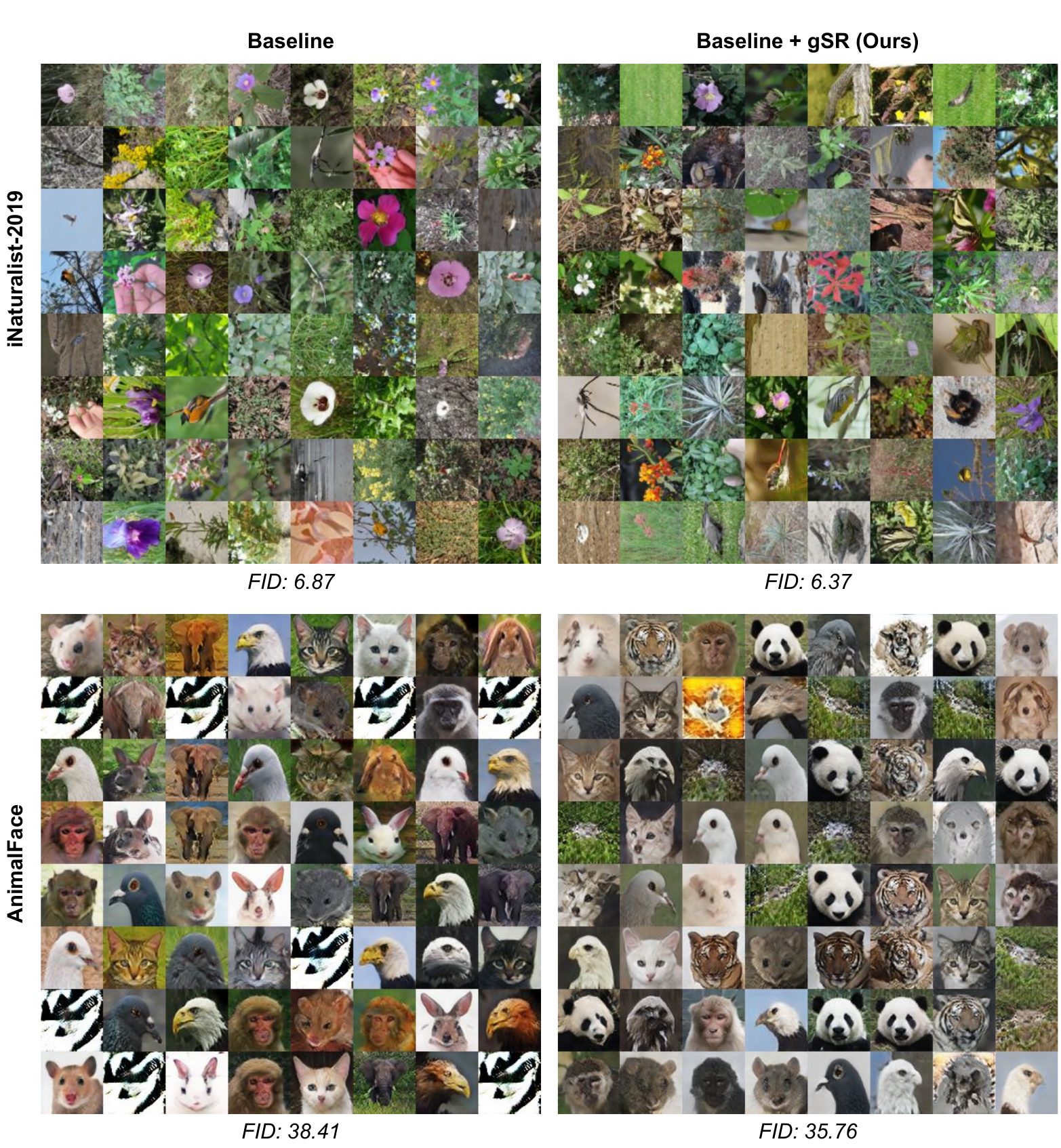}
    \caption{{Qualitative Results.} The baseline is composed of BigGAN~\cite{brock2018large}+LeCam  \cite{tseng2021regularizing} +DiffAug \cite{zhao2020differentiable}. gSR improves the quality and diversity of the images generated by baseline over challenging iNaturalist-19 and AnimalFace datasets.}
    \label{gsr_fig:qual_results_supp}
\end{figure*}

We show generated images on iNaturalist-2019 and AnimalFace in Fig.~\ref{gsr_fig:qual_results_supp} and Fig. \ref{gsr_fig:inat_64}. These are naturally occurring challenging data distributions for training a GAN. Sample diversity as well as quality is improved after applying our gSR regularizer. We also provide a video showing class specific collapse for BigGAN for CIFAR-10 in \texttt{gSR.mp4}.

\section{Experimental Details}
In this section, we elaborate on the technical and implementation details provided in Sec. \textcolor{red}{4} of the chapter.
\subsection{Datasets}
We describe the datasets used in our work below:

\vspace{1mm}\noindent\textbf{CIFAR-10}: We use CIFAR-10 \cite{krizhevsky2009learning} dataset which comprises of 32 $\times$ 32 images. The dataset is split into $50$k training images and $10$k test images. We use the training images for GAN training and the $10$k test set for calculation of FID. 

\vspace{1mm}\noindent\textbf{LSUN}:
We use a 250k subset of LSUN~\cite{yu2015lsun} dataset as followed by \cite{rangwani2021class, santurkar2018classification}, which is split across the classes of bedroom, conference room, dining room, kitchen and living room classes. We use a balanced subset of 10k images balanced across classes for FID calculation.

\vspace{1mm}\noindent\textbf{iNaturalist-2019}:
The iNaturalist-2019~\cite{inat19} is a long-tailed dataset composed of 268,243 images present across 1010 classes in the training set. The validation set contains 3030 images balanced across classes, used for FID calculation.

\vspace{1mm}\noindent\textbf{AnimalFace~\cite{si2011learning}}: The AnimalFace dataset contains 2,200 RGB images across 20 different categories with images containing animal faces. We use the training set for calculation of FID as there is no seperate validation set provided for baselines. Our results on this dataset show that our regularizer can also help in preventing collapse in extremely low data (i.e. few shot) scenario's as well.

\subsection{LeCam Regularizer}
We use LeCam regularizer~\cite{tseng2021regularizing} for all our experiments.
{\small
\begin{align}
R_{LC} = \mathop{\mathbb{E}}_{\mathbf{x}\sim\mathcal{T}}[\Vert D(\mathbf{x})-\alpha_F\Vert^2] + \mathop{\mathbb{E}}_{\mathbf{z}\sim p_{\mathbf{z}}}[\Vert D(G(\mathbf{x}))-\alpha_R\Vert^2]
\label{gsr_eq:lc_reg}
\vspace{-1mm}
\end{align}
}
\noindent \vspace{1mm} LeCam regularizer computes exponential moving average of discriminator outputs for real and generated images. The difference between discriminator outputs for real and generated images is taken against the moving averages of discriminator outputs of generated images ($\alpha_F$) and real images ($\alpha_R$) respectively. This does not allow the discriminator to output predictions with very high confidence, thereby preventing overfitting by keeping the predictions in a particular range. We use the $\lambda_{LC}$ value of 0.1, 0.3 and 0.01 as suggested by the authors \cite{tseng2021regularizing} ,which is specified in Table~\ref{gsr_tab:hyper}. The term $\lambda_{LC}R_{LC}$ is then added to discriminator loss for regularization.

\subsection{Spectral Norm Computation Time}
Since our regularizer involves estimating largest singular value for $\mathbf{{\Gamma}_y^l}$, this can be done through either power iteration or SVD. We use power iterations method to calculate the singular values of $\mathbf{\Gamma^l_y}$ and $\mathbf{B^l_y}$. We use 4 power iterations for estimating the largest singular value. For perfect decorrelation, other techniques like Group Whitening \cite{huang2021group} can also be used, but they involve full SVD computation. We provide a comparison of time for 100 generator steps of training for baseline, baseline (w/ power iteration (piter)) and baseline (with full SVD) computation for iNaturalist 2019 dataset in table below. All the runs were done on NVIDIA RTX 3090 GPU on the same machine.  
\begin{table}[h]
\centering
\begin{tabular}{l|c}
\toprule
                  & Time (in secs) \\ \midrule
BigGAN            & 68             \\ 
BigGAN (w/ piter) & 77             \\ 
BigGAN (w/ SVD)   & 1126         \\ \bottomrule
\end{tabular}
\caption{Comparison of time taken for 100 updates of generator(G) on iNaturalist-2019 dataset. }
\label{gsr_tab:time_comp}
\end{table}

 As for each class seperate SVD computation is performed we find that the SVD computation becomes very expensive (Table~\ref{gsr_tab:time_comp}) for large datasets like iNaturalist-2019. Whereas as the power iteration can be done in parallel there is not much computation overhead with addition of each class. Hence, techniques like Group Whitening \cite{huang2021group} which use SVD are not a viable baseline for our case. It can be observed that despite having large number of classes in iNaturalist there is only addition of 9 sec, which shows the scalability and viability of proposed gSR. We provide a PyTorch implementation of cBN, detailing the process of spectral norm calculation as part of the supplemental material.

\subsection{Sanity Checks}
We build our experiments over the PyTorch-StudioGAN framework, which provides a simple framework over standard GAN architectures and setups. Since we are not using the official code for the LeCam Regularizer baseline \cite{tseng2021regularizing}, we first reproduce the BigGAN (+ LeCam + DiffAug) results on CIFAR-10 to ensure that our codebase is on par with the official codebase of the LeCam GAN. Our code obtains an FID of 7.59$_{\pm 0.04}$ vs. 8.31$_{\pm 0.03}$ reported in same setting by \textit{Tseng}~\etal~\cite{tseng2021regularizing}, which verifies the authenticity of our experiments. Hence, we compare our results to a stronger baseline which is due to improved implementation of BigGAN in the framework.

\begin{table}[!t]
\centering
\caption{Hyperparameter setups for all the reported experiments. $\alpha_D$, and $\alpha_G$ denote the learning rates for Discriminator and Generator respectively.}
\label{gsr_tab:hyper}
\resizebox{0.75\linewidth}{!}{
\begin{tabular}{@{}lllllll}
\toprule
\multirow{2}{*}{Setting} & Adam & \multirow{2}{*}{n$_{dis}$} & \multirow{2}{*}{$\lambda_{LC}$} & \multirow{2}{*}{G$_{EMA}$} & EMA & Total \\ 
& ($\alpha_D$, $\alpha_G$, $\beta_1$, $\beta_2$) & & & & Start & Iterations \\ \midrule
A      & 2e-4, 2e-4, 0.5, 0.9                                & 5         & 0.3            & False     &           & 120k             \\
B      & 2e-4, 2e-4, 0.5, 0.999                              & 5         & 0.1            & True      & 1k        & 120k             \\
C      & 2e-4, 2e-4, 0.5, 0.9                                & 5         & 0.3           & True      & 1k        & 200k             \\
D       & 2e-4, 2e-4, 0.0, 0.999                                & 2         & 0.01            & True     & 20k           & 120k             \\
E       & 2e-4, 2e-4, 0.5, 0.999                              & 5         & 0.01           & True      & 1k        & 120k             \\ 
F       & 4e-4, 1e-4, 0.5, 0.9                                & 5         & 0.5             & True     & 1k           & 120k             \\

\bottomrule
\end{tabular}}

\bigskip 

\resizebox{0.75\linewidth}{!}{
\begin{tabular}{l|c|c|c|c}
    \toprule
    & CIFAR-10 & LSUN & iNaturalist-19 & AnimalFace\\ \midrule
    LSGAN~\cite{mao2017least} & \multicolumn{2}{c|}{\multirow{3}{*}{A}} & \multicolumn{2}{c}{\multirow{3}{*}{---}} \\
    SNGAN~\cite{miyato2018spectral} & \multicolumn{2}{c|}{} & \multicolumn{2}{c}{}\\
    \; + gSR (Ours) & \multicolumn{2}{c|}{} & \multicolumn{2}{c}{} \\ \hline
    BigGAN~\cite{brock2018large} & \multirow{2}{*}{B}  & \multirow{2}{*}{C|F} & \multirow{2}{*}{D}  & \multirow{2}{*}{E}\\
     \; + gSR (Ours) &  &  & &\\ \bottomrule
    \end{tabular}}

\end{table}

\subsection{Hyperparameters}
We provide the details of the hyperparameters used in the experiments in Table \textcolor{red}{1} and \textcolor{red}{2} of the chapter in Table~\ref{gsr_tab:hyper}. For CBGAN~\cite{rangwani2021class} based experiments we follow the same setup as reported in the chapter (except using a ResNet~\cite{gulrajani2017improved} architecture for fairness in experiments). For BigGAN on LSUN dataset we use configuration C for the imbalance factor ($\rho$ = 100) and F for imbalance factor ($\rho$ = 1000). In our tuning experiments we explored the configurations in Table \ref{gsr_tab:hyper} and use the configuration which produces best FID for baseline. Then we add gSR regularizer to obtain our results. 

\noindent \vspace{1mm} \textbf{High-Resolution Experiments:} For the high resolution ($128 \times 128$) image synthesis on LSUN we find that we only require very small change in hyperparameters for obtaining results. For SNGAN, we use configuration A in Table~\ref{gsr_tab:hyper} with EMA starting at 1k along with $\lambda_{LC} = 0.5$. For the BigGAN we use the same configuration as in the Table \ref{gsr_tab:hyper}. We find that for higher resolutions a larger $\lambda_{LC}$ helps the purpose.

\subsection{Intuition about $n_c$ and $n_g$}
As we group the parameters $\gamma_{\mathbf{y}}^{\mathbf{l}}$ (Eq. {\color{red} 3} in chapter) to a matrix $\mathbf{\Gamma_{\mathbf{y}}^{\mathbf{l}}}$ of $n_c \times n_g$. The matrix can be decomposed into $\min(n_c,n_g)$ (matrix rank) number of independent and diverse components through SVD. As the scope of attaining maximal orthogonal and diverse components (matrix rank) is when $n_c \approx n_g$, it helps gSR to ensure maximal diversity and performance (as seen in chapter Table {\color{red} 5}). In case of gSR we find that almost all eigen values of $\mathbf{\Gamma_{\mathbf{y}}^{\mathbf{l}}}$ have a similar value, which demonstrates orthogonality and diversity. \\

\section{Analysis of gSR}

\vspace{1mm}\noindent\textbf{How much should be gSR's strength ($\lambda_{gSR}$)? }\begin{wrapfigure}{r}{0.4\textwidth}

    \includegraphics[width=0.4\textwidth]{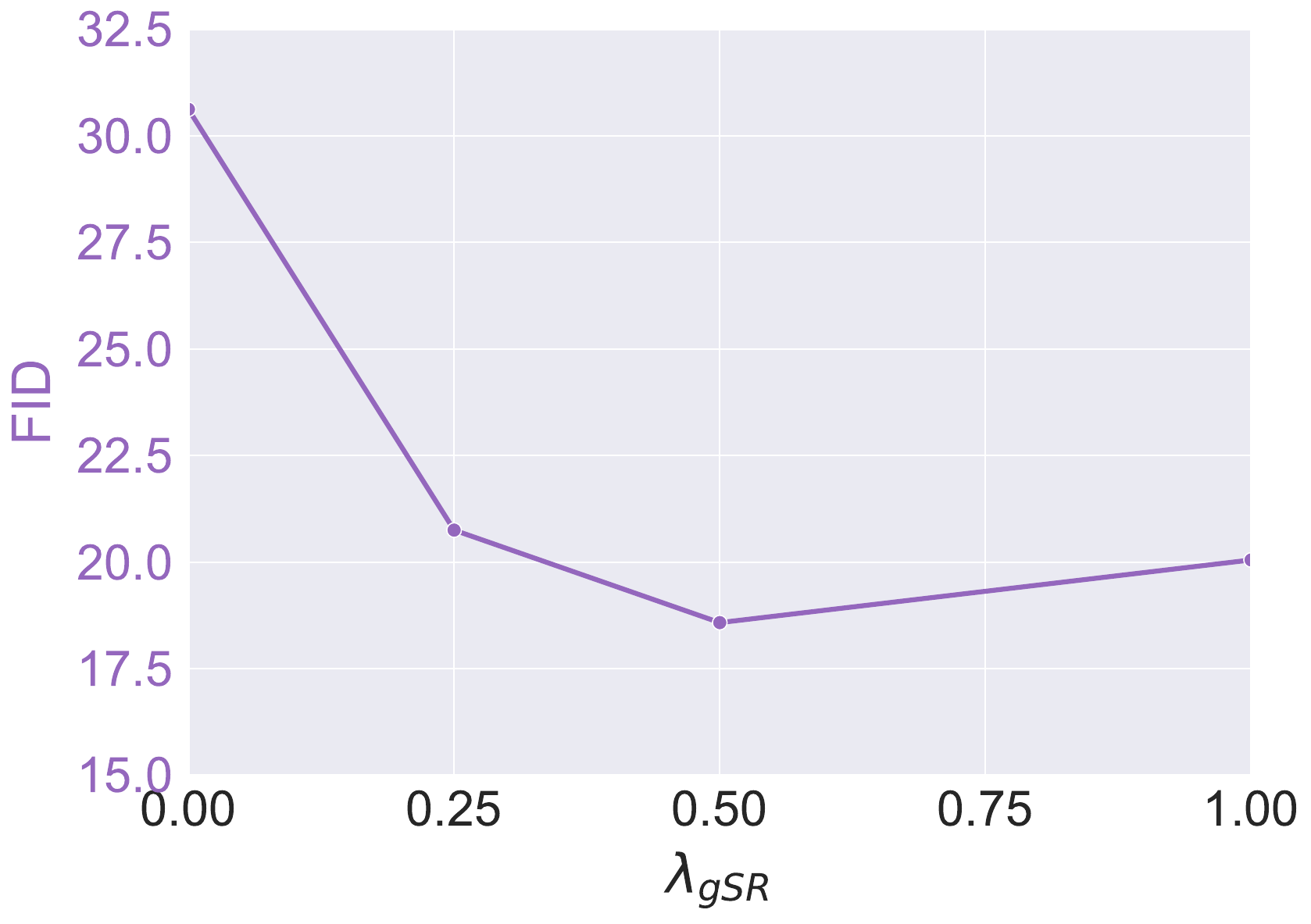}
    \caption{{Sensitivity to $\lambda_{gSR}$.}  On CIFAR-10, the FID marginally changes with $\lambda_{gSR}$ (0.25 to 1).}
    \label{gsr_fig:fid_lambda}
\end{wrapfigure}
\noindent\vspace{1mm} We experiment with different values of $\lambda_{gSR}$ for gSR in SNGAN as shown in Fig.~\ref{gsr_fig:fid_lambda}. $\lambda_{gSR}$ value of 0.5 attains best FID scores, hence we use it for all our experiments. 
 The value of FID changes marginally when $\lambda_{gSR}$ goes from 0.25 to 1 which highlights its robustness (\ie less sensitivity). 

\begin{figure}[t]
    \centering
    \includegraphics[width=0.8\textwidth]{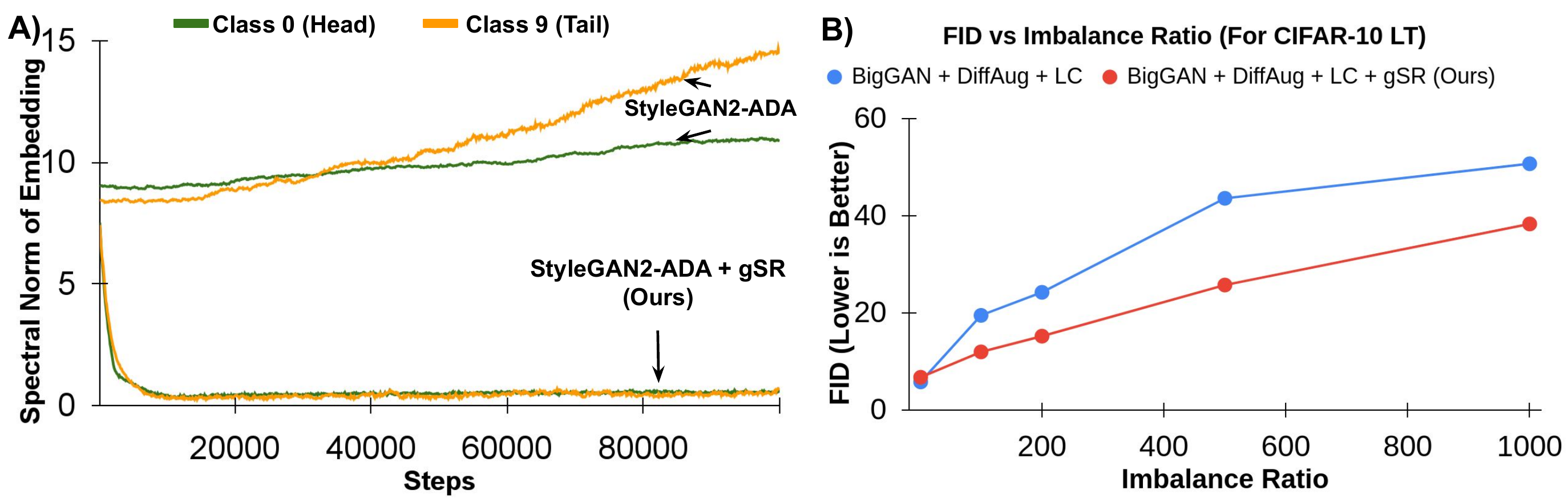}
    \caption{A) Spectral Norm comparison of StyleGAN2-ADA with and without gSR B) mean FID comparison of baseline with gSR across degrees of imbalance}
    \label{gsr_fig:spec_imb}
\end{figure}

\vspace{1mm}\noindent\textbf{How does performance of gSR change with degree of imbalance?}
Fig. \ref{gsr_fig:spec_imb}\textcolor{red}{.B} shows the comparison of mean FID of SOTA BigGAN (with DiffAug+LeCAM) and BigGAN (with gSR) in which we find that addition of \emph{gSR significantly improves performance across degrees of imbalance ratio}. Also, in the balanced case the performance with gSR is only slightly worse (by 0.95 FID) to the baseline.

\section{gSR for StyleGAN2}
 We train and analyze the spectral norm of class-conditional embeddings in StyleGAN2-ADA implementation available ~\cite{kang2020contrastive} on long-tailed datasets (CIFAR10 and LSUN), to find that it also suffers from spectral collapse of tail class embedding parameters (Fig.\ \ref{gsr_fig:spec_imb}\textcolor{red}{.A}) as BigGAN and SNGAN. 
  
\begin{figure}
    \centering
    \includegraphics[width=0.5\textwidth]{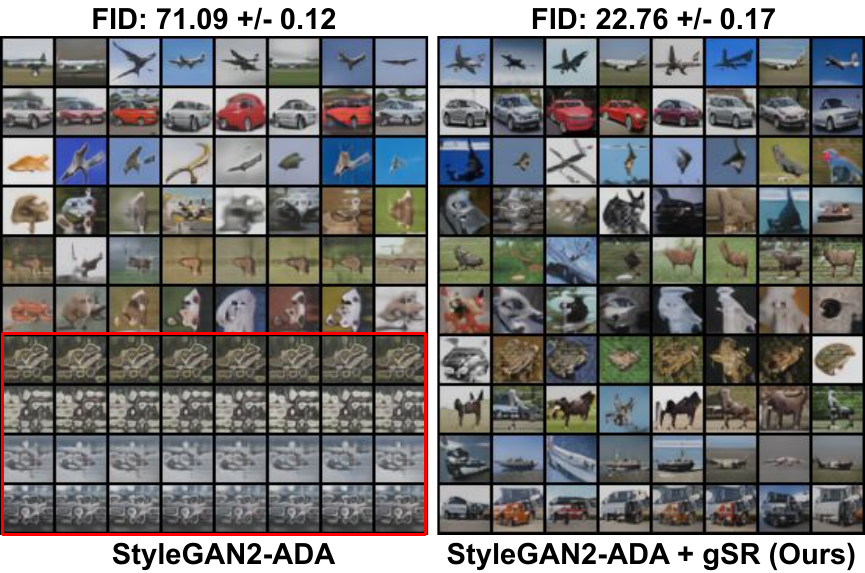}
    \caption{{StyleGAN2-ADA}  On CIFAR-10 ($\rho = 100$), comparison of gSR with the baseline.}
    \label{gsr_fig:style_gan_qual_results}
\end{figure}

 We then implement gSR for StyleGAN2 generator by grouping 512 dimensional class conditional embeddings to 16x32 and calculating their spectral norm  which is added to loss (Eq.\ \textcolor{red}{5}) as $R_{gSR}$. We find that gSR is able to effectively prevent the mode collapse (Fig.\ \ref{gsr_fig:style_gan_qual_results}) and also results in significant improvement in FID (Tab.~\ref{gsr_tab:quant_sg2}) in comparison to StyleGAN2-ADA baseline.

 \begin{table}[t]
    \centering
    \caption{Quantitative comparison of  gSR over StyleGAN2-ADA baseline.}
    \label{gsr_tab:quant_sg2}
    \begin{tabular}{lcccc} \hline
     & \multicolumn{2}{c}{CIFAR10-LT ($\rho = 100$)} & \multicolumn{2}{c}{LSUN-LT ($\rho = 100$)} \\ \hline
    & FID-10k ~\cite{zhao2020differentiable}  ($\downarrow$)&      IS($\uparrow$)& FID-10k ~\cite{zhao2020differentiable} ($\downarrow$)&      IS($\uparrow$)  \\\hline
  StyleGAN2-ADA & 71.09$_{\pm0.12}$ & 5.66$_{\pm0.03}$ & 55.04$_{\pm0.07}$ & 3.92$_{\pm0.02}$ \\
  +gSR(Ours) & \textbf{22.76}$_{\pm0.17}$ & \textbf{7.55}$_{\pm0.01}$ & \textbf{27.85}$_{\pm0.06}$ & \textbf{4.32}$_{\pm0.01}$ \\ \hline
  \end{tabular}
\end{table}

\clearpage

 \renewcommand{\thesection}{C}

\supersection{Class-Consistent and Diverse Image Generation through StyleGANs (Chapter-4)}

\section{Notations and Code}
We summarize the notations used throughout the chapter in Table \ref{nt_tab:notations}.  

\begin{table}[!b]
\centering
\caption{{Notation Table}}
\label{nt_tab:notations}
\resizebox{\linewidth}{!}{%
\begin{tabular}{p{0.17\linewidth} p{0.15
\linewidth} p{0.68\linewidth}}
\toprule
Symbol                & Space                              & Meaning                                                                               \\ \midrule
$\mb{c}$                     & $\mathbb{R}^{d}$                    & Class Embedding                                                                           \\
$\mb{z}$            & $\mathbb{R}^{d}$                          & Noise vector                                                                          \\
$\mathbf{w}$            & $\mathbb{R}^{d}$                   & Vector in $\mc{W}$ latent Space   \\
$\mc{D}$                     &                                    & Discriminator    \\
$\mc{G}$                     &                                    & Generator      \\
$BS$                    &  $\mathbb{R}^+$                      & Batch Size        \\
$\mathbf{x_i}$          & $\mathbb{R}^{3 \times H \times W}$ & Image  \\
$\mb{\tilde{c}}$          & $\mathbb{R}^{d}$                   & Noise Augmented Class Embedding     \\
$n_c$                    & $\mathbb{R}^{+}$                   & Frequency of training samples in class $\mb{c}$      \\
$\sigma_c$                & $\mathbb{R}^{+}$                   & Effective number of samples based noise standard deviation     \\
$\sigma$                & $\mathbb{R}^+$                        & Hyperparameter for scaling noise      \\
$\mb{\mu_{c}}$      & $\mathbb{R}^d$                            & Mean embedding parameters of class $\mb{c}$      \\
$\tilde{\mb{W}}_A$ $\tilde{\mb{W}}_B$ & $\mathbb{R}^{BS \times d} $          & Batches of augmented latents      \\
$\mb{C}_{j,k}$     &    $\mathbb{R}$                            & Cross-correlation between $j$th and $k$th latent variables        \\
$\lambda$          & $\mathbb{R}^{+}$                           & Strength of NoisyTwins regularization    \\
$\gamma$            & $\mathbb{R}^{+}$                          & Relative importance of the two terms of NoisyTwins loss     \\
$\rho$                  & $\mathbb{R}^{+}$                       & Imbalance ratio of dataset: Ratio between the most and the least frequent classes\\ \bottomrule
\end{tabular}}
\end{table}

\begin{figure}[h]
    \centering
    \includegraphics[width=\linewidth]{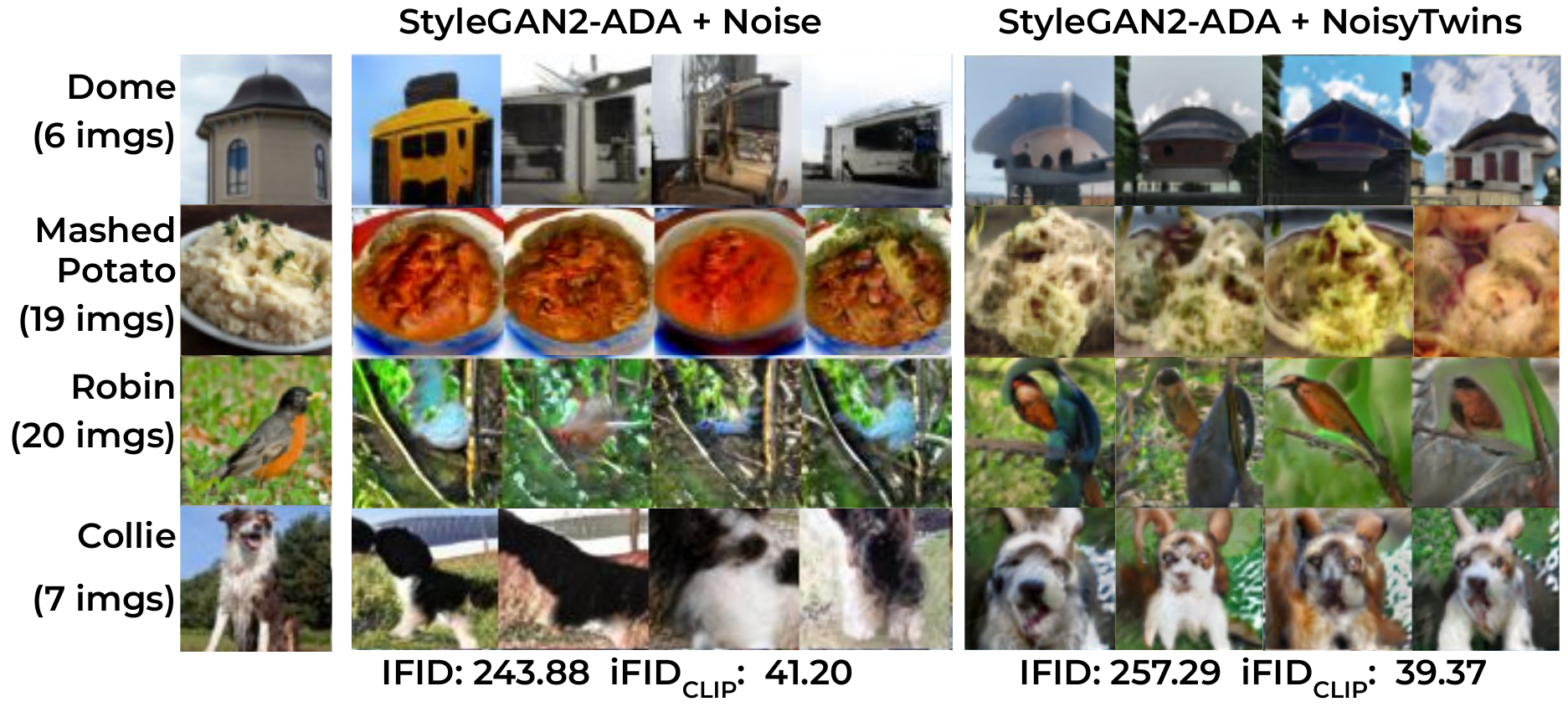}
    \caption{{Qualitative Results and iFID.} We observe that the noise-only baseline suffers from the mode collapse and class confusion for tail categories as shown on (\emph{left}). Despite this, it is found that the mean iFID based on Inception V3 shows a smaller value for StyleGAN2ADA+Noise, whereas a higher value for diverse and class-consistent NoisyTwins. Hence, this metric does not align with qualitative results. On the other hand, the proposed mean iFID$_\mathrm{CLIP}$ is lower for NoisyTwins, demonstrating its reliability for evaluating GAN models.}
    \label{nt_fig:mean_ifid_imagenet}
\end{figure}

\section{Comparison of iFID and \texorpdfstring{iFID$_\mathrm{CLIP}$}{iFID-CLIP}}

\begin{figure*}[t]
    \centering
    \includegraphics[width=\linewidth]{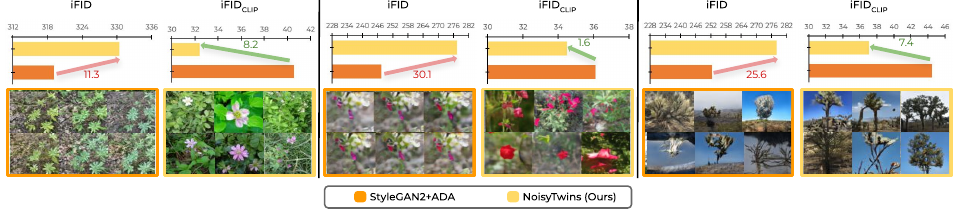}
    \caption{{iFID Comparison on iNaturalist 2019 dataset.} We provide examples of classes where the quality of images generated by StyleGAN2-ADA is worse, which either suffers from mode collapse or artifacts in generation. Yet iFID based on Inception V3 ranks it higher in terms of quality, which doesn't align with human judgement. On the other hand the proposed iFID$_{\mathrm{CLIP}}$ is able to rank the models correctly and gives a lower value to diverse generations from NoisyTwins.}
    \label{nt_fig:iFID Comparison_3_class}
\end{figure*}

In this section, we present failure cases of InceptionV3-based iFID in the detection of mode collapse, and show how CLIP-based iFID can detect these cases. InceptionV3-based iFID assigns a lower value to a generator with mode collapse, compared to another generator which creates diverse and class-consistent images. In addition to the example given in the main text (Fig. \textcolor{red}{5}), we provide examples from three different classes (Fig.~\ref{nt_fig:iFID Comparison_3_class}). In all the four cases, the InceptionV3-based iFID is better for mode collapsed classes. \emph{Whereas  iFID$_\mathrm{CLIP}$ follows the correct behavior, where the class consistent and diverse model is ranked better}. Due to this inconsistent behavior, mean iFID (mean across classes) which is a commonly used as a metric for quantifying class confusion~\cite{kang2021ReACGAN} can be incorrect. %

 For example, we observe that the StyleGAN2-ADA baseline with proposed noise augmentation achieves mean iFID (243.88) on ImageNet-LT, compared to 257.29 for the NoisyTwins model (Table \textcolor{red}{1} in main text). However, while examining the tail class samples (Fig. \ref{nt_fig:mean_ifid_imagenet}), we find that noise augmented baseline suffers from mode collapse and class confusion, whereas NoisyTwins generates diverse and class-consistent images. Hence, the mean iFID based on Inception-V3 does not align well with qualitative results. On the contrary, the iFID$_\mathrm{CLIP}$ value is 41.20 for the noise-augmented model compared to 39.37 for NoisyTwins, which correlates with the human observation that the NoisyTwins model should have a lower FID as it is diverse and class-consistent. Hence, the proposed metric iFID$_\mathrm{CLIP}$ can be used to to evaluate models for class-conditional image generation reliably.\

\section{Experimental Details}

\begin{table*}[!h]
    \parbox{\textwidth}{
        \centering
        \caption{{HyperParameter Configurations used for experiments.} We provide a detailed list of hyperparameters used for the experiments across datasets for NoisyTwins on StyleGANs.}
        \label{nt_tab:hyperparameters}}
        \resizebox{\textwidth}{!}
        {
        \begin{tabular}{r|c|c|c||c|c}
        \toprule
            \multicolumn{1}{c}{}& \multicolumn{3}{c}{Long-Tail Datasets} & \multicolumn{2}{c}{Few-Shot Datasets} \\
            \midrule
              & iNaturalist-2019 & ImageNet-LT & CIFAR10-LT ($\rho$=100) & ImageNet Carnivores & AnimalFaces  \\
            \midrule
            Resolution & 64 & 64 & 32 & 64 & 64 \\
            Augmentation & ADA & ADA & DiffAug & ADA & ADA \\
            \midrule 
             \multicolumn{1}{c}{}& \multicolumn{5}{c}{Regularizers}\\
            \midrule
            Effective Samples $\alpha$ & 0 & 0 & 0.99 & 0 & 0  \\
            Noise Scaling $\sigma$ & 0.1 & 0.25 & 0.75 & 0.5 & 0.5 \\
            NoisyTwins Start Iter. & 25k & 60k & 0 &  0 & 0 \\
            NoisyTwins Weights ($\lambda$, $\gamma$) & 0.001, 0.005 & 0.001, 0.005 & 0.01, 0.05 & 0.001, 0.05 & 0.001, 0.05 \\
            LeCam Reg Weight & 0.01 & 0 & 0 & 0 & 0 \\ 
            R1 Regularization $\gamma_{R1}$ & 0.2048 & 0.2048 & 0.01 & 0.01 & 0.01 \\
            PLR Start Iter. & 0 & 60k & No PLR & 0 & 0 \\
            \midrule 
            \multicolumn{1}{c}{} & \multicolumn{5}{c}{StyleGAN}\\
            \midrule
            Mapping Net Layers & 2 & 8 & 8 & 2 & 2 \\
            $\mc{D}$ Backbone & ResNet & ResNet & Orig & ResNet & ResNet \\
            Style Mixing & 0.9 & 0.9 & 0 & 0 & 0\\
            $\mc{G}$ EMA Rampup & None & None & 0.05 & 0.05 & 0.05\\
            $\mc{G}$ EMA Kimg & 20 & 20 & 500 & 500 & 500\\
            MiniBatch Group & 8 & 8 & 32 & 32 & 32 \\
            
            \bottomrule
        \end{tabular}
        }
\end{table*}

\begin{table*}[!t]
    \centering
    \parbox{\textwidth}{
    \caption{
    {Statistical Analysis for CIFAR10-LT.} This table provides the mean and one standard deviation of metrics for all methods on CIFAR10-LT performed on three independent evaluation runs by generating 50k samples across random seeds. 
    }
        \label{nt_tab:CIFAR10_LT Std Table}}
    \resizebox{0.9\textwidth}{!}{
    \begin{tabular}{lccccc}
    \toprule
         & \multicolumn{5}{c}{CIFAR10-LT ($\rho$=100)} \\ \hline
         Method & FID($\downarrow$) & FID$_\mathrm{CLIP}$($\downarrow$) & iFID$_\mathrm{CLIP}$($\downarrow$) & Precision($\uparrow$) & Recall($\uparrow$)\\
         \midrule
         SG2+DiffAug~\cite{Karras2020ada}& 31.72$_{\pm{0.16}}$ & 6.24$_{\pm{0.02}}$ & 11.63$_{\pm{0.03}}$ & 0.63$_{\pm{0.00}}$ & 0.35$_{\pm{0.00}}$ \\
         SG2+D2D-CE~\cite{kang2021ReACGAN}& 20.08$_{\pm{0.15}}$ & 4.75$_{\pm{0.04}}$ & 11.35$_{\pm{0.01}}$ & \textbf{0.73}$_{\pm{0.00}}$ & 0.43$_{\pm{0.00}}$ \\
         gSR~\cite{rangwani2022gsr}& 22.50$_{\pm{0.29}}$ & 5.55$_{\pm{0.01}}$ & 9.94$_{\pm{0.00}}$ & 0.70$_{\pm{0.00}}$ & 0.28$_{\pm{0.01}}$ \\
         \midrule
         
        \rowcolor{gray!10}  SG2+DiffAug+Noise (Ours)& 28.85$_{\pm{0.18}}$ & 5.29$_{\pm{0.02}}$ & 10.64$_{\pm{0.01}}$ & 0.71$_{\pm{0.00}}$ & 0.38$_{\pm{0.00}}$ \\
        \rowcolor{gray!10} \; + NoisyTwins (Ours)& \textbf{17.72}$_{\pm{0.08}}$ & \textbf{3.56}$_{\pm{0.01}}$ & \textbf{7.27}$_{\pm{0.02}}$ & 0.69$_{\pm{0.01}}$ & \textbf{0.52} $_{\pm{0.01}}$ \\ \bottomrule
    \end{tabular}
    }
    
\end{table*}

\begin{figure}[!t]
    \centering
    \includegraphics[width=0.6\linewidth]{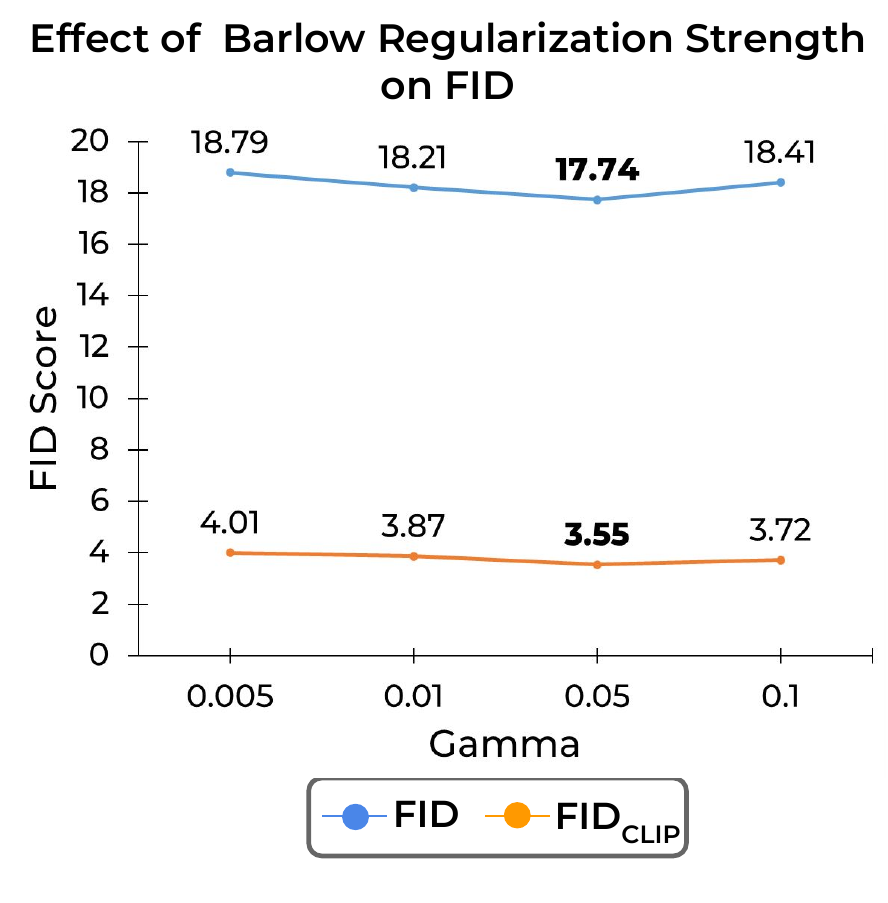}
    \caption{{Ablation on \texorpdfstring{$\gamma$}{Gamma}:} Quantitative comparison on CIFAR10-LT for the strength of hyperparameter (\texorpdfstring{$\gamma$}{Gamma}) in NoisyTwins loss function.}
    \label{nt_fig:Barlow Momentum Ablation}
    \vspace{-5mm}
\end{figure}

We run our experiments using PyTorchStudioGAN~\cite{kang2022StudioGAN} as the base framework. For most baseline experiments, we use the standard StyleGAN configurations present in the framework. We use a learning rate of 0.0025 for the discriminator ($\mc{D}$) and the generator ($\mc{G}$) network. We use a batch size of 128 for all our experiments. In addition, following the observations of previous work~\cite{Sauer2021ARXIV}, we apply a delayed Path Length Regularization (PLR) starting at 60k iterations for all our experiments on ImageNet-LT. For NoisyTwins, the most important hyperparameters are $\lambda$ (regularization strength) and $\sigma$ (noise variance). We perform a grid search on $\lambda$ values of \{0, 0.001, 0.01, 0.1\} and $\sigma$ values of \{0.10, 0.25, 0.50, 0.75\}.  We provide a detailed list of optimal hyperparameters used in Table \ref{nt_tab:hyperparameters}. All the models trained on a particular dataset use the same hyperparameters, to maintain fairness in the comparison of models. We summarize all the hyperparameters used for respective datasets in Table~\ref{nt_tab:hyperparameters}.%

For our experiments on few-shot datasets with SotA transitional-cGAN, we use the author's official code implementation available on GitHub~\footnote{https://github.com/mshahbazi72/transitional-cGAN}. We use the same configuration specified to first evaluate on ImageNet Carnivores and AnimalFaces datasets. To integrate NoisyTwins, we generate the noise augmentations by augmenting the class embeddings and then apply NoisyTwins regularization in $\mc{W}$ space. We use the same hyperparameter setting used by the authors and NoisyTwins with $\lambda = 0.001$ and $\gamma = 0.05$.

\subsection{Statistical Significance of the Experiments}

We report mean and standard deviation over three evaluation runs for all baselines on the CIFAR10-LT (Table~\ref{nt_tab:CIFAR10_LT Std Table}). It can be observed that most metrics that we have reported have a low standard deviation, and metrics are close to the mean value across runs. As we find standard deviation to be low across the metrics evaluated and the process of evaluating iFID to be expensive, we do not explicitly report them on large multi-class datasets. 

\begin{table*}[!t]
    \centering
    \parbox{\textwidth}{
    \caption{{Evaluation of NoisyTwins by varying degree of imbalance.} NoisyTwins can produce diverse and class-consistent results across imbalance ratios. 
    }
        \label{nt_tab:CIFAR10-LT imb factor ablation}}
    \resizebox{0.9\textwidth}{!}{
    \begin{tabular}{lcccccc}
    \toprule
         & \multicolumn{5}{c}{CIFAR10-LT} \\ \hline
         Method & $\rho$ & FID($\downarrow$) & FID$_\mathrm{CLIP}$($\downarrow$) & iFID$_\mathrm{CLIP}$($\downarrow$) & Precision($\uparrow$) & Recall($\uparrow$)\\
         \midrule
         SG2+DiffAug~\cite{Karras2020ada}& \multirow{2}{*}{50} & 26.79 & 5.83 & 9.61 & 0.65 & 0.38\\
        \;+NoisyTwins (Ours) & & \textbf{14.92} & \textbf{2.99} & \textbf{6.38} & \textbf{0.71} & \textbf{0.57}\\ 
        \midrule
        SG2+DiffAug~\cite{Karras2020ada}& \multirow{2}{*}{100} & 31.73 & 6.27 & 11.59 & 0.63 & 0.35  \\
        \;+NoisyTwins (Ours)& & \textbf{17.74} & \textbf{3.55} & \textbf{7.24} & \textbf{0.70} & \textbf{0.51}\\ 
        \midrule
        SG2+DiffAug~\cite{Karras2020ada}& \multirow{2}{*}{200} & 55.48 & 10.59 & 19.49 & 0.65 & 0.36 \\
        \;+NoisyTwins (Ours)& & \textbf{23.57} & \textbf{4.91} & \textbf{9.17} & \textbf{0.68} & \textbf{0.46}\\ 
        \bottomrule
    \end{tabular}
    }
    
\end{table*}

\begin{figure}[!t]
    \centering
    \includegraphics[width=0.75\linewidth]{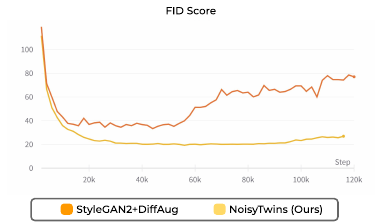}
    \caption{{Comparison of FID curves for CIFAR10-LT ({\boldmath$\rho$}=100).} NoisyTwins leads to stable training with decreasing FID with iterations.}
    \label{nt_fig:FID Comparison}
    \vspace{-4mm}
\end{figure}

\section{Additional Details of Analysis}
We perform our ablation experiments on CIFAR10-LT using the same configuration as mentioned in Table~\ref{nt_tab:hyperparameters}. We provide ablation experiments on the standard deviation of noise ($\sigma$) and the strength of regularization loss ($\lambda$) (Sec. \textcolor{red}{6}), as we observe that they influence the performance of the system most. We further provide ablation on the parameter $\gamma$ in Fig. \ref{nt_fig:Barlow Momentum Ablation}, which controls the relative importance between the invariance enforcement and decorrelation enhancement terms in Eq. \textcolor{red}{6} of the main text. We find that performance remains almost the same while varying $\gamma$ from 0.005 to 0.1, with optimal value occurring around 0.05 for CIFAR10-LT. Hence, the model is robust to $\gamma$. 

We further analyze our method for a range of imbalance ratios (i.e., $\rho$, ratio of the most frequent to least frequent class) in the class distribution. We present results for CIFAR10-LT with imbalance factors ($\rho$) values of 50, 100, and 200 in Table \ref{nt_tab:CIFAR10-LT imb factor ablation}. Our method can prevent mode collapse and improves the baseline FID significantly in all cases. Also note that the baseline gets more unstable (high FID) as the imbalance ratio increases, which shows the necessity of using NoisyTwins as it stabilizes the training even when large imbalances are present in the dataset (Fig. \ref{nt_fig:FID Comparison}).

\section{Additional Results}
\begin{figure}[!t]
    \centering
    \includegraphics[width=0.75\linewidth]{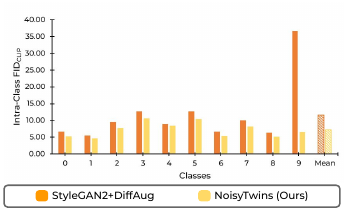}
    \caption{{Class-wise iFID$_{\textbf{CLIP}}$} comparison of models on CIFAR10-LT (\boldmath$\rho$=100) dataset.}
    \label{nt_fig:iFID Class-Wise Comparison}
        \vspace{-4mm}
\end{figure}

Fig. \ref{nt_fig:iFID Class-Wise Comparison}  provides the class-wise comparison of the proposed iFID$_\mathrm{CLIP}$ for the baseline and after adding NoisyTwins. NoisyTwins produces better iFID$_\mathrm{CLIP}$ for all classes, hence does not lead to performance degradation for head classes while improving performance on tail classes.
      
\begin{figure*}[!t]
    \centering
    \includegraphics[width=\linewidth]{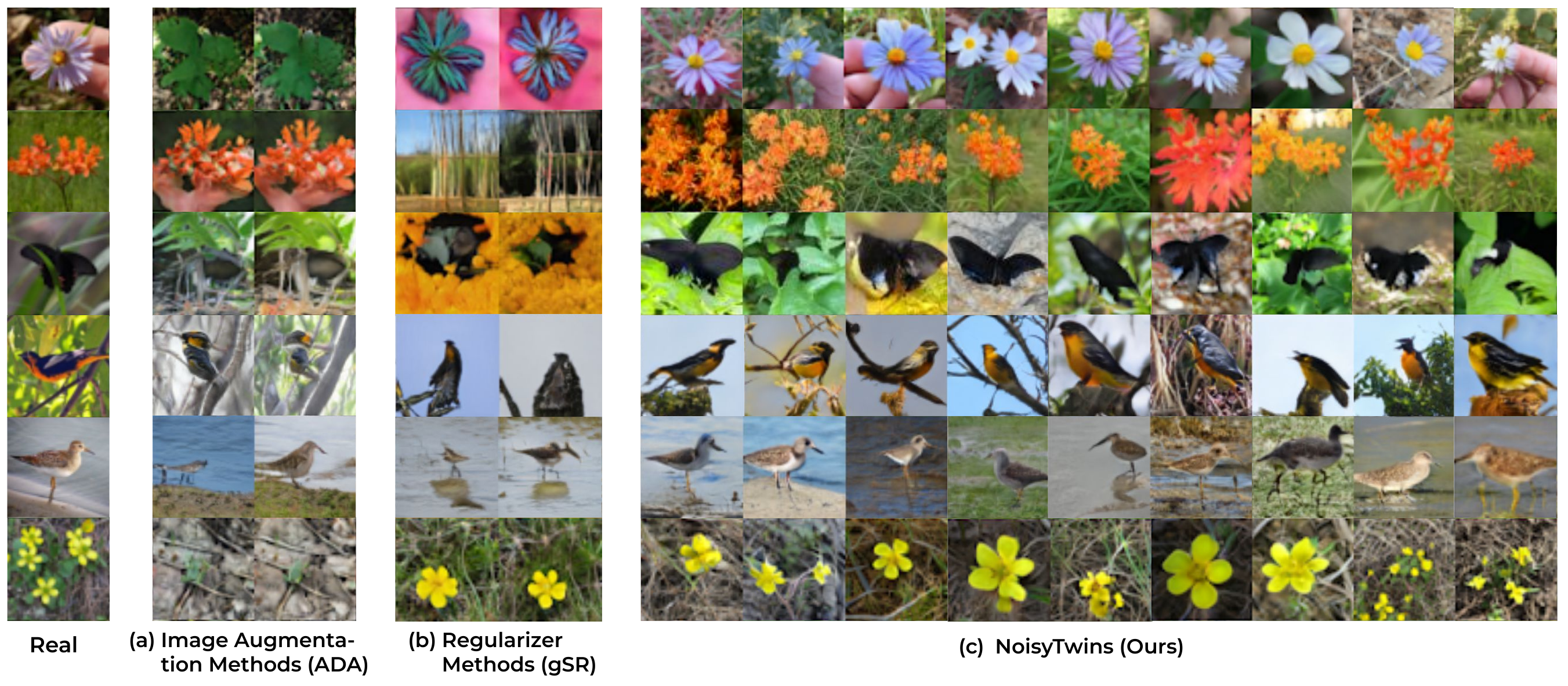}
    \caption{{Qualitative Analysis on iNaturalist2019 (1010 classes).} Examples of generations from various classes for evaluated baselines (Table \textcolor{red}{1}). The baseline ADA suffers from mode collapse, whereas gSR suffers from class confusion particularly for tail classes, particularly for tail classes as seen above on the left. NoisyTwins  generates diverse and class-consistent images across all categories.}
    \label{nt_fig:iNat_qualitative}
\end{figure*}

We now provide additional qualitative results for models. Similar to ImageNet-LT, we also provide a full-scale comparison of images from different methods in Fig. \ref{nt_fig:iNat_qualitative} for iNaturalist-2019. In addition to the images from the tail classes, we also show generations from the head and middle classes. In Fig. \ref{nt_fig:iNat_qualitative}, it is clearly shown that NoisyTwins can obtain high-quality and diverse samples compared to the baseline. We find that the StyleGAN2-ADA baseline produces similar images across a class for tail classes, which confirms the occurrence of class-wise mode collapse even in large datasets. Further, it can be seen that the regularizer-based method (gSR) is unable to capture the identity of the real class and suffers from the issue of class confusion (as also seen in t-SNE of Fig. \textcolor{red}{2} of the main text). Our method NoisyTwins, can produce realistic-looking diverse images even for tail classes, which shows the successful transfer of knowledge from head classes. Training a class-conditioned GAN on long-tailed datasets leads to class confusion when the extent of knowledge transfer is not controlled. NoisyTwins strikes the right balance between knowledge transfer from the head classes to benefit the quality of generation in the tail classes, thus not allowing class confusion. This would not be possible if we train GAN independently on tail classes ($\sim$ 30 images), which shows the practical usefulness of joint training on complete long-tailed data (i.e., our setup).

\begin{figure*}[h]
    \centering
    \includegraphics[width=\linewidth]{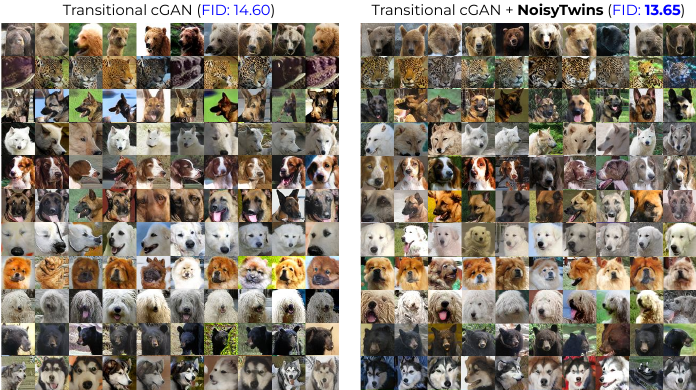}
    \caption{{Qualitative comparison on few-shot ImageNet Carnivores dataset.}}
    \label{nt_fig:in_carnivore}
\end{figure*}
\begin{figure*}[h]
    \centering
    \includegraphics[width=\linewidth]{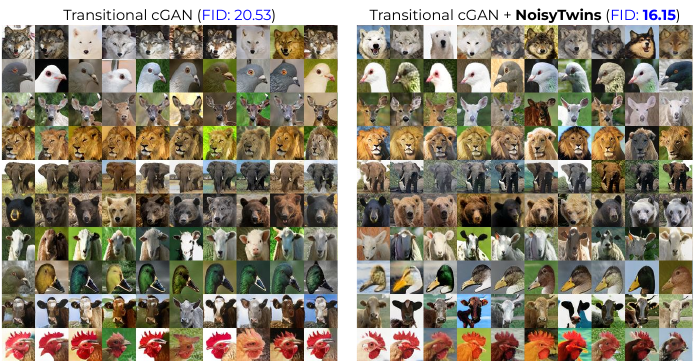}
    \caption{{Qualitative comparison on few-shot AnimalFaces dataset.}}
    \label{nt_fig:animalface}
\end{figure*}
\vspace{10mm}

We showcase qualitative results of generations from few-shot datasets (i.e., ImageNet Carnivore and AnimalFaces). Fig. \ref{nt_fig:in_carnivore} and \ref{nt_fig:animalface} show the results of the SotA few-shot baseline of Transitional-cGAN  (\textit{left}) and after augmenting it with our proposed NoisyTwins (\textit{right}). Our proposed method, NoisyTwins, can further stabilize the training of Transitional-cGAN and improve the quality and diversity of the generated samples on both datasets of ImageNet Carnivores and AnimalFaces.

\begin{table}[!t]
    \centering
   
    \caption{Results for Large Resolutions on Animal Faces dataset}
     \label{nt_tab:AF-high-res}

    \begin{adjustbox}{max width=\columnwidth}
    \begin{tabular}{l|ccc}
        \hline
     { FID ($\downarrow$)} & AF (128 $\times$ 128) & AF (256 $\times$ 256) \\ %
    \hline

    \textbf{Transitional-cGAN}~\cite{shahbazi2022collapse}  & 22.59 & 22.28 \\  %
    
     \textbf{+NoisyTwins (Ours)}& \textbf{16.79} & \textbf{19.14} \\ %

     \hline\hline

    \end{tabular}

\end{adjustbox}
\end{table}

\begin{table}[!t]
    \centering

     \caption{Results for large iNaturalist 2019 dataset (128 $\times$ 128)}
     \label{nt_tab:inat-128}
    \begin{adjustbox}{max width=\columnwidth}
    \begin{tabular}{l|ccc}
    \hline
         &  FID (80k) ($\downarrow$) & FID ($\downarrow$) & FID$_\mathrm{CLIP}$ ($\downarrow$)  \\ %
    \hline
    \textbf{StyleGAN2-ADA}~\cite{Karras2020ada}  & 16.58 & 12.31 & 2.18 \\  %
    
     \textbf{+NoisyTwins (Ours)} & \textbf{15.29} & \textbf{12.01} &  \textbf{1.93}\\ %
     \hline \hline
    
    \end{tabular}
 \end{adjustbox}
\end{table}
\vspace{1mm} \noindent \textbf{Results across other Resolutions:} NoisyTwins scales well on larger resolutions as demonstrated on few-shot AnimalFaces (AF) dataset using Transitional-cGAN~\cite{shahbazi2022collapse} in Table~\ref{nt_tab:AF-high-res}, where we observe a significant improvement if FID for both $128 \times 128$ and $64 \times 64$ resolution data. Further, on large-scale iNat-19 StyleGAN2-ADA baseline in Tab. ~\ref{nt_tab:inat-128}, we also find that NoisyTwins is able to improve performance. The NoisyTwins method also converges faster as at intermediate stage of 80k iterations in full run of 150k iterations, the FID for NoisyTwins is lower than baseline.  As NoisyTwins method is based on the information maximization principle~\cite{zbontar2021barlow} and generalizes on datasets, we expect it benefits other large resolutions of StyleGAN too, similar to what is observed in Sauer \etal ~\cite{Sauer2021ARXIV}.

\newpage

 \renewcommand{\thesection}{D}

\supersection{Escaping Saddle Points for Effective Generalization on Class-Imbalanced Data
(Chapter-5)}

\section{Limitations of Our Work} 
We would like to highlight that our theoretical results are based on \citet{daneshmand2018escaping} which verified CNC condition for small scale neural networks, verifying the CNC condition for large networks and exactly characterizing the saddle point solutions obtained by SAM for minority classes, are good directions for future work. 

Also empirically, we propose to use Sharpness-Aware Minimization with high $\rho$ for tail classes to escape from saddle points. Although the general guideline is to use a higher $\rho$ value like 0.5 or 0.8 to achieve the best result, we do find that $\rho$ as a hyperparameter still requires tuning to obtain the best results. We believe making SAM hyper-parameter free is an interesting direction to pursue in the future.

\section{Proof of Theorem}
\label{saddlesam_saddlesam_app:proof_theorem}
In this section, we re-state Theorem \textcolor{red}{2} and provide it's proof. The theorem analyzes the variance of stochastic gradient for SAM along the direction of negative curvature and shows that SAM amplifies the variance by a factor, which signals that it has a stronger component in direction of negative curvature under certain conditions. Hence, SAM can be used for effectively escaping saddle points in the loss landscape. This is based on Correlated Negative Curvature (CNC) Assumption for stochastic gradients (Assumption \textcolor{red}{1}). The 
$\mathbf{v_{w}}, {\nabla f(w)} \in \mathbb{R}^{p \times 1}$ whereas the Hessian denoted by $H(f(w)) \; (\text{also} \nabla^2 f(w)) \in   \mathbb{R}^{p \times p}$ where $p$ is the number of parameters in the model.
\begin{theorem}
Let $\mathbf{v_{w}}$ be the minimum eigenvector corresponding to the minimum eigenvalue $\lambda_{\min}$ of the Hessian matrix $\nabla^2 f(w)$. The $\nabla f_{\bm{z}}^{\text{SAM}}({{w}})$ satisfies that it's second moment of projection in ${v_{w}}$ is atleast $(1 + \rho\lambda_{min})^2$ times the original (component of $\nabla f_{\bm{z}}({{w}})$):

\begin{equation}
    \exists \; \gamma \geq 0 \; s.t. \; \forall w : \mathbf{E}[<\mathbf{v_w}, \nabla f_{\bm{z}}^{\text{SAM}}({{w}})>^2] \geq (1 + \rho \lambda_{min})^2\gamma
\end{equation}
\end{theorem}
\begin{proof}

Using the first-order approximation of a vector valued function through Taylor series:
\begin{equation}
    f(w + \epsilon) = f(w) + J(\nabla f(w))\epsilon
\end{equation}
here $J$ is the jacobian operator. After considering $\rho$ to be small we have the following approximation for the SAM gradient:
\begin{align} 
        \nabla f^{\text{SAM}}({\bm{w}}) &= \nabla f(w + \rho \nabla f(w)) \\
        & = \nabla f(w) + \rho  H(f(w)) \nabla f(w) 
\end{align}
Here, we have used the following property that  $J(\nabla f(w))$ is the Hessian matrix $H(f(w))$ (also written as $\nabla^2 f(w)$). Also, as we now want to work with stochastic gradients, we replace gradient $\nabla f(w)$ with it's stochastic version $\nabla f_{z}(w)$ and introduce an expectation expression.
Now, we analyze the second-moment of the SAM gradient along the direction of most negative curvature $\bm{v_w}$:
\begin{align*}
    \bm{E}[< \mathbf{v_w}, \nabla f_{\bm{z}}^{\text{SAM}}({{w}})> ^2] &=    \bm{E}[<\mathbf{v_w}, \nabla f_{z} (w) + \rho  H(f(w))\nabla f_{z} (w))>^2] \\
    &= \bm{E}[(<\mathbf{v_w}, \nabla f_{z} (w)> + \rho <\mathbf{v_w}, H(f(w))\nabla f_{z} (w)>)^2] \\
    &= \bm{E}[(<\mathbf{v_w}, \nabla f_{z} (w)> + \rho \mathbf{v_w}^{T}H(f(w))\nabla f_{z} (w))^2] \\
\end{align*}
Here, we use the matrix notation for dot product $<x,y> = x^{T}y$. Using the property of the eigen vector: $\mathbf{v^{T}_w}H(f(w)) = \lambda_{min} \mathbf{v^{T}_w}$, we substitute the value below:
\begin{align*}
    \bm{E}[< \mathbf{v_w}, \nabla f_{\bm{z}}^{\text{SAM}}(w)>^2] &= \bm{E}[(<\mathbf{v_w}, \nabla f_{z} (w)> + \rho \lambda_{min} \mathbf{v^{T}_w}\nabla f_{z} (w) >)^2] \\
    &= \bm{E}[(<\mathbf{v_w}, \nabla f_{z} (w)> + \rho \lambda_{min} <\mathbf{v_w}, \nabla f_{z} (w)>)^2] \\
    &= \bm{E}[ ((1 + \rho\lambda_{min})<\mathbf{v_w}, \nabla f_{z} (w)>)^2] \\
    &= (1 + \rho\lambda_{min})^2 \bm{E}[<\mathbf{v_w}, \nabla f_{z} (w)>^2] \\
    & \geq (1 + \rho\lambda_{min})^2 \gamma \label{saddlesam_eq:cnc_ass}
\end{align*}
The last step follows from the CNC Assumption \textcolor{red}{1}. This completes the proof.
\end{proof}
\section{Experimental Details}
\label{saddlesam_app:experimental_details}

\textbf{Imbalanced CIFAR-10 and CIFAR-100}:
For the long-tailed imbalance (CIFAR-10 LT and CIFAR-100 LT), the sample size across classes decays exponentially with $\beta$ = 100. CIFAR-10 LT holds 5000 samples in the most frequent class and 50 in the least, whereas CIFAR-100 LT decays from 500 samples in the most frequent class to 5 in the least. The classes are divided into three subcategories: \textit{Head }(Many),\textit{ Mid} (Medium), and \textit{Tail} (Few). For CIFAR-10 LT, the first 3 classes (> 1500 images each) fall into the head classes, following 4 classes (> 250 images each) into the mid classes, and the final 3 classes (< 250 images each) into the tail classes. Whereas for CIFAR-100 LT, head classes consist of the initial 36 classes, mid classes contain the following 35 classes, and the tail classes consist of the remaining 29 classes.

In the step imbalance setting, both CIFAR-10 and CIFAR-100 are split into two classes, i.e., \textit{Head} (Frequent) and \textit{Tail} (Minority), with $\beta$ = 100. The first 5 (Head) classes of CIFAR-10 contain 5000 samples each, along with 50 samples each in the remaining 5 (Tail) classes. On the other hand, the top first 50 (Head) classes of CIFAR-100 contain 500 samples each, and the remaining 50 (Tail) classes consist of 5 samples each. 

All the experiments on imbalanced CIFAR-10 and CIFAR-100 are run with ResNet-32 backbone and SGD with momentum 0.9 as the base optimizer. All the methods train on imbalanced CIFAR-10 and CIFAR-100 with a batch size of 128 for 200 epochs, except for VS Loss, which runs for 300 epochs. We follow the learning rate schedule mentioned in \citet{cao2019learning}. In the initial 5 epochs, we linearly increase the learning rate to reach 0.1. Following that, a multi-step learning rate schedule decays the learning rate by scaling it with 0.001 and 0.0001 at 160th and 180th epoch, respectively. For LDAM runs on imbalanced CIFAR, the value of $C$ is tuned so that $\Delta_j$ is normalised to set maximum margin of 0.5 (refer to Equation. \textcolor{red}{1} in main text). In the case of VS Loss, we use $\gamma$ as 0.05 and $\tau$ as 0.75 for imbalanced CIFAR-10 and CIFAR-100 datasets (refer to Equation. \textcolor{red}{3} in main text).

\vspace{1mm}\noindent \textbf{ImageNet-LT and iNaturalist 2018}:
The classes in ImageNet-LT and iNaturalist 2018 datasets are also divided into three subcategories, i.e., \textit{Head }(Many),\textit{ Mid} (Medium), and \textit{Tail} (Few). For ImageNet-LT, the head classes consist of the first 390 classes, mid classes contain the subsequent 445 classes, and the tail classes hold the remaining 165 classes. Whereas for iNaturalist 2018, first 842 classes fall into the head classes, subsequent 3701 classes into the mid classes, and the remaining 3599 into the tail classes. 

For ImageNet-LT and iNaturalist 2018, all the models are trained for 90 epochs with a batch size of 256. We use ResNet-50 architecture as the backbone and SGD with momentum 0.9 as the base optimizer. A cosine learning rate schedule is deployed with an initial learning rate of 0.1 and 0.2 for iNaturalist 2018 and ImageNet-LT, respectively. For LDAM runs on ImageNet-LT and iNaturalist 2018, the value of $C$ is tuned so that $\Delta_j$ is normalised to set maximum margin of 0.3 (refer to Equation. \textcolor{red}{1} in main text).  
\begin{table}
  \caption{$\rho$ value for used for reporting the results with SAM.}
  \label{saddlesam_tab:app_rho}
  \centering
  \begin{adjustbox}{max width=\linewidth}
  \begin{tabular}{l|c|c|c|c}
    \toprule
    \multicolumn{1}{c}{}  & \multicolumn{2}{c}{CIFAR-10}& \multicolumn{2}{c}{CIFAR-100}\\
    
    \cmidrule(r){1-5}
     &LT ($\beta$ = 100) & Step ($\beta$ = 100) & LT ($\beta$ = 100) & Step ($\beta$ = 100)  \\
    \midrule
    CE + SAM & 0.1 & 0.1 & 0.2 & 0.5  \Tstrut{} \Bstrut{}\\
\hline

 CE + DRW + SAM & 0.5  & 0.2 & 0.8 & 0.2 \Tstrut{} \Bstrut{}\\
    \hline

LDAM + DRW + SAM & 0.8 & 0.1 & 0.8 & 0.5 \Tstrut{}  \Bstrut{}\\
    \hline

VS + SAM & 0.5 & 0.2 & 0.8 & 0.2 \Tstrut{} \Bstrut{}\\

    \bottomrule
  \end{tabular}

   \end{adjustbox}

\end{table}

\vspace{1mm}\noindent \textbf{Optimum \textbf{$\rho$} value}: Table \ref{saddlesam_tab:app_rho} compiles the $\rho$ value used by SAM across various methods on imbalanced CIFAR-10 and CIFAR-100 datasets. The $\rho$ value in these runs is kept constant throughout the duration of training. 
We adopt a common step $\rho$ schedule for the SAM runs on both ImageNet-LT and iNaturalist 2018. We initialise the $\rho$ with 0.05 for the initial 5 epochs and change it to 0.1 till the 60th epoch. Following that, we increase the $\rho$ value to 0.5 for the final 30 epochs. 

\vspace{1mm}\noindent \textbf{How to select \textbf{$\rho$} ?} $\rho$ is an hyperparameter in the SAM algorithm and it is important to choose the right value of $\rho$ for best performance on long-tailed learning. We observe that default value of $\rho$ (0.05) as suggested in \citet{foret2021sharpnessaware} does not lead to significant gain in accuracy (Refer Fig. \textcolor{red}{4} in chapter), as it is not able to escape the region of negative curvature. On long-tail CIFAR-10 and CIFAR-100 setting with re-weighting (DRW), a large value of $\rho$ (0.5 or 0.8) seems to work best instead, as in this work our objective to escape saddle points instead of improving generalization. This can be intuitively understood as large regularization ($\rho$) is required for highly imbalanced datasets to escape saddle points as suggested by Theorem \textcolor{red}{2}. In Table \ref{saddlesam_tab:app_rho}, we have reported the $\rho$ value used in every experiment. For the large scale datasets like ImageNet-LT and iNaturalist 18, we found that progressively increasing the $\rho$ value gives the best results. This is based on the idea that, as the training progresses, more flatter regions can be recovered from the loss landscape \cite{bisla2022low}. In our experiments on ImageNet-LT, we use a large $\rho$ of 0.5 in the last 30 epochs of training and we observe that the tail accuracy significantly increases at this stage of training. For using the proposed method on a new imbalanced dataset, we suggest starting with $\rho$ = 0.05 and increasing $\rho$ till the overall accuracy starts to decrease.

\vspace{1mm}\noindent \textbf{LPF-SGD and PGD}:
We use the official implementation of LPF-SGD \cite{bisla2022low} \footnote{https://github.com/devansh20la/LPF-SGD} to report the results on CIFAR-10 LT and CIFAR-100 LT. For LPF-SGD, we use Monte Carlo iterations ($M$) = 8 and a constant filter radius ($\gamma$) of 0.001 (as defined in Algorithm 4.1 in \citet{bisla2022low}). We implement the stochastic PGD method \cite{jin2017escape, Jin2019StochasticGD} on our own since there is no official PyTorch implementation available. We sample the perturbation (noise) from a Gaussian distribution with zero mean and ($\sigma$) standard deviation. We use a $\sigma$ of 0.0001 for CIFAR-10 and CIFAR-100 LT experiments.

\vspace{1mm}\noindent \textbf{Hessian Experiments}: For calculating the Eigen Spectral Density, we use the PyHessian library \cite{yao2020pyhessian}. PyHessian uses Lanczos algorithm for fast and efficient computation of the complete Hessian eigenvalue density. The Hessian is calculated on the average loss of the training samples as done in \cite{pmlr-v97-ghorbani19b, yao2020pyhessian}. $\lambda_{min}$ and $\lambda_{max}$ are extracted from the complete Hessian eigenvalue density. It has been shown that the estimated spectral density calculated with the Lanczos algorithm can be used as an approximate to the exact spectral density \cite{pmlr-v97-ghorbani19b}. Several works \cite{foret2021sharpnessaware, pmlr-v97-ghorbani19b, gilmer2021loss, yao2020pyhessian} have used the same method to calculate spectral density and analyze the loss landscape of neural networks.

\vspace{1mm}\noindent All of our implementations are based on PyTorch \citep{NEURIPS2019_9015}. For experiments pertaining to imbalanced CIFAR, we use NVIDIA GeForce RTX 2080 Ti, whereas for the large scale ImageNet-LT and iNaturalist 2018, we use NVIDIA A100 GPUs. We log our experiments with Wandb~\cite{wandb}.

\begin{figure}[!t]
\begin{subfigure}{.32\textwidth}
  \centering
  \includegraphics[width=1.0\linewidth]{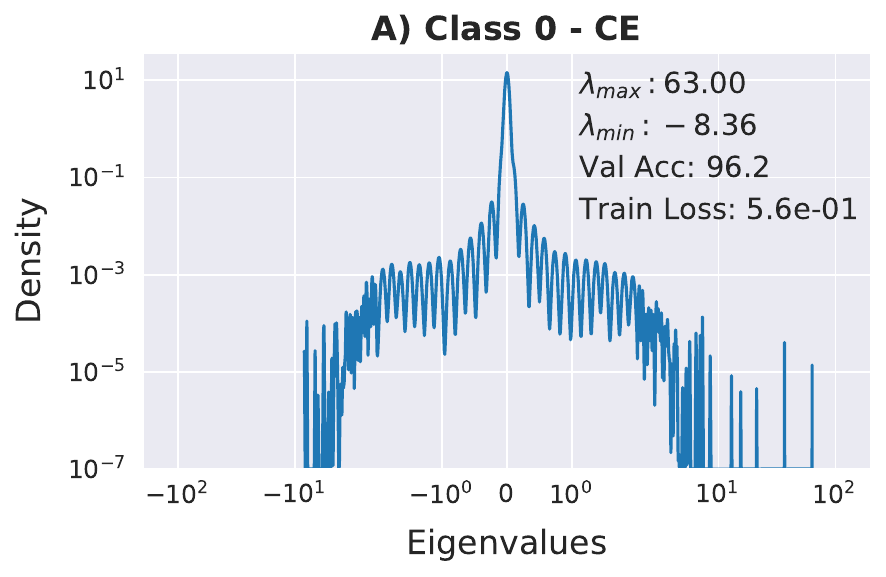}  

  \label{saddlesam_fig:sub-first3}
\end{subfigure}
\begin{subfigure}{.32\textwidth}
  \centering
  \includegraphics[width=1.0\linewidth]{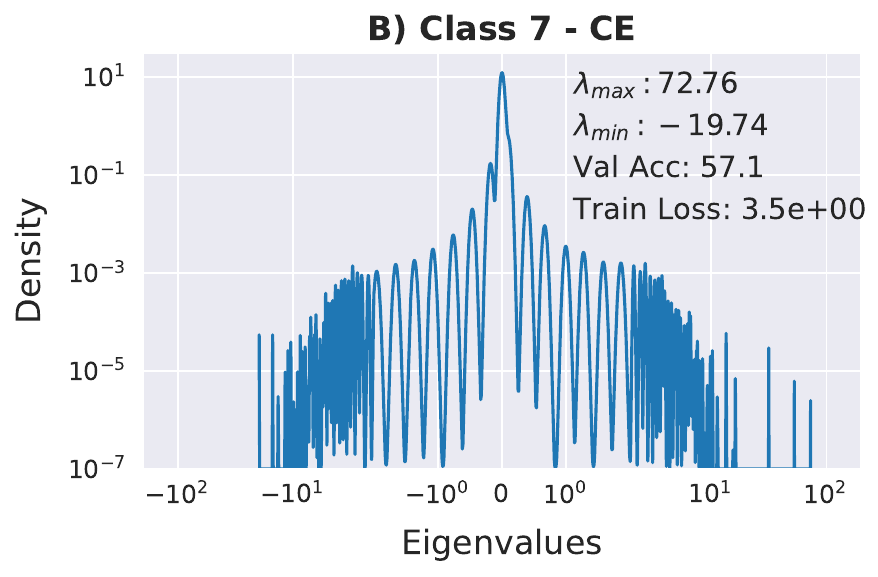}  

  \label{saddlesam_fig:sub-second4}
\end{subfigure}
\begin{subfigure}{.32\textwidth}
  \centering
  \includegraphics[width=1.0\linewidth]{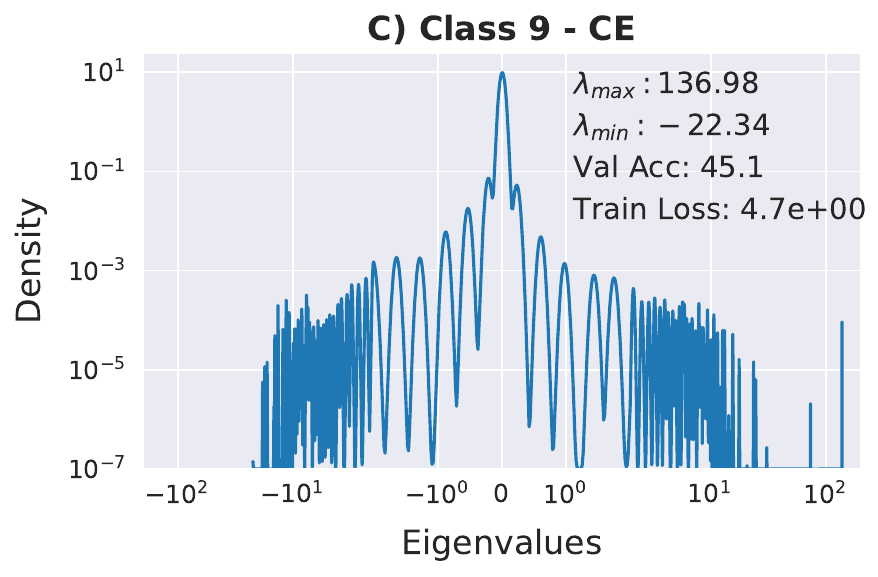}  

  \label{saddlesam_fig:sub-second5}
\end{subfigure}

\caption{Eigen Spectral Density of Head (Class 0) and Tail (Class 7 and Class 9) with standard CE (without re-weighting). Since CE minimizes the average loss, it can be seen that the loss on the tail class samples (B and C) is quite high. On the head class (A), the loss is low and $\lambda_{min}$ is close to 0.}
\label{saddlesam_fig:app_ce_none}

\end{figure}

\begin{table}
  \caption{Results on ImageNet-LT (ResNet-50) with LDAM+DRW and comparison with other methods. The numbers for methods marked with $\dag$ are taken from \cite{zhong2021improving}.}
  \label{saddlesam_tab:app_imagenetlt}
  \centering
  \begin{adjustbox}{max width=\linewidth}
  \begin{tabular}{l|c|llll}
    \toprule
     &Two stage &Acc & Head & Mid & Tail  \\
    \midrule
    CE &\Cross & 42.7 & 62.5 & 36.6 & 12.5 \\
    \hline
    cRT \cite{Kang2020Decoupling} $\dag$&\checkmark &50.3 & \underline{62.5} & 47.4 & 29.5 \Tstrut{}\\
    LWS \cite{Kang2020Decoupling} $\dag$ &\checkmark &51.2 & 61.8 & 48.6 & 33.5\\

    MisLAS \cite{zhong2021improving} &\checkmark &52.7 &61.7 & 51.3 & \textbf{35.8}\\
    DisAlign \cite{zhang2021distribution}& \checkmark&52.9 & 61.3 & \textbf{52.2} & 31.4\\
    DRO-LT* \cite{Samuel_2021_ICCV} & \Cross&\textbf{53.5} & \textbf{64.0} & 49.8 & 33.1 \\
\hline
    LDAM + DRW &\Cross &49.9 & 61.1 & 48.2 & 28.3 \Tstrut{} \\
    \rowcolor{mygray} LDAM + DRW + SAM &\Cross& \underline{53.1} & 62.0 & \underline{52.1} & \underline{34.8}  \\
    \bottomrule
  \end{tabular}
  \end{adjustbox}
\end{table}

\section{Additional Eigen Spectral Density Plots}
\label{saddlesam_app:eigen_spectral_density_plots}
We find that the spectral density of a class is representative of the other classes in same category (Head, Mid or Tail), hence for brevity we only display the eigen spectrum of one class per category for analysis.

\begin{figure}[t]
\begin{subfigure}{.32\textwidth}
  \centering
  \includegraphics[width=1.0\linewidth]{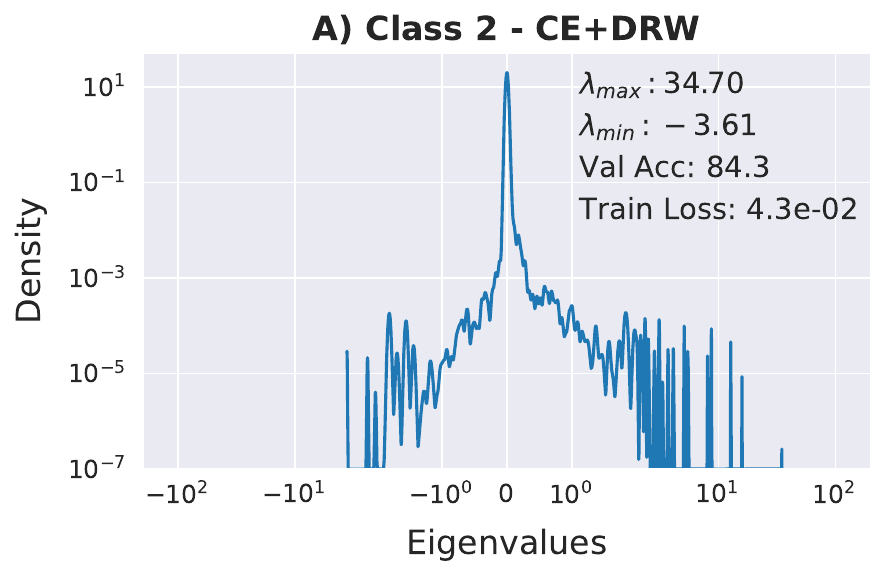}  

  \label{saddlesam_fig:sub-first4}
\end{subfigure}
\begin{subfigure}{.32\textwidth}
  \centering
  \includegraphics[width=1.0\linewidth]{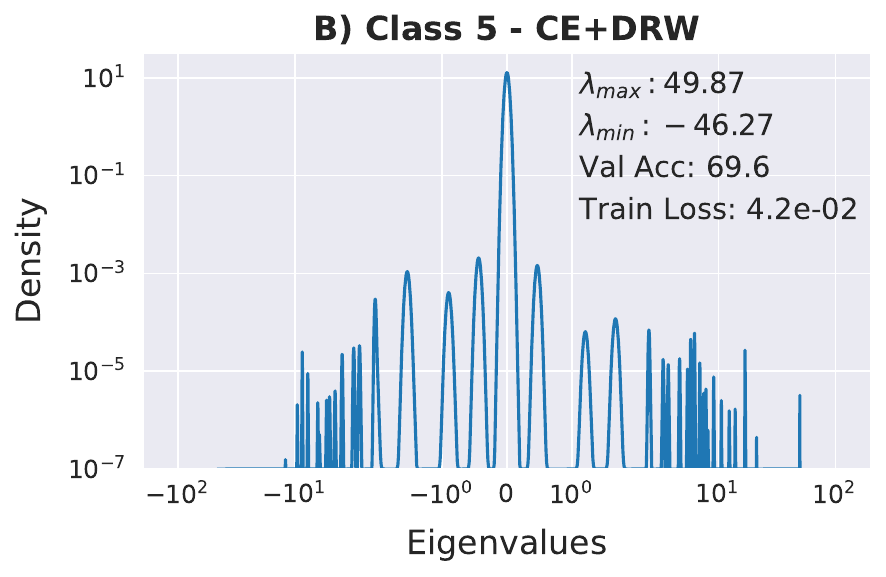}  

  \label{saddlesam_fig:sub-second6}
\end{subfigure}
\begin{subfigure}{.32\textwidth}
  \centering
  \includegraphics[width=1.0\linewidth]{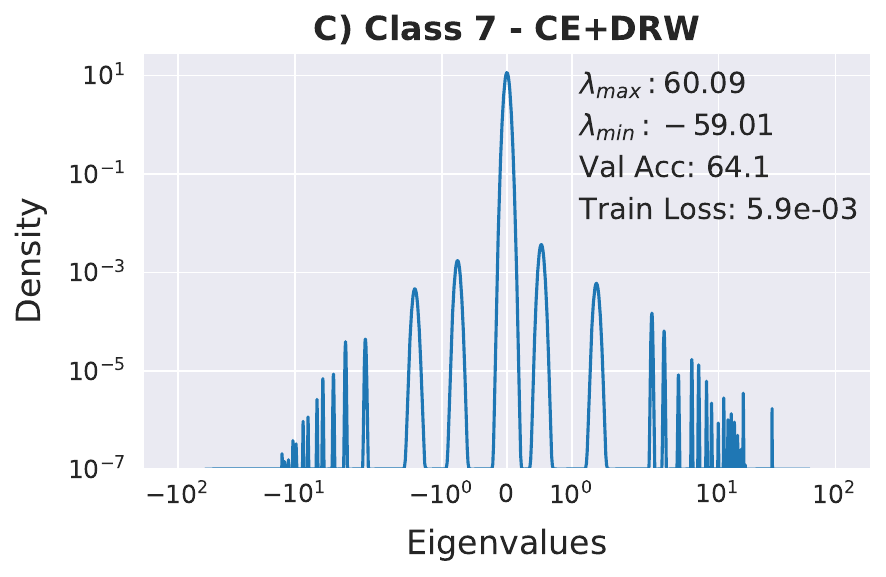}  

  \label{saddlesam_fig:sub-second7}
\end{subfigure}
\newline

\begin{subfigure}{.32\textwidth}
  \centering
  \includegraphics[width=1.0\linewidth]{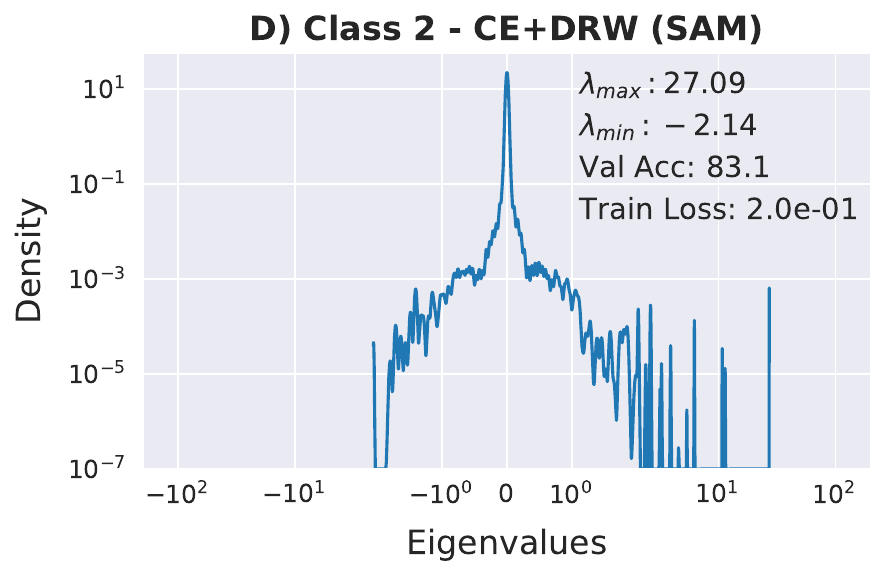}  

  \label{saddlesam_fig:sub-third2}
\end{subfigure}
\begin{subfigure}{.32\textwidth}
  \centering
  \includegraphics[width=1.0\linewidth]{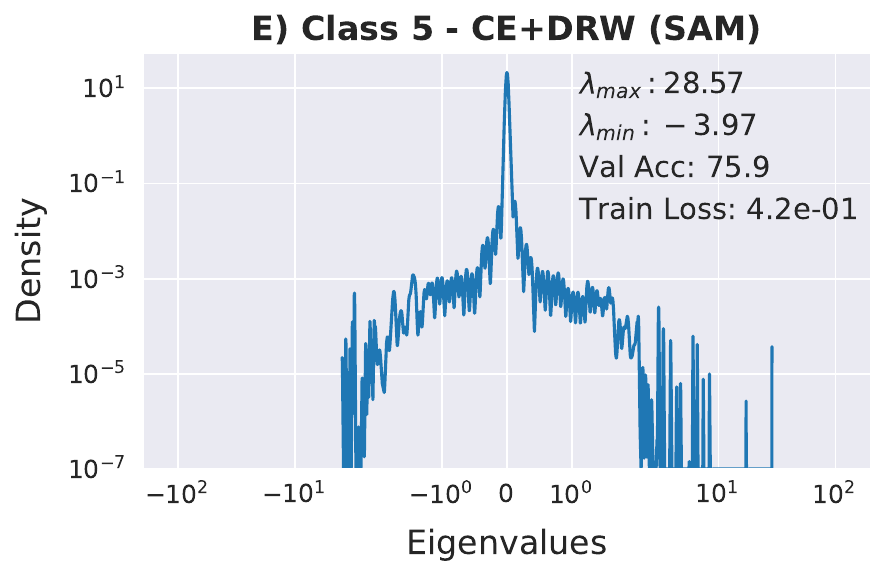}  

  \label{saddlesam_fig:sub-fourth1}
\end{subfigure}
\begin{subfigure}{.32\textwidth}
  \centering
  \includegraphics[width=1.0\linewidth]{    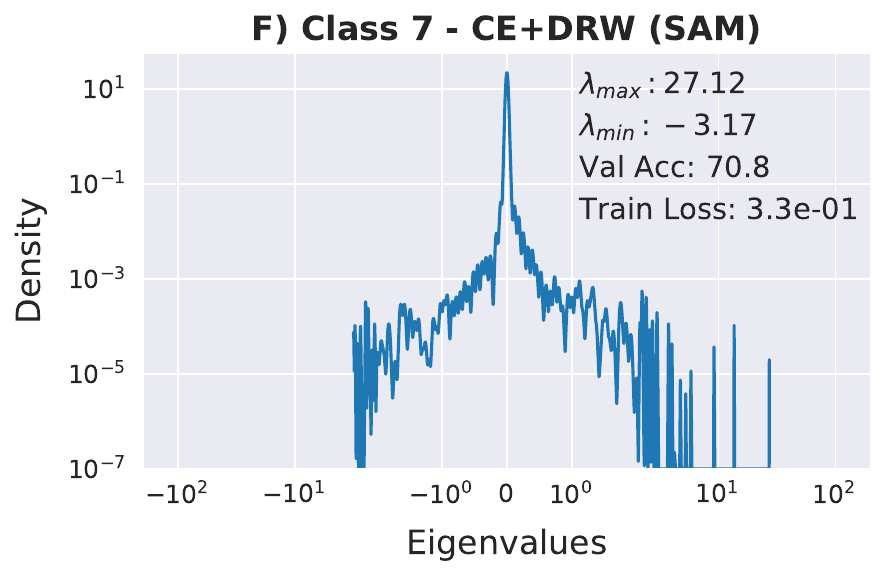}  

  \label{saddlesam_fig:sub-second9}
\end{subfigure}
\caption{Eigen Spectral Density of the Head (Class 2), Mid (Class 5) and Tail classes (Class 7) with CE+DRW and CE+DRW+SAM.}
\label{saddlesam_fig:app_head-tail-esd}
\end{figure}

\textbf{CE}: The spectral density on the standard CE loss (without re-weighting) can be seen in Fig. \ref{saddlesam_fig:app_ce_none}. We notice that the density and magnitude of negative eigenvalues is much larger for the tail class (Class 7 and Class 9 in Fig. \ref{saddlesam_fig:app_ce_none}\textcolor{red}{B} and \ref{saddlesam_fig:app_ce_none}\textcolor{red}{C}) compared to the head classes (Fig. \ref{saddlesam_fig:app_ce_none}\textcolor{red}{A}). On the other hand, the spectral density of the head class (Class 0) is very different from that of the tail class, with $\lambda_{min}$ of the head class very close to 0 indicating convergence to minima.

It must be noted that without re-weighting, the loss on the tail class samples is high because CE minimizes the average loss. Hence, the solution may not converge for tail class loss. However, in CE+DRW after re-weighting, we observe that the loss on tail class samples is very low, which indicates convergence to a stationary point. Thus, in CE+DRW, we can evidently conclude that the presence of large negative curvature indicates convergence to a saddle point. In summary, we find that just using CE converges to a point with significant negative curvature in tail class loss landscape. Further, though DRW is able to decrease the loss on tail classes, it still does converge to a point with significant negative curvature. This indicates that it converges to a saddle point instead of a minima. Hence, both CE and CE+DRW do not converge to local minima in tail class loss landscape.

\vspace{1mm}\noindent \textbf{CE+DRW}: We show additional class wise Eigen Spectral Density plots with CE+DRW and CE+DRW with SAM in Fig. \ref{saddlesam_fig:app_head-tail-esd}. We analyze the spectral density plots on Head (Class 2), Mid (Class 5) and Tail (Class 7). It can be seen that the magnitude of $\lambda_{max}$ and $\lambda_{min}$ is much lower with SAM in all the classes (Fig. \ref{saddlesam_fig:app_head-tail-esd} \textcolor{red}{D}, \textcolor{red}{E}, \textcolor{red}{F}). This indicates that SAM reaches a flatter local minima with no significant presence of negative eigenvalues, escaping saddle points. 

\begin{figure}[t]
\begin{subfigure}{.49\textwidth}
  \centering
  \includegraphics[width=1.0\linewidth]{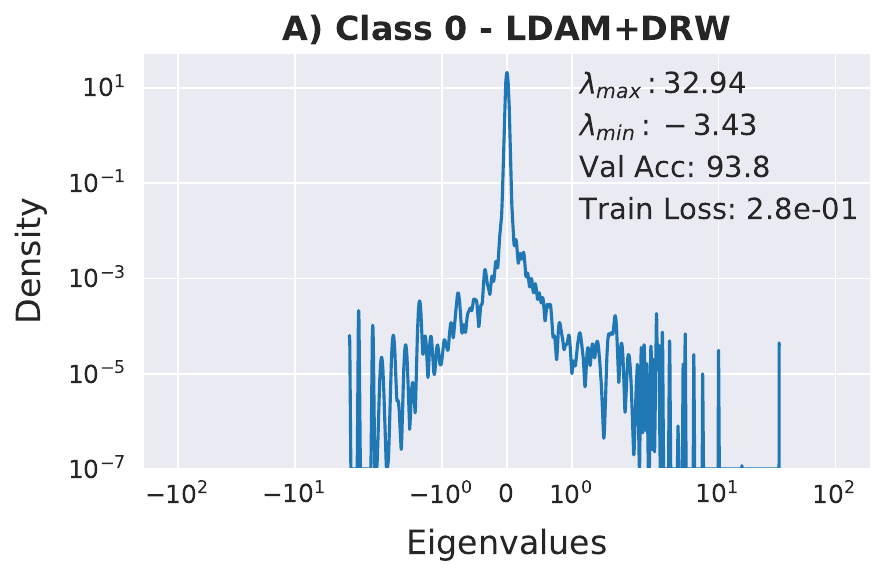}  

  \label{saddlesam_fig:sub-first1}
\end{subfigure}
\begin{subfigure}{.49\textwidth}
  \centering
  \includegraphics[width=1.0\linewidth]{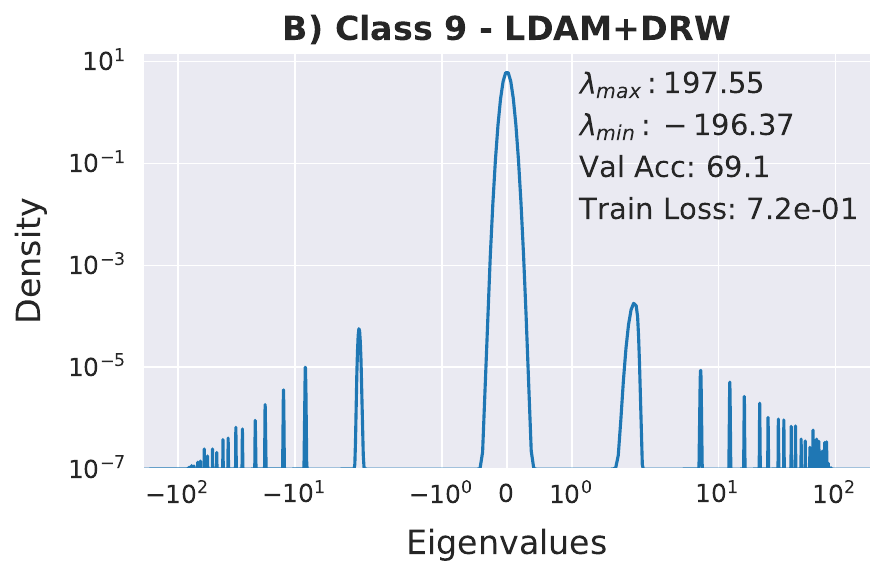}  

  \label{saddlesam_fig:sub-second10}
\end{subfigure}

\begin{subfigure}{.49\textwidth}
  \centering
  \includegraphics[width=1.0\linewidth]{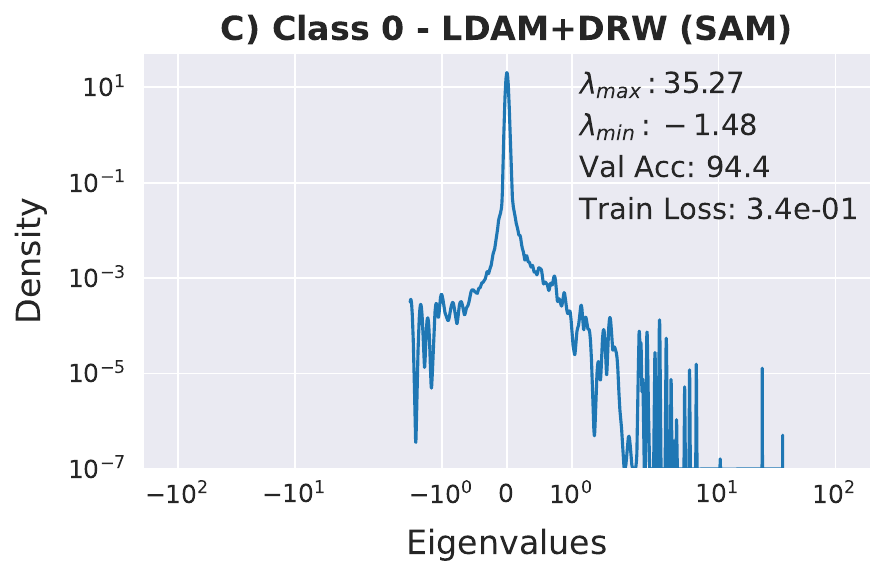}  

  \label{saddlesam_fig:sub-third3}
\end{subfigure}
\begin{subfigure}{.49\textwidth}
  \centering
  \includegraphics[width=1.0\linewidth]{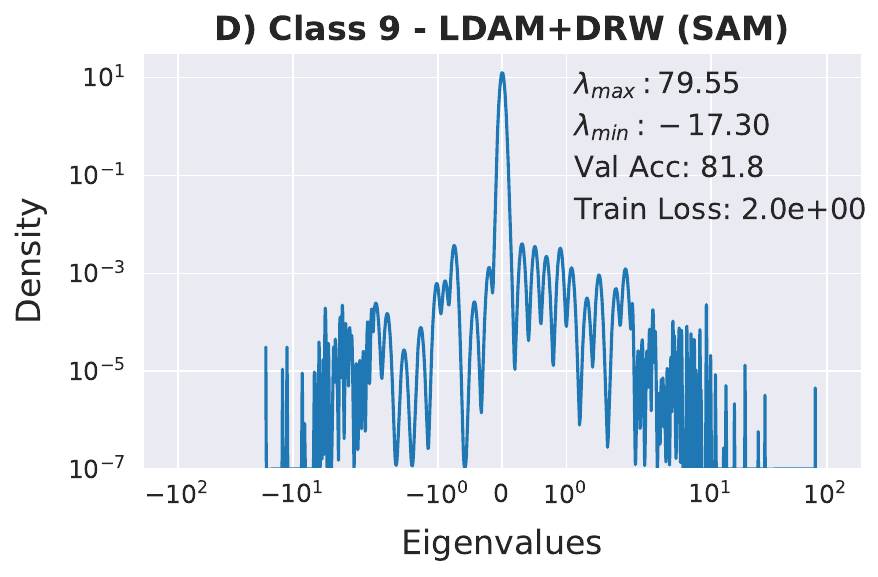}  

  \label{saddlesam_fig:sub-fourth5}
\end{subfigure}

\caption{Eigen Spectral Density of the Head (Class 0) and Tail (Class 9) class trained with LDAM. Even with LDAM, we observe existence of negative eigenvalues in the loss landscape for the tail class, which reduce in magnitude when LDAM is used with SAM.}
\label{saddlesam_fig:app_ldam}

\end{figure}

\vspace{1mm}\noindent \textbf{LDAM}: We also show Spectral density plots of Class 0 (Fig. \ref{saddlesam_fig:app_ldam} \textcolor{red}{A}, \textcolor{red}{C}) and Class 9 (Fig. \ref{saddlesam_fig:app_ldam} \textcolor{red}{B}, \textcolor{red}{D}) with LDAM+DRW method (SGD and SAM) in Fig. \ref{saddlesam_fig:app_ldam}. The existence of negative eigenvalues in the tail class spectral density (Fig. \ref{saddlesam_fig:app_ldam}\textcolor{red}{B}) indicates that even for LDAM loss (a regularized margin based loss), the solutions do converge to a saddle point. This also indicates that observations with CE+DRW hold good for long-tailed learning methods like LDAM which use margins instead of re-weighting directly. Hence, this gives evidence of the reason why SAM can be combined easily with  LDAM, VS Loss etc. to effectively improve performance.

The spectral density of the tail class of LDAM with SAM (Fig. \ref{saddlesam_fig:app_ldam}\textcolor{red}{D}) contains fewer negative eigenvalues compared to SGD (Fig. \ref{saddlesam_fig:app_ldam}\textcolor{red}{B}). This indicates convergence to local minima and clearly explains why SAM improves the performance of LDAM by 12.7\%.
\\

\section{Additional Results}
\label{saddlesam_app:additional_results}

For further establishing the generality of our method, we choose two recent orthogonal method Influence-Balanced Loss \cite{Park_2021_ICCV} (IB-Loss) and Parametric Contrastive Learning (PaCo) \cite{cui2021parametric} and apply proposed high $\rho$ SAM over them. We use the open-source implementations of IB-Loss \footnote{https://github.com/pseulki/IB-Loss} and PaCo \footnote{https://github.com/dvlab-research/Parametric-Contrastive-Learning} to reproduce the results and add our proposed method (high $\rho$ SAM) to that setup to obtain the results reported in the table below. We show results on CIFAR-100 LT with an imbalance factor ($\beta$) of 100 and 200. We observe that SAM with high $\rho$ significantly improves overall performance along with the performance on tail classes with both IB-Loss and PaCo method (Table \ref{saddlesam_tab:ib_cf100}). Despite PaCo baseline achieving close to state-of-the-art performance, the addition of high $\rho$ SAM is able to further improve the accuracy. This indicates the generality and applicability of proposed method across various long-tailed learning algorithms.

We show additional results on the large scale ImageNet-LT (Table \ref{saddlesam_tab:app_imagenetlt}) and iNaturalist 2018 (Table \textcolor{red}{3} in main text) dataset with LDAM-DRW. We also compare with recent long-tail learning methods:  cRT \cite{Kang2020Decoupling}, MisLAS \cite{zhong2021improving}, DisAlign \cite{zhang2021distribution} and DRO-LT \cite{Samuel_2021_ICCV}. On ImageNet-LT, LDAM+DRW with SAM leads to a 3.2\% gain in overall accuracy with 6.5\% increase in tail class accuracy. It can be seen that LDAM+DRW+SAM outperforms most other methods, including MisLAS which uses mixup. Also, it is important to note that MisLAS is trained for 180 epochs unlike LDAM+DRW which is trained only for 90 epochs. We observe that LDAM+DRW+SAM surpasses the performance of two-stage training methods including MisLAS, cRT, LWS, and DisAlign. Compared to these two-stage methods, our method is a single stage method and outperforms these two-stage methods. We want to add that we were not able to reproduce the numbers reported in DRO-LT* \cite{Samuel_2021_ICCV} when we were trying to incorporate SAM with DRO-LT.

With LDAM+DRW, the addition of SAM results in an increase in Head, Mid and Tail categories on iNaturalist 2018 (Table \textcolor{red}{3} in chapter). Specifically, LDAM+DRW+SAM outperforms all other methods in the tail class accuracy. 

This further emphasizes that our analysis is applicable to large scale imbalanced datasets like ImageNet-LT and iNaturalist 2018. We also want to highlight that our analysis shows that high $\rho$ SAM with re-weighting can be used as a \emph{strong baseline} in long tailed visual recognition problem. We also find that SAM is highly compatible with different loss-based methods (like LDAM, VS) for tackling imbalance and can be used to achieve significantly better performance.
\begin{table}
  \caption{Results on CIFAR-10 LT with different Imbalance Factor ($\beta$).}
  \label{saddlesam_tab:cifar10_imbalance_lt}
  \centering
  \begin{adjustbox}{max width=\linewidth}
  \begin{tabular}{l|llll | llll}
    \toprule
    \multicolumn{1}{c}{}  & \multicolumn{4}{c}{$\beta$ = 10}& \multicolumn{4}{c}{$\beta$ = 50}\\
    
    \cmidrule(r){1-9}
     &Acc & Head & Mid & Tail & Acc & Head  & Mid & Tail  \\
    \midrule
    CE + DRW \cite{cao2019learning} & 88.3 & 93.6  & 85.3 & 86.9 & 79.9 & 92.2 & 76.5 & 72.0 \Tstrut{}\\
    \rowcolor{mygray}
    \rowcolor{mygray} CE + DRW + SAM & 89.7 & 93.4 & 86.1  & 90.8 & 83.8 & 91.3 & 80.5 & 80.8  \Bstrut{}\\
    \hline
    LDAM + DRW \cite{cao2019learning} & 87.8 & 91.9  & 85.0 & 87.5 & 82.0 & 90.9 & 78.7 & 77.5 \Tstrut{}\\
    \rowcolor{mygray} LDAM + DRW + SAM & 89.4 & 93.4  & 86.2 & 89.8 & 84.8 & 92.8 & 82.1 & 80.4  \Bstrut{}\\
    \hline
    \toprule
    \multicolumn{1}{c}{}  & \multicolumn{4}{c}{$\beta$ = 100}& \multicolumn{4}{c}{$\beta$ = 200}\\
    
    \cmidrule(r){1-9}
     &Acc & Head & Mid & Tail & Acc & Head  & Mid & Tail  \\
    \midrule
    CE + DRW \cite{cao2019learning} & 75.5 & 91.6  & 74.1 & 61.4 & 69.9 & 91.1 & 70.0 & 48.4 \Tstrut{}\\
    \rowcolor{mygray}
    \rowcolor{mygray} CE + DRW + SAM & 80.6 & 91.4 & 78.0  & 73.1 & 76.6 & 91.5 & 74.9 & 64.0 \Bstrut{}\\
    \hline
    LDAM + DRW \cite{cao2019learning} & 77.5 & 91.1  & 75.7 & 66.4 & 72.5 & 90.2 & 72.3 & 54.9 \Tstrut{}\\
    \rowcolor{mygray} LDAM + DRW + SAM & 81.9 & 91.0  & 79.2 & 76.4 & 78.1 & 91.2 & 75.6 & 68.4  \Bstrut{}\\
    \bottomrule
  \end{tabular}

   \end{adjustbox}

\end{table}
\section{Additional Results with Varying Imbalance Factor}
\label{saddlesam_app:different_if}
We show the results with different imbalance factors ($\beta$ = 10, 50, 100 and 200) on CIFAR-10 LT (Table \ref{saddlesam_tab:cifar10_imbalance_lt}) and CIFAR-100 LT (Table \ref{saddlesam_tab:cifar100_imbalance_lt}) datasets with two methods. It can be seen that the observations in Table \textcolor{red}{1} are applicable with different degrees of imbalance. SAM with re-weighting improves upon the performance of CE and LDAM losses in all the experiments with varied imbalance factor. We observe an average increase of 3.9\% and 3.2\% on CIFAR-10 LT and CIFAR-100 LT datasets, respectively. This gain in performance is primarily due to the improvement in the tail accuracy, which increases by 8.6\% on CIFAR-10 LT and 3.9\% on CIFAR-100 LT.

As the dataset becomes more imbalanced ($\beta$ increases), the gain in accuracy with SAM on the tail classes improves significantly. For instance, on CIFAR-10 LT with $\beta$ = 10 (Table \ref{saddlesam_tab:cifar10_imbalance_lt}), CE+DRW+SAM improves upon CE+DRW by 1.2\% with a 3.9\% increase in tail class accuracy. However, with a more imbalanced dataset (\ie CIFAR-10 LT $\beta$ = 200), SAM leads to a 6.7\% boost in overall accuracy with a massive 15.6\% increase in the tail class performance.

\begin{table}[!t]
  \caption{Results on CIFAR-100 LT with different Imbalance Factor ($\beta$).}
  \label{saddlesam_tab:cifar100_imbalance_lt}
  \centering
  \begin{adjustbox}{max width=\linewidth}
  \begin{tabular}{l|llll | llll}
    \toprule
    \multicolumn{1}{c}{}  & \multicolumn{4}{c}{$\beta$ = 10}& \multicolumn{4}{c}{$\beta$ = 50}\\
    
    \cmidrule(r){1-9}
     &Acc & Head & Mid & Tail & Acc & Head  & Mid & Tail  \\
    \midrule
    CE + DRW \cite{cao2019learning} & 58.1 & 65.6 & 58.5 & 48.2 & 46.5 & 63.3 & 47.5 & 24.4 \Tstrut{}\\
    \rowcolor{mygray}
    \rowcolor{mygray} CE + DRW + SAM & 60.7 & 66.0 & 60.5 & 54.4 & 50.0 & 61.9 & 50.9 & 33.7 \Bstrut{}\\
    \hline
    LDAM + DRW \cite{cao2019learning} & 57.8 & 67.5 & 58.9 & 44.5 & 47.1 & 62.9 & 48.2 & 26.1\Tstrut{}\\
    \rowcolor{mygray} LDAM + DRW + SAM & 60.1 & 70.2 & 61.3 & 46.1 & 49.4 & 66.1 & 50.2 & 27.8  \Bstrut{}\\
    \hline
    \toprule
    \multicolumn{1}{c}{}  & \multicolumn{4}{c}{$\beta$ = 100}& \multicolumn{4}{c}{$\beta$ = 200}\\
    
    \cmidrule(r){1-9}
     &Acc & Head & Mid & Tail & Acc & Head  & Mid & Tail  \\
    \midrule
    CE + DRW \cite{cao2019learning} & 41.0 & 61.3 & 41.7 & 14.7 & 36.9 & 59.7 & 36.1 & 9.6 \Tstrut{}\\
    \rowcolor{mygray}
    \rowcolor{mygray} CE + DRW + SAM & 44.6 & 61.2 & 47.5 & 20.7 & 41.7 & 63.4 & 43.0 & 13.1 \Bstrut{}\\
    \hline
    LDAM + DRW \cite{cao2019learning} & 42.7 & 61.8 & 42.2 & 19.4 & 38.3 & 58.8 & 36.3 & 15.1 \Tstrut{}\\
    \rowcolor{mygray} LDAM + DRW + SAM & 45.4 & 64.4 & 46.2 & 20.8 & 42.0 & 63.0 & 41.4 & 16.6 \Bstrut{}\\
    \bottomrule
  \end{tabular}

   \end{adjustbox}

\end{table}

\begin{table}[t]
  \caption{Results on CIFAR-100 LT with IB-Loss and PaCo.}
  \label{saddlesam_tab:ib_cf100}
  \centering
  \begin{adjustbox}{max width=\linewidth}
  \begin{tabular}{l|ll|ll}
    \toprule
    \multicolumn{1}{c}{}  & \multicolumn{2}{c}{$\beta$ = 100}& \multicolumn{2}{c}{$\beta$ = 200}\\
    
     & Acc & Tail & Acc & Tail  \\

    \hline

    IB \cite{cao2019learning} &   40.4  & 14.9 & 36.7 & 10.3\Tstrut{}\\

    \cellcolor{mygray}{IB + SAM} & \cellcolor{mygray}42.8 &  \cellcolor{mygray}25.0 &\cellcolor{mygray}37.7  &\cellcolor{mygray}17.8 \Bstrut{}\\
    \hline
    PaCo \cite{cui2021parametric} & 51.5 & 33.9 & 47.0 & 26.9\Tstrut{}\\
    \cellcolor{mygray}{PaCo + SAM} & \cellcolor{mygray}53.0 & \cellcolor{mygray}36.0 &\cellcolor{mygray}48.0  &\cellcolor{mygray}27.8 \Bstrut{}\\

    \bottomrule
  \end{tabular}

   \end{adjustbox}

\end{table}
\section{Algorithm}
\label{saddlesam_app:algorithm}
We describe our method in detail in Algorithm \ref{saddlesam_algo:DRW_SAM}.  On the large scale ImageNet-LT and iNaturalist-18 dataset, we use $\rho_{drw} \geq \rho$. For CIFAR-10 LT and CIFAR-100 LT, we find that $\rho$ = $\rho_{drw}$ works well.
\begin{algorithm}[!h]
\caption{DRW + SAM}
\label{saddlesam_algo:DRW_SAM}
\begin{algorithmic}[1]
\REQUIRE  Network $g$ with parameters $w$; Training set $\sS$; Batch size $b$; Learning rate $\eta>0$; Neighborhood size $\rho>0$, Neighborhood size for re-weighted loss $\rho_{drw} >= \rho$; Total Number of Iterations $E$; Deferred Reweighting Threshold $T$; Number of samples in class $y$: $n_y$; Loss Function $\mathcal{L}$ (Cross-Entropy, LDAM).

	\FOR{$i=1$ to $E$ }
    \STATE Sample a mini-batch $\sB\subset \sS$  with size $b$. %
    \IF{$E$ < $T$}
      \STATE Compute Loss $\mathcal{L} \leftarrow \frac{1}{b} \sum_{(x,y)\in \sB}\mathcal{L}(y;g_w(x)) $ 
      \STATE Compute $\epsilon \leftarrow \rho *  \nabla_{w} {\mathcal{L}}/ ||\nabla_w {\mathcal{L}} ||$ \COMMENT{Compute Sharp-Maximal Point}
      \STATE Compute Loss at $ w + \epsilon; \mathcal{L} \leftarrow \frac{1}{b} \sum_{(x,y)\in \sB}\mathcal{L}(y;g_{w + \epsilon}(x))$ 
      \STATE Calculate gradient $d$: $d \leftarrow  \nabla_w{\mathcal{L}}$
    \ELSE %
    
    \STATE Compute re-weighted Loss $\mathcal{L}_\mathrm{\scriptscriptstyle RW} \leftarrow \frac{1}{b} \sum_{(x,y)\in \sB} n_y^{-1}\cdot \mathcal{L}(y;g_w(x)) $ 
      \STATE Compute $\epsilon \leftarrow \rho_{drw} *  \nabla_w {\mathcal{L}_\mathrm{\scriptscriptstyle RW}}/ ||\nabla_w {\mathcal{L}_\mathrm{\scriptscriptstyle RW}} ||$
      \STATE Compute re-weighted Loss at $ w + \epsilon; \mathcal{L}_\mathrm{\scriptscriptstyle RW} \leftarrow \frac{1}{b} \sum_{(x,y)\in \sB}n_y^{-1}\cdot \mathcal{L}(y;g_{w + \epsilon}(x))$ 
      \STATE Calculate gradient $d$: $d \leftarrow  \nabla_w{\mathcal{L}_\mathrm{\scriptscriptstyle RW}}$
     \ENDIF
    
       \STATE Update weights $w_{i+1} \leftarrow w_{i} - \eta d$
    
\ENDFOR
\end{algorithmic}
\end{algorithm}

 \renewcommand{\thesection}{E}

\supersection{DeiT-LT: Distillation Strikes Back for Vision Transformer Training on
Long-Tailed Datasets (Chapter-6)}

\section{Experimental Details}
\label{deit-lt_suppl:experimental}
\subsection{Datasets}
\label{deit-lt_suppl:datasets}
\textbf{CIFAR-10 LT and CIFAR-100 LT.} We use the imbalanced CIFAR-10 and CIFAR-100 datasets with an exponential decay in sample size across classes. This decay is guided by the Imbalance Ratio ($\rho =\frac{\max_i N_{i}}{\min_j N_j}$). For our experiments on CIFAR-10 LT and CIFAR-100 LT, we show the results on $\rho =100$ and $\rho =50$. CIFAR-10 LT comprises 12,406 training images across 10 classes ($\rho$ = 100). Out of the 10 classes, the first 3 classes are considered \textit{Head} classes with more than 1500 images per class, the following 4 classes are \textit{Mid} (medium) classes with more than 250 images each class, and the last 3 classes account for the \textit{Tail} classes, with each class containing less than 250 images each. Following a similar decay, the 100 classes of CIFAR-100 LT (10,847 training samples with $\rho$ = 100) are also divided into three subcategories: the first 36 classes are considered as the \textit{Head} classes, \textit{Mid} contains the following 35 classes, and the remaining 29 classes are labeled as \textit{Tail} classes. Both CIFAR-10 LT and CIFAR-100 LT datasets are evaluated on held-out sets of 10,000 images each, equally distributed across classes.

\vspace{1mm} \noindent \textbf{ImageNet-LT.} We use the standard LT dataset created out of ImageNet~\cite{russakovsky2015imagenet}. ImageNet-LT consists of 115,846 training images, with 1280 images in the class with the most images and 5 images in the class with the least images. Out of the 1,000 classes sorted in the descending order of sample frequency, we consider classes with more than 100 samples as \textit{Head} classes, the classes with samples between 20 and 100 to be \textit{Mid} classes and the classes with less than 20 samples as the \textit{Tail} classes as done in \citet{cui2021parametric}.

\vspace{1mm} \noindent \textbf{iNaturalist-2018.} iNaturalist-2018~\cite{van2018inaturalist} is a real-world imbalanced dataset with 437,513 training images. Out of the 8,142 classes sorted in the descending order of sample frequency, we consider classes with more than 100 samples as \textit{Head} classes, the classes with samples between 20 and 100 to be \textit{Mid} classes and the classes with less than 20 samples as the \textit{Tail} classes, similar to ImageNet-LT.

\begin{table*}[t]
    \centering
    
    \caption{Summary of our training procedures used to train DeiT-LT Base (B) from scratch on CIFAR-10 LT, CIFAR-100 LT, ImageNet-LT and iNaturalist-2018. 
    }
    \adjustbox{max width=\textwidth}{
    \begin{tabular}{c|c|c|c|c}
        \toprule[1pt]
        \rowcolor{Gray} Procedure & CIFAR-10 LT & CIFAR-100 LT & ImageNet-LT & iNaturalist-2018 \\
        \midrule
        Epochs   &1200&1200&1400&1000 \\
        Optimizer & AdamW &AdamW&AdamW& AdamW\\
        Effective Batch Size  & 1024 & 1024 &2048&2048\\
        LR  & 5$\times10^{-4}$ & 5$\times10^{-4}$ & 5$\times10^{-4}$ & 5$\times10^{-4}$\\
        LR schedule & cosine & cosine & cosine & cosine \\
        Warmup Epochs   & 5 & 5 & 5 & 5 \\
        DRW starting epoch & 1100 & 1100 & 1200 & 900\\
        \midrule
        Mixup ($\alpha$) & 0.8 & 0.8 &0.8&0.8 \\
        Cutmix ($\alpha$) & 1.0 & 1.0 & 1.0 & 1.0 \\
        Mixup and Cutmix during DRW &$\times$& $\times$ &$ \checkmark $&$ \checkmark $ \\
        Horizontal Flip & $\checkmark$ &$\checkmark$&$ \checkmark $&$ \checkmark $ \\
        Color Jitter & $\checkmark$ & $\checkmark$ &$ \checkmark $& $ \checkmark $\\
        Random Erase & $\checkmark$ & $\checkmark$ &$ \times $& $ \times $\\
       Label smoothing & 0.1 & 0.1 &0.1& 0.1\\
       Solarization & $\times$ & $\times$ & $\checkmark$& $\checkmark$\\ 
       Random Grayscale & $\times$ & $\times$ & $\checkmark$ & $\checkmark$ \\
        Repeated Aug & $ \checkmark $ &$ \checkmark $&$\times$&$\times$ \\
        Auto Aug  & $ \checkmark $ &$ \checkmark $&$ \times $&$ \times $ \\

\bottomrule      
    \end{tabular}}
    \label{deit-lt_tab:hyperparams}

\end{table*}

\subsection{Training Configuration}
\label{deit-lt_suppl:training_config}
In this subsection, we detail the strategies adopted to train DeiT-LT Base (B) model on four benchmark datasets, namely CIFAR-10 LT, CIFAR-100 LT, ImageNet-LT, and iNaturalist-2018. We use the AdamW optimizer to train DeiT-LT from scratch across all the datasets. These runs use a cosine learning rate decay schedule with an initial learning rate of 5$\times10^{-4}$. All the runs use a linear learning rate warm-up schedule for the initial five epochs. Furthermore, we deploy label smoothing with $\varepsilon=0.1$ for all our experiments where the ground truth labels are used to train the \texttt{CLS} expert. Under label smoothing, the true label is assigned a $(1-\varepsilon)$ probability, and the remaining $\varepsilon$ is distributed amongst the other labels. We use hard labels as distillation targets from the teacher network to train the \ttt{DIST} expert classifier via distillation from CNN teacher (Fig.~\ref{deit-lt_fig:overview}). For training the teacher networks with SAM optimizer, we follow the setup mentioned in ~\cite{rangwani_escapingsaddle}

\vspace{1mm} \noindent \textbf{CIFAR-10 LT and CIFAR-100 LT }: We train DeiT-LT for 1200 epochs on imbalanced versions of CIFAR datasets. DRW loss is added to the training of the \ttt{DIST} expert classifier after 1100 epochs. Mixup and Cutmix are used during the initial 1100 epochs of the training. As suggested in~\cite{touvron2021deit}, we use Repeated Augmentation to improve the performance of the DeiT-LT training. The (32$\times$ 32) images of CIFAR datasets are resized to (224 $\times$ 224) before feeding into the transformer architecture. For CIFAR-10 LT and CIFAR-100 LT datasets, ResNet-32 is used as the teacher network. The teacher is trained from scratch on these imbalanced datasets using LDAM+DRW+SAM~\cite{rangwani_escapingsaddle} and contrastive PaCo+SAM (training PaCo~\cite{cui2021parametric} with SAM~\cite{foret2020sharpness} optimizer) frameworks. The input images to the teacher are of size (32$\times$ 32), with the same augmentation used as input images to the teacher network during DeiT-LT training.  

\vspace{1mm} \noindent \textbf{ImageNet-LT and iNaturalist-2018.} DeiT-LT is trained from scratch for 1400 epochs on ImageNet-LT and for 1000 epochs on iNaturalist-2018. DRW loss for distillation head (\ttt{DIST} expert classifier) is initialized from epochs 1200 and 900 for ImageNet-LT and iNaturalist-2018, respectively. Mixup and Cutmix are used throughout the training, including the DRW training phase. More details regarding the training configuration can be found in Table \ref{deit-lt_tab:hyperparams}.

For the ImageNet-LT and iNaturalist-2018 datasets, the ResNet-50 teacher is trained from scratch on the respective datasets using the LDAM+DRW+SAM~\cite{rangwani_escapingsaddle} and contrastive PaCo+SAM (training PaCo~\cite{cui2021parametric} with SAM~\cite{foret2020sharpness} optimizer) methods. The input image size is (224 $\times$ 224) for both the student and teacher network.

\subsection{Additional Baselines}
\label{deit-lt_suppl:additional_baselines}
\begin{figure}[!t]
    \centering
    \caption{ Comparison of training loss for vanilla ViT and ViT+LDAM training on CIFAR-10 LT}
    \includegraphics[width=0.6\columnwidth]{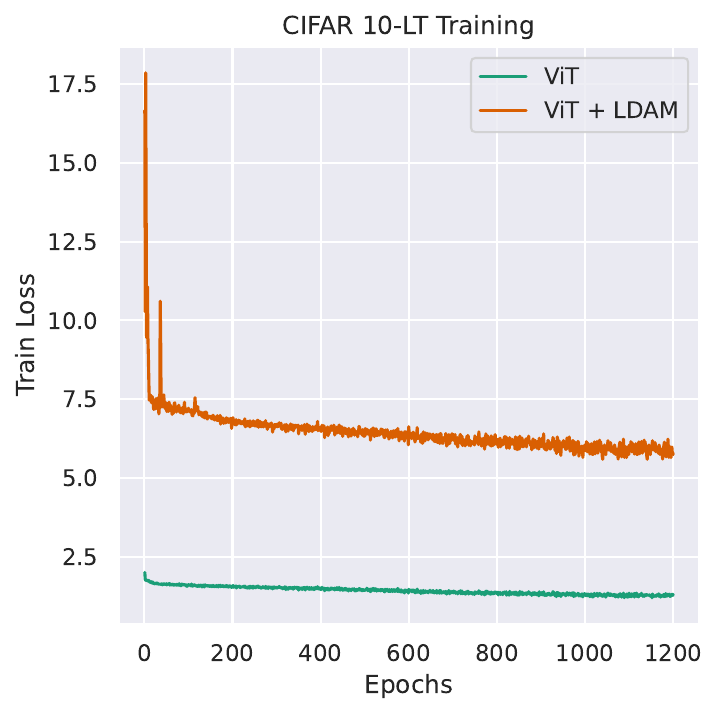}
    
    \label{deit-lt_fig:vit_ldam}
\end{figure}

We want to highlight that we attempted training baselines, like LDAM for vanilla ViT. However, we find that the LDAM baseline (52.75\%) performs inferiorly to the vanilla ViT baselines (62.62\%). We find that the loss for the LDAM baseline gets plateaued very early, and the model does not fit to the training dataset (Fig.~\ref{deit-lt_fig:vit_ldam}). To make the comparison fair with DeiT baselines, we used similar augmentation and other hyperparameters for the ViT Baselines. We think this can be one reason for the non-convergence of the ViT-LDAM baseline. We find that similar abysmal performance for LDAM baseline is also reported by the recent work~\cite{xu2023rethink}, which also resonates with our finding. We think that investigation into this behavior is a good direction for future work.  

 Additionally, for a fair comparison, we do not compare against baselines that use pre-training for long-tailed recognition tasks. RAC~\cite{tian2022vl} uses a ViT-B encoder for their retrieval module with weights obtained from pre-training on ImageNet-21K. The authors do not report on small-scale datasets, as they acknowledge the unfair advantage of using the information present in the pretrained encoder. Similarly, for small-scale datasets, LiVT~\cite{LiVT} method pretrains the encoder via Masked Generative Pretraining on ImageNet-1k. On the contrary, our DeiT-LT method enables training \textit{ViT from scratch} for both small-scale and large-scale datasets.

\subsection{Augmentations for OOD distillation}

\label{deit-lt_suppl:training_aug}

While both DeiT and our DeiT-LT pass images with strong augmentations to the teacher network for distilling into the student network, the set of augmentations used to train the teacher network itself differs between the two approaches. DeiT first trains a large teacher CNN (RegNetY-16GF) using the same set of strong augmentations as that used for the student network. However, we find that distilling from a small teacher CNN (such as ResNet32) trained with weak augmentations gives better performance (see Sec.~\ref{deit-lt_sebsec:out-of-dis-dist})
for more details). Table \ref{deit-lt_tab:teacher_comparison} compares the augmentations used to train the teacher for DeiT (RegNetY-16GF) and for our method DeiT-LT. Our experiments use ResNet32 as the teacher network for CIFAR-10 LT and CIFAR-100 LT, and ResNet50 for the Imagenet-LT and iNaturalist-2018 datasets. For the PaCo teacher, we utilize the mildly strong augmentations used by the PaCo~\cite{cui2021parametric} method itself. We would like to convey, that the PaCo training does not utilize the Mixup and CutMix augmentation in particular while training, which helps us to create OOD samples for this using Mixup and CutMix itself. Distilling via out-of-distribution (OOD) images enables the student to learn the inductive biases of the teacher effectively. This is particularly helpful in improving the performance on the tail classes that have significantly fewer training images. 
\begin{table}[t]
    \centering
    \caption{Comparing augmentation used to train RegNetY-16GF (teacher for DeiT training) and ResNet32 (teacher for DeiT-LT training) for CIFAR-10 LT.}
    \begin{tabular}{c|c|c}
        \toprule[1pt]
         \rowcolor{Gray}& RegNetY-16GF & ResNet32  \\
        \rowcolor{Gray}\multirow{-2}{*}{Procedure}& (Strong) & (Weak) \\
        \midrule
        Image Size & 224$\times$224& 32$\times$32\\
        Random Crop &$\checkmark$&$\checkmark$\\ 
        Horizontal Flip&$\checkmark$ &$\checkmark$ \\
        Mixup ($\alpha$)&0.8&$\times$\\
        Cutmix ($\alpha$) &1.0&$\times$\\
        Color Jitter &0.3&$\times$\\
        Random Erase &$\checkmark$&$\times$\\
        Auto Aug &$\checkmark$&$\times$\\
        Repeated Aug &$\checkmark$&$\times$\\
\bottomrule      
    \end{tabular}
    \label{deit-lt_tab:teacher_comparison}
\end{table}

\section{Detailed Results}
\label{deit-lt_suppl:detailed_results}

\begin{table*}
    \centering
    \caption{Accuracy of expert classifiers on Head, Mid, and Tail classes for CIFAR-10(100) LT.}
    \adjustbox{max width=\textwidth}{
    \begin{tabular}{c|c|c|ccc|c|ccc}
    \toprule[1pt]
         \rowcolor{Gray}& & \multicolumn{4}{c|}{CIFAR-10 LT} & \multicolumn{4}{c}{CIFAR-100 LT} \Tstrut\Bstrut\\
         \cline{3-10}
          \rowcolor{Gray}\multirow{-2}{*}{Imbalance} & \multirow{-2}{*}{Expert} & Overall & Head & Mid & Tail & Overall & Head & Mid & Tail \Tstrut\Bstrut\\
         \midrule[1pt]
         &Average & 87.3$_{\pm0.10}$&93.8$_{\pm0.33}$&83.7$_{\pm0.26}$&85.7$_{\pm0.33}$&54.8$_{\pm0.42}$&72.8$_{\pm0.16}$&55.9$_{\pm0.51}$&31.0$_{\pm0.73}$\Tstrut\Bstrut\\
         \cline{2-10}
         &\texttt{CLS}&78.6$_{\pm0.15}$&96.5$_{\pm0.06}$&79.4$_{\pm0.39}$&59.7$_{\pm0.20}$&43.3$_{\pm0.39}$&73.7$_{\pm0.19}$&41.7$_{\pm0.73}$&7.5$_{\pm0.26}$ \Tstrut\Bstrut\\
         \cline{2-10}
         \multirow{-3}{*}{100}&\texttt{DIST} &79.9$_{\pm0.31}$&72.8$_{\pm0.92}$&75.4$_{\pm0.18}$&93.0$_{\pm0.15}$&42.5$_{\pm0.48}$&39.3$_{\pm1.64}$&45.1$_{\pm0.47}$&43.1$_{\pm0.33}$\Tstrut\Bstrut\\
        \midrule[1pt]
        &Average&89.9$_{\pm0.17}$&94.5$_{\pm0.18}$&87.2$_{\pm0.26}$&88.8$_{\pm0.34}$&60.6$_{\pm0.03}$&74.6$_{\pm0.10}$&60.5$_{\pm0.10}$&43.1$_{\pm0.06}$  \Tstrut\Bstrut\\
         \cline{2-10}
         &\texttt{CLS}&84.1$_{\pm0.33}$&96.5$_{\pm0.12}$&83.3$_{\pm0.66}$&72.8$_{\pm0.55}$&49.6$_{\pm0.21}$&76.0$_{\pm0.31}$&50.5$_{\pm0.46}$&15.9$_{\pm0.41}$  \Tstrut\Bstrut\\
         \cline{2-10}
         \multirow{-3}{*}{50}&\texttt{DIST}&83.2$_{\pm0.23}$&74.6$_{\pm0.51}$&81.8$_{\pm0.21}$&93.6$_{\pm0.08}$&48.0$_{\pm0.20}$&44.0$_{\pm0.25}$&48.4$_{\pm0.36}$&52.6$_{\pm0.07}$  \Tstrut\Bstrut\\

         \bottomrule
         
    \end{tabular}}
    
    \label{deit-lt_tab:expert_performance_CF}
\end{table*}

\begin{table*}[h]
    \centering
    \caption{Accuracy of experts on Head, Mid and Tail classes for ImageNet-LT and iNaturalist-2018.}
    \adjustbox{max width=0.6\textwidth}{
    \begin{tabular}{c|c|ccc|c|ccc}
    \toprule[1pt]
         \rowcolor{Gray}& \multicolumn{4}{c|}{ImageNet-LT} & \multicolumn{4}{c}{iNaturalist-2018} \Tstrut\Bstrut\\
         \cline{2-9}
          \rowcolor{Gray}\multirow{-2}{*}{Expert} & Overall & Head & Mid & Tail & Overall & Head & Mid & Tail \Tstrut\Bstrut\\
          \midrule
         Average & 59.1 & 66.7 & 58.3 & 40.0 & 75.1 & 70.3 & 75.2 & 76.2 \Tstrut\Bstrut\\
         \hline
         \texttt{CLS} expert classifier & 47.5 & 68.3 & 40.0 & 13.5 & 65.6 & 73.8 & 65.8 & 63.1 \Tstrut\Bstrut\\
         \hline
         \texttt{DIST} expert classifier & 56.4 & 57.2 & 58.6 & 46.6 & 72.9 & 56.1 & 73.2 & 77.0 \Tstrut\Bstrut\\
         \bottomrule
    \end{tabular}}
    \label{deit-lt_tab:expert_performance_imnet}
\end{table*}

\textbf{Performance of individual experts}:
Our approach focuses on training diverse experts, where the \ttt{CLS} expert classifier is able to perform well on \textit{Head} (majority) classes, while the \ttt{DIST} expert classifier is able to perform well on the \textit{Tail} (minority) classes. By averaging the output of the individual classifiers, we are able to exploit the benefit of both. 

In this portion, we discuss the individual performance of the \ttt{CLS} and \ttt{DIST} expert classifiers of our proposed DeiT-LT method on CIFAR-10 LT, CIFAR-100 LT, ImageNet-LT, and iNaturalist-2018. As can be seen in Table \ref{deit-lt_tab:expert_performance_CF} and Table \ref{deit-lt_tab:expert_performance_imnet}, the \ttt{CLS} and \ttt{DIST} classifiers give a contrasting performance on the head and tail classes, supporting our claim of expert classifiers. For CIFAR-10 LT ($\rho$ = 100), the \ttt{CLS} expert classifier is able to report an accuracy of 96.5\% on images of the head classes, whereas the \ttt{DIST} expert classifier settles with 72.8\% on the same set of classes. On the other hand, the \ttt{DIST} expert classifier reports 93.0\% accuracy on the tail classes, which is almost 33\% more than that of the \ttt{CLS} expert classifier. Like CIFAR-10 LT, the \ttt{CLS} expert classifier performs better on the head classes of CIFAR-100 LT ($\rho$ = 100) than the \ttt{DIST}, whereas the \ttt{DIST} expert classifier reports much higher accuracy on the tail classes. The \ttt{CLS} classifier achieves an accuracy of 73.7\% on the head classes, and the \ttt{DIST} expert classifier secures 43.1\% accuracy on the tail classes. 
We notice that by averaging the output of the classifiers, we are able to report good performance in both the majority and the minority classes. CIFAR-10 LT reaches an overall accuracy of 87.3\%, with 93.8\% on head classes and 85.7\% on tail classes. Similarly, with 72.8\% on the head and 31.0\% on the tail, DeiT-LT is able to secure an overall 54.8\% on CIFAR-100 LT. The results demonstrate that there is a parallel trend in the performance of experts for both CIFAR-10 LT and CIFAR-100 LT when $\rho$ is set to 50.

A similar trend is seen for large-scale ImageNet-LT and iNaturalist-2018 in Table \ref{deit-lt_tab:expert_performance_imnet}. For ImageNet-LT, the \ttt{CLS} expert classifier reports 68.3\% accuracy on the head classes, which is approximately 11\% more reported by the \ttt{DIST} expert classifier. At the same time, we observe that the \ttt{DIST} expert classifier is able to get an accuracy of 46.6\% on the tail, which is significantly higher than the 13.5\% of the \ttt{CLS} expert classifier. For iNaturalist-2018 as well, the \ttt{CLS} expert classifier achieves a high accuracy of 73.8\% on the head classes, and the \ttt{DIST} expert classifier reaches 77.0\% on the tail classes. After averaging the outputs of the two classifiers, DeiT-LT reports an overall accuracy of 59.1\% for ImageNet-LT and 75.1\% for iNaturalist-2018, which would not have been possible by training a standard Vision Transformer (ViT) with a single classifier.

\section{Comparison with CLIP based methods}
\label{deit-lt_suppl:clip}
Recently, some approaches such as VL-LTR \citep{tian2022vl} and PEL \citep{shi2023parameter} have adopted a pre-trained CLIP backbone to address long-tailed recognition challenges. As indicated originally, and also reinforced by \citep{xu2023demystifying}, CLIP is trained on large-scale balanced dataset (400 M Image-Text pair). As there is a lot of \emph{overlapping concepts between balanced CLIP data and long-tailed datasets (ImageNet-LT and iNat-18)}, the performance of the CLIP fine-tuned methods  \emph{does not indicate meaningful progress on long-tail learning tasks}, as CLIP has already seen tail concepts in abundance. Due to this \underline{unfairness} in training datasets used, we refrain from comparing the CLIP fine-tuned models (i.e., VL-LTR, PEL etc.) with DeiT-LT models trained from scratch.

\section{Visualization of Attention}
\label{deit-lt_suppl:vis_attention}
\begin{figure*}[ht]
    \centering
    \includegraphics[width=\textwidth]{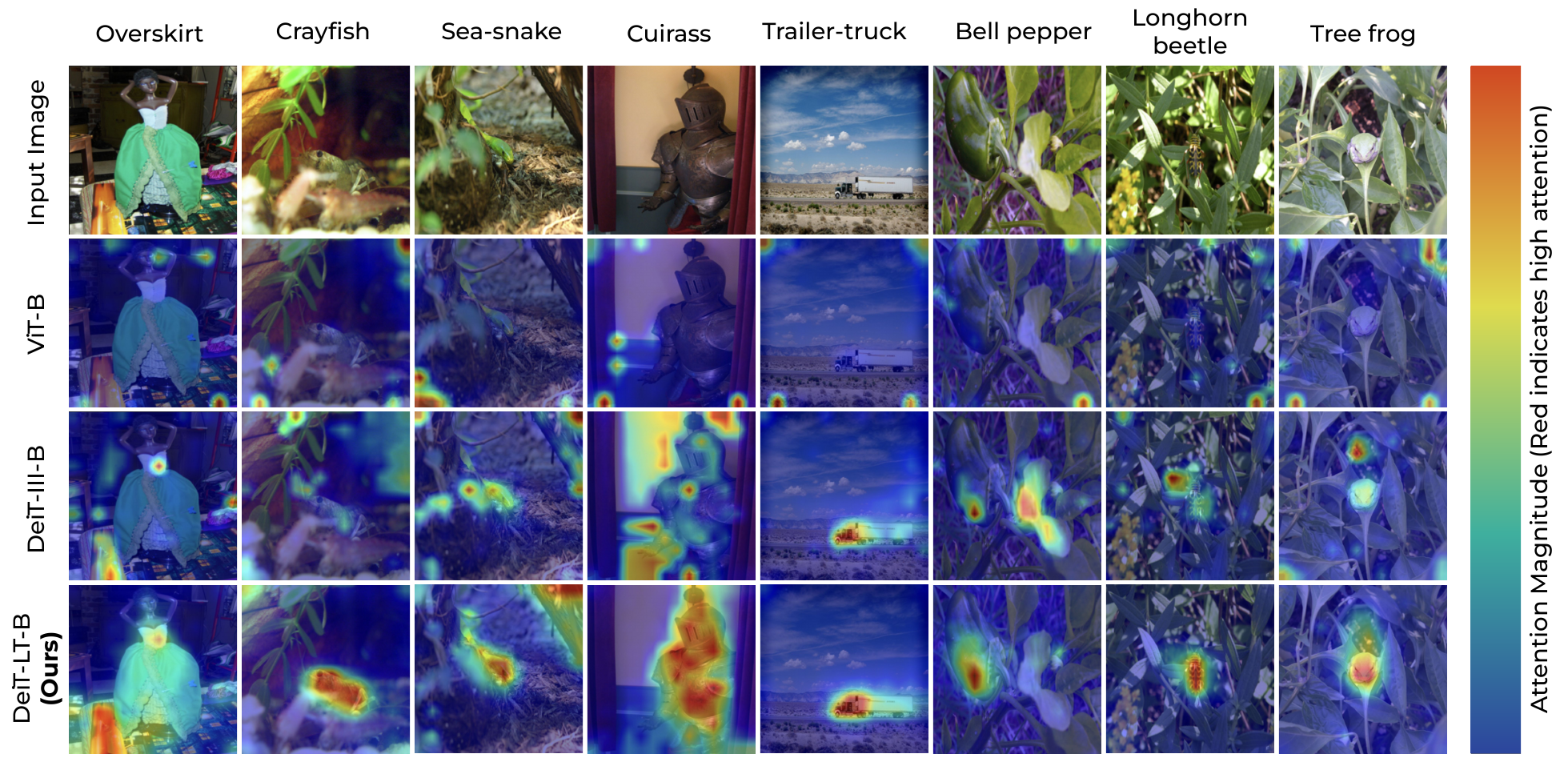}
    \caption{ Visual comparison of the attention maps from ViT-B, DeiT-III \cite{touvron2022deit} and DeiT-LT \emph{(ours)} on the ImageNet-LT dataset,  computed using the method of \emph{Attention Rollout} \cite{abnar2020quantifying}.}
    \label{deit-lt_fig:attention-vis-suppl}
\end{figure*}

To demonstrate the effect of distillation in DeiT-LT, we visualize the attention of baseline methods on ImageNet-LT without distillation (ViT and DeiT-III) and compare it with DeiT-LT. As DeiT-LT contains both the \ttt{DIST} token and the \ttt{CLS} token, for visualization we average the attention across both.  We use the Attention Rollout~\cite{tan2019attention} method for visualization. Fig.~\ref{deit-lt_fig:attention-vis-suppl} shows the result of attention for different methods. It can be clearly observed that DeiT-LT is able to localize attention at the correct position of objects, across almost all cases. We find that DeiT-III attention maps are better in comparison to ViT, but it also often gets confused (eg. Bell Pepper, Sea Snake etc.) compared to DeiT-LT.

\section{Statistical Significance of Experiments}
\label{deit-lt_suppl:signifance}
In this section, we present the results of our experiments on CIFAR-10 LT and CIFAR-100 LT ($\rho$ = 100, 50)(as in Table~\ref{deit-lt_tab:expert_performance_CF}), with three different random seeds. In Table~\ref{deit-lt_tab:expert_performance_CF}, we report the average performance of the expert classifiers along with the standard error for each. The low error demonstrates that the DeiT-LT training procedure is stable and quite robust across random seeds.

\section{Details on Local Connectivity Analysis}
\label{deit-lt_suppl:local_connectivity}
\begin{figure}[!t]
    \centering
    \includegraphics[width=0.6\columnwidth]{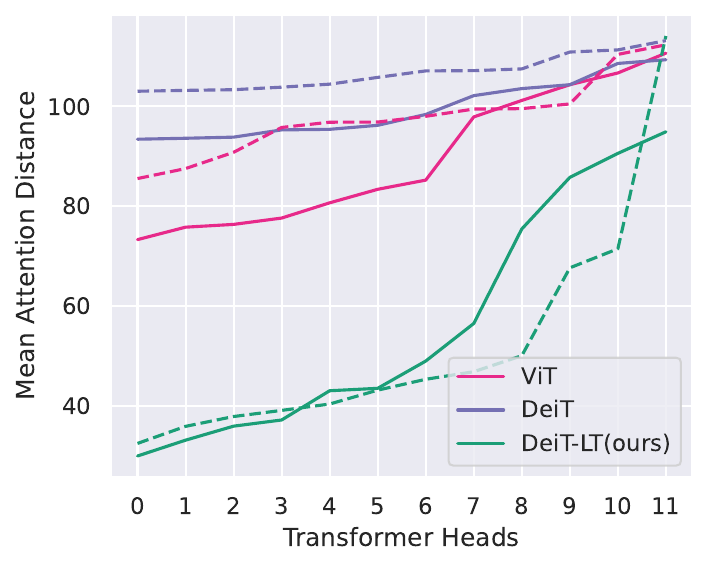}
    \caption{Mean attention distance for early blocks (1,2) for CIFAR-100 LT tail images.}
    \label{deit-lt_fig:local_cifar-100}
\end{figure}

We compute the mean attention distance for samples of tail classes (i.e. 7,8,9 class for CIFAR-10) using the method proposed by \citet{raghu2021vision}. For each head present in self-attention blocks, we calculate the distance of the patches it attends to. More specifically, we weigh the distance in the pixel space with the attention value and then average it. This is averaged for all the images present in the tail classes. We utilize the code provided here as our reference~\footnote{https://github.com/sayakpaul/probing-vits}. We show in Fig. ~\ref{deit-lt_fig:ViT_locality}
that for early blocks (1 and 2) of ViT, the proposed DeiT-LT method contains local features.
As we go from ViT to distilled DeiT to proposed DeiT-LT, we find that features become more local, which explains the generalizability of DeiT-LT for tail classes. To further confirm our observations, we also provide local connectivity plots for the tail classes of the CIFAR-100 dataset (Fig.~\ref{deit-lt_fig:local_cifar-100}). We observe that DeiT-LT produces highly local features. Further, we find that the DeiT baseline (Table~\ref{deit-lt_tab:cifar10_cifar100}), which is inferior to ViT for CIFAR-100, shows the presence of global features. Hence, the local connectivity correlates well with generalization on tail classes. The correlation of locality of features to generalization has also been observed by ~\cite{raghu2021vision}, who find that using the ImageNet-21k dataset for pre-training leads to more local and generalizable features in comparison to networks pre-trained on ImageNet-1k.

\begin{figure*}[t]
    \begin{subfigure}{0.49\textwidth}
    \centering
         \centering
         \includegraphics[width=\textwidth]{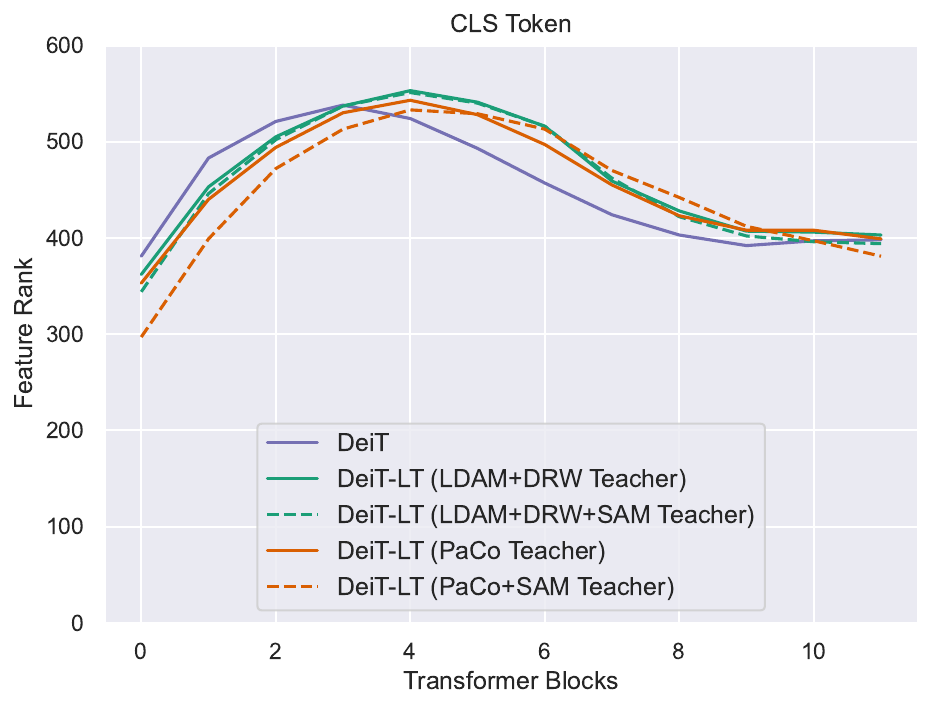}
         \caption{Rank of ViT from Distillation of CNN teachers using \texttt{CLS} token}
         \label{deit-lt_fig:supp_cls_vit_rank}
    
    \end{subfigure}%
   \begin{subfigure}{0.49\textwidth}
         \centering
         \includegraphics[width=\textwidth]{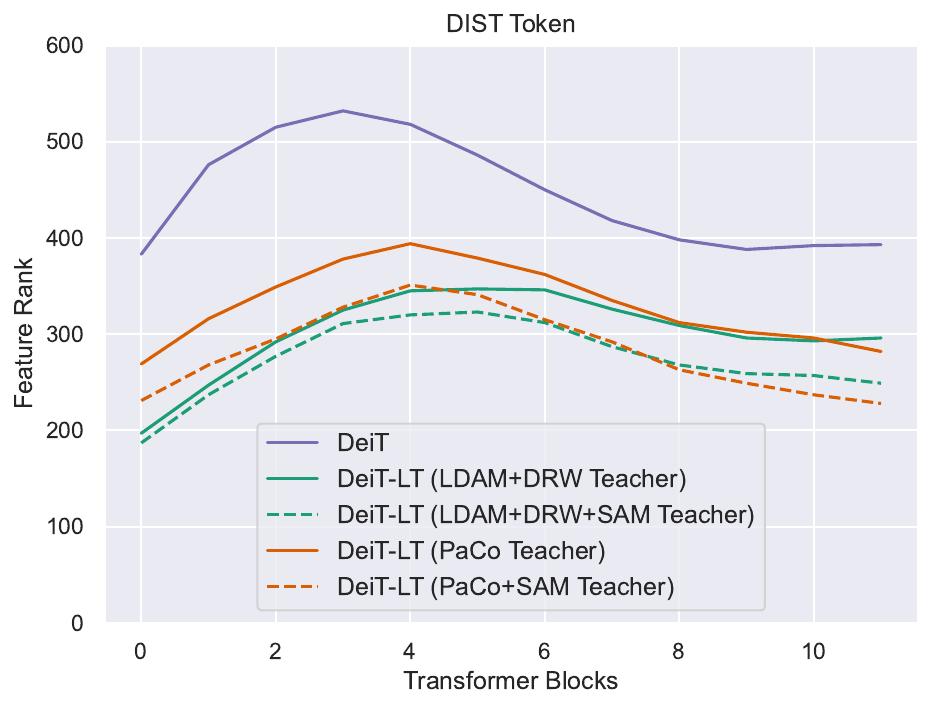}
         \caption{Rank of ViT from Distillation of CNN teachers using \texttt{DIST} token}
         \label{deit-lt_fig:supp_dist_ViT_rank}
    \end{subfigure} 
    \caption{We compare the rank calculated using features from the a) \texttt{CLS} token and b) \texttt{DIST} token when trained on CIFAR-10 LT. Our DeiT-LT captures both fine-grained features (from high-rank \texttt{CLS} token) and generalizable features (from low-rank \texttt{DIST} token).  }
    \label{deit-lt_fig:supp_rank}
\end{figure*}

\begin{figure*}[h]
\vspace{5mm}
\centering

\includegraphics[width=7cm,height=5.5cm]{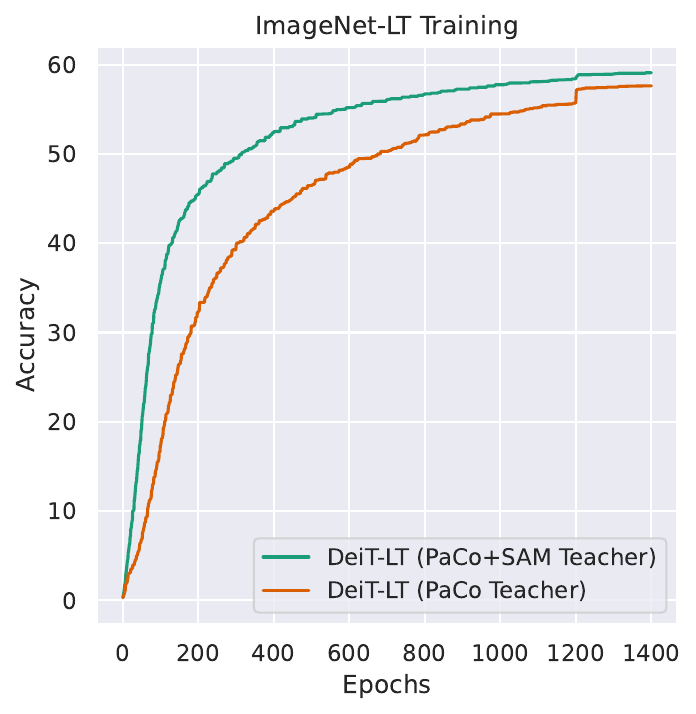}
\label{deit-lt_fig:test2}
\includegraphics[width=7cm,height=5.5cm]{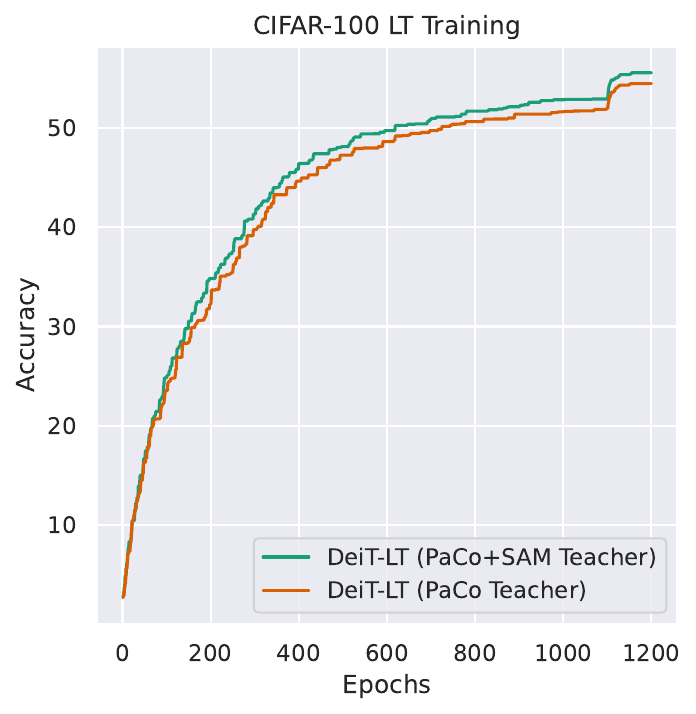}
\caption{Validation Accuracy Plots for the ImageNet-LT \emph{(left)} and CIFAR-100 LT \emph{(right)}.  DeiT-LT trained with SAM teachers converges faster than vanilla teachers.}
\vspace{5mm}
\label{deit-lt_fig:convg-plots}
\end{figure*}
\section{Distilling low-rank features}
\label{deit-lt_suppl:low_rank}
In our proposed method, as the \texttt{DIST} token serves as the expert on tail classes, it is important to ensure that it learns generalizable features for minority classes that are less prone to overfitting. As stated in \cite{andriushchenko2023sharpness}, training a network with SAM optimizer leads to low-rank features. In this subsection, we investigate the feature rank of the \texttt{DIST} token that is distilled via a SAM-based teacher.

\vspace{1mm} \noindent \textbf{Calculating Feature Rank.} Consider two sets of images \(\mathcal{X}_{all}, \mathcal{X}_{min} \subset \mathcal{X}\), where \(\mathcal{X}_{all}, \mathcal{X}_{min}\) refer to the set of images from all the classes and minority (tail) classes, respectively, with
\(\mathcal{X}\) being the set of all images.  We construct feature matrices \(F_{n_{h}, d}^{all}\) and \(F_{n_{t}, d}^{min}\), where \(n_h\) and \(n_t\) are the number of images in \(\mathcal{X}_{all}\) and \(\mathcal{X}_{min}\) respectively, and \(d\) is the dimension of the feature representation from \texttt{DIST} token.

Upon centering the columns of \(F_{n_{h}, d}^{all}\), we decompose the  feature matrix as \(U, S, V^T = \mathsf{SVD}(F_{n_{h}, d}^{all})\), and project \(F_{n_{t}, d}^{min}\) using the right singular vectors \(V\) as
\begin{equation*}
F_{proj}^{min}(k) = F_{n_{t}, d}^{min} * V_k
\end{equation*}

where \(V_k\) contains the top \(k\) singular vectors (principal componenets). We calculate our rank as the least \(k\) that satisfies

\begin{equation*}
\frac{{|| F_{n_{t}, d}^{min} - F_{recon}^{min}(k)||}^2 }{{|| F_{n_{t}, d}^{min} ||}^2} 	\le 0.01
\end{equation*}

where \(F_{recon}^{min}(k)\) is an approximate reconstructed feature matrix given by \(F_{recon}^{min} (k)= F_{proj}^{min}(k) * {V_k}^T\). \\

As shown in Fig.~\ref{deit-lt_fig:supp_dist_ViT_rank}, we find that the \texttt{DIST} token trained with a SAM-based teacher reports a lower rank. As we are able to use the same principal components to represent both the majority and minority classes' feature representation, it signifies that the \texttt{DIST} token learns generalizable characteristics relevant across different categories of images in an imbalanced dataset. By learning semantic similar features, our training of \texttt{DIST} token ensures good representation learning for minority classes by leveraging the discriminative features learned from majority classes. 

On the other hand, we observe that \texttt{CLS} token learns high-rank feature representations (Fig.~\ref{deit-lt_fig:supp_cls_vit_rank}), signifying that it captures intricately detailed information. Our DeiT-LT, thus, captures a wide range of information by using the predictions made using both fine-grained details from \texttt{CLS} token and generalizable features from \texttt{DIST} token.

\subsection{Convergence Analysis with SAM Teachers}
\label{deit-lt_suppl:sam_convergence}
We find that models distilled from the teachers trained using SAM~\cite{foret2020sharpness} converge faster than the usual CNN teachers. We provide the analysis for the Deit-LT(PaCo+SAM) and DeiT-LT(PaCo) on the ImageNet-LT and CIFAR-100 datasets in Fig.~\ref{deit-lt_fig:convg-plots}. We observe that models with SAM, coverage much faster, particularly for the ImageNet-LT dataset, demonstrating the increased convergence speed for the distillation. This can be attributed to the fact that low-rank models are simpler in structure and are much easier to distill to the transformer.

\section{Computation Requirement}

For training our proposed DeiT-LT method on CIFAR-10 LT and CIFAR-100 LT, we use two NVIDIA RTX 3090 GPU cards with 24 GiB memory each, with both datasets requiring about 15 hours to train to train the ViT student. We train the DeiT-LT student network on four NVIDIA RTX A5000 GPU cards for the large-scale ImageNet-LT dataset and on four NVIDIA A100 GPU cards for the iNaturalist-2018 dataset, in 61 and 63 hours, respectively.

 \renewcommand{\thesection}{F}

\supersection{Cost-Sensitive Self-Training for Optimizing \\ Non-Decomposable Metrics (Chapter-7)}

\section{Limitations and Negative Societal Impacts}
\label{csst_sec:limitations-and-negetive-societal}
\subsection{Limitations of our Work}
At this point we only consider optimizing objectives through \ttt{CSST} which can be written as a linear combination of entries of confusion matrix. Although there are important metrics like Recall, Coverage etc. which can be expressed as a linear form of confusion matrix. However, there do exist important metrics like Intersection over Union (IoU), Q-Mean etc. which we don't consider in the current work.
We leave this as an open direction for further work.

Also in this work we considered datasets where unlabeled data distribution doesn't significantly differ from the labeled data distribution, developing robust methods which can also take into account the distribution shift between unlabeled and labeled is an interesting direction for future work.

\subsection{Negative Societal Impact}
Our work has application in fairness domain~\cite{cotter2019optimization}, where it can be used to improve performance of minority sub-groups present in data. These fairness objectives can be practically enforced on neural networks through the proposed \ttt{CSST} framework.
However these same algorithms can be tweaked to artifically induce bias in decision making of trained neural networks, for example by ignoring performance of models on certain subgroups. Hence, we suggest deployment of these models after through testing on all sub groups of data.

\section{Connection between Minimization of Weighted Consistency Regularizer Loss (Eq. \eqref{csst_eq:wt-consistency-reg}) and 
Theoretical Weighted Consistency ($R_{\mB, w}(F)$ in Sec. \ref{csst_subsec:weighted_consistency})}
\label{csst_sec:appendix-conn-between}

The theoretical weighted consistency regularization can be approximated under strong augmentation $\mA$ as $R_{\mB, w}(F) \approx \sum_{i,j\in[K]}w_{ij}\ex[x \sim P_i]{\bone(F(\mA(x)) \neq F(x))}$. Noting that $\bone(F(\mA(x)) \neq F(x)) \le \bone(F(\mA(x)) \neq j) + \bone(F(x) \neq j)$ for any class label $j$, 
this value of $R_{\mB, w}(F)$ is bounded as follows:
\begin{align*}
    R_{\mB, w}(F) &\approx \sum_{i,j\in[K]}w_{ij}\ex[x \sim P_i]{\bone(F(\mA(x)) \neq F(x))}\\
    & \le 
    \sum_{i, j \in [K]} w_{ij}\ex[x \sim P_i]{\bone(F(\mA(x)) \neq j))} + 
\sum_{i, j \in [K]} w_{ij}\ex[x \sim P_i]{\bone(F(x) \neq j)}.
\end{align*}

Assuming $F(x)$ to be composed of a classifier made of neural network model output $p_{m}(x)$. For consistency regularization, we choose samples which have a low second term by thresholding on confidence for which we obtain the pseudo-label $F(x) = \hat{p}_{m}(x)$. In that case the minimization of (the upper bound of) $R_{\mB, w}(F)$ is approximately equivalent to cost-sensitive learning (CSL):
\begin{equation*}
\min_{F}\sum_{i, j \in [K]} w_{ij}\ex[x]{\bone(F(\mA(x)) \neq j))}.
\end{equation*}
The above objective can be equivalently expressed in form of gain matrix $\bG$ where $G_{i,j} = w_{i,j}/(\pi_{i})$. Hence, the above minimization can be effectively done for different gain matrices based on corresponding non-decomposable metric~\ref{csst_sec:ndo-loss-fun}. Following \citet{narasimhan2021training}, we can effectively minimize the $R_{\mB, w}(F)$ using the calibrated CSL loss function by plugging in the pseudo label $\hat{p}_m(x)$ in place of ground truth label ($\mathbf{y}$). The final weighted regularization function can be seen as:
\begin{equation}
    \ell^{hyb} (\hat{p}_{m}(x), p_{m}(\mA(x)); \bG)
\end{equation}
The above regularization loss is same as the weighted consistency regularization loss in Eq. \eqref{csst_eq:wt-consistency-reg}, which leads to minimization of $\mathcal{D}_{KL}(\norm(\bG^{\mathbf{T}}\hat{p}_{m}(x)) || p_{m}(\mathcal{A}(x))) \; \forall x \in \mX $ as seen in Prop. \ref{csst_prop:kl-weighted}. Hence, the minimization of theoretical consistency regularization term is achieved by minimizing the weighted consistency loss.

\section{Additional Examples and Proof of Theorem \ref{csst_thm:main-err-bd}}
\label{csst_sec:appendix-proofs}
In this section, we provide some examples for assumptions introduced in Sec. \ref{csst_sec:theoretical_results}
and a proof of Theorem \ref{csst_thm:main-err-bd}.
We provide proof of these examples in Sec. \ref{csst_sec:misc-proofs}.

\subsection{Examples for Theoretical Assumptions}
\label{csst_sec:appendix-examples-for-theoretical}
\addedtext{
The following example (Example \ref{csst_exa:c-exp-mix-gauss}) shows that the $c$-expansion (Definition \ref{csst_def:c-exp-def}) property is satisfied for mixtures of Gaussians and mixtures of manifolds.
\begin{example}
    \label{csst_exa:c-exp-mix-gauss}
    By \cite[Examples 3.4, 3.5]{wei2020theoretical},
    the $c$-expansion property is satisfied for mixtures of 
    isotropic Gaussian distributions and mixtures of manifolds.
    More precisely, in the case of mixtures of isotropic Gaussian distributions, 
    i.e., if $Q$ is given as mixtures of $\mN(\tau_i, \frac{1}{d}I_{d \times d})$
    for $i=1, \dots, n$ with some $n \in \ZZ_{\ge 1}$
    and $\tau_i \in \RR^d$,
    and $\mB(x)$ is an $\ell_2$-ball with radius $r$
    then by \cite[(13)]{bobkov1997isoperimetric} and \cite[Section B.2]{wei2020theoretical},
    $Q$ satisfies the $c$-expansion property
    with 
    $c(p) = R_h(p)/p$ for $p > 0$
    and $h = 2r\sqrt{d}$
    (c.f., \cite[section B.2]{wei2020theoretical}).
    Here $R_h(p) = \Phi(\Phi^{-1}(p) + h)$
    and $\Phi$ is the cumulative distribution function of the 
    standard normal distribution on $\RR$. 
\end{example}
}

In Sec. \ref{csst_subsec:weighted_consistency}, we required the assumption that $\gamma > 3$ (Assumption \ref{csst_assump:pseudo-labeler})
and remarked that it roughly requires $\err_w(\gpl)$ is ``small''.
The following example provides explicit conditions for $\err_w(\gpl)$ that satisfy the assumption using a toy example.
\begin{example}
    \label{csst_exa:gamma-3}
    Using a toy example provided in Example \ref{csst_exa:c-exp-mix-gauss}, 
    we provide conditions that satisfy the assumption $\gamma > 3$ approximately.
    To explain the assumption $\gamma > 3$, we assume that 
    $\modifiedtext{\Pw}$ is given as a mixture of isotropic Gaussians and $\mB(x)$ is $\ell_2$-ball with radius $r$ 
    as in Example \ref{csst_exa:c-exp-mix-gauss}.
    Furthermore, we assume that
    $|w|_1 = 1$ and $\err_w(\gstar)$ is sufficiently small compared to $\err_w(\gpl)$.
    Then, $p_w = \err_w(\gpl) + \err_w(\gstar) \approx \err_w(\gpl)$.
    Using this approximation,
    since $\modifiedtext{\Pw}$ satisfies the $c$-expansion property with $c(p) = R_{2r\sqrt{d}}(p)/p$,
    if $r = \frac{1}{2\sqrt{d}}$ then, the condition $\gamma > 3$ 
    is satisfied when $\err_w(\gpl) < 0.17$.
    If $r = \frac{3}{2\sqrt{d}}$ then, the condition $\gamma > 3$
     is satisfied when $\err_w(\gpl) < 0.33$.
\end{example}

In Assumption \ref{csst_assump:mu-small}, we assumed that both of $\err_w(\gstar)$ and $R_{\mB, w}(\gstar)$ are small.
The following example suggests the validity of this assumption. 
\begin{example}
    \label{csst_exa:r_gopt_bound}
    In this example, 
    we assume $w$ is a diagonal matrix $\diag(w_1, \dots, w_K)$.
    For simplicity, we normalize $w$ so that $\sum_{i\in [K]}w_i = 1$.
    As in \cite[Example 3.4]{wei2020theoretical}, we assume 
    that $P_i$ is given as isotropic Gaussian distribution $\mN(\tau_i, \frac{1}{d}I_{d\times d})$
    with $\tau_i \in \RR^d$ for $i = 1,\dots, K$ and $\mB(x)$ is an $\ell^2$-ball with radius $\frac{1}{2\sqrt{d}}$.
    Furthermore, 
    we assume $\inf_{1 \le i < j \le K}\|\tau_i - \tau_j\|_2 \gtrsim \frac{\sqrt{\log d}}{\sqrt{d}}$
    and $\sup_{i, j \in [K]}\frac{w_i}{w_j} = o(d)$, where the latter assumption is valid for high dimensional 
    datasets (e.g., image datasets).
    Then it can be proved that there exists a classifier $F$
    such that $R_{\mB, w}(F) = O(\frac{1}{d^c})$ and $ \err_w(F) = O(\frac{1}{d^c})$,
    where $c > 0$ is a constant (we can take $F$ as the Bayes-optimal classifier for $\err_w$).
    Thus, this suggests that Assumption \ref{csst_assump:mu-small} is valid for 
    datasets with high dimensional instances.
\end{example}

\addedtext{The statement of Example \ref{csst_exa:gamma-3} follows from numerical computation of $R_{2r\sqrt{d}}(p)/p$. We provide proofs of Examples \ref{csst_exa:c-exp-mix-gauss} and \ref{csst_exa:r_gopt_bound} in Sec. \ref{csst_sec:misc-proofs}.}

\subsection{Proof of Theorem \ref{csst_thm:main-err-bd} Assuming a Lemma}
Theorem \ref{csst_thm:main-err-bd} can be deduced from the following lemma (by taking $H = \gstar$ and 
$\mL_{Q, H}(\widehat{F}) \le \mL_{Q, H}(\gstar)$),
which provides a similar result to \cite[Lemma A.8]{wei2020theoretical}.
\begin{lemma}
    \label{csst_lem:err-bd}
    Let $H$ be a classifier and 
    $Q$ a probability measure on $\mX$ satisfying $c$-expansion property.
    We put $\gamma_H = c(Q(\{x \in \mX: \gpl(x) \ne H(x)\}))$.
    For a classifier $F$, we define $\mS_{\mB}(F)$ by 
    \begin{math}
    \mS_{\mB}(F) = \{x \in \mX: F(x) = F(x') \quad \forall x' \in \mB(x)\}.
    \end{math}
    For a classifier $F$, we define $\mL_{Q, H}(F)$ by
    \begin{multline*}
        \frac{\gamma_H + 1}{\gamma_H - 1}
        Q(\{x \in \mX : F(x) \ne \gpl(x)\})\\
        + \frac{2\gamma_H}{\gamma_H - 1}
        Q(\mS_{\mB}^c(F)) + 
        \frac{2\gamma_H}{\gamma_H - 1}Q(\mS_\mB^c(H))
         - Q(\{x \in \mX: \gpl(x) \ne H(x)\}),
    \end{multline*}
    where $\mS_\mB^c(F)$ denotes the complement of $\mS_\mB(F)$.
    Then, we have $Q(\{x \in \mX: F(x) \ne H(x)\}) \le \mL_{Q, H}(F)$
    for any classifier $F$.
\end{lemma}
In this subsection,  we provide a proof of Theorem \ref{csst_thm:main-err-bd}
assuming Lemma \ref{csst_lem:err-bd}.
We provide a proof of the lemma in the next subsection.
For a classifier $F$, we define $\mM(F)$ as $\{x \in \mX: F(x) \ne \gstar(x)\}$
and $\mpl(F)$ as $\{x \in \mX: F(x) \neq \gpl(x)\}$.
We define $\widetilde{\mL}_w(F)$ by 
\begin{align*}
    \widetilde{\mL}_w(F)
    &= \mL_w(F) +  
    \frac{2\gamma}{\gamma - 1} R_{\mB, w}(\gstar) - \modifiedtext{\Pw}(\{x \in \mX: \gpl(x) \ne \gstar(x)\}).
\end{align*}
We note that $\widetilde{\mL}_w(F) - \mL_w(F)$ does not depend on $F$.

\begin{proof}[Proof of Theorem \ref{csst_thm:main-err-bd}]
    We let $Q = \modifiedtext{\Pw}$ and $H = \gstar$ in Lemma \ref{csst_lem:err-bd}
    and denote $\gamma_H$ in the lemma by $\gamma'$.
    Since $w_{ij}\ge 0$ and $\ex[x \sim P_i]{\indc(\gpl(x) \ne \gstar(x))} 
    \le \ex[x \sim P_i]{\indc(\gpl(x) \ne j)} + \ex[x \sim P_i]{\indc(\gstar(x) \ne j)}$
     for any $i, j$, we have the following: 
    \begin{align*}
        |w|_1 \modifiedtext{\Pw}(\mM(\gpl)) &= \sum_{i, j\in[K]}w_{ij}\ex[x \sim P_i]{\indc(\gpl(x) \ne \gstar(x))}\\
        &\le \sum_{i, j \in [K]}w_{ij}
        \left\{\ex[x \sim P_i]{\indc(\gpl(x) \ne j)} + \ex[x \sim P_i]{\indc(\gstar(x) \ne j)}\right\}\\
        &= \err_w(\gpl) + \err_w(\gstar).
    \end{align*}
    Here, the first equation follows from the definition of $\Pw$.
    Thus, we obtain $\modifiedtext{\Pw}(\mM(\gpl)) \le p_w$.
    By definition of $\gamma$ and $\gamma'$
    ($\gamma = c(p_w)$ and $\gamma' = c\left(\Pw(\mM(\gpl))\right)$)
    and the assumption that $c$ is non-increasing, we have the following: \begin{equation}
        \gamma \le \gamma'.
        \label{csst_eq:gamma-gamma'}
    \end{equation}
    We note that 
    \begin{align}
        \err_w(F) 
        &= \sum_{i, j\in [K]}
        w_{ij}\ex[x\sim P_i]{\indc(F(x) \ne j)}\notag \\
        &\le 
        \sum_{i, j \in [K]} 
        w_{ij} \ex[x\sim P_i]{\indc(F(x) \ne \gstar(x))}
        + 
        \sum_{i, j \in [K]} 
        w_{ij} \ex[x\sim P_i]{\indc(\gstar(x) \ne j)}\notag \\
        &= |w|_1 \modifiedtext{\Pw}(\mM(F)) + \err_w(\gstar).
        \label{csst_eq:errw_pw_errw_gstar}
    \end{align}
    Here the first inequality holds since 
    $\indc(F(x) \neq j) \le \indc(F(x) \neq \gstar(x)) + \indc(\gstar(x) \neq j)$
    for any $x$ and $j$.
    By \eqref{csst_eq:errw_pw_errw_gstar} and Lemma \ref{csst_lem:err-bd}, the error is upper bounded as follows:
    \begin{multline*}
        \err_w(F)
        \le 
        \err_w(\gstar) + 
        \frac{\gamma' + 1}{\gamma' - 1} |w|_1 \modifiedtext{\Pw}(\mpl(F)) \\
        +\frac{2\gamma'}{\gamma' - 1} |w|_1 \modifiedtext{\Pw} (\mS_\mB^c(F)) +
        \frac{2\gamma'}{\gamma' - 1}|w|_1 \modifiedtext{\Pw}(\mS_\mB^c(\gstar)) - |w|_1 \modifiedtext{\Pw}(\mM(\gpl)).
    \end{multline*}
    By \eqref{csst_eq:gamma-gamma'}, we obtain
    \begin{multline}
        \label{csst_eq:errw-lem-applied}
        \err_w(F)
        \le 
        \err_w(\gstar) + 
        \frac{\gamma + 1}{\gamma - 1} |w|_1 \modifiedtext{\Pw}(\mpl(F)) \\
        +\frac{2\gamma}{\gamma - 1} |w|_1 \modifiedtext{\Pw} (\mS_\mB^c(F)) +
        \frac{2\gamma}{\gamma - 1}|w|_1 \modifiedtext{\Pw}(\mS_\mB^c(\gstar)) - |w|_1 \modifiedtext{\Pw}(\mM(\gpl)).
    \end{multline}
    By definition of $\mL_w$ and letting $F = \widehat{F}$, we have the following:
    \begin{align*}
        & \err_w(\widehat{F})
        \le \err_w(\gstar)  + \mL_w(\widehat{F}) + \frac{2\gamma}{\gamma - 1} R_{\mB, w}(\gstar) - |w|_1 \modifiedtext{\Pw}(\{x \in \mX: \gpl(x) \ne \gstar(x)\})\\
        &\le \err_w(\gstar) + \mL_w(\gstar) + \frac{2\gamma}{\gamma - 1} R_{\mB, w}(\gstar) - |w|_1 \modifiedtext{\Pw}(\{x \in \mX: \gpl(x) \ne \gstar(x)\})\\
        &= \err_w(\gstar) + \frac{2}{\gamma - 1}|w|_1 \modifiedtext{\Pw}(\mM(\gpl)) + \frac{4\gamma}{\gamma - 1} R_{\mB, w}(\gstar)\\
        &\le \err_w(\gstar) + \frac{2}{\gamma - 1}(\err_w(\gpl) + \err_w(\gstar)) + \frac{4\gamma}{\gamma - 1}R_{\mB, w}(\gstar)\\
        & = 
        \frac{2}{\gamma - 1}
        \err_w(\gpl)
        + \frac{\gamma + 1}{\gamma - 1} \err_w(\gstar)
        + \frac{4\gamma}{\gamma - 1} R_{\mB, w}(\gstar).
    \end{align*}
    Here, the second inequality holds since $\widehat{F}$ is a minimizer of $\mL_w$,
    the third inequality follows from 
    $\indc(\gstar(x) \ne \gpl(x)) \le \indc(\gstar(x) \ne j) + \indc(\gpl(x) \ne j)$ for any $j$.
    Thus, we have the assertion of the theorem.
\end{proof}

\subsection{Proof of Lemma \ref{csst_lem:err-bd}}
We decompose $\mM(F) \cap \mS_\mB(F) \cap \mS_\mB(H)$ into the following three sets:
\begin{align*}
    \mN_1 &= \{x \in \mS_\mB(F) \cap \mS_\mB(H): F(x) = \gpl(x), \text{ and } \gpl(x) \ne H(x)\},\\
    \mN_2 &= \{x \in \mS_\mB(F) \cap \mS_\mB(H): F(x) \ne \gpl(x), \gpl(x) \ne H(x), \text{ and } F(x) \ne H(x)\},\\
    \mN_3 &= \{x \in \mS_\mB(F) \cap \mS_\mB(H): F(x) \ne \gpl(x) \text{ and } \gpl(x) = H(x)\}.
\end{align*}

\begin{lemma}
    \label{csst_lem:incl}
    Let $S = \mS_\mB(F) \cap \mS_\mB(H)$
    and $V = \mM(F) \cap \mM(\gpl)\cap S$.
    Then, we have $\mN(V) \cap \mM^c(F) \cap S = \emptyset$
    and $\mN(V) \cap \mM^c(\gpl) \cap S \subseteq \mpl(F)$.
    Here $\mpl(F)$ is defined as $\{x \in \mX : F(x) \neq \gpl(x)\}$.
\end{lemma}
\begin{proof}
    We take any element $x$ in $\mN(V) \cap S$.
    Since $x \in \mN\left(V\right)$ and definition of neighborhoods,
    there exists $x' \in \mM(\gpl) \cap \mM(F) \cap S$ such that 
    $\mB(x) \cap \mB(x') \neq \emptyset$.
    Since $x, x' \in \mS_\mB(F)$, 
    $F$ takes the same values on $\mB(x)$ and $\mB(x')$.
    By $\mB(x) \cap \mB(x') \neq \emptyset$, $F$ takes the same value on $\mB(x) \cup \mB(x')$.
    It follows that $F(x) = F(x')$.
    Since we have $x, x' \in \mS_\mB(H)$, similarly, we see that $H(x) = H(x')$.
    By $x' \in \mM(F)$, we have $F(x) = F(x') \neq H(x') = H(x)$.
    Therefore, we have $x \in \mM(F)$.
    Thus, we have proved that 
    $x \in \mN(V) \cap S$ implies $x \in \mM(F)$, i.e.,
    $ \mN(V) \cap S \subseteq \mM(F)$.
    Therefore, we have $\mN(V) \cap \mM^c(F) \cap S = \emptyset$.
    This completes the proof of the first statement.
    Next, we assume 
    $x \in \mN(V) \cap \mM^c(\gpl) \cap S$.
    Then, we have $F(x)\neq H(x)$ and $\gpl(x) = H(x)$.
    Therefore, we obtain $F(x) \neq \gpl(x)$.
    This completes the proof.
\end{proof}

\begin{lemma}
    \label{csst_lem:n1_cup_n2}
    Suppose that assumptions of Lemma \ref{csst_lem:err-bd} hold.
    We define $q$ as follows:
    \begin{equation}
        \label{csst_eq:def-of-q}
        q = \frac{Q(\mpl(F) \cup \mS_\mB^c(F) \cup \mS_\mB^c(H))}{\gamma_H - 1}.
    \end{equation}
    Then, we have $Q(\mS_\mB(F) \cap \mS_\mB(H) \cap \mM(\gpl) \cap \mM(F)) \le q$.
    In particular, noting that 
    $\mN_1 \cup \mN_2 \subseteq \mS_\mB(F) \cap \mS_\mB(H) \cap \mM(\gpl)\cap \mM(F)$, 
    we have $Q(\mN_1 \cup \mN_2) \le q$.
\end{lemma}
\begin{proof}
    We let $S = \mS_\mB(F) \cap \mS_\mB(H)$ and $V = \mM(F) \cap \mM(\gpl)\cap S$ as before.
    Then by Lemma \ref{csst_lem:incl}, we have
    \begin{align*}
        \mN(V) \cap V^c \cap S &= 
        \left(\mN(V) \cap \mM^c(F) \cap S\right)\cup
        \left(\mN(V) \cap \mM^c(\gpl) \cap S\right)\\
        &\subseteq
        \emptyset \cup \mpl(F) = \mpl(F).
    \end{align*}
    Therefore, we have
    \begin{align*}
        \mN(V) \cap V^c &= \mN(V) \cap V^c \cap (S \cup S^c)\\
        &= \left(\mN(V)  \cap V^c \cap S \right)
        \cup \left(
            \mN(V)  \cap V^c \cap S^c
        \right)\\
        &\subseteq 
        \mpl(F) \cup S^c.
    \end{align*}
    Thus, by the $c$-expansion property, we have
    \begin{align*}
        Q(\mpl(F) \cup S^c) &\ge Q(\mN(V) \cap V^c)\\
        & \ge Q(\mN(V)) - Q(V)\\
        & \ge \left(c(Q(V)) - 1\right)Q(V).
    \end{align*}
    Since $V \subseteq \mM(\gpl)$, $c$ is non-increasing, and $\gamma_H > 1$,
    we have $Q(V) \le Q(\mpl(F) \cup S^c)/(\gamma_H - 1) \le q$.
    This completes the proof.
\end{proof}

The following lemma provides an upper bound of $Q(\mN_3)$.
\begin{lemma}
    \label{csst_lem:m3-upper-bd}
    Suppose that the assumptions of Lemma \ref{csst_lem:err-bd} hold.
    We have 
    \begin{equation*}
        Q(\mN_3) \le q + Q(\mS_\mB^c(F) \cup \mS_\mB^c(H)) + Q(\mpl(F)) - Q(\mM(\gpl)),
    \end{equation*}
    where $q$ is defined by \eqref{csst_eq:def-of-q}.
\end{lemma}
\begin{proof}
    We let $S = \mS_\mB(F) \cap \mS_\mB(H)$.
   First, we prove 
   \begin{equation}
    \label{csst_eq:m3_m1_disjoint_union}
       \mN_3 \sqcup \left(\mpl^c(F) \cap S \right) 
       = \mN_1 \sqcup \left(\mM^c(\gpl) \cap S \right).
   \end{equation} 
   Here, for sets $A, B$, we denote union $A \cup B$ by $A \sqcup B$ if the union is disjoint. 
   By definition, we have $\mN_1 = S \cap \mpl^c(F) \cap \mM(\gpl)$
   and $\mN_3 = S \cap \mpl(F) \cap \mM^c(\gpl)$.
   Thus, we have
   \begin{align*}
       &\mN_3 \cup \left(\mpl^c(F) \cap S \right)\\
       &= \left(
        S \cap \mpl(F) \cap \mM^c(\gpl)
       \right)  \cup \left(\mpl^c(F) \cap S \right)\\
       &= S \cap 
       \left\{
       \left(\mpl(F)\cap \mM^c(\gpl) \right)
       \cup \mpl^c(F)
       \right\}\\
       &= S\cap \left(\mM^c(\gpl) \cup \mpl^c(F) \right).
   \end{align*}
   Similarly, we have the following:
   \begin{align*}
       &\mN_1 \cup     \left(\mM^c(\gpl) \cap S\right)\\
       &= \left(
        S \cap \mpl^c(F) \cap \mM(\gpl)
       \right)
       \cup
     \left(\mM^c(\gpl) \cap S \right)\\
       &= S \cap \left(\mM^c(\gpl) \cup \mpl^c(F) \right).
   \end{align*}
   Since disjointness is obvious by definition, we obtain \eqref{csst_eq:m3_m1_disjoint_union}.
   Next, we note that the following holds:
   \begin{align}
      Q(\mpl^c(F) \cap S) &= Q(\mpl^c(F)) - Q(\mpl^c(F) \cap S^c) \notag \\
      & \ge Q(\mpl^c(F)) - Q(S^c).
      \label{csst_eq:lem-m3-upper-bd1}
   \end{align}
   By \eqref{csst_eq:m3_m1_disjoint_union}, we obtain the following:
   \begin{align*}
       Q(\mN_3) &= Q(\mN_1) + Q(\mM^c(\gpl) \cap S)
       - Q(\mpl^c(F) \cap S)\\
       &\le Q(\mN_1) + Q(\mM^c(\gpl)) - Q(\mM^c(\gpl) \cap S)\\
       & \le Q(\mN_1) + Q(\mM^c(\gpl)) - Q(\mpl^c(F)) + Q(S^c)\\
       & \le q + Q(\mM^c(\gpl)) - Q(\mpl^c(F)) +Q(S^c)\\
       & = q - Q(\mM(\gpl)) + Q(\mpl(F))  + Q(S^c).
   \end{align*} 
   Here the second inequality follows from \eqref{csst_eq:lem-m3-upper-bd1}
   and the third inequality follows from Lemma \ref{csst_lem:n1_cup_n2}.
   This completes the proof.
\end{proof}

Now, we can prove Lemma \ref{csst_lem:err-bd} as follows.

\begin{proof}[Proof of Lemma \ref{csst_lem:err-bd}]
   \begin{align*}
       Q(\mM(F)) 
       &= Q(\mM(F) \cap \mS_\mB(F) \cap \mS_\mB(H)) + 
       Q\left(\mM(F) \cap \left(\mS_\mB^c(F) \cup \mS_\mB^c(H) \right)\right)\\
       &\le 
       Q(\mN_1 \cup \mN_2) + Q(\mN_3) + Q(\mS_\mB^c(F) \cup \mS_\mB^c(H))\\
       &\le 2q + 2Q(\mS_\mB^c(F) \cup \mS_\mB^c(H)) + Q(\mpl(F)) - Q(\mM(\gpl)).
   \end{align*} 
   Here, the last inequality follows from Lemmas \ref{csst_lem:n1_cup_n2} and \ref{csst_lem:m3-upper-bd}.
   
   Since $q$ satisfies $q \le \frac{Q(\mpl(F)) + Q(\mS_\mB^c(F) + Q(\mS_\mB^c(H)))}{\gamma_H - 1}$
   by \eqref{csst_eq:def-of-q}, we have our assertion.
\end{proof}
\subsection{Miscellaneous Proofs for Examples}
\label{csst_sec:misc-proofs}
\begin{proof}[Proof of Example \ref{csst_exa:c-exp-mix-gauss}]
    In Example \ref{csst_exa:c-exp-mix-gauss}, we stated that 
    $p \mapsto R_h(p)/p$ is non-increasing.
    This follows from the concavity of $R_h$
    and $\lim_{p \rightarrow +0} R_h(p) = 0$.
    In fact, we can prove the concavity of $R_h$
    by $\frac{d^2R_h}{dp^2}(p) = - h \exp\left(\frac{\xi^2}{2} - h\xi - \frac{1}{2}h^2\right) \le 0$,
    where $\xi = \Phi^{-1}(p)$.
\end{proof}

\begin{proof}[Proof of Example \ref{csst_exa:r_gopt_bound}]
    For each $i, j \in [K]$, $w_i P_i(x) \ge w_j P_j(x)$ is equivalent to
    $(x - \tau_i) \cdot v_{ji} \le \frac{\|\tau_i - \tau_j\|}{2} +
     \frac{2 (\log w_i - \log w_j)}{d \|\tau_i - \tau_j\|}$,
     where $v_{ji} = \frac{\tau_j - \tau_i}{\|\tau_i - \tau_j\|}$.
    Thus, for each $i \in [K]$,
    we have $\bigcap_{j \in [K]\setminus \{i\}}X_{ij} \subseteq \mS_\mB(\gopt)$.
    Here $X_{ij}$ is defined as $\{x \in \mX : (x - \tau_i)\cdot v_{ji}  \le 
    \frac{\|\tau_i - \tau_j\|}{2} + 
    \frac{2 (\log w_i - \log w_j)}{d \|\tau_i - \tau_j\|} - \frac{r}{2}
    \}$.
    For any $w \in \RR^d$ with $\|w\|_2 = \sqrt{d}$ and $a > 0$,
    we have $P_i(\{x \in \mX: (x - \tau_i)\cdot w > a\}) = 1- \Phi(a) \le \frac{1}{2}\exp(-a^2/2)$
    (c.f., \cite{chiani2003new}).
    Thus, $P_i(X_{ij}^c) \le \frac{1}{2}\exp(-d a_{ij}^2/2)$, where 
    $a_{ij} = \frac{\|\tau_i - \tau_j\|}{2} + 
    \frac{2 (\log w_i - \log w_j)}{d \|\tau_i - \tau_j\|} - \frac{r}{2}$.
    By assumptions, we have $a_{ij}\sqrt{d} \gtrsim \sqrt{\log d}$.
    Therefore, $P_i(X_{ij}^c) = O(\frac{1}{poly(d)})$.
    It follows that
    \begin{math}
        P_i(\mS_\mB^c(\gopt)) \le \sum_{j\in [K]\setminus \{i\}} P_i(X_{ij}^c)
        = O(\frac{1}{poly(d)})
    \end{math}.
    Thus, we have $R_{\mB, w}(\gopt) = O(\frac{1}{poly(d)})$.
    By the same way, we can prove that $\err_w(\gopt) = O(\frac{1}{poly(d)})$.
\end{proof}

\section{All-Layer Margin Generalization Bounds}
\label{csst_sec:appendix-all-layer}
Following \cite{wei2020theoretical,wei2019improved}, we introduce 
all layer margin of neural networks and provide generalization bounds of \csst{}. In this section, we assume that classifier $F(x)$ is given as $F(x) = \argmax_{1\le i \le K} \Phi_i(x)$,
where $\Phi$ is a neural network of the form
\begin{equation*}
\Phi(x) = (f_p \circ f_{p-1} \circ \cdots \circ f_1) (x).
\end{equation*}
Here $f_i : \RR^{d_{i-1}} \rightarrow \RR^{d_{i}}$ with $d_0 = d$ and $d_p = K$.
We assume that each $f_i$ belongs to a function class $\mF_i \subset \mathrm{Map}(\RR^{d_{i-1}}, \RR^{d_{i}})$.
We define a function class $\mF$ to which $\Phi$ belongs 
by 
\begin{equation*}
  \mF = \{\Phi: \RR^{d} \rightarrow \RR^K : \Phi(x) = (f_p \circ \dots \circ f_1)(x), \quad f_i \in \mF_i,  
  \forall i \}.
\end{equation*}
For example, for $b > 0$, 
$\mF_i$ is given as $\{h \mapsto W \phi(h) : W \in \RR^{d_{i-1} \times d_{i}}, \|W\|_\fro \le b\}$ if $i > 1$
and $\{h \mapsto W h : W \in \RR^{d_0 \times d_{1}}, \|W\|_\fro \le b\}$ if $i = 1$,
where $\phi$ is a link function (applied on $\RR^{d_i}$ entry-wise) with bounded operator norm
(i.e., $\|\phi\|_\op := \sup_{x \in \RR^{d_i} \setminus \{0\}}\|\phi(x)\|_2/\|x \|_2 < \infty$) 
and $\|W\|_\fro$ denotes the Frobenius norm of the matrix.
However, we do not assume the function class $\mF_i$ does not have this specific form.
We assume that each function class $\mF_i$ is a normed vector space with norm $\| \cdot \|$.
In the example above, we consider the operator norm, i.e., if $f(h) = \phi(Wh)$, $\|f\|$ is defined 
as $\| f\|_\op$.
Let $x_1, \dots, x_n$ be a finite i.i.d. sequence of samples drawn from $\modifiedtext{\Pw}$.
We denote the corresponding empirical distribution by $\pwhat$, i.e.,
for a measurable function $f$ on $\mX$, $\ex[x \sim \pwhat]{f} = \sum_{i=1}^{n}f(x_i)$.

For $\xi = (\xi_1, \dots, \xi_p) \in \prod_{i=1}^p\RR^{d_i}$, we define the perturbed output 
$\Phi(x, \xi)$ as $\Phi(x, \xi) = h_p(x, \xi)$, where
\begin{align*}
    h_1(x, \xi) &= f_1(x) + \xi_1 \|x \|_2,\\ 
    h_i(x, \xi) &= f_i(h_{i-1}(x, \xi)) + \xi_{i} \|h_{i-1}(x, \xi)\|_2, \quad \text{for } 2 \le i \le p.
\end{align*}
Let $x \in \mX$ and $y \in [K]$. 
We define $\Xi(\Phi, x, y)$ by $\{\xi \in \prod_{i=1}^p \RR^{d_i}: \argmax_{i}\Phi_i(x, \xi) \neq y\}$.
Then, the all-layer margin $m(\Phi, x, y)$ is defined as 
\begin{equation*}
    m(\Phi, x, y) = \min_{\xi \in \Xi(\Phi, x, y)}  \|\xi\|_2,
\end{equation*}
where $\|\xi\|_2$ is given by $\sqrt{\sum_{i=1}^p\|\xi_i\|^2_2}$.
Following \cite{wei2020theoretical}, we define 
a variant of the all-layer margin that measures 
robustness of $\Phi$ with respect to input transformations defined by $\mB(x)$ as follows:
\begin{equation*}
    m_{\mB}(\Phi, x) := \min_{x' \in \mB(x)} m(F, x', \argmax_{i} \Phi_i(x)).
\end{equation*}
\begin{assumption}[c.f. \cite{wei2019improved}, Condition A.1]
    \label{csst_assump:cov-cond}
    Let $\mG$ be a normed space with norm $\| \cdot \|$
    and $\epsilon > 0$.
    We say $\mG$  satisfies the $\epsilon^{-2}$ covering condition with complexity $\mcnorm(\mG)$ if 
    for all $\epsilon > 0$, we have
    \begin{equation*}
        \log \mnnorm(\epsilon, \mG)  \le \frac{\mcnorm(\mG)}{\epsilon^2}.
    \end{equation*}
    Here $\mnnorm(\epsilon, \mG)$ the $\epsilon$-covering number of $\mG$.
    We assume function class $\mF_i$ satisfies the $\epsilon^{-2}$ covering condition with complexity $\mcnorm(\mF_i)$
    for each $1 \le i \le p$.
\end{assumption}
Throughout this section, we suppose that Assumption \ref{csst_assump:cov-cond} holds.
Essentially, the following two propositions follows were proved by \citet{wei2020theoretical}:
\begin{proposition}[c.f., \cite{wei2020theoretical}, Lemma D.6]
    \label{csst_prop:rw-finite-bd}
   With probability at least $1-\delta$ over the draw of the training data, for all $t \in (0, \infty)$,
   any $\Phi \in \mF$ satisfies the following:
   \begin{equation*}
       R_{\mB, w}(F) = \ex[\pwhat]{\bone(m_{\mB}(\Phi, x) \le t)} +
       \otilde\left(\frac{\sum_{i=1}^p \mcop(\mF_i)}{t \sqrt{n}}\right) +
       \zeta,
   \end{equation*} 
   where $\zeta = O\left(\sqrt{\frac{\log(1/\delta)+ \log n}{n}}\right)$ is a lower order term
   and $F(x) = \argmax_{i \in [K]}\Phi_i(x)$.
\end{proposition}

\begin{proposition}[c.f., \cite{wei2020theoretical}, Theorem D.3]
    \label{csst_prop:lw-finite-bd}
    With probability at least $1-\delta$  over the draw of the training data,
    for all $t \in (0, \infty)$, any $\Phi \in \mF$ satisfies the following:
    \begin{equation*}
        L_w(F, \gpl) = \ex[\pwhat]{\bone(m(\Phi, x, \gpl(x)) \le t)} + 
        \otilde\left(\frac{\sum_{i=1}^p \mcop(\mF_i)}{t \sqrt{n}} \right) +
        \zeta,
    \end{equation*}
   where $\zeta = O\left(\sqrt{\frac{\log(1/\delta)+ \log n}{n}}\right)$ is a lower order term
   and $F(x) = \argmax_{i \in [K]}\Phi_i(x)$.
\end{proposition}
\begin{remark}
    Although we have proved Theorem \ref{csst_thm:main-err-bd} following \citep{wei2020theoretical},
    we had to provide our own proof due to some differences in theoretical assumptions
    (e.g., in our case there does not necessarily exist the ground-truth classifier, 
    although they assumed the expansion property for each $P_i$,
    we assume the expansion property for the weighted probability measure).
    On the other hand, the proofs of \citep[Lemma D.6]{wei2020theoretical}
    and \citep[Theorem D.3]{wei2020theoretical} work for any distribution $P$ on $\mX$ and its empirical distribution 
    $\widehat{P}$. 
    Since $\|w\|_1 \modifiedtext{\Pw}(\mS^c_{\mB}(F)) = R_{\mB, w}(F)$ and 
    $\|w\|_1 \modifiedtext{\Pw}(\{x: F(x) \neq \gpl(x)\}) = L_w(F, \gpl)$,
    Proposition \ref{csst_prop:rw-finite-bd} and Proposition \ref{csst_prop:lw-finite-bd} follow from 
    the corresponding results in \cite{wei2020theoretical}.
\end{remark}

\begin{theorem}
    Suppose Assumptions \ref{csst_assump:c-exp}, \ref{csst_assump:pseudo-labeler}, and \ref{csst_assump:cov-cond} hold.
    Then, with probability at least $1 - \delta$ over the draw of the training data,
    for all $t_1, t_2 \in (0, \infty)$, and any neural network $\Phi$ in $\mF$, 
    we have the following:
    \begin{multline*}
        \err_w(F) = 
        \frac{\gamma + 1}{\gamma - 1}
        \ex[\pwhat]{\bone(m(\Phi, x, \gpl(x)) \le t_1)}
        + \frac{2\gamma}{\gamma - 1}\ex[\pwhat]{\bone(m_{\mB}(\Phi, x) \le t_2)} 
        \\
        - \err_w(\gpl) + 2\err_w(\gstar)+
        \frac{2\gamma}{\gamma - 1}R_{\mB, w}(\gstar)\\
        +\otilde\left(\frac{\sum_{i=1}^p \mcop(\mF_i)}{t_1 \sqrt{n}} \right) +
        \otilde\left(\frac{\sum_{i=1}^p \mcop(\mF_i)}{t_2 \sqrt{n}} \right) +
        \zeta,
    \end{multline*}
   where $\zeta = O\left(\sqrt{\frac{\log(1/\delta)+ \log n}{n}}\right)$ is a lower order term
   and $F(x) = \argmax_{i \in [K]}\Phi_i(x)$.
\end{theorem}
\begin{proof}
    By \eqref{csst_eq:errw-lem-applied} with $H = \gstar$ and 
    $-\bone(\gpl(x) \neq \gstar(x)) \le -\bone(\gpl(x) \neq j) + \bone(\gstar(x) \neq j)$
    for any $x \in \mX$ and $j \in [K]$, we obtain the following inequality:
    \begin{equation*}
        \err_w(F) \le \frac{\gamma + 1}{\gamma - 1} L_w(F, \gpl) + \frac{2\gamma}{\gamma - 1} R_{\mB, w}(F)
        + \frac{2\gamma}{\gamma - 1}R_{\mB, w}(\gstar) - \err_w(\gpl) + 2\err_w(\gstar).
    \end{equation*}
    Then, the statement of the theorem follows from Proposition \ref{csst_prop:rw-finite-bd} and Proposition \ref{csst_prop:lw-finite-bd}.
\end{proof}

\section{Proof of Proposition \ref{csst_prop:kl-weighted}}
\label{csst_sec:appendix-proof-prop}

\begin{proof} Let the average weighted consistency loss be $\mathcal{L}^{wt}_{u} = \frac{1}{|B|}\sum_{x \in B} \ell^{wt}_{u}(\hat{p}_{m}(x),p_{m}(\mathcal{A}(x)), \bG)$ this will be minimized if for each of $x \in B$ the $\ell^{wt}_{u}(\hat{p}_{m}(x),p_{m}(\mathcal{A}(x)), \bG)$ is minimized. This expression can be expanded as:
\begin{align*}
    \ell^{wt}_{u}(\hat{p}_{m}(x), p_{m}(\mathcal{A}(x)), \bG) &= -\sum_{i=1}^{K}(\bG^{T}\hat{p}_{m}(x))_i \log(p_{m}(\mathcal{A}(x))_i) \\
    &= -C\sum_{i=1}^{K}\frac{(\bG^{T}\hat{p}_{m}(x))_i}{\sum_{j = 1}^{m}\bG^{T}\hat{p}_{m}(x))_j} \log(p_{m}(\mathcal{A}(x))_i) \\
    & = C \times \mathrm{H}(\norm(\bG^T\hat{p}_{m}(x)) \; || \; p_{m}(\mathcal{A}(x))).
\end{align*}
Here we use H to denote the cross entropy between two distributions. As we don't backpropogate gradients from the $\hat{p}_{m}(x)$ (pseudo-label) branch of prediction network we can consider $C = \sum_{j = 1}^{m}\bG^{T}\hat{p}_{m}(x))_j$ as a constant in our analysis. Also adding a constant term of entropy $\mathrm{H}(\norm(\bG^{T}\hat{p}_{m}(x)))$ to cross entropy term and dropping constant $C$ doesn't change the outcome of minimization. Hence we have the following:
\begin{align*}
    \min_{p_{m}} \mathrm{H}(\norm(\bG^T\hat{p}_{m}(x)) \; || \; p_{m}(\mathcal{A}(x))) &= \min_{p_{m}} \mathrm{H}(\norm(\bG^T\hat{p}_{m}(x)) \; || \; p_{m}(\mathcal{A}(x))) \\
    & \qquad \qquad + \mathrm{H}(\norm(\bG^{T}\hat{p}_{m}(x))) \\
    &= \min_{p_{m}} \mathcal{D}_{KL}(\norm(\bG^{T}\hat{p}_{m}(x)) || p_{m}(\mathcal{A}(x))).
\end{align*}

This final term is the $\mathcal{D}_{KL}(\bG^{T}\hat{p}_{m}(x) || p_{m}(\mathcal{A}(x))$ which is obtained by using the identity $\mathcal{D}_{KL}(p,q) = \mathrm{H}(p,q) + \mathrm{H}(p)$ where $p, q$ are the two distributions.
\end{proof}

\section{Notation}
We provide the list of notations commonly used in the chapter in Table \ref{csst_tab:notations}. 
\vspace{-5mm}
\begin{table}[htbp]\caption{Table of Notations used in chapter}
\vspace{-7.5mm}
\begin{center}%
\begin{tabular}{r c p{10cm} }
\toprule
    \centering
    $\mY$ & $:$ & Label space\\
    $\mX$ & $:$ & Instance space \\
    $K$ & $:$ & Number of classes\\
    $\pi_i$ & $:$ & prior for class i\\  
    $F$ & $:$ & a classifier model\\
    $D$ & $:$ & data distribution \\
    $\blambda$ & $:$ & Lagrange multiplier \\
    $\lambda_u$ & $:$ & coefficient of unlabeled loss \\
    $\text{G}$ & $:$ & a $K \times K$ matrix\\
    $\text{D}$ & $:$ & a $K \times K$ diagonal matrix\\
    $\text{M}$ & $:$ & a $K \times K$ matrix\\
    $\mu$ & $:$ & ratio of labelled to unlabelled samples\\
    $B$ & $:$ & batch size for FixMatch\\
    $\ell_u^{\text{wt}}$ & $:$ & loss for unlabelled data using pseudo label\\
    $\ell_s^{\text{hyb}}$ & $:$ & loss for labelled data\\
    $\mathcal{L}_u^{\text{hyb}}$ & $:$ & average batch loss for unlabelled data using pseudo labels\\
    $\mathcal{L}_s^{\text{hyb}}$ & $:$ & average loss for labelled data on a batch of samples\\
    $H$ & $:$ & cross entropy function\\
    $\mA$ & $:$ & a $\mX \rightarrow \mX$ function a strong augmentation applied to it \\
    $\alpha$ & $:$ & a $\mX \rightarrow \mX$ function a weak augmentation applied to it \\
    $\rho$ & $:$ & imbalance factor \\
    $B$ & $:$ & batch size of samples\\
    $B_s$ & $:$ & batch of labelled samples\\
    $B_u$ & $:$ & batch of unlabelled samples\\
    $x$ & $:$ & an input sample, $x \in \mX$\\
    $\hat{p}_{m}$ & $:$ & a pseudo label generating function \\
    $p_{m}$ & $:$ & distribution of confidence of a model on a given sample \\
    $w$ & $:$ & a $K \times K$ weight matrix that corresponds to Gain $\bG$\\
    $\err_w(F)$ & $:$ & weighted error of $F$ corresponding to CSL\\
    $\addedtext{\mathcal{P}_w}$ & $:$ & weighted distribution on $\mX$\\
    $\addedtext{P_i}$ & $:$ & \addedtext{ class conditional distribution of samples for class $i$}\\
    $R_{\mB, w}(F)$ & $:$ & theoretical weighted (cost sensitive) regularizer\\
    $\gpl$ & $:$ & a pseudo labeler\\
    $L_w(F, \gpl)$ & $:$ & weighted error between $F$ and $\gpl$\\
    $\mL_w(F)$ & $:$ & theoretical \csst{} loss\\
    $c$ & $:$  & a non-increasing function used in the definition of the $c$-expansion property (Definition \ref{csst_def:c-exp-def})\\
    $\gamma$  & $:$ & a value of $c$ defined in Assumption \ref{csst_assump:pseudo-labeler}\\
    $\regub$ & $:$ & an upper bound of $R_{\mB, w}(F)$ in the optimization \eqref{csst_eq:cons-cnst-obj}\\
    $S^c$ & $:$ & the complement of a set $S$\\

\bottomrule

\end{tabular}
\end{center}
\label{csst_tab:notations}
\end{table}

 \section{Objective}
\subsection{Logit Adjusted Weighted Consistency Regularizer}
\label{csst_sec:La-hyb-loss}
As we have introduced weighted consistency regularizer in Eq. \eqref{csst_eq:wt-consistency-reg} for utilizing unlabeled data, we now provide  logit adjusted variant of it for training deep networks in this section. We provide logit adjusted term for $ \ell^{\mathrm{wt}}_{u}(\hat{p}_{m}(x), p_{m}(\mathcal{A}(x), \bG)$ below:
\begin{align*}
      \ell^{\mathrm{wt}}_{u}(\hat{p}_{m}(x), p_{m}(\mathcal{A}(x), \bG) &= -\sum_{i=1}^{K}(\bG^{\mathbf{T}}\hat{p}_{m}(x))_i \log(p_{m}(\mathcal{A}(x))_i) \\
      &= -\sum_{i=1}^{K}(\bG^{\mathbf{T}}\hat{p}_{m}(x))_i \log\left(\frac{\exp({p_{m}(\mathcal{A}(x))_i})}{\sum_{j=1}^{K}\exp({p_{m}(\mathcal{A}(x))_j})}\right) \\
      &= -\sum_{i=1}^{K}(\mathrm{\bm{D}}^{\mathbf{T}} \mathrm{\bm{M}}^{\mathbf{T}}\hat{p}_{m}(x))_i \log\left(\frac{\exp({p_{m}(\mathcal{A}(x))_i})}{\sum_{j=1}^{K}\exp({p_{m}(\mathcal{A}(x))_j})}\right) 
\end{align*}
The above expression comes from the decomposition $\bG=\mathrm{MD}$. The above loss function can be converted into it's logit adjusted equivalent variant by following transformation as suggested by \citet{narasimhan2021training} which is equivalent in terms of optimization of deep neural networks:
\begin{align}
      \ell^{\mathrm{wt}}_{u}(\hat{p}_{m}(x), p_{m}(\mathcal{A}(x), \bG)  \equiv -\sum_{i=1}^{K}( \mathrm{\bm{M}}^{\mathbf{T}}\hat{p}_{m}(x))_i \log\left(\frac{\exp({p_{m}(\mathcal{A}(x))_i -\log( \bm{D}_{ii}}))}{\sum_{j=1}^{K}\exp({p_{m}(\mathcal{A}(x))_j  -\log(\bm{D}_{jj}}))}\right)
      \label{csst_eq:lgt-adj-wt-cons}
\end{align}
The above loss is the consistency loss $\ell^{\mathrm{wt}}_{u}$ that we practically implement for \ttt{CSST}. Further in case $\hat{p}_{m}(x)$ is a hard pseudo label $y$ as in FixMatch, the above weighted consistency loss reduces to $\ell^{\mathrm{hyb}}(y, p_{m}(\mathcal{A}(x)))$. Further in case the gain matrix $G$ is diagonal the above loss will converge to $\ell^{\mathrm{LA}}(y, p_{m}(\mathcal{A}(x)))$. Thus the weighted consistency regularizer can be converted to logit adjusted variants $\ell^{\mathrm{LA}}$ and $\ell^{\mathrm{hyb}}$ based on $\bG$ matrix.

\subsection{\texttt{CSST}(FixMatch)}
\label{csst_sec:appendix-fixmatch-obj}
In FixMatch, we use the prediction made by the model on a sample $x$ after applying a weak augmentation $\alpha$ and is used to get a hard pseudo label for the models prediction on a strongly augmented sample i.e. $\mathcal{A}(x)$. The set of weak augmentations include horizontal flip,  We shall refer to this pseudo label as $\hat{p}_{m}(x)$. The list of strong augmentations are given in Table 12 of \citet{sohn2020fixmatch}. Weak augmentations include padding, random horizontal flip and cropping to the desired dimensions (32X32 for CIFAR and 224X224 for ImageNet).
Given a batch of labeled and unlabeled samples $B_s$ and $B_u$, \texttt{CSST} modifies the supervised and un-supervised component of the loss function depending upon the non-decomposable objective and its corresponding gain matrix $\bG$ at a given time during training. We assume that in the dataset, a sample $x$, be it labeled or unlabeled is already weakly augmented.  \texttt{vanilla} FixMatch's supervised componenet of the loss function is a simple cross entropy loss whereas in our \texttt{CSST}(FixMatch) it is replaced by $\ell_s^{\text{hyb}}$ . 
\begin{equation}
    \label{csst_eq:hyb-sup}
    \mathcal{L}_s^{\text{hyb}} =  \frac{1}{|B_s|}\sum_{x,y \in B_s}{\ell^{\text{hyb}}( y, s(x))}.
\end{equation}
\vspace{-3mm}
\begin{equation}
    \mathcal{L}_u^{\mathrm{wt}} = \frac{1}{|B_u|}\sum_{x \in B_u}\indc_{(\mathcal{D}_{KL} (\norm(\bG^{T}\hat{p}_{m}(x))  \; || \; p_m(x))\leq \tau)}
     \ell^{\mathrm{wt}}_{u}(\hat{p}_{m}(x), p_{m}(\mathcal{A}(x)), \bG)).
     \label{csst_loss:supp-hybloss}
\end{equation}

The component of the loss function for unlabeled data (i.e. consistency regularization) is where one of our contributions w.r.t the novel thresholding mechanism comes into light. \texttt{vanilla} FixMatch selects unlabeled samples for which consistency loss is non-zero, such that the model's confidence on the most likely predicted class is above a certain threshold. We rather go for a threshold mechanism that select based on the basis of degree of distribution match to a target distribution based on $\bG$. The final loss function $\mathcal{L} = \mathcal{L}_s^{hyb} + \lambda_u \mathcal{L}_u^{\mathrm{wt}}$, i.e. a linear combination of $\mathcal{L}_s^{\mathrm{hyb}}$ and $\mathcal{L}_u^{wt}$. Since for FixMatch we are dealing with Wide-ResNets and ResNets which are deep networks, as mentioned in Section \ref{csst_sec:La-hyb-loss}, we shall use the alternate logit adjusted formulation as mentioned in Eq. \eqref{csst_eq:lgt-adj-wt-cons} as substitute for $\ell_u^{\mathrm{wt}}$ in Eq. \eqref{csst_loss:supp-hybloss}.

\subsection{\texttt{CSST}(UDA)}
\label{csst_sec:appendix-uda-obj}
The loss function of UDA is a linear combination of supervised loss and consistency loss on unlabeled samples. The former is the cross entropy (CE) loss, while the latter for the unlabeled samples minimizes the KL-divergence between the model's predicted label distribution on an input sample and its augmented sample. Often the predicted label distribution on the unaugmented sample is sharpened. The augmentation we used was a English-French-English backtranslation based on the MarianMT~\cite{junczys-dowmunt-etal-2018-marian} fast neural machine translation model. In UDA supervised component of the loss is annealed using a method described as Training Signal Annealing (TSA), where the CE loss is considered only for those labeled samples whose $\max_i p_m(x)_i < \tau_t$, where $t$ is a training time step. We observed that using TSA in a long tailed setting leads to overfitting on the head classes and hence chose to not include the same in our final implementation.

\texttt{CSST} modifies the supervised and unsupervised component of the loss function in UDA depending upon a given objective and its corresponding gain matrix $\bG$ at a given time during training. The supervised component of the loss function for a given constrained optimization problem and a gain matrix $\bG$, is the hybrid loss $\ell_s^{\text{hyb}}$. For the consistency regularizer part of the loss function, we minimize the KL-divergence between a target distribution and the model's prediction label distribution on its augmented version.
The target distribution is $\norm(\bG^{T}\hat{p}_m(x))$, where $\hat{p}_m(x)$ is the sharpened prediction of the label distribution by the model. Given a batch of labeled and unlabeled samples $B_s$ and $B_u$, the final loss function in \ttt{CSST}(UDA) is a linear combination of $\mathcal{L}_s^{hyb}$ and $\mathcal{L}_u^{wt}$, i.e $\mathcal{L} = \mathcal{L}_s^{hyb} + \lambda_u \mathcal{L}_u^{wt}$. 
\begin{gather}
    \mathcal{L}_s^{\text{hyb}} =  \frac{1}{|B_s|}\sum_{x,y \in B_s}{\ell^{\text{hyb}}(p_m(x), y)}. \\
    \mathcal{L}_u^{wt} = \frac{1}{|B_u|}\sum_{x \in B_u}\indc_{(\mathcal{D}_{KL} (\norm(\bG^{T}\hat{p}_{m}(x))  \; || \; p_m(x))\leq \tau)}
     \ell^{\mathrm{wt}}_{u}(\hat{p}_{m}(x), p_{m}(\mathcal{A}(x), \bG)).
\end{gather}
Since for UDA, we are dealing with DistilBERT, as mentioned in Section \ref{csst_sec:La-hyb-loss}, we shall use the alternate formulation as mentioned in Eq. \eqref{csst_eq:lgt-adj-wt-cons} as substitute for $\ell_u^{\mathrm{wt}}$ in Eq. above. 

 \section{\addedtext{Threshold mechanism for diagonal Gain Matrix }}
\label{csst_Diagonal-G-hard-PL}
\addedtext{Consider the case when the gain matrix is a diagonal matrix. The loss function $\mathcal{L}_u^{wt}(B_u)$ as defined in \eqref{csst_loss:CSST-KL} makes uses of a threshold function that selects samples based on the KL divergence based threshold between the target distribution as defined by the gain matrix $\bG$ and the models predicted distribution of confidence over the classes.
\begin{equation}
    \text{Threshold function }\coloneqq \indc_{(\mathcal{D}_{KL} (\norm(\bG^{T}\hat{p}_{m}(x))  \; || \; p_m(x))\leq \tau)}
\end{equation}
Since $\bG$ is a diagonal matrix and the pseudo-label $\hat{p}_m(x)$ is one hot, the $\text{norm}(\bG^T\hat{p}_m(x))$ is a one-hot vector. The threshold function's KL divergence based criterion can be expanded as follows where $\hat{y}$ is the pseudo-label's maximum class's index:
\begin{equation}
    \mathcal{D}_{KL}(\text{norm}(\bG^T\hat{p}_m(x)) || p_{m}(x))   = - \log{p_m(x)}_{\hat{y}} < \tau \\  
\end{equation}
The above equations represents a threshold on the negative log-confidence of the model's prediction for a given unlabeled sample, for the pseudo-label class ($\hat{y}$). This can be further simplified to $p_m(x)_{\hat{y}} \geq \exp{(- \tau)} $ which is simply a threshold based on the model's confidence. Since pseudo-label is generated from the model's prediction, this threshold is nothing but a selection criterion to select only those samples whose maximum confidence for a predicted hard pseudo-label is above a fixed threshold. This is identical to the threshold function which is used in Fixmatch~\cite{sohn2020fixmatch} i.e. $\max(p_m(x)) \geq \exp(-\tau)$. In FixMatch this $\exp(-\tau)$ is set to 0.95.}

 \section{Dataset}
\textbf{CIFAR-10 and CIFAR-100~\cite{krizhevsky2009learning}.} are image classification datasets of images of size 32 X 32. Both the datasets have a size of 50k samples and by default, they have a uniform sample distribution among its classes. CIFAR-10 has 10 classes while CIFAR-100 has 100 classes. The test set is a balanced set of 10k images.

\noindent \textbf{ImageNet-100~\cite{russakovsky2015imagenet}.} is an image classification dataset carved out of ImageNet-1k by selecting the first 100 classes. The distribution of samples is uniform with 1.3k samples per class. The test set contains 50 images per class. All have a resolution of 224X224, the same as the original ImageNet-1k dataset.

\noindent \textbf{IMDb\cite{maas-EtAl:2011:ACL-HLT2011}.} dataset is a binary text sentiment classification dataset. The data distribution is uniform by default and has a total 25k samples in both trainset and testset. In this work, we converted the dataset into a longtailed version of $\rho=10, 100$ and selected 1k labeled samples while truncating the labels of the rest and using them as unlabeled samples.

\noindent \textbf{DBpedia-14\cite{lehmann2015dbpedia}.} is a topic classification dataset with a uniform distribution of labeled samples. The dataset has 14 classes and has a total of 560k samples in the trainset and 70k samples in the test set. Each sample, apart from the content, also has title of the article that could be used for the task of topic classification. In our experiments, we only make use of the content.

 \section{Algorithms}
We provide a detailed description of algorithms used for optimizing non decomposable objectives through \ttt{CSST}(FixMatch) ans \texttt{CSST}(UDA). Algorithm \ref{csst_algo:worst-case} is used for experiments in Section \ref{csst_subsec:worst-case-recall} for maximizing worst-case recall (i.e. min recall using \ttt{CSST}(FixMatch) and \ttt{CSST}(UDA)). Algorithm \ref{csst_algo:coverage} is used for experiments in Section \ref{csst_subsec:worst-case-recall} for maximizing recall under coverage constraints (i.e. min coverage experiments on CIFAR10-LT, CIFAR100-LT and ImageNet100-LT).
\begin{algorithm}[H]
\caption{\texttt{CSST}-based Algorithm for Maximizing Worst-case Recall }
\label{csst_algo:worst-case}
\begin{algorithmic}
\STATE Inputs: Training set $S_s$(labeled) and $S_u$(unlabeled) , Validation set $S^{\text{val}}$, Step-size $\omega \in \mathbb{R}_{+}$, Class priors $\pi$
\STATE Initialize: Classifier $F^0$, Multipliers $\blambda^0 \in \Delta_{K -1}$
\FOR{$t = 0 $ to $T-1$}
\STATE \textbf{Update $\blambda$:}~\\[2pt]
\STATE ~~~~$\lambda^{t+1}_i = \lambda^t_i\exp\left(-\omega \cdot \text{recall}_i[F^{t}] \right), \forall i,$\\
\STATE ~~~~$\blambda = \norm (\blambda)$
\STATE ~~~~$\bG =  \diag(\lambda^{t+1}_1/\pi_1, \ldots, \lambda^{t+1}_K/\pi_K)$~\\
\STATE \text{Compute $\ell_u^{\text{wt}}$, $\ell_s^{\text{hyb}}$ using $\bG$ }\\
\STATE \textbf{Cost-sensitive Learning (CSL) for FixMatch:}~\\[2pt]
\STATE  ~~~~$B_u \sim S_u , B_s \sim S_s$ ~// Sample batches of data
\STATE ~~~~$F^{t+1} \,\in\, \arg \min_{F} \sum_{B_u, B_s}\lambda_u\mathcal{L}_{u}^{\mathrm{wt}} + \mathcal{L}_s^{\mathrm{hyb}}$ ~// Replaced by few steps of SGD

\ENDFOR
\RETURN $F^T$
\end{algorithmic}
\end{algorithm}

\begin{algorithm}[H]
\caption{\texttt{CSST}-based Algorithm for Maximizing Mean Recall s.t. per class coverage > 0.95/K }
\label{csst_algo:coverage}
\begin{algorithmic}
\STATE Inputs: Training set $S_s$(labeled) and $S_u$(unlabeled) , Validation set $S^{\text{val}}$, Step-size $\omega \in \mathbb{R}_{+}$, Class priors $\pi$
\STATE Initialize: Classifier $F^0$, Multipliers $\blambda^0 \in \RR_+^{K}$
\FOR{$t = 0 $ to $T-1$}
\STATE \textbf{Update $\blambda$:}~\\[2pt]
\STATE ~~~$\lambda^{t+1}_i = \lambda^t_i - \omega\big(\text{cov}_i[F^{t}] - \frac{0.95}{K}\big), \forall i$\\
\STATE ~~~~$\lambda^{t+1}_i \,=\, \max\{0, \lambda^{t+1}_i\}, \forall i \in [K]$ ~~~// Projection to $\mathbb{R}_+$
\STATE ~~~~$\bG =  \diag(\lambda^{t+1}_1/\pi_1, \ldots, \lambda^{t+1}_K/\pi_K) + \bf{1}_K \blambda^{\trn}$~\\
\STATE \text{Compute $\ell_u^{\text{wt}}$, $\ell_s^{\text{hyb}}$ using $\bG$ }\\

\STATE \textbf{Cost-sensitive Learning (CSL) for FixMatch:}~\\[2pt]
\STATE  ~~~~$B_u \sim S_u , B_s \sim S_s$ ~// Sample batches of data
\STATE ~~~~$F^{t+1} \,\in\, \arg \min_{F} \sum_{B_u, B_s}\lambda_u\mathcal{L}_{u}^{\mathrm{wt}} + \mathcal{L}_s^{\mathrm{hyb}}$ ~// Replaced by few steps of SGD

\ENDFOR
\RETURN $F^T$
\end{algorithmic}
\end{algorithm}

\section{Details of Experiments and Hyper-parameters}
\label{csst_sec:appendix-details-of-experiments}
The experiment of $\max_F \min_i \text{recall}_i[f]$ and $\max_F \text{recall}[F] \text{ s.t. } \text{cov}_i[F] > \frac{0.95}{K}, \forall i \in [K]$  was performed on the long tailed version of CIFAR-10, IMDb($\rho=10,100$) and DBpedia-14 datasets. This was because the optimization of the aforementioned 2 objectives is stable for cases with low number of classes. Hence the objective of $\max_F \min(\text{recall}_{\mH}[F], \text{ recall}_{\mT}[F])$  and  \\ $\max_F \text{recall}[F] \text{ s.t. } \min_{\mH, \mT} \text{cov}_{\mH,\mT}[F] > \frac{0.95}{K}$ is a relatively easier objective for datasets with large number of classes, hence were the optimization objectives for CIFAR-100 and ImageNet-100 long tailed datasets. For all experiments for a given dataset, we used the same values for a given common hyperparameter. We ablated the threshold for our novel unlabeled sample selection criterion($\tau$) and the ratio of labeled and unlabeled samples, given fixed number of unlabeled samples($\mu$) and are available in Fig. \textcolor{red}{4b}.

\begin{table}[t]
\centering
\begin{adjustbox}{max width=\textwidth}
\begin{tabular}{cccccccc} 
\toprule
    Parameter &  CIFAR-10 & CIFAR-100 & ImageNet-100 & \begin{tabular}{@{}c@{}}IMDb \\ ($\rho=10$)\end{tabular} & \begin{tabular}{@{}c@{}}IMDb \\ ($\rho=100$)\end{tabular}  &  DBpedia-14\\
     \hline
    $\tau$ &  0.05 & 0.05 & 0.05 & 0.1 & 0.1 & 0.1\\
    $\lambda_u$ & 1.0 & 1.0 & 1.0 & 0.1 & 0.1 & 0.1\\
    $\mu$ & 4.0 & 4.0 & 4.0 & 13.8 & 12.6 & 133\\
    $|B_s|$ & 64 & 64 & 64 & 32 & 32 & 32\\
    $|B_u|$ & 256 & 256 & 256 & 128 & 128 & 128\\
    lr & 3e-3 & 3e-3 & 0.1 & 1e-5 & 1e-5 & 1e-5\\
    $\omega$ & 0.25 & 0.25 & 0.1 & 0.5 & 0.5 & 0.5\\
    \begin{tabular}{@{}c@{}}SGD steps \\ before eval\end{tabular} &  32 & 100 & 500 & 50 & 50 & 100\\
    optimizer & SGD & SGD & SGD & AdamW & AdamW & AdamW \\
    KL-Thresh & 0.95 & 0.95 & 0.95 & 0.9 & 0.9 & 0.9 \\
    Weight Decay & 1e-4 & 1e-3 & 1e-4 & 1e-2 & 1e-2 & 1e-2\\
    $\rho$ &  100 & 10 & 10 & 10 & 100 & 100\\
    $\lambda_u$ &  1.0 & 1.0 & 1.0 & 0.1 & 0.1 & 0.1 \\
    $\mu$ &  4.0 & 4.0 & 4.0 & 11.0 & 11.0 & 11.0 \\
    Arch. &  WRN-28-2 & WRN-28-8 & ResNet50 & DistilBERT & DistilBERT & DistilBERT \\
    \hline
\end{tabular}
\end{adjustbox}
\caption{This table shows us the detailed hyper parameters used for \texttt{CSST}(FixMatch) for the long tailed datasets CIFAR-10, CIFAR-100, ImageNet-100 and \ttt{CSST}(UDA) on IMDb, DBpedia-14. All the datasets were converted to their respective long tailed versions based on the imbalance factor $\rho$, and a fraction of the samples were used along with their labels for supervision.}
\label{csst_tab:hyperparams}
\end{table}
\vspace{-3mm}

 \section{Code, License, Assets and Computation Requirements}
\label{csst_sec:code-licence-assets-comp-req}
\subsection{Code and Licenses of Assets}
In this work, we use the open source implementation of FixMatch~\cite{sohn2020fixmatch} \footnote{https://github.com/LeeDoYup/FixMatch-pytorch} in PyTorch, which is licensed under \ttt{MIT} License for educational purpose. Also for NLP experiments we make use of DistillBERT~\cite{sanh2019distilbert} pretrained model available in the HuggingFace~\cite{wolf2020transformers} library. The code to reproduce the main experiments results can be found on \href{https://github.com/val-iisc/CostSensitiveSelfTraining}{Github}.

\subsection{Computational Requirements}
\begin{table}[htb]
\centering
\captionsetup{width=.68\textwidth}
\caption{Computational requirements and training time (d:days, h:hours, m:minutes) for  experiments relevant to vision datasets. As we can see some of the experiments on the larger datasets such as ImageNet requires long compute times.}
\label{csst_tab:GPU-usage-vision}
\begin{tabular}{cccccccc} 
\toprule
    Method &  CIFAR-10 & CIFAR-100 & ImageNet-100 \\
     \hline
   \texttt{ERM}   &  \begin{tabular}{@{}c@{}}A5000 \\ 49m\end{tabular}  & \begin{tabular}{@{}c@{}}A5000 \\ 6h 47m\end{tabular} & \begin{tabular}{@{}c@{}} RTX3090 \\ 15h 8m\end{tabular} \\
   \texttt{LA}   &  \begin{tabular}{@{}c@{}}RTX3090 \\ 
39m\end{tabular}  & \begin{tabular}{@{}c@{}}A5000 \\ 6h 9m\end{tabular} & \begin{tabular}{@{}c@{}}A5000 \\ 15h 7m \end{tabular} \\
   \texttt{CSL}   &  \begin{tabular}{@{}c@{}}A5000 \\ 47m\end{tabular}  & \begin{tabular}{@{}c@{}}A5000 \\ 6h 40m\end{tabular} & \begin{tabular}{@{}c@{}}A5000 \\ 12h\end{tabular} \\
   \begin{tabular}{@{}c@{}}\texttt{CSST}(FixMatch)\\ w/o KL-Threshold\end{tabular} &  \begin{tabular}{@{}c@{}}4 X A5000 \\ 21h 0m\end{tabular}  & \begin{tabular}{@{}c@{}}4 X A100 \\ 2d 19h 16 m\end{tabular} & \begin{tabular}{@{}c@{}}4 X A5000 \\ 2d 13h 19m\end{tabular} \\
   \texttt{CSST}(FixMatch)    &  \begin{tabular}{@{}c@{}}4 X A5000 \\  21h 41m \end{tabular}  & \begin{tabular}{@{}c@{}}4 X A5000 \\2d 11h 52m \end{tabular} & \begin{tabular}{@{}c@{}}4 X A5000 \\ 2d 4m\end{tabular} \\
    \hline
\end{tabular}

\end{table}

\begin{table}[htb]
\centering
\captionsetup{width=.68\textwidth}
\caption{Computational requirements and training time(d:days, h:hours, m:minutes)  for experiments done on NLP datasets. The DistilBERT model which we are using is pretrained on a language modeling task, hence it requires much less time for training in comparison to vision models trained from scratch.}
\label{csst_tab:GPU-usage-nlp}
\begin{tabular}{cccccccc} 
\toprule
    Method &  IMDb($\rho=10$) & IMDb($\rho=100$) & DBpedia-14\\
     \hline
   \texttt{ERM}   &  \begin{tabular}{@{}c@{}}4 X A5000 \\ 25m \end{tabular}  & \begin{tabular}{@{}c@{} }4 X A5000 \\ 29m \end{tabular} & \begin{tabular}{@{}c@{}}4 X A5000 \\2h 44m\end{tabular} \\
   \texttt{UDA}   &  \begin{tabular}{@{}c@{}}4 X A5000 \\ 44m\end{tabular}  & \begin{tabular}{@{}c@{}}4 X A5000 \\ 32m\end{tabular} & \begin{tabular}{@{}c@{}}4 X A5000 \\ 10h 18m\end{tabular} \\
   \texttt{CSST}(UDA)    &  \begin{tabular}{@{}c@{}}4 X A5000 \\ 49m\end{tabular}  & \begin{tabular}{@{}c@{}}4 X A5000 \\ 35m\end{tabular} & \begin{tabular}{@{}c@{}}4 X A5000 \\ 13h 12m\end{tabular} \\
    \hline
\end{tabular}

\end{table}

All experiments were done on a variety of GPUs, with primarily Nvidia A5000 (24GB) with occasional use of Nvidia A100 (80GB) and Nvidia RTX3090 (24GB). For finetuning DistilBERT and all experiments with ImageNet-100 dataset we used PyTorch data parallel over 4 A5000s. Training was done till no significant change in metrics was observed. The detailed list of computation used per experiment type and dataset have been tabulated in  Table \ref{csst_tab:GPU-usage-vision} and Table \ref{csst_tab:GPU-usage-nlp}.

 \begin{table}[!t]
    \centering
    \small
  
    \captionsetup{width=.55\textwidth}
    \caption{Avg. and std. deviation of Mean Recall and Min. Recall for CIFAR-10 LT}
      \label{csst_tab:statistical-analysis}
    \begin{tabular}{lcccccc}
        \hline
        \multicolumn{1}{c}{\textbf{Method}} & 
        \multicolumn{1}{c}{\textbf{Mean Recall}} & 
        \multicolumn{1}{c}{\textbf{Min Recall}} &
        \\
        \hline
        \texttt{ERM} &  
            0.52 $\pm$ 0.01 &	0.27 $\pm$ 0.02
            \\
        \texttt{LA} &  
            0.54 $\pm$ 0.02 &	0.37 $\pm$ 0.01 
            \\
        \texttt{CSL} &  
            0.63 $\pm$ 0.01 &	0.43 $\pm$ 0.04 
            \\
        \hline
        \begin{tabular}{@{}c@{}}\texttt{Vanilla} (FixMatch)
        \end{tabular}
            & 0.78 $\pm$ 0.01 & 0.47 $\pm$ 0.02 
            \\
        \begin{tabular}{@{}c@{}}\texttt{CSST}(FixMatch) 
        \end{tabular}
            & 0.75 $\pm$ 0.01 &	0.72 $\pm$ 0.01 
            \\
        \hline
    \end{tabular}
    \vspace{0.5em}

    \vspace{-5mm}
\end{table}
\section{Statistical Analysis}
We establish the statistical soundness and validity of our results we ran our experiments on 3 different seeds. Due to the computational requirements for some of the experiments ($\approx $2days) we chose to run the experiments on multiple seeds for a subset of tasks i.e. for maximising the minimum recall among all classes for CIFAR-10 LT. We observe that the std. deviation is significantly smaller than the average values for mean recall and min. recall and our performance metrics fall within our std. deviation hence validating the stability and soundness of training.

 \section{\addedtext{Additional Details}}
\subsection{\addedtext{Formal Statement Omitted in Sec. \ref{csst_sec:loss-fn-for-ndo}}}
\label{csst_sec:appendix-formal-statement}
\addedtext{
In Sec. \ref{csst_sec:loss-fn-for-ndo}, 
we stated that learning with the hybrid loss $\hybloss$ gives the Bayes-optimal classifier
for the CSL \eqref{csst_eq:csl-obj}.
However, due to space constraint, we did not provide a formal statement.
In this section, we provide a formal statement of it for clarity.
}
\begin{proposition}[\addedtext{\cite{narasimhan2021training} Proposition 4}]
    \addedtext{For any diagonal matrix $\bD \in \RR^{K \times K}$ with $D_{ii} > 0, \forall i$, 
    $\bM \in \RR^{K \times K}$,
    and $\bG =  \bM \bD$, the hybrid loss $\hybloss$ is calibrated for $\bG$.
    That is, for any model-prediction $\widehat{p}_m: \mX \rightarrow \RR^K$ that minimizes 
    $\ex[(x, y) \sim D]{\hybloss(y, \widehat{p}_m(x))}$, the associated classifier $F(x) = \argmax_{y \in [K]} 
    \left(\widehat{p}_{m}\right)_i(x)$ is the Bayes-optimal classifier for CSL \eqref{csst_eq:csl-obj}.}
\end{proposition}
\subsection{Comparison with the $(a, \widetilde{c})$-expansion Property in \cite{wei2020theoretical}}
\label{csst_sec:appendix-a-c-expansion}
\addedtext{We compare the $c$-expansion property with $(a, \widetilde{c})$-expansion property proposed by \cite{wei2020theoretical},
where $a \in (0, 1)$ and $\widetilde{c} > 1$.
Here we say a distribution $Q$ on $\mX$ satisfies the $(a, \widetilde{c})$-expansion property 
if $Q(\mN(S)) \ge \widetilde{c}$ for any $S \subset \mX$ with $Q(S) \le a$.
If $Q$ satisfies $(a, \widetilde{c})$-expansion property \cite{wei2020theoretical} 
with $\widetilde{c} > 1$, then
$Q$ satisfies the $c$-expansion property, where 
the function $c$ is defined as follows.
$c(p) = \widetilde{c}$ if $p \le a$ and $c(p) = 1$ otherwise.
On the other hand, if $Q$-satisfies $c$-expansion property, 
then for any $a \in (0, 1)$ and $S \subseteq \mX$ with $Q(S) \le a$,
we have $Q(\mN(S)) \ge c(Q(S)) Q(S) \ge c(a) Q(S)$ since $c$ is non-increasing.
Therefore, $Q$ satisfies the $(a, c(a))$-expansion property.
Thus, we could say these two conditions are equivalent.
To simplify our analysis, we use our definition of the expansion property.
}

\addedtext{In addition, \citet{wei2020theoretical} showed that the $(a, \widetilde{c})$-expansion property is realistic for vision. Although they assumed the $(a, c)$-expansion property for each $P_i$ $(1 \le i \le K)$ and we assume the $c$-expansion property for $\Pw$, it follows that the $c$-expansion property for $\Pw$ is also realistic for vision, since $\Pw$ is a linear combination of $P_i$}.

\section{Additional Experiments}
\label{csst_sec: additional-experiments}
In this section, we compare \texttt{CSST}(FixMatch) against contemporary semi-supervised learning techniques namely CReST~\cite{Wei_2021_CVPR}, DARP~\cite{10.5555/3495724.3496945},  \texttt{vanilla}(FixMatch)~\cite{sohn2020fixmatch} and FlexMatch~\cite{NEURIPS2021_995693c1}. We compared these methods on the long-tailed CIFAR-10 ($\rho$ = 100) and CIFAR-100 ($\rho$ = 10) datasets. The objective for long-tailed CIFAR-10 dataset was to maximise the worst-case recall \eqref{csst_eq:min-recall-obj} and average recall, subject to a per-class coverage constraint \eqref{csst_eq: cov-const-obj}.  For CIFAR-100  LT dataset, we compare these methods for the objectives maximizing HT recall \eqref{csst_eq:min-HT-recall-obj} and recall under HT coverage constraints\eqref{csst_eq: HT-cov-const-obj}. For the objectives \eqref{csst_eq:min-recall-obj} and  \eqref{csst_eq:min-HT-recall-obj}, DARP achieves the best average recall yet it suffers on the worst-case recall. \texttt{CSST}(FixMatch) outperforms it on CIFAR-100 and  the has superior worst-case recall on CIFAR-10. A similar observation can be made for the objectives \eqref{csst_eq: cov-const-obj} and \eqref{csst_eq: HT-cov-const-obj} where it is only \texttt{CSST}(FixMatch) that has the highest min. coverage and either has superior mean recall or has negligible decrease in the avg. recall.
\begin{table}[!t]
    \centering
    \small
    \caption{Comparing CSST(FixMatch) against other Semi-Supervised Learning Methods for long tailed data distribution for the objectives \eqref{csst_eq:min-recall-obj} and \eqref{csst_eq:min-HT-recall-obj}. Although \texttt{CSST}(FixMatch) does not achieve the highest mean recall, it at very little cost to mean recall, achieves the best worst-case recall              }
    \begin{tabular}{lcc|cccccc}
        \hline
        \multicolumn{1}{c}{\textbf{Method}} & 
        \multicolumn{2}{c}{\textbf{CIFAR10-LT }} & 
        \multicolumn{2}{c}{\textbf{CIFAR100-LT }} &
        \\
        & \multicolumn{2}{c}{($\rho = 100$)} & 
        \multicolumn{2}{c}{($\rho = 10$)} &
        \\
        \hline
        & \textbf{Avg. Rec} & \textbf{Min Rec} & 
        \textbf{Avg. Rec } & \textbf{Min HT Rec} &
        \\
        \hline
        CReST &	0.72 &	0.47 &	0.52 &	0.46 \\
        DARP  &	\highlight{0.81} &	0.64 &	0.55 &	0.54 \\
        FlexMatch &	0.80 &	0.48 &	0.61 &	0.39 \\
        \begin{tabular}{@{}c@{}}\texttt{Vanilla} (FixMatch)
        \end{tabular}
            & 0.78 &	0.48 &
            \highlight{0.63} &	0.36 \\
        \begin{tabular}{@{}c@{}}\texttt{CSST}(FixMatch) 
        \end{tabular}
            & 0.76 &	\highlight{0.72} &
            \highlight{0.63} &	\highlight{0.61} \\
        
        \hline
    \end{tabular}
    \vspace{0.5em}

    \label{csst_tab:add_expts_min_rec}
    \vspace{-5mm}
\end{table}

\begin{table}[!t]
    \centering
    \small
    \caption{Comparing CSST(FixMatch) against other Semi-Supervised Learning Methods for long tailed data distribution for the objectives \eqref{csst_eq: cov-const-obj}, \eqref{csst_eq: HT-cov-const-obj}. Only \texttt{CSST}(FixMatch) is the closest to satisfying the constraint yet suffers very little on the avg. recall. }
    \begin{tabular}{lcc|cccccc}
        \hline
        \multicolumn{1}{c}{\textbf{Method}} & 
        \multicolumn{2}{c}{\textbf{CIFAR10-LT }} & 
        \multicolumn{2}{c}{\textbf{CIFAR100-LT }} &
        \\
        & \multicolumn{2}{c}{($\rho = 100$, tgt : 0.1)} & 
        \multicolumn{2}{c}{($\rho = 10$, tgt : 0.01)} &
        \\
        \hline
        & \textbf{Avg. Rec} & \textbf{Min Cov} & 
        \textbf{Avg. Rec } & \textbf{Min HT Cov} &
        \\
        \hline
        CReST &	0.72 &	0.052 &	0.52 &	0.009 \\
        DARP  &	\highlight{0.81} &	0.063 &	0.55 &	0.006 \\
        FlexMatch &	0.80 &	0.046 &	0.61 & 	0.006 \\
        \begin{tabular}{@{}c@{}}\texttt{Vanilla} (FixMatch)
        \end{tabular}
            & 0.78 &	0.055 &
            \highlight{0.63} &	0.004 \\
        \begin{tabular}{@{}c@{}}\texttt{CSST}(FixMatch) 
        \end{tabular}
            & 0.80 &	\highlight{0.092} &	\highlight{0.63} &	\highlight{0.010} \\
        
        \hline
    \end{tabular}
    \vspace{0.5em}

    \label{csst_tab:add_expts_min_cov}
    \vspace{-5mm}
\end{table}

\renewcommand{\thesection}{G}

\supersection{Selective Mixup Fine-Tuning for Optimizing \\ Non-Decomposable Objectives (Chapter-8)}

\section{Notation}
We provide a summary of notations in Table~\ref{tab:notations}.

\begin{table}[htbp]
\vspace{-5mm}
\caption{Table of Notations used in chapter.}
\vspace{-7.5mm}
\label{tab:notations}
\begin{center}%

\begin{adjustbox}{valign=t, max height=\textheight}
\begin{tabular}{c p{10cm} }
\toprule
    \centering
    $K$  & Number of classes\\
    $\mY := [K]$  & Label space\\
    $x$ & instance \\
    $y$  & label \\
    $\eta$  & learning rate \\
    $\mX$  & instance space \\
    $d$  & feature space \\
    $h: \mX \rightarrow \RR^{K}$  & a neural network based scoring function \\
    $C[h]$  & confusion matrix for the classifier $h$\\
    $\Delta_{n-1} \subset \RR^{n}$ & the $n-1$-dimensional probability simplex\\
    $\psi: \Delta_{K^2-1} (\subset \RR^{K \times K}) \rightarrow \RR$  & a function defined on the set of confusion matrices ($\psi(C[h])$ is the metric of $h$)\\
    $\pi_i$  & prior for class $i \in [K]$\\  
    $g: \mX \rightarrow \RR^d$  & a feature extractor (backbone)\\
    $f: \RR^d \rightarrow \Delta_{K-1}$  & the final classifier such that $h = \argmax_{i}f_i \circ g$\\
    $W \in \RR^{d \times K}$  & the weight of the final layer\\
    $z_k$  & the centroid of features of class samples given as $\ex[x \sim D^{\mathrm{val}}_{k}]{g(x)}$\\
    $\mL_{\text{mixup}}(x_i, x_j, y_i; W)$  & the loss for mixup between labeled sample $(x_i, y_i)$ and unlabeled sample $x_j$\\
    ${\mL}_{(ij)}^{\text{mix}}$   & the expected loss due to $(i, j)$ mixup\\
    $G_{ij}$  & gain upon performing $(i, j)$ mixup\\
    $V_{ij}$  & the directional vector (matrix) defined by the $(i, j)$ mixup\\
    $\langle A, B \rangle$ \ (where $A, B \in \RR^{m \times n}$) &  $\Tr A B^\trn$\\
    $\nabla_{A} \chi$ \ (where $A \in \RR^{ d \times K}$)  & the directional derivative of a function $\chi$ with the directional vector (matrix) $A$ \\
    $s$  & the inverse temperature parameter for the softmax\\
    $\tilde{C}$   & unconstrained extension for confusion matrix $C$\\
    $D_i$  & subset of data with label $i$\\
    $\tilde{D}_i$  & subset of data with pseudo-label $i$\\
    $\mP$ & a distribution on $[K] \times [K]$\\
    $\bmP = (\mP^t)_{t=1}^T$ & a policy (a sequence of distributions $\mP^t$)\\
    $\overline{G}(\bmP)$ & the expected average gain of $\bmP$\\
    $N_k$ & the number of samples in the $k$-th labeled class \\
    $M_k$ & the number of samples in the $k$-th unlabeled class\\
    $\rho_l$ & the class imbalanced factor of the labeled dataset ($\max_{1\le i, j \le K} N_i/N_j$)\\
    $\rho_u$ & the class imbalanced factor of the unlabeled dataset\\
    $\mathcal{H}$ & The set of first 90\% classes that contains the majority\\ 
    $\mathcal{T}$ & The set of last 10\% classes that contains the minority\\ 
    $\|A\|_\frb$ & the Frobenius norm of a matrix\\

\bottomrule

\end{tabular}
\end{adjustbox}

\end{center}

\end{table}

\section{Computational Complexity}
\label{app:complexity}

We discuss the computational complexity of SelMix and that of an existing method \cite{rangwani2022costsensitive} for 
non-decomposable metric optimization
in terms of the class number $K$.
We note that to the best of our knowledge, CSST \cite{rangwani2022costsensitive} is the only existing method for 
non-decomposable metric optimization 
in the SSL setting.

\begin{proposition}
    \label{prop:computational-complexity}
    The following statements hold:
    \begin{enumerate}
        \item In each iteration $t$ in Algorithm \ref{alg:ours}, 
        computational complexity for $\mP^{(t)}_\selmix$ is given 
        as $O(K^3)$.
        \item In each iteration of CSST \cite{rangwani2022costsensitive}, it needs procedure that takes $O(K^3)$ time.
    \end{enumerate}
    Here, the Big-O notation hides sizes of parameters of the network other than $K$ (i.e., the number of rows of $W$)
    and
    the size of the validation dataset.
\end{proposition}
\begin{proof}
    \textbf{1.}
    Computational complexity for the confusion matrix is given as $O(K^3)$ 
    since there are $K^2$ entries and for each entry, evaluating $h^{(t)}(x)$ takes $O(K)$ time for each validation data $x$.
    For each $1 \le k \le K$, 
    computational complexity for $z_k$ is $O(K)$.
    We compute $\{\softmax(z_k)\}_{1 \le k \le K}$, which takes $O(K^2)$ time.
    The $(m,l)$-th entry of the matrix $\nu_{ij}$ is given as 
    $-\eta \zeta_m(\delta_{il} - \softmax_i(\zeta))$,
    where 
    $1 \le m \le d$, $1 \le l \le K$, and
    $\zeta = \beta z_i + (1-\beta)z_j \in \RR^d$.
    Therefore, once we compute $\{\softmax(z_k)\}_{1 \le k \le K}$,
    computational complexity for 
    $\{\nu_{ij}\}_{1\le i, j \le K}$ is $O(K^3)$.
    For each $1 \le l \le K$,
    we put 
    $v_l = \sum_{k=1}^{K}
    \frac{\partial \psi(C^{(t)})}{\partial \tilde{C}_{kl}} z_k$.
    Then computational complexity for $\{v_l\}_{1 \le l \le K}$ is $O(K^2)$.
    Since 
    $G^{(t)}_{ij} =  \sum_{l=1}^{K} (\nu_{ij}^{(t)})^\top_l \cdot v_l$ is a sum of $K$ dot products of $d$-dimensional vectors,
    once we compute $\{v_l\}_{l}$, computational complexity for $\{G^{(t)}_{ij}\}_{1\le i,j\le K}$ is $O(K^3)$.
    Thus, computational complexity for $\mP^{(t)}_\selmix$ is given as $O(K^3)$.

    \textbf{2.}
    In each iteration $t$, CSST needs computation of a confusion matrix at validation dataset.
    Since there are $K^2$ entries and for each entry,
    $h^{(t)}(x)$ takes $O(K)$ time for each validation data $x$,
    computational complexity for the confusion matrix is given as $O(K^3)$.
    Thus, we have our assertion.
\end{proof}

\section{Additional Theoretical Results and Proofs omitted in the Chapter}

\subsection{Convergence Analysis}
\label{sec:app-convergence-analysis-proof}
We provide convergence analysis of Algorithm \ref{alg:ours}.
For each iteration $t=1, \dots, T$, Algorithm \ref{alg:ours} updates parameter $W$ as follows:
\begin{enumerate}
    \item It selects a mixup pair $(i, j)$ from the distribution $\mP_{\mathrm{SelMix}}^{(t)}$.
    \item and updates parameter $W$ by $W^{(t + 1)} = W^{(t)} + \eta_t \vtt$, 
    where $\vtt = V_{ij}^{(t)}/\|V_{ij}^{(t)}\|$.
\end{enumerate}
Here, we consider the normalized directional vector $\vtt$ instead of $V_{ij}^{(t)}$ and $\| \cdot \|$ denotes the Euclidean norm.
We denote by $\ex[t-1]{\cdot}$ the conditional expectation conditioned on randomness with respect to 
mixup pairs up to time step $t-1$.
\begin{assumption}
    The function $\psi$ (as a function of $W$) is concave, differentiable, the gradient $\frac{\partial \psi}{\partial W}$ is $\gamma$-Lipschitz, i.e., $\|\frac{\partial \psi}{\partial W} - \frac{\partial \psi}{\partial W'}\| \le \gamma \|W -W'\|$ where $\gamma > 0$.
    There exists a constant $c > 0$ independent of $t$ satisfying
    \begin{equation}
        \label{eq:assump-cosine}
        \ex[t-1]{\vtt} \cdot \frac{\nabla \psi(W^{(t)})}{\|\nabla \psi(W^{(t)})\|}  > c,
    \end{equation}
    where $\nabla \psi (W^{(t)})=\frac{\partial \psi (W^{(t)})}{\partial W}$,
    that is, $\vtt$ \addedtext{vector has sufficient alignment with the gradient}.
    Moreover, we make the following technical assumption.
    We assume that in the optimization process, 
    $\|W^{(t)}\|$ does not diverge, i.e., we assume that for any $t \ge 1$, we have
    \begin{equation*}
        \| W^{(t)} \| < R
    \end{equation*}
    with a constant $R> 0$.
    In practice, this can be satisfied by adding $\ell^2$-regularization of $W$ to the optimization.
    We define a constant $R_0>0$ as $R_0 = \| W^*\| + R$,
    where $W^* = \argmax_W \psi(W)$.
    We note that a similar boundedness condition using a level set is assumed by 
    \cite{wright2015coordinate}.
\end{assumption}

\begin{theorem}
    Under the above assumptions and notations, we have the following result.
For any $t > 1$, we have
\begin{equation*}
  \sup_{W}\psi(W) -  \ex{\psi(W^{(t)})} \le \frac{4\gamma R_0^2}{c^2 (t -1)},
\end{equation*}
with an appropriate choice of the learning rate $\eta_t$.
\end{theorem}
\begin{proof}
By the Taylor's theorem, for each iteration $t$, there exists $s = s_t \in [0, 1]$ such that
$\psi(W^{(t+1)}) = \psi(W^{(t)}) + \eta_t \nabla \psi(\xi) \cdot \vtt$,
where $\xi = W^{(t)} + s\eta_t \vtt$.
We decompose $\eta_t \nabla \psi(\xi) \cdot \vtt$ as 
$\eta_t (\nabla \psi(\xi) - \nabla \psi(W^{(t)}))\cdot \vtt + 
\eta_t \nabla \psi(W^{(t)}) \cdot \vtt$.

First, we provide a lower bound of the first term.
Since $\nabla \psi$ is Lipschitz, we have 
$\|\nabla \psi(\xi) - \nabla \psi(W^{(t)}) \| \le \gamma \| \xi - W^{(t)}\|
\le \gamma \eta_t$.
Thus, by the Cauchy-Schwartz, we have
$\eta_t (\nabla \psi(\xi) - \nabla \psi(W^{(t)}))\cdot \vtt 
\ge -\gamma \eta_t^2.$
Next, we consider the second term. 
By the assumption on the cosine similarity \eqref{eq:assump-cosine}, we have
\begin{equation*}
  \eta_t \nabla \psi(W^{(t)}) \cdot \ex[t-1]{\vtt} \ge c \eta_t \| \nabla \psi(W^{(t)})\|  
\end{equation*}
where $c > 0$
is a constant independent of $t$. Here we note that when taking the expectation $\ex[t-1]{\cdot}$, 
we can regard $W^{(t)}$ as a non-random variable.
Thus, we have 
\begin{align*}
\ex{\psi(W^{(t + 1)})} &= \ex{\psi(W^{(t)}) + \eta_t \nabla \psi(\xi) \cdot \vtt}\\
&\ge \ex{\psi(W^{(t)})} - \gamma \eta_t ^2 + c \eta_t \ex{\| \nabla \psi(W^{(t)})\|}
\end{align*}
By letting $\eta_t = \frac{c}{2\gamma}\ex{\| \nabla \psi(W^{(t)})\|}$, we see that 
\begin{equation*}
\ex{\psi(W^{(t + 1)})} \ge \ex{\psi(W^{(t)})} + \frac{c^2}{4\gamma} \left(\ex{\|\nabla \psi(W^{(t)})\|}\right)^2
\end{equation*}

We define $\phi_{t} = \sup_W \psi(W) - \ex{\psi(W^{(t)})}$.
By the above argument, we have 
\begin{equation}
    \label{app:eq-convergence-analysis}
  \phi_{t+1} \le \phi_{t}- \frac{c^2}{4\gamma} \left(\ex{\|\nabla \psi(W^{(t)})\|}\right)^2  .
\end{equation}
Then, we can prove the statement by a similar argument to Theorem 1 in \cite{wright2015coordinate} as follows.
Let $W^{*} = argmax_{W}\psi(W)$.
By the concavity of $\psi$ and the definition of $R_0$, we have
\begin{equation*}
  \psi(W^{*}) - \psi(W^{(t)}) \le \|\nabla \psi(W^{(t)})\| \|W^{*} - W^{(t)} \| \le \| \nabla \psi(W^{(t)})\| R_0 .
\end{equation*}
Therefore, we have
\begin{equation*}
    \phi_{t} \le  R_0  \ex{\| \nabla \psi(W^{(t)})\|}.
\end{equation*}
By this inequality and \eqref{app:eq-convergence-analysis},
we have 
\begin{equation*}
    \phi_t - \phi_{t+1} \ge A \phi_{t}^2,
\end{equation*}
where $A = \frac{c^2}{4\gamma R_0^2}$.
Thus, noting that $\phi_{t+1} \le \phi_{t}$ holds by \eqref{app:eq-convergence-analysis}, we have
\begin{align*}
 \frac{1}{\phi_{t+1}} - \frac{1}{\phi_t} 
 = \frac{\phi_t - \phi_{t+1}}{\phi_t \phi_{t+1}}
 \ge \frac{\phi_{t}-\phi_{t+1}}{\phi_t^2}\ge A.
\end{align*}
Therefore, it follows that 
\begin{equation*}
    \frac{1}{\phi_t} \ge \frac{1}{\phi_1} + (t -1) A \ge (t -1) A.
\end{equation*}
This completes the proof.
\end{proof}

\subsection{A Formal Statement of Theorem \ref{thm:gain-approx} and Remarks on Non-differentiability}
\label{app:sec-proof-of-approx}
We provide a more formal statement of Theorem \ref{thm:gain-approx} (Theorem \ref{thm:approx-formula}) 
and provide its proof.
\begin{theorem}
    \label{thm:approx-formula}
    For a matrix $A \in \RR^{n\times m}$, we denote by $\|A\|_\frb$ the Frobenius norm of $A$. 
    We fix the iteration of the gradient descent and assume that the weight $\weightmat$ takes the value $\weightmat^{(0)}$
    and $\tC$ takes the value $\tC^{(0)}$. 

    We assume that the following inequality holds for all $k \in [K]$ and $l \in [K]$ uniformly $\weightmat \in \mN_0$,
    where $\mN_0$ is an open neighbourhood of $\weightmat^{(0)}$:
    \begin{equation*}
        |\ex[x_k]{\softmax_l(\weightmat^\trn g(x_k))} - \softmax_l(\tC_k)| \le \varepsilon.
    \end{equation*}
    We also assume that on $\mN_0$, $\tC$ can be regarded as a smooth function of $\weightmat$
    and the Frobenius norm of the Hessian is bounded on $\mN_0$.
    Furthermore, we assume that the following small variance assumption with $\widetilde{\varepsilon} > 0$ for all $k$:
    \begin{equation*}
        \sum_{m=1}^{K}\var[x_k]{(\weightmat^\trn g(x_k))_m} \le \widetilde{\varepsilon}.
    \end{equation*}
    Then if $\|\Delta \weightmat\|_\frb$ is sufficiently small, there exist a positive constant $c > 0$ depending only on $K$ 
    with $c = O(\mathrm{poly}(K))$
    and a positive constant $c' > 0$ such that the following inequality holds:
    \begin{equation*}
       \left|
        G_{ij} - \sum_{k=1}^{K}
        \frac{\partial \psi}{\partial \tC_k} (\Delta \weightmat)^\trn z_k
        \right| 
        \le 
        c\left \|\frac{\partial \psi}{\partial C}\right \|_{\frb }
        (\varepsilon + \widetilde{\varepsilon}) + 
        c'(\|\Delta \weightmat\|_\frb^2 + \|\Delta \tC\|_\frb^2 ).
    \end{equation*}
    Here $\Delta \weightmat = \tilde{\nu}_{ij}^t$ and $\tC_k$ is a column vector such that the $k$-th row of $\tC$ is given as $\tC_k$,
    and we consider Jacobi matrices at $\tC=\tC^{(0)}$ and the corresponding value of $C$.
\end{theorem}

We provide some remarks below.

\noindent \textbf{Independence of the choices of $\widetilde{C}$}.
Although the matrix $\widetilde{C}$ is not uniquely determined since the softmax function is not one-to-one,
the approximation (the first term of the RHS of \eqref{eq:gain-approx-exact}) is unique as the derivative of all these are unique.
We explain this more in detail.
In the approximation formula, we only need jacobian
$\partial \psi/\partial \widetilde{C} = \frac{\partial \psi}{\partial {C}} \frac{\partial {C}}{\partial{\widetilde{C}}}$.
We note that gradient of the softmax function can also be written as a function of the softmax function (i.e., $\frac{\partial \sigma_l}{\partial \xi_m} = \delta_{lm} \sigma_l(\xi) - \sigma_l(\xi) \sigma_m(\xi)$, where $\sigma_l(\xi) = \softmax_l(\xi)$
for $\xi \in \mathbb{R}^K$, $1 \le l, m \le K$).
Therefore, the first term of the RHS of eq \eqref{eq:gain-approx-exact} is uniquely determined even if $\widetilde{C}$ is not uniquely determined.

\noindent \textbf{Non-smoothness of the indicator functions}.
In Theorem \ref{thm:approx-formula}, we assume that $C$ and $\tC$ as smooth functions $W$, however strictly speaking this assumption does not hold
since the indicator functions are not differentiable.
Thus, in the definition of $G_{ij}$, we used surrogate functions of the indictor functions.
In the following, we provide more detailed explanation.
In Eq. \eqref{eq:exact-gains}, we define the gain $G_{ij}$ by a directional derivative of $\psi(C)$ 
with respect to weight $W$. However, strictly speaking, since the definition of the confusion matrix $C$
involves the indicator function, $\psi(C)$ is not a differentiable function of $W$. 
Moreover, even if gradients are defined, 
they vanish because of the definition of the indicator function.
In the assumption of Theorem \ref{thm:approx-formula} (a formal version of Theorem \ref{thm:gain-approx}), we assume $\widetilde{C}$ 
is a smooth function of $W$ and it implies $C$ is a differentiable function of $W$.
This assumption can be satisfied if
we replace the indicator function by surrogate functions of the indicator functions 
in the definition of the confusion matrix $C$.
More precisely, 
we replace the definition of $C_{ij}[h] = \pi_i \ex[x \sim P_i]{\indc(h(x) = j)}$ by $\pi_i \ex[x \sim P_i]{s_{j}(f(x))}$. Here $h(x) = \argmax_{k} f_k(x)$ as before, $P_i$ is the class conditional distribution $P(x | y=i)$ and 
$s_j$ is a surrogate function of $p \mapsto \indc(\argmax_{i} p_i = j)$ satisfying $0 \le s_j(p) \le 1$ for any $1 \le j \le K$, $p \in \Delta_{K-1}$
and $\sum_{j=1}^{K}s_j(p) = 1$ for any $p \in \Delta_{K-1}$.
To compute $G_{ij}$, one can directly use the definition of Eq. \eqref{eq:exact-gains} with the smoothed confusion matrix using surrogate functions of the 
indicator function. However, an optimal choice of the surrogate function is unknown.
Therefore, in this chapter, we introduce an unconstrained confusion matrix $\tilde{C}$ 
and the approximation formula Theorem \ref{thm:gain-approx} (Theorem \ref{thm:approx-formula}).
An advantage of introducing $\widetilde{C}$ and the approximation formula is that 
the RHS of the approximation formula
$\sum_{k,l}\frac{\partial \psi(\Tilde{C})}{\partial \tilde{C}_{kl}}\left(({\nu}_{ij})^{\top}_l \cdot z_k \right)$
does not depend on the choice of the surrogate function if we use formulas provided in Sec. \ref{app:unconstrained-derivative} with the original (non-differentiable) definition of $C$ (error terms such as $\varepsilon$ in Theorem \ref{thm:approx-formula} do depend on the surrogate function).
Since the optimal choice of the surrogate function is unknown, this gives a reliable approximation.

\subsection{Proof of Theorem \ref{thm:approx-formula}}
\begin{proof}
    In this proof, to simplify notation, we denote $\softmax(z)$ by $\sigma(z)$ for $z \in \RR^K$.
    In this proof, 
    we fix the iteration of the gradient descent and assume that the weight $\weightmat$ takes the value $\weightmat^{(0)}$
    and $\tC$ takes the value $\tC^{(0)}$. 
    We assume in an open neighborhood of $\weightmat^{(0)}$, we have a smooth correspondence $\weightmat \mapsto \tC$ and
    that if the value of $\weightmat$ changes from $\weightmat_0$ to $\weightmat_0 + \Delta \weightmat$, then $\tC$ changes from
    $\tC_0$ to $\tC_0 + \Delta \tC$.
    To prove the theorem, we introduce the following three lemmas.
    We note that by the assumption of the theorem and Lemma \ref{lem:approx-exchange}, 
    the assumption \eqref{eq:lem-ineq-zi-tc} of Lemma \ref{lem:approx-lem3} can be satisfied with 
    \begin{equation*}
        \varepsilon_1 = c'' (\varepsilon + \widetilde{\varepsilon}),
    \end{equation*}
    where $c''>0$ is a constant depending only on $K$ with $c'' = O(\mathrm{poly}(K))$.
    Then by Lemma \ref{lem:approx-lem3}, 
    there exist constants $c_1 = c_1(K)$ and $c_2 = c_2(K)$ depending on only $K$ with $c_1, c_2 = O(\mathrm{poly}(K))$
    such that the following inequality holds for all $k$:
    \begin{equation}
        \label{eq:lem-proof-ctilde-nuzk}
        \left\|
            \left. \frac{\partial C}{\partial \tC_k}\right|_{\tC_k = \tC_k^{(0)}}
        \left( 
            \Delta \tC_k - (\Delta \weightmat)^\trn z_k
        \right)\right\|_\frb \le c_1 \varepsilon_1 + c_2 (\|\Delta \tC_k\|^2_\frb + \|(\Delta \weightmat)^\trn z_k\|_\frb^2).
    \end{equation}
    Then, we have the following: 
    \begin{align*}
        \left|\frac{\partial \psi}{\partial \tC}\Delta \tC - 
        \sum_{k=1}^{K}
        \frac{\partial \psi}{\partial \tC_k} (\Delta \weightmat)^\trn z_k
        \right| &=
        \left|
        \frac{\partial \psi}{\partial C} 
        \frac{\partial C}{\partial \tC}
        \Delta \tC
        -
        \sum_{k=1}^{K}
        \frac{\partial \psi}{\partial C}\frac{\partial C}{\partial \tC_k} (\Delta \weightmat)^\trn z_k
        \right|\\
        &=
        \left|
            \frac{\partial \psi}{\partial C}\sum_{k=1}^{K} 
            \frac{\partial C}{\partial \tC_k} \Delta \tC_k-
        \sum_{k=1}^{K}
        \frac{\partial \psi}{\partial C}\frac{\partial C}{\partial \tC_k} (\Delta \weightmat)^\trn z_k
        \right|\\
        &\le 
        \left\|\frac{\partial \psi}{\partial C}\right\|_{\frb }
        \left\|
            \sum_{k=1}^{K}
            \frac{\partial C}{\partial \tC_k} \left(
                \Delta \tC_k-  (\Delta \weightmat)^\trn z_k
            \right)
        \right\|_\frb
    \end{align*}
    Here, by fixing an order on $[K] \times [K]$, we regard $\frac{\partial \psi}{\partial \tC}$,  
    $\frac{\partial C}{\partial \tC}$ and $\Delta \tC$
    as a $K^2$-dimensional row vector, a $K^2\times K^2$-matrix, and a $K^2$-dimensional column vector, respectively.
    Moreover, we consider Jacobi matrices at $\tC = \tC^{(0)}$.
    Then, the assertion of the theorem from this inequality, \eqref{eq:lem-proof-ctilde-nuzk}, 
    Lemma \ref{lem:approx-jac}.
\end{proof}

\begin{lemma}
    \label{lem:approx-jac}
    Under assumptions and notations in the proof of Theorem \ref{thm:approx-formula},
    there exists a constant $c > 0$ 
    such that 
    \begin{equation*}
         \left|G_{ij} -  \left.\frac{\partial \psi}{\partial \tC}\right|_{\tC=\tC^{(0)}}\Delta \tC \right| 
         \le
         c \| \Delta \weightmat\|_{\frb}^2.
    \end{equation*}
\end{lemma}
\begin{proof}
    By the assumption of the mapping $\weightmat \mapsto \tC$ and the Taylor's theorem, there exists $c_1 > 0$ such that 
    \begin{equation}
        \label{eq:ctilde-nu-jac-approx}
        \left\| \Delta \tC - 
        \left(\left.\frac{\partial \tC}{\partial \weightmat} \right)\right |_{\weightmat = \weightmat_0} \Delta \weightmat \right\|_{\frb}
        \le c_1 \| \Delta \weightmat \|_\frb^2.
    \end{equation}
    By definition of $G_{ij}$, we have the following:
    \begin{align*}
        \left|G_{ij} -  \left.\frac{\partial \psi}{\partial \tC}\right|_{\tC=\tC^{(0)}}\Delta \tC \right| 
        &=
      \left |\frac{\partial \psi}{\partial \weightmat} \Delta \weightmat - \frac{\partial \psi}{\partial \tC}\Delta \tC  \right|
      \\
      &= 
      \left | \frac{\partial \psi}{\partial \tC} 
      \frac{\partial \tC}{\partial \weightmat} \Delta \weightmat 
      - \frac{\partial \psi}{\partial \tC} \Delta \tC
      \right | \\
      &\le 
      c_1 \left \| \frac{\partial \psi}{\partial \tC} \right \|_\frb  \left \| \Delta \weightmat \right \|_\frb^2.
    \end{align*}
    Here we consider Jacobi matrices at $\weightmat = \weightmat_0$ and corresponding values.
    The last inequality follows from the fact that 
    the matrix norm $\| \cdot\|_\frb$ is sub-multiplicative and Eq. \eqref{eq:ctilde-nu-jac-approx}.
\end{proof}

\begin{lemma}
    \label{lem:approx-exchange}
    Under assumptions and notations in the proof of Theorem \ref{thm:approx-formula},
    there exist a positive constant $c = c(K)$ depending only on $K$ with $c = O(\mathrm{poly}(K))$
    such that:
    \begin{equation*}
        |\ex{\sigma_l(\weightmat^\trn g(x_k))} -
        \sigma_l(\weightmat^\trn z_k)
        | \le c \sum_{m=1}^{K}\var{(\weightmat^\trn g(x_k))_m},
    \end{equation*}
    for any $1 \le k, l \le K$.
\end{lemma}
\begin{proof}
    This can be proved by applying the Taylor's theorem to $\sigma_l$.
    We fix $k, l$ and apply the Taylor's theorem to the function $\xi \mapsto \sigma_l(\xi)$ 
    at $\xi = \weightmat^\trn z_k = \weightmat^\trn \ex{g(x_k)}$.
    Then there exists $\xi_0 \in \RR^K$ such that 
    \begin{equation*}
        \sigma_l(\xi) = 
        \sigma_l(\weightmat^\trn z_k) + 
        \left. \frac{\partial \sigma_l}{\partial \xi}\right|_{\xi=\weightmat^\trn z_k}(\xi - \weightmat^\trn z_k) + 
        \frac{1}{2}
        (\xi - \weightmat^\trn z_k)^\trn H_k
        (\xi - \weightmat^\trn z_k),
    \end{equation*}
    where $H_k = 
    \left. \frac{\partial^2 \sigma_l}{\partial \xi^2} \right|_{\xi=\xi_0}$.
    By noting that $\frac{\partial \sigma_l}{\partial \xi_m} = \delta_{lm} \sigma_l(\xi) - \sigma_l(\xi) \sigma_m(\xi)$ 
    (here $\delta_{lm}$ is the Kronecker's delta), it is easy to see that there exists a constant $c'_{l}$ 
    depending only on $l$ and $K$
    such that $\|H\|_\frb < c'_{l}$ and $c'_{l} = O(\mathrm{poly}(K))$.
    By letting $\xi = \weightmat^\trn g(x_k)$ in the above equation and taking the expectation of the both sides,
    we obtain the assertion of the lemma with $c=\frac{1}{2}\max_{l \le [K]}c_{l}'$.
\end{proof}

\begin{lemma}
    \label{lem:approx-lem3}
    Under assumptions and notations in the proof of Theorem \ref{thm:approx-formula},
    we assume there exists $\varepsilon_1 > 0$ such that 
    the following inequality holds for all $k$ and $l$ for any $\weightmat$ in an open neighborhood of $\weightmat^{(0)}$ and corresponding $\tC$:
    \begin{equation}
        \label{eq:lem-ineq-zi-tc}
        \left | \sigma_l(\weightmat^\trn z_k) -  
        \sigma_l(\tC_k)
        \right | \le \varepsilon_1.
    \end{equation}
    Furthermore, we assume that $\|(\Delta \weightmat)^\trn z_k\|_\frb$ is sufficiently small for all $k$.
    Then there exist constants $c_1 = c_1(K)$ and $c_2 = c_2(K)$ depending on only $K$ with $c_1, c_2 = O(\mathrm{poly}(K))$
    such that 
    \begin{equation*}
        \left\|
            \left. \frac{\partial C}{\partial \tC_k}\right|_{\tC_k = \tC_k^{(0)}}
        \left( 
            \Delta \tC_k - (\Delta \weightmat)^\trn z_k
        \right)\right\|_\frb \le c_1 \varepsilon_1 + c_2 (\|\Delta \tC_k\|^2_\frb + \|(\Delta \weightmat)^\trn z_k\|_\frb^2).
    \end{equation*}
    Here, 
    $\tC_k$ (resp. $\Delta \tC_k$) is a column vector such that the $k$-th row vector of $\tC$ (resp. $\Delta \tC$) is given as
    $\tC_k$ (resp. $\Delta \tC_k$).
    Moreover, 
    when defining Jacobi matrices,
    we regard $C$ as a $K^2$-vector
    and consider a $K^2\times K$ Jacobi matrix $\left. \frac{\partial C}{\partial \tC_k}\right|_{\tC_k = \tC_k^{(0)}}$ at $\tC_k = \tC_k^{(0)}$.
\end{lemma}
\begin{proof}
    Since \eqref{eq:lem-ineq-zi-tc} holds all $\weightmat$ in an open neighborhood of $\weightmat^{(0)}$ and corresponding $\tC$, 
    we apply the Taylor's theorem to the function $\xi \mapsto \sigma_l(\xi)$ at $\xi = (\weightmat^{(0)})^\trn z_k$ and 
    $\xi = \tC_k^{(0)}$. Then by \eqref{eq:lem-ineq-zi-tc} and the same argument in the proof of Lemma \ref{lem:approx-exchange},
    we have
    \begin{equation*}
        \left |
        \left.
            \frac{\partial \sigma_l} {\partial \xi} 
        \right|_{\xi=\mu_k} \Delta \mu_k -
        \left. \frac{\partial \sigma_l}{\partial_\xi}\right|_{\xi=\tC_k^{(0)}} \Delta \tC_k
        \right | \le \varepsilon_1 + c_2'(\|\Delta \mu_k\|_\frb^2 + \|\Delta \tC_k\|_\frb^2),
    \end{equation*}
    where $\mu_k = (\weightmat^{(0)})^\trn z_k$,
    $\Delta \mu_k = (\Delta \weightmat)^\trn z_k$.
    Noting that $(\frac{\partial \sigma_l}{\partial \xi})_m$ is given as $\delta_{ml} \sigma_l(\xi) - \sigma_m(\xi)\sigma_l(\xi)$,
    \eqref{eq:lem-ineq-zi-tc} and the assumption that $\|\Delta \mu_k\|_\frb$ is sufficiently small, 
    we see that there exists a constant $c_1', c_2'>0$ depending only on $K$
    with $c_1', c_2' = O(\mathrm{poly}(K))$ such that the following inequality holds:
    \begin{equation}
        \label{eq:lem-delmu-deltilde}
        \left|
        \left. 
            \frac{\partial \sigma_l} {\partial \xi} 
        \right|_{\xi=\tC_k^{(0)}}
        \left(
        \Delta \mu_k - \Delta \tC_k
        \right)
        \right|
        \le c_1' \varepsilon_1 + c_2'(\|\Delta \mu_k\|_\frb^2 + \|\Delta \tC_k\|_\frb^2).
    \end{equation}
    Next, we consider entries of the $K^2$-vector 
    $
    \left.  \frac{\partial C}{\partial \tC_k}
        \right|_{\tC_k=\tC_k^{(0)}} (\Delta \tC_k - \Delta \mu_k)$.
    Here as previously mentioned by fixing an order on $[K] \times [K]$, we regard 
    $        \frac{\partial C}{\partial \tC_k}$ as a $K^2 \times K$-matrix.
    For $(k, l) \in [K] \times [K]$, by the definition of the mapping $\tC \mapsto C$,
    $(k, l)$-th entry of 
    $\left.  \frac{\partial C}{\partial \tC_k}
    \right|_{\tC_k=\tC_k^{(0)}} (\Delta \tC_k - \Delta \mu_k)$ 
    is given as $\pi_k \left.\frac{\partial \sigma_l}{\partial \xi}\right|_{\xi=\tC_k^{(0)}} (\Delta \tC_k - \Delta \mu_k)$.
    By \eqref{eq:lem-delmu-deltilde}, we see that there exist constants $c_1'', c_2''$ depending only on $K$ and 
    $c_1'', c_2'' = O(\mathrm{poly}(K))$ such that 
    \begin{equation*}
        \left\|
        \left. \frac{\partial C}{\partial \tC_k}\right|_{\tC_k = \tC_k^{(0)}}
        \left( 
            \Delta \tC_k - (\Delta \weightmat)^\trn z_k
        \right)\right\|_\frb \le c_1'' \varepsilon_1 + c_2'' (\|\Delta \tC_k\|^2_\frb + \|(\Delta \weightmat)^\trn z_k\|_\frb^2).
    \end{equation*}
    Since constants $c_1'', c_2''$ may depend on $(k, l)$ by taking $c_1 = \max_{(k, l)}c_1''$ and $c_2 = \max_{(i, l)} c_2''$, 
    we have the assertion of the lemma.
\end{proof}

\subsection{Validity of the Mixup Sampling Distribution}
\label{sec:app-analysis-of-var-of-selmix-policy}
In this section, 
Motivated by Algorithm \ref{alg:ours}, we consider the following online learning problem and 
prove validity of our method.
For each time step $t=1,\dots, T$, an agent selects pairs $(i^{(t)}, j^{(t)}) \in [K] \times [K]$,
where random variables $(i^{(t)}, j^{(t)})$ follows a distribution $\mathcal{P}^{(t)}$ on $[K] \times [K]$.
We call a sequence of distributions $(\mathcal{P}^{(t)})_{t=1}^{T}$ a policy.
For $(i, j) \in [K] \times [K]$ and $1 \le t \le T$, we assume that random variable $G^{(t)}_{ij}$ 
is defined. 
We regard $G^{(t)}_{ij}$ as the gain in the metric when performing $(i, j)$-mixup at iteration $t$
in Algorithm \ref{alg:ours}.
We assume that $G^{(t)}_{ij}$ is random variable 
due to randomness of the validation dataset, $X_1, X_2$, and the policy.
Furthermore, we assume that when selecting $(i^{(t)}, j^{(t)})$, the agent observes random variables 
$G^{(t)}_{ij}$ for $(i, j) \in [K] \times [K]$
but cannot observe the true gain defined by $\ex{G^{(t)}_{ij}}$.
The average gain $\overline{G}^{(T)}(\boldsymbol{\mathcal{P}})$ of a policy $\boldsymbol{\mathcal{P}} = (\mathcal{P}^{(t)})_{t=1}^{T}$ is defined as 
\begin{math}
   \overline{G}^{(T)}(\boldsymbol{\mathcal{P}}) = \frac{1}{T}\sum_{t=1}^T \ex{G^{(t)}_{i^{(t)}j^{(t)}}},
\end{math}
where $(i^{(t)}, j^{(t)})$ follows the distribution $\mathcal{P}^{(t)}$ and 
the expectation is taken with respect to the randomness of the policy,
validation dataset, $X_1, X_2$.
This problem setting is similar to that of Hedge
\cite{freund1997decision} (i.e., online convex optimization).
However, in the problem setting of Hedge, the agent observes gains (or losses) 
after performing an action but in our problem setting, the agent have random estimations of the gains before performing an action.
We note that even in this setting, methods such as $\mathrm{argmax}$ with respect to $G^{(t)}_{ij}$
may not perform well due to randomness of $G^{(t)}_{ij}$ and errors in the approximation (Refer Sec. \ref{sec:empirical_results} for evidence).

We call a policy $\boldsymbol{\mathcal{P}} =(\mathcal{P}^{(t)})_{t=1}^{T}$ non-adaptive (or stationary) if $\mathcal{P}^{(t)}$ is the same for all $t=1, \dots, T$, i.e,
if there exists a distribution $\mathcal{P}^{(0)}$ on $[K] \times [K]$ such that $\mathcal{P}^{(t)} = \mathcal{P}^{(0)}$ for all $t=1, \dots, T$.
A typical example of non-adaptive policies is the uniform mixup, i.e.,  $\mathcal{P}^{(t)}$ is the uniform distribution on $[K] \times  [K]$.
Another example is $\mathcal{P}^{(t)} = \delta_{(i^{(0)}, j^{(0)})}$ 
for a fixed $(i^{(0)}, j^{(0)}) \in [K]\times [K]$ (i.e., the agent performs the fixed $(i^{(0)}, j^{(0)})$-mixup in each iteration).
Similarly to Hedge \citep{freund1997decision} and EXP3 \cite{auer2002}, we define 
$\boldsymbol{\mathcal{P}}_\text{SelMix} = (\mathcal{P}^{(t)}_\text{SelMix})_{t=1}^T$ by 
$\mathcal{P}^{(t)}_\text{SelMix} = \softmax((\invtemp \sum_{\tau=1}^t G^{(\tau)}_{ij})_{1\le i, j \le K})$,
where $\invtemp > 0$ is the inverse temperature parameter.
The following theorem states that $\bmP_\selmix$ is better than any non-adaptive policy in terms of the average expected gain
if $T$ is large:
\begin{theorem}
    \label{thm:regbound}
    We assume that $G^{(t)}_{ij}$ is normalized so that $|G^{(t)}_{ij}| \le 1$.
    Then, 
    for any non-adaptive policy $\boldsymbol{\mathcal{P}}^{(0)} = (\mathcal{P}^{(0)})_{t=1}^{T}$, we have
    \begin{math}
        \overline{G}^{(T)}(\boldsymbol{\mathcal{P}}_\text{SelMix}) + 
        \frac{2 \log K}{s T}
        \ge \overline{G}^{(T)}(\boldsymbol{\mathcal{P}}^{(0)}).
    \end{math}
\end{theorem}

\begin{proof}[Proof of Theorem \ref{thm:regbound}]
    This can be proved by standard argument of the proof of the mirror descent method (see e.g. \citep{lattimore2020bandit}, chapter 28).

    Denote by $\Delta \subset \RR^{K\times K}$ the probability simplex of dimension $K^2 - 1$.
    Let $(i_0, j_0) \in K \times K$ 
    be the best fixed mixup hindsight.
    Since any non-adaptive policy is no better than
    the best fixed mixup in terms of $\overline{G}$,
    we may assume that $\bmP^{(0)} = (\pi_0)_t$,
    where 
    $\pi_0$ is 
    the one-hot vector in $ \Delta$ 
    defined 
    as $(\pi_0)_{ij} = 1$ if $(i, j) = (i_0, j_0)$ and $0$ otherwise
    for $1 \le i, j \le K$.
    Let $F$ be the negative entropy function, i.e., $F(p) = \sum_{i,j=1}^{K}p_{ij} \log p_{ij}$.
    For $p \in \Delta$ and $G \in \RR^{K \times K}$, we define $\langle p, G \rangle = \sum_{i,j=1}^{K}p_{ij}G_{ij}$.
    Then, it is easy to see that $p^{(t)}=\mP_\selmix^{(t)}$ defined above is given as the solution of the following:
    \begin{equation}
        \label{eq:mirror-descent}
        p^{(t)} = \argmin_{p \in \Delta}
        - s\langle p, G^{(t)} \rangle + D(p, p^{(t-1)}).
    \end{equation}
    Here $D$ denotes the KL-divergence
    and we define $p^{(0)}=(1/K^2)_{1\le i ,j \le K} = \argmin_{p \in \Delta} F(p)$.
    Since the optimization problem \eqref{eq:mirror-descent} is a convex optimization problem,
    by the first order optimality condition, we have
    \begin{equation*}
        \langle \pi_0 - p^{(t)} ,  G^{(t)}\rangle 
        \le \frac{1}{s}
        \left\{
            D(\pi_0, p^{(t-1)}) - D(\pi_0, p^{(t)}) - D(p^{(t)}, p^{(t-1)})
        \right\}.
    \end{equation*}
    By summing the both sides and taking expectation, we have
    \begin{align*}
        T\overline{G}^{(T)}(\bmP^{(0)}) - T\overline{G}^{(T)}(\bmP_\selmix) 
        &\le 
        \frac{1}{s} \left\{ 
            D(\pi_0, p^{(0)}) - D(\pi_0, p^{(T)})
            - \sum_{t=1}^{T} D(p^{(t)}, p^{(t-1)})
        \right\}\\
        &\le 
        \frac{1}{s} D(\pi_0, p^{(0)}).
    \end{align*}
    Here the second inequality follows from the non-negativity of the KL-divergence. 
    Since $p^{(0)} = \argmin_{p} F(p)$, by the first-order optimality condition, we have
    $D(\pi_0, p^{(0)}) \le F(\pi_0) - F(p^{(0)})$.
    Noting that $F(\pi_0) \le 0$, we have the following
    \begin{equation*}
        T\overline{G}^{(T)}(\bmP^{(0)}) - T\overline{G}^{(T)}(\bmP_\selmix) 
        \le 
        \frac{-F(p^{(0)})}{s}
        = \frac{\log K^2}{s}.
    \end{equation*}
    This completes the proof.
\end{proof}

\subsection{A Variant of Theorem \ref{thm:regbound}}
\label{app:proof-reg}
In the case when $\bmP_{\selmix}^{(t)}$ is defined similarly to Hedge \cite{freund1997decision},
i.e., 
$\boldsymbol{\mathcal{P}}^{(t)}_\text{SelMix} = \softmax((\invtemp \sum_{\tau=1}^{t-1} G^{(\tau)}_{ij})_{ij})$,
then by the standard analysis, we can prove the following.

\begin{theorem}
    \label{thm:regbound-variant}
    We assume that $G^{(t)}_{ij}$ is normalized so that $|G^{(t)}_{ij}| \le 1$.
    Then, with an appropriate choice of the parameter $\invtemp$, 
    for any non-adaptive policy $\boldsymbol{\mathcal{P}}^{(0)} = (\mathcal{P}^{(0)})_{t=1}^{T}$, we have
    \begin{math}
        \overline{G}^{(T)}(\boldsymbol{\mathcal{P}}_\text{SelMix}) + 2\sqrt{\log K}/\sqrt{T} 
        \ge \overline{G}^{(T)}(\boldsymbol{\mathcal{P}}^{(0)}).
    \end{math}
\end{theorem}

First we introduce the following lemma, which is due to \cite{freund1997decision}.
Although, one can prove the following result by a standard argument, 
we provide a proof for the sake of completeness.
\begin{lemma}[c.f. \cite{freund1997decision}]
    \label{lem:hedge-lem}
    We assume that $G_{i, j}^{(t)} \in [0, 1]$ for all $t$ and $1\le i, j \le K$.
    For $(i, j) \in [K] \times [K]$, we define
    $\overline{S}_{i, j} = \sum_{t=1}^T \ex{G_{i, j}^{(t)}}$.
    For a policy $\bmP = (\mP_t)_{t=1}^{T}$, we define 
    $\overline{S}_{\bmP}:= \sum_{t=1}^T \ex{G_{i_t, j_t}^{(t)}}$.
    Then, we have the following inequality:
    \begin{equation*}
        - 2 \log K + \invtemp \max_{(i, j)\in [K] \times [K]} \overline{S}_{i, j} 
        \le (\exp(\invtemp) - 1) \overline{S}_{\bmP_{\selmix}}.
    \end{equation*}
\end{lemma}
\begin{proof}
    This lemma can be proved by a standard argument, but for the sake of completeness, we provide a proof.
    We put $\mA = [K] \times [K]$, $a_t = (i^{(t)}, j^{(t)})$ 
    and in the proof we simply denote $\bmP_{\selmix}$ by $\bmP$.
    For $a \in \mA$ and $1 \le t \le T + 1$, we define $w_{a, t}$ as follows.
    We define $w_{a, 1} = 1/K^2$ for all $a \in \mA$
    and $w_{a, t + 1} = w_{a, t} \exp(\invtemp G_{a}^{(t)})$.
    We also define $W_t = \sum_{a \in \mA}w_{a, t}$.
    Then, the distribution $\mP_t$ is given as the probability $(w_{a, t}/W_t)_{a \in \mA}$ by definition.
    Noting that $\exp(\invtemp x) \le 1 + (\exp(\invtemp) - 1)x$ for $x \in [0, 1]$, 
    we have the following inequality:
    \begin{align*}
        W_{t+1} &= \sum_{a\in \mA}w_{a, t+1}
        = \sum_{a \in \mA}w_{a, t} \exp(\invtemp G_{a}^{(t)})\\
        & \le \sum_{a \in \mA} w_{a, t} (1 + \exp(\invtemp- 1) G_{a}^{(t)}).
    \end{align*}
    Thus, we have
    \begin{align*}
        W_{t+1}  &\le 
        \sum_{a \in \mA} w_{a, t} (1 + \exp(\invtemp- 1) G_{a}^{(t)})\\
        &= W_t \left(1 + (\exp(\invtemp) - 1)\ex[\mP_t]{G_{a_t}^{(t)}}\right),
    \end{align*}
    where $\ex[\mP_t]{\cdot}$ denotes the expectation with respect to $a_t$.
    By repeatedly apply the inequality above, we obtain:
    \begin{equation*}
        W_{T+1} \le \prod_{t=1}^{T}
        \left( 1 + (\exp(\invtemp) - 1)\ex[\mP_t]{G_{a_t}^{(t)}} \right).
    \end{equation*}
    Let $a \in \mA$ be any pair. 
    By this inequality and $W_{T + 1} \ge w_{a, T + 1} = \frac{1}{K^2}\exp(\invtemp \sum_{t=1}^{T} G_{a}^{(t)})$,  
    we have the following:
    \begin{align*}
        \frac{1}{K^2}\exp(\invtemp \sum_{t=1}^{T} G_{a}^{(t)})
        \le \prod_{t=1}^{T} \left( 1 + (\exp(\invtemp)-1) \ex[\mP_t] {G_{a_t}^{(t)}}\right).
    \end{align*}
    By taking $\log$ of both sides and $\log(1 + x) \le x$, we have
    \begin{align*}
        -2\log K + \invtemp \sum_{t=1}^{T} G_{a}^{(t)} &\le \sum_{t=1}^{T}\log\left( 1 + (\exp(\invtemp)-1) \ex[\mP_t] {G_{a_t}^{(t)}}\right)\\
        &\le 
        (\exp(\invtemp) - 1) \sum_{t=1}^T\ex[\mP_t]{G_{a_t}^{(t)}}.
    \end{align*}
    By taking the expectation with respect to the randomness of $G_{i, j}^{(t)}$, we obtain the following:
    \begin{equation*}
        -2\log K + \invtemp \overline{S}_a \le 
        (\exp(\invtemp) -1) \overline{S}_{\bmP}.
    \end{equation*}
    Since $a \in [K] \times [K]$ is arbitrary, we have the assertion of the lemma.
\end{proof}

We can prove Theorem \ref{thm:regbound} by Lemma \ref{lem:hedge-lem} as follows:
\begin{proof}[Proof of Theorem \ref{thm:regbound}]
   Let $(i^*, j^*)$ be the best fixed Mixup pair hindsight, i.e.,
    $(i^*, j^*) = \argmax_{(i, j) \in [K] \times [K]} \overline{S}_{i, j}$.
    Since any non-adaptive (or stationary) policy is no better than $\delta_{(i^*, j^*)}$, to prove the theorem, 
    it is enough to prove the following:
    \begin{equation}
        \label{eq:restatement-regbound}
        \overline{S}_{i^*, j^*} \le \overline{S}_{\bmP} + 2\sqrt{ T\log K}.
    \end{equation}
    Here in this proof, we simply denote $\bmP_{\selmix}$ by $\bmP$.
    To prove \eqref{eq:restatement-regbound}, we define the pseudo regret $R_T$ by $R_T = \overline{S}_{i^*, j^*} - \overline{S}_{\bmP}$.
    Then by Lemma \ref{lem:hedge-lem}, we have
    \begin{equation*}
        R_T \le 
        \frac{(\exp(\invtemp) - 1 - \invtemp)\overline{S}_{i^*, j^*} + 2\log K} {\exp(\invtemp) - 1}.
    \end{equation*}
    We put $\invtemp = \log(1 + \alpha)$ with $\alpha > 0$.
    Then we have
    \begin{equation*}
        R_T \le \frac{(\alpha - \log(1 + \alpha)) \overline{S}_{i^*, j^*} + 2 \log K}{\alpha}.
    \end{equation*}
    We note that the following inequality holds for $\alpha > 0$:
    \begin{equation*}
        \frac{\alpha - \log(1 + \alpha)}{\alpha} \le \frac{1}{2}\alpha
    \end{equation*}
    Then it follows that:
    \begin{equation*}
        R_T \le \frac{1}{2} \alpha \overline{S}_{i^*, j^*} + \frac{2\log K}{\alpha}
        \le \frac{1}{2}\alpha T + \frac{2\log K}{\alpha}.
    \end{equation*}
    Here the second inequality follows from $\overline{S}_{i^*, j^*} \le T$.
    We take $\alpha = 2\sqrt{\frac{\log K}{T}}$. Then we have $R_T \le 2\sqrt{T\log K}$.
    Thus, we have the assertion of the theorem.
\end{proof}

\section{Unconstrained Derivatives of metric}
\label{app:unconstrained-derivative}
For any general metric $\psi(C[h])$ the derivative w.r.t the unconstrained confusion matrix $\Tilde{C}[h]$ is expressible purely in terms of the entries of the confusion matrix. This is because $C[h] = \text{softmax}(\Tilde{C}[h]) $  The derivative using the chain rule is expressed as follows,
\begin{equation}
    \label{eq:general-metric-derivative}
     \frac{\partial \psi(C[h])}{\partial \Tilde{C}_{ij}[h]} = \sum_{k,l} \frac{\partial \psi(C[h])}{\partial C_{kl}[h]} \cdot
     \frac{\partial C_{kl}[h]}{\partial \Tilde{C}_{ij}[h]}
\end{equation}

We observe that in Eq. \ref{eq:general-metric-derivative} the partial derivative $\frac{\partial \psi(C[h])}{\partial C_{kl}[h]}$ is purely a function of entries of $C[h]$ since $\psi(C[h])$ itself is a function of entries of $C[h]$. The second term is the partial derivative of our confusion matrix w.r.t the unconstrained confusion matrix. Since $C$ and $\Tilde{C}$ are related by the following relation $C_{ij}[h] = \text{softmax}(\Tilde{C}_i[h])_j$. By virtue of the aforementioned map $\frac{\partial C_{kl}[h]}{\partial \Tilde{C}_{ij}[h]}$  also happens to be expressible in terms of $C[h]$:
\begin{equation}
    \label{eq:map-derivative}
     \frac{\partial C_{kl}[h]}{\partial \Tilde{C}_{ij}[h]} = 
    \begin{cases}
    \qquad  0, \qquad \quad  \quad  \qquad k \neq i \\
    
   -\frac{C_{il}[h] \cdot C_{ij}[h]}{ \pi^{\text{val}}_{i}}, \qquad \quad i= k, l \neq j \\
   
    C_{ij}[h] - \frac{C_{ij}^2[h]}{ (\pi^{\text{val}}_{i})^2}, \quad \quad \ i =  k, l = j \\
    \end{cases}
\end{equation}

Let us consider the metric mean recall $\psi^{\text{AM}}(C[h]) = \frac{1}{K}\sum_i \frac{C_{ii}[h]}{\sum_j C_{ij}[h]}$. The derivative of $\psi^{\text{AM}}(C[h])$ w.r.t the unconstrained confusion matrix $\Tilde{C}$ can be expressed in terms of the entries of the confusion matrix. This is a useful property of this partial derivative since we need not infer the inverse map from $C \rightarrow \Tilde{C}$ inorder to evaluate the partial derivative in terms of $\Tilde{C}$. It can be expressed follows:
\begin{equation}
    \label{eq:mean-recall-derivative}
     \frac{\partial \psi^{\text{AM}}(C[h])}{\partial \Tilde{C}_{ij}[h]} = 
    \begin{cases}
   -\frac{C_{ij}[h] \cdot C_{ii}[h]}{K (\pi^{\text{val}}_{i})^2}, \qquad \qquad m\neq n \\
    \frac{C_{ii}[h]}{K \cdot \pi^{\text{val}}_{i}}-\frac{C_{ii}^2[h]}{K \cdot (\pi^{\text{val}}_{i})^2}, \ \ \qquad m=n \\
    \end{cases}
\end{equation}
Hence we can conclude that for a metric defined as a function of the entries of the confusion matrix, the derivative w.r.t the unconstrained confusion matrix ($\Tilde{C}$) is easily expressible using the entries of the confusion matrix ($C$). 

\section{Comparison of SelMix with Other Frameworks}
\label{app:comparison}
In our work, we optimize the non-decomposable objective function by using Mixup~\cite{zhang2018mixup}. In recent works, Mixup training has been shown to be effective specifically for real-world scenarios where the long-tailed imbalance is present in the dataset~\cite{zhong2021mislas, fan2021cossl}. Further, mixup has been demonstrated to have a data-dependent regularization effect~\cite{zhang2021does}. Hence, this provides us the motivation to consider optimization of the non-decomposable objectives which are important for long-tailed imbalanced datasets, in terms of directions induced by Mixups.However, this mixup-induced data-dependent regularization is not present for works~\cite{narasimhan2021training, rangwani2022costsensitive}, which use consistent loss functions without mixup. Hence, this explains the superior generalization demonstrated by SelMix (mixup based) on non-decomposable objective optimization for long-tailed datasets.

\section{Updating the Lagrange multipliers}
\subsection{Min. Recall and Min of Head and Tail Recall}
Consider the objective of optimizing the worst-case(Min.) recall, 
$$\psi^{\text{MR}}(C[h]) =  \min_{\bm{\lambda} \in\Delta_{K-1}} \sum_{i \in [K]}\lambda_i \text{Rec}_i[h] =  \min_{\bm{\lambda} \in\Delta_{K-1}} \sum_{i \in [K]}\lambda_i \frac{C_{ii}[h]}{\sum_{j\in[K]}{C_{ij}[h]}}$$, as in Table \ref{tab:metric_full}. the Lagrange multipliers are sampled from a $K-1$ dimensional simplex and $\lambda_i = 1$ if recall of $i^{\text{th}}$ class is the lowest and the remaining lagrange multiplers are zero. Hence, a good approximation of the lagrange multipliers at a given time step $t$ can be expressed as:

\begin{equation}
    \label{eq:MinRecall-lagrange}
    \lambda_i^{(t)} = \frac{e^{-\omega \text{Rec}_i^{(t)}[h]}}{\sum_{j \in [K]} e^{-\omega \text{Rec}_j^{(t)}[h]}}
\end{equation}

This has some nice properties such as the Lagrange multipliers being a soft and momentum-free approximation of their hard counterpart. This enables SelMix to compute a sampling distribution $\mathcal{P}_{\text{SelMix}}^{(t)}$ that neither over-corrects nor undercorrects based on the feedback from the validation set. For sufficiently high $\omega$ this approximates the objective to the min recall.

\subsection{Mean recall under Coverage constraints}
For the objective where we wish to optimize for the mean recall, subject to the constraint that all the classes have a coverage above $\frac{\alpha}{K}$, where if $\alpha =1$ 
is the ideal case for a balanced validation set. We shall relax this constraint and set $\alpha=0.95$, $\psi^{\text{AM}}_{\text{cons.}}(C[h]) \text{ = } \min_{\bm{\lambda} \in \mathbb{R}^K_+} \frac{1}{K} \sum_{i \in [K]} \text{Rec}_i[h] + \sum_{j \in [K]}\lambda_j \left(\text{Cov}_j[h] - \frac{\alpha}{K} \right) \text{ = } \min_{\bm{\lambda} \in \mathbb{R}^K_+}\frac{1}{K} \sum_{i \in [K]} \frac{C_{ii}[h]}{\sum_{j\in[K]}{C_{ij}[h]}}$ $+ \sum_{j \in [K]}\lambda_j \left(\sum_{i \in [K]} C_{ij}[h] - \frac{\alpha}{K} \right)$.
For practical purposes, we look at a related constrained optimization problem, 

$$\psi^{\text{AM}}_{\text{cons.}}(C[h]) =  \min_{\bm{\lambda} \in \mathbb{R}^K_+} \frac{1}{\lambda_{max}+1}\left(\frac{1}{K}\sum_{i \in [K]} \text{Rec}_i[h] + \sum_{j \in [K]}\lambda_j \left(\text{Cov}_j[h] - \frac{\alpha}{K} \right) \right)$$
Such that if $\left(\text{Cov}_i[h] - \frac{\alpha}{K} \right) < 0$, then $\lambda_i$ increases, and vice-versa for the converse case. Also, if $\exists i$ s.t. $\left(\text{Cov}_i[h] - \frac{\alpha}{K} \right) < 0$, then this implies that $\frac{1}{\lambda_{\text{max}}+1} \rightarrow 0^{+} \text{ and } \frac{\lambda_{\text{max}}}{\lambda_{\text{max}+1}} \rightarrow 1^-$, which forces $h$ to satisfy the constraint $\left(\text{Cov}_i[h] - \frac{\alpha}{K} \right) > 0$. 
Based on this, a momentum free formulation for updating the lagrangian multipliers is as follows:
$$\lambda_i = \text{max} \left( 0, \Lambda_{\text{max}}\left(1-e^{ \frac{\text{Cov}_i[h] - \frac{\alpha}{K}}{\tau}}\right)\right)$$

Here, $\lambda_{max}$ is the maximum value that the Lagrange multiplier can take. A large value of  $\lambda_{max}$ forces the model to focus more on the coverage constraints that to be biased towards mean recall optimization. $\tau$ is a hyperparameter that is usually kept small, say $0.01$ or so, which acts as sort of a tolerance factor to keep the constraint violation in check.

\section{Experimental details}

The baselines (Table~\ref{tab:matched-results}) are evaluated with the SotA base pre-training method of FixMatch + LA using DASO codebase~\cite{oh2022daso}, whereas CSST~\cite{rangwani2022costsensitive} is done through their official codebase .
\subsection{Hyperparameter Table}
The detailed values of all hyperparameters specific to each dataset are in Table \ref{tab:hyperparams}.
\begin{table}[H]
\centering
\caption{Table of Hyperparameters for Semi-Supervised datasets.}
\begin{adjustbox}{max width=\textwidth}
\begin{tabular}{cccccc} 
\toprule
    Parameter & \begin{tabular}{c} CIFAR-10 \\ (All distributions) \end{tabular}    & \begin{tabular}{c} CIFAR-100 \\ ($\rho_l=10, \rho_u=10$) \end{tabular} & STL-10 &  Imagenet-100($\rho_l=\rho_u=10$)\\
     \hline
    Gain scaling ($s$) & 10.0 & 10.0 & 2.0 & 10.0\\
    $\omega_{\text{Min. Rec}}$ & 40 & 20 & 20 & 20\\
    $\lambda_{\max}$ & 100 & 100 & 100 & 100\\
    $\tau$ & 0.01 & 0.01 & 0.01 & 0.01\\
    $\alpha$ & 0.95 & 0.95 & 0.95 & 0.95  \\
    Batch Size & 64 & 64 & 64 & 64 \\
    Learning Rate($f$) & 3e-4 & 3e-4 & 3e-4 & 0.1\\
    Learning Rate($g$) & 3e-5 & 3e-5 & 3e-5 & 0.01 \\
    Optimizer & SGD & SGD & SGD & SGD \\
    Scheduler & Cosine & Cosine & Cosine & Cosine \\
    Total SGD Steps & 10k & 10k & 10k & 10k  \\
    Resolution & 32 X 32 & 32 X 32 & 32 X 32 & 224 X 224 \\
    Arch. & WRN-28-2 & WRN-28-2 & WRN-28-2 & WRN-28-2 \\ 
    \hline
\end{tabular}
\end{adjustbox}

\label{tab:hyperparams}
\end{table}

\begin{table}[H]
\centering
\caption{Table of Hyperparameters for Supervised Datasets.}
\begin{adjustbox}{max width=\textwidth}
\begin{tabular}{cccc} 
\toprule
    Parameter &  \begin{tabular}{c} CIFAR-10 \\ ($\rho=100$) \end{tabular} & \begin{tabular}{c} CIFAR-100 \\ ($\rho=100$) \end{tabular} & Imagenet-1k LT\\
     \hline
    Gain scaling ($s$) & 10.0 & 10.0 & 10.0\\
    $\omega_{\text{Min. Rec}}$ & 50 & 25 & 100 \\
    $\lambda_{\max}$ & 100 & 100  & 100\\
    $\tau$ & 0.01 & 0.01 & 0.001 \\
    $\alpha$ & 0.95 & 0.95 & 0.95 \\
    Batch Size & 128 & 128 & 256 \\
    Learning Rate($f$) & 3e-3 & 3e-3 & 0.1\\
    Learning Rate($g$) & 3e-4 & 3e-4 & 0.01 \\
    Optimizer & SGD & SGD & SGD \\
    Scheduler & Cosine & Cosine & Cosine \\
    Total SGD Steps & 2k & 2k & 2.5k \\
    Resolution & 32 X 32 & 32 X 32 & 224 X 224 \\
    Arch. & ResNet-32 & ResNet-32 & ResNet-50 \\
    \hline
\end{tabular}
\end{adjustbox}

\label{tab:hyperparams-sup}
\end{table}

\subsection{Experimental Details for Supervised Learning}
 We show our results on 3 datasets: CIFAR-10 LT ($\rho=100$), CIFAR-100 LT ($\rho=100$)~\cite{krizhevsky2009learning} and Imagenet-1k~\cite{russakovsky2015imagenet} LT. For our pre-trained model, we use the model trained by stage-1 of MiSLAS\cite{zhong2021mislas}, which uses a mixup-based pre-training as it improves calibration. For CIFAR-10,100 LT ($\rho=100$) we use ResNet-32 while for Imagenet-1k LT, we use ResNet-50. The detailed list of hyperparameters have been provided in Tab. \ref{tab:hyperparams-sup}. Unlike semi-supervised fine-tuning, we do not require to refresh the pseudo-labels for the unlabelled samples since we already have the true labels. The backbone is trained at a learning rate $\frac{1}{10}^{th}$ of the linear classifier learning rate. The batch norm is frozen across all the layers. The detailed algorithm can be found in Alg. \ref{alg:ours-sup} and is very similar to its semi-supervised variant. We report the performance obtained at the end of fine-tuning.

\begin{algorithm}[H]
    \caption{Training through SelMix .}
    \label{alg:ours-sup}
    \begin{algorithmic}
        \STATE {\bfseries Input:} Data $(D, D^{\mathrm{val}})$, iterations $T$, classifier $h^{(0)}$, metric function $\psi$
        \FOR{$t=1$ {\bfseries to} $T$}
        \STATE $h^{(t)}$ = $h^{(t-1)}$, $C^{(t)} = \mathbb{E}_{(x,y) \sim D^{\mathrm{val}}}[C[h^{(t)}]] \quad$
        \STATE $V_{ij}^{(t)} = - \eta \frac{\partial \mathcal{L}^{\text{mix}}_{ij}}{\partial W} \quad \forall \ i,j$  \qquad \eqref{eq:exact-gains} 
        \STATE $G^{(t)}_{ij} =  \sum_{k,l}\frac{\partial \psi(C^{(t)})}{\partial \tilde{C}_{kl}}(V_{ij}^{(t)})^\top_l \cdot z_k \quad \forall \ i,j $
        \STATE $\mP^{(t)}_{\text{SelMix}} = \text{softmax}(s\textbf{G}^{(t)})$
        \begin{scriptsize}\emph{// Compute Sampling distribution and update pseudo-label }\end{scriptsize}
        \FOR{$n$ SGD steps}
        \STATE $Y_1, Y_2 \sim \mP^{(t)}_{\text{SelMix}}$
        \STATE  $X_1 \sim \mathcal{U}(D_{Y_1}) \ , \ X_2 \sim \mathcal{U}(D_{Y_2})$ \  \begin{scriptsize}\emph{ // sample batches of data}\end{scriptsize}
        \STATE $h^{(t)} := \text{SGD-Update}(h^{(t)},\mathcal{L}_{\text{mixup}},(X_1, Y_1, X_2))$
        \ENDFOR
        \ENDFOR
        \STATE {\bfseries Output:} $h^{(T)}$
    \end{algorithmic}
\end{algorithm}

\begin{table*}[ht]
\caption{The Expression of Non-Decomposable Objectives we consider in our work.}
\label{tab:metric_full}
\begin{adjustbox}{width=\columnwidth,center}
\begin{tabular}{cc}\toprule
    \textbf{Metric} & \textbf{Definition}
    \\\midrule \\
    Mean Recall $(\psi^{AM})$ &  $\frac{1}{K}\sum_{i \in [K]}\frac{C_{ii}[h]}{\sum_{j\in[K]}{C_{ij}[h]}}$ \Tstrut \Bstrut \\ \\ 
      
      G-mean $(\psi^{GM})$ &   $\left ( \prod_{i \in [K]} \frac{C_{ii}[h]}{\sum_{j\in [K]} C_{ij}[h]} \right )^{\frac{1}{k}}$ \Tstrut \Bstrut \\ \\

       H-mean $(\psi^{HM})$&  $K\left( \sum_{i \in [K]} \frac{\sum_{j\in [K]} C_{ij}[h]}{C_{ii}[h]}\right)^{-1}$ \Tstrut \Bstrut \\ \\

      Min. Recall $(\psi^{MR})$& $  \min_{\bm{\lambda} \in\Delta_{K-1}} \sum_{i \in [K]}\lambda_i \frac{C_{ii}[h]}{\sum_{j\in[K]}{C_{ij}[h]}}  $  \Tstrut \Bstrut \\ \\

\multirow{2}{*}{Min of Head and Tail class recall $(\psi^{MR}_{\text{HT}})$} &  $\min_{(\lambda_{\mathcal{H}}, \lambda_{\mathcal{T}}) \in \Delta_{1}} \frac{\lambda_{\mathcal{H}}}{|\mathcal{H}|} \sum_{i \in \mathcal{H}}\frac{C_{ii}[h]}{\sum_{j\in[K]}{C_{ij}[h]}} $  \\ &  $+ \frac{\lambda_{\mathcal{T}}}{|\mathcal{T}|} \sum_{i \in \mathcal{T}}\frac{C_{ii}[h]}{\sum_{j\in[K]}{C_{ij}[h]}}$ \Tstrut \Bstrut \\ \\

     Mean Recall s.t.  per class coverage $\geq \frac{\alpha}{K} $ $(\psi^{AM}_{\text{cons.}})$& 
    $\min_{\bm{\lambda} \in \mathbb{R}^K_+} \frac{1}{K}\sum_{i \in [K]}\frac{C_{ii}[h]}{\sum_{j\in[K]}{C_{ij}[h]}}+ \sum_{j \in [K]}\lambda_j \left(\sum_{i \in [K]} C_{ij}[h] - \frac{\alpha}{K} \right)$ \Tstrut \Bstrut \\ \\

   \multirow{2}{*}{Mean Recall s.t.  minimum of head  }& 
$\min_{(\lambda_{\mathcal{H}}, \lambda_{\mathcal{T}}) \in \RR^2_{\ge 0}} 
  \frac{1}{K}\sum_{i \in [K]}\frac{C_{ii}[h]}{\sum_{j\in[K]}{C_{ij}[h]}} + \lambda_{\mathcal{H}}\left(
      \sum_{i \in [K], j \in \mathcal{H}} \frac{C_{ij}[h]}{|\mathcal{H}|}  - \frac{0.95}{K}
   \right)  $ \\ and tail class coverage $\geq \frac{\alpha}{K}$ $(\psi^{AM}_{\text{cons.(HT)}})$& $ + \lambda_{\mathcal{T}}  \left(
      \sum_{i \in [K], j \in \mathcal{T}} \frac{C_{ij}[h]}{|\mathcal{T}|} - \frac{0.95}{K}
   \right) $ \Tstrut \Bstrut \\ \\
   
    H-mean s.t. per class coverage $\geq \frac{\alpha}{K} $ $(\psi^{HM}_{\text{cons.}})$& 
    $\min_{\bm{\lambda} \in \mathbb{R}^K_+} K\left( \sum_{i \in [K]} \frac{\sum_{j\in [K]} C_{ij}[h]}{C_{ii}[h]}\right)^{-1}+\sum_{j \in [K]}\lambda_j \left(\sum_{i \in [K]} C_{ij}[h] - \frac{\alpha}{K} \right)$ \Tstrut \Bstrut \\ \\

   \multirow{2}{*}{H-mean s.t. minimum of head} & $\min_{(\lambda_{\mathcal{H}}, \lambda_{\mathcal{T}}) \in \RR^2_{\ge 0}} K\left( \sum_{i \in [K]} \frac{\sum_{j\in [K]} C_{ij}[h]}{C_{ii}[h]}\right)^{-1} + \lambda_{\mathcal{H}}\left(
      \sum_{i \in [K], j \in \mathcal{H}} \frac{C_{ij}[h]}{|\mathcal{H}|}  - \frac{0.95}{K}
   \right)$ \\ and tail class coverage $\geq \frac{\alpha}{K}$ $(\psi^{HM}_{\text{cons.(HT)}})$& $ + \lambda_{\mathcal{T}}  \left(
      \sum_{i \in [K], j \in \mathcal{T}} \frac{C_{ij}[h]}{|\mathcal{T}|} - \frac{0.95}{K}
   \right)$ \\\bottomrule
\end{tabular}

\end{adjustbox}

\end{table*}

\section{Results for Case With Unknown Labeled Distribution}
In this section, we provide a full-scale comparison of all the methods of the case when the labeled distribution does not match the unlabeled data distribution, simulating the scenario when label distribution is unknown. The  chapter presents the summary plots for these results in Fig.~\ref{fig:summary_mismatched_radar}. The Table~\ref{tab:cf10-mismatched-results}, we present results for the case when for CIFAR-10 once the unlabeled data follow an inverse label distribution i.e. ($\rho_l = 100, \rho_{u} = \frac{1}{100}$) and the case when the unlabeled data is distributed uniformly across all classes ($\rho_l = 100, \rho_u = 1$). In both cases, we find that SelMix can produce significant improvement across metrics. Further, we also compare our method in the practical setup where unlabeled data distribution is unknown. This situation is perfectly emulated by the STL-10 dataset, which also contains an unlabeled set of 100k images. Table~\ref{tab:stl10-mismatched} presents results for different approaches on the STL-10 case. We observe that SelMix produces superior results compared to baselines and is robust to the mismatch in distribution between labeled and unlabeled data.

\label{app:mismatched_comparison}
\begin{table*}[ht!]

\caption{Comparison on metric objectives for CIFAR-10 LT under $\rho_l \neq \rho_u$ assumption. Our experiments involve $\rho_u = 100, \rho_l = 1
\; (\text{uniform}) \; \text{and} \; \rho_u = 100, \rho_l = \frac{1}{100} (\text{inverted})$. SelMix achieves significant gains over other SSL-LT methods across all the metrics.}
\label{tab:cf10-mismatched-results}
\begin{adjustbox}{width=\columnwidth,center}
\begin{tabular}{lccccc|ccccc}\toprule

& \multicolumn{5}{c}{\textbf{CIFAR-10} \quad $(\rho_l=100, \rho_u=\frac{1}{100}, N_1=1500, M_1=30)$} & \multicolumn{5}{c}{\textbf{CIFAR-10} \quad $(\rho_l=100, \rho_u=1, N_1=1500, M_1=3000)$}
\\ \cmidrule(lr){2-6}\cmidrule(lr){7-11}
           & Mean Rec.  & Min Rec. & HM & GM  & Mean Rec./Min Cov. & Mean Rec.  & Min Rec. & HM & GM  & Mean Rec./Min Cov.\\\midrule
FixMatch    & \s{71.3}{1.1} & \s{28.5}{2.6} & \s{61.3}{2.7} & \s{67.1}{1.7} & \s{71.3}{1.1} / \s{0.030}{2e-3}  & \s{82.8}{1.3} & \s{59.1}{5.8} & \s{80.6}{2.1} & \s{82.3}{1.5} & \s{82.8}{1.3} / \s{0.059}{6e-3} \\
 \addedtext{DARP} & \s{79.7}{0.8} & \s{60.7}{2.4} & \s{78.1}{0.9} & \s{78.9}{0.9} & \s{79.7}{0.8} / \s{0.065}{2e-3}  & \s{84.8}{0.7} & \s{66.9}{3.1} & \s{83.5}{0.8} & \s{85.2}{0.7} & \s{84.8}{0.7} / \s{0.067}{3e-3} \\
 \addedtext{CReST} & \s{71.3}{0.9} & \s{40.3}{3.0} & \s{65.8}{1.5} & \s{68.6}{1.2} & \s{71.3}{0.9} / \s{0.040}{5e-3}  & \s{85.7}{0.3} & \s{68.7}{1.7} & \s{84.6}{0.14} & \s{85.1}{0.1} & \s{85.7}{0.3} / \s{0.075}{7e-4} \\
\addedtext{CReST+}   & \s{72.8}{0.8} & \s{45.2}{2.5} & \s{68.4}{1.3} & \s{70.6}{1.1} & \s{72.8}{0.8} / \s{0.047}{3e-3}  & \s{86.4}{0.2} & \s{71.7}{1.9} & \s{85.6}{0.2} & \s{86.1}{0.1} & \s{86.4}{0.2} / \s{0.078}{1e-3} \\
\addedtext{DASO} & \s{79.2}{0.2} & \addedtext{\s{64.6}{1.9}} & \s{78.1}{0.1} & \s{78.6}{0.8} & \s{79.2}{0.2} / \s{0.072}{3e-3}  & \s{88.6}{0.4} & \s{78.2}{1.6} & \s{88.4}{0.5} & \s{88.5}{0.4} & \s{88.6}{0.4} / \s{0.089}{1e-3} \\
\addedtext{ABC} & \s{80.8}{0.4} & \s{65.1}{0.8} & \s{79.6}{0.3} & \s{80.7}{0.6} & \s{80.8}{0.4} / \s{0.073}{5e-3}  & \s{88.6}{0.4} & \s{74.8}{2.9} & \s{88.2}{0.7} & \s{88.6}{0.3} & \s{88.6}{0.4} / \s{0.086}{4e-3} \\
\addedtext{CoSSL} & \s{78.6}{1.0} & \s{66.3}{2.9} & \s{77.2}{1.2} & \s{77.8}{1.1} & \s{78.6}{1.0} / \s{0.070}{1e-3}  & \s{88.7}{0.9} & \s{76.1}{2.9} & \s{88.2}{1.1} & \s{88.5}{1.0} & \s{88.7}{0.9} / \s{0.084}{8e-3} \\
\addedtext{CSST} & \s{77.5}{1.5} & \s{72.1}{0.2} & \s{76.5}{4.9} & \s{76.8}{5.2}  & \s{77.5}{1.5} / \s{0.091}{3e-3} & \s{87.6}{0.7} & \s{78.1}{0.3} & \s{86.1}{0.7} & \s{87.1}{0.2} & \s{87.6}{0.7} / \s{0.091}{1e-3}   \\
FixMatch(LA) & \s{75.5}{1.5} & \s{45.1}{4.4} & \s{71.1}{2.5} & \s{73.3}{1.9} & \s{75.5}{1.5} / \s{0.046}{4e-3} & \s{90.1}{0.4} & \s{75.8}{2.1} & \s{89.5}{0.7} & \s{89.7}{0.5} & \s{90.1}{0.4} / \s{0.083}{1e-3} \\
\qquad \textbf{w/SelMix (Ours)} & \s{\textbf{81.3}}{0.5} & \s{\textbf{74.3}}{1.2} & \s{\textbf{81.0}}{0.8} & \s{\textbf{80.9}}{0.5}  & \s{\textbf{81.7}}{0.8} / \s{\textbf{0.091}}{3e-3} & \s{\textbf{91.4}}{1.2} & \s{\textbf{84.7}}{0.7} & \s{\textbf{91.3}}{1.1} & \s{\textbf{91.3}}{1.2}  & \s{\textbf{91.4}}{1.2} / \s{\textbf{0.096}}{1e-3}\\\bottomrule
\end{tabular}
\end{adjustbox}
\end{table*}

\begin{table}[ht!]
\caption{Comparison across methods when label distribution $\rho_u$ is
unknown. We use the STL-10 dataset for comparison in such a case.}
\label{tab:stl10-mismatched}
\begin{adjustbox}{width=\columnwidth,center}
\begin{tabular}{lcccccc}\toprule
& \multicolumn{6}{c}{\textbf{STL-10} \quad $(\rho_l=10, \rho_u=\text{NA}, N_1=450, \sum_i M_i = 100\text{k})$}
\\\cmidrule(lr){2-7}
           & Mean Rec.  & Min Rec. & HM & GM & HM/Min Cov. & Mean Rec./Min Cov.\\\midrule
FixMatch    & \s{72.7}{0.7} & \s{43.2}{7.1} & \s{67.7}{1.5} & \s{71.6}{1.3} & \s{67.7}{1.5} / \s{0.048}{1e-2} & \s{72.7}{0.7} / \s{0.048}{1e-2}\\
 \addedtext{DARP} & \s{76.5}{0.3} & \s{54.7}{1.9} & \s{74.0}{0.5} & \s{75.3}{0.4} & \s{74.0}{0.5} / \s{0.058}{2e-3} & \s{76.5}{0.3} / \s{0.058}{2e-3} \\
 \addedtext{CReST} & \s{70.1}{0.3} & \s{48.2}{2.2} & \s{67.1}{1.1} & \s{67.8}{1.1} & \s{67.1}{1.1} / \s{0.066}{2e-3} & \s{70.1}{0.3} / \s{0.066}{2e-3}\\
\addedtext{DASO} & \s{78.1}{0.5} & \s{55.8}{3.7} & \s{76.6}{1.1} & \s{77.2}{0.2} & \s{76.6}{1.1} / \s{0.083}{3e-3} & \s{78.1}{0.5} / \s{0.083}{3e-3}\\
\addedtext{ABC} & \s{77.5}{0.4} & \s{55.4}{6.7} & \s{74.7}{1.5} & \s{76.3}{0.9} & \s{74.7}{1.5} / \s{0.079}{7e-3} & \s{77.5}{0.4} / \s{0.079}{7e-3}\\
\addedtext{CSST} & \s{79.2}{1.5} & \s{50.8}{2.9} & \s{78.3}{2.6} & \s{78.9}{2.1} & \s{78.3}{2.6} / \s{0.081}{6e-3} & \s{79.2}{1.5} / \s{0.081}{6e-3}\\
FixMatch(LA) & \s{78.9}{0.4} & \s{56.4}{1.9} & \s{76.5}{1.1} & \s{77.8}{0.8} & \s{76.5}{1.1} / \s{0.066}{5e-3} & \s{78.9}{0.4} /  \s{0.066}{5e-3} \\
\qquad \textbf{w/SelMix (Ours)} & \s{\textbf{80.9}}{0.5} & \s{\textbf{68.5}}{1.8} & \s{\textbf{79.1}}{1.2} & \s{\textbf{80.1}}{0.4} & \s{\textbf{79.1}}{1.2} /  \s{\textbf{0.088}}{1e-3} & \s{\textbf{80.9}}{0.1} / \s{\textbf{0.088}}{1e-3}\\\bottomrule
\end{tabular}
\end{adjustbox}
\end{table}

\section{Optimization of H-mean with Coverage Constraints}
\label{app:hmean-coverage}
We consider the objective of optimizing H-mean subject to the constraint that all classes must have a coverage $\geq \frac{\alpha}{K}$. For CIFAR-10, when the unlabeled data distribution matches the labeled data distribution, uniform or inverted, SelMix is able to satisfy the coverage constraints. A similar observation could be made for CIFAR-100, where the constraint is to have the minimum head and tail class coverage above $\frac{0.95}{K}$. For STL-10, SelMix fails to satisfy the constraint because the validation dataset is minimal (500 samples compared to 5000 in CIFAR). We want to convey here that as CSST is only able to optimize for linear metrics like min. recall its performance is inferior on complex objectives like optimizing H-mean with constraints. This shows the superiority of the proposed SelMix framework.

\begin{table*}[ht!]
\caption{Comparison of methods for optimization of H-mean with coverage constraints.}
\begin{adjustbox}{width=\columnwidth,center}
\begin{tabular}{lcc|cc|cc|cc|cc}\toprule

& \multicolumn{2}{c}{\textbf{CIFAR-10}} & \multicolumn{2}{c}{\textbf{CIFAR-10}} & \multicolumn{2}{c}{\textbf{CIFAR-10}} & \multicolumn{2}{c}{\textbf{CIFAR-100}} & \multicolumn{2}{c}{\textbf{STL-10}} \\
& \multicolumn{2}{c}{$\rho_l=100, \rho_u=\frac{1}{100}$} & \multicolumn{2}{c}{$\rho_l=\rho_u=100$} & \multicolumn{2}{c}{$\rho_l=100, \rho_u=1$} & \multicolumn{2}{c}{$\rho_l=\rho_u=10$} & \multicolumn{2}{c}{$\rho_l=10, \rho_u=\text{NA}$}\\
& \multicolumn{2}{c}{$N_1=1500, M_1=30$} & \multicolumn{2}{c}{$N_1=1500, M_1=3000$} & \multicolumn{2}{c}{$N_1=1500, M_1=3000$} & \multicolumn{2}{c}{$N_1=150, M_1=300$} & \multicolumn{2}{c}{$N_1=450, \sum_i M_i = 100\text{k}$} \\
\cmidrule(lr){2-3}\cmidrule(lr){4-5}\cmidrule(lr){6-7}\cmidrule(lr){8-9}\cmidrule(lr){10-11}
           & HM  & Min Cov. & HM  & Min Cov. & HM  & Min Cov. & HM & Min H-T Cov.  & HM & Min Cov.\\\midrule
DARP & \s{78.1}{0.9} & \s{0.065}{3e-3} & \s{81.9}{0.5} & \s{0.070}{3e-3} & \s{83.5}{0.8} & \s{0.067}{3e-3} & \s{48.7}{1.3} & \s{0.0040}{2e-3} & \s{74.0}{0.5} & \s{0.058}{2e-3}\\
CReST & \s{65.8}{1.5} & \s{0.040}{5e-3} & \s{81.0}{0.7} & \s{0.073}{5e-3} & \s{84.6}{0.2} & \s{0.075}{7e-4} & \s{48.3}{0.2} & \s{0.0083}{2e-4} & \s{67.1}{1.1} & \s{0.066}{2e-3}\\
DASO & \s{78.1}{0.1} & \s{0.072}{3e-3} & \s{83.5}{0.3} & \s{0.083}{1e-3} & \s{88.4}{0.5} & \s{0.089}{1e-3} & \s{49.1}{0.7} & \s{0.0063}{3e-4} & \s{76.6}{1.1} & \s{\underline{0.083}}{3e-3}\\
ABC & \s{\underline{79.6}}{0.3} & \s{0.073}{5e-3}  &\s{\underline{84.6}}{0.5} & \s{0.086}{3e-3} & \s{88.2}{0.7} & \s{0.086}{1e-3} & \s{\underline{50.1}}{1.2} & \s{0.0089}{2e-4} & \s{74.7}{1.5} & \s{0.079}{7e-3}\\\midrule
CSST & \s{76.5}{4.9} & \s{\underline{0.081}}{6e-3} & \s{76.9}{0.2} & \s{\underline{0.093}}{3e-4} & \s{86.7}{0.7} & \s{\underline{0.092}}{1e-3} & \s{47.7}{0.8} & \s{\underline{0.0098}}{2e-4} & \s{\underline{78.3}}{2.6} & \s{0.081}{6e-3}\\
FixMatch (LA)    & \s{78.3}{0.8} &  \s{0.064}{1e-3} & \s{76.7}{0.1} & \s{0.056}{3e-3} & \s{\underline{89.3}}{0.2} & \s{0.086}{1e-3} & \s{45.5}{2.1} & \s{0.0053}{1e-4} & \s{74.6}{1.7} & \s{0.066}{5e-3}\\
\quad \textbf{w/SelMix (Ours)} & \s{\textbf{81.0}}{0.8} & \s{\textbf{0.095}}{1e-3} &\s{\textbf{85.1}}{0.1} & \s{\textbf{0.095}}{1e-3} & \s{\textbf{91.3}}{0.7} & \s{\textbf{0.096}}{1e-3} & \s{\textbf{53.8}}{0.5} & \s{\textbf{0.0098}}{1e-4} & \s{\textbf{79.1}}{1.2} & \s{\textbf{0.088}}{1e-3}\\\bottomrule
\end{tabular}
\end{adjustbox}

\end{table*}

\section{Evolution of Gain Matrix with Training }
\label{app:gain-matrix-train}

From the above collection of gain matrices, which are taken from different time steps of the training phase, we observe that  ($|\text{max}(\bG^{(t)})|)$ of the gain matrix decreases with increase in SGD steps $t$, and settles on a negligible value by the time training is finished. This could be attributed to the fact that as the training progresses, the marginal improvement of the gain matrix decreases. 

Another phenomenon we observe is that initially, during training, only a few mixups (particularly tail class ones) have a disproportionate amount of gain associated with them. A downstream consequence of this is that the sampling function $\mathcal{P}_{\text{SelMix}}$ prefers only a few $(i, j)$ mixups. Whereas, as the training continues, it becomes more exploratory rather than greedily exploiting the mixups that give the maximum gain at a particular timestep. 
\begin{figure}[!t]
    \centering
    \subfloat[\centering Initial Stage ($t=0$ SGD steps)]{{\includegraphics[width=0.3\textwidth]{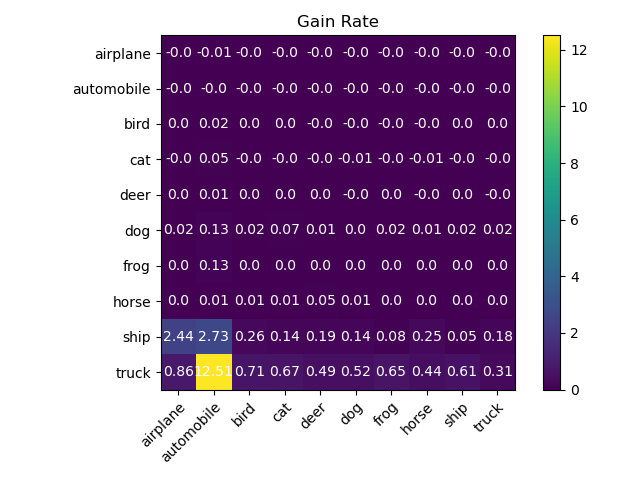} }}%
    \quad
    \subfloat[\centering Intermediate Stage ($t=5k$ SGD steps)]{{\includegraphics[width=0.3\textwidth]{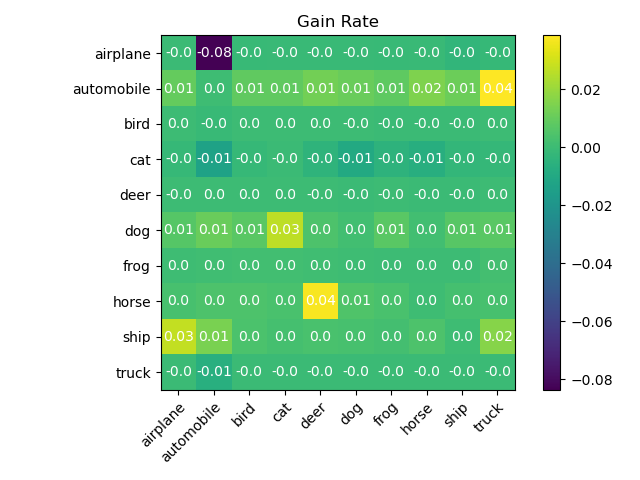} }}%
    \quad
    \subfloat[\centering Final Stage ($t=10k$ SGD steps)]{{\includegraphics[width=0.3\textwidth]{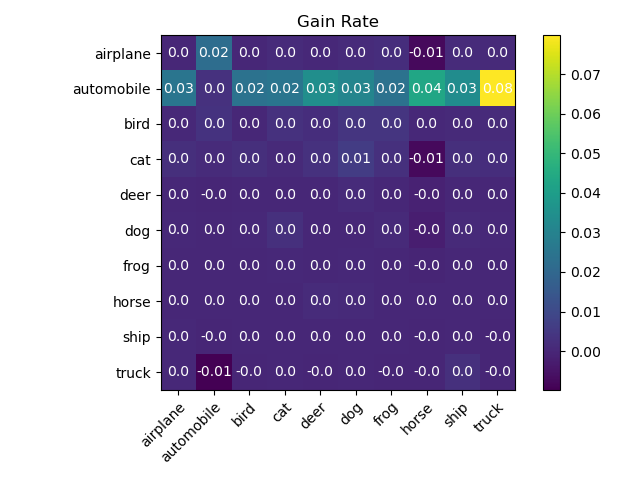} }}%
    \caption{Evolution of gain matrix for mean recall optimized run for CIFAR-10 LT ($\rho_l = \rho_u$).}%
    \label{fig:mean_gain}%
\end{figure}

\begin{figure}[!t]
    \centering
    \subfloat[\centering Initial Stage ($t=0$ SGD steps)]{{\includegraphics[width=0.3\textwidth]{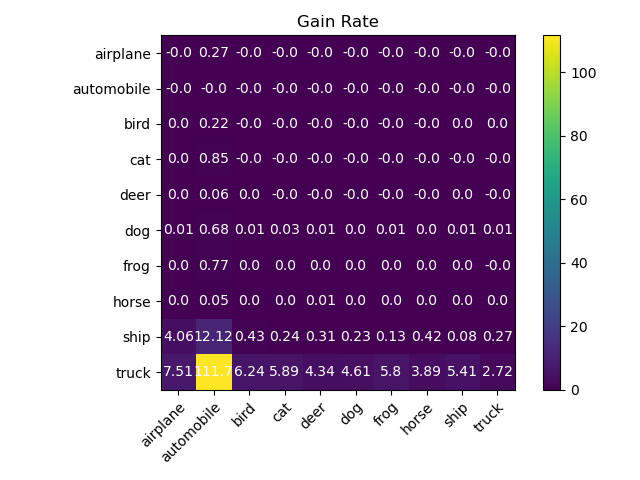} }}%
    \quad
    \subfloat[\centering Intermediate Stage ($t=3k$ SGD steps)]{{\includegraphics[width=0.3\textwidth]{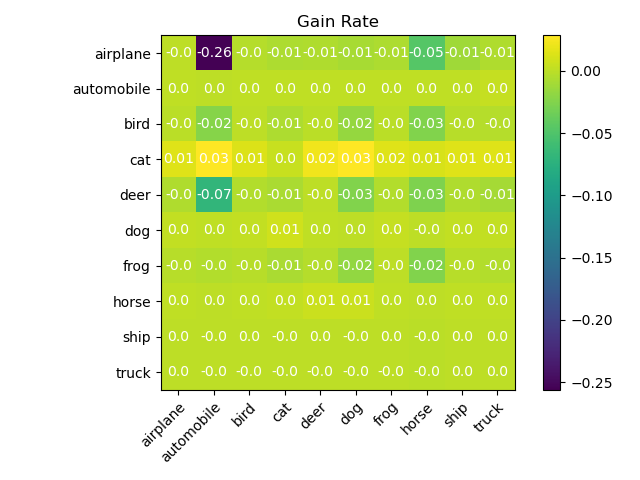} }}%
    \quad
    \subfloat[\centering Final Stage ($t=10k$ SGD steps)]{{\includegraphics[width=0.3\textwidth]{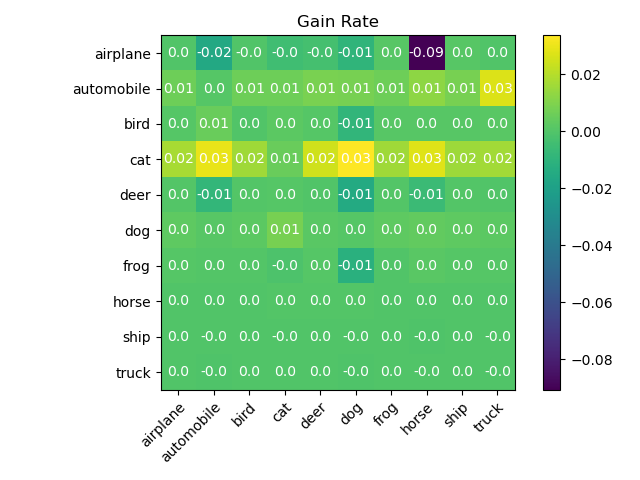} }}%
    \caption{Evolution of gain matrix for min. recall optimized run for CIFAR-10 LT ($\rho_u = \rho_l$).}%
    \label{fig:min_gain}%
\end{figure}

\section{Analysis}
\label{app:analysis}

We use a subset of Min. Recall, Mean Recall and Mean Recall with constraints for analysis.

\begin{table}[!t]
    \begin{minipage}{0.5\textwidth}
        \caption{Results for sampling policies for $\mP_{\text{Mix}}$ (CIFAR-10 LT, semi-supervised) $\rho=100$. }
    
    \centering
    \begin{adjustbox}{max width=0.9\columnwidth}
    \begin{tabular}{lcccc}
    \hline
    \multirow{2}{*}{Method} & Mean & Min &  Min  &  Mean  \Tstrut\\ 
    & \textit{ Recall } & \textit{Coverage} & \textit{Recall } &  \textit{ Recall }\\
    \hline 
        Uniform Policy  & 83.3 & 0.072 & 70.5 &  83.3 \\ 

        Greedy Policy & 83.6 & 0.093 & 78.2 &  81.8 \\
                SelMix Policy & \textbf{84.9} &\textbf{0.094} & \textbf{79.1} & \textbf{84.1} \\ 
    \hline
    \end{tabular}
    \end{adjustbox}
    \label{tab:policy-comparison}
    \end{minipage}
    \begin{minipage}{0.5\textwidth}
         \caption{Comparison of finetuning the feature extractor vs only the linear classification layer (CIFAR-10 LT, semi-supervised) $\rho=100$}
    
    \centering
    \begin{adjustbox}{max width=0.8\columnwidth}
    \begin{tabular}{lcccc}
    \hline
    \multirow{2}{*}{Method} & Mean & Min &  Min  &  Mean  \Tstrut\\ & \textit{ Recall } & \textit{Coverage} & \textit{Recall } &  \textit{ Recall }\\
    \hline 
        Frozen $g$  & 83.5 & 0.089 & 77.3 &  84.1 \\ 
        Finetuning $g$ & \textbf{85.4} &\textbf{0.095} & \textbf{79.1} &  \textbf{84.5} \\
    \hline
    \end{tabular}
    \end{adjustbox}
    \label{tab:finetune-vs-frozen}
   
    \end{minipage}
    
\end{table}

\noindent \textbf{a) Sclability and Computation.} To demonstrate scalability of our method we show results on ImageNet100-LT for semi and ImageNet-LT for fully supervised settings. Similar to other datasets, we find in Table~\ref{tab:large-data} SelMix is able to improve over SotA methods across objectives. Further, SelMix has the same time complexity as CSST~\cite{rangwani2022costsensitive} baseline. (Refer Sec.~\ref{app:computation} \& Sec.~\ref{app:complexity}). 

\begin{figure}[!ht]
    \centering
    \includegraphics[width= 0.8\textwidth]{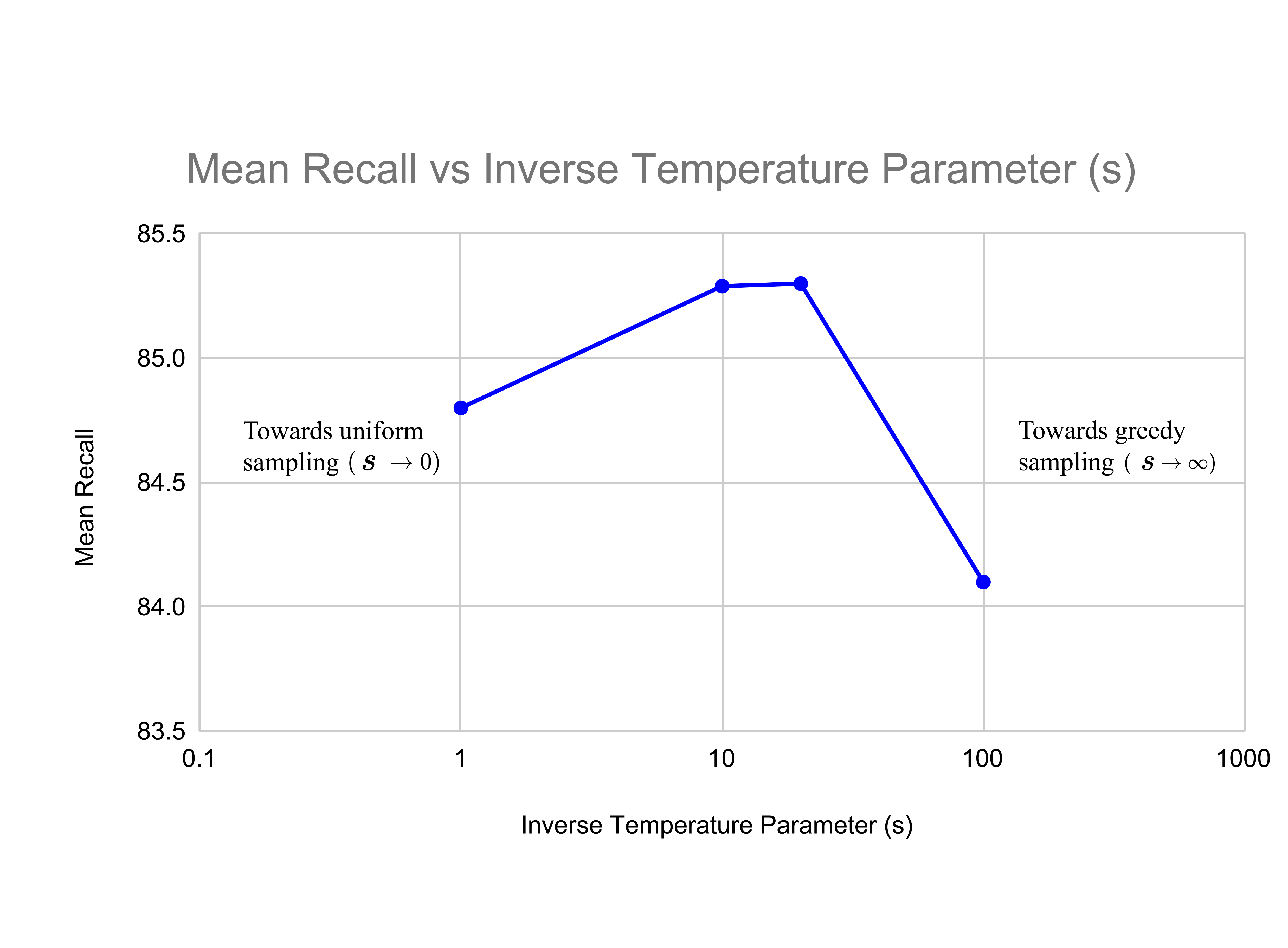}
    \caption{We show ablation on the inverse temperature parameter ($s$) v/s the performance on the mean recall. For a mean recall optimized run a very small $s$ yields a sampling function close to a uniform sampling, whereas a very large $s$ ends up close to a greedy sampling strategy. }
    \label{fig:mixup_comparison}
\end{figure}

\noindent \textbf{b) Comparison of Sampling Policies.}
We compare the sampling policies ($\mP_{\text{Mix}}$): the uniform mixup policy, the greedy policy of selecting the max gain mixup and the SelMix policy (Table \ref{tab:policy-comparison}). We find that other policies in comparison to SelMix are unstable and lead to inferior results. We compare furtherthe performance of a range of sampling distribution by varying the inverse distribution temperature $s$ in $\mathcal{P}_{\mathrm{SelMix}}$. We find that intermediate values of $s = 10$ work better in practice (Fig.~\ref{fig:mixup_comparison}). 

\noindent \textbf{c) Fine-Tuning.} In Table \ref{tab:finetune-vs-frozen} we observe that fine-tuning $g$ leads to improved results in comparison to keeping it frozen. We further show in App. \ref{app:computation} that fine-tuning with SelMix is computationally cheap compared to CSST for optimizing a particular non-decomposable objective.

\subsection{Detailed Performance Analysis of SelMix Models}
\label{app:detailed-analysis}
As in SelMix, we have provided results for fine-tuned models for optimizing specific metrics on CIFAR-10 ($\rho_l = \rho_u$) in Table~\ref{tab:matched-results}. In this section, we analyze all these specific models on all other sets of metrics. We tabulate our results in Table \ref{tab:full_metric_runs}. It can be observed that when the model is trained for the particular metric for the diagonal entries, it performs the best on it. Also, we generally find that all models trained through SelMix reasonably perform on other metrics. This demonstrates that the models produced are balanced and fair in general. As a rule of thumb, we would like the users to utilize models trained for constrained objectives as they perform better than others cumulatively.

\begin{table}[!ht]

\caption{Values of all metric values for individually optimized runs for CIFAR-10 LT ($\rho_l = \rho_u$)}
\label{tab:full_metric_runs}
\begin{adjustbox}{width=0.8\columnwidth,center}
\begin{tabular}{c|cccccc}\toprule
     \diagbox{Optimized On}{Observed Metric}      & Mean Rec.  & Min. Rec. & HM & GM &  Mean Rec./Min Cov. & HM/Min Cov.\\\midrule
Mean Rec.    & 85.4 & 77.6 & 85.0 & 85.1 & 85.4/0.089 & 85.0/0.089\\
Min. Rec. & 84.2 & 79.1 & 84.1 & 84.2 &  84.2/0.091 & 84.1/0.091 \\
HM & 85.3 & 77.7 & 85.1 & 85.2 &  85.3/0.091 & 85.1/0.091 \\
GM & 85.3 & 77.5 & 85.1 & 85.3 & 85.3/0.091 & 85.1/0.091\\
Mean Rec./Min. Cov. & 85.7 & 75.9 & 84.7 & 84.8 &  85.7/0.095 & 84.7/0.095\\
HM/Min Cov. & 85.1 & 76.2 & 84.8 & 84.9 &  85.1/0.095 & 84.8/0.095\\\bottomrule
\end{tabular}
\end{adjustbox}

\end{table}

\subsection{Comparison between FixMatch and FixMatch (LA)}
\label{app:fix-vs-fixla}
We find that using logit-adjusted loss helps in training feature extractors, which perform much superior in comparison to the vanilla FixMatch Algorithm (Table \ref{tab:backbone-scaling}). However, our method SelMix is able to improve both the FixMatch and the FixMatch (LA) variant. We advise users to use the FixMatch (LA) algorithm for better results.

 \begin{table}[!ht]
    
    \caption{Comparison of the FixMatch and FixMatch (LA) methods with SelMix.}
    
    \centering
    \begin{adjustbox}{max width=0.8\columnwidth}
    \begin{tabular}{lcccc}
    \hline
    \multirow{2}{*}{Method} & Mean & Min &  Min  &  Mean  \Tstrut\\ & \textit{ Recall } & \textit{Coverage} & \textit{Recall } &  \textit{ Recall }\\
    \hline 
        FixMatch & 76.8 & 0.037 & 36.7 &  76.8 \\
        w/ SelMix  & 84.7 & 0.094 & 78.8 &  82.7 \\
        FixMatch (LA) & 82.6 & 0.065 &  63.6 &  82.6 \\ 
        w/ SelMix  & 85.4 & 0.095 & 79.1 &  84.1 \\
    \hline
    \end{tabular}
    \end{adjustbox}
    \label{tab:backbone-scaling}
    
 \end{table}

\subsection{Variants of Mixup}
\label{app:variants-mixup}
As SelMix is a distribution on which class samples $({i,j})$ to be mixed up, it can be easily be combined with different variants of mixup~\cite{yun2019cutmix, kim2020puzzle}. To demonstrate this, we replace the feature mixup that we perform in SelMix, with CutMix and PuzzleMix. Table~\ref{tab:mixup_variants} contains results for various combinations for optimizing the Mean Recall and Min Recall across cases. We observe that SelMix can optimize the desired metric, even with CutMix and PuzzleMix. However, the feature mixup we performed originally in SelMix works best in comparison to other variants. This establishes the complementarity of SelMix with the different variants of Mixup like CutMix, PuzzleMix, etc., which re-design the procedure of mixing up images.

 \begin{table}[!ht]
  
    \caption{Comparison of SelMix when applied to various Mixup variants.}
    \label{tab:mixup_variants}
    \centering
    \begin{adjustbox}{max width=0.8\columnwidth}
    \begin{tabular}{lcc}
    \hline
    \multirow{2}{*}{Method} & Mean & Min  \Tstrut\\ & \textit{ Recall } &  \textit{Recall } \\
    \hline 
        FixMatch & 79.7$_{\pm 0.6}$ & 55.9$_{\pm 1.9}$   \\
        w/ SelMix (CutMix)  & 84.8$_{\pm 0.2}$ & 75.3$_{\pm 0.1}$  \\
        w/ SelMix (PuzzleMix) & 85.1$_{\pm 0.3}$ & 75.2$_{\pm 0.1}$  \\ 
        w/ SelMix (Features-Ours)  & 85.4$_{\pm 0.1}$ & 79.1$_{\pm 0.1}$  \\
    \hline
    \end{tabular}
    \end{adjustbox}
    
 \end{table}

\section{Computational Requirements}
\label{app:comp-req}
\begin{table*}[ht!]
\centering
\caption{Comparison of time taken across datasets, for the calculation of Gain using SelMix (Alg.~\ref{alg:ours}) was done on GPU (NVIDIA RTX A5000).}
\label{tab:time-required}
\begin{tabular}{l*{3}{c}}\toprule
{Dataset} & CIFAR-10 LT ($\rho=100$) & (CIFAR-100 LT $\rho=100$) & Imagenet-1k LT  \\ \midrule
 & 0.02 sec. & 1.3 sec. & 124 sec. \\ \bottomrule
\end{tabular}
\end{table*}
\label{app:computation}
The experiments were done on Nvidia A5000 GPU (24 GB). While the fine-tuning was done on a single A5000, the pre-training was done using PyTorch data parallel on 4XA5000. The pre-training was done until no significant change in metrics was observed and the fine-tuning was done for 10k steps of SGD with a validation step every 50 steps. A major advantage of SelMix over  CSST is that the process of training a model optimized for a specific objective requires end to end training which is computationally expensive($\sim$10 hrs on CIFAR datasets). Our finetuning method takes a fraction ($\sim$1hr on CIFAR datasets) of  what it requires in computing time compared to CSST. An analysis for computing the Gain through Alg.~\ref{alg:ours}, is provided in Table \ref{tab:time-required}. We observe that even for the ImageNet-1k dataset, the gain calculation doesn't require large amount of GPU time. Further, an efficient parallel implementation across classes can further reduce time significantly.

\section{Limitation of Our Work}
\label{app:limitations}
In our current work, we mostly focus on our algorithm's correctness and empirical validity of SelMix across datasets.
Another direction that could be further pursued is improving the algorithm's performance by efficiently parallelizing the operations across GPU cores, as the operations for each class are independent of each other. The other direction for future work could be characterization of the classifier obtained through SelMix, using a generalization bound.
Existing work \cite{zhang2021does} on the mixup method for accuracy optimization 
showed that 
learning with the vicinal risk minimization using mixup leads to
a better generalization error bound
than the empirical risk minimization.
It would be interesting future work to show a similar result for SelMix.

\addedtext{

\section{Additional Related Works}
\label{app:rel_work_extra}
\vspace{1mm} \noindent \textbf{Selective Mixup for Robustness.} The paper SelecMix~\cite{hwang2022selecmix} creates samples for robust learning in the presence of bias by mixing up samples with conflicting features, with samples with bias-aligned features. The paper demonstrates that learning on these samples leads to improved out-of-distribution accuracy, even in the presence of label noise. This paper selects the samples to mixup based on the similarity of the labels, to improve mixup performance on regression tasks. Existing works \cite{hwang2022selecmix, yao2022improving,palakkadavath2022improving} show that mixups help train classifiers with better domain generalization. \citet{teney2023selective} show that resampling-based techniques often come close in performance to mixup-based methods when utilizing the implicit sampling technique used in these methods. It has been shown that mixup improves the feature extractor and can also be used to train more robust classifiers \cite{hwang2022selecmix}. \citet{palakkadavath2022improving} show that it is possible to generalize better to unknown domains by making the model's feature extractor invariant to both variation in the domain and any interpolation of a sample with similar semantics but different domains.

\vspace{1mm} \noindent \textbf{Black-Box Optimization.} We further note that there are some recent works~\cite{li2023earning, wierstra2014natural} which aim to fine-tune a model based on local target data for an specific objective, however these methods operate in a black-box setup whereas SelMix works in a white-box setup with full access to model and its gradients.

}

\renewcommand{\thesection}{I}

\supersection{S$^3$VAADA: Submodular Subset Selection for Virtual Adversarial Active Domain Adaptation (Chapter-9)}

\section{t-SNE Analysis for AADA}
\label{s3vaada_aada}
\begin{figure}[h]
    \centering
    \includegraphics[width=0.49\textwidth]{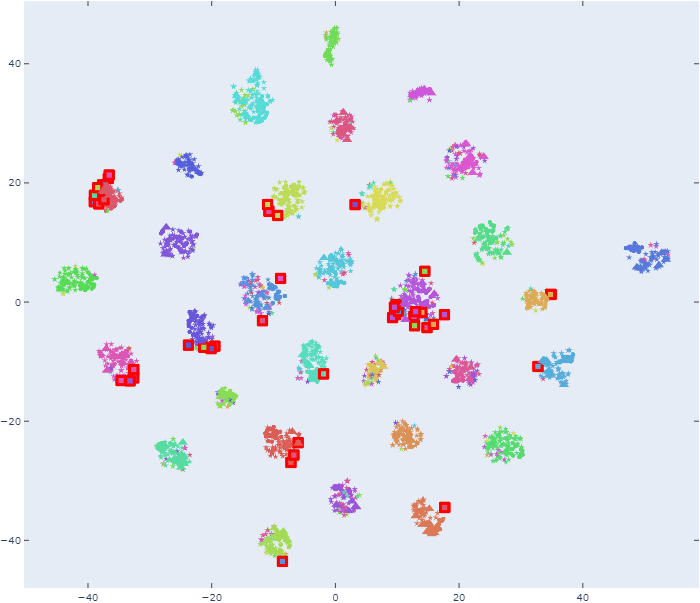}
    \caption{t-SNE analysis of AADA sampling. The selected samples are represented by the red boxes. We see clusters of samples being selected which depict \textit{redundancy} in selection. }
    \label{s3vaada_fig:tsne-aada}
\end{figure}

We give experimental evidence of the redundancy issue present in the AADA sampling. We perform the training with VAADA training method with the implementation details present in Sec. \ref{s3vaada_experimental_details} on Webcam $\rightarrow$ Amazon. Fig. \ref{s3vaada_fig:tsne-aada} shows the selected samples in the intermediate cycle, which clearly depicts clusters of the samples selected. The existence of clusters confirms the presence of \textit{redundancy} in selection.

\section{Proofs}
\label{s3vaada_proof}
\subsection{Lemma 1}

We present proof for lemma 1 which is stated in Section {\color{red} 4.1.4} of the chapter.

\noindent \textbf{Lemma 1} The set function $f(S)$ defined by equation below is submodular.  \\
\begin{equation}
        f(S \cup \{x_i\}) - f(S) = \alpha VAP(x_i) + \beta d(S,x_i)  +
     (1 - \alpha - \beta) R(S, x_i)
\end{equation}

We first prove that all the three individual components in the above expression are submodular and then prove that the convex combination of the three terms is submodular.

\noindent \textbf{Submodularity of the VAP Score $VAP(x_i)$:}
The gain value of the VAP score is given as the following below:
$$
f(S \cup \{x_i\}) - f(S) = VAP(x_i)
$$
We give below the proof for the submodularity which is based on the \textit{diminishing returns} property as stated in Sec. {\color{red} 3.1}, in the main text.

\begin{proof}
  For two sets $S_1, S_2$ such that $S_1 \subseteq S_2$ and $x_i \in \Omega \backslash S_2$, if the function is submodular it should satisfy the following property in Sec {\color{red} 3.1}.
      \begin{align*}
        f(S_1 \cup \{x_i\}) - f(S_1) &\geq f(S_2 \cup \{x_i\}) - f(S_2)  \\
        VAP(x_i) &\geq  VAP(x_i)
    \end{align*}
    As the left hand side is equal to right hand side, the inequality is satisfied, hence the VAP score function is submodular.
\end{proof}

\noindent \textbf{Submodularity of Diversity Score $d(S, x_i)$:}
The gain in value for the diversity function is given as:
$$
f(S \cup \{x_i\}) - f(S) = \min_{x \in S} D(x, x_i)
$$
We provide the proof that the above gain function corresponds to a submodular function $f(S)$:

\begin{proof}
For two sets $S_1, S_2$ such that $S_1 \subseteq S_2$ and $x_i \in \Omega \backslash S_2$, if the function is submodular it should satisfy the following property in Sec {\color{red} 3.1}:
    \begin{align*}
        f(S_1 \cup \{x_i\}) - f(S_1) &\geq f(S_2 \cup \{x_i\}) - f(S_2) \\
        min_{x \in S_1} D(x, x_i) &\geq  min_{x \in S_2} D(x, x_i)
    \end{align*}
$D(x, x_i) \geq 0$ for every $x$ and $x_i$ as it is a divergence function. As $S_2$ contains more elements than $S_1$, the minimum of $D(x, x_i)$  will be less then for $S_2$ in comparison to that of $S_1$. Hence the final inequality is satisfied which shows that $f(S)$ is submodular.
\end{proof}
\textbf{Submodularity of Representativeness Score $R(S, x_i)$:}
We first prove one property which we will use for analysis of Representativeness Score. \\
\begin{lemma}
\textit{\textbf{Property}: The sum of two submodular set functions $f(S)$ = $f_1(S) + f_2(S)$, is submodular.}
\end{lemma}
\begin{proof}
Let A and B be any two random sets.
  \begin{align*}
    f(A) + f(B) &= f_1(A) + f_2(A) + f_1(B) + f_2(B) \\
                &\geq f_1(A \cup B) + f_2(A \cup B) + f_1(A \cap B) + \\ &\; f_2(A \cap B) \\
                &= f(A \cup B) + f(A \cap B).
  \end{align*}
  Hence the sum of the two submodular functions is also submodular. The result can be generalized to a sum of arbitrary number of submodular functions.
\end{proof}

The representativeness score can be seen as the following set function below:

\begin{align*}
 f(S)  &= \underset{x_i \in \mathcal{D}_u}{\sum} \underset{x_j \in S}{\max}\ s_{ij}  
\end{align*}
We calculate the gain for each sample through this function which is equal to $R(S, x_i)$:
\begin{align*}
    f(S \cup \{x_i\}) - f(S) &=  \underset{x_k \in \mathcal{D}_u}{\sum} \underset{x_j \in S \cup \{x_i\}}{\max} s_{kj} - \underset{x_k \in \mathcal{D}_u}{\sum} \underset{x_j \in S}{\max }\ s_{kj} 
    R(S, x_i) &= \underset{x_k \in \mathcal{D}_u}{\sum} \max(s_{ik} - \underset{x_j \in S}{\max}\ s_{kj}, 0).
\end{align*}
\begin{lemma}
\textit{\textbf{Property}: The set function defined below is submodular:}
$$f(S) = \underset{x_i \in \mathcal{D}_u}{\sum} \underset{x_j \in S}{\max}\ s_{ij}$$
\end{lemma}
\begin{proof}
  We first show that the function $f_i(S) = \underset{x_j \in S}{\max}\ s_{ij}$ is submodular. We first use the property, $f(A) + f(B) \geq f(A \cup B) + f(A \cap B)$ where $A, B$ are two sets, sufficient to show that $f(S)$ is submodular: 
\begin{align}
    f_i(A) + f_i(B) &\geq f_i(A \cup B) + f_i(A \cap B) \\
    \underset{x_j \in A}{\max}\ s_{ij} +  \underset{x_j \in B}{\max}\ s_{ij} &\geq \underset{x_j \in A \cup B}{\max} s_{ij} + \underset{x_j \in A \cap B}{\max} s_{ij} 
 \end{align}
 which follows due to the following:
 $$
 \max(\underset{x_j \in A}{\max}\ s_{ij}, \underset{x_j \in B}{\max}\ s_{ij}) =  \max_{x_j \in A \cup B} s_{ij}
 $$
 and 
 $$
 \min(\underset{x_j \in A}{\max}\ s_{ij}, \underset{x_j \in B}{\max}\ s_{ij}) \geq \max_{x_j \in A \cap B} s_{ij}
 $$
As $f_i(S)$ is submodular, the $f(S)$ can be seen as:
$$
f(S) = \sum_{x_i \in \mathcal{D}_u} f_i(S)
$$
which is submodular according to the property that sum of submodular functions is also submodular proved above.
\end{proof}

\noindent \textbf{Combining the Submodular Functions:} We use the property that a convex combination of the submodular functions is also submodular.
Hence our sampling function which is the convex combination given by:
\begin{equation*}
    f(S \cup \{x_i\}) - f(S) = \alpha VAP(x_i) + \beta d(S, x_i)   + (1 - \alpha - \beta) R(S, x_i).
\end{equation*}

Also follows the property of submodularity.

\subsection{Lemma 2}
Here we present proof of lemma 2 stated in Sec. {\color{red} 4.1.4} of chapter. \\
\textbf{Lemma 2} The set function f(S) defined by equation below is a non-decreasing, monotone function:
   $$ 
    f(S \cup \{x_i\}) - f(S) = \alpha VAP(x_i) + \beta d(S,x_i)  +
     (1 - \alpha - \beta) R(S, x_i)
   $$
    
\begin{proof}
  For the function to be non-decreasing monotone for every set $S$ the addition of a new element should increase value of $f(S)$. The gain function for $f(S)$ is given below:
 \begin{align*}
     &f(S \cup \{ x_i \}) - f(S)  \geq 0 \\
     & \alpha VAP(x_i) + \beta d(S,x_i) +
     (1 - \alpha - \beta) R(S, x_i)  \geq 0
 \end{align*}
 as the $VAP(x_i)$ and $d(S,x_i)$ are KL-Divergence terms, they have value $\geq$ 0. The third term $R(S, x_i) = \underset{x_k \in \mathcal{D}_u}{\sum} \max(s_{ik} - \underset{x_j \in S}{\max}\ s_{kj}, 0)$ is also $\geq$ 0.
 As $0 \leq \alpha, \beta, \alpha + \beta \leq 1$, the value of gain is positive, this shows that the function $f(S)$ is a non-decreasing monotone.
 
\end{proof}

\subsection{Theorem 1}
\textbf{Theorem 1}: Let $S^*$ be the optimal set that maximizes the objective in Eq. \ref{s3vaada_eq:submod_obj} then the solution $S$ found by the greedy algorithm has the following approximation guarantee:
\begin{equation}
    f(S) \geq \left(1 - \frac{1}{e}\right)f(S^*).
\end{equation}
\textbf{Proof: } As $f(S)$ is submodular according to Lemma 1 and is also non-decreasing, monotone according to Lemma 2. Hence the approximation result directly follows from Theorem 4.3 in \cite{nemhauser1978analysis}. The approximation result shows that the algorithm is guaranteed to get at least $63\%$ of the score of the optimal function $f(S^*)$. However, in practice, this algorithm is often able to achieve $98\%$ of the optimal value in certain applications \cite{krause2009optimizing}.
As it's a worst-case result, in practice, we get better performance than in the worst-case scenario. 

\section{Insight for Diversity Score}
\label{s3vaada_diversity}
When the $\alpha = 0$ and $\beta=1$ the gain function $f(S \cup \{x_i\}) - f(S)$ is just $min_{x \in S} D(x,x_i)$. The greedy algorithm described for sampling in Algorithm {\color{red} 1} in chapter, leads to following objective
for selecting sample $x{^*}$.
\begin{equation*}
    x{^*} = \underset{x_i \in \mathcal{D}_u \setminus S}{argmax} \; \underset{x \in S}{\min} \; D(x,x_i)
\end{equation*}
This objective exactly resembles the $K$-Center Greedy method objective which is used by Core-Set method \cite{sener2018active} and is shown to select samples which cover the entire dataset. The $K$-Center Greedy method is very effective in practice. This connection shows that diversity component in our framework also tries to cover the dataset as done by Core-Set \cite{sener2016learning} method which is one of the very effective diversity based active learning method.

\section{Additional Analysis for S$^3$VAADA}
In this sections we provide additional experiments for analysis of the proposed S$^3$VAADA. Unless specified, we run the experiments with single random seed and report the performance. In case the performance difference is small, we provide average results of three runs with different random seeds.
\label{s3vaada_add_analysis}

\begin{figure}[]
  \centering
  \includegraphics[width=0.5\linewidth]{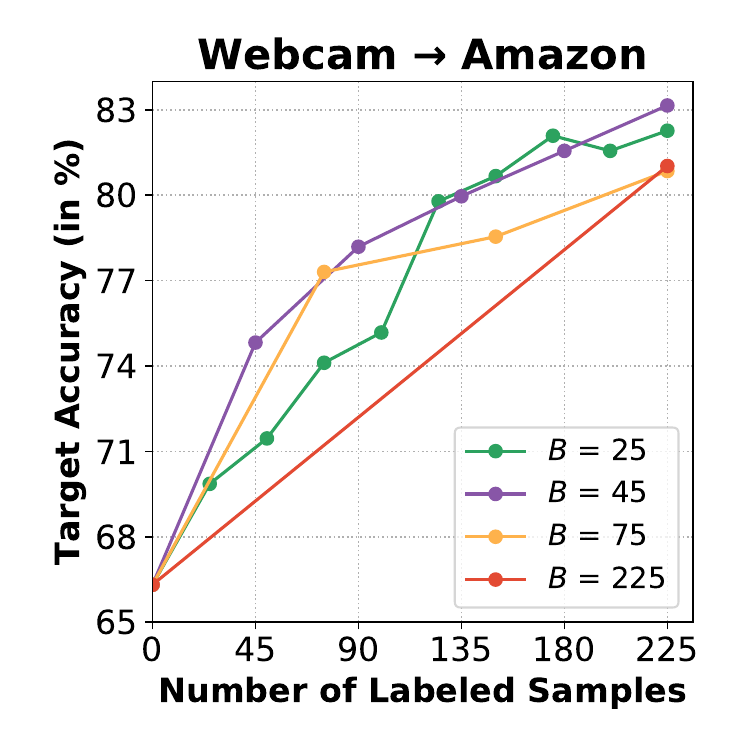}
  \caption{Analysis of S$^3$VAADA for different budget sizes on Webcam $\rightarrow$ Amazon shift of Office-31 dataset.}
  \label{s3vaada_fig:budget-ablation}
\end{figure}

\subsection{Budget Ablation}
Keeping in mind the practical constraint of only having a small amount of labeling budget in the target domain, we restrict ourselves to having a budget size of $2\%$ of the labeled target data. Due to different size of target data in each dataset, the sampling algorithm needs to work robustly under different budget scenario's. For further analysis, we provide results on Webcam to Amazon with different budget sizes $B$ for sampling in Fig. \ref{s3vaada_fig:budget-ablation}. We find that S$^3$VAADA is quite robust for budget sizes greater than 45. We find that small budget of 25 results in more stochasticity in the results.

\begin{figure}[!t]
  \centering
  \includegraphics[width=0.5\linewidth]{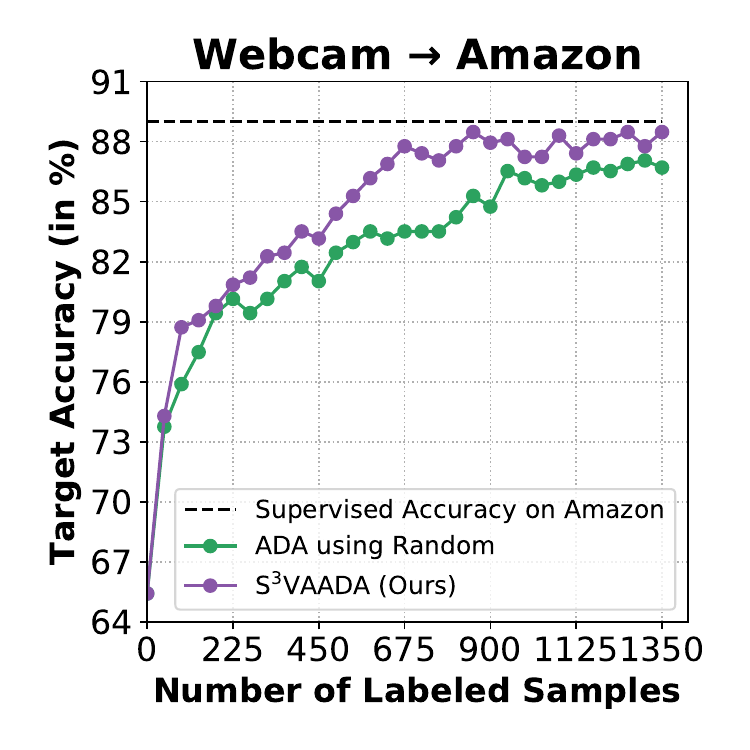}
  \caption{Active DA performance on Webcam $\rightarrow$ Amazon for 30 cycles. We find that the performance converges to supervised learning performance after around 15 cycles.}
  \label{s3vaada_fig:conv}
\end{figure}
\subsection{Convergence: When does the Active DA performance stop improving?}
In all the experiments, we have used a budget of 2\% for 5 rounds which corresponds to 10\% of the target dataset. We find that the performance of algorithm improves in majority of cycles.  This brings up the question, \textit{When does the performance of the model stop improving even after adding more labeled samples?}. For answering this question, we perform experiments on Webcam $\rightarrow$ Amazon and perform active DA for 30 rounds. Fig.~\ref{s3vaada_fig:conv} shows the results on Webcam $\rightarrow$ Amazon with S$^3$VAADA and Random sampling. It can be seen that after around 15 cycles, the gains due to additional samples being added decrease significantly and the performance seems to converge. The performance of the proposed S$^3$VAADA is much better than random sampling in all the rounds. It must also be noted that S$^3$VAADA reaches an accuracy of 89\% with 20 rounds (40\% of the dataset) which is equal to the performance when trained on all the target data.

\section{Analysis of VAADA training}
\label{s3vaada_improved_vada}
We propose VAADA method which is an enhanced version of VADA, suitable for Active DA. We find that proposed improvements in VAADA have a significant effect on the final active DA performance, which we analyse in detail in the following sections. We have done all our analysis using source dataset as Webcam and target dataset as Amazon which is a part of Office-31.
\subsection{Analysis of Learning Rate}
\begin{figure}[h]
  \centering
  \includegraphics[width=0.50\linewidth]{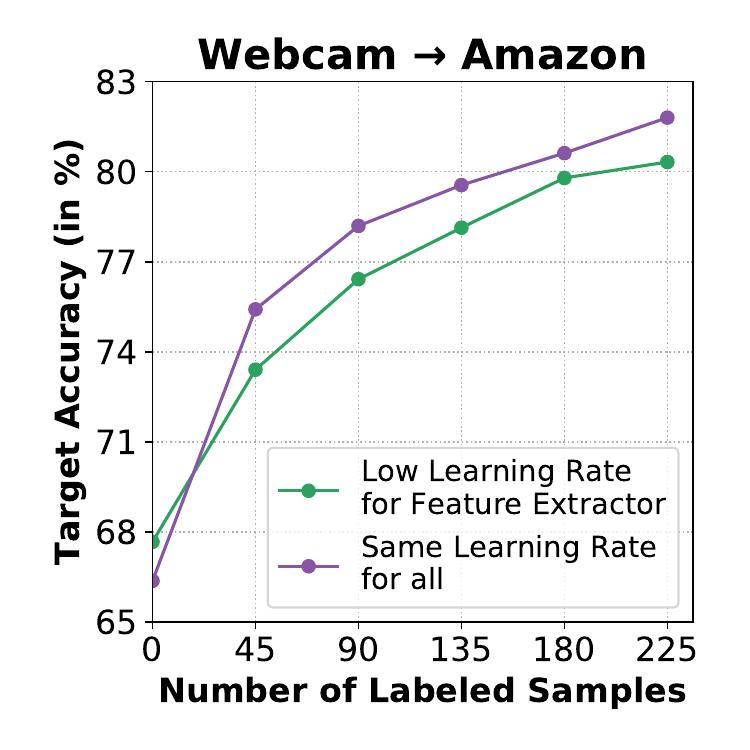}
  \caption{Comparison between Active DA with lower learning rate and a higher learning rate for backbone. The results are the average across three runs with different random seeds.}
  \label{s3vaada_fig:low-lr}
\end{figure}

It is a common practice ~\cite{ganin2015unsupervised, long2018conditional} in domain adaptation (DA) to use a relatively lower learning rate (usually decreased by a multiplying a factor of 0.1) for convolutional backbone which is ResNet-50 in our case. We find that though this practice helps for Unsupervised DA performance, it was not useful in the case of Active DA. In Fig.~\ref{s3vaada_fig:low-lr}, we show the comparison of using same learning rate for the backbone network (as proposed in VAADA), to using a smaller learning rate for backbone. The results clearly show that not lowering the learning is specially helpful for Active DA, whereas it is not for Unsupervised DA.

\subsection{Analysis of using Gradient Clipping}
\begin{figure}[h]
  \centering
  \includegraphics[width=0.5\linewidth]{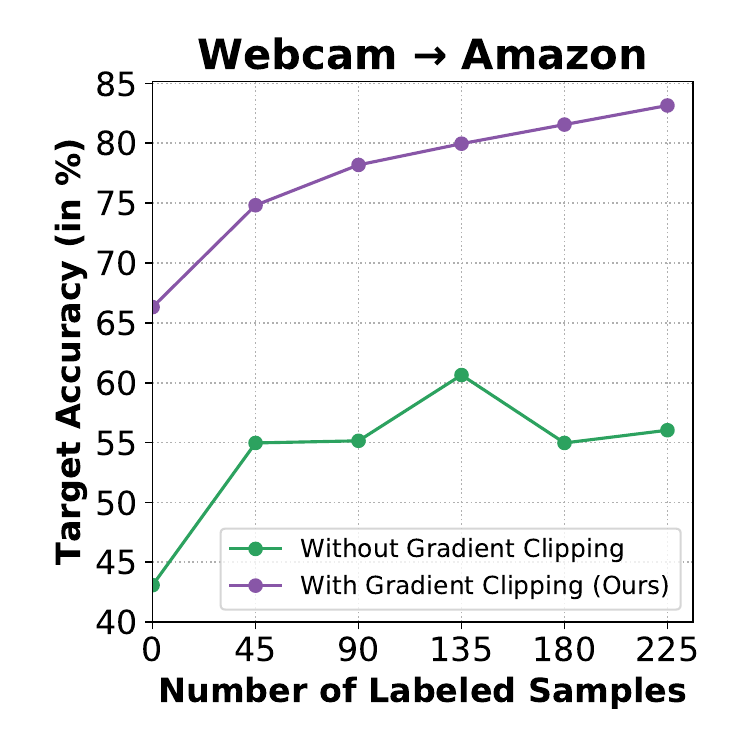}
  \caption{Ablating Gradient clipping on VAADA}
  \label{s3vaada_fig:no-gc}
\end{figure}

In the original implementation of VADA ~\cite{shu2018dirt} the authors use the method of Exponential Moving Average (EMA) (also known as Polyak Averaging ~\cite{polyak1992acceleration}) of model weights, which increases the stability of results. In place of EMA, we find that using proposed Gradient Clipping in VAADA works  better for stabilizing the training. In Gradient Clipping, we scale the gradients such that the gradient vector norm has magnitude 1. We find that Gradient Clipping allows the network to train stably, with a relatively high learning rate of 0.01. For showcasing the stabilising effect of Gradient Clipping, in Fig. \ref{s3vaada_fig:no-gc} we compare the performance of the model with and without gradient clipping. We find that Gradient Clipping leads to a increase of accuracy of above 10\% for each active learning cycle, with achieving stable increase in performance with the addition of more labels. On the other hand the model without clipping is unable to produce stable increase in performance with addition of labels. 

\subsection{Comparison of VADA with VAADA}
\begin{figure}[htp]
  \centering
  \includegraphics[width=\linewidth]{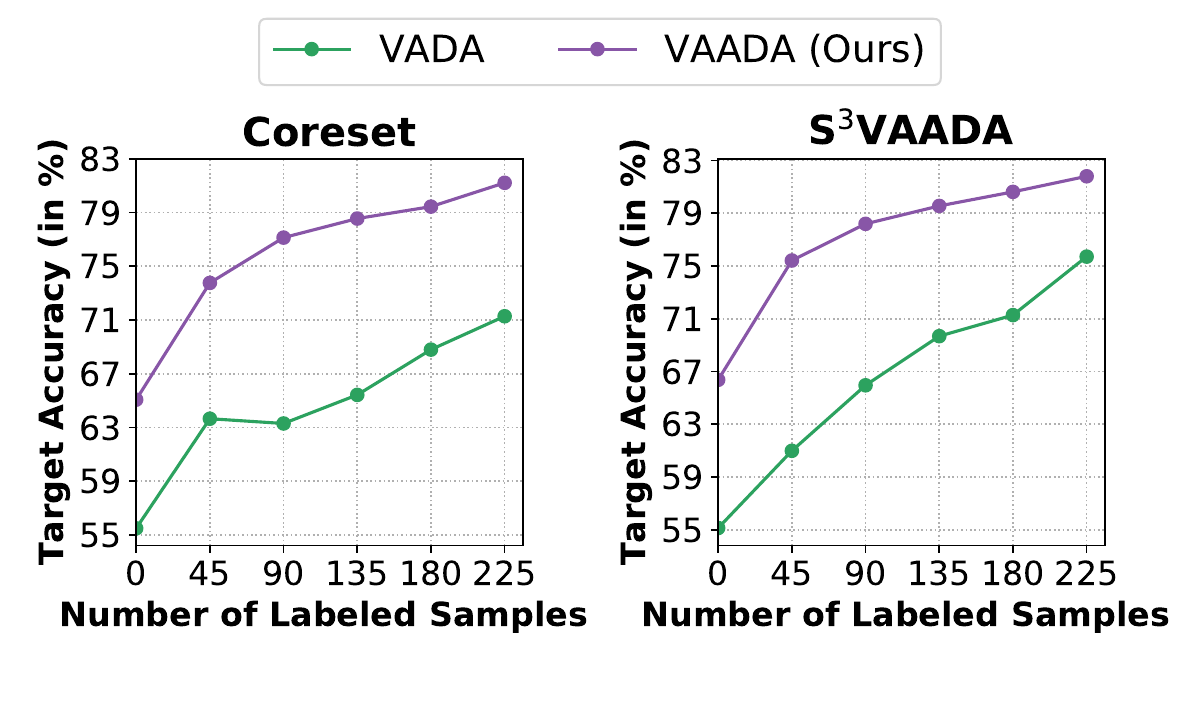}
  \vspace{-10mm}
  \caption{We show the comparison of VAADA and VADA. We see consistent improvement of VAADA over VADA  across all cycles.}
  \label{s3vaada_fig:ivada}
\end{figure}

In this section we provide additional implementation details and analysis, continuing from Sec. {\color{red} 6} of chapter. The comparison shown with the VADA method corresponds to the original VADA configuration specified in \cite{shu2018dirt}. In the original implementation, the authors propose to use Adam optimizer and EMA for training. We use Adam with learning rate of 0.0001 and use the exact same settings as in \cite{shu2018dirt}. It can be seen in Fig.~\ref{s3vaada_fig:ivada} that VAADA consistently outperforms the VADA training in Active DA for CoreSet and S$^3$VAADA as well.

\subsection{Visualizing clusters using t-SNE}
\begin{figure*}[h]
  \centering
   \subcaptionbox{DANN}{\includegraphics[width=0.45\linewidth]{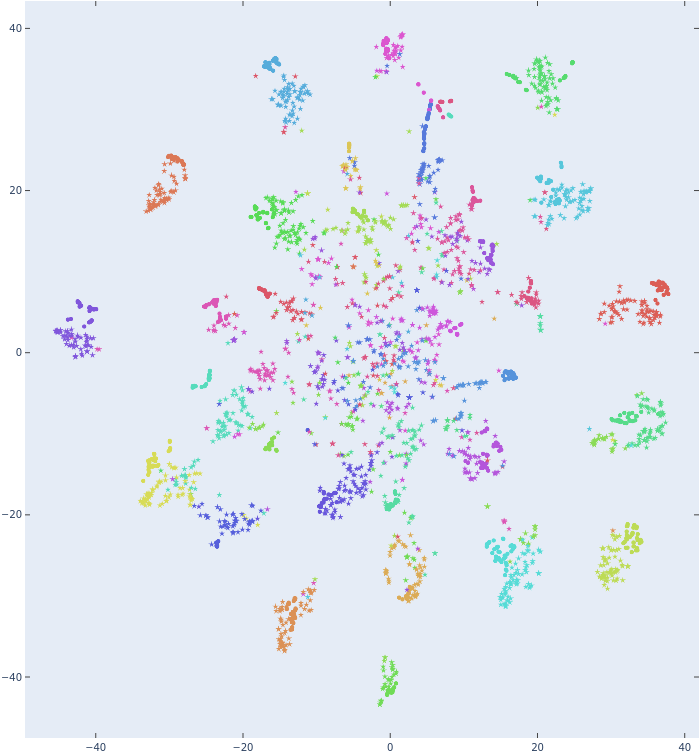}}
   \subcaptionbox{VAADA}{\includegraphics[width=0.45\linewidth]{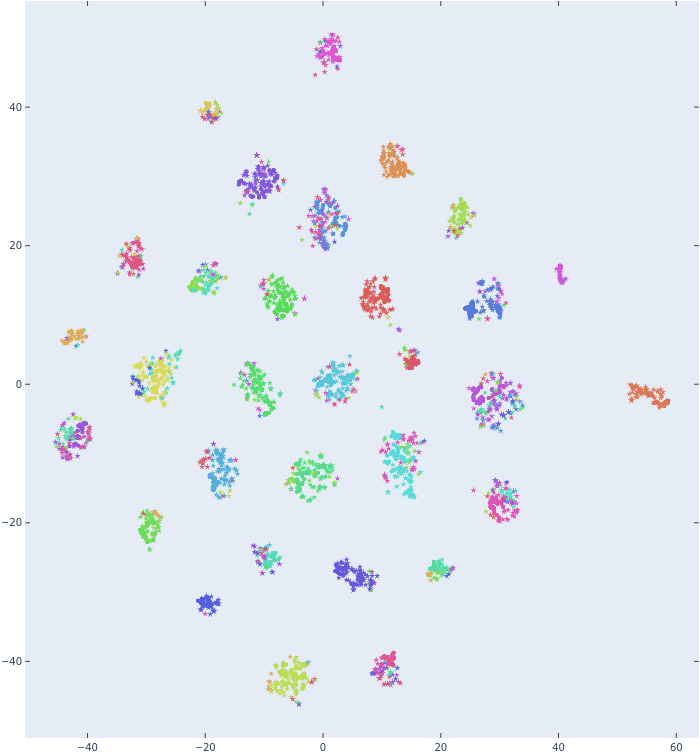}}
  \caption{Visualization of clusters of data points formed by DANN and VAADA on DA for Webcam $\rightarrow$ Amazon. Different colors represent different classes. It can seen that VAADA forms much distinct clusters data than DANN.}
  \label{s3vaada_fig:dann-vada}
\end{figure*}

In this section, we analyse the t-SNE plot (Fig.~\ref{s3vaada_fig:dann-vada}) of the two different training methods i.e., DANN and VAADA. We find that in VAADA training, there is formation of distinct clusters and also the cluster sizes are similar.  Whereas in DANN t-SNE, there is no formation of distinct clusters, and a large portion of sample are clustered in between. This shows that additional losses of conditional entropy and smoothing through Virtual Adversarial Perturbation loss are necessary to enforce the cluster assumption.

\subsection{Hyper-Parameter Sensitivity of VAADA}
\begin{figure}[h]
  \centering
  \includegraphics[width=0.50\linewidth]{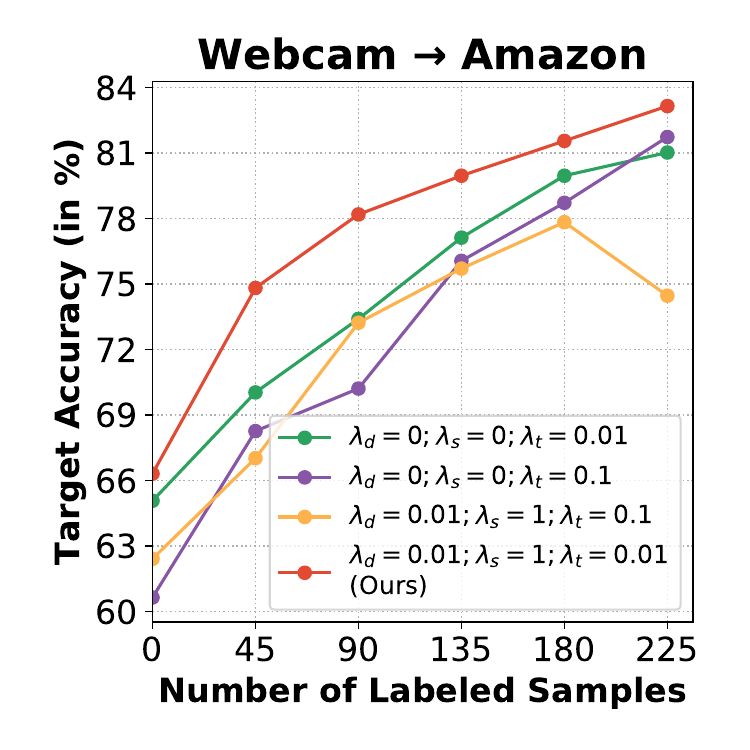}
  \caption{Different Hyperparameters on Webcam $\rightarrow$ Amazon dataset.}
  \label{s3vaada_fig:lambda-ablation}
\end{figure}

We used the same $\lambda$ values mentioned as a robust choice by VADA \cite{shu2018dirt} authors, for VAADA training, setting $\lambda_d = 0.01$, $\lambda_s = 1$ and $\lambda_t = 0.01$ across all datasets. For analysing the sensitivity of the performance of VAADA across different hyper-parameter choices, we provide results with varying $\lambda$ parameters in Fig. \ref{s3vaada_fig:lambda-ablation}. We also find that the robust choice recommended for VADA, also works the best for VAADA. Hence, this \textit{fixed-set} of robust $\lambda$ parameters can be used across datasets with varying degree of domain shifts. This is also enforced by the fact, that in all our experiments these \textit{fixed} hyperparameters were able to achieve state-of-the-art performance across datasets. This decreases the need for hyper-parameter tuning specific to each dataset. 

\section{Implementation Details}
\label{s3vaada_experimental_details}
\subsection{Configuration for DANN}
For the DANN experiments, we use a batch size of 36 with a learning rate of 0.01 for all the linear layers. We use a smaller learning rate of 0.001 for the ResNet-50 backbone. DANN is trained with SGD with a momentum of 0.9 and weight decay value of 0.0005 following the schedule described in \cite{ganin2015unsupervised}. The model architecture and hyperparameters are same as in \cite{long2018conditional}. The model is trained for 10,000 iterations as done in \cite{long2018conditional} and the best validation accuracy is reported in the graphs.
\subsection{Configuration for SSDA (MME*)}
We use the author's implementation\footnote{https://github.com/VisionLearningGroup/SSDA\_MME} for experiments on Office dataset. We used ResNet-50 as backbone and used same parameters as used in their implementation. For Active DA, we initially train the model with no labeled target data and keeps on adding 2\% of the unlabeled target data to labeled target set for 5 cycles. We train the model for 20,000 iterations. A similar procedure of reporting the best validation accuracy on the fixed validation set, as done for other baselines is followed.
\subsection{Configuration for VAADA}
The model is trained with a batch size of 16 and a learning rate of 0.01 for all the layers using the SGD Optimizer with a momentum of 0.9. A weight decay of 0.0005 was used. The model is trained for 100 epochs and the best accuracy is reported in the graphs. A ResNet-50 backbone is used with pretrained ImageNet weights. The architecture for various model components used are shown in Table \ref{s3vaada_tab:gen} and \ref{s3vaada_tab:clf}. Same architecture is used for all experiments in the chapter. 

\begin{table}[!t]
    \centering
    \begin{tabular}{c||c}
    \hline
       Layer/Component  &  Output Shape\\
       \hline
        - & 224 $\times$ 224 $\times$ 3 \\
        ResNet-50 & 2048 \\
        Linear & 256 \\
        \hline
    \end{tabular}
    \caption{\textbf{Feature Generation $g_{\theta}$}: Architecture used for generating the features}
    \label{s3vaada_tab:gen}
\end{table}

\begin{table}[!t]
    \centering
    \begin{tabular}{c||c}
    \hline
       Layer  &  Output Shape\\
       \hline
       \multicolumn{2}{c}{\textbf{Feature Classifier ($f_{\theta}$)}} \\
       \hline
        - & 256 \\
        Linear & $C$ \\
        \hline
        \multicolumn{2}{c}{\textbf{Domain Classifier} ($D_\phi$)} \\
        \hline
        - & 256 \\
        Linear & 1024 \\
        ReLU & 1024 \\
        Linear & 1024 \\
        ReLU & 1024 \\
        Linear & 2 \\
        \hline
    \end{tabular}
    \caption{Architecture used for feature classifier and Domain classifier. $C$ is the number of classes. Both classifiers will take input from feature generator ($g_\theta$).}
    \label{s3vaada_tab:clf}
\end{table}

The above hyper parameters are used for all our experiments on Office-Home and Office-31 datasets. We just change the batch size to 128 and use the learning rate decay schedule of DANN for experiments on VisDA-18 dataset.

\begin{equation}
\begin{split}
\begin{aligned}
                     L(\theta ; \mathcal{D}_{s}, \mathcal{D}_t, \mathcal{D}_u) = L_y(\theta; \mathcal{D}_s, \mathcal{D}_t) + \lambda_dL_d(\theta; \mathcal{D}_s, \mathcal{D}_t, \mathcal{D}_u) \notag\\ + \lambda_sL_{v}(\theta; \mathcal{D}_s \cup \mathcal{D}_t) + \lambda_t(L_v(\theta; \mathcal{D}_u) + L_c(\theta; \mathcal{D}_u)) \end{aligned}
\end{split}
\end{equation}
The $\epsilon$ used in Eq. 3 and 4 in the chapter refer to the maximum norm of the virtual adversarial perturbation,  was set it to 5 in our experiments.
The value of the number of random restarts ($N$) to generate virtual adversarial perturbation for the proposed sampling is set to 5. 
The $\alpha$ value is set to 0.5 and $\beta$ value is set to 0.3 across all experiments. We use Gradient Clipping to clip the norm of the gradient vector to 1 to stabilize and accelerate VAADA.%
\\
We used Weights \& Biases \cite{wandb} to track our experiments. 
\begin{figure}[h]
  \centering
  \includegraphics[width=0.50\linewidth,]{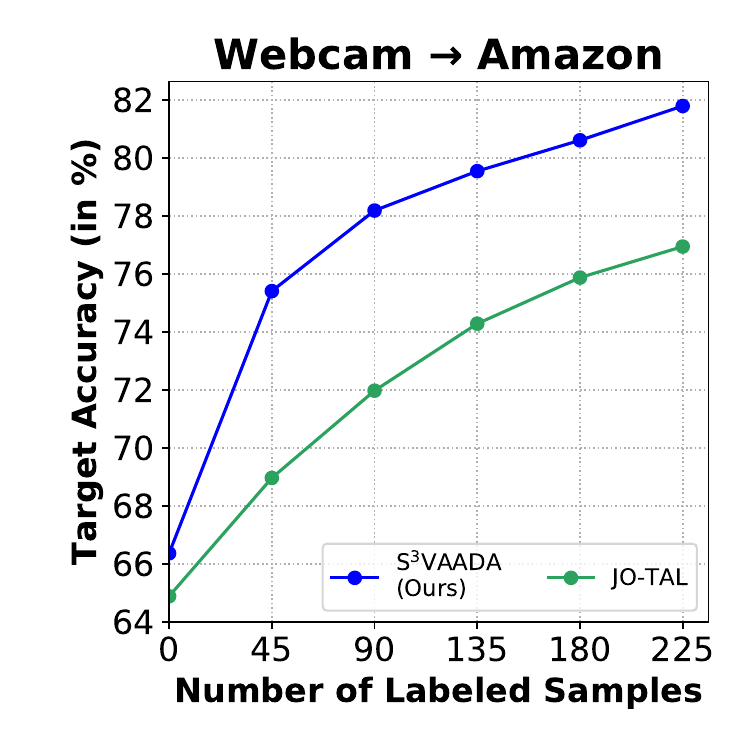}
  \caption{S$^3$VAADA vs. JO-TAL}
  \label{s3vaada_fig:jotal}
\end{figure}
\section{Comparison with JO-TAL}
We compare our results with JO-TAL by Chattopadhyay et al. \cite{chattopadhyay2013joint} (by implementing in \texttt{cvxopt}). JO-TAL performs both active learning and domain adaptation in a single step. Since, JO-TAL was not proposed in the context of deep learning, we use deep features from ImageNet pretraind model and train an SVM classifier on top of them. The optimization problem was implemented in \texttt{cvxopt}. Fig. \ref{s3vaada_fig:jotal} shows S$^3$VAADA achieves significant performance gains across cycles when compared to JO-TAL.\\

\section{Comparison with Alternate Adversarial Perturbation based sampling}
\label{s3vaada_Comparison with alternate Adversarial Sampling}
\begin{figure}[h]
  \centering
  \includegraphics[width=0.50\linewidth]{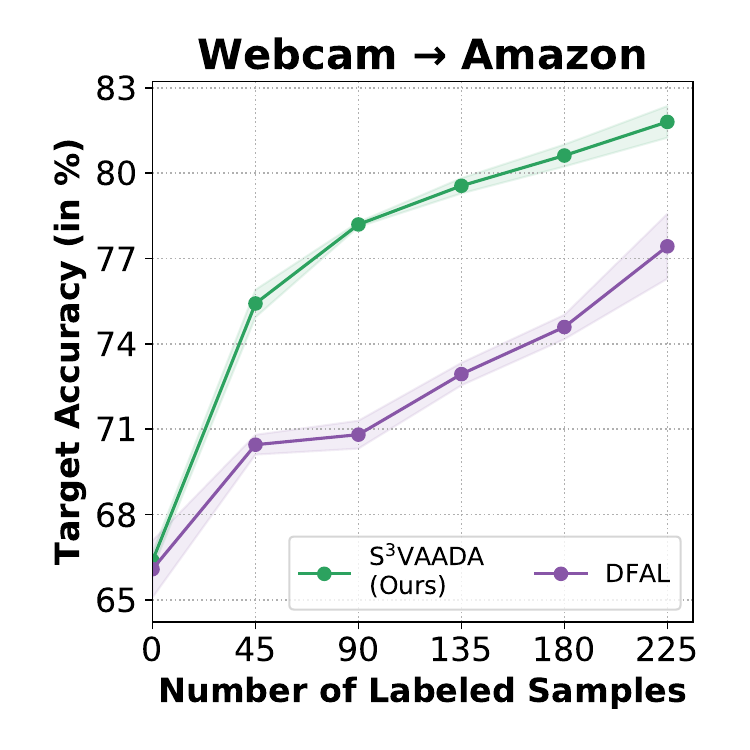}
  \caption{S$^3$VAADA outperforms DFAL in all the cycles, even though both attain same initial accuracy. It shows that S$^3$VAADA selects much more informative samples compared to DFAL.}
  \label{s3vaada_fig:dfal-svap}
\end{figure}

There also exists a sampling method \cite{ducoffe2018adversarial} based on DeepFool adversarial perturbations \cite{dezfooli2016deepfool} Active Learning (DFAL) but due to its higher complexity and computation time, it was unfeasible for us to use it as a baseline for all experiments. We provide the comparison of DFAL with S$^3$VAADA in terms of accuracy on Active DA from Webcam $\rightarrow$ Amazon in Fig. \ref{s3vaada_fig:dfal-svap}. Training is done through VAADA for both sampling methods. %
We find that S$^3$VAADA significantly outperforms DFAL sampling achieving better results in all cycles.
\begin{figure}[!t]
  \centering
  \includegraphics[width=0.50\linewidth]{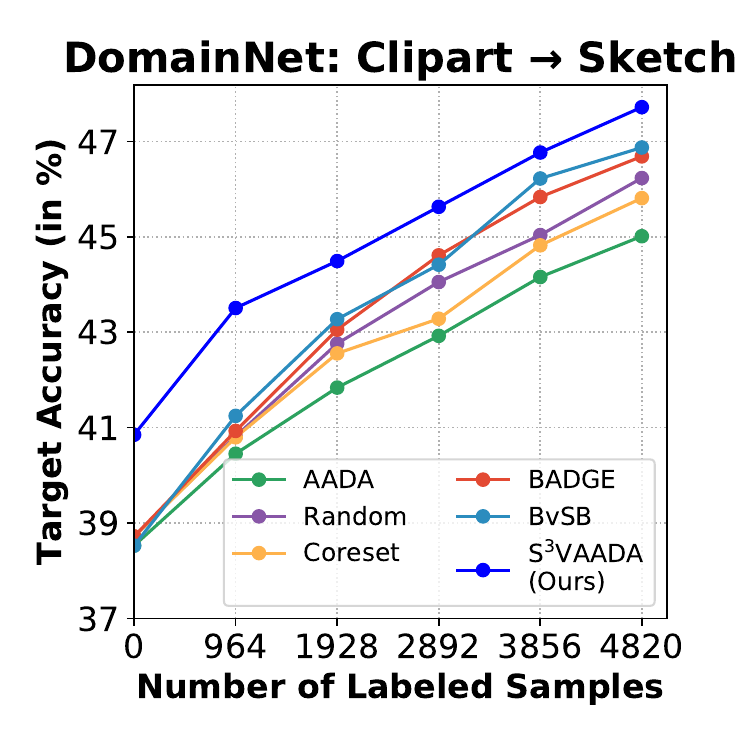}
  \caption{Active Domain Adaptation results on Clipart $\rightarrow$ Sketch dataset. This shows the proposed method is scalable to larger datasets.}
  \label{s3vaada_fig:c2s}
\end{figure}

\begin{figure}[]
  \centering
  \subcaptionbox{Webcam}{\fbox{\includegraphics[width=0.45\linewidth]{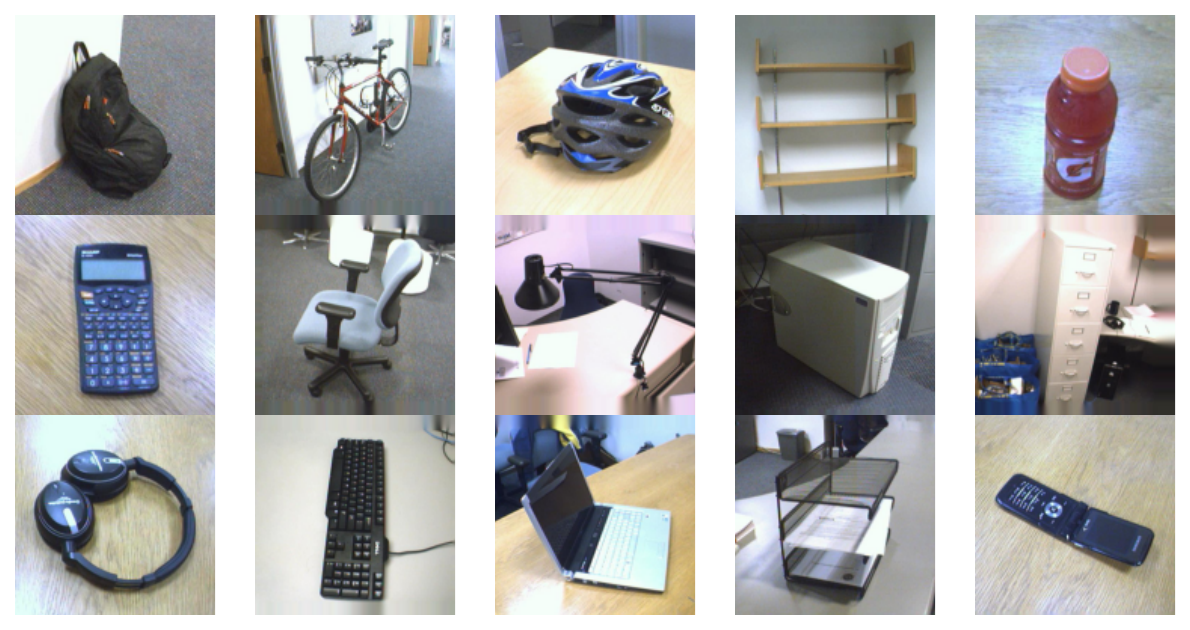}}}
  \subcaptionbox{DSLR}{\fbox{\includegraphics[width=0.45\linewidth]{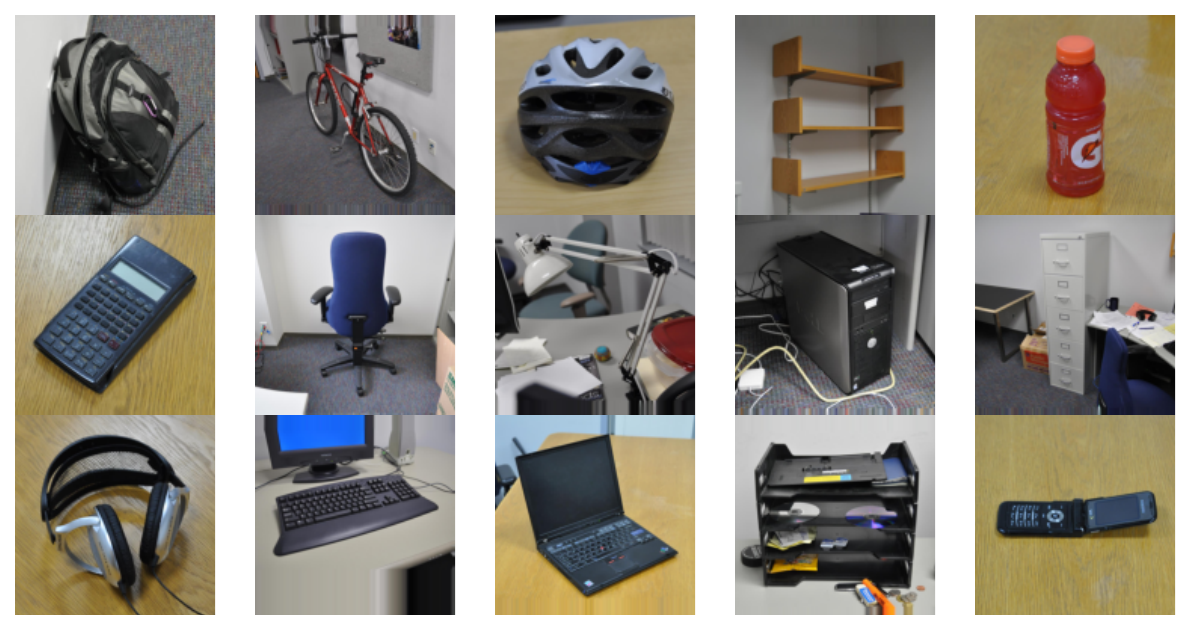}}}
  \subcaptionbox{Amazon}{\fbox{\includegraphics[width=0.45\linewidth]{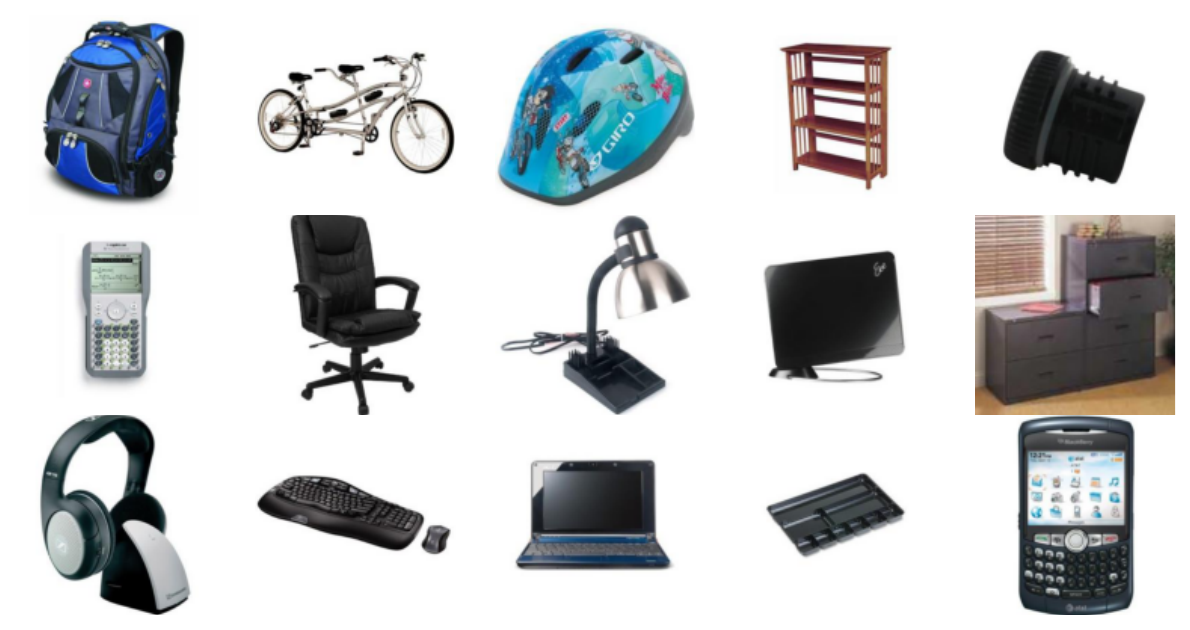}}}
  \caption{Some Office-31 Dataset examples}
  \label{s3vaada_fig:office-31-images}
\end{figure}

\begin{figure}[h]
  \centering
  \subcaptionbox{Art}{\fbox{\includegraphics[width=0.45\linewidth]{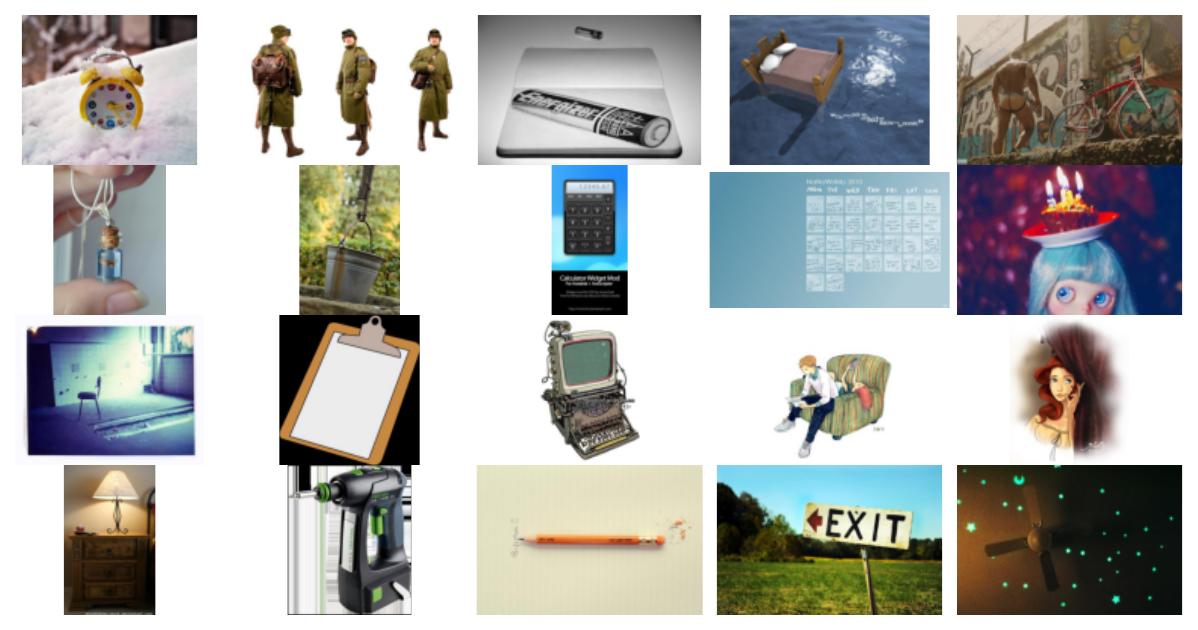}}}
  \subcaptionbox{Product}{\fbox{\includegraphics[width=0.45\linewidth]{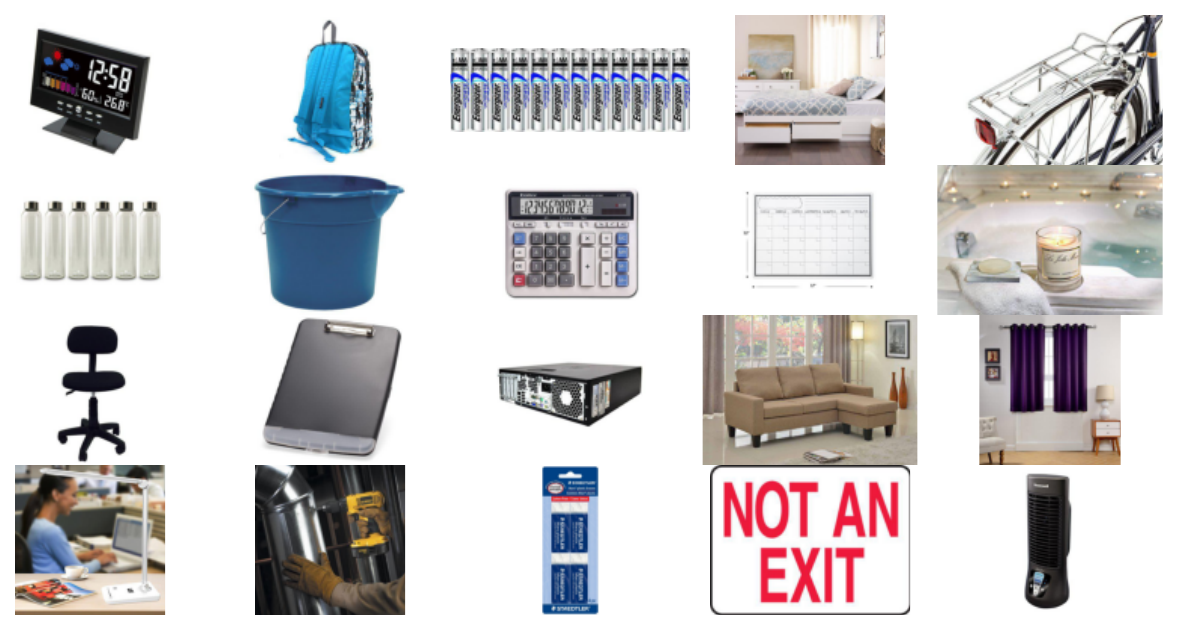}}}
  \subcaptionbox{Clipart}{\fbox{\includegraphics[width=0.45\linewidth]{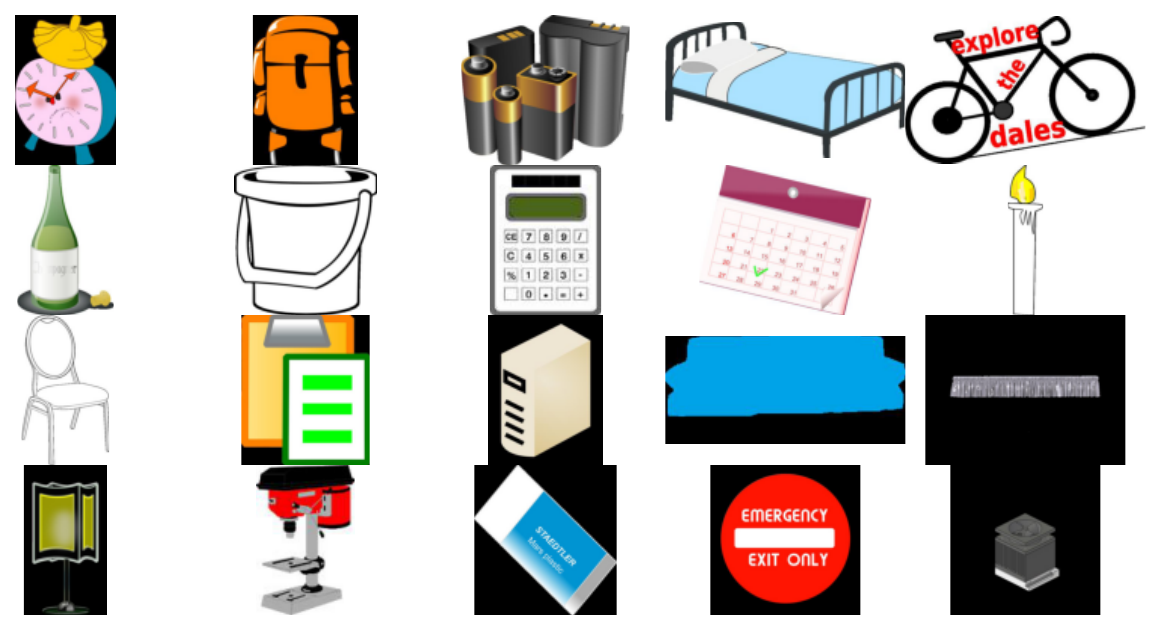}}}
  \subcaptionbox{Real World}{\fbox{\includegraphics[width=0.45\linewidth]{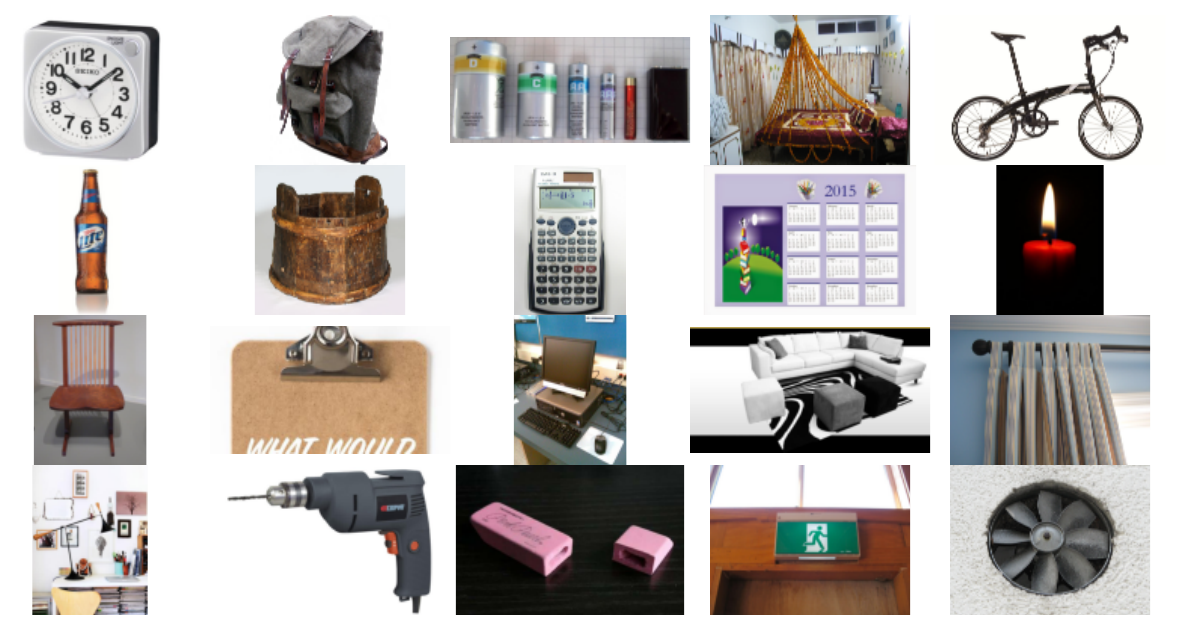}}}
  \caption{Some Office-Home Dataset examples}
  \label{s3vaada_fig:office-home-images}
\end{figure}

\begin{figure}[h]
  \centering
  \subcaptionbox{Real}{\includegraphics[width=0.45\linewidth]{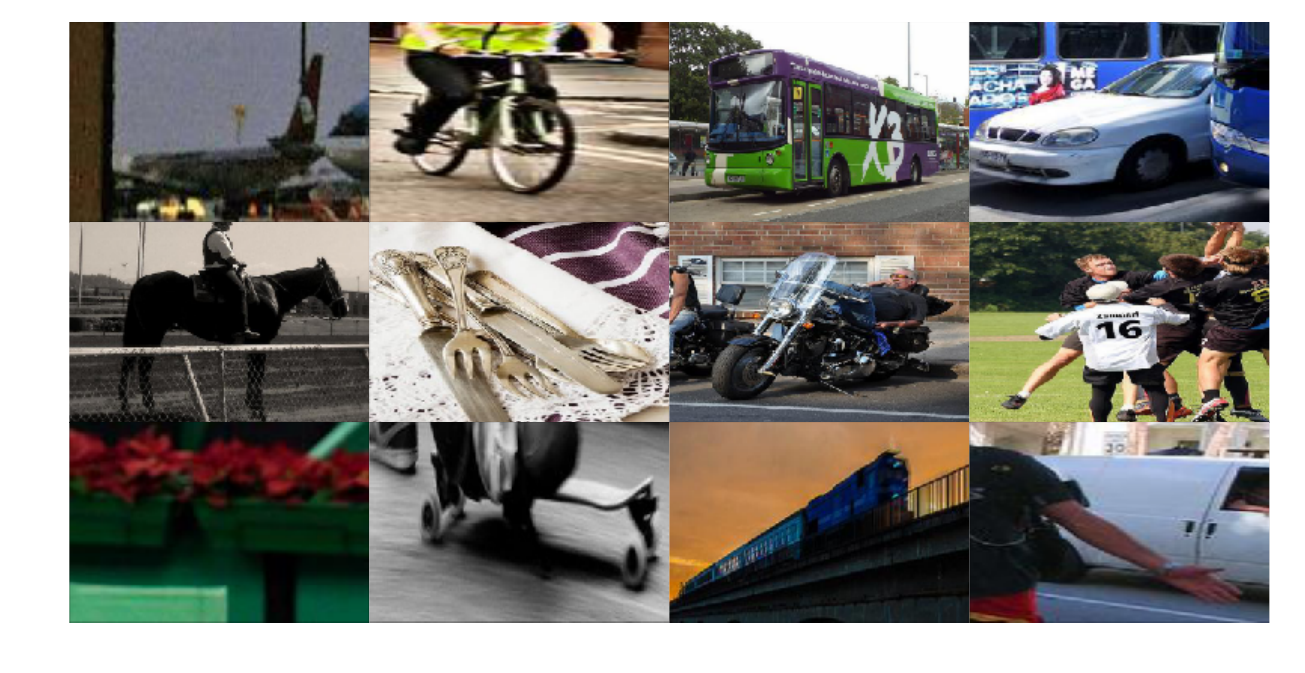}}
  \subcaptionbox{Synthetic}{\includegraphics[width=0.45\linewidth]{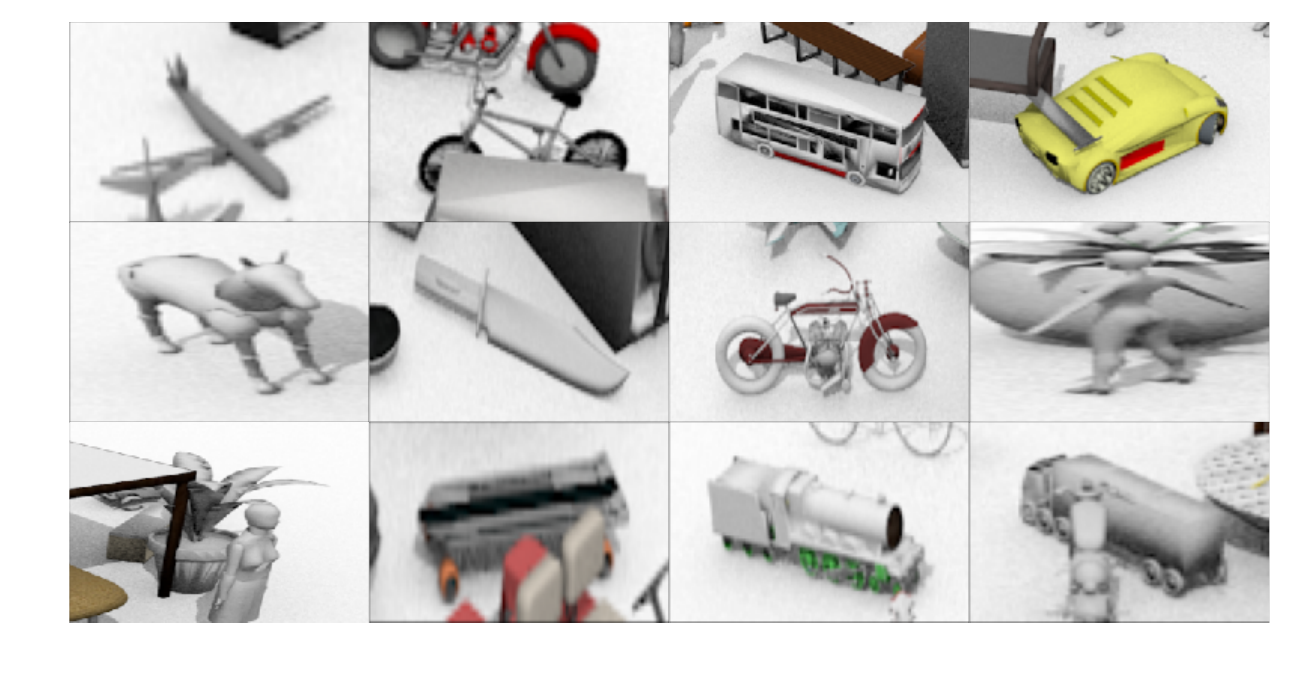}}
  \caption{Some Visual DA (VisDA-18) Dataset examples}
  \label{s3vaada_fig:visda-images}
\end{figure}

\section{Description of Datasets Used}
\label{s3vaada_datasets}
\textbf{Office-31 \cite{saenko2010adapting}:} It has images from 3 domains i.e., Webcam, DSLR and Amazon, belonging to 31 classes.

\noindent \textbf{Office-Home \cite{venkateswara2017Deep}:} This dataset has a more severe domain shift across domains  compared to Office-31. It is a 65 class dataset and contains images from 4 domains namely, Art, Clipart, Product and Real World. 

\noindent \textbf{VisDA-18 \cite{8575439}:} This dataset consist of images from synthetic and real domains. The dataset has annotations for for two tasks: image classification and image segmentation. We used dataset of image classification task. It has 12 different object categories.

Some example images of each dataset are shown in Figs. \ref{s3vaada_fig:office-31-images}, \ref{s3vaada_fig:office-home-images} and \ref{s3vaada_fig:visda-images}.

\section{DomainNet Experiments}
DomainNet \cite{peng2019moment} consists of about 0.6 million images belonging to 345 classes. The images belong to 5 domains: Clipart, Sketch, Quickdraw, Painting, Real. For showing the scalability of our method, we use Clipart as the source and Sketch as the target domain. The Clipart domain consists of 33,525 images in the train set and 14,604 images in the test set. The Sketch domain consists of 50,416 images in the train set and 21,850 images in the test set. Due to computational limitations we are not able to provide results on different possible domains.

For the DomainNet experiments, we use a batch size of 36 with a learning rate scheduler same as DANN and run each baseline for 30 epochs. We use Gradient Clipping and clip the norm to 10. In this case we find that performance of S$^3$VAADA does not stagnate at 30 epochs but due to limitations of compute we only train for 30 epochs. Hence, there exist scope for improvement in results with parameter tuning and more computational budget.

Fig.~\ref{s3vaada_fig:c2s} shows the results on Clipart $\rightarrow$ Sketch domain shift. The performance of S$^3$VAADA outperforms all the other techniques in all the cycles. This shows that the efficacy of the proposed method on a large dataset containing 345 classes. 

\section{Future Extension to Other Applications}
The S$^3$VAADA technique is based on the idea of cluster assumption i.e., aligning of clusters of different classes, which is used in the sampling method. Some recent DA techniques for Object Detection \cite{xu2020cross} and Image Segmentation \cite{wang2020classes} which aim for classwise alignment of features, can be seen as methods which satisfy the cluster assumption. Hence, we hope that combining such techniques with our method can yield good Active DA techniques tailored for these specific applications. In the current work, we focused on diverse image classification tasks, leaving these applications for future work.

\renewcommand{\thesection}{H}

\supersection{A Closer Look at Smoothness in Domain Adversarial Training (Chapter-10)}

\label{app:s3vaada}

\section{Notation Table}
\label{sdat_app:notn_tab}
Table \ref{sdat_tab:notation} contains all the notations used in the chapter and the proofs of theorems. 
\begin{table}[h!]
 \caption{The notations used in the chapter and the corresponding meaning.}
    \centering
    \begin{tabular}{l|l}
    \hline
         \textbf{Notation} & \textbf{Meaning} \\
        \hline \hline
         $S$ & Labeled Source Data \\
         $T$ & Unlabelled Target Data \\
         $P_S$ (or $P_T$) & Source (or Target) Distribution \\
         $\mathcal{X}$ & Input space \\
         $\mathcal{Y}$ & Label space \\
         $y(\cdot)$ & Maps image to labels \\
         $h_\theta$ & Hypothesis function \\
         $R_S^l(h_\theta)$ (or $R_T^l(h_\theta)$) & Source (or Target) risk \\
         $\hat{R}_S^l(h_\theta)$ (or $\hat{R}_T^l(h_\theta)$) & Empirical Source (or Target) risk \\
         $\mathcal{H}$ & Hypothesis space \\
         $ D_{h_{\theta},\mathcal{H}}^\phi(P_S || P_T)$ & Discrepancy between two domains $P_S$ and $P_T$ \\
         $g_{\psi}$ & Feature extractor \\
         $f_{\Theta}$ & Classifier \\
         $\mathcal{D}_{\Phi}$ & Domain Discriminator \\
         $d_{S,T}^{\Phi}$ & Tractable Discrepancy Estimate\\
         $\nabla^2_{\theta} \hat{R}_S^l(h_{\theta})$ (or $H$) & Hessian of classification loss\\
         $Tr(H)$ & Trace of Hessian\\
         $\lambda_{max}$ & Maximum eigenvalue of Hessian\\
         $\epsilon$ & Perturbation \\
         $\rho$ & Maximum norm of $\epsilon$\\
         
    \end{tabular}
   
    \label{sdat_tab:notation}
\end{table}

\section{Connection of Discrepancy to $d_{S,T}^\Phi$ (Eq. \textcolor{red}{4}) in chapter}\label{sdat_app:discrepancy}
We refer reader to Appendix \textcolor{red}{C.2} of \citet{acuna2021f} for relation of $d_{S,T}^\Phi$. The $d_{S,T}^\Phi$ term defined in Eq. \textcolor{red}{4} given as:
\begin{equation}
    d_{S,T}^{\Phi} = \mathbb{E}_{x \sim P_S}[\log(\mathcal{D}_{\Phi}(g_{\psi}(x)))] + \mathbb{E}_{x \sim P_T}[ \log(1-\mathcal{D}_{\Phi}(g_{\psi}(x)))]
\end{equation}
the above term is exactly the Eq. \textcolor{red}{C.1} in \citet{acuna2021f} where they show that optimal $d_{S,T}^\Phi$ i.e.:
\begin{equation}
    \max_{\Phi} d_{S,T}^{\Phi} = D_{JS}(P_S||P_T) - 2\log(2).
\end{equation}
 Hence we can say from result in Eq. \textcolor{red}{4} is a consequence of Lemma 1 and Proposition 1 in \citep{acuna2021f}, assuming that $D_{\Phi}$ satisfies the constraints in Proposition 1.

\section{Proof of Theorems}\label{sdat_app:proof}
In this section we provide proofs for the theoretical results present in the chapter:
\setcounter{theorem}{0}
\begin{theorem}[\textbf{Generalization bound}]
Suppose $l: \mathcal{Y} \times \mathcal{Y} \rightarrow [0,1] \subset dom \; \phi^*$. Let $h^*$ be the ideal joint classifier with error $\lambda^* = R_S^l(h^*) +  R_T^l(h^*)$. We have the following relation between source and target risk:
\begin{equation}
    R_{T}^l(h_{\theta}) \leq R_{S}^{l}(h_{\theta}) + D_{h_{\theta}, \mathcal{H}}^{\phi} (P_S || P_T) + \lambda^*
\end{equation}
\end{theorem}
\begin{proof}
We refer the reader to Theorem 2 in Appendix \textcolor{red}{B} of \citet{acuna2021f} for the detailed proof the theorem.
\end{proof}
We now introduce a Lemma for smooth functions which we will use in the proofs subsequently:
\setcounter{theorem}{0}
\begin{lemma}
\label{sdat_lem:lsmooth}
For an L-smooth function $f(w)$ the following holds where $w^*$ is the optimal minima:
$$
f(w) - f(w^*) \geq \frac{1}{2L}  || \nabla f(w) ||^2
$$
\end{lemma}
\begin{proof}
The L-smooth function by definition satisfies the following:
$$
f(w^*) \leq f(v) \leq f(w) + \nabla f(w) (v - w) + \frac{L}{2}||v - w||^2
$$
Now we minimize the upper bound wrt $v$ to get a tight bound on $f(w^*)$.
$$
D(v) = f(w) + \nabla f(w) (v - w) + \frac{L}{2}||v - w||^2
$$
after doing $\nabla_{v} D(v) = 0$ we get:
$$
v = w - \frac{1}{L} \nabla f(w) 
$$
By substituting the value of $v$ in the upper bound we get:
$$
f(w^*) \leq f(w) - \frac{1}{2L} || \nabla f(w) ||^2
$$
Hence rearranging the above term gives the desired result:
$$
f(w) - f(w^*) \geq \frac{1}{2L}  || \nabla f(w) ||^2.
$$

\end{proof}

\setcounter{theorem}{1}
\begin{theorem}
For a given classifier $h_{\theta}$ and one step of (steepest) gradient ascent i.e. $\Phi' = \Phi + \eta (\nabla d_{S,T}^{\Phi}/||\nabla d_{S,T}^{\Phi}||)$ and $\Phi'' = \Phi + \eta (\nabla d_{S,T}^{\Phi}|_{\Phi + \hat{\epsilon}(\Phi)}/||\nabla d_{S,T}^{\Phi}|_{\Phi + \hat{\epsilon}(\Phi)}||)$ for maximizing
\begin{equation}
\begin{split}
         d_{S,T}^{\Phi'} - d_{S,T}^{\Phi''} \leq  \eta(1 - \cos \alpha)\sqrt{2L(d^*_{S,T} - d^{\Phi}_{S,T}) }
\end{split}
\end{equation}
where $\alpha$ is the angle between $\nabla d_{S,T}^{\Phi}$ and $\nabla d_{S,T}^{\Phi}|_{\Phi + \hat{\epsilon}(\Phi)}$. 
\end{theorem}

\begin{proof}[Proof of Theorem \ref{sdat_th:suboptimality}]
We assume that the function is $L$-smooth (the assumption of L-smoothness is the basis of many results in non-convex optimization \citep{carmon2020lower}) in terms of input $x$. As for a fixed $h_{\theta}$ as we use a reverse gradient procedure for measuring the discrepancy, only one step analysis is shown. This is because only a single step of gradient is used for estimating discrepancy $d^{\Phi}_{S,T}$ i.e. one step of each min and max optimization is performed alternatively for optimization. After this the $h_{\theta}$ is updated to decrease the discrepancy. Any differential function can be approximated by the linear approximation in case of small $\eta$:
\begin{equation}
    d^{\Phi + \eta v}_{S,T} \approx d^{\Phi}_{S,T} + \eta \nabla {{d^{\Phi^{\mathbf{T}}}_{S,T}}}v
\end{equation}
the dot product between two vectors can be written as the following function of norms and angle $\theta$ between those:
\begin{equation}
    \nabla d^{\Phi^{\mathbf{T}}}_{S,T}v = || \nabla d^{\Phi}_{S,T} || \; ||v|| \; cos \theta 
\end{equation}
The steepest value will be achieved when $\cos \theta = 1$ which is actually $v =  \frac{\nabla d^{\Phi}_{S,T}(x)}{||\nabla d^{\Phi}_{S,T}(x)||}$. Now we compare the descent in another direction $v_2 = \frac{\nabla d^{\Phi}_{S,T}|_{w + \epsilon(w)}}{||\nabla d^{\Phi}_{S,T}|_{w + \epsilon(w)}||}$ from the gradient descent. The difference in value can be characterized by:
\begin{equation}
     d^{\Phi + \eta v}_{S,T} - d^{\Phi + \eta v_2}_{S,T}= \eta ||\nabla d^{\Phi}_{S,T}||(1 - \cos \alpha) 
\end{equation}
As $\alpha$ is an angle between $\nabla d^{\Phi}_{S,T}|_{w + \epsilon(w)} \; (v_2)$ and $\nabla d^{\Phi}_{S,T}(X) \; (v)$. The suboptimality is dependent on the gradient magnitude. We use the following result to show that when optimality gap $ d^{*}_{S,T} - d^{\Phi}_{S,T}(x)$ is large the difference between two directions is also large.

For an L-smooth function  the following holds according to Lemma \ref{sdat_lem:lsmooth}:
$$
f(w) - f(w^*) \geq \frac{1}{2L}  || \nabla f(w) ||^2
$$

As we are performing gradient ascent $f(w) = -d^{\Phi}_{s,t}$, we get the following result:
$$
 (d^{*}_{S,T} - d^{\Phi}_{S,T}) \geq   \frac{1}{2L}|| \nabla d^{\Phi}_{S,T}(x) ||^2
$$
$$
2L (d^{*}_{S,T} - d^{\Phi}_{S,T} ) \geq \frac{(d^{\Phi + \eta v_2}_{S,T} - d^{\Phi + \eta v}_{S,T})^2}{(\eta(1 - \cos \alpha))^2}
$$
$$
\eta(1 - \cos \alpha)\sqrt{2L(d^*_{S,T} - d^{\Phi}_{S,T}) } \geq {(d^{\Phi'}_{S,T} - d^{\Phi''}_{S,T})}.
$$
This shows that difference in value of by taking a step in direction of gradient $v$ vs taking the step in a different direction $v_2$ is upper bounded by the $ d^{*}_{S,T} - d^{\Phi}_{S,T}(x)$, hence if we are far from minima the difference can be potentially large. As we are only doing one step of gradient ascent $ d^{*}_{S,T} - d^{\Phi}_{S,T}$ will be potentially large, hence can lead to suboptimal measure of discrepancy. 
\end{proof}

\begin{theorem}
\label{sdat_th:th3}
Suppose l is the loss function, we denote $\lambda^* := R_S^l(h^*) + R_T^l(h^*)$ and let $h^*$ be the ideal joint hypothesis:
\begin{equation}
         R_{T}^l(h_{\theta}) \leq \; \max_{||\epsilon|| \leq \rho}\hat{R}_S^l(h_{\theta + \epsilon}) + D_{h_{\theta}, H}^{\phi}(P_S||P_T)  \\ + \gamma(||\theta||_2^2/\rho^2) + \lambda^* 
\end{equation}
where $\gamma: \mathbb{R}^{+} \rightarrow \mathbb{R}^{+}$ is a strictly increasing function.
\end{theorem}
\begin{proof}[Proof of Theorem \ref{sdat_th:th3}: ]
 In this case we make use of Theorem 2 in the chapter sharpness aware minimization \citep{foret2021sharpnessaware} which states the following:
The source risk $R_S(h)$ is bounded using the following PAC-Bayes generalization bound for any $\rho$ with probability $1 - \delta$:
\begin{equation}
\begin{split}
      R_S(h_{\theta}) \leq \max_{||\epsilon|| \leq \rho} \hat{R}_S(h_{\theta})  +\sqrt{\frac{k\log\left(1+\frac{\|\boldsymbol{\theta}\|_2^2}{\rho^2}\left(1+\sqrt{\frac{\log(n)}{k}}\right)^2\right) + 4\log\frac{n}{\delta} + \tilde{O}(1)}{n-1}}  
\end{split}
\end{equation}
here $n$ is the training set size used for calculation of empirical risk $\hat{R}_S(h)$, $k$ is the number of parameters and $||\theta||_2$ is the norm of the weight parameters. The second term in equation can be abbreviated as $\gamma(||\theta||_2)$. Hence,
\begin{equation}
    R_S(h_{\theta}) \leq     \max_{||\epsilon|| \leq \rho} \hat{R}_S(h_{\theta}) + \gamma(||\theta||_2^2/\rho^2) 
\end{equation}
from the generalization bound for domain adaptation for any f-divergence  \citep{acuna2021f} (Theorem 2) we have the following result.
\begin{equation}
R_T^l(h_{\theta}) \leq R_{S}^l(h_{\theta}) + \mathcal{D}_{h_{\theta}, H}^{\phi}(P_S||P_T) + \lambda^*
\end{equation}
Combining the above two inequalities gives us the required result we wanted to prove i.e.
\begin{equation}
         R_{T}^l(h_\theta) \leq \; \tilde{R}_S^l(h_{\theta}) + D_{h_{\theta}, H}^{\phi}(P_S||P_T)  + \gamma(||\theta||_2^2/\rho^2) + \lambda^* .
\end{equation}

\end{proof}
\begin{table*}[!t]

    \begin{minipage}{.4\linewidth}
      \caption{Architecture used for feature classifier and Domain classifier. $C$ is the number of classes. Both classifiers will take input from feature generator ($g_\theta$).}
      \vskip 0.15in
      \label{sdat_tab:clf}
      \centering
    \begin{tabular}{c||c}
    \hline
      Layer  &  Output Shape\\
      \hline
      \multicolumn{2}{c}{\textbf{Feature Classifier ($f_{\Theta}$)}} \\
      \hline
        - & Bottleneck Dimension \\
        Linear & $C$ \\
        \hline
        \multicolumn{2}{c}{\textbf{Domain Classifier} ($\mathcal{D}_\Phi$)} \\
        \hline
        - & Bottleneck Dimension \\
        Linear & 1024 \\
        BatchNorm & 1024 \\
        ReLU & 1024 \\
        Linear & 1024 \\
        BatchNorm & 1024 \\
        ReLU & 1024 \\
        Linear & 1 \\
       
    \end{tabular}
    \end{minipage}%
     \hspace{2em}
    \begin{minipage}{.59\linewidth}
      \centering
        \caption{ Accuracy (\%) on {VisDA-2017} (ResNet-101 and ViT backbone).}
        
        \label{sdat_table:visda}
        \begin{tabular}{l|c|c}
        \hline
		\textbf{Method}& &\textbf{Syn $\rightarrow$ Real} \Bstrut\\
		\hline \hline
		DANN \citep{ganin2016domain} &\parbox[t]{2mm}{\multirow{7}{*}{\rotatebox[origin=c]{90}{ResNet-101}}}& 57.4 \\
		MCD \citep{saito2018maximum}  && 71.4\\
		CDAN* \citep{long2018conditional} && 73.7
		\Bstrut 
		\\
		\cline{1-1}\cline{3-3}
		CDAN && 76.6
		\\
		CDAN w/ SDAT && 78.3\Bstrut
		\\
		\cline{1-1}\cline{3-3}
		CDAN+MCC \citep{jin2020minimum} && \underline{80.4}
		\\
		CDAN+MCC w/ SDAT && \textbf{81.2}
		\\
		\hline \hline
		CDAN &\parbox[t]{2mm}{\multirow{4}{*}{\rotatebox[origin=c]{90}{ViT}}}& 76.7
		\\
		CDAN w/ SDAT && 81.1\Bstrut
		\\
		\cline{1-1}\cline{3-3}
		CDAN+MCC \citep{jin2020minimum} && \underline{85.1}
		\\
		CDAN+MCC w/ SDAT && \textbf{87.8}		
	\end{tabular}%
    \end{minipage} 
\end{table*}

\section{Hessian Analysis}\label{sdat_app:hess}
We use the PyHessian library \citep{yao2020pyhessian} to calculate the Hessian eigenvalues and the Hessian Eigen Spectral Density. For Office-Home experiments, all the calculations are performed using 50\% of the source data at the last checkpoint. For DomainNet experiments (Fig. \ref{sdat_fig:hessian}D), we use 10\% of the source data for Hessian calculation. The Maximum Eigenvalue is calculated at the checkpoint with the best validation accuracy ($\lambda_{\max}^{best}$) and the last checkpoint ($\lambda_{\max}^{last}$). Only the source class loss is used for calculating to clearly illustrate our point. {The partition was selected randomly, and the same partition was used across all the runs. We also made sure to use the same environment to run all the Hessian experiments. A subset of the data was used for Hessian calculation mainly because the hessian calculation is computationally expensive \citep{yao2020pyhessian}. This is commonly done in hessian experiments. For example, \citep{chen2021vision} (refer Appendix D) uses 10\% of training data for Hessian Eigenvalue calculation}. 
The PyHessian library uses Lanczos algorithm \citep{ghorbani2019investigation} for calculating the Eigen Spectral density of the Hessian and uses the Hutchinson method to calculate the trace of the Hessian efficiently. 

\section{Smoothness of Discriminator in SNGAN}
\label{sdat_app:gan_exp}
\begin{figure*}[!t]
  \centering
  \includegraphics[scale=0.5]{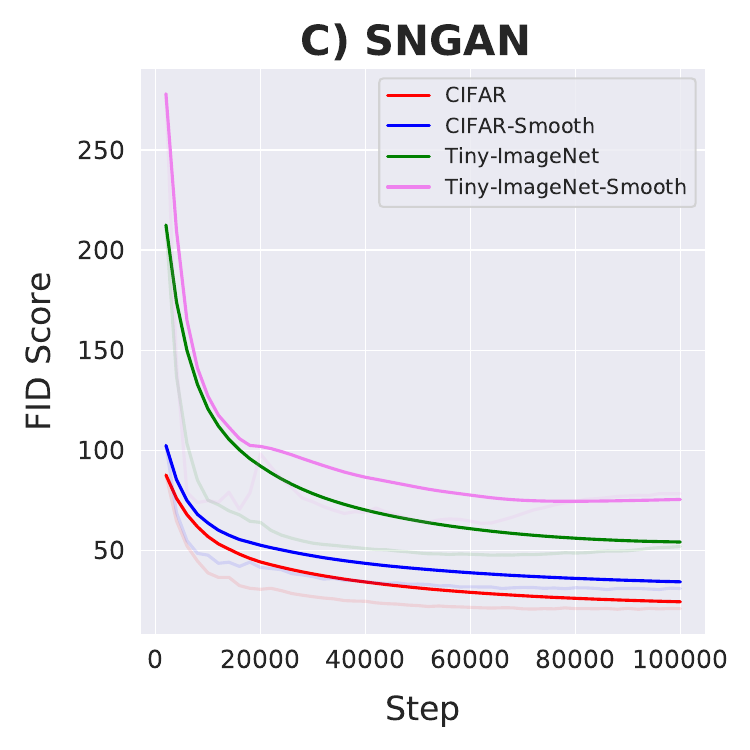}
  \caption{SNGAN performance on different datasets, smoothing discriminator in GAN also leads to inferior GAN performance (higher FID) across both datasets.}
  \label{sdat_fig:gan}
\end{figure*}
For further establishing the generality of sub-optimality of smooth adversarial loss, we also perform experiments on Spectral Normalised Generative Adversarial Networks (SNGAN) \citep{miyato2018spectral}. In case of SNGAN we also find that smoothing discriminator through SAM leads to suboptimal performance (higher FID) as in Fig. \ref{sdat_fig:gan}. The above evidences indicates that \textit{smoothing the adversarial loss leads to sub-optimality}, hence it should not be done in practice. 
We use the same configuration for SNGAN as described in PyTorch\-StudioGAN \citep{kang2020ContraGAN}  for both CIFAR10 \citep{krizhevsky2009learning} and TinyImageNet \footnote{https://www.kaggle.com/c/tiny-imagenet} with batch size of 256 in both cases. We then smooth the discriminator while discriminator is trained by using the same formulation as in Eq. \ref{sdat_eq:smooth_disc}. We find that smoothing discriminator leads to higher (suboptimal) Fr\'{e}chet Inception Distance in case of GANs as well, shown in Fig. \ref{sdat_fig:gan}.

\section{Experimental Details}\label{sdat_app:experimental_deets}
\label{sdat_exp_details}
\subsection{Image Classification}
\textbf{Office-Home}: For CDAN methods with ResNet-50 backbone, we train the models using mini-batch stochastic gradient descent (SGD) with a batch size of 32 and a learning rate of 0.01. The learning rate schedule is the same as \citep{ganin2016domain}. We train it for a total of 30 epochs with 1000 iterations per epoch. The momentum parameter in SGD is set to 0.9 and a weight decay of 0.001 is used. For CDAN+MCC experiments with ResNet-50 backbone, we use a temperature parameter \citep{jin2020minimum} of 2.5. The bottleneck dimension for the features is set to 2048.

\noindent \textbf{VisDA-2017}:
We use a ResNet-101 backbone initialized with ImageNet weights for VisDA-2017 experiments. Center Crop is also used as an augmentation during training. We use a bottleneck dimension of 256 for both algorithms.
For CDAN runs, we train the model for 30 epochs with same optimizer setting as that of Office-Home.
For CDAN+MCC runs, we use a temperature parameter of 3.0 and a learning rate of 0.002. 

\noindent \textbf{DomainNet}:
We use a ResNet-101 backbone initialized with ImageNet weights for DomainNet experiments.  We run all the experiments for 30 epochs with 2500 iterations per epoch. The other parameters are the same as that of Office-Home.

Additional experiments with a ViT backbone are performed on Office-Home and VisDA-2017 datasets. We use the ViT-B/16 architecture pretrained on ImageNet-1k, the implementation of which is borrowed from \cite{rw2019timm}. For all CDAN runs on Office-Home and VisDA, we use an initial learning rate of 0.01, whereas for CDAN+MCC runs, the initial learning rate of 0.002 is used. $\rho$ value of 0.02 is shared across all the splits on both the datasets for the ViT backbone. A batch-size of 24 is used for Office-Home and 32 for VisDA-2017. 

To show the effectiveness of SDAT fairly and promote reproducibility, we run with and without SDAT on the same GPU and environment and with the same seed. All the above experiments were run on Nvidia V100, RTX 2080 and RTX A5000 GPUs. We used Wandb \citep{wandb} to track our experiments. We will be releasing the code to promote reproducible research.

\vspace{1mm} \noindent \textbf{Architecture of Domain Discriminator}
One of the major reasons for increased accuracy in Office-Home baseline CDAN compared to reported numbers in the chapter is the architecture of domain classifier. The main difference is the use of batch normalization layer in domain classifier, which was done in the library \citep{dalib}. Table \ref{sdat_tab:clf} shows the architecture of the feature classifier and domain classifier.

\subsection{Additional Implementations Details for DA for Object detection}
In SDAT, we modified the loss function present in \citet{chen2018domaindafaster} by adding classification loss smoothing, i.e. smoothing classification loss of RPN and ROI, used in {Faster R-CNN} \citep{ren2015faster}, by training with source data. Similarly, we applied smoothing to regression loss and found it to be less effective. We implemented SDAT for object detection using Detectron2 \citep{wu2019detectron2}. The training is done via SGD with momentum 0.9 for 70k iterations with the learning rate of 0.001, and then dropped to 0.0001 after 50k iterations. We split the target data into train and validation sets and report the best mAP on validation data. 
We fixed $\rho$ to 0.15 for object detection experiments.

\section{Additional Results}\label{sdat_app:add_results}
\textbf{VisDA-2017}:
Table \ref{sdat_table:visda} shows the overall accuracy on the VisDA-2017 with ResNet-101 and ViT backbone. The accuracy reported in this table is the overall accuracy of the dataset, whereas the accuracy reported in the Table \textcolor{red}{5} of the chapter refers to the mean of the accuracy across classes. CDAN w/ SDAT outperforms CDAN by 1.7\% with ResNet-101 and by 4.4\% with ViT backbone, showing the effectiveness of SDAT in large scale Synthetic $\rightarrow$ Real shifts. With CDAN+MCC as the DA method, adding SDAT improves the performance of the method to 81.2\% with ResNet-101 backbone.\\
\textbf{DomainNet}:
Table \ref{sdat_tab:domainnet} shows the results of the proposed method on DomainNet across five domains. We compare our results with ADDA and MCD and show that CDAN achieves much higher performance on DomainNet compared to other techniques. It can be seen that CDAN w/ SDAT further improves the overall accuracy on DomainNet by 1.8\%. 
\\
We have shown results with three different domain adaptation algorithms namely DANN \citep{ganin2015unsupervised}, CDAN \citep{long2018conditional} and CDAN+MCC \citep{jin2020minimum}. SDAT has shown to improve the performance of all the three DA methods. This shows that SDAT is a generic method that can applied on top of any domain adversarial training based method to get better performance.

 \begin{table*}[t]

    \centering
    \caption{Accuracy(\%) on \textbf{DomainNet} dataset for unsupervised domain adaptation (ResNet-101) across five distinct domains. The row indicates the source domain and the columns indicate the target domain.}
\vskip 0.15in
    \footnotesize
    \begin{adjustbox}{max width=\textwidth} 
    \begin{tabular}{c|c c c c c c ||c|c c c c c c  }
    \hline
    ADDA & clp & inf & pnt  & rel & skt & Avg & MCD &    clp & inf & pnt  & rel & skt & Avg\\
    \hline
    clp & - & 11.2 & 24.1 & 41.9 & 30.7 & 27.0 & clp & - & 14.2 & 26.1 & 45.0 & 33.8 & 29.8\\
    inf & 19.1 & - & 16.4 & 26.9 & 14.6 & 19.2 & inf & 23.6 & - & 21.2 & 36.7 & 18.0 & 24.9\\
    pnt & 31.2 & 9.5 & - & 39.1 & 25.4 & 26.3  & pnt & 34.4 & 14.8 & - & 50.5 & 28.4 & 32.0\\
    rel & 39.5 & 14.5 & 29.1 & - & 25.7 & 	27.2 & rel & 42.6 & 19.6 & 42.6 & - & 29.3 & 33.5\\
    skt & 35.3 & 8.9 & 25.2 & 37.6 & - & 	26.7 & skt & 41.2 & 13.7 & 27.6 & 34.8 & - & 29.3 \\
    Avg & 31.3 & 11.0 & 23.7 & 36.4 & 24.1 &25.3 & Avg & 35.4 & 	15.6 & 	29.4 & 41.7 & 27.4  & 29.9\\
    \hline
    \textbf{CDAN} &   clp & inf & pnt  & rel & skt & Avg & \textbf{CDAN w/ SDAT}&    clp & inf & pnt  & rel & skt & Avg  \\\hline
    clp &     -  & 20.6& 38.9 &  56.0 & 44.9 & 40.1& clp &     -  & 22.0 & 41.5 &   57.5 & 47.2 & 42.1 \\
    inf &    31.5 &  -  & 29.3   & 43.6 & 26.3 & 32.7  & inf &    33.9 &  -  & 30.3 &  48.1 & 27.9 & 35.0\\
    pnt &    44.1 & 19.8 &   - & 57.2 & 39.9 & 40.2 & pnt &    47.5 & 20.7 &   - &   58.0 & 41.8 & 42.0\\
    rel &    55.8& 24.4 & 53.2  &  -  & 42.3 & 43.9  & rel &    56.7 & 25.1 & 53.6 &  -  & 43.9 & 44.8\\
     skt &    56.0 & 20.7 & 45.3 &   54.9&  -  & 44.2 & skt &    58.7 & 21.8 & 48.1  & 57.1 &  -  & 46.4 \\
     Avg&    46.9 & 21.4 & 41.7    & 52.9 & 38.3 & 40.2 & Avg &   49.2 & 22.4 & 43.4 &  55.2 & 40.2 & \textbf{42.1}\\

    \hline

    \hline 
\end{tabular}
\end{adjustbox}
    \label{sdat_tab:domainnet}

\end{table*}

\noindent \textbf{Source-only}: Source-only setting measures the performance of a model trained only on source domain directly on unseen target data with no further target adaptation. We compare the performance of models with and without smoothing the loss landscape for source-only experiments on VisDA-2017 (Table \ref{sdat_table:visda_uda_vit_erm}) and Office-Home (Table \ref{sdat_tab:officehome_erm}) datasets with a ViT backbone pretrained on ImageNet. Initial learning rate of 0.001 and 0.002 is used for Office-Home and VisDA-2017 dataset, respectively. $\rho$ value of 0.002 is used for ERM w/SAM run for both the datasets. It can be seen that ERM w/ SAM does not directly lead to better performance on the target domain.

\begin{table*}[t]
\caption{Accuracy (\%) of source-only model trained with SGD (ERM) and SAM (ERM w/SAM) on VisDA-2017 for unsupervised DA with ViT-B/16 backbone }
\vskip 0.15in
	\centering
	\label{sdat_table:visda_uda_vit_erm}
	\begin{adjustbox}{max width=\textwidth}

	\begin{tabular}{l|cccccccccccc|c}
        \hline
		\textbf{Method} & \textbf{plane} & \textbf{bcybl} & \textbf{bus} & \textbf{car} & \textbf{horse} & \textbf{knife} & \textbf{mcyle} & \textbf{persn} & \textbf{plant} & \textbf{sktb} & \textbf{train} & \textbf{truck} & \textbf{mean} \Tstrut\Bstrut\\
		\hline \hline

		{ERM} & 98.4  & 58.3 & 80.2 & 60.7 & 89.3 & 53.6 & 88.4 & 40.8 & 62.8 & 87.4 & 94.7 & 19.1 & 69.5\\
		ERM w/ SAM  & 98.6 & 33.1 & 80.0 & 76.9 & 90.1 & 35.9 & 94.2 & 22.8 & 77.8 & 89.0 & 95.3 & 11.6 & 67.1 \\
 
		\hline\hline
		
	\end{tabular}%
	\end{adjustbox}
	
\end{table*}

\begin{table*}[t]
  \centering     
  \caption{Accuracy (\%) of source-only model trained with SGD (ERM) and SAM (ERM w/SAM) on Office-Home for unsupervised DA with ViT-B/16 backbone}  
  \vskip 0.15in
  \resizebox{\textwidth}{!}{%
  \begin{tabular}{l|cccccccccccc|c}
    \hline
    \textbf{Method} & \textbf{Ar$\veryshortarrow$Cl} & \textbf{Ar$\veryshortarrow$Pr} & \textbf{Ar$\veryshortarrow$Rw} & \textbf{Cl$\veryshortarrow$Ar} & \textbf{Cl$\veryshortarrow$Pr} & \textbf{Cl$\veryshortarrow$Rw} & \textbf{Pr$\veryshortarrow$Ar} & \textbf{Pr$\veryshortarrow$Cl} & \textbf{Pr$\veryshortarrow$Rw} & \textbf{Rw$\veryshortarrow$Ar} & \textbf{Rw$\veryshortarrow$Cl} & \textbf{Rw$\veryshortarrow$Pr} & \textbf{Avg} \Tstrut
    \Bstrut\\\hline \hline
	{ERM} & 51.5 & 80.8 & 86.0 & 74.8 & 80.2 & 82.6 & 71.8 & 51.0  & 85.5 & 79.5 & 55.0 & 87.9 & 73.9 \\
	{ERM w/ SAM} & 50.8 & 79.5 & 85.2 & 72.6 & 78.4 & 81.4 & 71.8 & 49.6 & 85.2 & 79.0 & 52.8 & 87.2 & 72.8 \\
\hline \hline
  \end{tabular}%
      }

  \label{sdat_tab:officehome_erm}
\end{table*}

\section{Different Smoothing Techniques}\label{sdat_app:smooth_tech}
\textbf{Stochastic Weight Averaging (SWA)} \citep{izmailov2018averaging}: SWA is a widely popular technique to reach a flatter minima. The idea behind SWA is that averaging weights across epochs leads to better generalization because it reaches a wider optima. The recently proposed SWA-Densely (SWAD) \citep{cha2021swad} takes this a step further and proposes to average the weights across iterations instead of epochs. SWAD shows improved performance on domain generalization tasks.
We average every 400 iterations in the SWA instead of averaging per epochs. We tried averaging across 800 iterations as well and the performance was comparable. 
\\
\textbf{Difference between SWAD and SDAT}:
As SWAD performs Weight Averaging, it is not possible to selectively smooth only minimization (ERM) components with SWAD, as gradients for both the adversarial loss and ERM update weights of the backbone. Due to this, SWAD cannot reach optimal performance for DAT. For verifying this, we also compare our method by implementing SWAD for Domain Adaptation on four different source-target pairs of Office-Home dataset in Table \ref{sdat_tab:diff_smooth}. On average, SDAT (Ours) gets 61.6\% (+2.4\% over DAT) accuracy in comparison to 60.4\% (+1.2\% over DAT) for SWAD.
\\
\textbf{Virtual Adversarial Training (VAT)} \citep{miyato2019vat}: VAT is regularization technique which makes use of adversarial perturbations. Adversarial perturbations are created using Algo. 1 present in  \citep{miyato2019vat}. We added VAT by optimizing the following objective: 
\begin{equation}
\min_{\theta}  \mathbb{E}_{x \sim P_S}[\underset{||r|| \leq \epsilon}{\max} D_{KL}(h_\theta(x)|| h_{\theta}(x+r))]
\end{equation}
This value acts as a negative measure of smoothness and minimizing this will make the model smooth. For training, we set hyperparameters $\epsilon$ to 15.0, 
$\xi$ to 1e-6, and $\alpha$ as 0.1.\\
\textbf{Label Smoothing (LS)} \citep{szegedy2016rethinking}: The idea behind label smoothing is to have a distribution over outputs instead of one hot vectors. Assuming that there are $k$ classes, the correct class gets a probability of 1 - $\alpha$ and the other classes gets a probability of $\alpha/(k-1)$ . \citep{stutz2021relating} mention that label smoothing tends to avoid sharper minima during training. We use a smoothing parameter ($\alpha$) of 0.1 in all the experiments in Table \ref{sdat_tab:diff_smooth_appendix}. We also show results with smoothing parameter of 0.2 and observe comparable performance. We observe that label smoothing slightly improves the performance over DAT. \\
\textbf{SAM} \cite{foret2021sharpnessaware}: In this method, we apply SAM directly to both the task loss and adversarial loss with $\rho$ = 0.05 as suggested in the chapter. It can be seen that the performance improvement of SAM over DAT is minimal, thus indicating the need for SDAT.

\begin{table*}[h]
    \centering
    \caption{Different Smoothing techniques. We refer to \citep{stutz2021relating} to compare the proposed SDAT with other techniques to show the efficacy of SDAT. It can be seen that SDAT outperforms the other smoothing techniques significantly. Other smoothing techniques improve upon the performance of DAT showing that smoothing is indeed necessary for better adaptation.}
    \vskip 0.15in
     \begin{adjustbox}{max width=\columnwidth}
    \begin{tabular}{l|cccc}
    \hline
    {Method} & Ar$\veryshortarrow$Cl &  Cl$\veryshortarrow$Pr & Rw$\veryshortarrow$Cl &  Pr$\veryshortarrow$Cl \\
    \hline \hline
    {DAT} & 54.3 & 69.5 & 60.1 & 55.3\\
    
     {VAT} & 54.6 & 70.7 & 60.8 & 54.4 \\
      {SWAD-400} & 54.6 & 71.0 & 60.9 & 55.2 \\
       {LS ($\alpha$ = 0.1)} & 53.6 & 71.6 & 59.9 & 53.4\\
       {LS ($\alpha$ = 0.2)} & 53.5 & 71.2 & 60.5 & 53.2\\
       {SDAT} & \textbf{55.9} & \textbf{73.2} & \textbf{61.4} & \textbf{55.9} \\
    \end{tabular}
    
    \label{sdat_tab:diff_smooth_appendix}
    \end{adjustbox}
\end{table*}
\section{
{Optimum $\rho$ value}}
\label{sdat_app:opt_rho}
Table \ref{sdat_tab:rho_domainnet} and \ref{sdat_tab:rho_visda} show that $\rho$ = 0.02 works robustly across experiments providing an increase in performance (although it does not achieve the best result each time) and can be used as a rule of thumb.
\begin{table*}[h!]
    \centering
    \caption{{$\rho$ value for DomainNet}}
    \vskip 0.15in
     \begin{adjustbox}{max width=\columnwidth}
    \begin{tabular}{l|ccc}
    \hline
    {Split} & DAT &  SDAT($\rho$ = 0.02) & SDAT - Reported ($\rho$ = 0.05) \\
    \hline \hline
    \textbf{clp$\veryshortarrow$skt } & 44.9 & 46.7 & 47.2\\
    
     \textbf{skt$\veryshortarrow$clp} & 56.0 & 59.0 & 58.7 \\
      \textbf{skt$\veryshortarrow$pnt} & 45.3 & 47.8 & 48.1 \\
       \textbf{inf$\veryshortarrow$rel} & 43.6 & 47.3 & 48.1\\
    \end{tabular}
    
    \label{sdat_tab:rho_domainnet}
    \end{adjustbox}
\end{table*}
\begin{table*}[h!]
    \centering
    \caption{{$\rho$ value for VisDA-2017 Synthetic $\veryshortarrow$ Real} }
    \vskip 0.15in
     \begin{adjustbox}{max width=\columnwidth}
    \begin{tabular}{l|ccc}
    \hline
    {Backbone} & DAT &  SDAT ($\rho$ = 0.02) & SDAT Reported($\rho$ = 0.005)\\
    \hline \hline
    {CDAN} & 76.6 & 78.2 & 78.3\\
     {CDAN+MCC} & 80.4 & 80.9 & 81.2\\
    \end{tabular}
    
    \label{sdat_tab:rho_visda}
    \end{adjustbox}
\end{table*}

\section{
{Comparison with TVT}}
\label{sdat_app:comp_tvt}

TVT \cite{yang2021tvt} is a recent work that reports performance higher than the other contemporary unsupervised DA methods on the publicly available datasets. This method uses a ViT backbone and focuses on exploiting the intrinsic properties of ViT to achieve better results on domain adaptation. 
Like us, TVT uses an adversarial method for adaptation to perform well on the unseen target data. On the contrary, they introduce additional modules within their architecture. The Transferability Adaption Module (TAM) is introduced to assist the ViT backbone in capturing both discriminative and transferable features. Additionally, the Discriminative Clustering Module (DCM) is used to perform discriminative clustering to achieve diverse and clustered features. 

Even without using external modules to promote the transferability and discriminability in the features learned using ViT, we are able to report higher numbers than TVT. This advocates our efforts to show the efficacy of converging to a smooth minima w.r.t. task loss to achieve better domain alignment. Moreover, TVT uses a batch size of 64 to train the network, causing a memory requirement of more than 35GB for efficient training, which is significantly higher than the 11.5GB memory used by our method on a batch-size of 24 for Office-Home to obtain better results. This allows our method to be trained using a standard 12GB GPU, removing the need of an expensive hardware. The ViT backbone used by TVT is pretrained on a much larger ImageNet-21k dataset, whereas we use the backbone pretrained on ImageNet-1k dataset.   

\section{
{Significance and Stability of Empirical Results}}
\begin{figure*}[!t]
    \centering
    \begin{subfigure}[b]{0.3\linewidth}
  \centering
  \includegraphics[width=\textwidth]{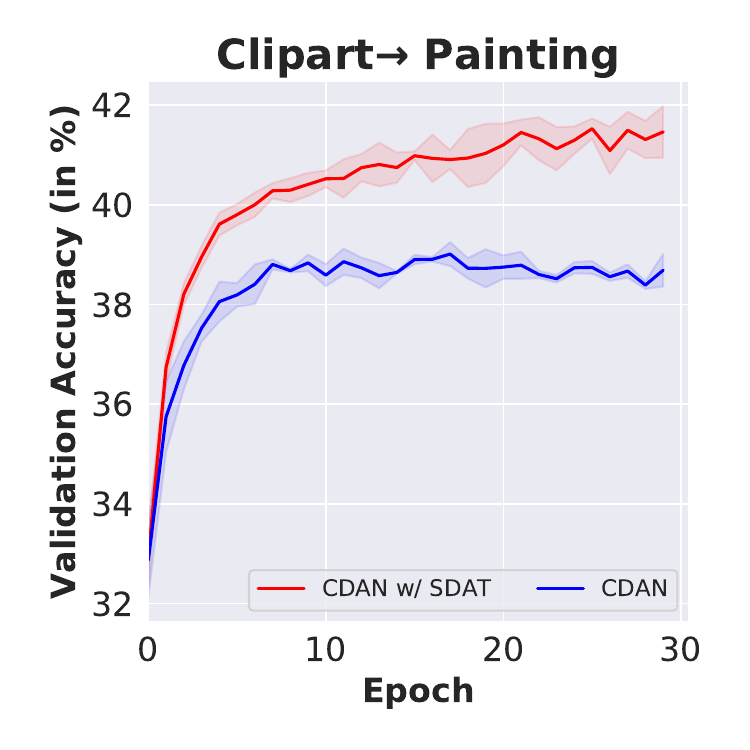}
  \label{sdat_fig:c2p}
\end{subfigure}
\begin{subfigure}[b]{0.3\linewidth}
  \centering
  \includegraphics[width=\textwidth]{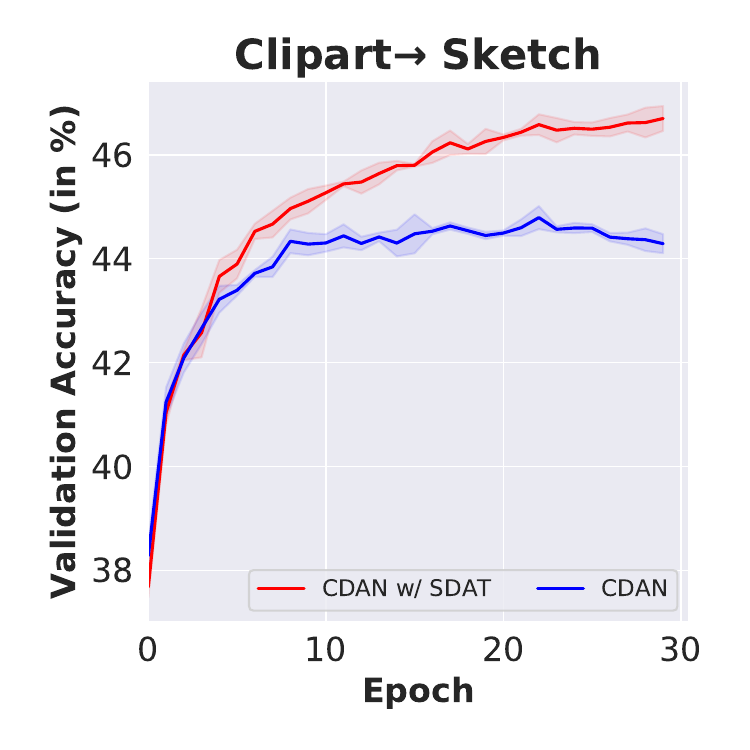}
  \label{sdat_fig:c2s}
\end{subfigure}
\begin{subfigure}[b]{0.3\linewidth}
  \centering
  \includegraphics[width=\textwidth]{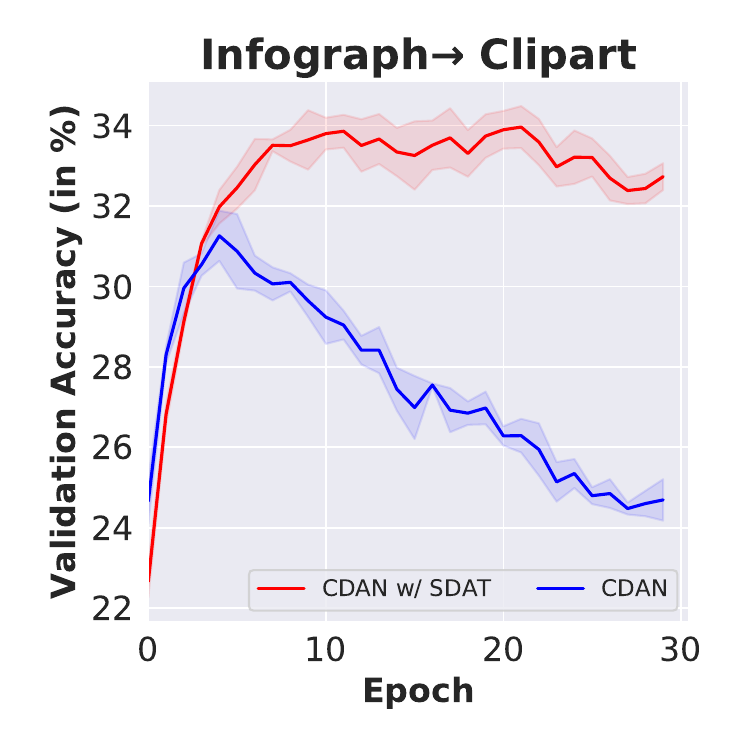}
  \label{sdat_fig:i2c}
\end{subfigure}\\
    \begin{subfigure}[b]{0.3\linewidth}
  \centering
  \includegraphics[width=\textwidth]{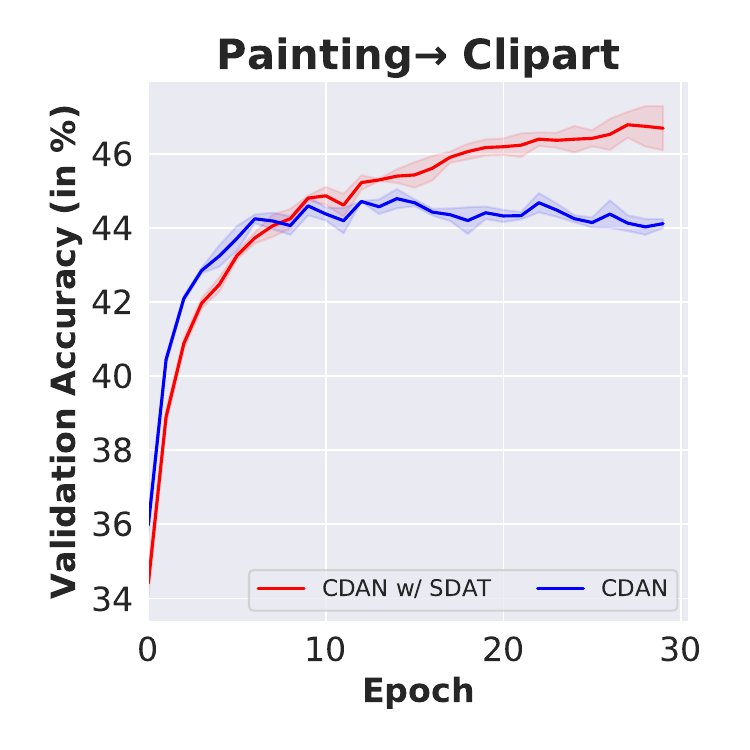}
  \label{sdat_fig:p2c}
\end{subfigure}
\begin{subfigure}[b]{0.3\linewidth}
  \centering
  \includegraphics[width=\textwidth]{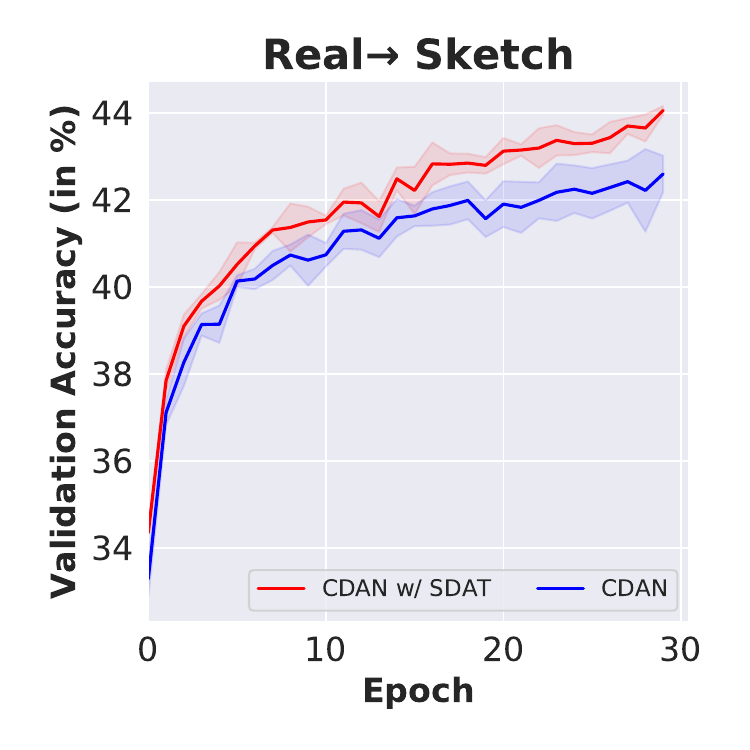}
  \label{sdat_fig:r2s}
\end{subfigure}
\begin{subfigure}[b]{0.3\linewidth}
  \centering
  \includegraphics[width=\textwidth]{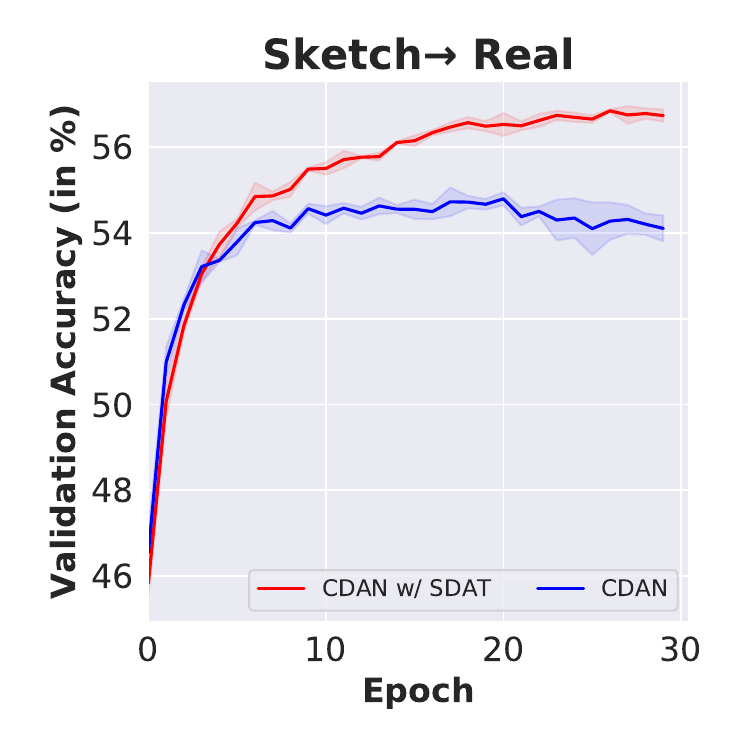}
  \label{sdat_fig:s2r}
\end{subfigure}
\caption{{Validation Accuracy across epochs on different splits of DomainNet. We run on three different random seeds and plot the error bar indicating standard deviation across runs. CDAN w/ SDAT consistently outperforms CDAN across different splits of DomainNet.}}
\label{sdat_fig:plotss}

\end{figure*}
\begin{table*}[!t]
    \centering
    \caption{ {DomainNet experiments over 3 different seeds (with ResNet backbone). We report the mean, standard deviation,  reported increase and average increase in the accuracy (in \%).}}
    \vskip 0.15in
     \begin{adjustbox}{max width=\columnwidth}
    \begin{tabular}{l|cc|cc}
    \hline
    {Split} &  CDAN & CDAN w/ SDAT & Reported Increase (Table \ref{sdat_table:domainnet}) & Average Increase\\
    \hline \hline
    \textbf{clp$\veryshortarrow$pnt} & 38.9 $\pm$  0.1 & 41.5 $\pm$ 0.3 & +2.6 & +2.6 \\
    \textbf{skt$\veryshortarrow$rel} & 55.1 $\pm$  0.2 & 57.1 $\pm$  0.1 & +2.2 & +2.0 \\
    \textbf{pnt$\veryshortarrow$clp } & 44.5 $\pm$  0.3 & 47.1 $\pm$  0.3 & +3.4 & +2.6\\
    \textbf{rel$\veryshortarrow$skt } & 42.4 $\pm$  0.4 & 43.9 $\pm$  0.1& +1.6 & +1.5\\
    \textbf{clp$\veryshortarrow$skt } & 44.9 $\pm$  0.2 & 47.3 $\pm$  0.1 & +2.3 & +2.4\\
    \textbf{inf$\veryshortarrow$clp} & 31.4 $\pm$  0.5 & 34.2 $\pm$  0.3 & +2.3 & +2.7\\
    \end{tabular}
    
    \label{sdat_tab:seed_exp}
    \end{adjustbox}
\end{table*}
\begin{table}[!t]
\centering
\captionsetup{width=\linewidth}
 \caption{{Median accuracy of last 5 epochs on DomainNet dataset with CDAN w/ SDAT. The number in the parenthesis indicates the increase in accuracy with respect to CDAN.}}
 \vskip 0.15in
\begin{adjustbox}{max width=\linewidth}
\begin{tabular}{c | c  c  c  c  c | c } 
 \hline
  \textbf{Target \textbf{($\rightarrow$)}} & \multirow{2}{*}{\textbf{clp}} & \multirow{2}{*}{\textbf{inf}} & \multirow{2}{*}{\textbf{pnt}} & \multirow{2}{*}{\textbf{real}} & \multirow{2}{*}{\textbf{skt}} & \multirow{2}{*}{\textbf{Avg}} \Tstrut\\  \textbf{Source ($\downarrow$)} &&&&&& \Bstrut\\
 \hline\hline
\multirow{2}{*}{\textbf{clp}} & - & 21.9  & 41.6 & 56.5 &  46.4 & 41.6 \Tstrut\\&& \textcolor{ForestGreen}{(+1.7)}& \textcolor{ForestGreen}{(+3.0)}&
 \textcolor{ForestGreen}{(+1.3)} & \textcolor{ForestGreen}{(+2.0)} & \textcolor{ForestGreen}{(+2.0)} \\ 

\multirow{2}{*}{\textbf{inf}} & 32.4 & - & 29.8 & 46.7 & 25.6 & 33.6 \Tstrut\\& \textcolor{ForestGreen}{(+7.9)}&&
  \textcolor{ForestGreen}{(+7.0)} & \textcolor{ForestGreen}{(+12.7)} & \textcolor{ForestGreen}{(+5.4)} & \textcolor{ForestGreen}{(+8.2)} \\

\multirow{2}{*}{\textbf{pnt}} &  47.2 & 21.0 & - & 57.6 & 41.5 & 41.8 \Tstrut\\ & \textcolor{ForestGreen}{(+2.9)}& 
\textcolor{ForestGreen}{(+1.1)} 
 &&
 \textcolor{ForestGreen}{(+1.0)} &
\textcolor{ForestGreen}{(+2.4)} &
 \textcolor{ForestGreen}{(+1.8)} \\

\multirow{2}{*}{\textbf{real}} &  56.5 & 25.5 & 53.9 & - & 43.5 & 44.8 \Tstrut \\ & \textcolor{ForestGreen}{(+0.7)} &
 \textcolor{ForestGreen}{(+0.9)} &
 \textcolor{ForestGreen}{(+0.5)} &
 &
 \textcolor{ForestGreen}{(+1.3)} &
 \textcolor{ForestGreen}{(+0.8)} \\

\multirow{2}{*}{\textbf{skt}} &  59.1 & 22.1 & 48.2 & 56.6 & - & 46.5 \Tstrut \\ &  \textcolor{ForestGreen}{(+3.0)} & 
 \textcolor{ForestGreen}{(+1.7)} &
 \textcolor{ForestGreen}{(+3.1)} &
 \textcolor{ForestGreen}{(+2.9)} &
&
 \textcolor{ForestGreen}{(+2.7)} 
\\\hline

\multirow{2}{*}{\textbf{Avg}} &  48.8 & 22.6 & 43.4 & 54.3 & 39.2 & 41.7 \Tstrut \\ & \textcolor{ForestGreen}{(+3.6)} &
 \textcolor{ForestGreen}{(+1.3)} &
 \textcolor{ForestGreen}{(+3.4)} &  \textcolor{ForestGreen}{(+4.5)} &
 \textcolor{ForestGreen}{(+2.8)} &
 \textcolor{ForestGreen}{(+3.1)}

\end{tabular}
\label{sdat_table:domainnet_median}
 \end{adjustbox}
 \end{table}
\label{sdat_app:stats_sig}
To establish the empirical results' soundness and reliability, we run a subset of experiments (representative of each different source domain) on DomainNet. The experiments are repeated with three different random seeds leading to overall 36 experimental runs (18 for CDAN w/ SDAT (Our proposed method) and 18 for CDAN baseline). 
Due to the large computational complexity of each experiment ($\approx$20 hrs each), we have presented results for multiple trials on a subset of splits. We find (in Table \ref{sdat_tab:seed_exp}) that our method can outperform the baseline average in each of the 6 cases, establishing significant improvement across all splits. However, we found that due to the large size of DomainNet, the average increase (across three different trials) is close to the reported increase in all cases (Table \ref{sdat_tab:seed_exp}), which also serves as evidence of the soundness of reported results (for remaining splits). We also present additional statistics below for establishing soundness.

If the proposed method is unstable, there is a large variance in the validation accuracy across epochs. For analyzing the stability of SDAT, we show the validation accuracy plots in Figure \ref{sdat_fig:plotss}  on six different splits of DomainNet. We find that our proposed SDAT improves over baselines consistently across epochs without overlap in confidence intervals in later epochs. This also provides evidence for the authenticity and stability of our results.  We also find that in some cases, like when using the Infographic domain as a source, our proposed SDAT also significantly \textit{stabilizes the training} (Figure \ref{sdat_fig:plotss} \textit{inf} $\veryshortarrow$ \textit{clp}).

One of the other ways of reporting results reliably proposed by the concurrent work \citep{berthelot2021adamatch} (Section 4.4) involves reporting the median of accuracy across the last few checkpoints. The median is a measure of central tendency which ignores outlier results. We also report the median of validation accuracy for our method \emph{across all splits} for the last five epochs. It is observed that we observe similar gains for median accuracy (in Table \ref{sdat_table:domainnet_median}) as reported in Table \ref{sdat_table:domainnet}.

\begin{table*}[!t]
    \centering
    \caption{ {Office-Home experiments over 3 different seeds (with ResNet-50 backbone). We report the mean, standard deviation,  reported increase and average increase in the accuracy (in \%).}}
    \vskip 0.15in
     \begin{adjustbox}{max width=\columnwidth}
    \begin{tabular}{l|cc|cc}
    \hline
    {Split} &  CDAN & CDAN w/ SDAT & Reported Increase (Table \ref{sdat_tab:officehome}) & Average Increase\\
    \hline \hline
    \textbf{Ar$\veryshortarrow$Cl} & 53.9 $\pm$  0.2 & 55.5 $\pm$ 0.2 & +1.7 & +1.6 \\
    \textbf{Ar$\veryshortarrow$Pr} & 70.6 $\pm$  0.4 & 72.1 $\pm$  0.4 & +1.6 & +1.5 \\
    \textbf{Rw$\veryshortarrow$Cl} & 60.7 $\pm$  0.5 & 61.8 $\pm$  0.4 & +1.3 & +1.1\\
    \textbf{Pr$\veryshortarrow$Cl} & 54.7 $\pm$  0.4 & 55.5 $\pm$  0.4& +0.6 & +0.8\\
    \end{tabular}
    
    \label{sdat_tab:seed_exp_office}
    \end{adjustbox}
\end{table*}
As the Office-Home dataset is smaller (i.e., 44 images per class) in comparison to DomainNet we find that there exists some variance in baseline CDAN results (This is also reported in the well-known benchmark for DA \citep{dalib}). For establishing the empirical soundness, we report results of 4 different dataset splits on 3 seeds. It can be seen in Table \ref{sdat_tab:seed_exp_office} that even though there is variance in baseline results, our combination of CDAN w/ SDAT can produce consistent improvement across different random seeds. This further establishes the empirical soundness of our procedure.

\bibliographystyle{plainnat}

{
\small
\bibliography{references}
}

\end{document}